\documentclass[runningheads]{llncs}

\usepackage{eccv}
\usepackage{eccvabbrv}

\usepackage{xcolor}
\usepackage{colortbl}
\usepackage{wrapfig}
\usepackage{graphicx}
\usepackage{booktabs}
\usepackage{graphicx}
\usepackage{amsmath}
\usepackage{amssymb}
\usepackage[bb=dsserif]{mathalpha}
\usepackage{booktabs}
\usepackage{hhline}
\usepackage{appendix}
\usepackage{tabularx}
\usepackage{scrextend}
\usepackage{mathtools}
\usepackage{caption}
\usepackage{subcaption}
\usepackage{multirow}
\usepackage[dvipsnames]{xcolor}
\usepackage{float}
\usepackage[frozencache,cachedir=.]{minted}
\usepackage{pgfplots}
\usepackage{tikz}
\usepackage{algorithm}
\usepackage{algpseudocode}
\usepackage{titletoc}
\usepackage{enumitem}

\usepackage[accsupp]{axessibility}  %

\definecolor{SteelBlue}{rgb}{0.27,0.51,0.71}
\definecolor{LightSteelBlue}{rgb}{0.69,0.77,0.87}
\definecolor{bg}{HTML}{111111}

\definecolor{tabgreen}{HTML}{AED581}
\definecolor{tabyellow}{HTML}{DCE775}

\definecolor{lightgrey}{RGB}{211,211,211}

\newcommand{\ra}[1]{\renewcommand{\arraystretch}{#1}}

\DeclareMathOperator*{\argmin}{arg\,min}

\usepackage{url}
\usepackage{hyperref}

\usepackage{orcidlink}

\begin{document}

\title{TetraDiffusion: Tetrahedral Diffusion Models for 3D Shape Generation}

\author{Nikolai Kalischek$^\ast$\inst{1}\and
Torben Peters$^\ast$\inst{1} \and
Jan D.\ Wegner\inst{2} \and
Konrad Schindler\inst{1}\\
\small{$^\ast$ equal contribution}
}

\authorrunning{N. Kalischek, T. Peters \etal}

\institute{ETH Zurich \and University of Zurich \\
\texttt{\{nkalischek,tpeters,schindler\}@ethz.ch} \\ \texttt{jandirk.wegner@uzh.ch}
}

\maketitle

\begin{center}
\vspace{-1em}

\includegraphics[width=0.1\linewidth]{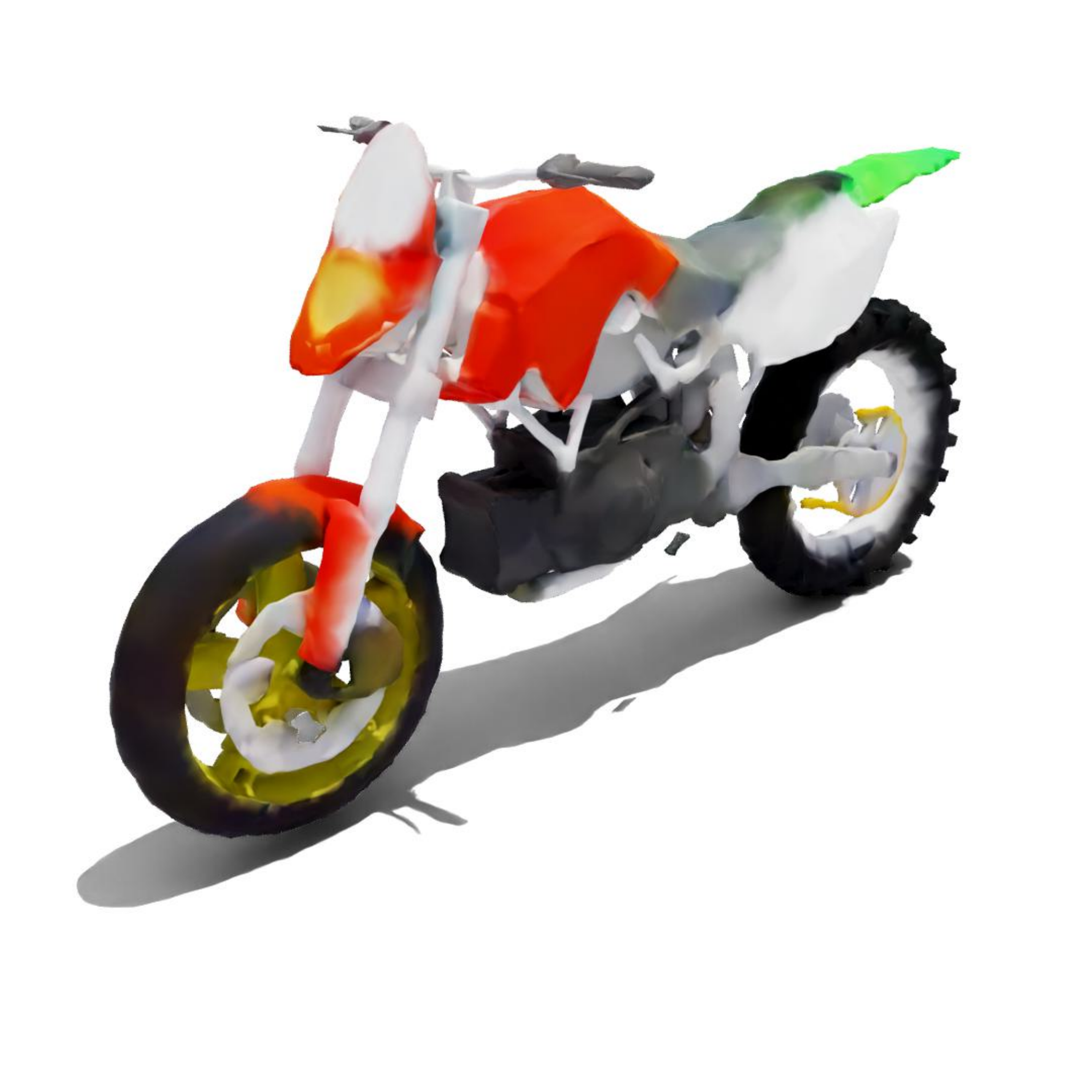}\includegraphics[width=0.1\linewidth]{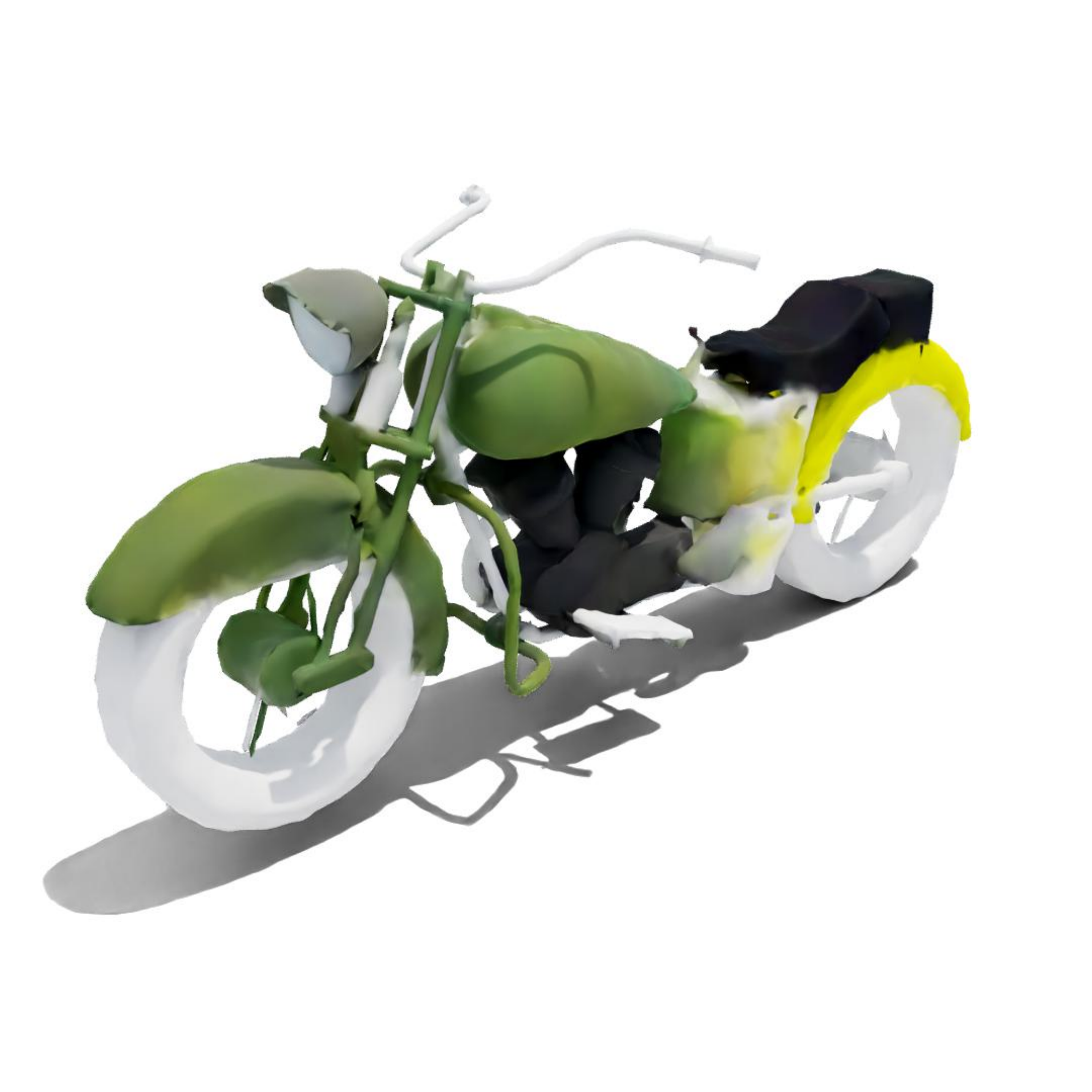}\includegraphics[width=0.1\linewidth]{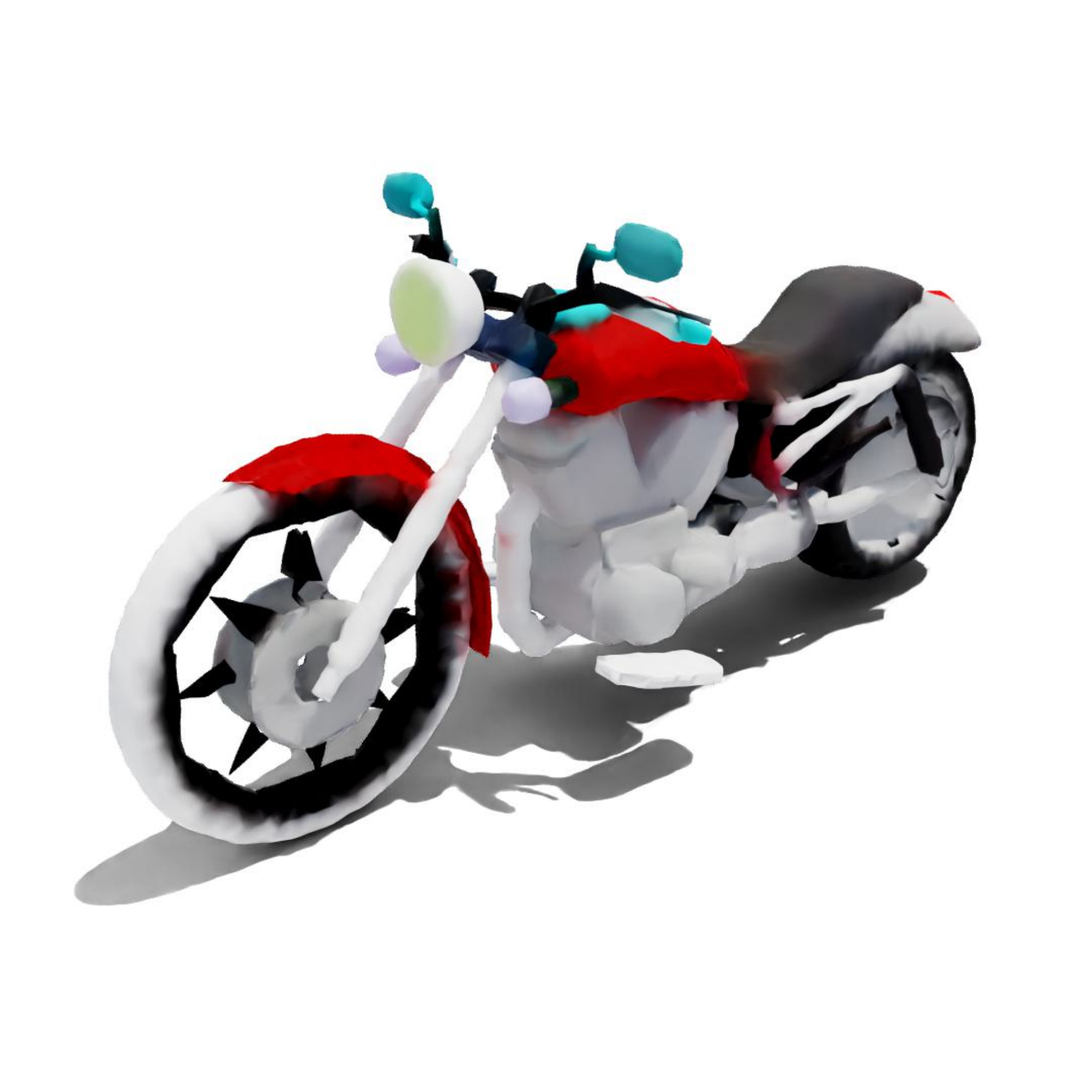}\includegraphics[width=0.1\linewidth]{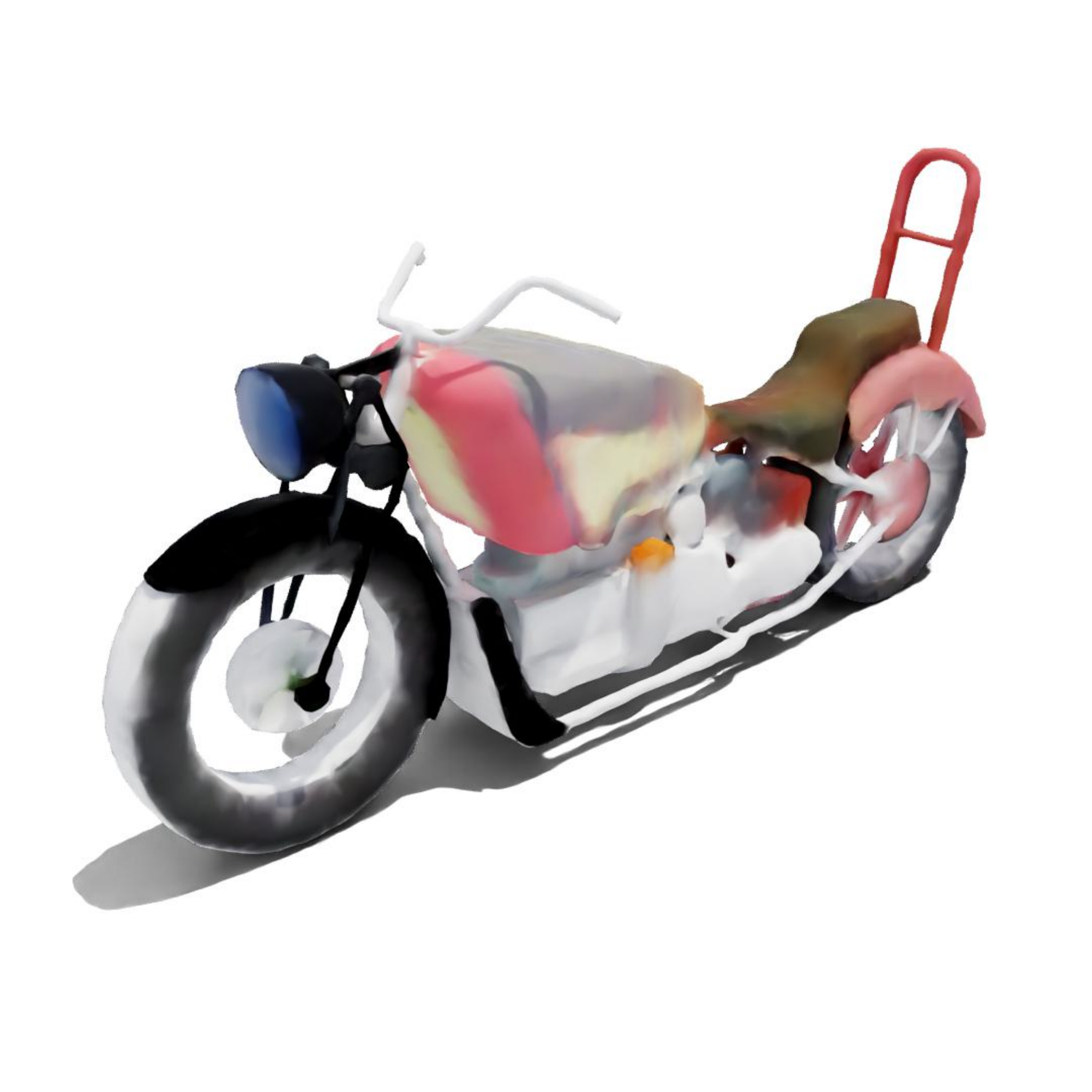}\includegraphics[width=0.1\linewidth]{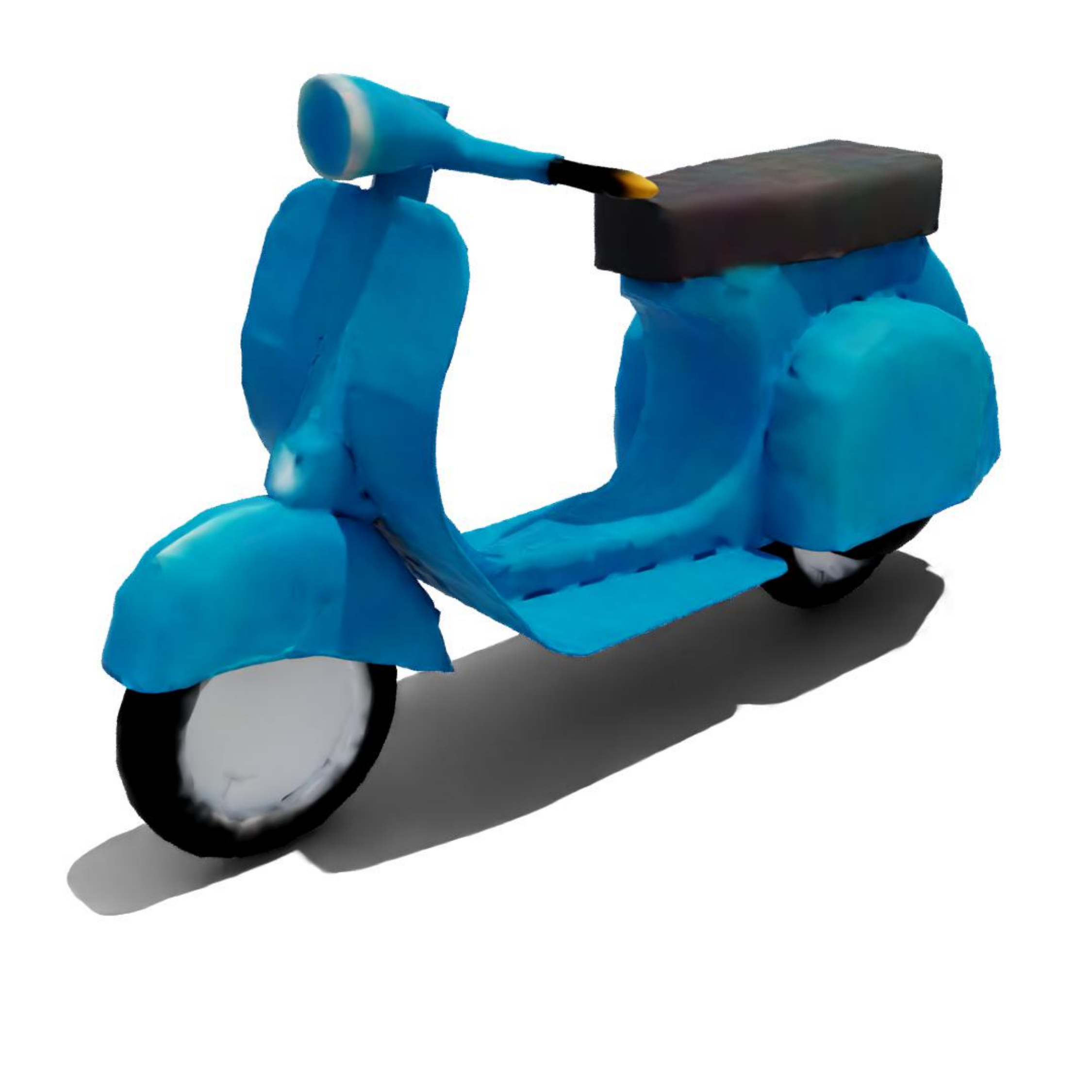}\includegraphics[width=0.1\linewidth]{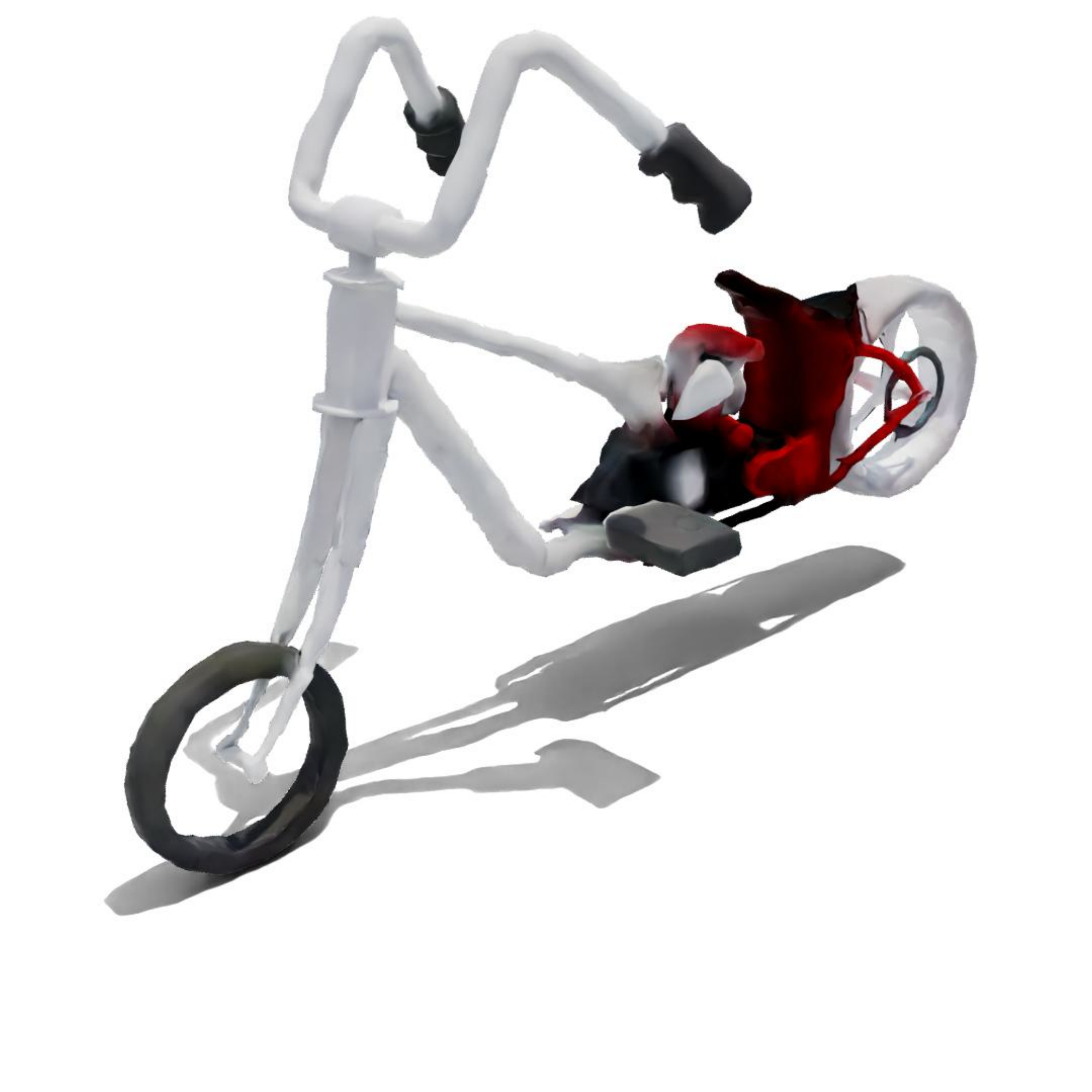}\includegraphics[width=0.1\linewidth]{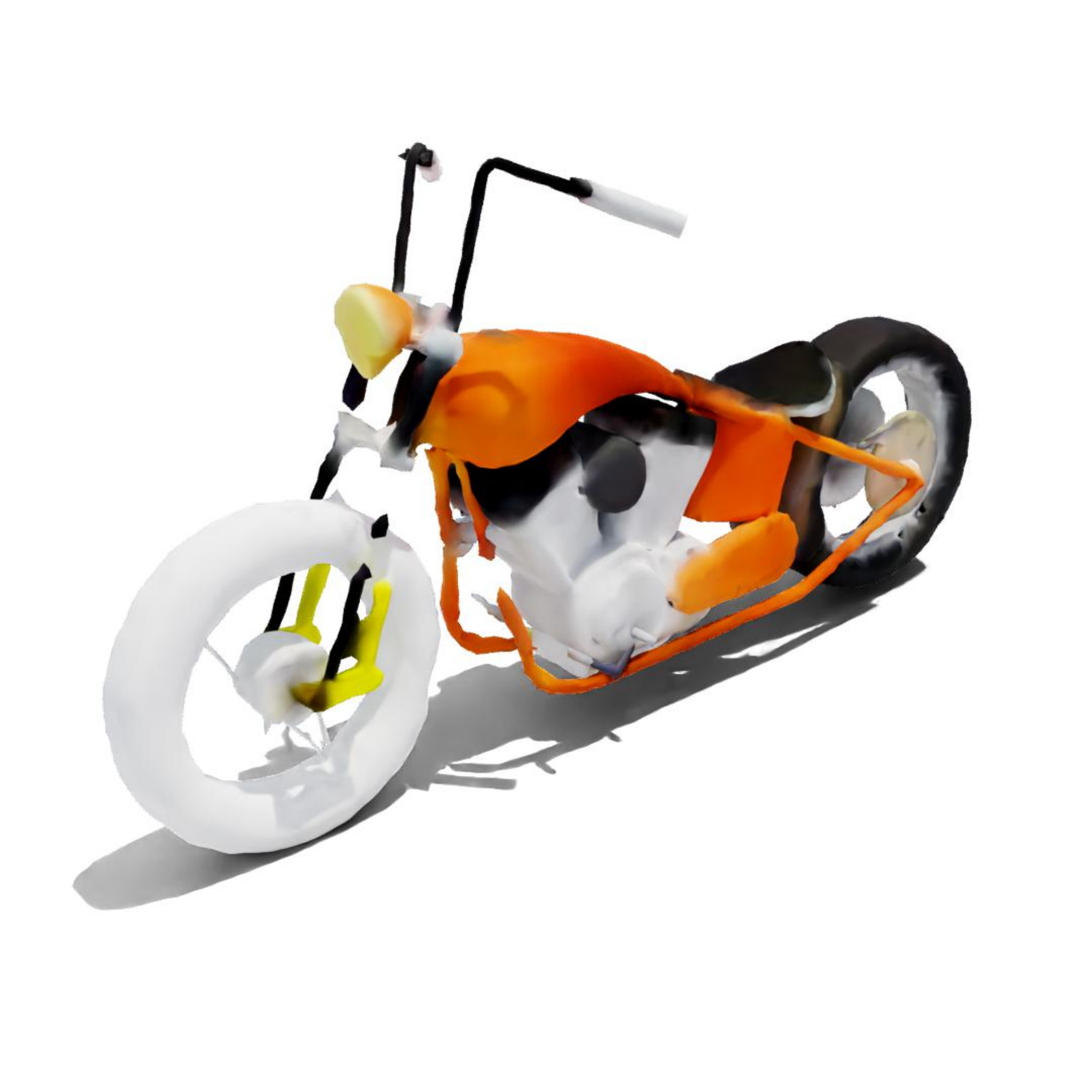}\includegraphics[width=0.1\linewidth]{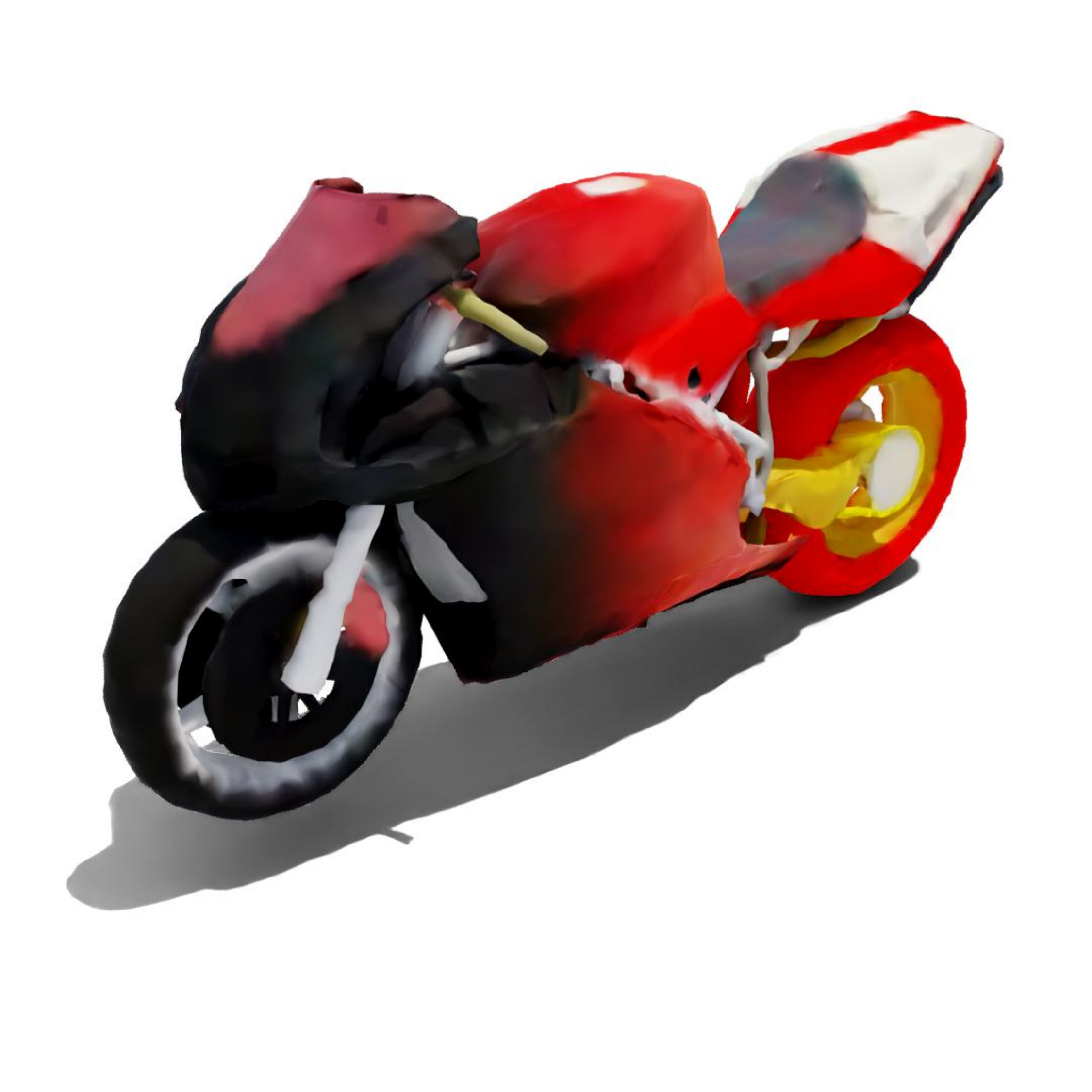}\includegraphics[width=0.1\linewidth]{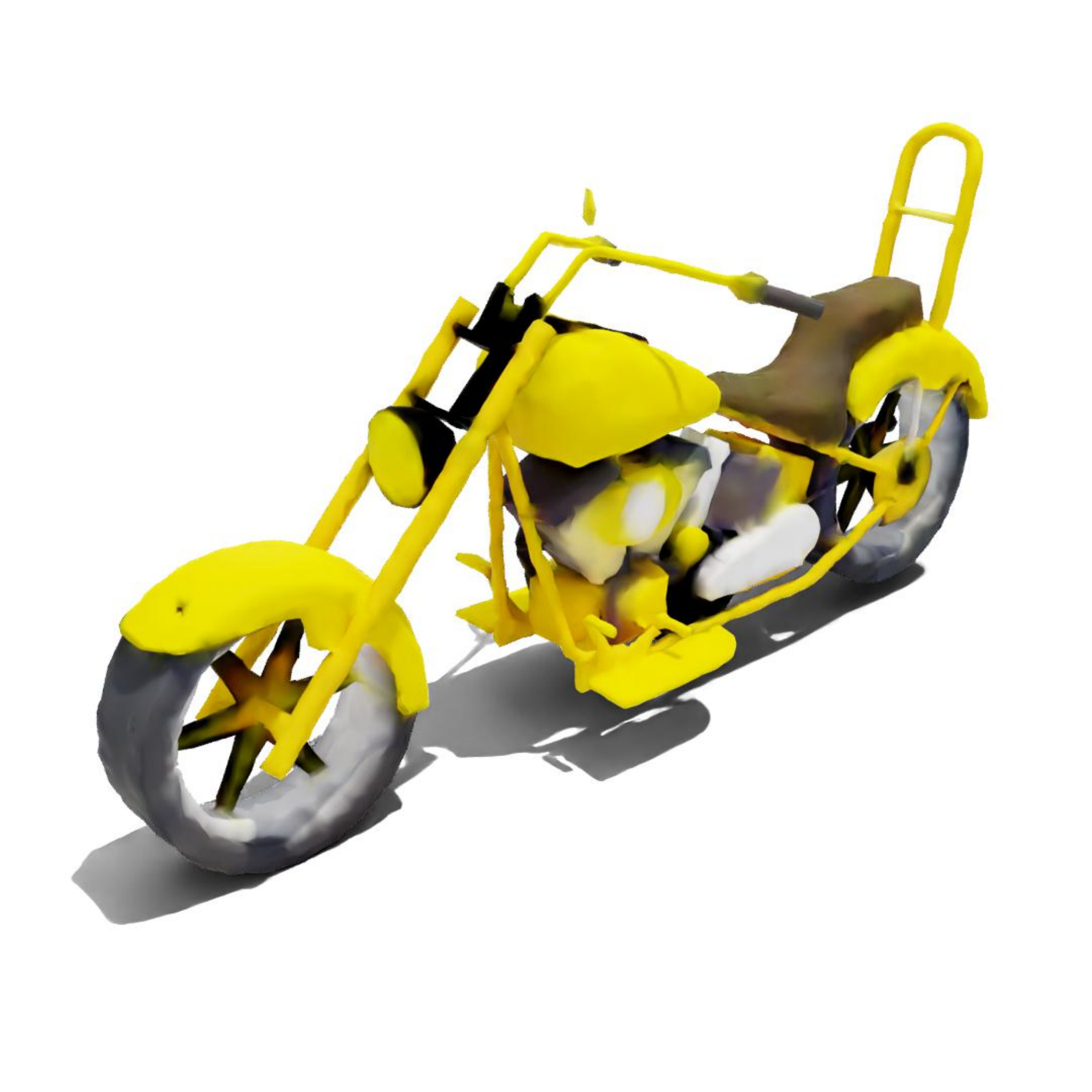}\includegraphics[width=0.1\linewidth]{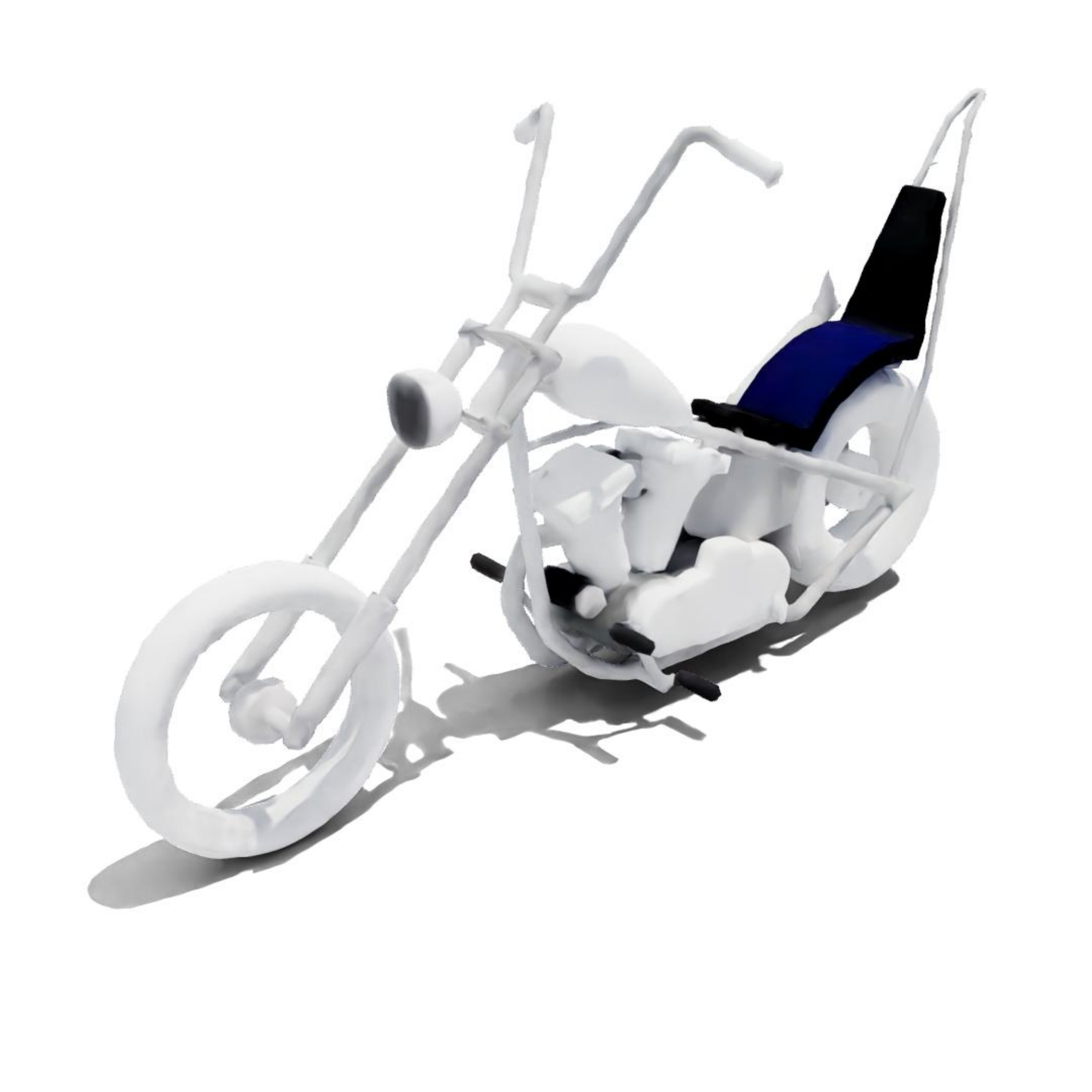}

\vspace{-0.25cm}
\includegraphics[width=0.1\linewidth]{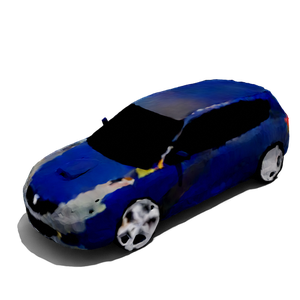}\includegraphics[width=0.1\linewidth]{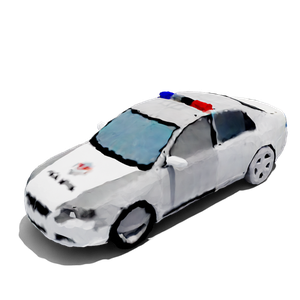}\includegraphics[width=0.1\linewidth]{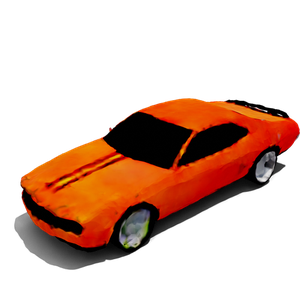}\includegraphics[width=0.1\linewidth]{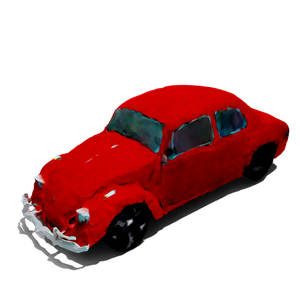}\includegraphics[width=0.1\linewidth]{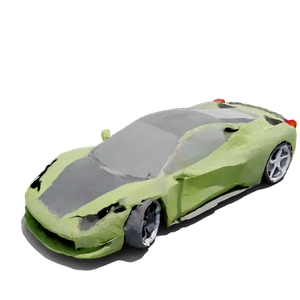}\includegraphics[width=0.1\linewidth]{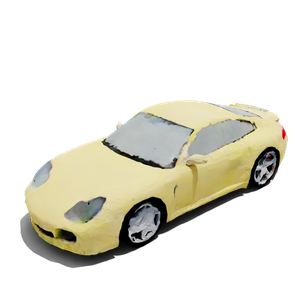}\includegraphics[width=0.1\linewidth]{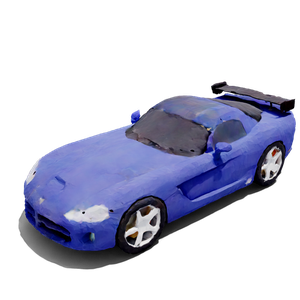}\includegraphics[width=0.1\linewidth]{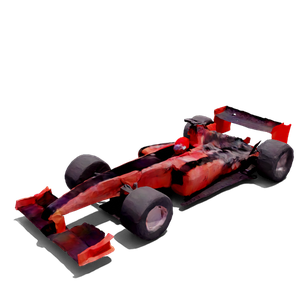}\includegraphics[width=0.1\linewidth]{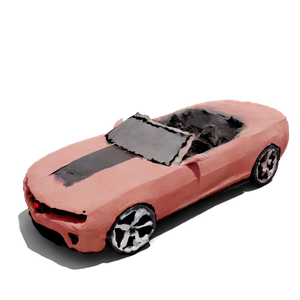}\includegraphics[width=0.1\linewidth]{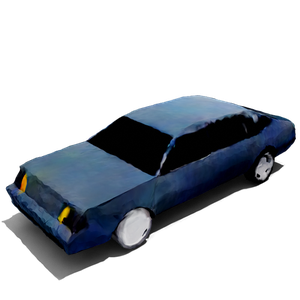}

\vspace{-0.25cm}
\includegraphics[width=0.1\linewidth]{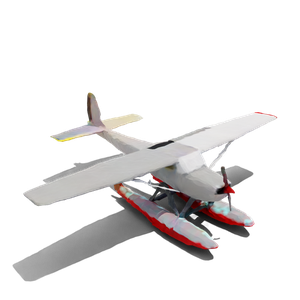}\includegraphics[width=0.1\linewidth]{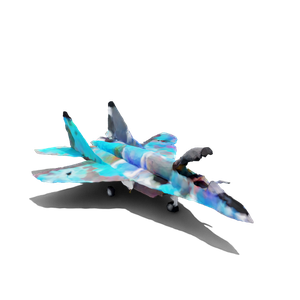}\includegraphics[width=0.1\linewidth]{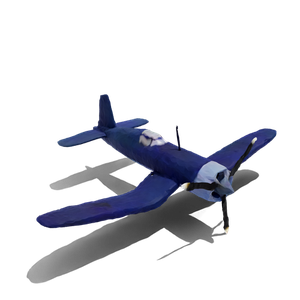}\includegraphics[width=0.1\linewidth]{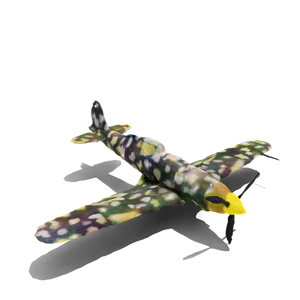}\includegraphics[width=0.1\linewidth]{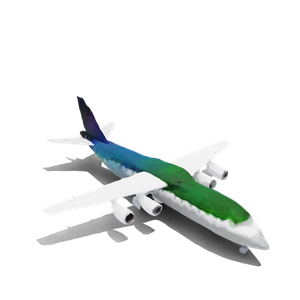}\includegraphics[width=0.1\linewidth]{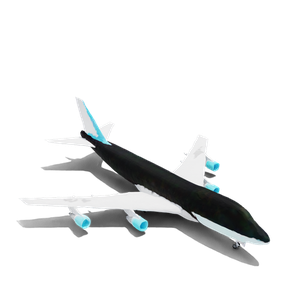}\includegraphics[width=0.1\linewidth]{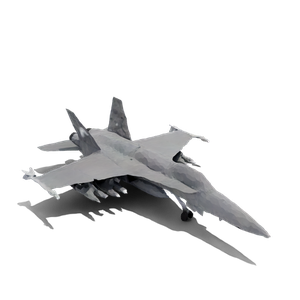}\includegraphics[width=0.1\linewidth] {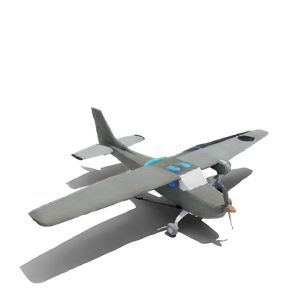}\includegraphics[width=0.1\linewidth]{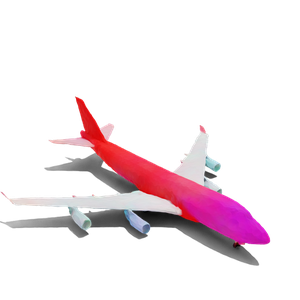}\includegraphics[width=0.1\linewidth]{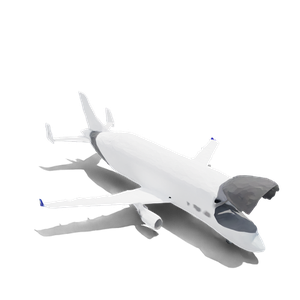}

\vspace{-0.25cm}
\hspace{3cm}
\includegraphics[width=0.1\linewidth]{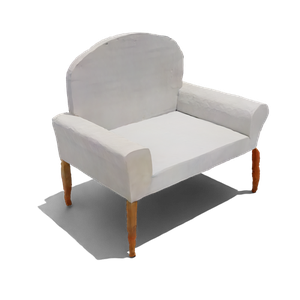}\includegraphics[width=0.1\linewidth]{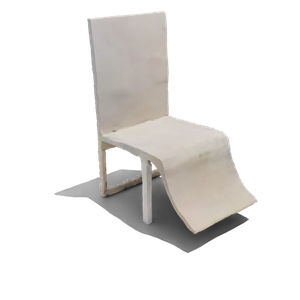}\includegraphics[width=0.1\linewidth]{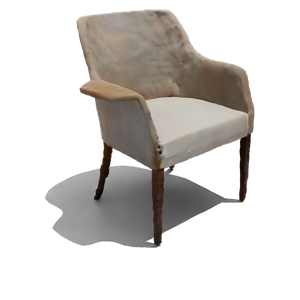}\includegraphics[width=0.1\linewidth]{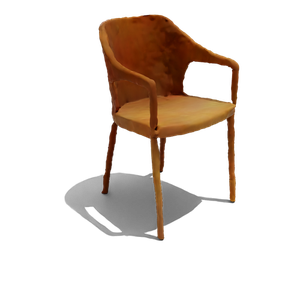}\includegraphics[width=0.1\linewidth]{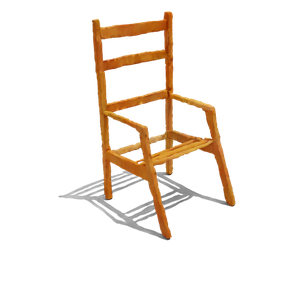}\includegraphics[width=0.1\linewidth]{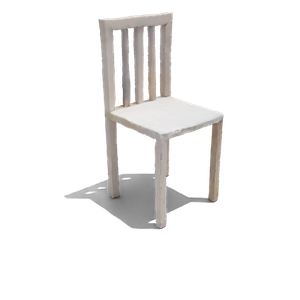}\includegraphics[width=0.1\linewidth]{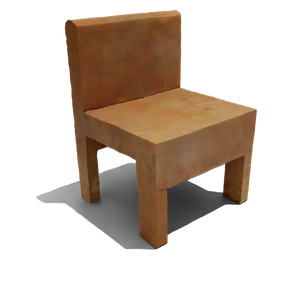}\includegraphics[width=0.1\linewidth]{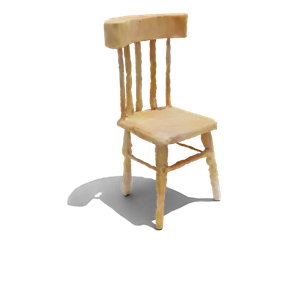}\includegraphics[width=0.1\linewidth]{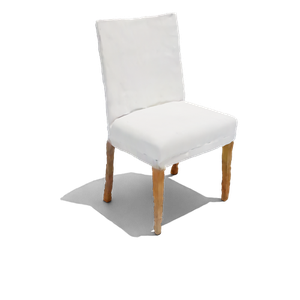}\includegraphics[width=0.1\linewidth]{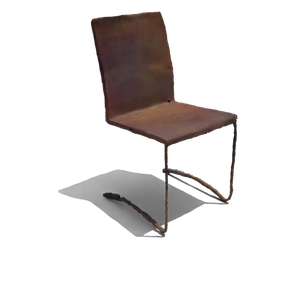}
\vspace{-1em}
  \captionsetup{type=figure}
  \captionof{figure}{\textbf{TetraDiffusion} is a 3D denoising diffusion model that operates on a tetrahedral grid to generate high-resolution 3D shapes in seconds. All depicted meshes are shown without any postprocessing, hole-filling or smoothing.}%
  \label{fig:teaser}
 \end{center}%

\begin{abstract}
Probabilistic denoising diffusion models (DDMs) have set a new standard for 2D image generation. Extending DDMs for 3D content creation is an active field of research. Here, we propose \emph{TetraDiffusion}, a diffusion model that operates on a tetrahedral partitioning of 3D space to enable efficient, high-resolution 3D shape generation. Our model introduces operators for convolution and transpose convolution that act directly on the tetrahedral partition, and seamlessly includes additional attributes like color. Our design generates mesh geometry much more efficiently: Compared to existing mesh diffusion techniques, \emph{TetraDiffusion} is up to 200$\times$ faster. At the same time, it reduces memory consumption and can operate at substantially higher resolution than existing mesh generators. Using only standard consumer hardware, it sets a new standard in terms of spatial detail and outperforms other mesh generators across a range of quality metrics. For additional results and code see our project page \url{tetradiffusion.github.io}.
\end{abstract}

\section{Introduction}

The growing demand for virtual content has sparked a wave of research to automate the laborious and costly generation of 3D assets. At the heart of this endeavor lies the search for powerful and flexible 3D representations to encode, store and manipulate the geometry and topology of 3-dimensional objects. In this work, we describe a novel, highly efficient generative model that is able to produce high-quality surface meshes in seconds. 

\begin{wrapfigure}{r}{0.5\textwidth}
\vspace{-2em}
\centering
\pgfplotsset{compat=1.5}
\begin{tikzpicture}[scale = 0.7]
\begin{axis}[
    xlabel={Execution Time (seconds)},
    ylabel={Average Clip-FID ($\downarrow$)},
    ymax=17,
    xmax=1000,
    xmode=log,
    log basis x=10,
    legend pos=north east,
    grid style=dashed,
    grid=both, %
    major grid style={dashed}, %
    minor grid style={dotted,gray}, %
    minor tick num=1, %
    scatter/classes={
        a={mark=square*,blue},
        b={mark=triangle*,red},
        c={mark=o,draw=black}
    },
]
\addplot[scatter,only marks,
    scatter src=explicit symbolic]
table[meta=label] {
x           y       label
0.8      9.51       a
714.3    7.37        b
9.1      3.03       b
3.4      4.35       b
0.24     14.48       a
3.6      7.29        b
};
\node[anchor=south] at (axis cs:0.8,9.51) {GET3D};
\node[anchor=east] at (axis cs:714.3,7.37) {Meshdiffusion};
\node[anchor=west] at (axis cs:9.1,3.03) {\textbf{\itshape Ours$_{192}$}};
\node[anchor=west] at (axis cs:3.4,4.35) {\textbf{\itshape Ours$_{128}$}};
\node[anchor=west] at (axis cs:0.24,14.48) {SDF-SG};
\node[anchor=south] at (axis cs:3.6,7.29) {NWD};
\legend{GAN based,Diffusion based}
\end{axis}
\end{tikzpicture}
\captionsetup{type=figure}
\vspace{-0.5em}
\caption{\textbf{Generation time per sample vs.\ mesh quality for different methods.} Clip-FID is averaged over the ShapeNet classes airplane, car and motorbike.}%
\vspace{-1.5em}
\label{fig:speedplot}
\end{wrapfigure}
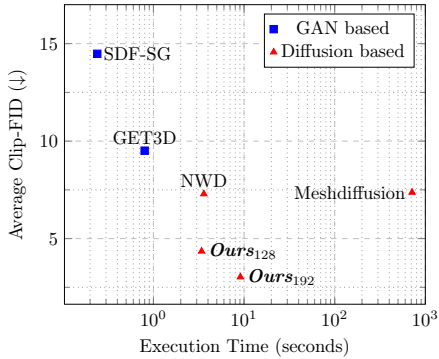
Following the rise of probabilistic denoising diffusion models (DDMs) for image generation \cite{ho2020denoising, song2020denoising, dhariwal2021diffusion, rombach2022high, Ramesh2022HierarchicalTI}, there have been several attempts to extend their generative capabilities to 3D \cite{zeng2022lion, zhou20213d, Liu2023MeshDiffusion, luo2021diffusion, gupta20233dgen, zhang20233dshape2vecset}.
It is relatively straightforward to transfer the DDM principle -- gradual per-point perturbations with Gaussian noise -- to voxels or points in 3D space, and this has already lead to interesting results \cite{zeng2022lion, zhou20213d, luo2021diffusion, zheng2023locally}.
However, these representations also come with their own disadvantages. Voxels are a natural extension of 2D pixels and amenable to well-established neural architectures based on discrete 3D convolutional operators, yet they are notoriously memory-hungry and therefore limited in terms of resolution. Point clouds, on the other hand, are sampled irregularly and avoid unnecessary discretization of empty space, but they lack connectivity information and have no direct notion of the underlying surfaces.
Both representations face additional challenges when converting them to a surface mesh, as one must trade off smoothing and loss of detail against surface noise and topological artifacts. An alternative could be to directly work with meshes, but handling explicit meshes is cumbersome and restricted by the fixed surface topology.

Instead, we turn to a hybrid representation that combines the advantages of both worlds, namely a tetrahedral decomposition of 3D space. We develop neural operators that directly act on the tetrahedral representation, and thus allow for fast and memory-efficient learning. This is in contrast to methods that embed the tetrahedral representation in a voxel grid of much higher resolution, thus inheriting the limitations of voxel models~\cite{Liu2023MeshDiffusion}.
Tetrahedral decompositions of 3D space have their origin in engineering and physics simulation and have also been used for volumetric modelling in graphics \cite{jacobson2011bounded, paille2015dihedral}, but they have only recently been adopted in the context of deep learning \cite{gao2020learning, shen2021deep, gao2022tetgan}. By combining the flexibility and structure of the tetrahedral grid with the generative power of DDMs, our model overcomes some of the limitations of existing 3D diffusion frameworks. A key ingredient of our method is a carefully predefined neighborhood topology of the space-filling tetrahedral decomposition~\cite{gao2020learning} that enables well-defined convolutional operators and makes it easy to extract a surface mesh with the help of a differentiable Marching Tetrahedra scheme \cite{gao2022tetgan}.

\begin{figure*}[t]
\centering
\includegraphics[width=0.14285714285714285\linewidth]{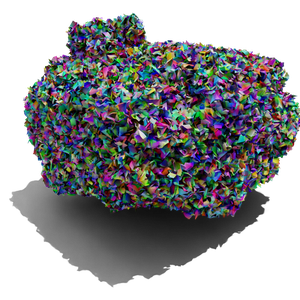}\includegraphics[width=0.14285714285714285\linewidth]{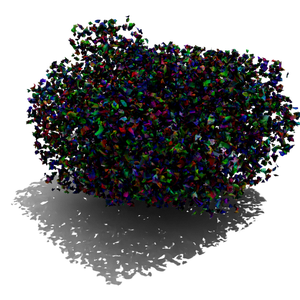}\includegraphics[width=0.14285714285714285\linewidth]{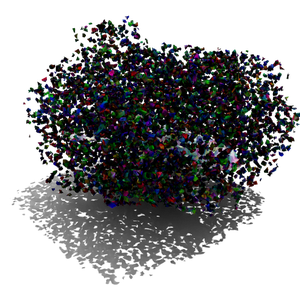}\includegraphics[width=0.14285714285714285\linewidth]{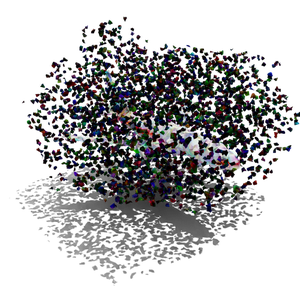}\includegraphics[width=0.14285714285714285\linewidth]{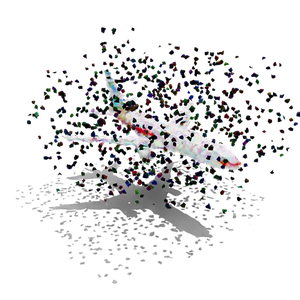}\includegraphics[width=0.14285714285714285\linewidth]{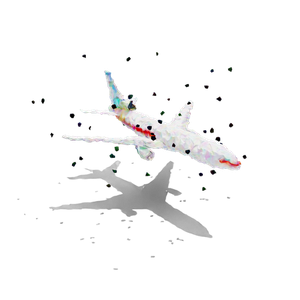}\includegraphics[width=0.14285714285714285\linewidth]{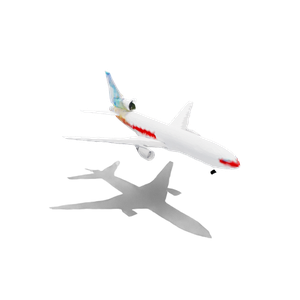}

\begin{tikzpicture}
    \draw[->, >=latex, thick] (0,0) -- (0.93\linewidth,0);

    \draw (0,3pt) -- (0,-3pt) node[anchor=north] {0s};
    \draw (0.93\linewidth,3pt) -- (0.93\linewidth,-3pt) node[anchor=north] {5.2s};
\end{tikzpicture}
\caption{\textbf{Reverse tetrahedral diffusion sequence.} Starting from a noisy tetrahedral grid, the model recovers a textured mesh in a few seconds. Note the shape of the initial grid: our method allows more targeted grid pruning than prior art.}\label{fig:trajectory}%
\vspace{-0.5cm}
\end{figure*}
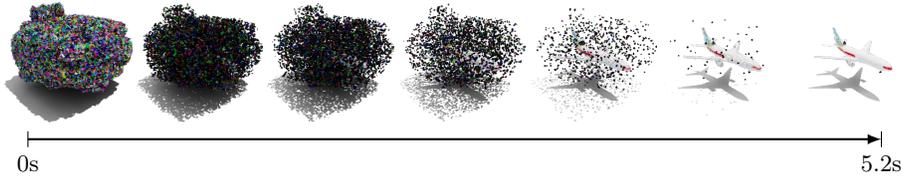

In the spirit of KPConv~\cite{thomas2019kpconv} and graph convolutions~\cite{kipf2016semi, wu2019simplifying, bruna2013spectral}, we equip that representation with convolution (and transposed convolution) kernels that operate directly on the deformable tetrahedral structure. These operations make it possible to construct a U-Net~\cite{ronneberger2015u} in the tetrahedralized space, which, in turn, is the computational core of a denoising diffusion model. That DDM transforms random noise into a 3D object shape (\cref{fig:trajectory}) by predicting both a per-vertex signed distance field and an individual per-vertex displacement.

Our formulation has several benefits. Like in a point cloud, the vertices of the tetrahedral grid can be moved around to align with object surfaces. At the same time they retain a uniquely defined neighborhood connectivity, which ensures that one can readily extract a topologically sound surface mesh, and obviates the need to train an additional surface reconstruction network~\cite{peng2021shape}.
Moreover, our network can be deployed on a multi-resolution hierarchy of tetrahedral decompositions, which makes both training and inference memory-efficient and computationally affordable, \cf~\cref{fig:speedplot}.
Empowered by the modest computational cost we are able to run diffusion at higher native resolution (in our experiments $>5\cdot10^6$ tetrahedra) and to reconstruct 3D shapes with unprecedented detail.
Moreover, our proposed tetrahedral convolution layers offer the possibility to exploit sparsity and prune tetrahedra in unoccupied regions of 3D space (\cref{fig:trajectory}), further reducing training time.

Going beyond purely geometric object properties, as captured by signed distance values and vertex offsets, we extend the tetrahedral representation and the associated Marching Tetrahedra scheme~\cite{shen2021deep} to unrestricted feature vectors, so as to diffuse and extract further attributes like color. The extended Marching Tetrahedra algorithm remains fully differentiable. Consequently, the model is fully compatible with guided and conditional diffusion schemes, as we show in the supplementary material, e.g. by guiding the sampled shapes in size, volume or specific colors. Moreover one can drive the shapes to resemble existing examples or renderings, similar to classifier-free guidance \cite{ho2022classifier} and regularization \cite{rombach2022high}.
In summary, our contributions are:
\begin{enumerate}
    \item To the best of our knowledge, we propose the first 3D denoising diffusion model that operates entirely on a tetrahedral representation.
    \item We design convolution operators and up- and downsampling kernels on the tetrahedral grid.
    \item We extend our tetrahedral DDM to include color, which enables the generation of textured assets.
    \item We show that \emph{TetraDiffusion} enables efficient training and fast inference at unprecedented resolution, on consumer hardware. 
\end{enumerate}

\section{Related work}

\paragraph{3D Generative Models.}

Compared to their 2D counterparts, 3D generative models have to chose among a wider range of data representations. Most dominant are voxels and point clouds, for which it is straightforward to adapt the 2D formulations~\cite{yang2019pointflow, zamorski2020adversarial, achlioptas2018learning}. Early work directly treats point clouds as matrices to make them amenable to standard neural architectures \cite{gadelha2018multiresolution}, or rasterizes 3D data into a voxel grid in order to apply conventional 3D convolutions \cite{wu20153d, brock2016generative}. To overcome fixed structures and to induce permutation invariance, PointFlow \cite{yang2019pointflow} first samples from a shape distribution and in a second step samples from the distribution of points given the corresponding shape prior. 

Another prominent stream of work relies on implicit representations. The authors of \cite{chen2019learning} introduce an implicit field decoder to learn a signed distance function. The network is trained in adversarial fashion to predict the distance from the surface when presented a point coordinate and its associated feature encoding. In \cite{cai2020learning}, shapes are represented as gradient fields over the logarithmic surface density, taking advantage of score-based generative models \cite{song2020denoising}. Similarly, \cite{achlioptas2018learning} learn a shape prior with an auto-encoder and GAN in latent space. Our model does not rely on adversarial training or latent interpolation, rather we train directly on input encodings.

GET3D~\cite{gao2022get3d} explores a combination of explicit and implicit representations. Similar to our work, 3D shapes are described via signed distances and deformation vectors on a tetrahedral grid with fixed topology. Their tetrahedral representation is encoded in separate triplane representations for geometry and texture that are trained in an adversarial manner with rendering losses, making use of the differentiable Marching Tetrahedra~\cite{shen2021deep}. Instead of projecting into triplanes, TetGAN~\cite{gao2022tetgan} directly operates on the tetrahedralized cuboid with an auto-encoder, supervised with ground truth features as well as global and local adversarial losses. Somewhat similar to our work they define convolutions and up-/ down-sampling operators on entire tetrahedra, which in their setting is straightforward as each tetrahedron has exactly four neighbours. In contrast, we extend those operations to act on vertices directly so as to allow for displacements, in the spirit of DMTet~\cite{shen2021deep}. Our more general formulation naturally offers more flexibility and representation power, since every vertex can be aligned individually in contrast to the tetrahedral occupancy field in TetGAN, which only allows aligning a tetrahedron as a whole. 

\paragraph{3D Diffusion Models.}

With the rise of DDMs, various 3D representations have been transferred to the diffusion setting~\cite{zeng2022lion, luo2021diffusion, zhou20213d, mo2023dit}. Point-voxel diffusion incorporates a point-voxel CNN \cite{zhou20213d} to directly apply diffusion on the hybrid representation. Similarly, \cite{luo2021diffusion} directly diffuse point clouds conditioned on a latent shape representation generated with a normalizing flow. On the other hand, PolyDiff \cite{alliegro2023polydiff} runs diffusion directly on a soup of (quantized) triangles. LION \cite{zeng2022lion} uses DDMs in latent space and maps the latent encoding back to a point cloud with a VAE. By itself this generates noisy point clouds, which is why they must train an additional network~\cite{peng2021shape} to turn those into smooth meshes. To circumvent that extra surface extractor, 3DGen~\cite{gupta20233dgen} introduces a two-stage training pipeline consisting of a triplane VAE and a separately trained latent diffusion model in triplane space. Somewhat similarly, \cite{shue20233d} first train a decoder that maps triplane features to occupancy grids, then train a 2D diffusion model to generate those triplanes. In contrast, our method only requires \textit{single-stage training} and no separate decoder, by natively operating on the tetrahedral 3D representation. 

Perhaps the closest work to ours is MeshDiffusion~\cite{Liu2023MeshDiffusion}, which also maps meshes to a deformable tetrahedral grid and performs diffusion on that representation. However, to circumvent the lack of convolutional operators for that grid they embed the tetrahedra in a higher-resolution regular voxel grid. This makes it possible to employ conventional 3D convolutions, but largely sacrifices the benefit of the tetrahedral representation: the voxelization is extremely inefficient, as it leads to a cubic increase in memory footprint and computation without adding any information. By defining convolutions directly on the spatially sparse vertices of the tetrahedral grid our model avoids that large overhead and allows for a finer tetrahedralization that captures higher-resolution details at much lower memory footprint (\cf \cref{fig:speedplot} and \cref{table:efficiency}). 

\begin{figure*}[!t]
    \begin{subfigure}[b]{0.70\textwidth}
        \includegraphics[width=\linewidth]{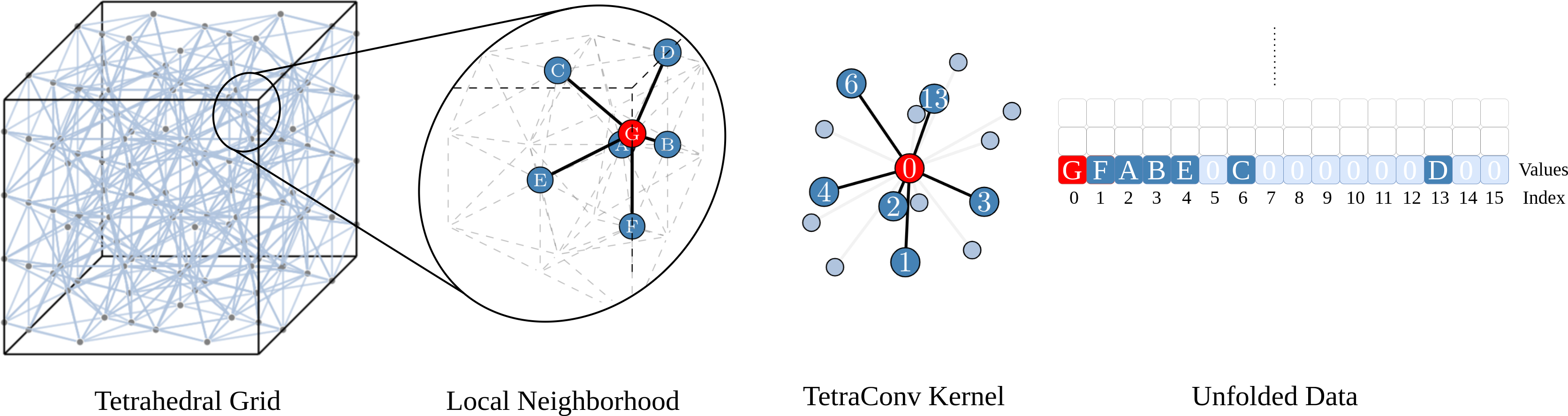}
        \caption{}\label{fig:tetraconv:a}
    \end{subfigure}
    \hspace{1mm}
    \begin{subfigure}[b]{0.26\textwidth}
    
        \resizebox{1.0\textwidth}{!}{
            \input{Sections/images/updown}
        }
        \caption{}\label{fig:tetraconv:b}
    \end{subfigure}
    \caption{\textbf{(a) Tetrahedral convolution for kernel size $m=16$.} The TetraConv kernel defines the ordering and padding for every local neighborhood. The center vertex, where the convolution aggregates, is marked in red and the  neighborhood in dark blue. Missing vertices (light blue) are padded with zeros.
    \textbf{(b) Strided convolution $H\!=\!\text{TetraConv}(A,B,C,D,E)$ and transposed convolution $K\!=\!\text{TetraConv}(F,G,H,I,J)$.} Both operators accumulate information from the nearest neighbors in their preceding layers. Operators are depicted in 2D. %
    }
    \label{fig:tetraconv}
\end{figure*}

\section{Tetrahedral Representation Learning}
\label{sec:tetrahedrallearning}

\paragraph{Tetrahedral Grid.}
Following \cite{shen2021deep}, we represent shapes within a given cuboid $\mathcal{T}$ with a signed distance field and a displacement field that are both defined on the vertices of the same, space-filling tetrahedral decomposition of $\mathcal{T}$. We refer to that structure as the \emph{tetrahedral grid}, with vertices $V_\mathcal{T}\in \mathbb{R}^{N\times 3}$ and tetrahedra $T\in \mathbb{N}^{K\times 4}$. A nearly regular decomposition can be found via close-packed tetrahedral tiling with the A15 lattice~\cite{doran2013isosurface}. Tetrahedral grid resolution refers to the grid spacing used in the iso-surface stuffing algorithm. A tetrahedralized cube of resolution $R$ contains $0.72\cdot R^3$ tetrahedra. Note that a given number of tetrahedra can capture more detail than the same number of voxels, as they can be deformed to better follow the surface. 
 Each tetrahedron $T_k$ consists of four vertices $\{v_{k1}, v_{k2}, v_{k3}, v_{k4}\}$ and corresponding edges to form a simplex, \ie, the connectivity is predefined and fixed during training and inference. Each vertex is assigned a displacement $\Delta_v$ and a signed distance value $s_v$. The SDF provides an implicit surface representation, whereas the displacements serve to precisely align it with the actual object surface. Conveniently, one can extract a surface \textit{mesh} from $\mathcal{T}$ with Deep Marching Tetrahedra (DMTet, \cite{shen2021deep}). 
In \Cref{sec:tetrahedrallearning}, we extend the tetrahedral representation to arbitrary feature vectors, such that additional vertex attributes (\eg, color) can be propagated to the final mesh.

\paragraph{Tetra Convolution.}
Our displacement and signed distance fields are not directly amenable to regular 3D convolutions due to varying neighborhoods and connectivity in the grid. However, the superimposed A15 lattice permits a collision-free spatial ordering of vertices within each neighborhood, and as such a well-defined, rigorous tetrahedral convolution. This is in contrast to graph convolutions that ignore the spatial configuration of the neighborhood, but still eliminates the need to search for the relevant points under the kernel and apply distance weighting, as in KPConv \cite{thomas2019kpconv}. We define our unique ordering, a discrete binning of the local edge orientations, by iteratively clustering the set of all outgoing edges with $k$-means until we find a collision-free ordering, \ie no edges within a neighborhood fall within the same bin. Interestingly, in practice that clustering always converges to a basis of at most $m+1$ reference directions, where $m$ is the maximal number of neighbors in the tetrahedral grid. Compared to 3D voxel convolutions, the resulting kernel size is more than 40\% lower (\eg, 15 compared to 26), a key property for efficient learning. A detailed description of our ordering scheme is given in the Appendix. %

Given a vertex $v_k$ and its neighbors $\{v_j,j\in\mathcal{U}_{v_k}\}$ in a fixed ordering, it is straightforward to define the result of a tetrahedral convolution layer $l$ as a weighted sum over their associated feature values on the previous layer, $\Phi_{v_j}^{l-1}$: 
\begin{equation}\label{eq:tetraconv}
    \Phi_{v_k}^l = W_0 \Phi_{v_k}^{l-1} + \sum_{j=1}^{m+1} \mathbb{1}_{j\in \mathcal{U}_{v_k}} W_j \Phi_{v_j}^{l-1},
\end{equation}
where $W_j$ are the kernel weights. We refer to this operation as \emph{TetraConv}. As illustrated in \cref{fig:tetraconv:a}, the varying size of the neighborhood simply corresponds to appropriately zero-padding the kernel. To fully exploit the power of convolutional learning, we need to define down- and upsampling operations on the tetrahedral grid. We construct tetrahedral tesselations of our cuboid $\mathcal{T}$ at multiple resolutions and establish parent-child relations between vertices in adjacent hierarchy levels via $k$-nearest neighbor search (using the Euclidean distance). As shown in \Cref{fig:tetraconv:b}, this enables flexible down- and upsampling rates ; and, with appropriate book-keeping about the neighborhood indices $\mathcal{U}_{v,k}$ in \cref{eq:tetraconv}, strided convolutions and transposed convolutions, thus going beyond naive upsampling.

\paragraph{Grid Pruning.}

Like any volumetric tesselation, the tetrahedral grid is in most cases sparsely populated, reflecting the fact that only a small part of 3D space lies near an object surface. Unlike voxel-based convolutions, where only axis-parallel cropping is efficient (in particular limited to the bounding box), our tetrahedral formalism facilitates pruning in a more targeted manner. Deleting all unoccupied vertices and their corresponding connections translates to two simple operations: (1) unused vertices are completely removed by deleting the corresponding row in the unfolded kernel-data matrix and (2) removed connections lead to additional zero padding in affected rows. As a result, we can truncate our grid to the convex hull of all training data in a lossless manner. Additionally, training and inference speed and memory consumption are easily reduced further by (lossy) pruning of vertices that are occupied by fewer than a user-defined number of instances.

\paragraph{Tetrahedral Diffusion.}
With all necessary building blocks at hand, we can now assemble our tetrahedral diffusion model. We directly diffuse in deformable tetrahedral space, \ie, our basic input consists of $N$ vertices with features $\{s_v, \Delta_v\}$, plus their associated neighborhood relations. However, we can seamlessly extend the feature vector with further vertex propoerties without losing the differentiability of the marching tetrahedra algorithm. In particular, surface information like color is a convex combination of the corresponding tetrahedral vertex features. In our experiments we therefore directly diffuse SDF values, deformation vectors and color vectors per vertex, \ie $\{s_v, \Delta_v, c_v\}$. 

Our network largely adheres to the standard U-ViT architecture of many diffusion models~\cite{hoogeboom2023simple}, but with the tetrahedral convolution operators introduced above. \Ie, residual convolution layers with group normalization~\cite{wu2018group} and SiLU activations~\cite{hendrycks2016gaussian}, alternating with attention layers. The network, hyperparameters and extended marching tetrahedra scheme are further described in the Appendix. %

\section{Experiments}
We now experimentally demonstrate the capabilities of TetraDiffusion. We train and evaluate our method on the classes \textit{airplane}, \textit{bike}, \textit{car} and \textit{chair} from the ShapeNet \cite{shapenet2015} dataset, using the official ShapeNet train/val/test split. Each shape is individually normalized to lie in $[-1, 1]$. Our network requires SDFs and displacement fields on a tetrahedral grid, \ie, ShapeNet meshes have to be converted to that format. We adapt the rendering pipeline of \cite{munkberg2022extracting} and fit each shape individually into the grid with a combination of rendering and volumetric losses. Similar to \cite{Liu2023MeshDiffusion}, the ground truth is created with a two-step procedure, fixing SDF values to $\{-1, 1\}$. For a comprehensive overview of the pre-processing protocol, please refer to the supplementary material.

\newpage
\subsection{Qualitative results}

\begin{figure}
\centering
\begin{tikzpicture}
\fill[lightgrey, opacity=0.25, rounded corners=4mm] (6.05,-0.1) rectangle (8.5,6.2);
\fill[lightgrey, opacity=0.25, rounded corners=4mm] (9.7,-0.1) rectangle (12.2,6.2);
\vspace{-0.15cm}
\node[anchor=south west,inner sep=0] (image) at (0,5) 
{\begin{minipage}{0.1\linewidth}\centering\vspace{-1.2cm}\tiny{not\\available}\end{minipage}\includegraphics[width=0.1\linewidth,trim={ 0.56cm 1.46cm 1.18cm 1.06cm },clip]{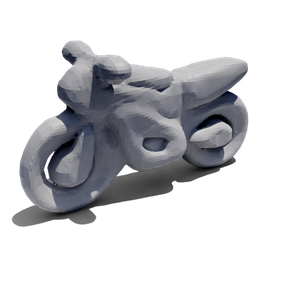}\includegraphics[width=0.1\linewidth,trim={ 0.56cm 1.46cm 1.18cm 1.06cm },clip]{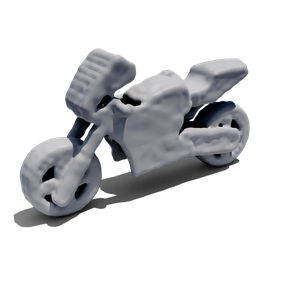}\includegraphics[width=0.1\linewidth,trim={ 0.56cm 1.46cm 1.18cm 1.06cm },clip]{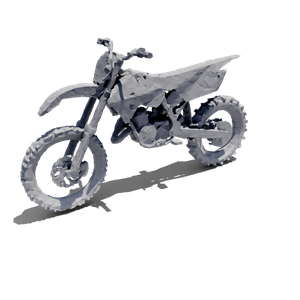}\includegraphics[width=0.1\linewidth,trim={ 0.56cm 1.46cm 1.18cm 1.06cm },clip]{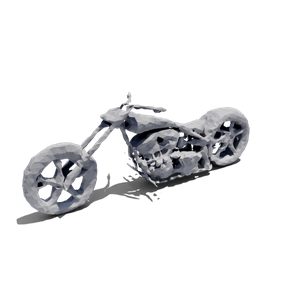}\includegraphics[width=0.1\linewidth,trim={ 0.56cm 1.46cm 1.18cm 1.06cm },clip]{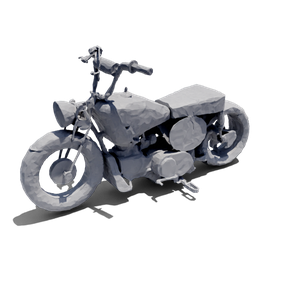}\includegraphics[width=0.1\linewidth,trim={ 0.56cm 1.46cm 1.18cm 1.06cm },clip]{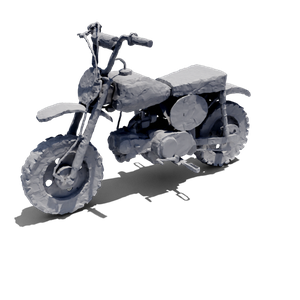}\includegraphics[width=0.1\linewidth,trim={ 0.56cm 1.46cm 1.18cm 1.06cm },clip]{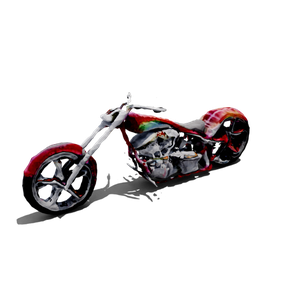}\includegraphics[width=0.1\linewidth,trim={ 0.56cm 1.46cm 1.18cm 1.06cm },clip]{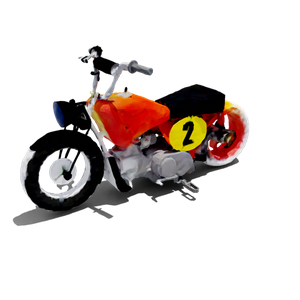}\includegraphics[width=0.1\linewidth,trim={ 0.56cm 1.46cm 1.18cm 1.06cm },clip]{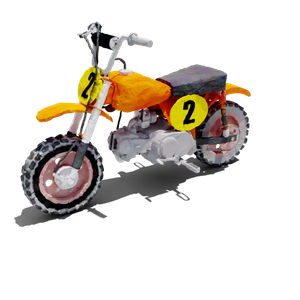}};

\vspace{-0.15cm}
\node[anchor=south west,inner sep=0] (image) at (0,4) 
{\begin{minipage}{0.1\linewidth}\centering\vspace{-1.2cm}\tiny{not\\available}\end{minipage}\includegraphics[width=0.1\linewidth,trim={ 0.56cm 1.46cm 1.18cm 1.06cm },clip]{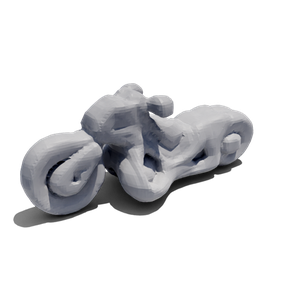}\includegraphics[width=0.1\linewidth,trim={ 0.56cm 1.46cm 1.18cm 1.06cm },clip]{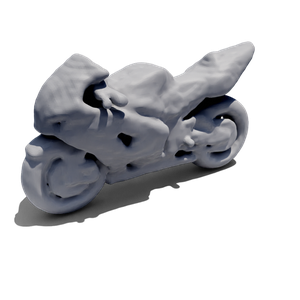}\includegraphics[width=0.1\linewidth,trim={ 0.56cm 1.46cm 1.18cm 1.06cm },clip]{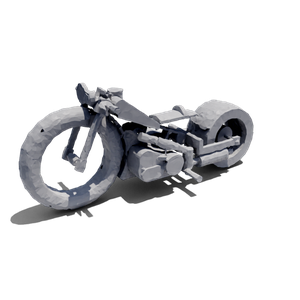}\includegraphics[width=0.1\linewidth,trim={ 0.56cm 1.46cm 1.18cm 1.06cm },clip]{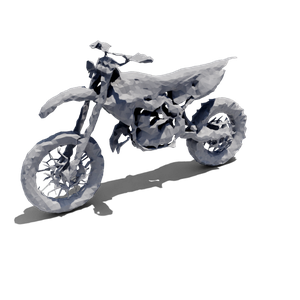}\includegraphics[width=0.1\linewidth,trim={ 0.56cm 1.46cm 1.18cm 1.06cm },clip]{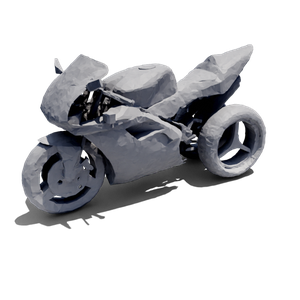}\includegraphics[width=0.1\linewidth,trim={ 0.56cm 1.46cm 1.18cm 1.06cm },clip]{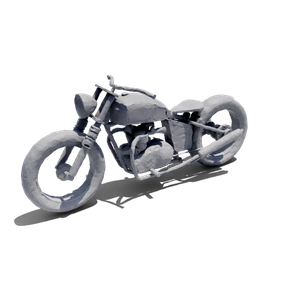}\includegraphics[width=0.1\linewidth,trim={ 0.56cm 1.46cm 1.18cm 1.06cm },clip]{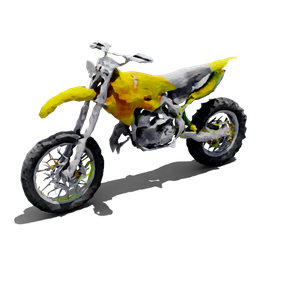}\includegraphics[width=0.1\linewidth,trim={ 0.56cm 1.46cm 1.18cm 1.06cm },clip]{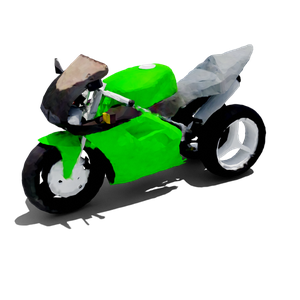}\includegraphics[width=0.1\linewidth,trim={ 0.56cm 1.46cm 1.18cm 1.06cm },clip]{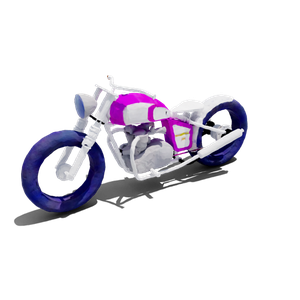}};

\vspace{-0.15cm}
\node[anchor=south west,inner sep=0] (image) at (0,3) 
{\includegraphics[width=0.1\linewidth,trim={ 0.84cm 1.40cm 1.40cm 1.40cm },clip]{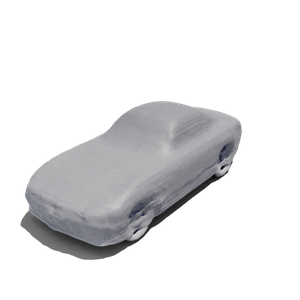}\includegraphics[width=0.1\linewidth,trim={ 0.84cm 1.40cm 1.40cm 1.40cm },clip]{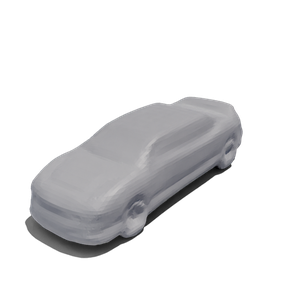}\includegraphics[width=0.1\linewidth,trim={ 0.84cm 1.40cm 1.40cm 1.40cm },clip]{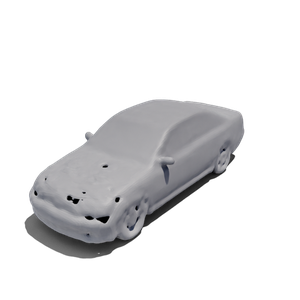}\includegraphics[width=0.1\linewidth,trim={ 0.84cm 1.40cm 1.40cm 1.40cm },clip]{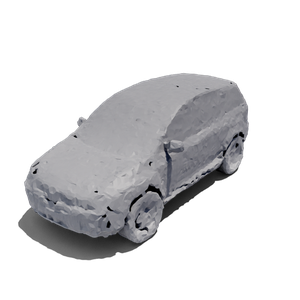}\includegraphics[width=0.1\linewidth,trim={ 0.84cm 1.40cm 1.40cm 1.40cm },clip]{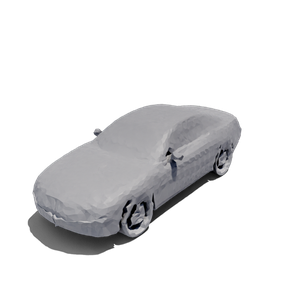}\includegraphics[width=0.1\linewidth,trim={ 0.84cm 1.40cm 1.40cm 1.40cm },clip]{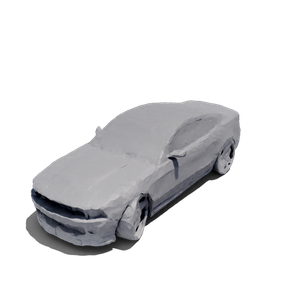}\includegraphics[width=0.1\linewidth,trim={ 0.84cm 1.40cm 1.40cm 1.40cm },clip]{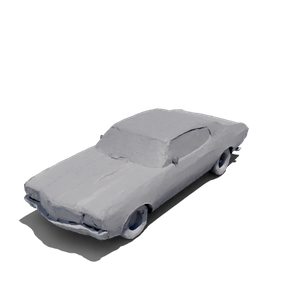}\includegraphics[width=0.1\linewidth,trim={ 0.84cm 1.40cm 1.40cm 1.40cm },clip]{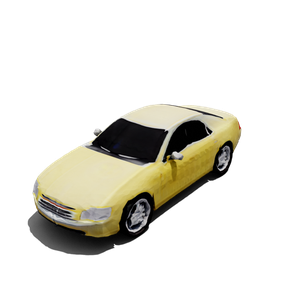}\includegraphics[width=0.1\linewidth,trim={ 0.84cm 1.40cm 1.40cm 1.40cm },clip]{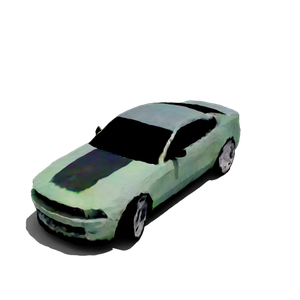}\includegraphics[width=0.1\linewidth,trim={ 0.84cm 1.40cm 1.40cm 1.40cm },clip]{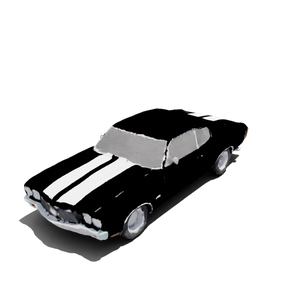}};

\vspace{-0.15cm}
\node[anchor=south west,inner sep=0] (image) at (0,2) 
{\includegraphics[width=0.1\linewidth,trim={ 0.84cm 1.40cm 1.40cm 1.40cm },clip]{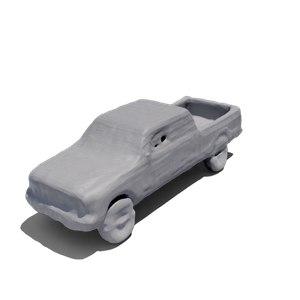}\includegraphics[width=0.1\linewidth,trim={ 0.84cm 1.40cm 1.40cm 1.40cm },clip]{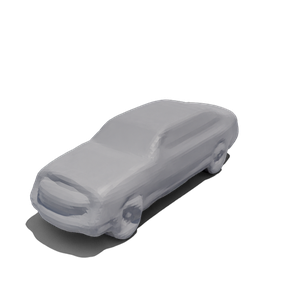}\includegraphics[width=0.1\linewidth,trim={ 0.84cm 1.40cm 1.40cm 1.40cm },clip]{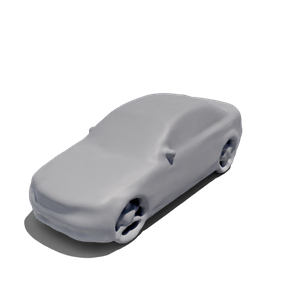}\includegraphics[width=0.1\linewidth,trim={ 0.84cm 1.40cm 1.40cm 1.40cm },clip]{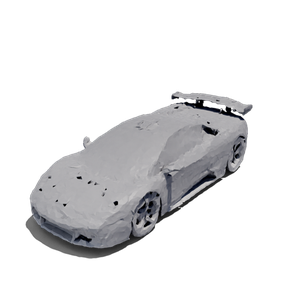}\includegraphics[width=0.1\linewidth,trim={ 0.84cm 1.40cm 1.40cm 1.40cm },clip]{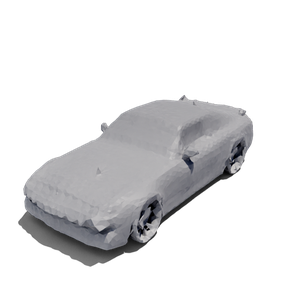}\includegraphics[width=0.1\linewidth,trim={ 0.84cm 1.40cm 1.40cm 1.40cm },clip]{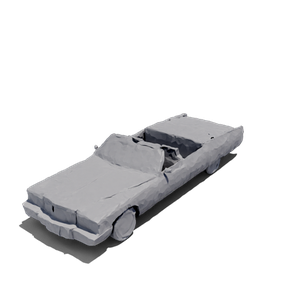}\includegraphics[width=0.1\linewidth,trim={ 0.84cm 1.40cm 1.40cm 1.40cm },clip]{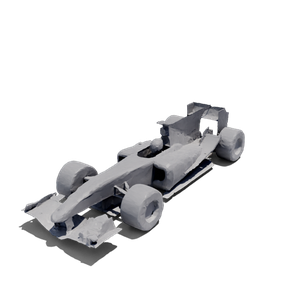}\includegraphics[width=0.1\linewidth,trim={ 0.84cm 1.40cm 1.40cm 1.40cm },clip]{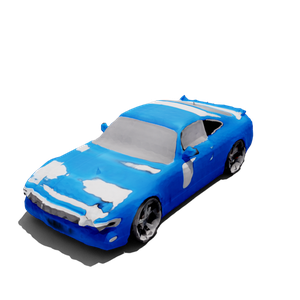}\includegraphics[width=0.1\linewidth,trim={ 0.84cm 1.40cm 1.40cm 1.40cm },clip]{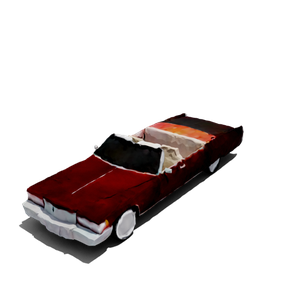}\includegraphics[width=0.1\linewidth,trim={ 0.84cm 1.40cm 1.40cm 1.40cm },clip]{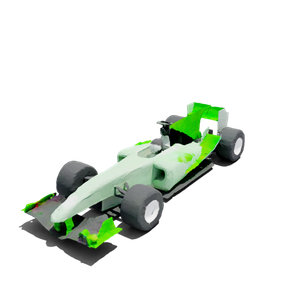}};

\vspace{-0.15cm}
\node[anchor=south west,inner sep=0] (image) at (0,1) 
{\includegraphics[width=0.1\linewidth,trim={ 0.87cm 1.43cm 1.15cm 1.43cm },clip]{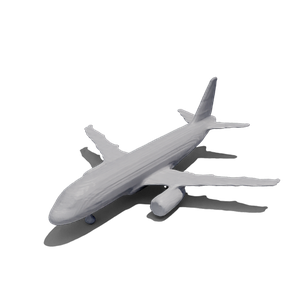}\includegraphics[width=0.1\linewidth,trim={ 0.87cm 1.43cm 1.15cm 1.43cm },clip]{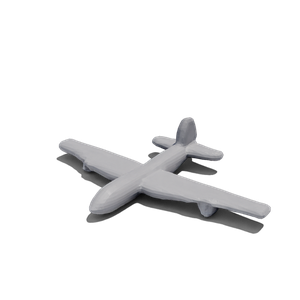}\includegraphics[width=0.1\linewidth,trim={ 0.87cm 1.43cm 1.15cm 1.43cm },clip]{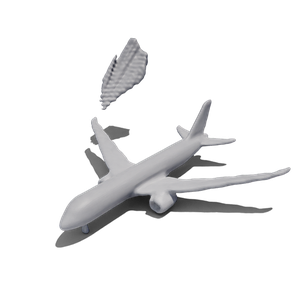}\includegraphics[width=0.1\linewidth,trim={ 0.87cm 1.43cm 1.15cm 1.43cm },clip]{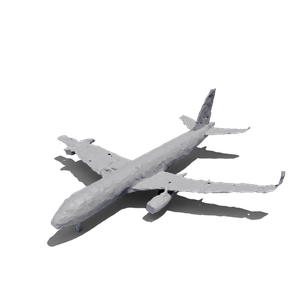}\includegraphics[width=0.1\linewidth,trim={ 0.87cm 1.43cm 1.15cm 1.43cm },clip]{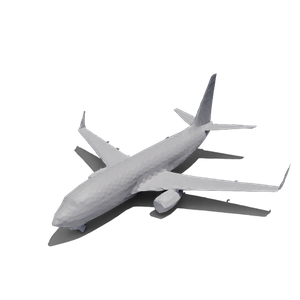}\includegraphics[width=0.1\linewidth,trim={ 0.87cm 1.43cm 1.15cm 1.43cm },clip]{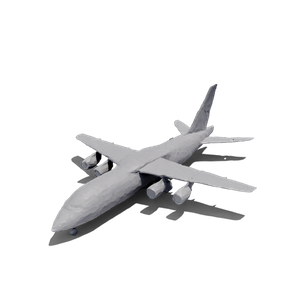}\includegraphics[width=0.1\linewidth,trim={ 0.87cm 1.43cm 1.15cm 1.43cm },clip]{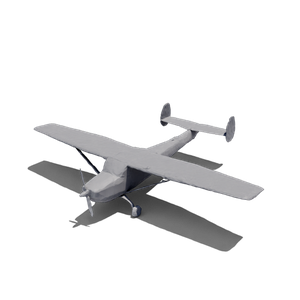}\includegraphics[width=0.1\linewidth,trim={ 0.87cm 1.43cm 1.15cm 1.43cm },clip]{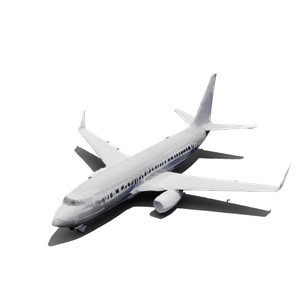}\includegraphics[width=0.1\linewidth,trim={ 0.87cm 1.43cm 1.15cm 1.43cm },clip]{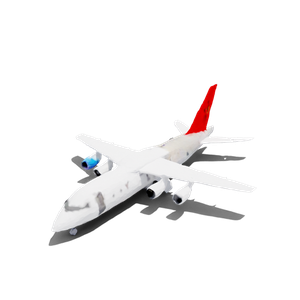}\includegraphics[width=0.1\linewidth,trim={ 0.87cm 1.43cm 1.15cm 1.43cm },clip]{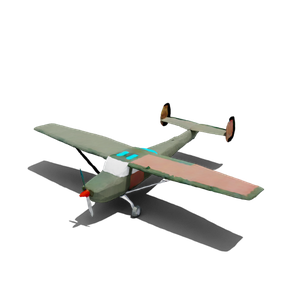}};

\vspace{-0.15cm}
\node[anchor=south west,inner sep=0] (image) at (0,0) 
{\includegraphics[width=0.1\linewidth,trim={ 0.87cm 1.43cm 1.15cm 1.43cm },clip]{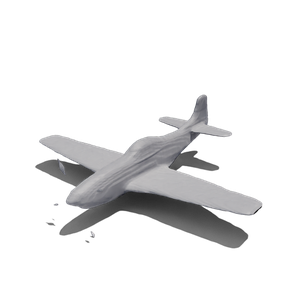}\includegraphics[width=0.1\linewidth,trim={ 0.87cm 1.43cm 1.15cm 1.43cm },clip]{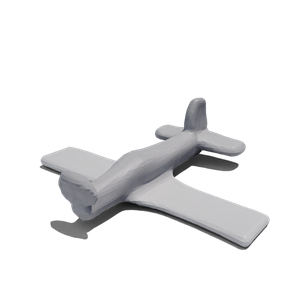}\includegraphics[width=0.1\linewidth,trim={ 0.87cm 1.43cm 1.15cm 1.43cm },clip]{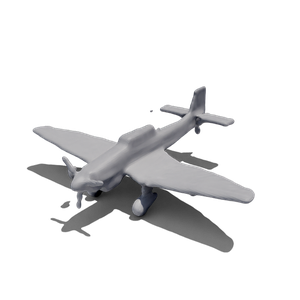}\includegraphics[width=0.1\linewidth,trim={ 0.87cm 1.43cm 1.15cm 1.43cm },clip]{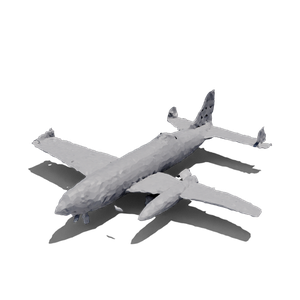}\includegraphics[width=0.1\linewidth,trim={ 0.87cm 1.43cm 1.15cm 1.43cm },clip]{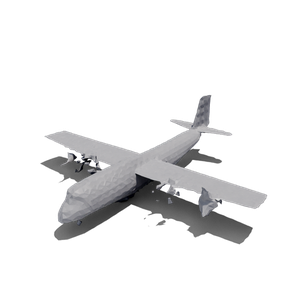}\includegraphics[width=0.1\linewidth,trim={ 0.87cm 1.43cm 1.15cm 1.43cm },clip]{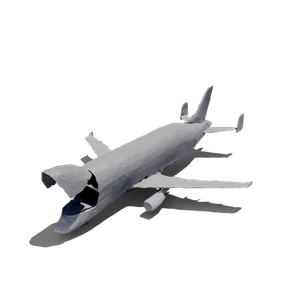}\includegraphics[width=0.1\linewidth,trim={ 0.87cm 1.43cm 1.15cm 1.43cm },clip]{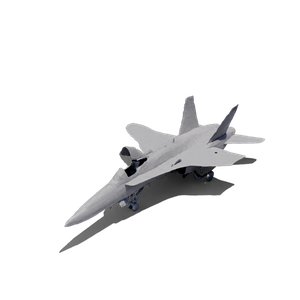}\includegraphics[width=0.1\linewidth,trim={ 0.87cm 1.43cm 1.15cm 1.43cm },clip]{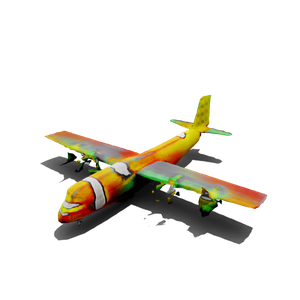}\includegraphics[width=0.1\linewidth,trim={ 0.87cm 1.43cm 1.15cm 1.43cm },clip]{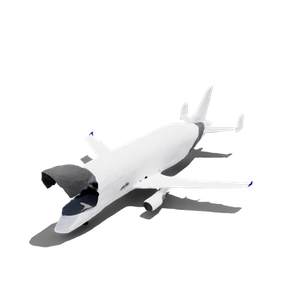}\includegraphics[width=0.1\linewidth,trim={ 0.87cm 1.43cm 1.15cm 1.43cm },clip]{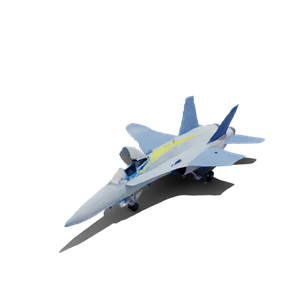}};

\node[anchor=south west,inner sep=0] (image) at (0,6.5) 
{\begin{minipage}{0.09\linewidth}
\centering
\tiny{NFD}
\end{minipage}
\begin{minipage}{0.09\linewidth}
\centering
\tiny{SDF-SG}
\end{minipage}
\begin{minipage}{0.09\linewidth}
\centering
\tiny{NWD}
\end{minipage}
\begin{minipage}{0.09\linewidth}
\centering
\tiny{MD$_{128}$}
\end{minipage}
\begin{minipage}{0.09\linewidth}
\centering
\tiny{GET3D}
\end{minipage}
\begin{minipage}{0.09\linewidth}
\centering
\tiny{\textbf{Ours\textsubscript{128}}}
\end{minipage}
\begin{minipage}{0.09\linewidth}
\centering
\tiny{\textbf{Ours\textsubscript{192}}}
\end{minipage}
\begin{minipage}{0.09\linewidth}
\centering
\tiny{GET3D}
\end{minipage}
\begin{minipage}{0.09\linewidth}
\centering
\tiny{\textbf{Ours\textsubscript{128}}}
\end{minipage}
\begin{minipage}{0.09\linewidth}
\centering
\tiny{\textbf{Ours\textsubscript{192}}}
\end{minipage}};

    \end{tikzpicture}
\caption{\textbf{Column-wise qualitative comparison between \emph{TetraDiffusion} and existing generative mesh models.} Our method generates the finest geometry while seamlessly integrating texture. Please zoom in to appreciate the high resolution quality.}
\label{fig:randomsamples}
\end{figure}

Figures \ref{fig:teaser} and \cref{fig:randomsamples} showcase randomly generated 3D shapes, demonstrating the high quality and level of detail achievable with our model. Note the diversity of shapes within each class, and the intricate details like chains and brake discs on motorbikes or propellers and wing appendages on airplanes.
The motorbikes nicely illustrate the model's generative abilities: although the training set is small ($\approx$ 200 examples), the results go beyond memorization and naive interpolation. Parts of different training samples are seamlessly blended into plausible assemblies, and curious designs emerge that clearly were not observed during training. We emphasize that the presented shapes have \emph{not} been post-processed in any way. Cleaning up the output of 3D generative models in post-processing is a common practice, but has side effects. \Eg, smoothing would remove fine structures and high-frequency details that the model may have correctly synthesized.

\paragraph{Geometry comparison.}

\begin{figure}
\newcommand{\includeplanes}[2][]{%
  \includegraphics[width=0.16\linewidth, trim=0.90cm 2.10cm 1.35cm 2.70cm, clip, #1]{#2}%
}
\newcommand{\includeplaneszoom}[2][]{%
  \includegraphics[width=0.16\linewidth, trim=5.40cm 3.15cm 3.30cm 5.55cm, clip, #1]{#2}%
}
\includeplanes{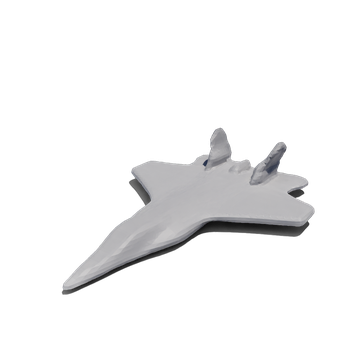}
\includeplanes{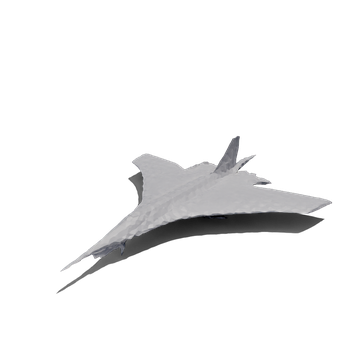}
\includeplanes{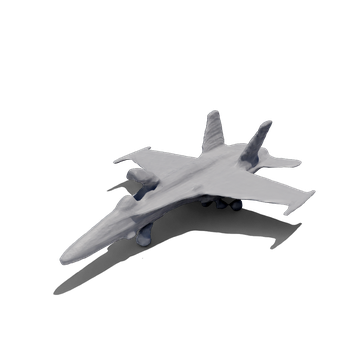}
\includeplanes{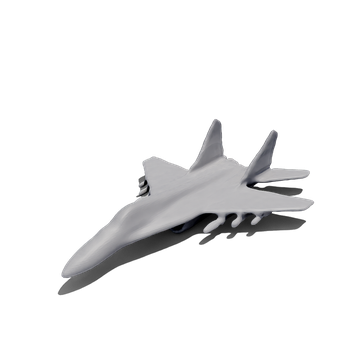}
\includeplanes{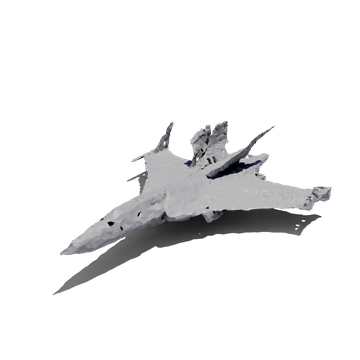}
\includeplanes{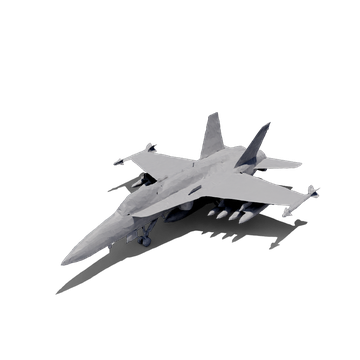}
\includeplaneszoom{Images/comparision/planes2/sdf_plane_details.png}
\vspace{-1.0em}
\includeplaneszoom{Images/comparision/planes2/15_get3d_0.png}
\includeplaneszoom{Images/comparision/planes2/15_nfd_0.png}
\includeplaneszoom{Images/comparision/planes2/41.png}
\includeplaneszoom{Images/comparision/planes2/15_md_0.png}
\includeplaneszoom{Images/comparision/planes2/15_192_1.png}

\newcommand{\includecars}[2][]{%
  \includegraphics[width=0.16\linewidth, trim=0.75cm 1.50cm 2.10cm 2.40cm, clip, #1]{#2}%
}
\newcommand{\includecarszoom}[2][]{%
  \includegraphics[width=0.16\linewidth, trim=3.60cm 1.95cm 5.10cm 6.75cm, clip, #1]{#2}%
}

\newcommand{\includecarssuvzoom}[2][]{%
  \includegraphics[width=0.16\linewidth, trim=3.00cm 1.95cm 5.70cm 6.75cm, clip, #1]{#2}%
}

\includecars{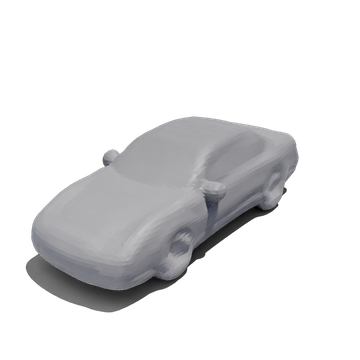}
\includecars{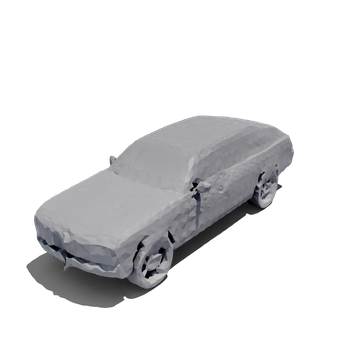}
\includecars{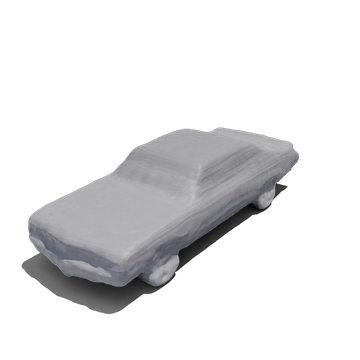}
\includecars{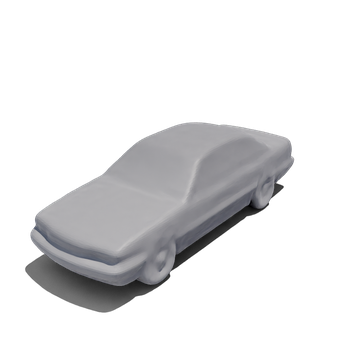}
\includecars{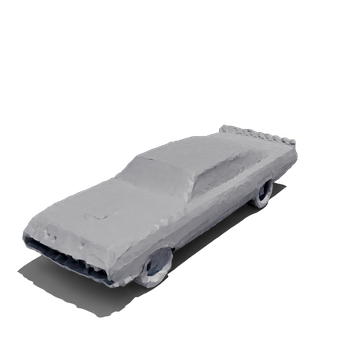}
\includecars{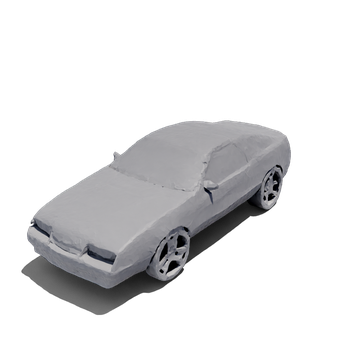}
\includecarszoom{Images/comparision/cars2/sdf_car_details.png}
\includecarszoom{Images/comparision/cars2/5_get3d_2.png}
\includecarszoom{Images/comparision/cars2/5_nfd_0.png}
\includecarszoom{Images/comparision/cars2/5_nwd.png}
\includecarszoom{Images/comparision/cars2/5_md_1.png}
\includecarszoom{Images/comparision/cars2/5_192_0.png}

\includecars{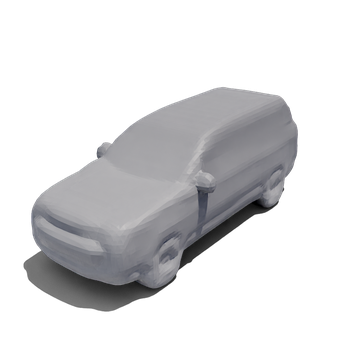}
\includecars{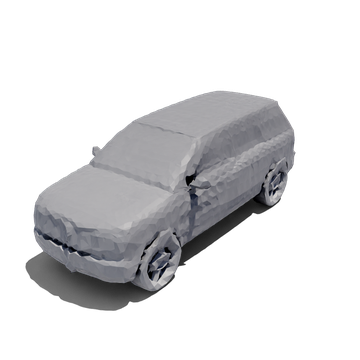}
\includecars{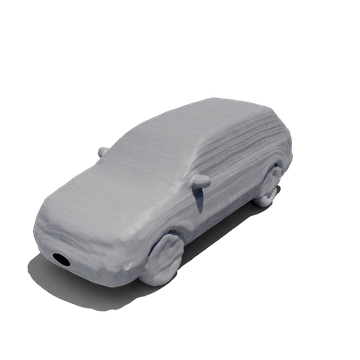}
\includecars{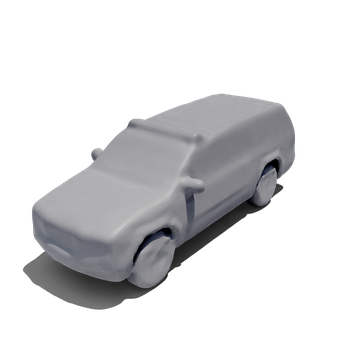}
\includecars{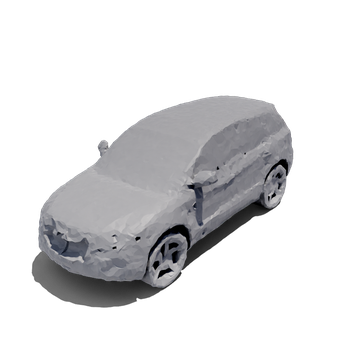}
\includecars{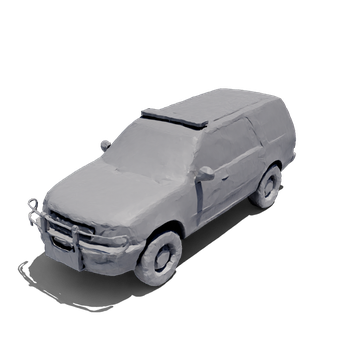}

\includecarssuvzoom{Images/comparision/cars2/sdf_car.png}
\includecarssuvzoom{Images/comparision/cars2/85_get3d_0.png}
\includecarssuvzoom{Images/comparision/cars2/85_nfd_0.png}
\includecarssuvzoom{Images/comparision/cars2/322.png}
\includecarssuvzoom{Images/comparision/cars2/85_md_0.png}
\includecarssuvzoom{Images/comparision/cars2/85_192_0.png}

\begin{minipage}{0.16\linewidth}
\centering
SDF-SG \cite{zheng2022sdf}
\end{minipage}
\begin{minipage}{0.16\linewidth}
\centering
GET3D \cite{gao2022get3d}
\end{minipage}
\begin{minipage}{0.16\linewidth}
\centering
NFD \cite{shue20233d}
\end{minipage}
\begin{minipage}{0.16\linewidth}
\centering
NWD \cite{hui2022neural}
\end{minipage}
\begin{minipage}{0.16\linewidth}
\centering
MD \cite{Liu2023MeshDiffusion}
\end{minipage}
\begin{minipage}{0.16\linewidth}
\centering
\textbf{Ours\textsubscript{192}}
\end{minipage}
\caption{\textbf{Detailed comparison of geometry among the different methods.} Every second row is a zoomed in crop out of the previous row. Samples are gathered based on nearest neighbor to a reference shape from the validation set.}
\label{fig:comall}
\end{figure}

\Cref{fig:comall} compares TetraDiffusion against other recent mesh generators, namely GET3D~\cite{gao2022get3d}, MeshDiffision (MD)~\cite{Liu2023MeshDiffusion}, Neural Field Generation (NFD)~\cite{shue20233d}, Neural Wavelet Diffusion (NWD)~\cite{hui2022neural} and SDF-~StyleGAN (SDF-SG)~\cite{zheng2022sdf}. 
The figure depicts examples from a large pool of generated shapes, matched across methods by finding the nearest neighbors to the same example from the test set. As general trends, we observe that 
\emph{(i)} the GAN-based methods GET3D and SDF-SG produce noisier, less intricate shapes and have a tendency to hallucinate implausible shape details; \emph{(ii)} TetraDiffusion excels in generating more precise and complex shapes, with sharper creases, detailed small structures, and smoother surfaces with fewer defects, compared to oversmoothed surfaces produced by e.g. NFD and NWD. \emph{(iii)} our hi-res version with $R=192$ clearly improves shape quality compared to MD or GET3D with $R=128$, supporting our claim that \emph{resolution limits has so far been a bottleneck}: existing 3D diffusion models run into hardware limits, and efficient use of memory and compute matters. 
\vspace{-1.5em}

\vspace{-1em}
In \cref{fig:comall} this is underlined by direct comparison between TetraDiffusion and other approaches. Our results are the only ones that produce detailed, realistic shapes that are neither oversmoothed nor have apparent geometric flaws.

\paragraph{Texture comparison.}

\begin{figure}[h]
\newcommand{\includecolorplanes}[2][]{%
  \includegraphics[width=0.155\linewidth, trim=0.54cm 1.35cm 1.22cm 2.43cm, clip, #1]{#2}%
}
\newcommand{\includecolorcars}[2][]{%
  \includegraphics[width=0.155\linewidth, trim=0.27cm 1.35cm 1.22cm 1.89cm, clip, #1]{#2}%
}

\begin{minipage}{0.01\linewidth}
\vspace{-0.9cm}
\rotatebox[origin=c]{90}{GET3D \cite{gao2022get3d}}
\end{minipage}
\includecolorplanes{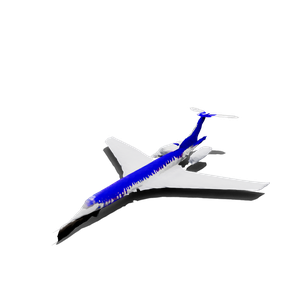}
\includecolorplanes{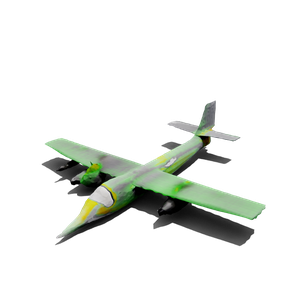}
\includecolorplanes{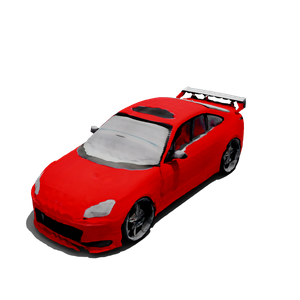}
\includecolorcars{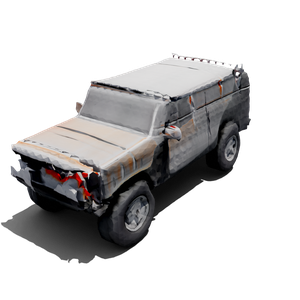}
\includecolorplanes{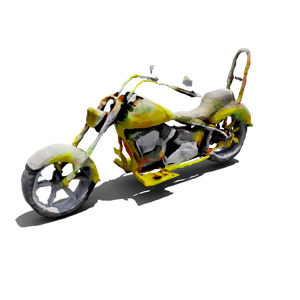}
\includecolorplanes{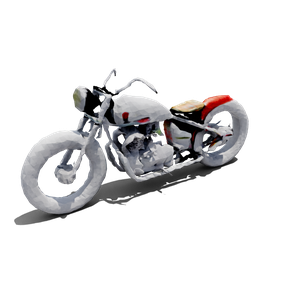}

\begin{minipage}{0.01\linewidth}
\vspace{-0.9cm}
\rotatebox[origin=c]{90}{\textbf{Ours\textsubscript{192}}}
\end{minipage}
\includecolorplanes{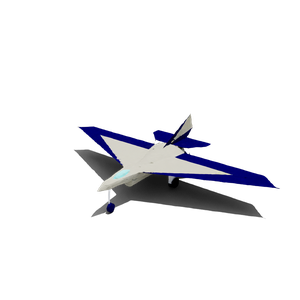}
\includecolorplanes{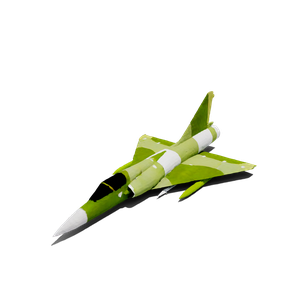}
\includecolorplanes{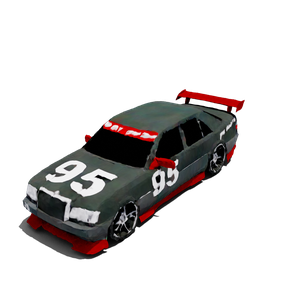}
\includecolorcars{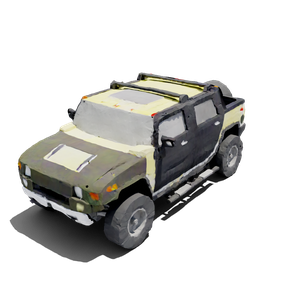}
\includecolorplanes{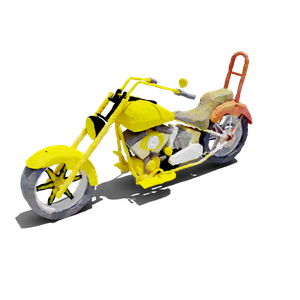}
\includecolorplanes{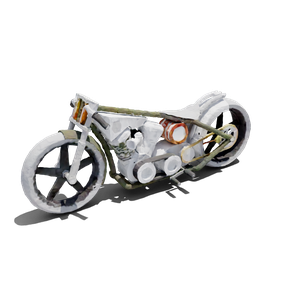}

\caption{\textbf{Comparison of colorized assets between \emph{TetraDiffusion} and GET3D.} Our method is able to produce highly consistent textures with abrupt color changes and diverse patterns. Pairs were selected by finding nearest neighbours in the feature space of a pre-trained ResNet18. Best viewed when zoomed in.}
\label{fig:comcolor}
\vspace{-0.4cm}
\end{figure}

GET3D is the only other model that generates colorized meshes, opting to predict color maps directly, whereas our approach assigns colors to vertices. The latter could be seen as a disadvantage, but TetraDiffusion's ability to produce high-resolution details also results in superior texture, see~\cref{fig:comcolor}. Whereas GET3D tends to create blurry textures (2\textsuperscript{nd}, 4\textsuperscript{th}, 5\textsuperscript{5}th columns), misses small features (front of the car, 4\textsuperscript{th} column), and suffers from misaligned color boundaries (1\textsuperscript{st}, 3\textsuperscript{rd}, 5\textsuperscript{th} columns). There are two reasons for this: On the one hand, a higher vertex density naturally leads to crisper color discontinuities such as numbers on a car body in \cref{fig:comcolor}, especially when vertex placement is trained to jointly reproduce geometry and color. Second, color maps are of limited use unless they are markedly sharper than the geometry \emph{and} well-aligned with shape details.

\begin{figure*}[b!]
\centering
\vspace{-0.1cm}
\includegraphics[width=0.1\linewidth]{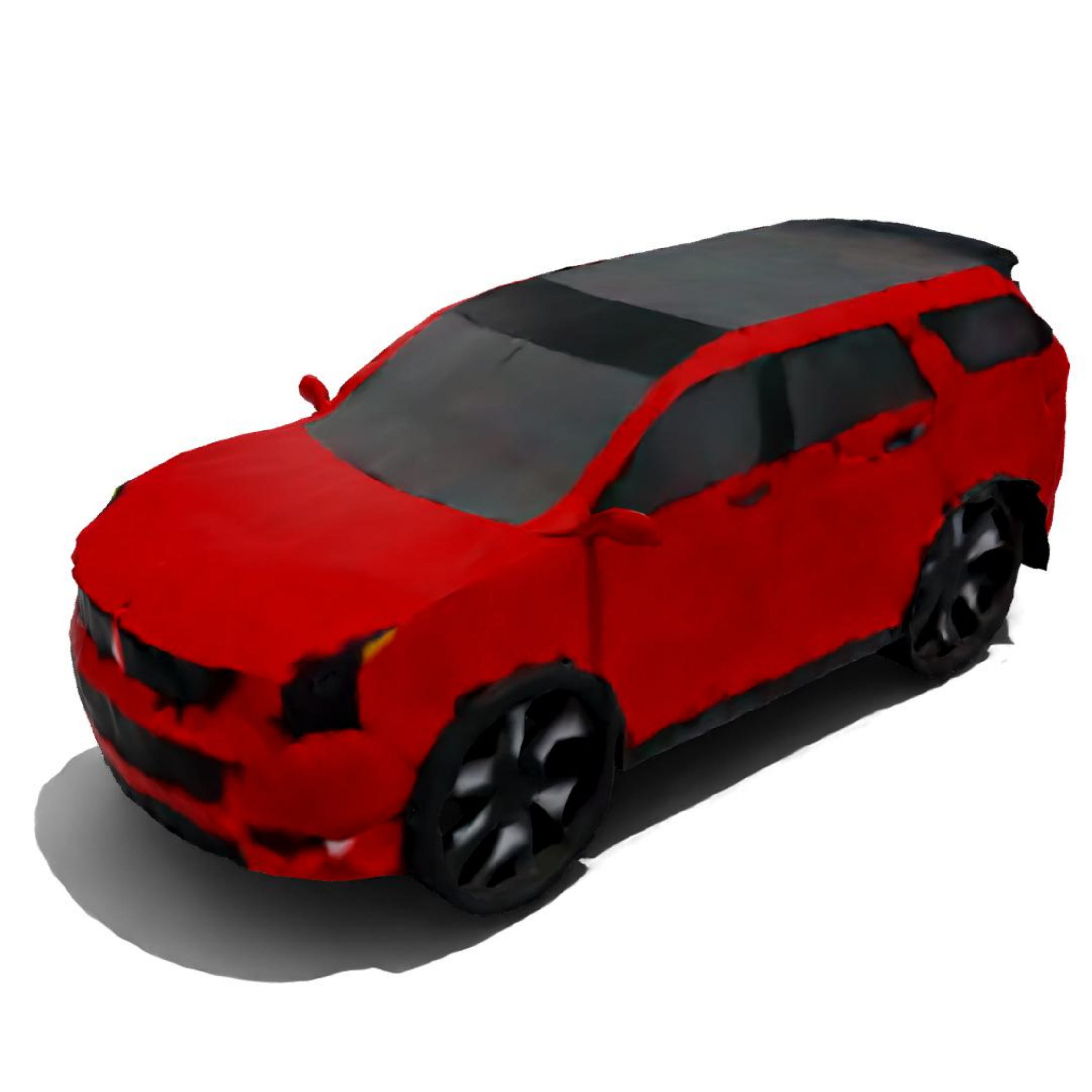}\includegraphics[width=0.1\linewidth]{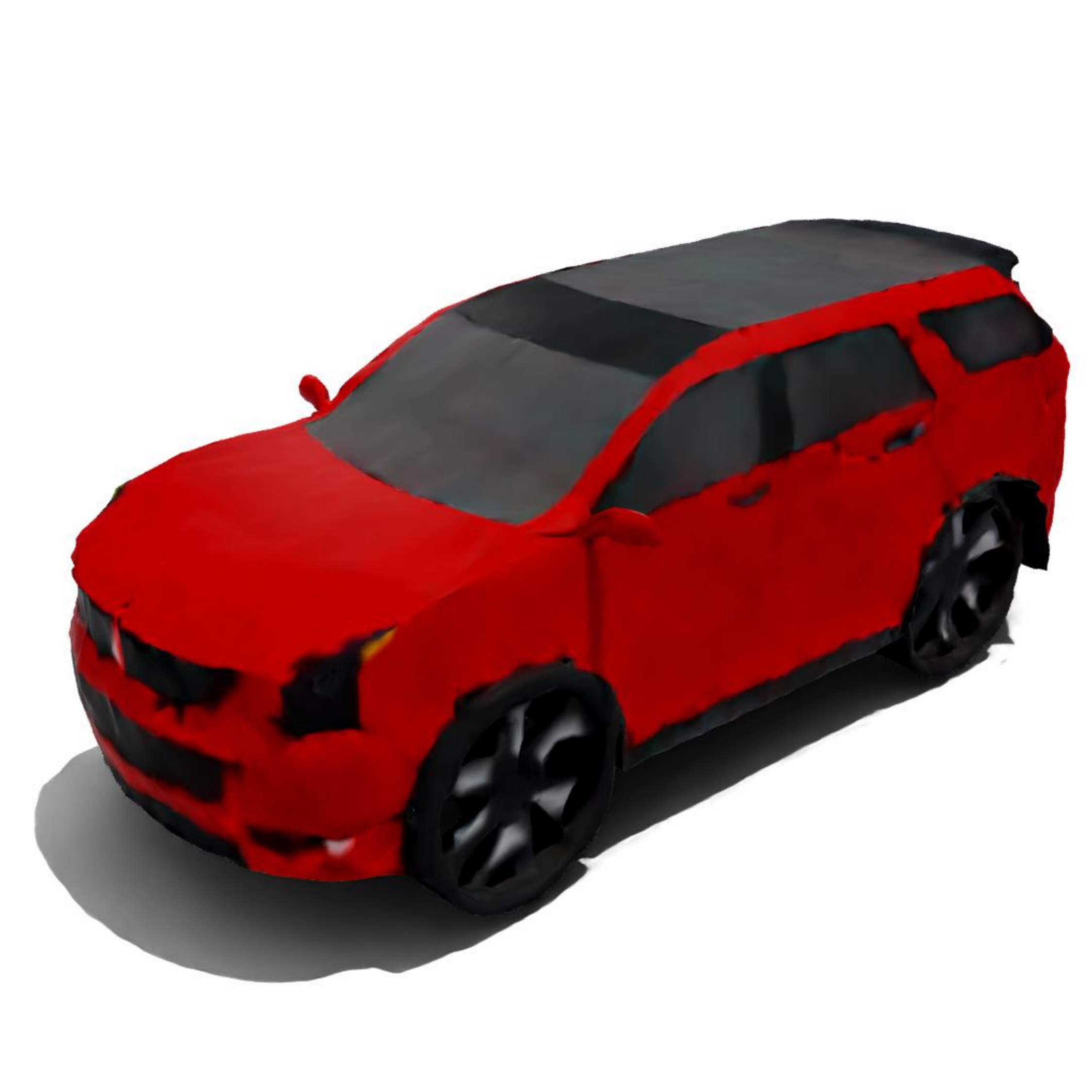}\includegraphics[width=0.1\linewidth]{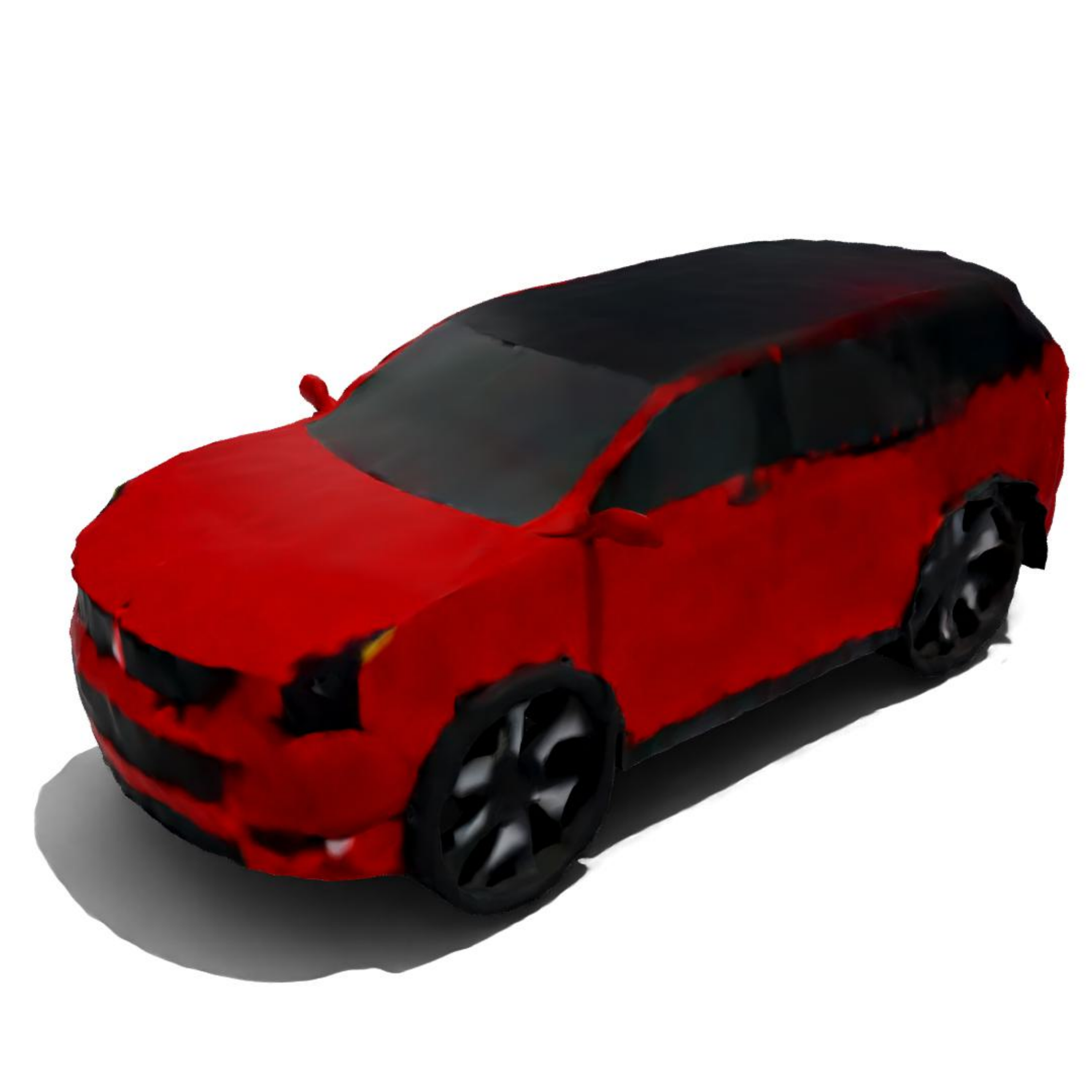}\includegraphics[width=0.1\linewidth]{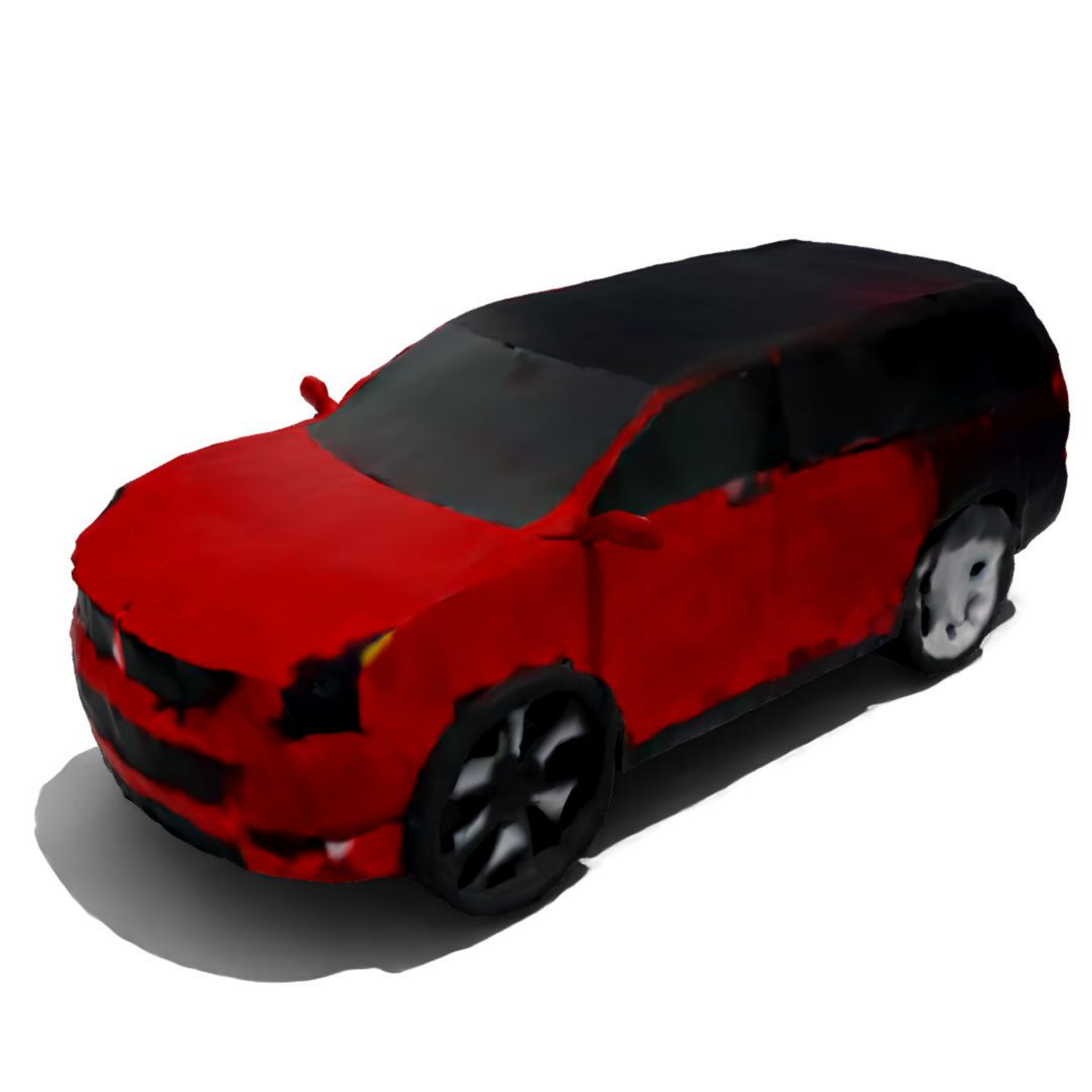}\includegraphics[width=0.1\linewidth]{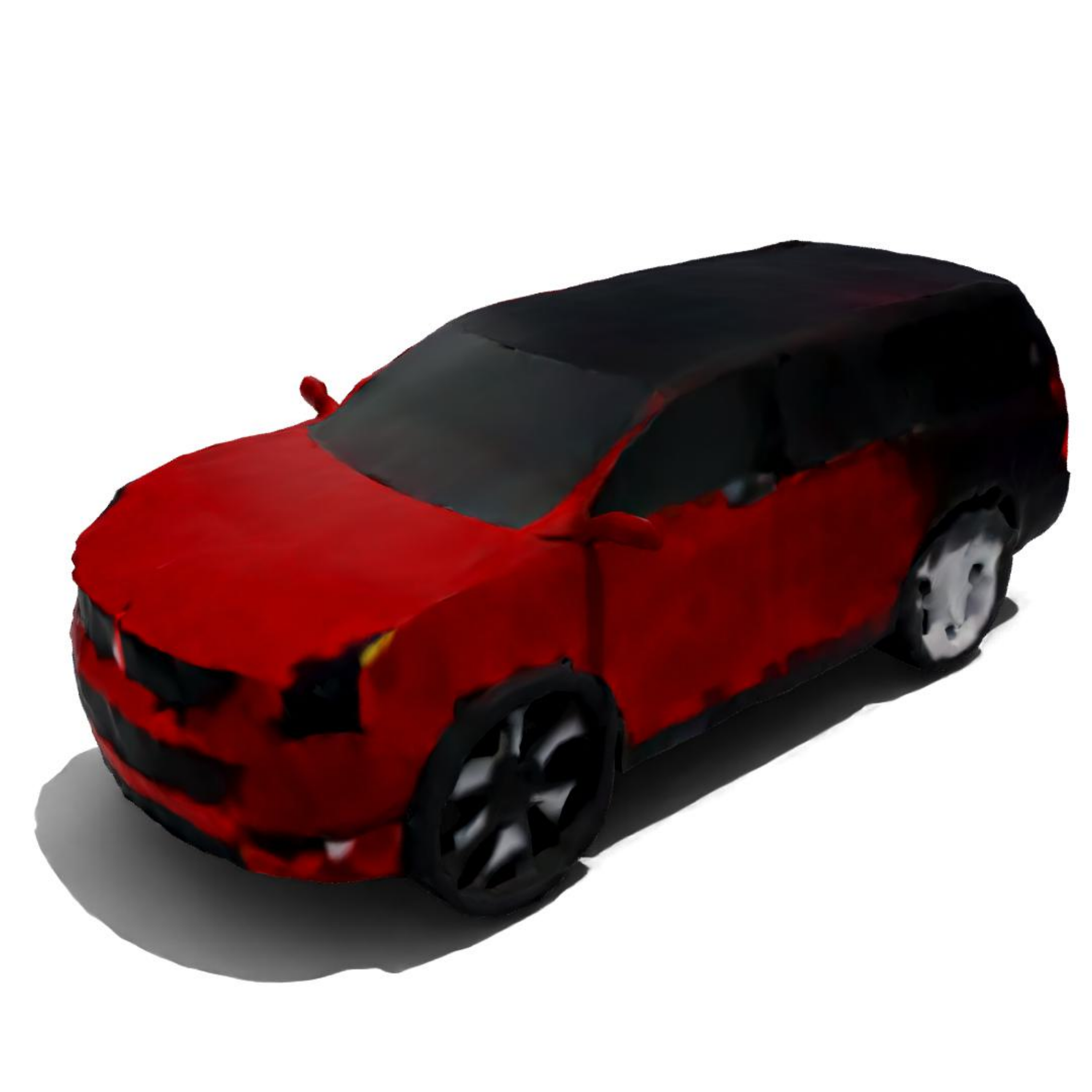}\includegraphics[width=0.1\linewidth]{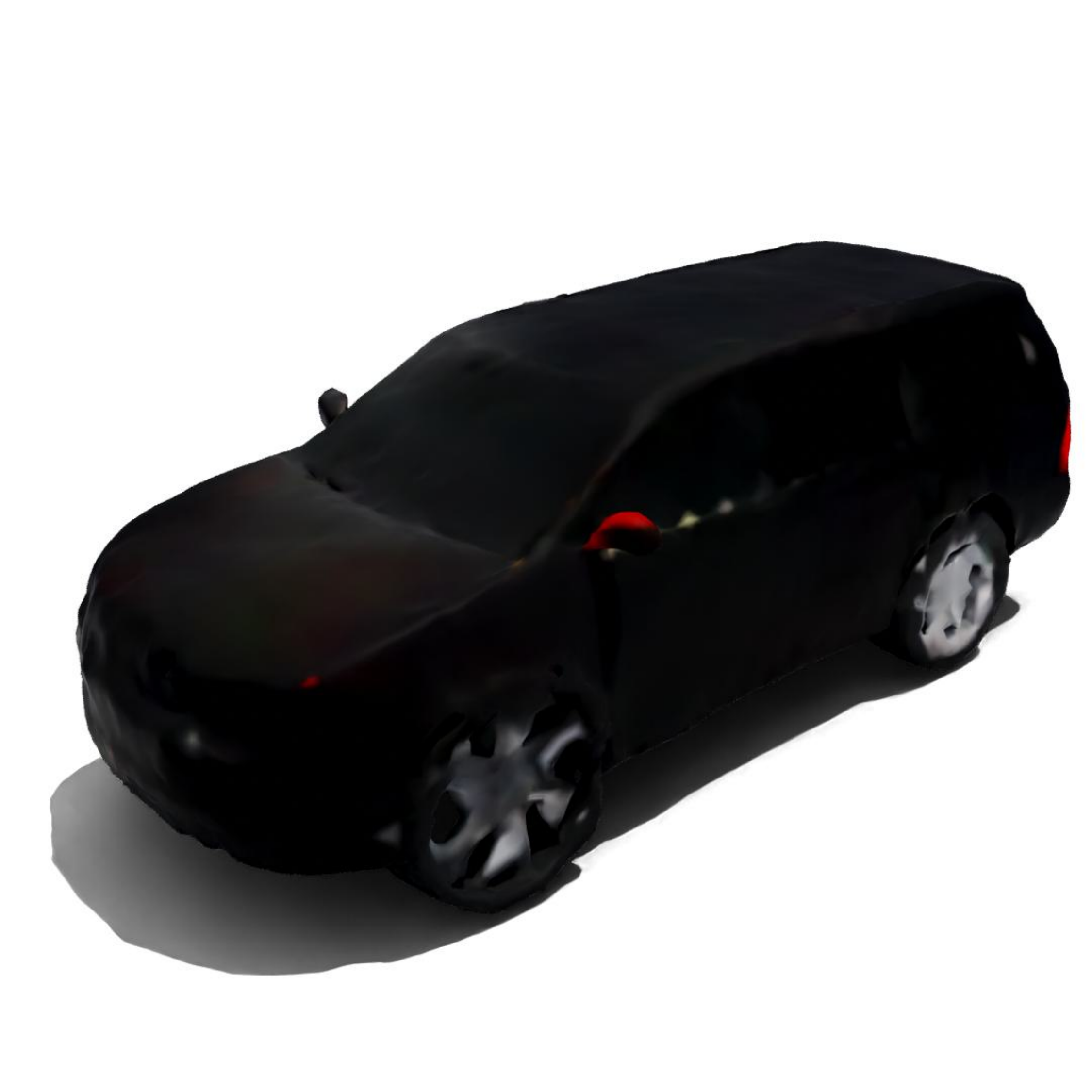}\includegraphics[width=0.1\linewidth]{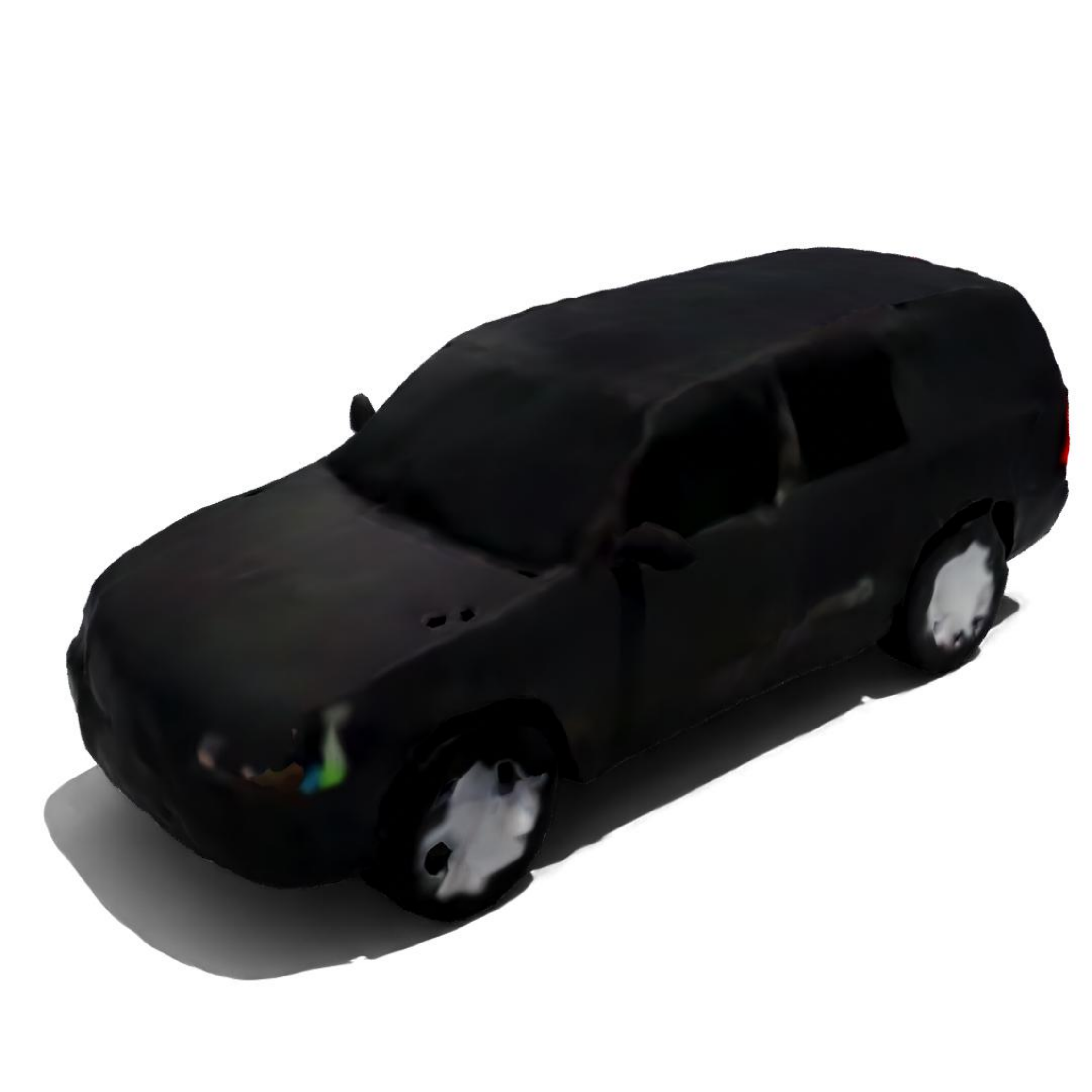}\includegraphics[width=0.1\linewidth]{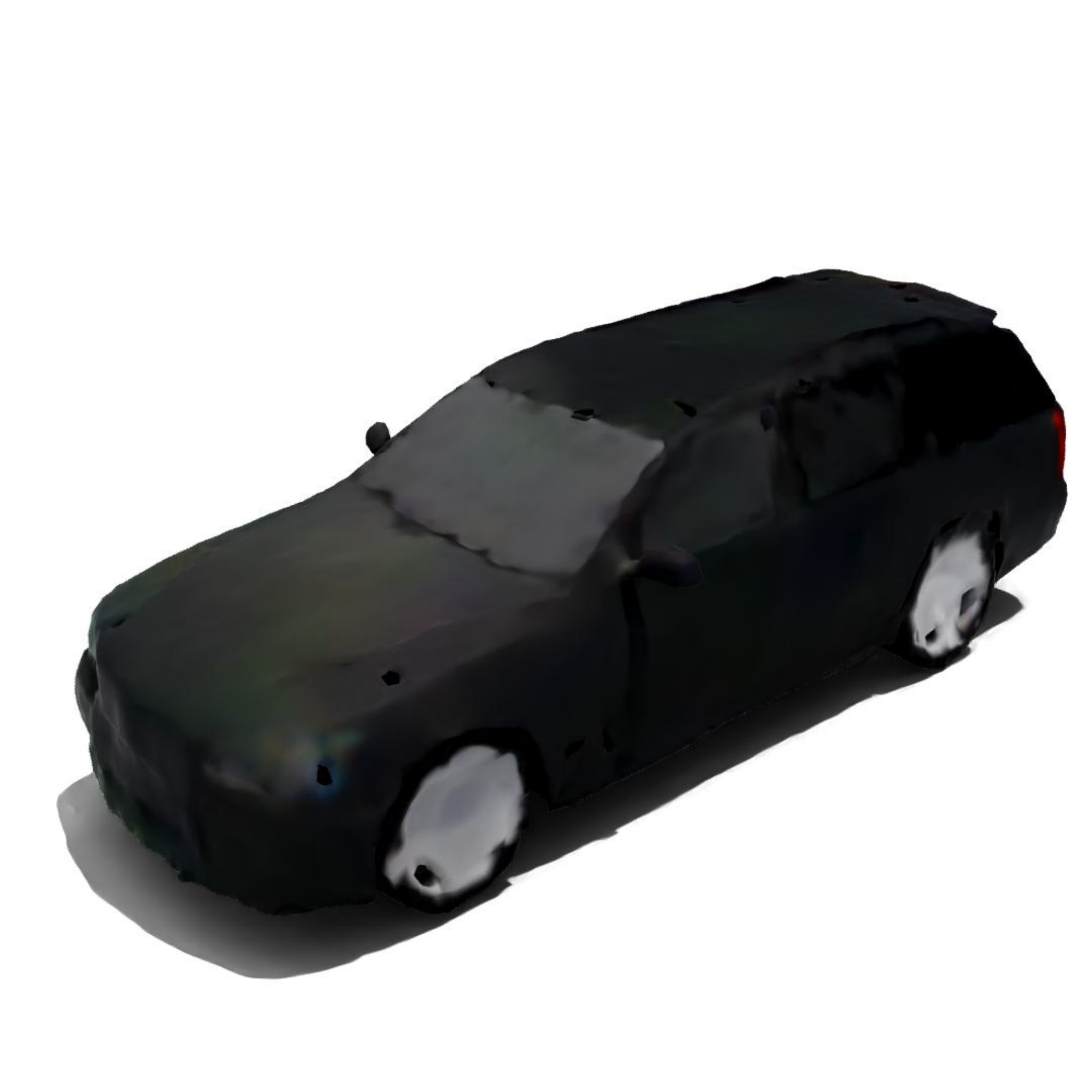}\includegraphics[width=0.1\linewidth]{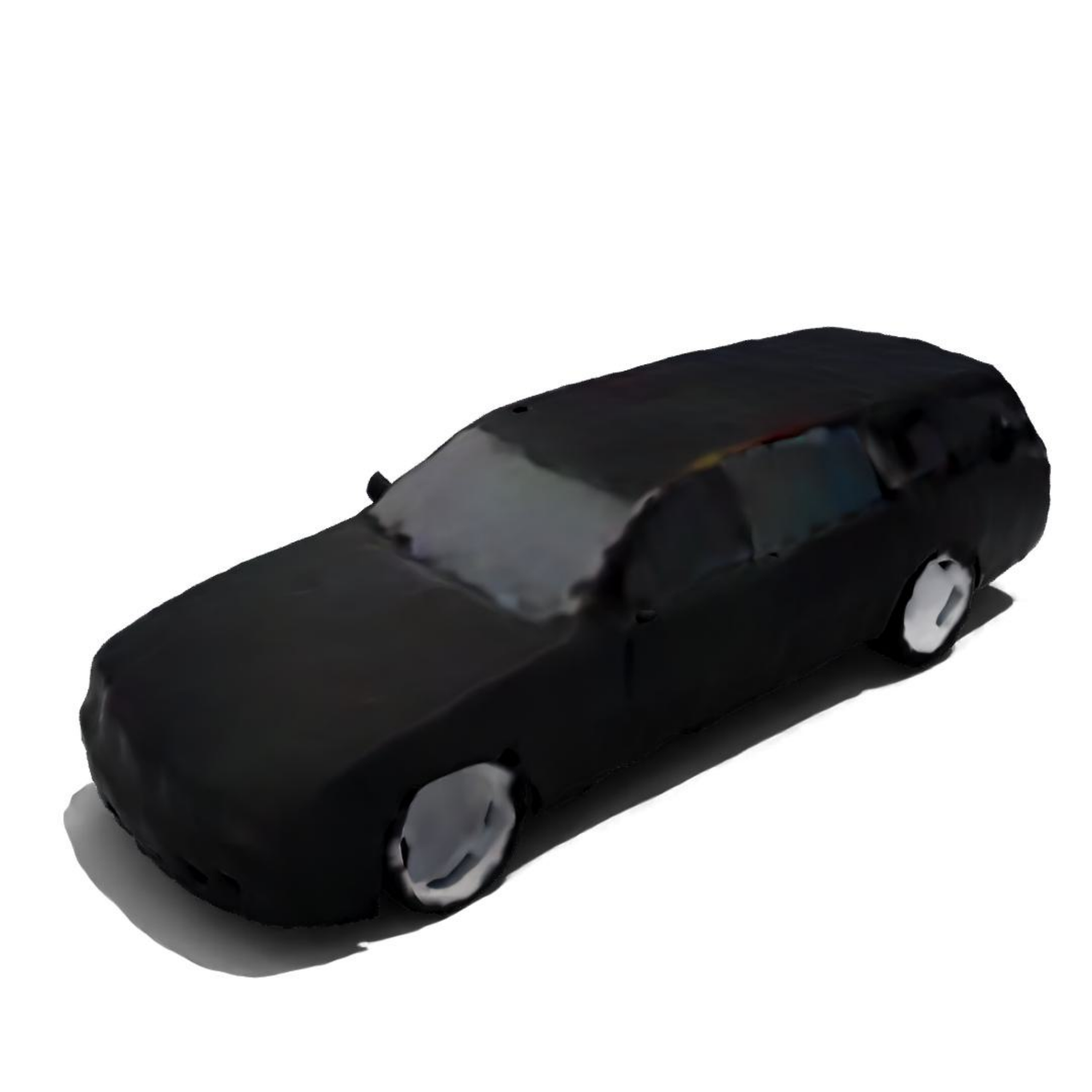}\includegraphics[width=0.1\linewidth]{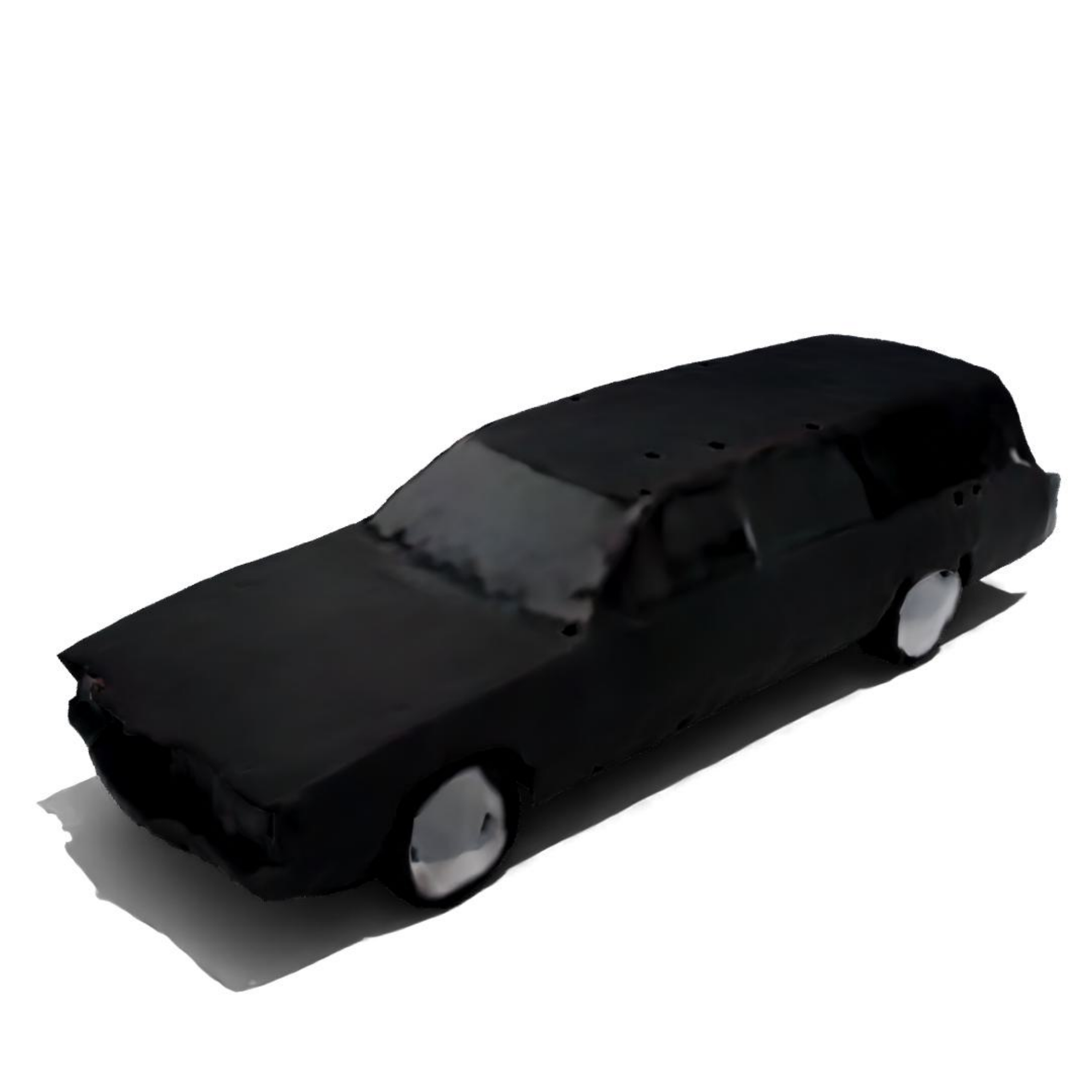}

\vspace{-0.15cm}
\includegraphics[width=0.1\linewidth]{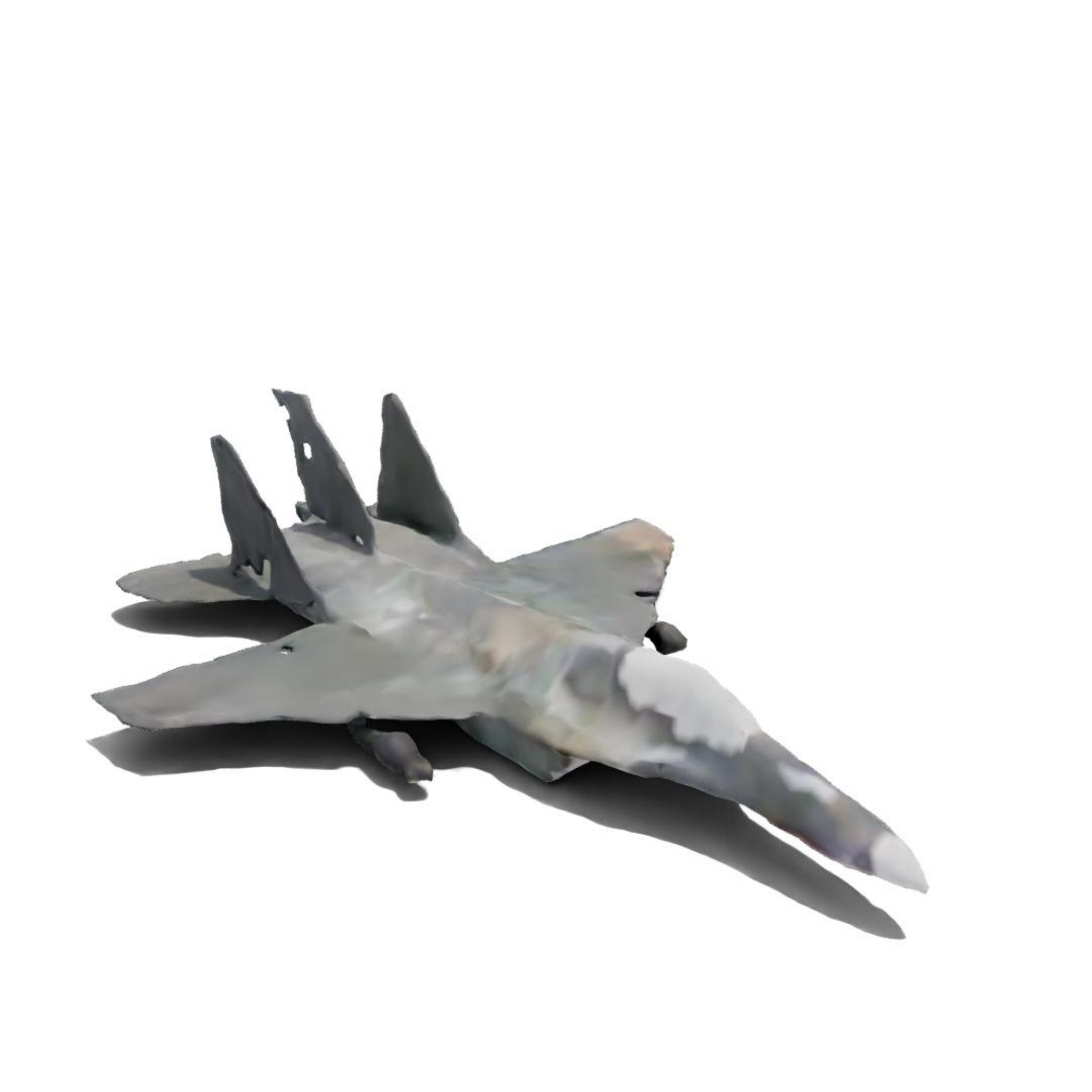}\includegraphics[width=0.1\linewidth]{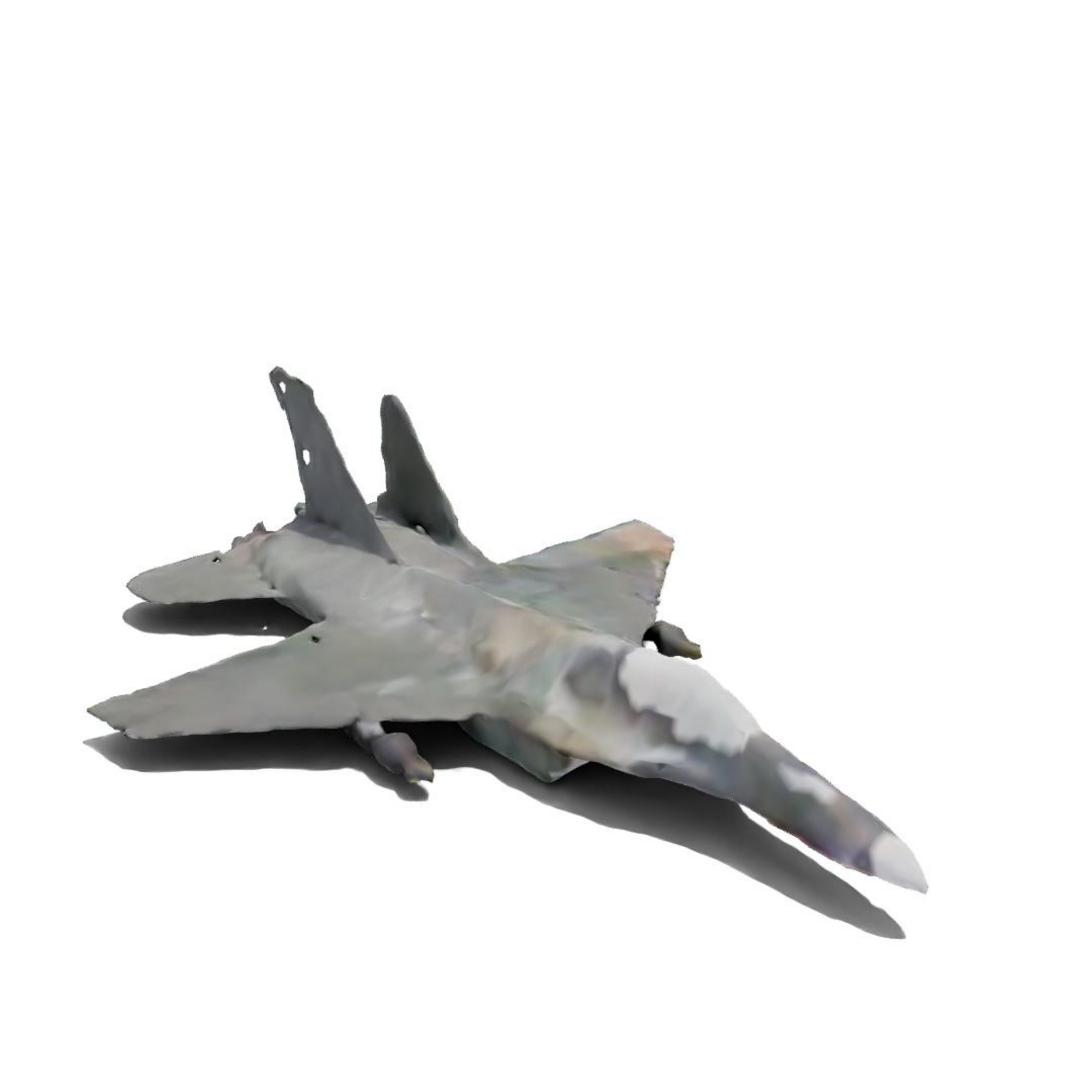}\includegraphics[width=0.1\linewidth]{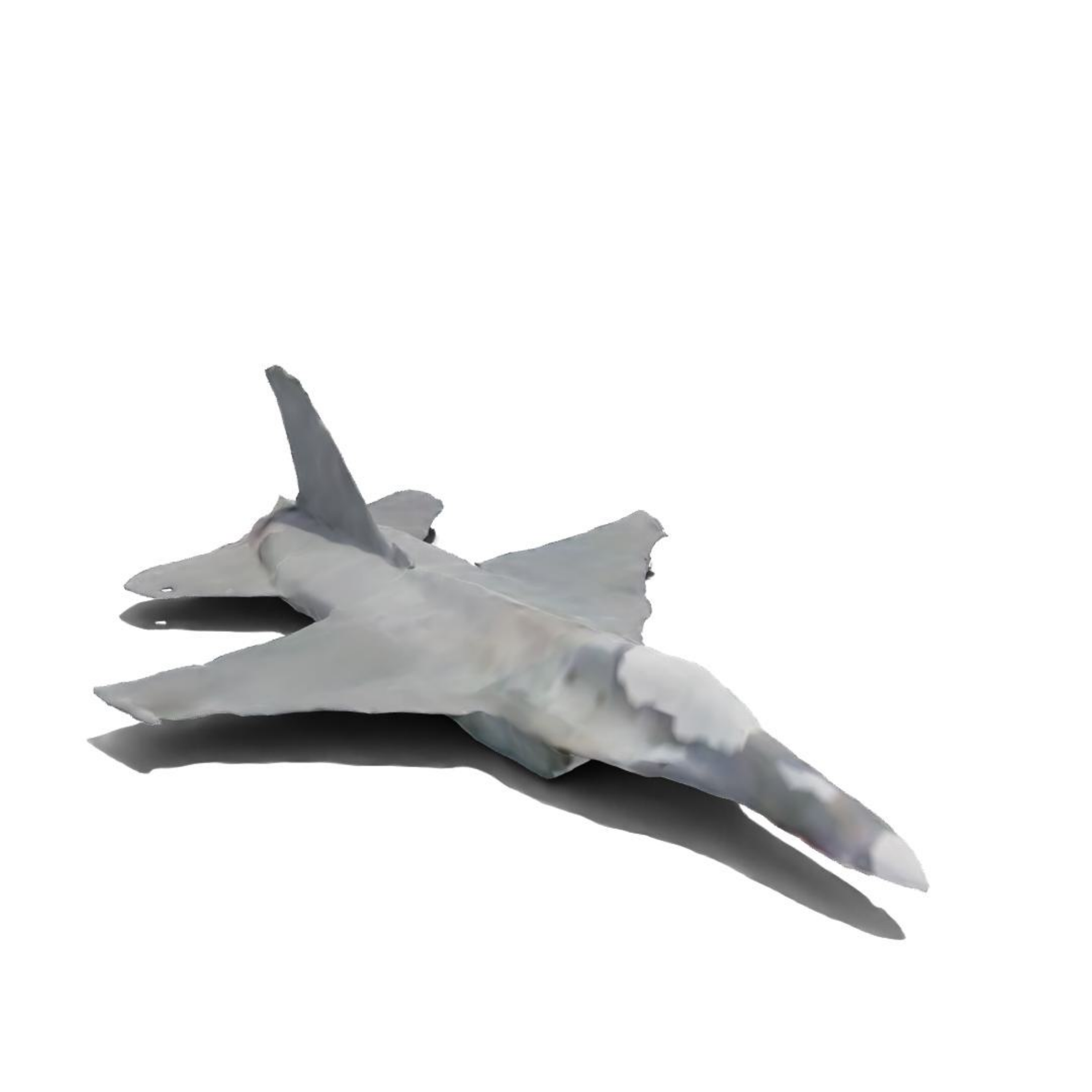}\includegraphics[width=0.1\linewidth]{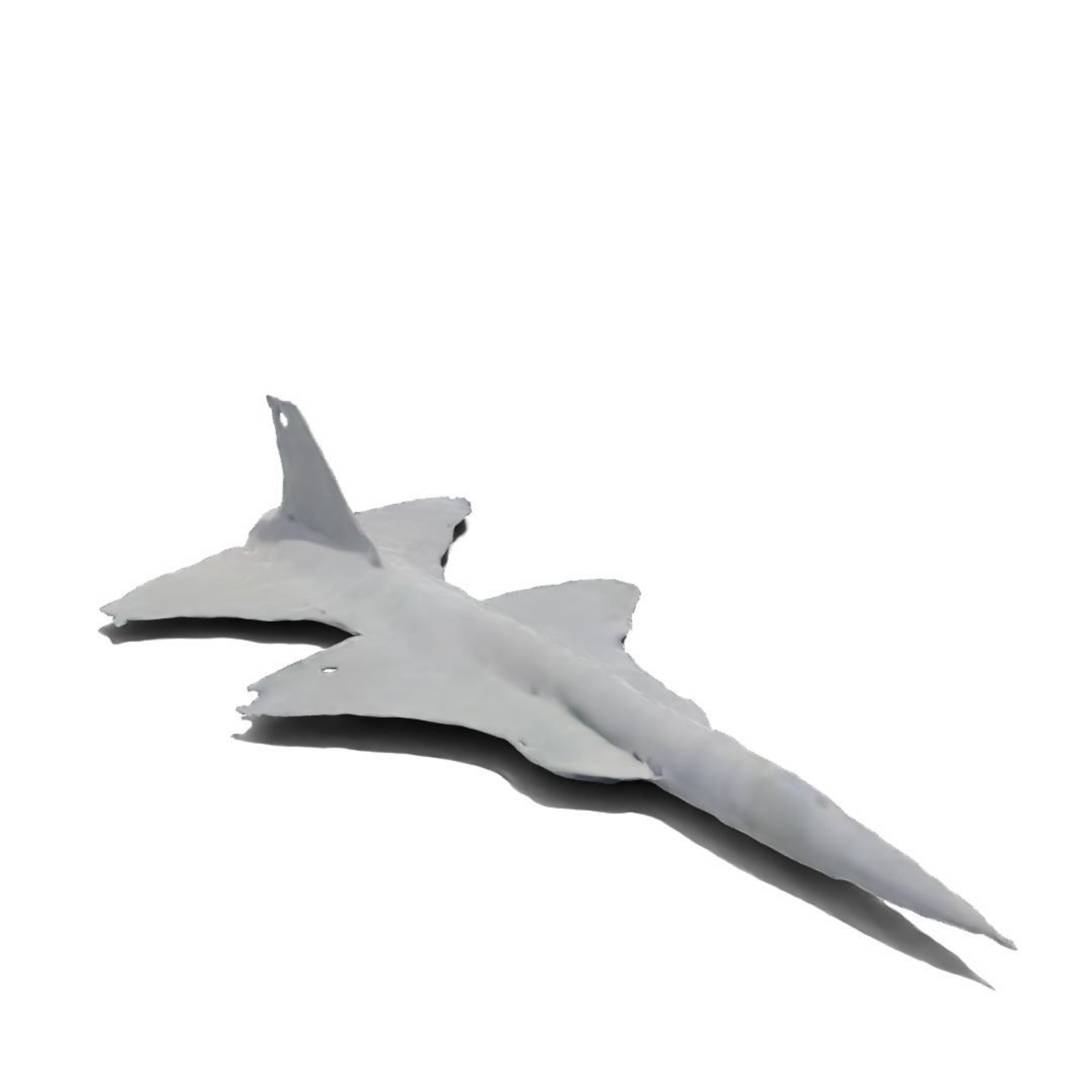}\includegraphics[width=0.1\linewidth]{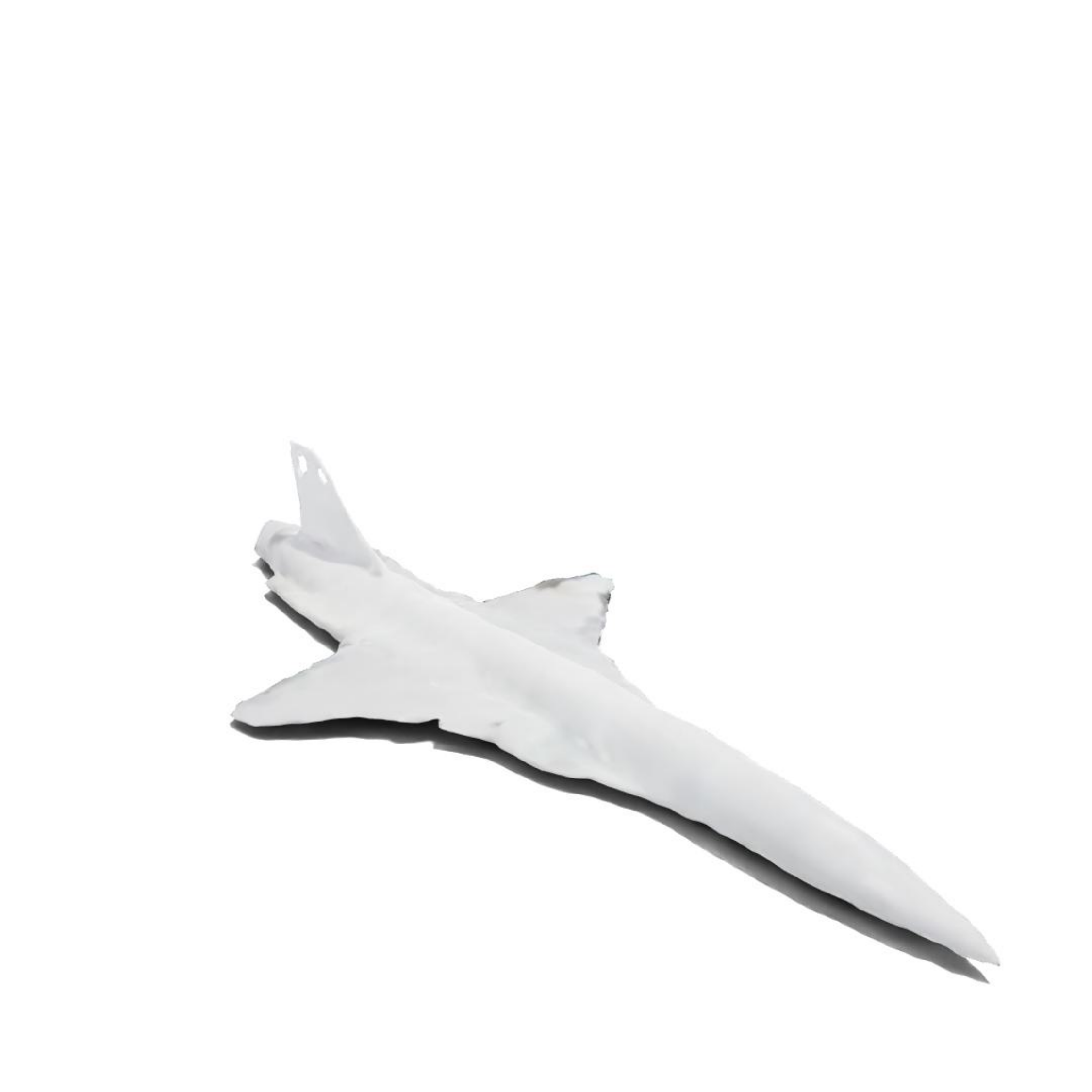}\includegraphics[width=0.1\linewidth]{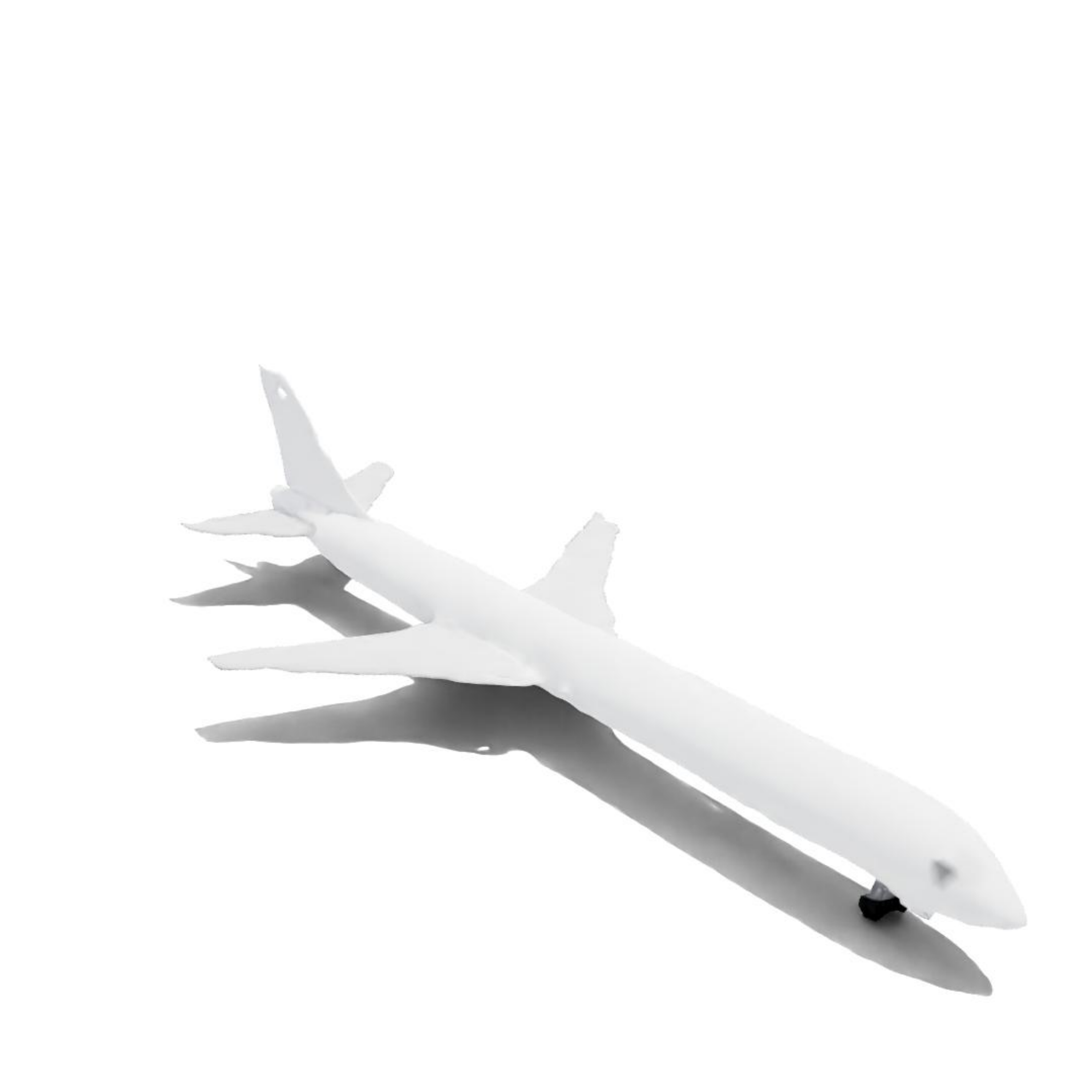}\includegraphics[width=0.1\linewidth]{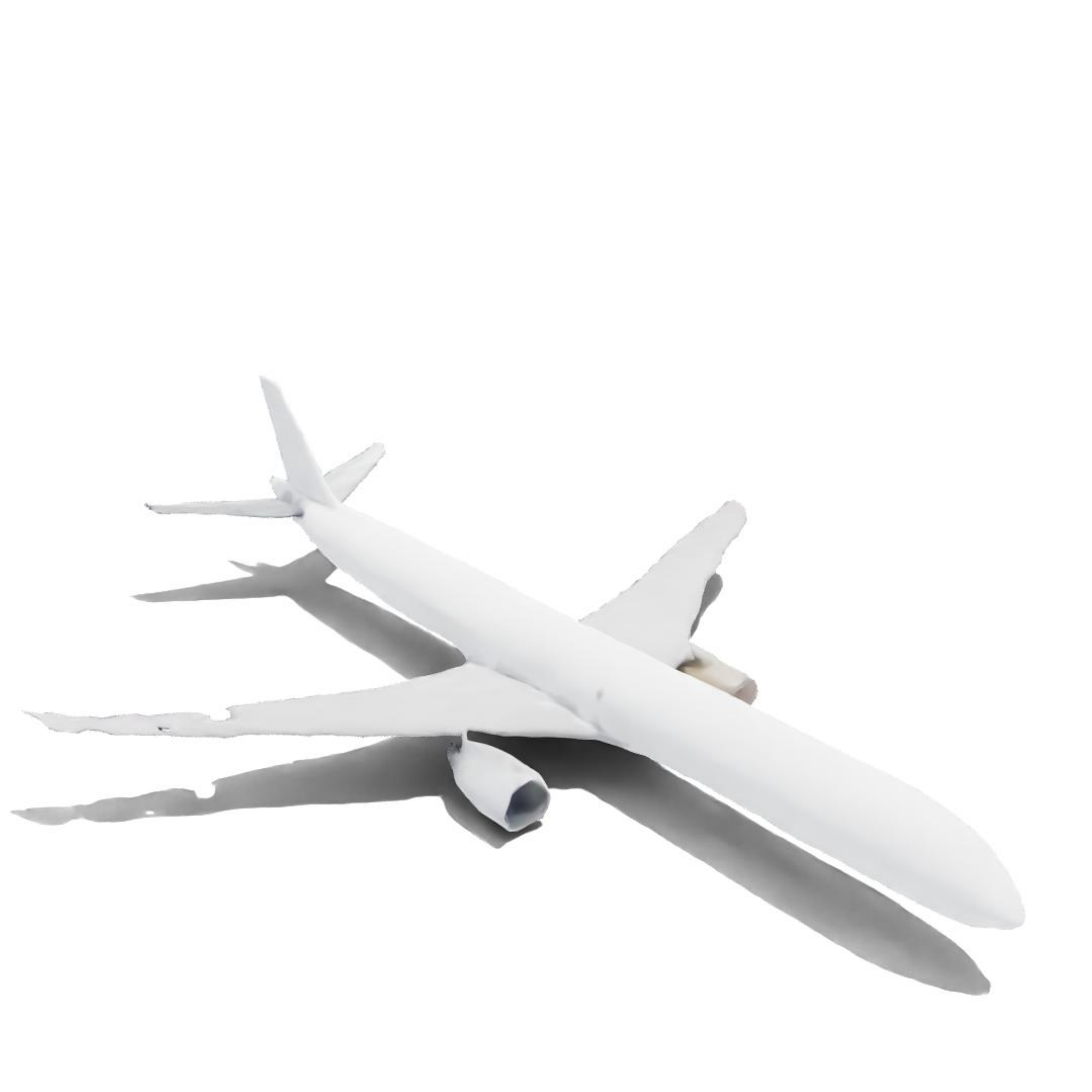}\includegraphics[width=0.1\linewidth]{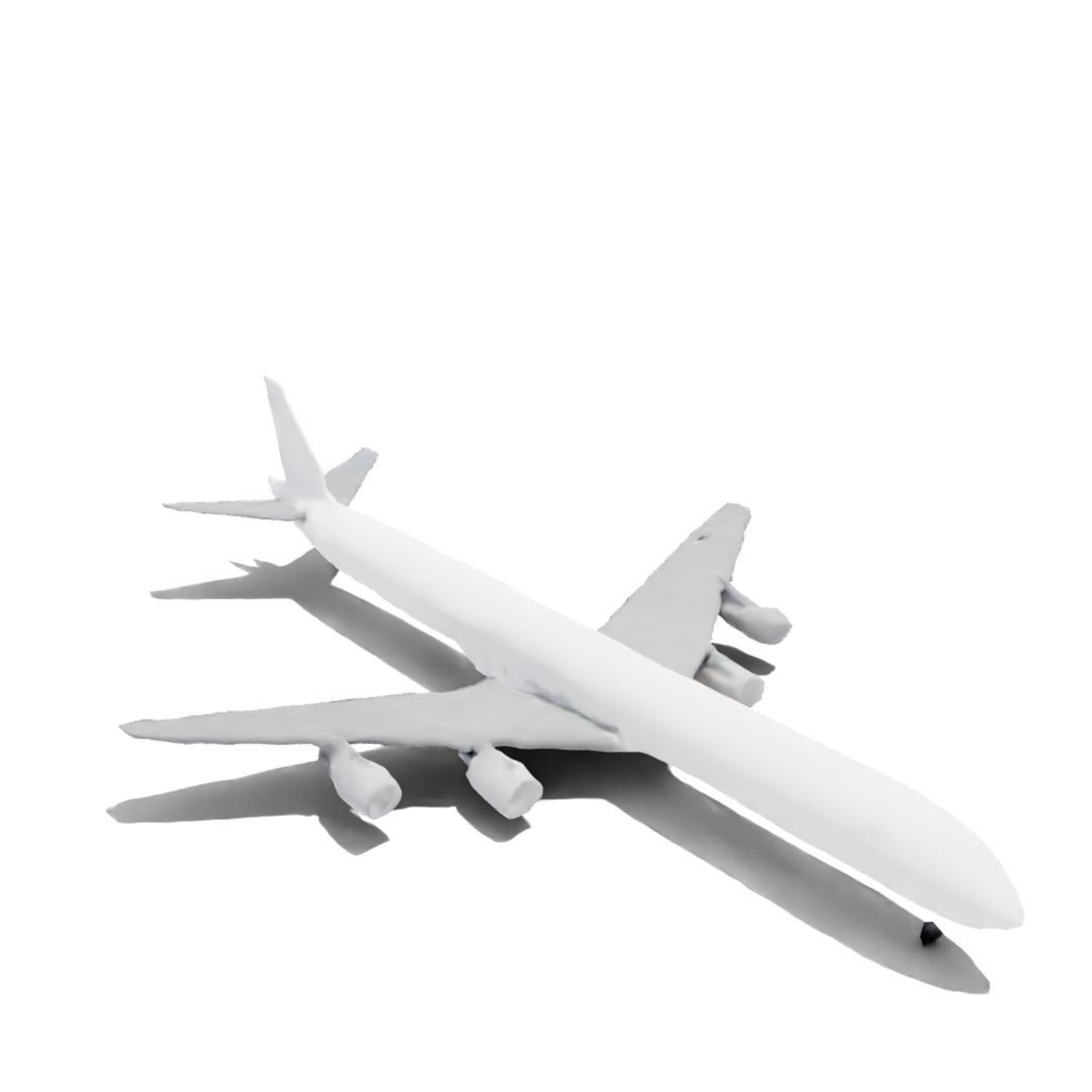}\includegraphics[width=0.1\linewidth]{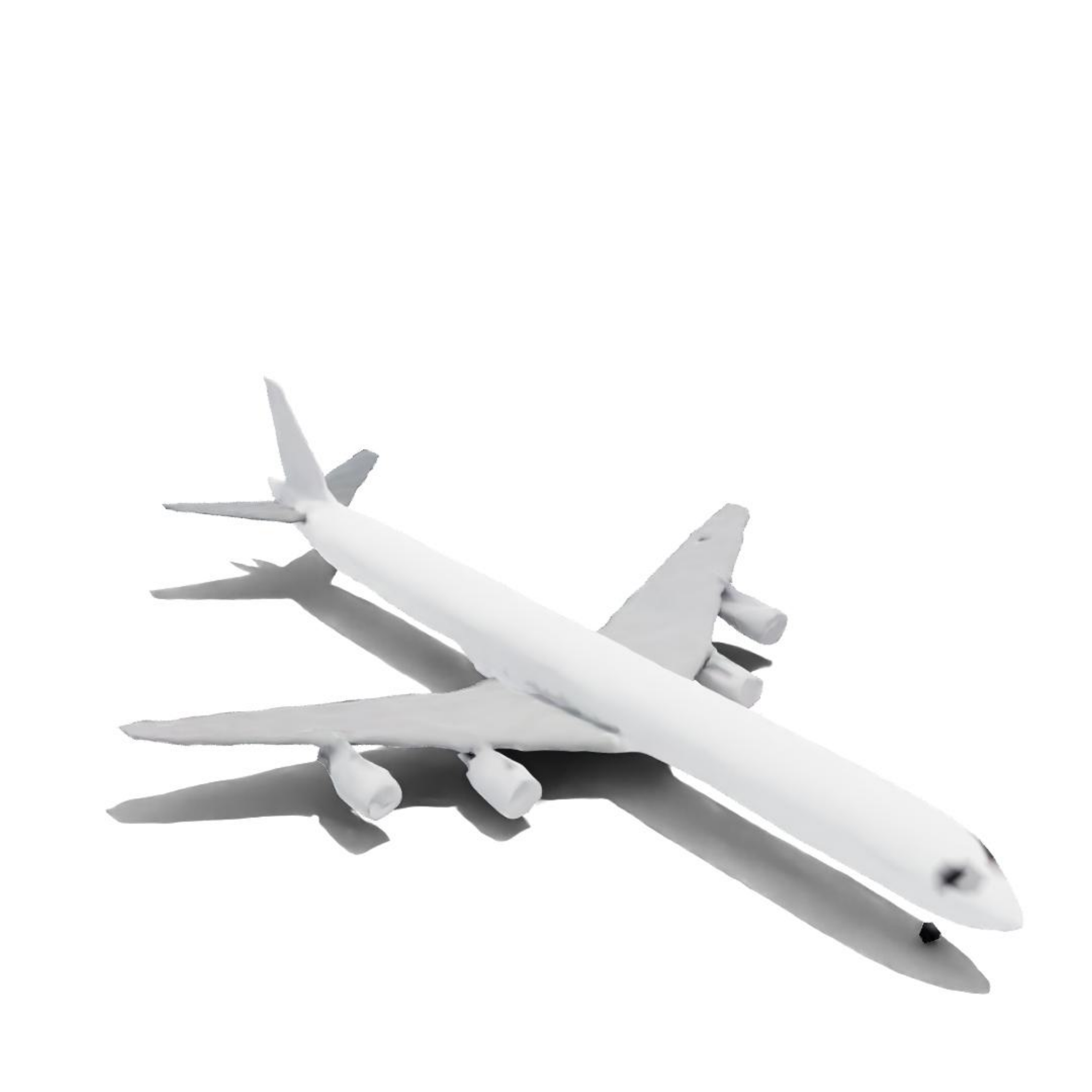}\includegraphics[width=0.1\linewidth]{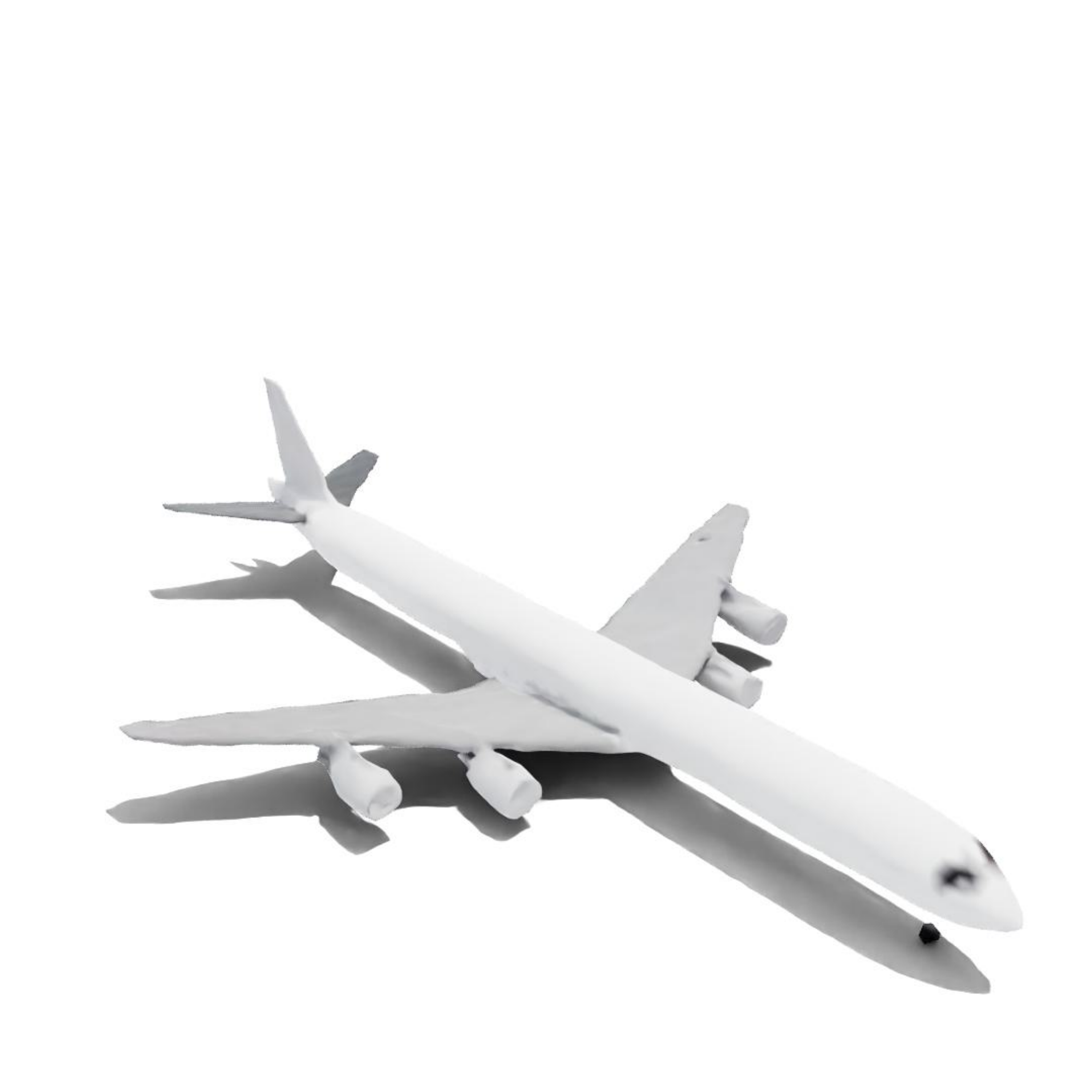}

\includegraphics[width=0.1\linewidth]{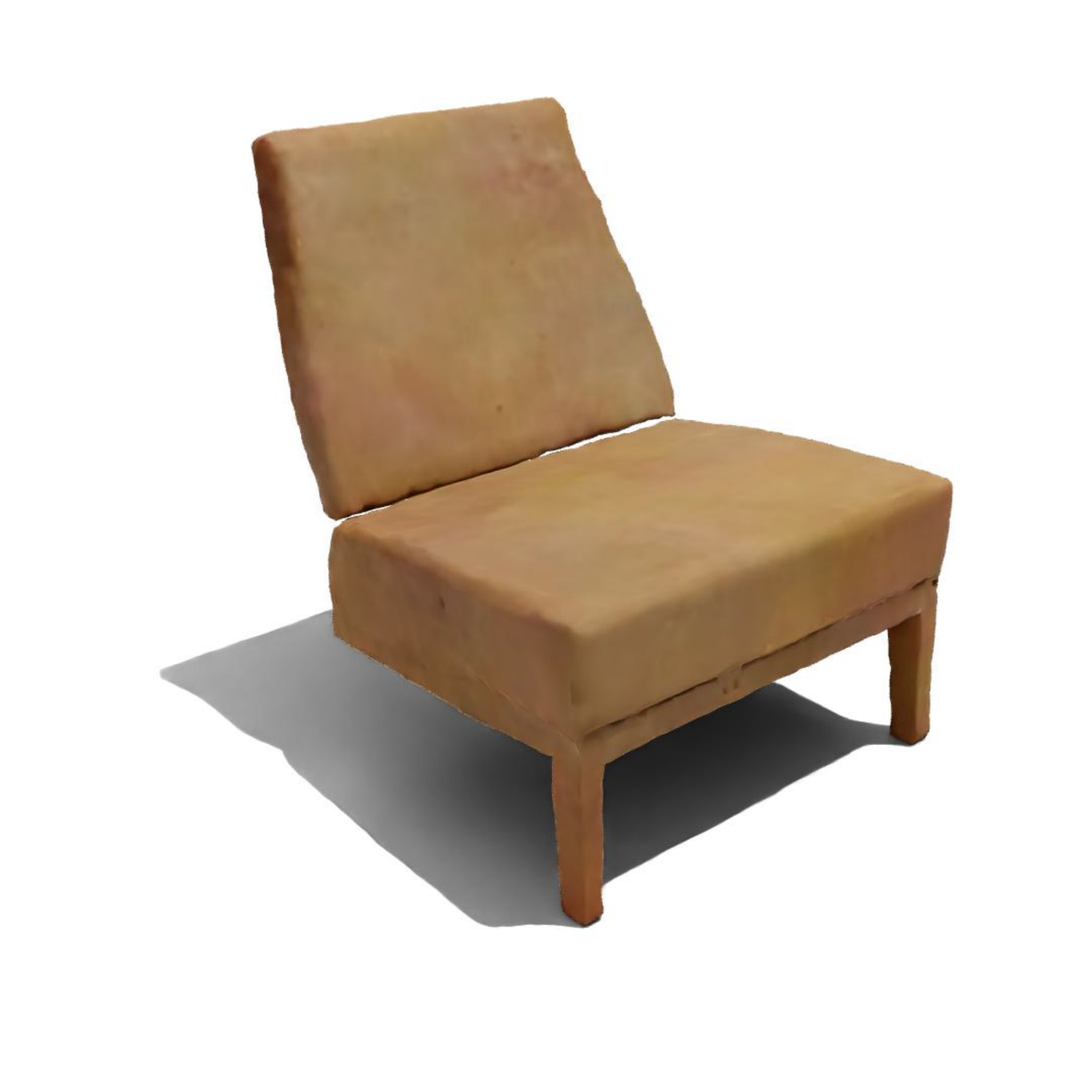}\includegraphics[width=0.1\linewidth]{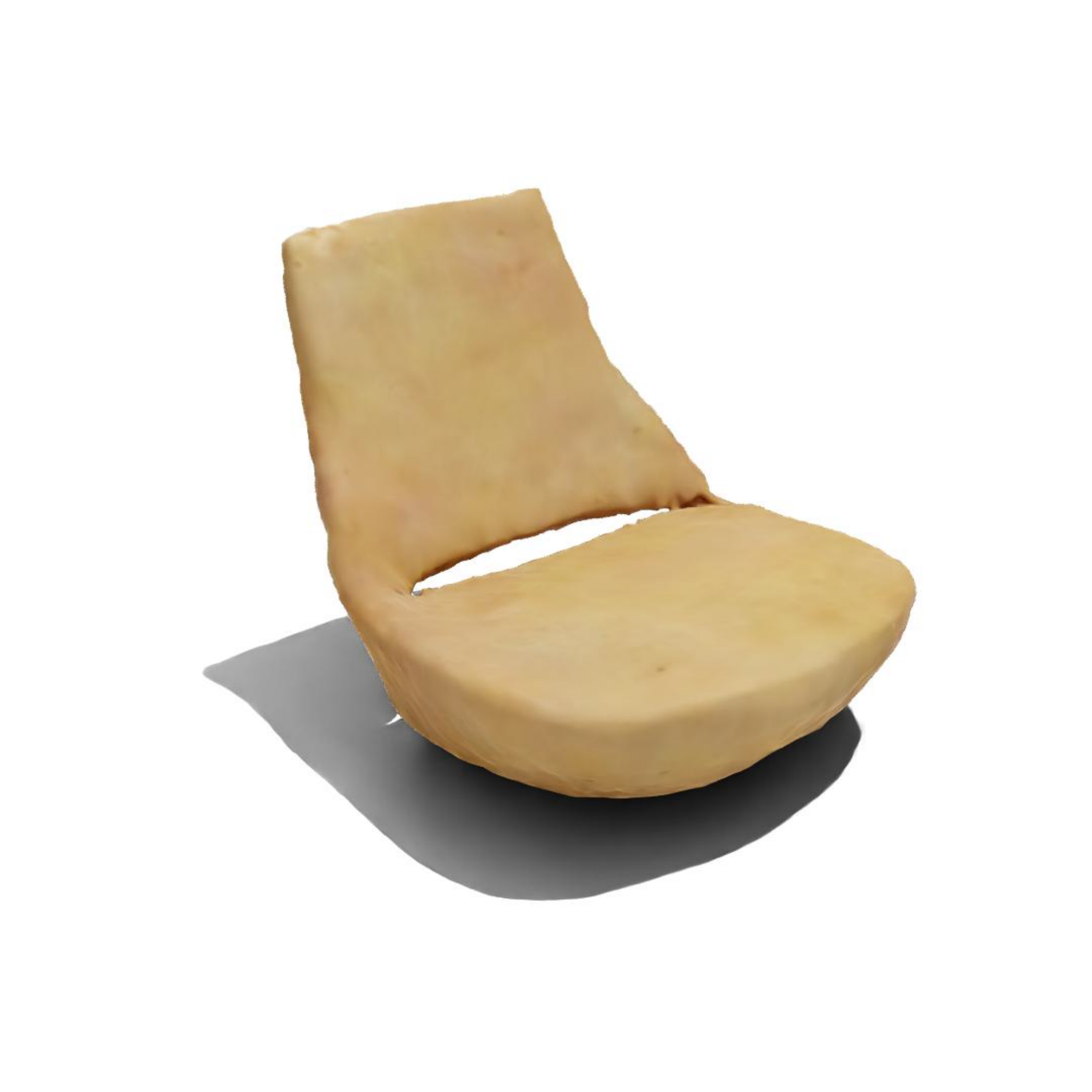}\includegraphics[width=0.1\linewidth]{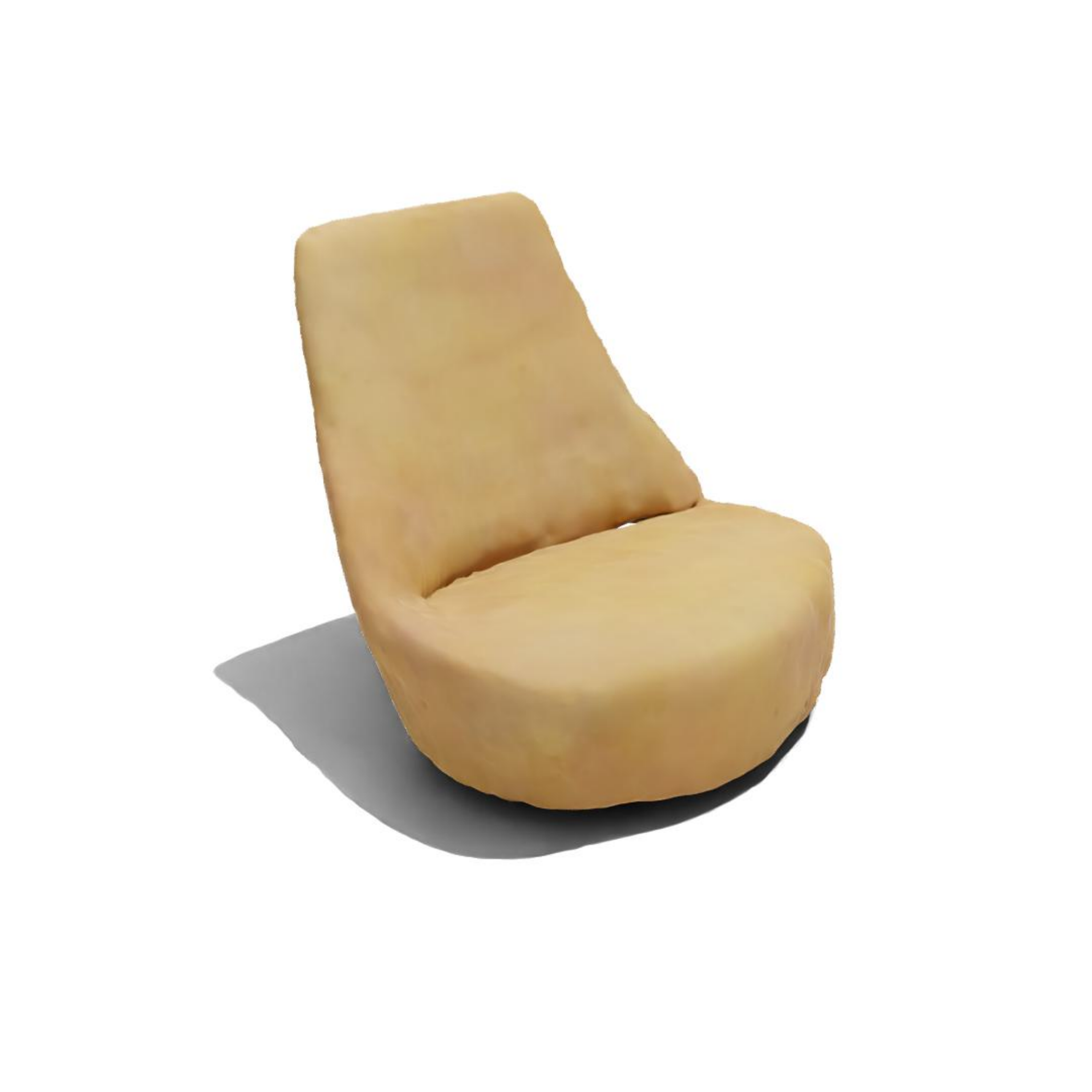}\includegraphics[width=0.1\linewidth]{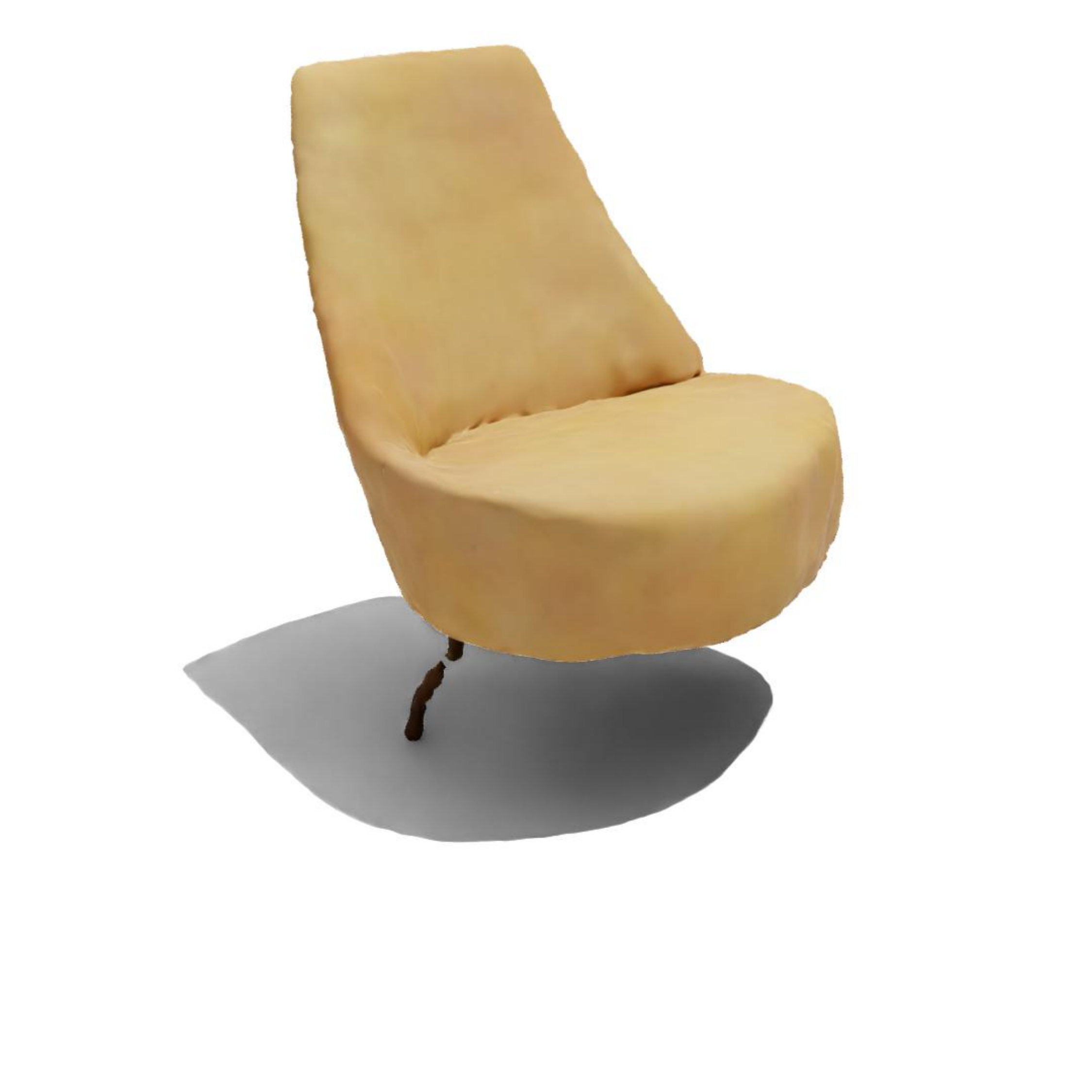}\includegraphics[width=0.1\linewidth]{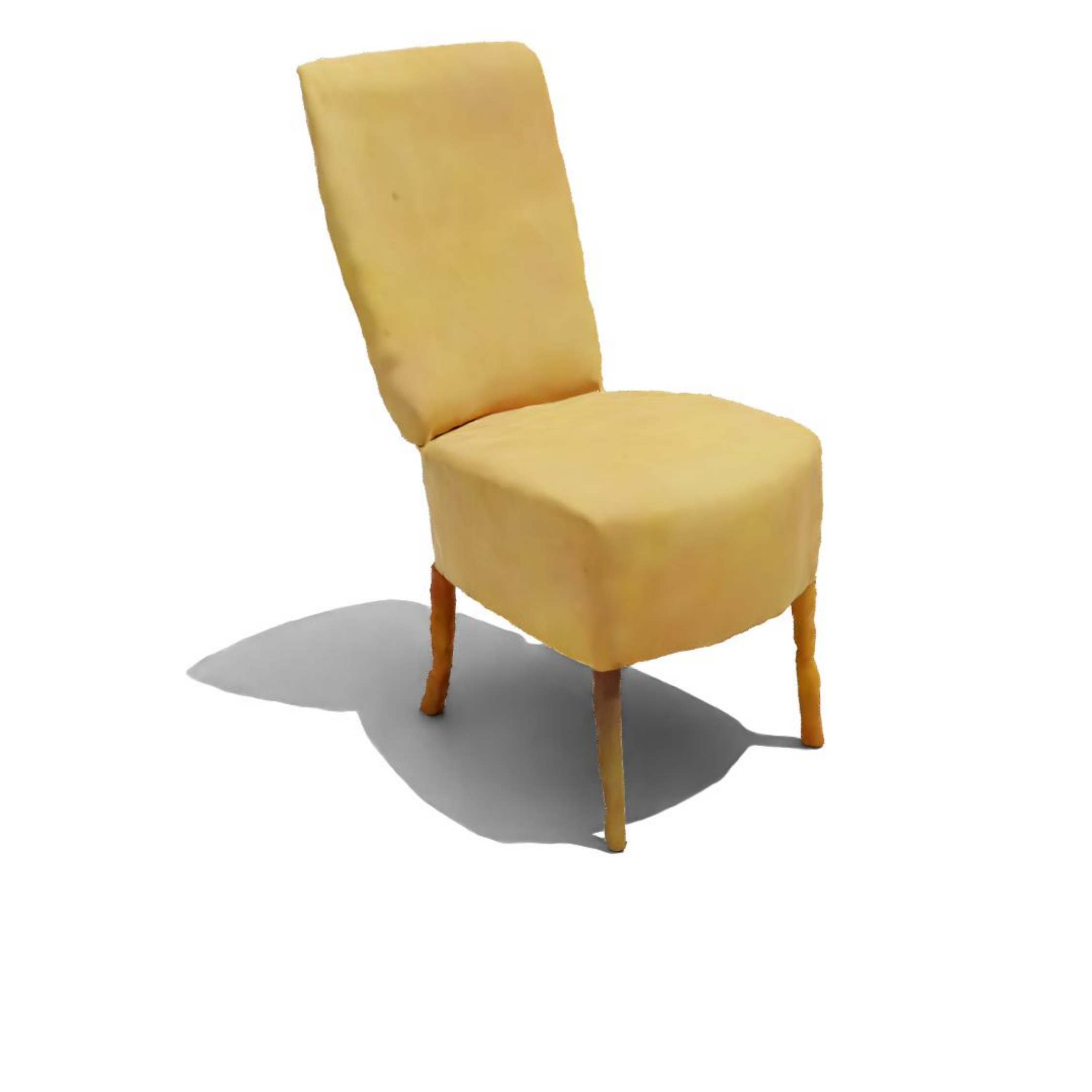}\includegraphics[width=0.1\linewidth]{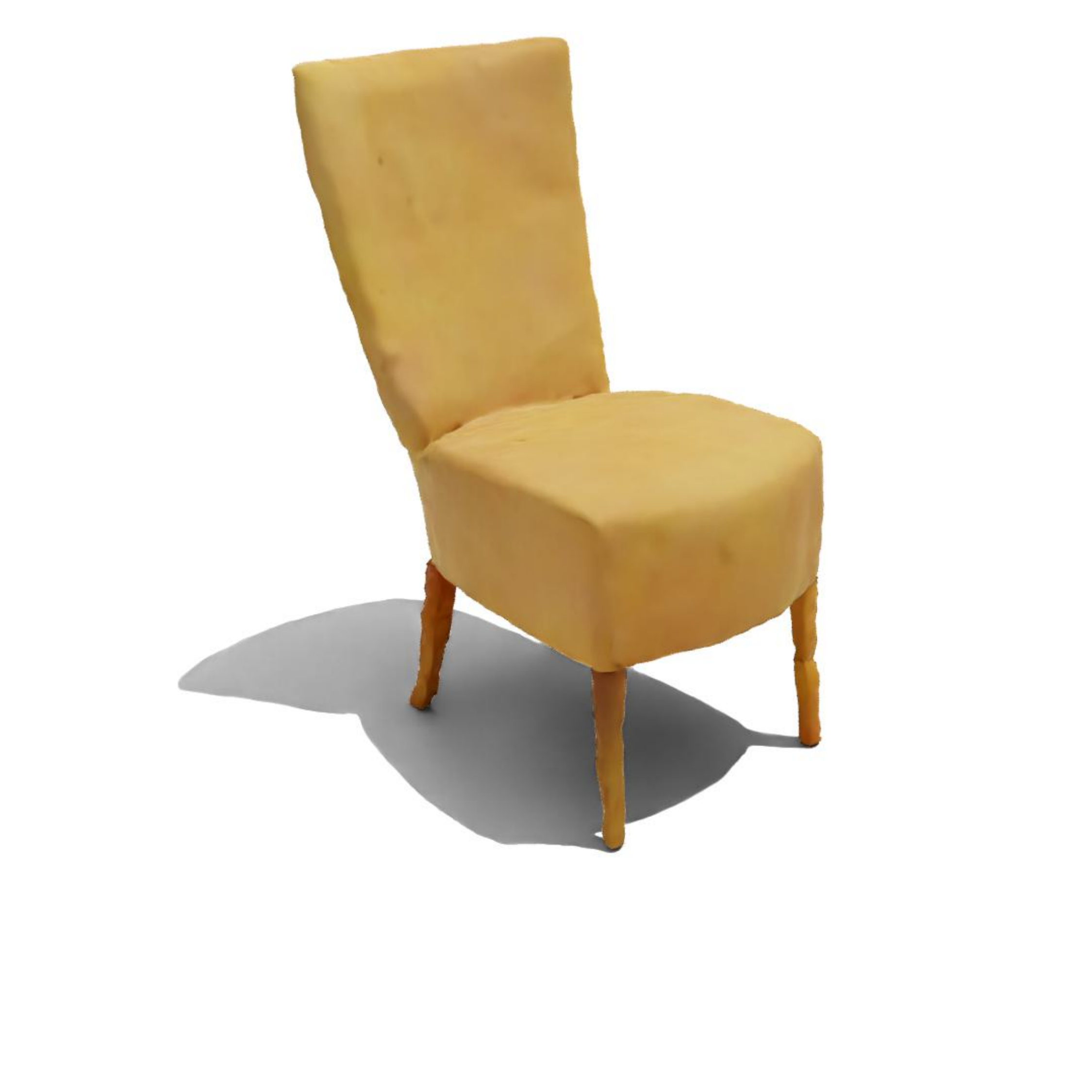}\includegraphics[width=0.1\linewidth]{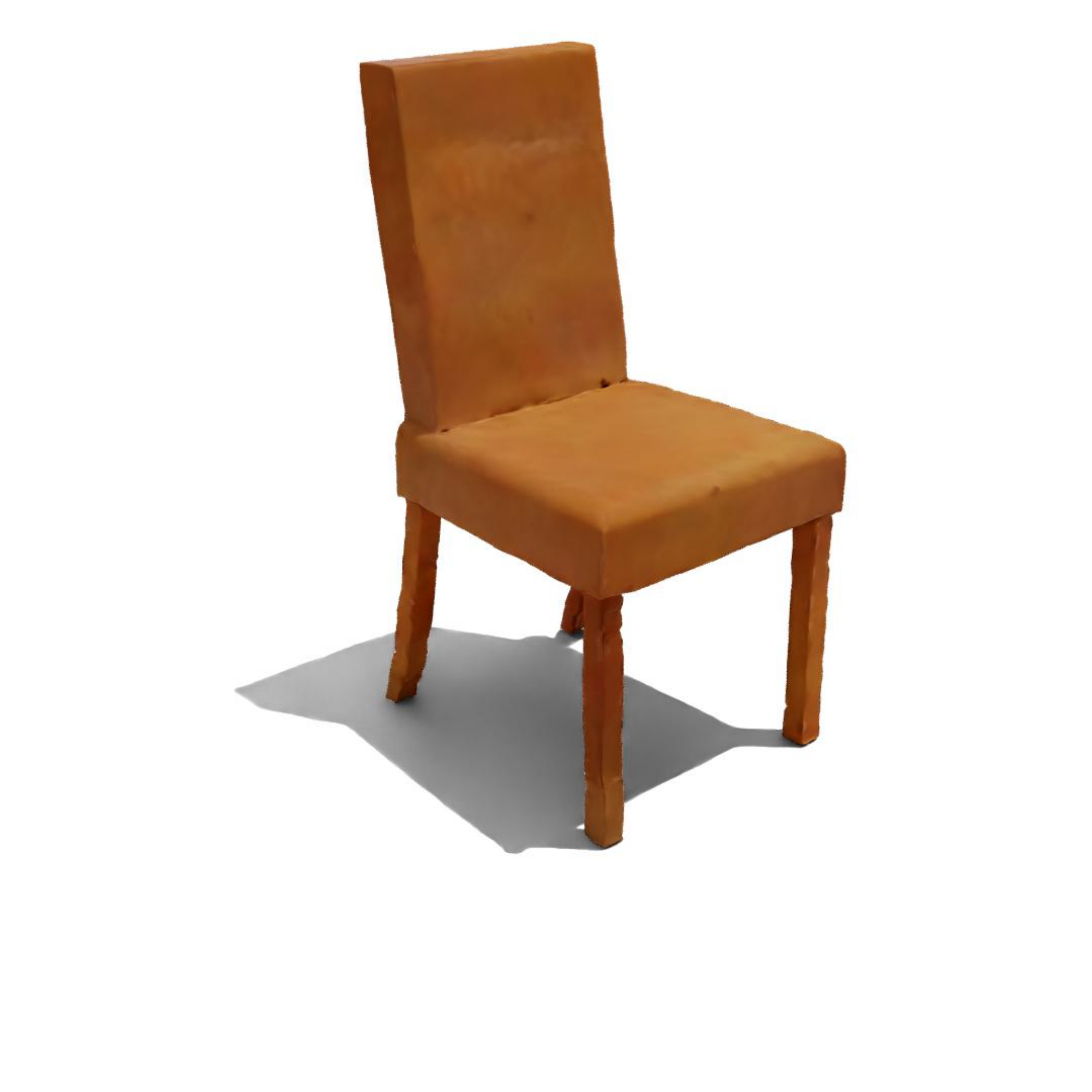}\includegraphics[width=0.1\linewidth]{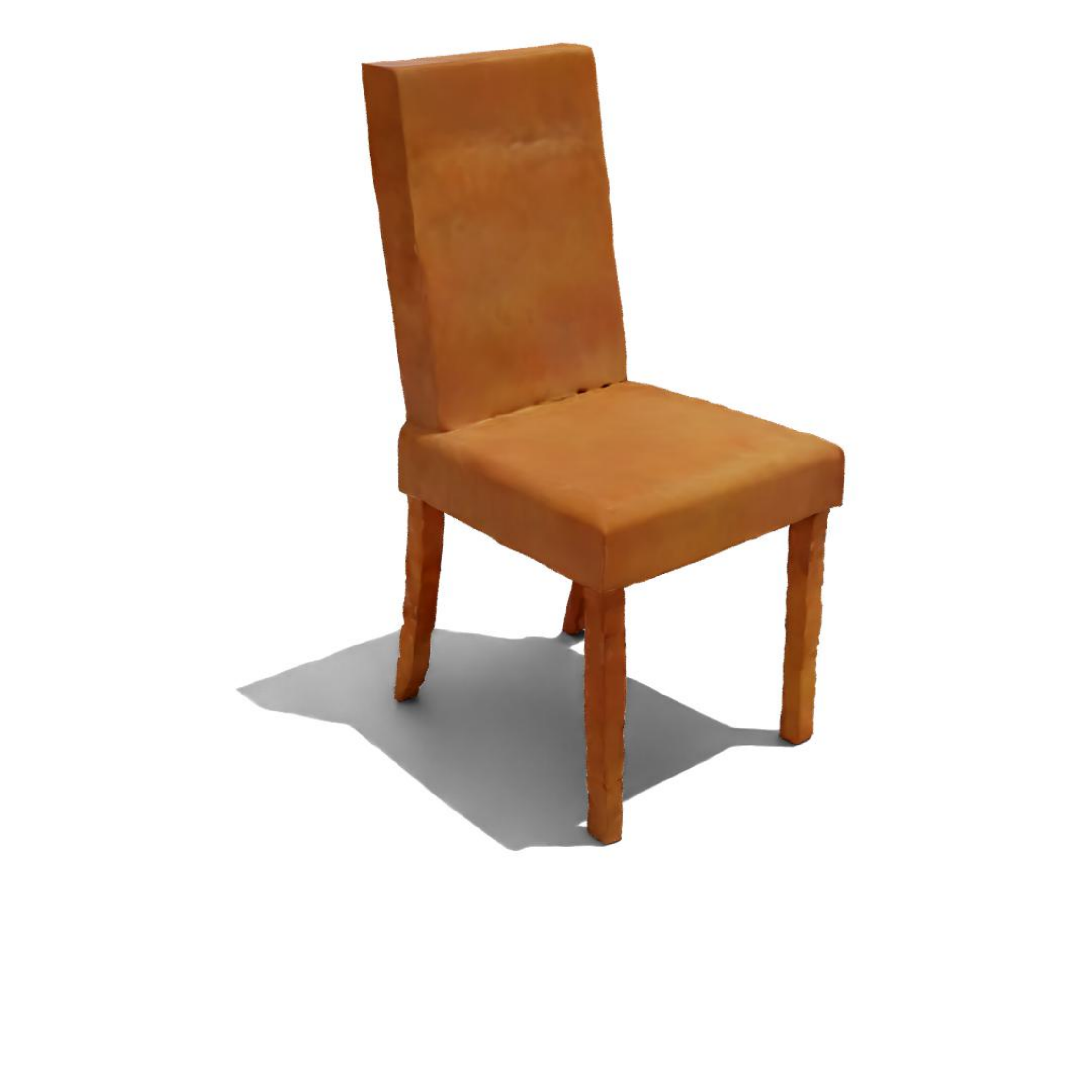}\includegraphics[width=0.1\linewidth]{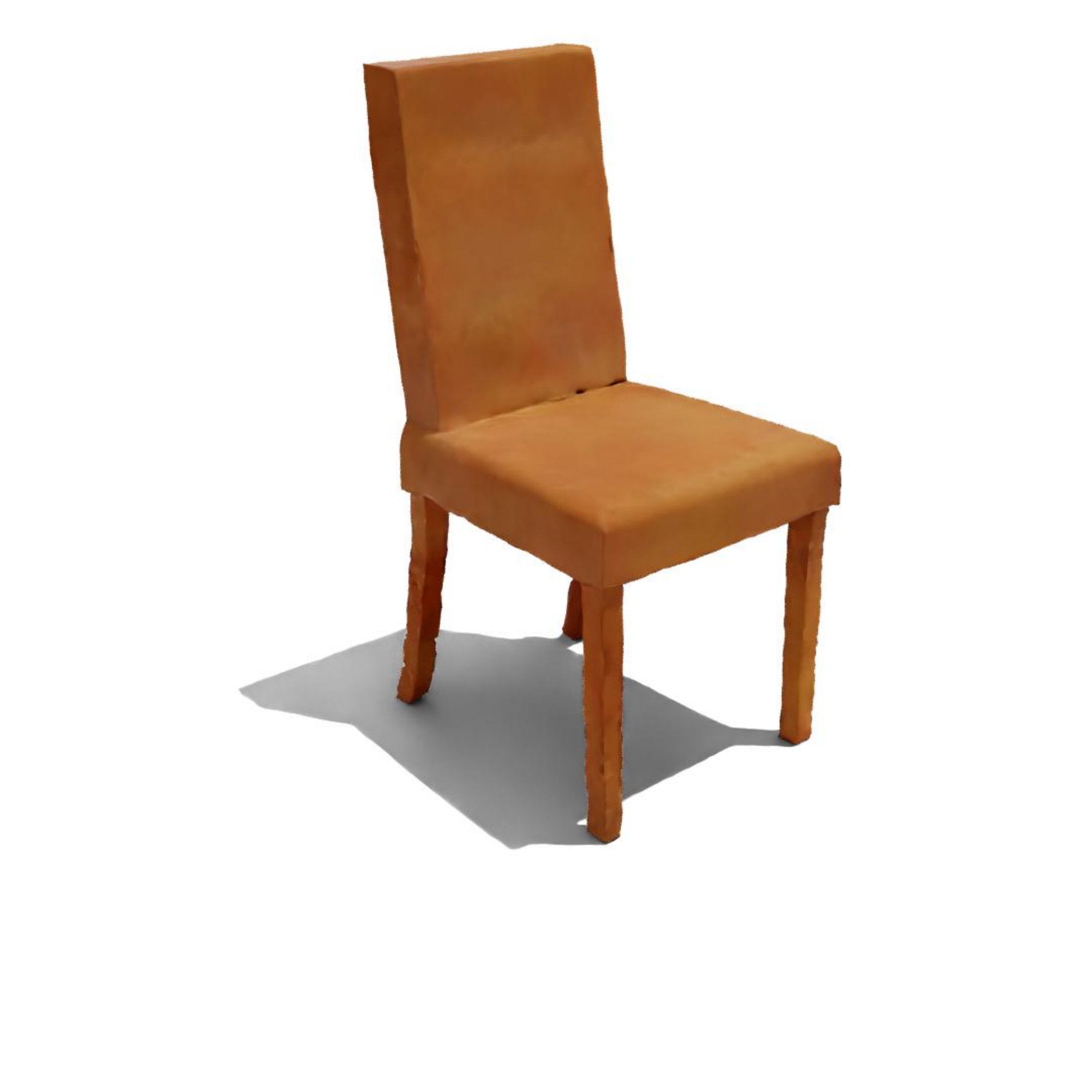}\includegraphics[width=0.1\linewidth]{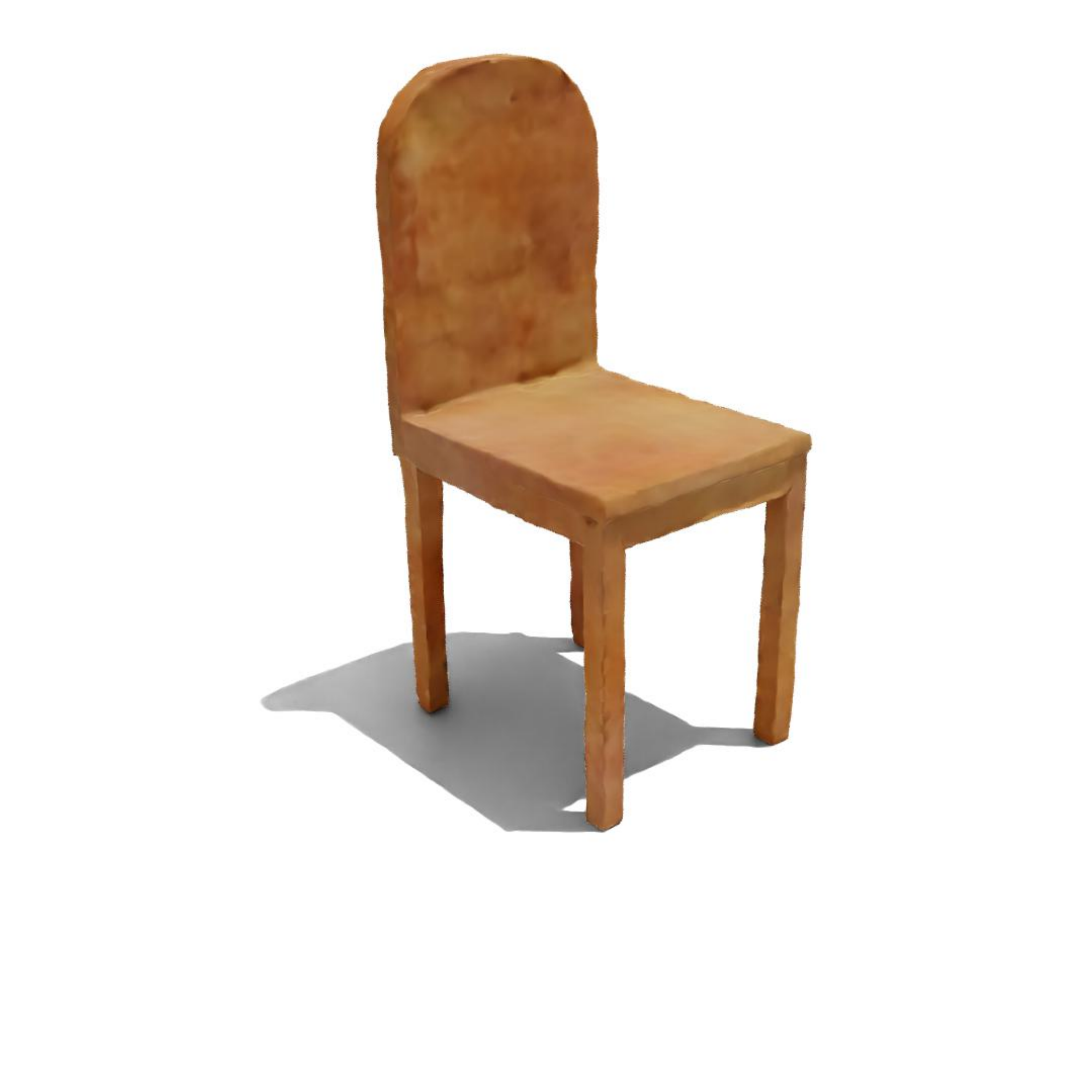}
\caption{\textbf{Row-wise interpolation between two reference shapes:} We first create the extreme left and right samples, storing the noise at each diffusion step. Then we use spherical linear interpolation of these noises to produce intermediate shapes.}
\vspace{-0.4cm}
\label{fig:interpolation}
\end{figure*}

\paragraph{Interpolation.}
In Fig.~\ref{fig:interpolation} we explore the latent shape space learned by TetraDiffision. The model allows for direct interpolation between two different shape instances by spherical interpolation of their generating noise patterns. The intermediate noises are fed into the diffusion model to generate interpolated meshes. The geometric integrity of the intermediate samples and the plausible, gradual transitions suggest that the model indeed disentangles functionally meaningful dimensions and represents objects in terms of those components.

\subsection{Quantitative results}

\paragraph{Metrics.}

\begin{table}[b!]

\caption{\textbf{Quantitative evaluation of \emph{TetraDiffusion} and current mesh generators.} Metrics are computed over ShapeNet classes airplane, bike, car and chair. MMD-CD is multiplied $\times 10^3$, MMD-EMD $\times 10^2$. We use our high resolution model for all experiments except chairs. $^\dag$ NFD could not be retrained as there is no public code, thus training and test split overlap.}
\label{tab:1nna}

\centering
\setlength{\tabcolsep}{6pt}

\resizebox{1.0\linewidth}{!}{
\begin{tabular}{@{}llccccccccc}
\toprule
    \multicolumn{2}{@{}l}{}  & \multicolumn{2}{c}{1-NNA $\downarrow$}&  \multicolumn{2}{c}{MMD $\downarrow$} &  \multicolumn{2}{c}{COV $\uparrow$} & \multicolumn{3}{c}{shading-FID}\\ 
    \hhline{~~---------}
    \multicolumn{1}{@{}l}{Category} & Method & CD & EMD & CD & EMD & CD & EMD & FID $\downarrow$ & CLIP-FID $\downarrow$ & KID $\downarrow$ \\
    \midrule
    \multirow{6}{*}{Airplane}
    & NFD$\dagger$ \cite{shue20233d}                 &\cellcolor{tabyellow}87.3 & 91.0& \cellcolor{tabyellow}0.40& \cellcolor{tabyellow}0.79 & \cellcolor{tabyellow}40.5 & 23.0 & 36.9 & 2.39 & 2.38  \\
    & SDF-SG \cite{zheng2022sdf}                       &98.0 & 97.3& 0.64& 1.14 & 32.8 & 15.3 & 80.7 & 6.66 & 7.17  \\
    & NWD \cite{hui2022neural}                          %
                                    &90.4 & \cellcolor{tabyellow}90.9& 0.44& \cellcolor{tabyellow}0.79 & 36.8 & \cellcolor{tabyellow}27.9 & \cellcolor{tabyellow}20.2 & \cellcolor{tabyellow}2.09 & \cellcolor{tabyellow}1.35 \\
    & GET3D \cite{gao2022get3d}                        &89.9 & 93.6& 0.49& \cellcolor{tabyellow}0.79 & 35.8 & 22.7 & 27.0 & 5.35 & 1.56  \\
    & MD \cite{Liu2023MeshDiffusion}                          &92.7 & 93.7& 0.73& 1.23 & 36.1 & 21.0 & 82.9 & 8.78 & 7.28\\
     & \textbf{Ours}                &\cellcolor{tabgreen}\textbf{64.9} & \cellcolor{tabgreen}\textbf{85.9}& \cellcolor{tabgreen}\textbf{0.36}& \cellcolor{tabgreen}\textbf{0.69} &\cellcolor{tabgreen} \textbf{44.2} &\cellcolor{tabgreen} \textbf{33.1}  &\cellcolor{tabgreen}\textbf{15.6} &\cellcolor{tabgreen}\textbf{1.02} &\cellcolor{tabgreen}\textbf{0.78} \\ %
    \midrule
    
     \multirow{6}{*}{Car} 
    & NFD$^\dag$ \cite{shue20233d}          &76.3 & \cellcolor{tabyellow}72.9& 1.36& \cellcolor{tabyellow}0.80 & \cellcolor{tabyellow}35.8 & \cellcolor{tabgreen}\textbf{44.6}  & 143.9 & 9.60 & 11.6 \\
    & SDF-SG \cite{zheng2022sdf}           &87.4 & 82.1& 1.47& 0.85 & 29.0 & 36.1  & 146.1 & 13.6 & 11.4 \\
    & NWD \cite{hui2022neural}              &75.9 & 77.7& 1.38& 0.82 & 26.7 & 31.8  & \cellcolor{tabyellow}116.3 & \cellcolor{tabyellow}8.37 & \cellcolor{tabyellow}9.09 \\
    & GET3D \cite{gao2022get3d}            &90.9 & 90.5& 1.69& 1.02 & 10.2 & 19.6  & 169.9 & 12.6 & 15.3 \\
    & MD \cite{Liu2023MeshDiffusion}               &\cellcolor{tabyellow}64.5 & 77.8& \cellcolor{tabyellow}1.29& 0.83 & \cellcolor{tabgreen}\textbf{38.1} &38.9  & 151.5 & 10.1 & 13.6\\

     & \textbf{Ours}      &\cellcolor{tabgreen}\textbf{60.1} & \cellcolor{tabgreen}\textbf{69.9}& \cellcolor{tabgreen}\textbf{1.26}&\cellcolor{tabgreen}\textbf{0.78} & 34.94 & \cellcolor{tabyellow}42.61  &\cellcolor{tabgreen}\textbf{ 107.2}&\cellcolor{tabgreen}\textbf{5.15} & \cellcolor{tabgreen}\textbf{8.63} \\ 

    \midrule

    \multirow{5}{*}{Chair} 
    & SDF-SG  \cite{zheng2022sdf}                      &85.2 & 83.4& \cellcolor{tabyellow}4.76& 2.82 & 30.9 & 29.0  & 74.1 & 9.57 & 4.91 \\
    & NWD \cite{hui2022neural}                          &\cellcolor{tabyellow}65.2 & \cellcolor{tabyellow}66.0& \cellcolor{tabgreen}\textbf{3.90}& \cellcolor{tabgreen}\textbf{2.18} & \cellcolor{tabyellow}43.4 & \cellcolor{tabyellow}45.3  & \cellcolor{tabgreen}\textbf{16.7} & \cellcolor{tabgreen}\textbf{2.23} & \cellcolor{tabgreen}\textbf{0.69 }\\
     & GET3D \cite{gao2022get3d}                       &74.7 & 71.8& 5.35& 2.68 & 31.6 & 36.4  & 41.3 & 6.09 & \cellcolor{tabyellow}2.36 \\
     & MD \cite{Liu2023MeshDiffusion}                          &74.9 & 70.4& 5.28& 2.75 & 38.5 & 43.3  & 60.9 & 9.19 & 4.33 \\
     & \textbf{Ours}                       &\cellcolor{tabgreen}\textbf{61.2} & \cellcolor{tabgreen}\textbf{63.2}& 4.90& \cellcolor{tabyellow}2.66 & \cellcolor{tabgreen}\textbf{45.6} & \cellcolor{tabgreen}\textbf{47.2}  & \cellcolor{tabyellow}39.9& \cellcolor{tabyellow}4.17 & 2.67 \\ %

    \midrule
    
    \multirow{4}{*}{Bike} 
     & SDF-SG \cite{zheng2022sdf}              &89.8 & 88.3& 1.77& 1.25 & 20.5 & 32.1  & 198.2 & 23.2 & 8.41 \\
     & NWD \cite{hui2022neural}                 & 69.1 & \cellcolor{tabyellow}71.1& \cellcolor{tabyellow}1.09 & 0.96 & \cellcolor{tabyellow}42.7 & \cellcolor{tabyellow}47.8 & 131.0 & 11.4 & 10.3\\
     & GET3D \cite{gao2022get3d}               &73.7 & 76.9& 1.45& 1.04 & 26.4 & 38.9 & 135.8 & 10.6 & 10.7 \\
     & MD \cite{Liu2023MeshDiffusion}                  &\cellcolor{tabgreen}\textbf{54.0} & 72.0& \cellcolor{tabgreen}\textbf{0.99}& \cellcolor{tabgreen}\textbf{0.93} &\cellcolor{tabgreen}\textbf{46.6} & 44.2  & \cellcolor{tabyellow}56.0 & \cellcolor{tabyellow}3.25 & \cellcolor{tabyellow}4.19\\
          
      &\textbf{Ours}           &\cellcolor{tabyellow}58.5 & \cellcolor{tabgreen}\textbf{66.9}& 1.17 & \cellcolor{tabyellow}0.95 & 40.1 & \cellcolor{tabgreen}\textbf{48.1}  &\cellcolor{tabgreen}\textbf{47.9} & \cellcolor{tabgreen}\textbf{2.94}& \cellcolor{tabgreen}\textbf{3.00}\\
\bottomrule
\bottomrule
\vspace{0.5em}
\end{tabular}%
} %
\end{table}
We quantitatively evaluate our model in terms of \textit{1-nearest neighbor accuracy} (1NNA), as proposed by Yang \etal \cite{yang2019pointflow}, as well as minimum matching distance (MMD) and coverage (COV) (\cref{tab:1nna}). Like~\cite{zeng2022lion}, we compute all metrics based on both Earth Mover's Distance (EMD) and Chamfer Distance (CD). 1-NNA evaluates how closely the generated samples match the distribution of real examples by counting how often the nearest neighbor of a generation result is a real, respectively generated one. 1-NNA scores should lie near 50\%, with higher scores suggesting underfitting and lower scores indicating overfitting. MMD measures how far generated examples are from real ones, via the average distance to the nearest reference sample. COV measures how well the generated samples represent the diversity of the ground truth, with higher values indicating better coverage of the ground truth's variability. Additionally, by rendering all the meshes, we evaluate the similarity to the original ShapeNet class through the Fréchet Inception Distance (FID), CLIP-FID, and Kernel Inception Distance (KID) of shaded meshes (\cref{tab:1nna}). This allows us to assess visual similarity between generated meshes (without color) and their respective ShapeNet categories. To assess color fidelity we do the same for colored meshes (\cref{tab:FID}).

We compare our model with leading 3D mesh generators, particularly noting GET3D~\cite{gao2022get3d} and MeshDiffusion (MD)~\cite{Liu2023MeshDiffusion} as direct competitors that also employ tetrahedral grids.
Additionally, we benchmark against a GAN-based mesh generator (SDF-SG, \cite{zheng2022sdf}), diffusion in the wavelet domain (NWD, \cite{hui2022neural}) and diffusion on triplanes (NFD, \cite{shue20233d}). For NFD we could not retrain and had to use the authors' weights as no training code is available, hence there is an overlap between training and test set. TetraDiffusion and GET3D are the only methods that also generate color. 

\begin{wrapfigure}{R}{0.58\textwidth}
\vspace{-2em}
\captionof{table}{\textbf{Quantitative comparison of generative texture quality}, evaluated via renderings of colored meshes. We use our high resolution model.}
\label{tab:FID}
\setlength{\tabcolsep}{6pt}
\resizebox{\linewidth}{!}{
\begin{tabular}{@{}llcccc}
\toprule
 & & \multicolumn{3}{c}{Texture-FID $\downarrow$}  \\

 &&FID&CLIP-FID&KID&\\
\hhline{~~---}
\multirow{2}{*}{Airplane} &GET3D \cite{gao2022get3d} &41.91&11.87&0.023\\
                & \textbf{Ours} &\textbf{22.51}&\textbf{2.35}&\textbf{0.013}\\
\midrule
\multirow{2}{*}{Car} & GET3D \cite{gao2022get3d} & \textbf{40.30} &\textbf{7.0}&\textbf{0.023}\\
& \textbf{Ours}   &61.05&8.5&0.041\\
\midrule
\multirow{2}{*}{Chair} & GET3D \cite{gao2022get3d} & 68.89&18.91&0.043\\
& \textbf{Ours}  &\textbf{41.63}&\textbf{15.34}&\textbf{0.026}\\
\midrule
\multirow{2}{*}{Bike} & GET3D \cite{gao2022get3d} &87.79&9.44&0.068\\
& \textbf{Ours}  &\textbf{52.24}&\textbf{8.65}&\textbf{0.029}\\                  
\bottomrule
\bottomrule
\end{tabular}
}
\vspace{-1em}
\end{wrapfigure}
The results presented in \cref{tab:1nna} show that our approach performs very well in a head-to-head comparison. It achieves the top 1-NNA scores (nearing 50) in all cases except for the bikes CD category, where it ranks second. Regarding COV scores, our method outperforms others in 5 out of 8 categories, achieving the highest scores, and takes the second position for Cars EMD.
In the evaluation of MMD, our approach leads by registering the smallest distances for both Airplane and Car categories, while it ranks second for Chair and Bike EMD. Additionally, our high-resolution variant, Ours$_{192}$, outclasses all competing methods in Shading-FID scores (\cref{tab:1nna}), and our Ours$_{128}$ variant secures the second-highest Shading-FID score in the Chair category.
When assessing the texture quality of our generated meshes, TetraDiffusion beats GET3D across all three FID scores for Airplane, Chair, and Bike, as detailed in \cref{tab:FID}. We also compared our method to LION and PVD, both of which generate point clouds only. See the supplementary material. TetraDiffusion outperforms both in most of the metrics, too, although the comparison across different native shape representations comes with its own challenges and should be taken with a grain of salt.

\subsection{Efficiency}

A central asset of TetraDiffusion is its efficiency. We argue that, despite the rapid development of GPU hardware, 3D generative models are at present held back by hardware limits; especially GPU memory, but also training time could become prohibitive when scaling current methods up to industrial scale. In other words, a lighter and more efficient design makes 3D generative models not only faster and cheaper, but also yields results of higher quality. 

By performing convolution natively on the tetrahedral grid, we can avoid overlaying a higher-resolution voxel grid as done, for instance, in MeshDiffusion. 
\begin{wraptable}{l}{0.6\textwidth} 
    \vspace{-1em}
    \captionof{table}{\textbf{Memory consumption and runtime of generative shape models.} For a fair comparison we also implemented 16-bit inference for MD. *$\,$denotes TetraDiffusion with pruning. NWD training numbers are for only on the first of two training stages. For NFD there is no public training code.} 
    \label{table:efficiency}
    \resizebox{\linewidth}{!}{
    \begin{tabular}{@{}lcccc@{}}
    \toprule
        & \multicolumn{2}{c}{Training} & \multicolumn{2}{c}{Inference} \\
        \cmidrule{2-5}
        Method\phantom{abc} & GPU (GB) & Speed (it/s) & GPU (GB) & Speed (s/obj) \\ 
        \midrule
        SDF \cite{zheng2022sdf}& 2.9 & 2.4 & \textcolor{white}{0}4.0 & \textcolor{white}{00}0.2 \\ 
        NWD \cite{hui2022neural}& (5.3) & (15.3) & \textcolor{white}{0}5.6 & \textcolor{white}{00}3.6 \\ 
        NFD \cite{shue20233d} & --- & --- & \textcolor{white}{0}4.9 & \textcolor{white}{0}23.3 \\ 
        GET3D \cite{gao2022get3d} & 13.3 & 0.1 & 11.3 & \textcolor{white}{00}0.8 \\ 
        MD(org) \cite{Liu2023MeshDiffusion} & 76.6 & 0.5 & 29.2 & 714.3 \\
        MD(16 bit) \cite{Liu2023MeshDiffusion} & 76.6 & 0.5 & 22.6 & 526.3 \\
        \midrule
        \textbf{Ours$_{\textbf{128}}^\textbf{*}$} & 12.0 & 2.8 & \textcolor{white}{1}7.4 & \textcolor{white}{00}3.4 \\ 
        \textbf{Ours\textsubscript{128}} & 20.8 & 1.0 & \textcolor{white}{1}9.7 & \textcolor{white}{0}11.2 \\
        \textbf{Ours$_\textbf{{192}}^\textbf{*}$} & 20.9 & 1.2 & 11.7 & \textcolor{white}{00}9.1 \\ 
        \textbf{Ours\textsubscript{192}} & 78.2 & 0.3 & 42.1 & \textcolor{white}{0}33.3 \\ 
        \bottomrule
        \bottomrule
    \end{tabular}
    }
\end{wraptable}
What is more, the voxel grid implies convolution kernels with at least $3\times 3\times 3=27$, whereas our operator works well with only 16 neighbors.
This, in turn, makes it possible to increase the network capacity under the same hardware constraints. For instance TetraDiffusion can, due to its smaller memory footprint, work with 1028 feature channels in the bottleneck, and can also be run at higher resolution.

Table \ref{table:efficiency} compares the computational demands of different 3D generative models. Our peak memory consumption at resolution $R=128$ (with batch size 1) is 20.8 GB, in contrast to MeshDiffusion's 76.6 GB. Pruning the data cube as described in \Cref{sec:tetrahedrallearning} further reduces the memory footprint to 12GB, without compromising prediction quality. Even at high resolution $R=192$, and assuming that shape variability is so large that no pruning is possible, our model consumes less than 80 GB and can be trained on a single high-end GPU.
As a consequence, TetraDiffusion is significantly faster: Inference takes $\approx3$ seconds per sample, $\approx200\times$ less than MeshDiffusion on the same hardware, and approaching GAN-based methods.
Furthermore, the time-continuous formulation of our diffusion model provides the flexibility to use a varying number of sampling steps during inference. In our experiments, described in the supplementary material, we have explored the range of steps and found that we can generate a mesh in 32 diffusion steps without compromising quality.

\subsection{Ablations}

\begin{figure}
    \centering
    \begin{subfigure}[b]{0.5\textwidth}
    \includegraphics[width=\linewidth]{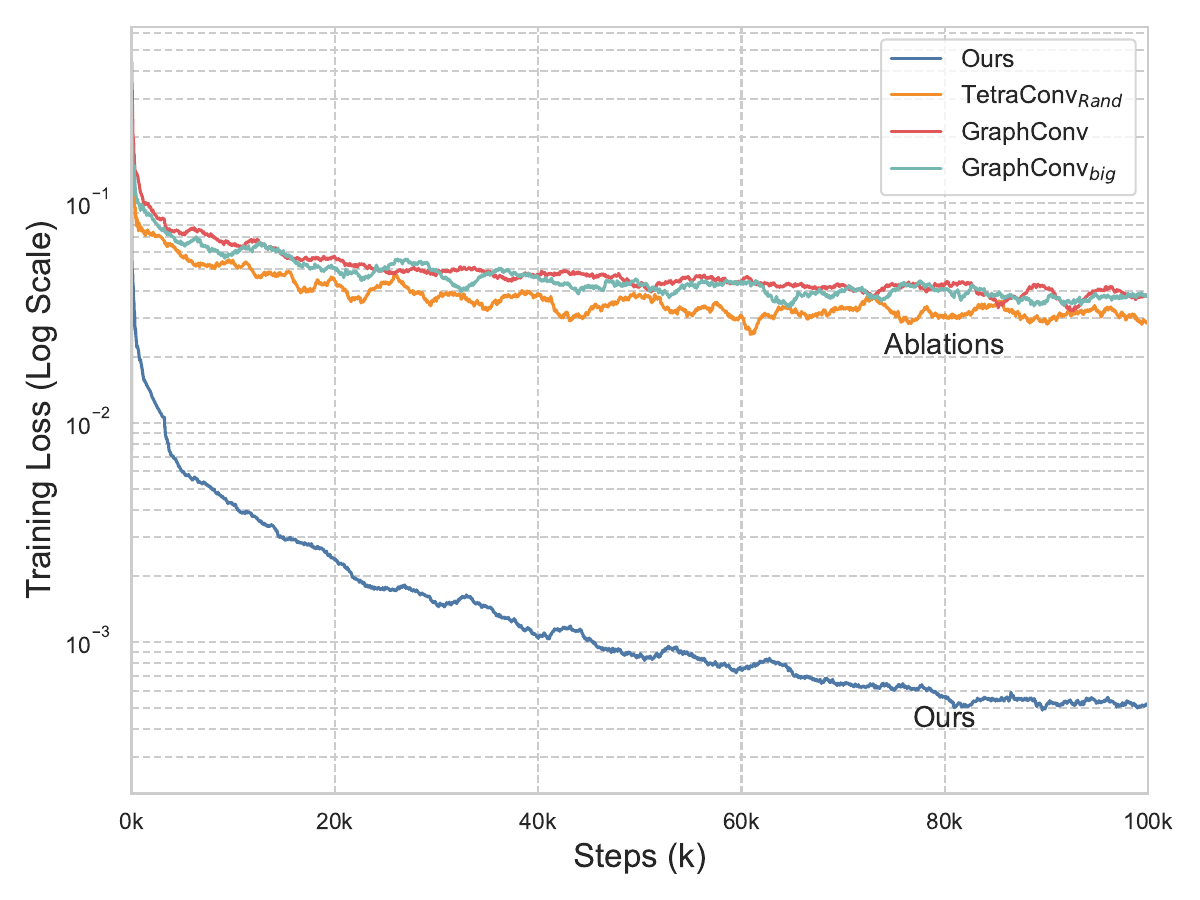}
        \subcaption{}
    \end{subfigure}
    \hspace{3mm}
    \begin{subfigure}[b]{0.40\textwidth}
        
        \includegraphics[width=0.45\textwidth, trim=0cm 1.35cm 0cm 0cm, clip]{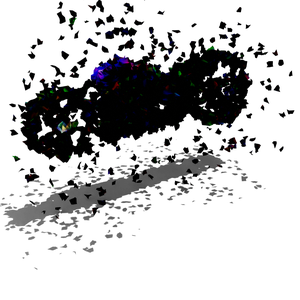}
        \includegraphics[width=0.45\textwidth, trim=0cm 1.35cm 0cm 0cm, clip]{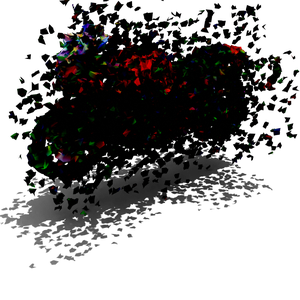}
        \begin{minipage}{0.45\linewidth}
        \centering
        GraphConv
        \end{minipage}
        \hfill
        \begin{minipage}{0.45\linewidth}
        \centering
        GraphConv$_{big}$
        \end{minipage}
        \includegraphics[width=0.45\textwidth, trim=0cm 1.35cm 0cm 0cm, clip]{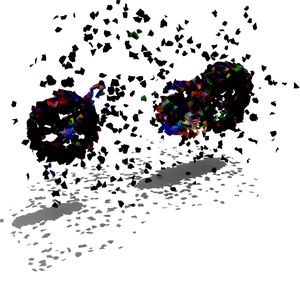}
        \includegraphics[width=0.45\textwidth, trim=0cm 1.35cm 0cm 0cm, clip]{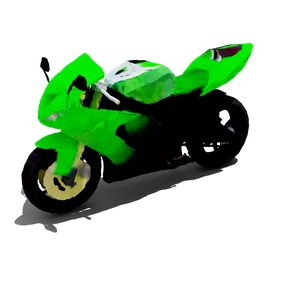}
        \begin{minipage}{0.45\linewidth}
        \centering
        TetraConv$_{Rand}$
        \end{minipage}
        \hfill
        \begin{minipage}{0.45\linewidth}
        \centering
        Ours
        \end{minipage}
        \subcaption{}
    \end{subfigure}
    \caption{\textbf{(a)} Training loss and \textbf{(b)} inference results using graph convolutions (GraphConv and GraphConv$_{big}$), randomized node ordering (TetraConv$_{R}$) and our spatially correct ordering (Ours).}
    \label{fig:ablation}
\end{figure}

To demonstrate the importance of the spatial ordering in our tetrahedral convolution, we run control experiments with random ordering of the neighboring nodes, and with graph convolutions that are invariant to the ordering. We define four otherwise identical networks: (1) our standard \textit{Ours}, equipped with TetraConv; (2) \textit{TetraConv$_{Rand}$}, with randomized ordering of the neighbors within each TetraConv layer; (3) \textit{GraphConv} graph convolutions instead of TetraConv; (4) another version of GraphConv, \textit{GraphConv$_{big}$}, with increased number of parameters to match the parameter count of our standard model with TetraConv. All variants were trained on the motorbike category, with the same training protocol.

As shown in \cref{fig:ablation}, TetraConv$_{Rand}$, GraphConv and GraphConv$_{big}$ all fail to converge to meaningful solutions. Only our standard model, which preserves the spatial ordering, is able to produce convincing results. The experiment supports our claim that a convolution operator should (as per its mathematical definition) account for the spatial layout of the function values. We note in particular that graph convolutions, despite their name, are not convolutions in the formal sense, since they combine information from the neighborhood in a permutation-invariant manner, thus discarding the spatial layout.

\section{Conclusion}
\vspace{-0.1em}

We have presented \emph{TetraDiffusion}, an innovative 3D diffusion framework capable of generating high-resolution, colored triangle meshes with arbitrary topology at unprecedented resolution. To our knowledge, our work is the first 3D diffusion model that natively operates on a tetrahedral data structure, enabling highly efficient training and sampling. Through a series of experiments, we have showcased its ability to synthesize diverse 3D shapes of high geometric quality and the advantages of our tetrahedral convolution over traditional methods such as 3D and graph convolutions.

We believe that a promising avenue for further exploration is to better leverage the extended differentiable tetrahedral marching algorithm and differentiable rendering \cite{munkberg2022extracting, laine2020modular}. This could pave the way towards a seamless integration of large-scale 2D data into the framework. Moreover, \emph{TetraDiffusion} has so far been trained for isolated objects, yet our framework seems a good fit for synthesising entire scenes, with larger extent and often high sparsity.

\clearpage

\bibliographystyle{splncs04}
\bibliography{main}

\begin{thebibliography}{10}
\providecommand{\url}[1]{\texttt{#1}}
\providecommand{\urlprefix}{URL }
\providecommand{\doi}[1]{https://doi.org/#1}

\bibitem{achlioptas2018learning}
Achlioptas, P., Diamanti, O., Mitliagkas, I., Guibas, L.: Learning representations and generative models for 3d point clouds. In: ICML (2018)

\bibitem{alliegro2023polydiff}
Alliegro, A., Siddiqui, Y., Tommasi, T., Nie{\ss}ner, M.: Polydiff: Generating 3d polygonal meshes with diffusion models. arXiv preprint arXiv:2312.11417  (2023)

\bibitem{brock2016generative}
Brock, A., Lim, T., Ritchie, J.M., Weston, N.: Generative and discriminative voxel modeling with convolutional neural networks. arXiv:1608.04236  (2016)

\bibitem{bruna2013spectral}
Bruna, J., Zaremba, W., Szlam, A., {LeCun}, Y.: Spectral networks and locally connected networks on graphs. arXiv:1312.6203  (2013)

\bibitem{cai2020learning}
Cai, R., Yang, G., Averbuch-Elor, H., Hao, Z., Belongie, S., Snavely, N., Hariharan, B.: Learning gradient fields for shape generation. In: ECCV (2020)

\bibitem{shapenet2015}
Chang, A.X., Funkhouser, T., Guibas, L., Hanrahan, P., Huang, Q., Li, Z., Savarese, S., Savva, M., Song, S., Su, H., Xiao, J., Yi, L., Yu, F.: {ShapeNet}: An information-rich 3d model repository. arXiv:1512.03012  (2015)

\bibitem{chen2021crossvit}
Chen, C.F.R., Fan, Q., Panda, R.: Crossvit: Cross-attention multi-scale vision transformer for image classification. In: ICCV (2021)

\bibitem{chen2019learning}
Chen, Z., Zhang, H.: Learning implicit fields for generative shape modeling. In: CVPR (2019)

\bibitem{crowson2021clip}
Crowson, K.: {CLIP} guided diffusion {HQ} 256x256. Colab Notebook. URL https://colab. research. google. com/drive/12a\_Wrfi2\_ gwwAuN3VvMTwVMz9TfqctNj  (2021)

\bibitem{dhariwal2021diffusion}
Dhariwal, P., Nichol, A.: Diffusion models beat {GANs} on image synthesis. In: NeurIPS (2021)

\bibitem{doran2013isosurface}
Doran, C., Chang, A., Bridson, R.: Isosurface stuffing improved: acute lattices and feature matching. In: ACM SIGGRAPH (2013)

\bibitem{gadelha2018multiresolution}
Gadelha, M., Wang, R., Maji, S.: Multiresolution tree networks for 3d point cloud processing. In: ECCV (2018)

\bibitem{gao2020learning}
Gao, J., Chen, W., Xiang, T., Jacobson, A., McGuire, M., Fidler, S.: Learning deformable tetrahedral meshes for 3d reconstruction. In: NeurIPS (2020)

\bibitem{gao2022get3d}
Gao, J., Shen, T., Wang, Z., Chen, W., Yin, K., Li, D., Litany, O., Gojcic, Z., Fidler, S.: Get3d: A generative model of high quality 3d textured shapes learned from images. NeurIPS  (2022)

\bibitem{gao2022tetgan}
Gao, W., Wang, A., Metzer, G., Yeh, R.A., Hanocka, R.: {TetGAN}: A convolutional neural network for tetrahedral mesh generation. arXiv:2210.05735  (2022)

\bibitem{gupta20233dgen}
Gupta, A., Xiong, W., Nie, Y., Jones, I., O{\u{g}}uz, B.: {3DGen}: Triplane latent diffusion for textured mesh generation. arXiv:2303.05371  (2023)

\bibitem{Guttenberg2023}
Guttenberg, N.: \url{https://www.crosslabs.org/blog/diffusion-with-offset-noise}

\bibitem{hendrycks2016gaussian}
Hendrycks, D., Gimpel, K.: Gaussian error linear units ({GELUs}). arXiv:1606.08415  (2016)

\bibitem{ho2020denoising}
Ho, J., Jain, A., Abbeel, P.: Denoising diffusion probabilistic models. In: NeurIPS (2020)

\bibitem{ho2022classifier}
Ho, J., Salimans, T.: Classifier-free diffusion guidance. arXiv:2207.12598  (2022)

\bibitem{hoogeboom2023simple}
Hoogeboom, E., Heek, J., Salimans, T.: Simple diffusion: End-to-end diffusion for high resolution images. arXiv:2301.11093  (2023)

\bibitem{hui2022neural}
Hui, K.H., Li, R., Hu, J., Fu, C.W.: Neural wavelet-domain diffusion for 3d shape generation. In: SIGGRAPH Asia (2022)

\bibitem{jacobson2011bounded}
Jacobson, A., Baran, I., Popovic, J., Sorkine, O.: Bounded biharmonic weights for real-time deformation. ACM ToG  \textbf{30}(4), ~78 (2011)

\bibitem{kingma2021variational}
Kingma, D.P., Salimans, T., Poole, B., Ho, J.: Variational diffusion models. In: NeurIPS (2021)

\bibitem{kipf2016semi}
Kipf, T.N., Welling, M.: Semi-supervised classification with graph convolutional networks. arXiv:1609.02907  (2016)

\bibitem{laine2020modular}
Laine, S., Hellsten, J., Karras, T., Seol, Y., Lehtinen, J., Aila, T.: Modular primitives for high-performance differentiable rendering. ACM ToG  \textbf{39}(6),  1--14 (2020)

\bibitem{lin2023common}
Lin, S., Liu, B., Li, J., Yang, X.: Common diffusion noise schedules and sample steps are flawed. arXiv:2305.08891  (2023)

\bibitem{Liu2023MeshDiffusion}
Liu, Z., Feng, Y., Black, M.J., Nowrouzezahrai, D., Paull, L., Liu, W.: {MeshDiffusion}: Score-based generative 3d mesh modeling. In: ICLR (2023)

\bibitem{luo2022understanding}
Luo, C.: Understanding diffusion models: A unified perspective. arXiv:2208.11970  (2022)

\bibitem{luo2021diffusion}
Luo, S., Hu, W.: Diffusion probabilistic models for 3d point cloud generation. In: CVPR (2021)

\bibitem{mo2023dit}
Mo, S., Xie, E., Chu, R., Yao, L., Hong, L., Nie{\ss}ner, M., Li, Z.: {DiT-3D}: Exploring plain diffusion transformers for 3d shape generation. arXiv:2307.01831  (2023)

\bibitem{munkberg2022extracting}
Munkberg, J., Hasselgren, J., Shen, T., Gao, J., Chen, W., Evans, A., M{\"u}ller, T., Fidler, S.: Extracting triangular 3d models, materials, and lighting from images. In: CVPR (2022)

\bibitem{nichol2021glide}
Nichol, A., Dhariwal, P., Ramesh, A., Shyam, P., Mishkin, P., {McGrew}, B., Sutskever, I., Chen, M.: Glide: Towards photorealistic image generation and editing with text-guided diffusion models. arXiv:2112.10741  (2021)

\bibitem{paille2015dihedral}
Paill{\'e}, G.P., Ray, N., Poulin, P., Sheffer, A., L{\'e}vy, B.: Dihedral angle-based maps of tetrahedral meshes. ACM ToG  \textbf{34}(4),  1--10 (2015)

\bibitem{peng2021shape}
Peng, S., Jiang, C., Liao, Y., Niemeyer, M., Pollefeys, M., Geiger, A.: Shape as points: A differentiable {Poisson} solver. In: NeurIPS (2021)

\bibitem{perez2018film}
Perez, E., Strub, F., De~Vries, H., Dumoulin, V., Courville, A.: Film: Visual reasoning with a general conditioning layer. In: AAAI (2018)

\bibitem{radford2021learning}
Radford, A., Kim, J.W., Hallacy, C., Ramesh, A., Goh, G., Agarwal, S., Sastry, G., Askell, A., Mishkin, P., Clark, J., et~al.: Learning transferable visual models from natural language supervision. In: ICML (2021)

\bibitem{Ramesh2022HierarchicalTI}
Ramesh, A., Dhariwal, P., Nichol, A., Chu, C., Chen, M.: Hierarchical text-conditional image generation with {CLIP} latents. arXiv:2204.06125  (2022)

\bibitem{rombach2022high}
Rombach, R., Blattmann, A., Lorenz, D., Esser, P., Ommer, B.: High-resolution image synthesis with latent diffusion models. In: CVPR (2022)

\bibitem{ronneberger2015u}
Ronneberger, O., Fischer, P., Brox, T.: U-net: Convolutional networks for biomedical image segmentation. In: MICCAI (2015)

\bibitem{saharia2022photorealistic}
Saharia, C., Chan, W., Saxena, S., Li, L., Whang, J., Denton, E.L., Ghasemipour, K., Gontijo~Lopes, R., Karagol~Ayan, B., Salimans, T., et~al.: Photorealistic text-to-image diffusion models with deep language understanding. NeurIPS  (2022)

\bibitem{salimans2022progressive}
Salimans, T., Ho, J.: Progressive distillation for fast sampling of diffusion models. arXiv:2202.00512  (2022)

\bibitem{shen2021deep}
Shen, T., Gao, J., Yin, K., Liu, M.Y., Fidler, S.: Deep marching tetrahedra: a hybrid representation for high-resolution 3d shape synthesis. In: NeurIPS (2021)

\bibitem{shue20233d}
Shue, J.R., Chan, E.R., Po, R., Ankner, Z., Wu, J., Wetzstein, G.: 3d neural field generation using triplane diffusion. In: CVPR (2023)

\bibitem{song2020denoising}
Song, J., Meng, C., Ermon, S.: Denoising diffusion implicit models. In: ICLR (2021)

\bibitem{song2020score}
Song, Y., Sohl-Dickstein, J., Kingma, D.P., Kumar, A., Ermon, S., Poole, B.: Score-based generative modeling through stochastic differential equations. arXiv:2011.13456  (2020)

\bibitem{thomas2019kpconv}
Thomas, H., Qi, C.R., Deschaud, J.E., Marcotegui, B., Goulette, F., Guibas, L.J.: {KPConv}: Flexible and deformable convolution for point clouds. In: ICCV (2019)

\bibitem{wang2020linformer}
Wang, S., Li, B.Z., Khabsa, M., Fang, H., Ma, H.: Linformer: Self-attention with linear complexity. arXiv:2006.04768  (2020)

\bibitem{wegner2021lecture}
Wegner, S.A.: Lecture notes on high-dimensional data. arXiv:2101.05841  (2021)

\bibitem{wu2019simplifying}
Wu, F., Souza, A., Zhang, T., Fifty, C., Yu, T., Weinberger, K.: Simplifying graph convolutional networks. In: ICML (2019)

\bibitem{wu2018group}
Wu, Y., He, K.: Group normalization. In: ECCV (2018)

\bibitem{wu20153d}
Wu, Z., Song, S., Khosla, A., Yu, F., Zhang, L., Tang, X., Xiao, J.: {3d ShapeNets}: A deep representation for volumetric shapes. In: CVPR (2015)

\bibitem{yang2019pointflow}
Yang, G., Huang, X., Hao, Z., Liu, M.Y., Belongie, S., Hariharan, B.: {PointFlow}: 3d point cloud generation with continuous normalizing flows. In: ICCV (2019)

\bibitem{zamorski2020adversarial}
Zamorski, M., Zieba, M., Klukowski, P., Nowak, R., Kurach, K., Stokowiec, W., Trzci{\'n}ski, T.: Adversarial autoencoders for compact representations of 3d point clouds. CVIU  \textbf{193},  102921 (2020)

\bibitem{zeng2022lion}
Zeng, X., Vahdat, A., Williams, F., Gojcic, Z., Litany, O., Fidler, S., Kreis, K.: {LION}: Latent point diffusion models for 3d shape generation. arXiv:2210.06978  (2022)

\bibitem{zhang20233dshape2vecset}
Zhang, B., Tang, J., Niessner, M., Wonka, P.: 3dshape2vecset: A 3d shape representation for neural fields and generative diffusion models. ACM Transactions on Graphics (TOG)  \textbf{42}(4),  1--16 (2023)

\bibitem{zheng2023locally}
Zheng, X.Y., Pan, H., Wang, P.S., Tong, X., Liu, Y., Shum, H.Y.: Locally attentional sdf diffusion for controllable 3d shape generation. ACM Transactions on Graphics (ToG)  \textbf{42}(4),  1--13 (2023)

\bibitem{zheng2022sdf}
Zheng, X., Liu, Y., Wang, P., Tong, X.: {SDF}-{StyleGAN}: Implicit {SDF}-based {StyleGAN} for 3d shape generation. In: Computer Graphics Forum. vol.~41, pp. 52--63 (2022)

\bibitem{zhou20213d}
Zhou, L., Du, Y., Wu, J.: 3d shape generation and completion through point-voxel diffusion. In: ICCV (2021)

\end{thebibliography}

\clearpage
\title{\large{Supplementary Material}}

\author{\phantom{a}
}

\institute{\phantom{a}
}

\maketitle

\makeatletter
\def\authcount#1{} %
\def\lastand{}     %
\def\and{} 

\newcommand{\supplementtoc}{
    \startcontents[sections] %
    \printcontents[sections]{l}{1}{\setcounter{tocdepth}{2}} %
}
\makeatother

\supplementtoc

\clearpage
\newpage

\section{Dataset preprocessing}\label{sec:dataprocessing}

We employ an isosurface stuffing algorithm with the A15 tile \cite{doran2013isosurface} to tessellate a cube $[-1, 1]^3$ at the necessary grid resolutions using Quartet \url{https://github.com/crawforddoran/quartet}. Since our algorithm can handle arbitrary edge configurations, there is no need for post-processing non-symmetrical boundaries. In our experiments, we generate grids with tetrahedral resolutions of 32, 48, 64, 96, 128, and 192.

Our network requires SDF values, deformation vectors and color vectors. In order to convert meshes with possible texture maps, we adapt the optimization pipeline of \cite{munkberg2022extracting} (\emph{nvdiffrec} \url{https://github.com/NVlabs/nvdiffrec/}). In particular, we include per vertex color information and extend the marching tetrahedra algorithm to extract the corresponding surface color (see \cref{sec:marchingtet}). Originally, \emph{nvdiffrec} jointly optimizes geometry and materials, stored in 2D textures, from multi-view object renderings. In our case, material information must be stored in the vertex features to jointly diffuse geometry and texture in a single U-Net, \ie we implement a custom render function that additionally rasters and interpolates the vertex color information extracted from our extended marching tetrahedra. We iteratively optimize SDF value, deformation and color, starting from a random initialization, rendering random views uniformly around a sphere in each step. The vertex features are updated by backward propagation of different 2D and 3D loss functions, making use of the differentiable marching algorithm and differentiable rendering. In particular, we define the following total loss with respect to an optimizable mesh $M_{\text{Opt}}$ extracted from the updated SDF, deformations and colors and the ground truth mesh $M_{\text{Gt}}$.

\begin{equation}
\begin{split}
    \mathcal{L}(M_{\text{Opt}}, M_{\text{Gt}}) = & \ \lambda_1 \mathcal{L}_{\text{Img}} + \lambda_2 \mathcal{L}_{\text{Normal}} + \lambda_3 \mathcal{L}_{\text{Depth}} + \\
    & \lambda_4 \mathcal{L}_{\text{Mask}} + \lambda_5 \mathbb{1}_{2^{nd}} \mathcal{L}_{\text{Laplace}} + \lambda_6  \mathcal{L}_{\text{Reg}}.
\end{split}
\end{equation}

Similar to \cite{Liu2023MeshDiffusion}, the vertex features are updated in a two-stage procedure, each consisting of 5000 iterations in total. In a first step, we limit the deformation vector to a range of 0.45 of the grid resolution, which results in an increase in the number of vertices for the extracted mesh. A higher number of vertices allows better alignment of the ground truth shape in the second step, where we fix all SDF values to their respective sign, \ie SDF $\in \{-1,1\}$. We compensate for the restriction by increasing the range of the deformation vector to twice the grid resolution in the second step.

\boldmath$\mathcal{L}_{\text{Img}}(M_{\text{Opt}}, M_{\text{Gt}})$\unboldmath: We extract the two closest surfaces for each pixel in a raster using depth peeling, which allows us to optimize inner structure as well (see \cref{sec:innerstructure}). After interpolation and antialiasing, this gives us two pairs of RGB images $\{(\text{Img}_{1}^{\text{gt}}, \text{Img}_{1}^{\text{opt}}), (\text{Img}_{2}^{\text{gt}}, \text{Img}_{2}^{\text{opt}})\}$ For which we compute gradients \wrt an L1-loss and $\lambda_1 = 10$.
\\

\boldmath$\mathcal{L}_{\text{Normal}}(M_{\text{Opt}}, M_{\text{Gt}})$\unboldmath: Similar to the RGB images, we extract vertex (smooth shading) normals and face normals (flat shading) for both depth peels and compute gradients of the L1-loss \wrt ground truth  normals. With a slight abuse of notation, we set $\lambda_2 = 10$ for the first layer and $\lambda_2 = 0.1$ for the second layer. Smooth vertex normals are computed as the normalized average of the surrounding surface normals.
\\

\boldmath$\mathcal{L}_{\text{Depth}}(M_{\text{Opt}}, M_{\text{Gt}})$\unboldmath: We compute normalized depth maps for both raster layers and compute the corresponding L1-loss. The depth map is weighted with $\lambda_3 = 100$.
\\

\boldmath$\mathcal{L}_{\text{Mask}}(M_{\text{Opt}}, M_{\text{Gt}})$\unboldmath: We render the silhouette of our shape For the first depth peel only, since the silhouette of the second layer does not provide much information. Again, we use the L1-loss.
\\

\boldmath$\mathcal{L}_{\text{Chamfer}}(M_{\text{Opt}}, M_{\text{Gt}})$\unboldmath: We compute the Chamfer distance \wrt to $M_{\text{Gt}}$ and $M_{\text{Opt}}$ and weight it with $\lambda_4 = 1$.
\\

\boldmath$\mathcal{L}_{\text{Laplace}}(M_{\text{Opt}}, M_{\text{Gt}})$\unboldmath: To regularize the extracted triangular mesh, we add Laplacian smoothing using the weighted umbrella operator in the second stage of our optimization. The loss is attenuated over time.  
\\

\boldmath$\mathcal{L}_{\text{Reg}}(M_{\text{Opt}}, M_{\text{Gt}})$\unboldmath: To diminish floaters, we penalize sign flips in the SDF values with a cross-entropy loss defined over neighboring vertices, following \cite{munkberg2022extracting, gao2022get3d}.
\\

In Addition to the aforementioned regularization loss, we encountered small floaters far away from the actual surface every now and then. In order to remove them, we simply compute the convex hull of our ground truth mesh and enforce all SDF values outside the hull to be positive.

\section{Background about the employed diffusion model}
Denoising diffusion models can be seen as restricted, hierarchical Markovian VAEs \cite{luo2022understanding} that learn to approximate the data distribution $p(\mathbf{x})$ with a sequence of steps. In the variational formulation~\cite{kingma2021variational, hoogeboom2023simple} the steps are indexed by a continuous time variable $t\in [0,1]$. At the final time step $t=1$ the latent variables $\mathbf{z}_t$  should be normally distributed, $q(\mathbf{z}_1) = \mathcal{N}(\mathbf{z}_1;\mathbf{0}, \mathbf{I})$.  The forward process of the marginals $q(\mathbf{z}_t|\mathbf{x})$ is Gaussian and given by
\begin{equation}
    q(\mathbf{z}_t|\mathbf{x}) = \mathcal{N}(\mathbf{z}_t;\alpha_t\mathbf{x},\sigma_t^2\mathbf{I}),
\end{equation}
where $\text{SNR}(t) = \frac{\alpha_t^2}{\sigma_t^2}$ is the signal-to-noise ratio, assumed to decrease strictly monotonically in time, and $\alpha_t$ and $\sigma_t^2$ are strictly positive for all $t$. Consequently, $\mathbf{z}_t$ will be increasingly noisy over time. We fix $\alpha_t^2 = 1 - \sigma_t^2$, corresponding to a variance-preserving process~\cite{song2020score}. Under the Markov assumption the forward transition kernels for $t>s$ are also Gaussian and given by:
\begin{equation}
    q(\mathbf{z}_t|\mathbf{z}_s) = \mathcal{N}(\mathbf{z}_t; \alpha_{t|s}\mathbf{z}_s, \sigma_{t|s}^2\mathbf{I}),
\end{equation}
where $\alpha_{t|s}=\frac{\alpha_t}{\alpha_s}$ and $\sigma_{t|s}^2 = \sigma_t^2 - \alpha_{t|s}^2\sigma_s^2$.
A common noise schedule is $\alpha_t = \cos(\pi t/2)$, which under variance preservation leads to a signal-to-noise ratio of $\text{SNR}(t)=\frac{1}{\tan(\pi t /2)}$. 

We are interested in learning the reverse diffusion process. While $q(\mathbf{z}_s|\mathbf{z}_t)$ is in general intractable as it requires integration over the whole dataset, conditioning it on a data sample $x$ gives rise to a closed-form solution:
\begin{equation}\label{eq:reverse}
    q(\mathbf{z}_s|\mathbf{z}_t, \mathbf{x}) = \mathcal{N}(\mathbf{z}_s; \mu_{s,t}(\mathbf{z}_t, \mathbf{x}), \sigma^2_{s,t} \mathbf{I}),
\end{equation}
where $\mu_{s,t}(\mathbf{z}_t, \mathbf{x}) = \frac{\alpha_{t|s}\sigma_s^2}{\sigma_t^2}\mathbf{z}_t + \frac{\alpha_s\sigma^2_{t|s}}{\sigma_t^2}\mathbf{x}$ and $\sigma^2_{s,t} = \sigma_{t|s}^2\frac{\sigma_s^2}{\sigma_t^2}$. As $\mathbf{x}$ is only available during training, it is replaced by a neural network prediction $\mathbf{\hat{x}}_\theta(\mathbf{z}_t; t) \approx \mathbf{x}$. An equivalent interpretation of the denoising model is as a score model, which in the infinite data limit coverges to the marginal distribution $q(\mathbf{z}_t)$~\cite{song2020denoising}.

To train $\hat{\mathbf{x}}_\theta(\mathbf{z}_t; t)$ one optimizes the variational lower bound of the marginal log-likelihood
\begin{equation}
    \mathcal{L}(\mathbf{x}) = -\frac{1}{2}
    \mathbb{E}_{\substack{\;\epsilon\sim\mathcal{N}(0, \mathbf{I}),\\t\sim\mathcal{N}(0, \mathbf{I})}}
    \bigg[ \text{SNR}'(t)\big\lVert \mathbf{x} - \hat{\mathbf{x}}_\theta(\mathbf{z}_t;t)\big\rVert_2^2\bigg],
\end{equation}
where $\text{SNR}'(t) = \textit{d}\,\text{SNR}/\textit{d}t$ and $\mathbf{z}_t = \alpha_t \mathbf{x} + \sigma_t \epsilon$.
Instead of parametrizing the model to directly recover $\mathbf{x}$ from its corrupted version $\mathbf{z}_t$, one can predict the noise $\hat{\epsilon}$ and recover $\hat{\mathbf{x}}$ from $\hat{\mathbf{x}} = \frac{\mathbf{z}_t}{\alpha_t} - \sigma_t\frac{\hat{\epsilon}_t}{\alpha_t}$. Since that objective tends to destabilize training near $t=1$, we use the more robust \textit{v}-parametrization~\cite{salimans2022progressive}, $\hat{\mathbf{v}}_t = \alpha_t \hat{\epsilon}_t - \sigma_t\hat{\mathbf{x}}$. Note that $\hat{\mathbf{x}} = \alpha_t\mathbf{z}_t-\sigma_t\hat{\mathbf{v}_t}$.
Once trained, we can sample from our data distribution with ancestral sampling. Setting $\epsilon\sim\mathcal{N}(\mathbf{0}, \mathbf{I})$, we starting at $\mathbf{z}_1 \sim \mathcal{N}(\mathbf{0}, \mathbf{I})$ and iteratively denoise it according to
\begin{equation}
    \mathbf{z_s} = \frac{\alpha_t\sigma_s^2}{\alpha_s\sigma_t^2}\mathbf{z}_t + \frac{\alpha_s\sigma_{t|s}^2}{\sigma_t^s}\mathbf{\hat{x}}_\theta(\mathbf{z}_t; t) + \sqrt{\frac{\sigma_{t|s}^2\sigma_s^2}{\sigma_t^2}}\epsilon.
\end{equation}

\section{Additional qualitative results}

\subsection{High resolution and inner structure}\label{sec:innerstructure}

\begin{figure*}[!ht]
\centering
\includegraphics[width=0.30\linewidth, trim={1cm 5cm 9cm 5cm}, clip]{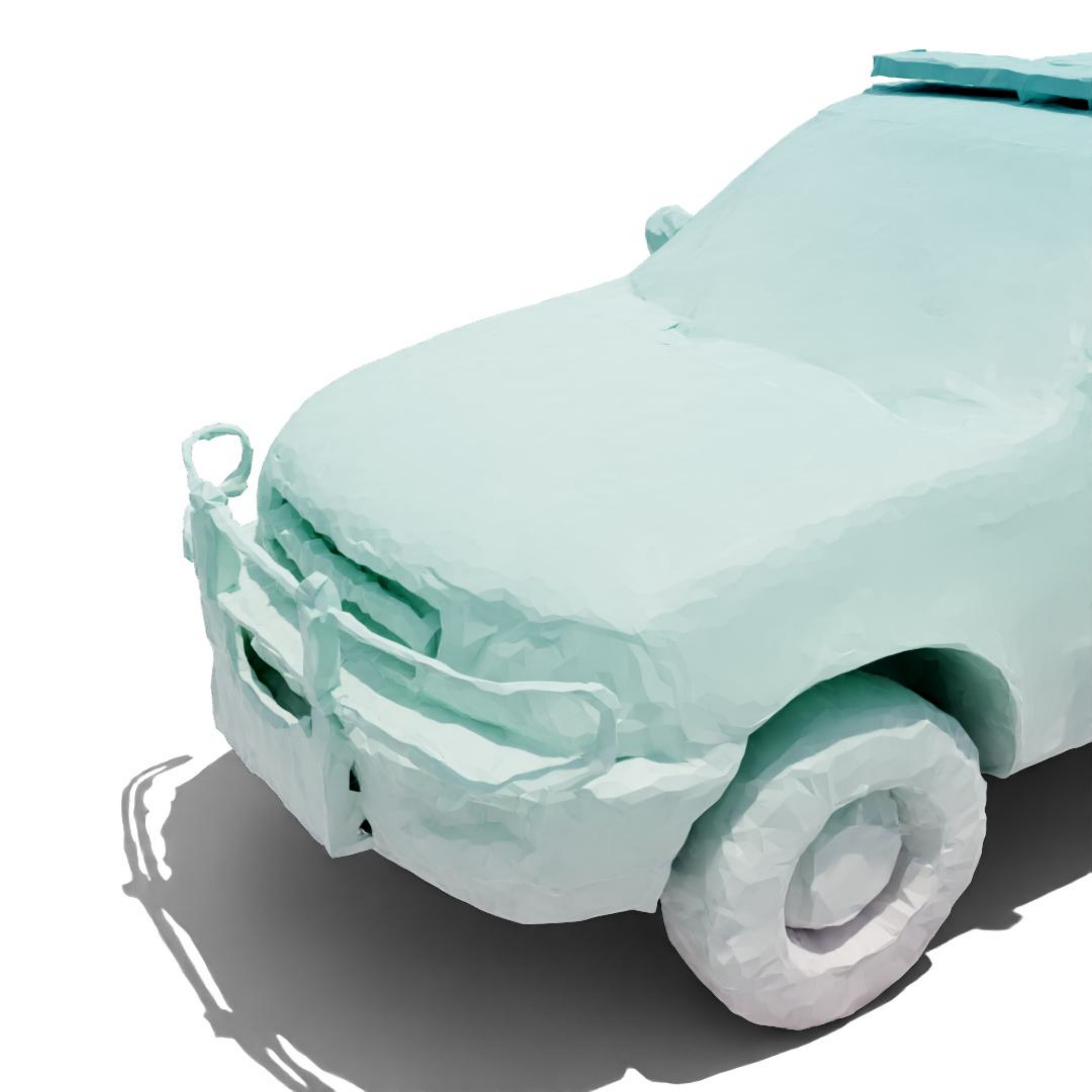}
\hspace{0.01\linewidth}
\includegraphics[width=0.30\linewidth]
{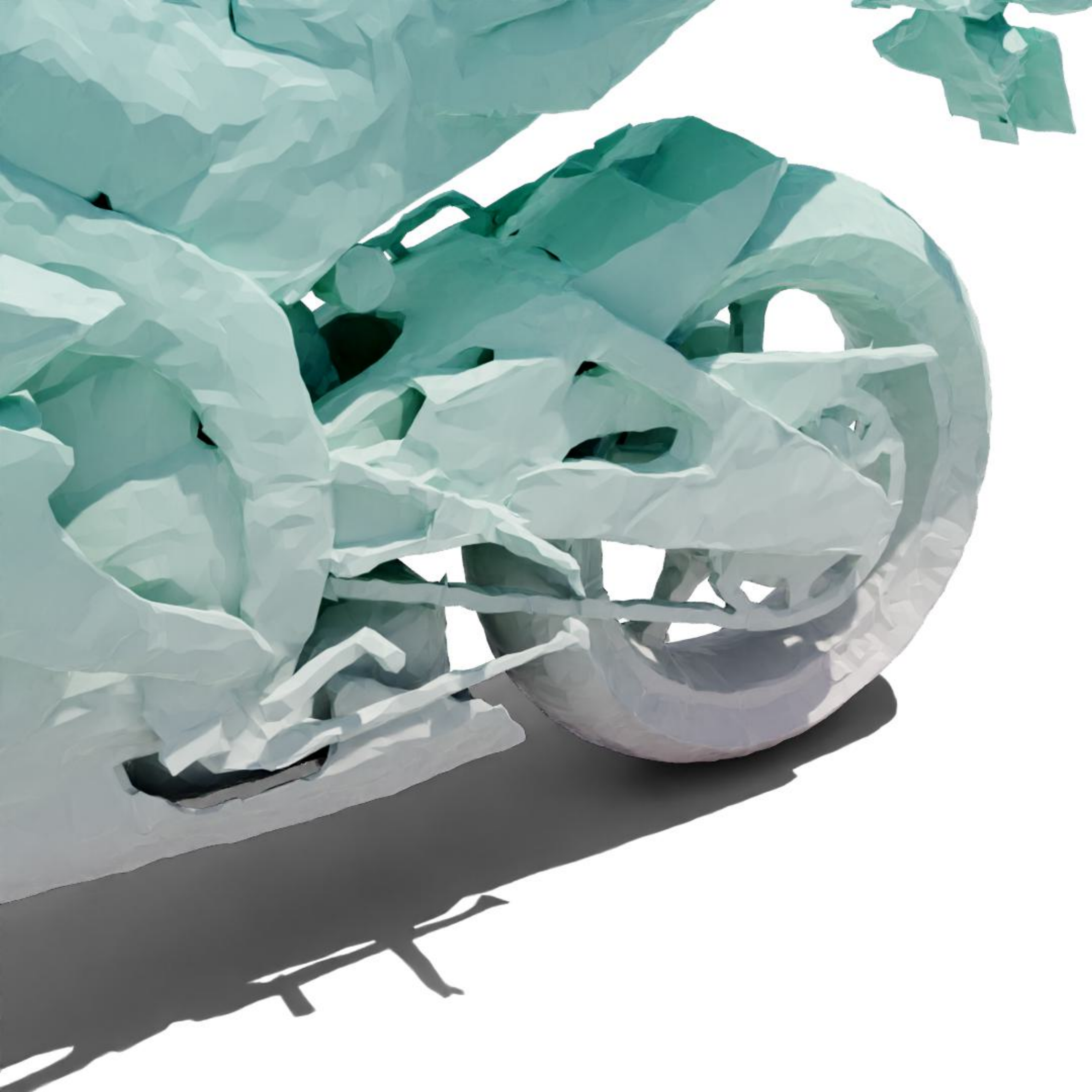}
\hspace{0.01\linewidth}
\includegraphics[width=0.30\linewidth]{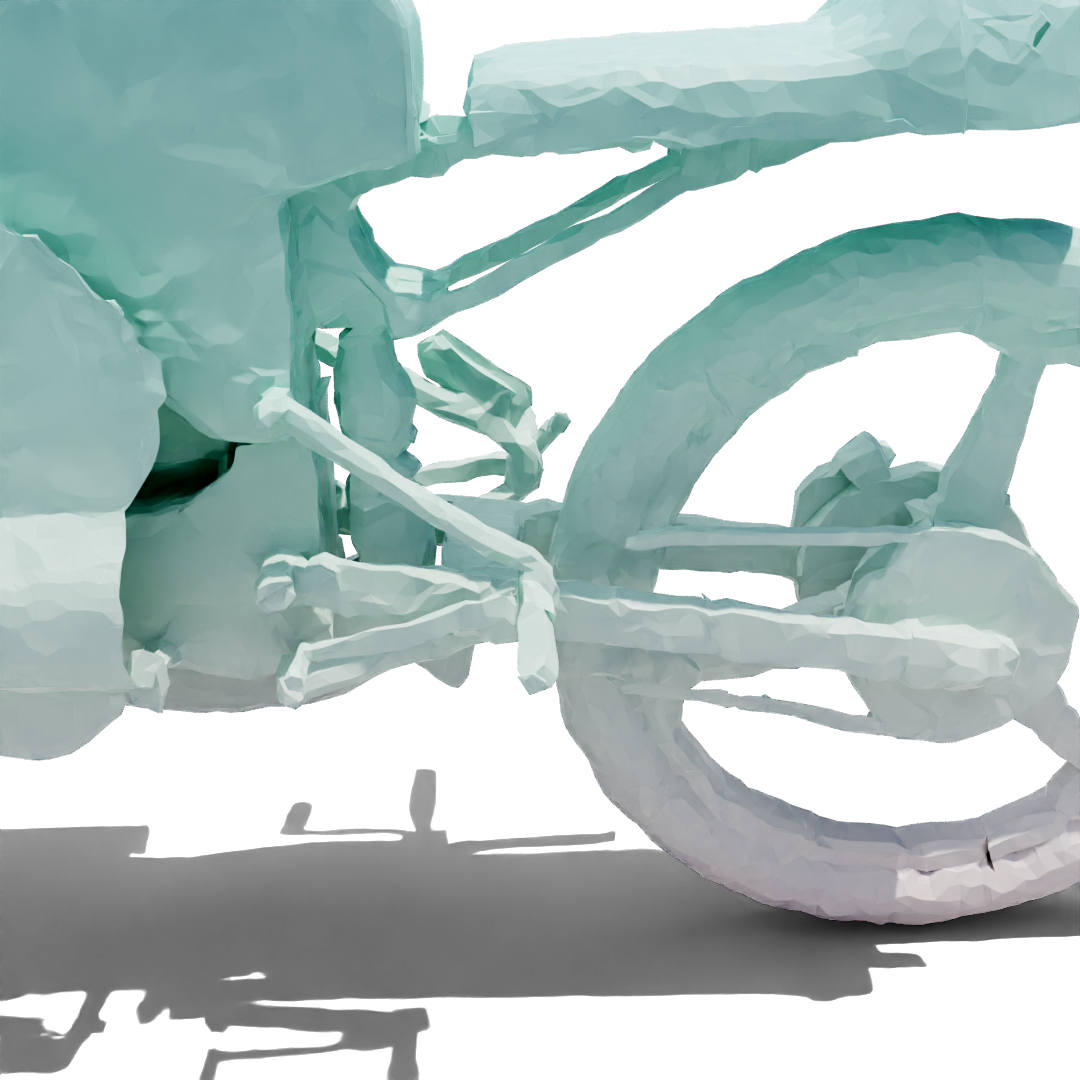}\\
\vspace{0.02\linewidth}
\includegraphics[width=0.30\linewidth]
{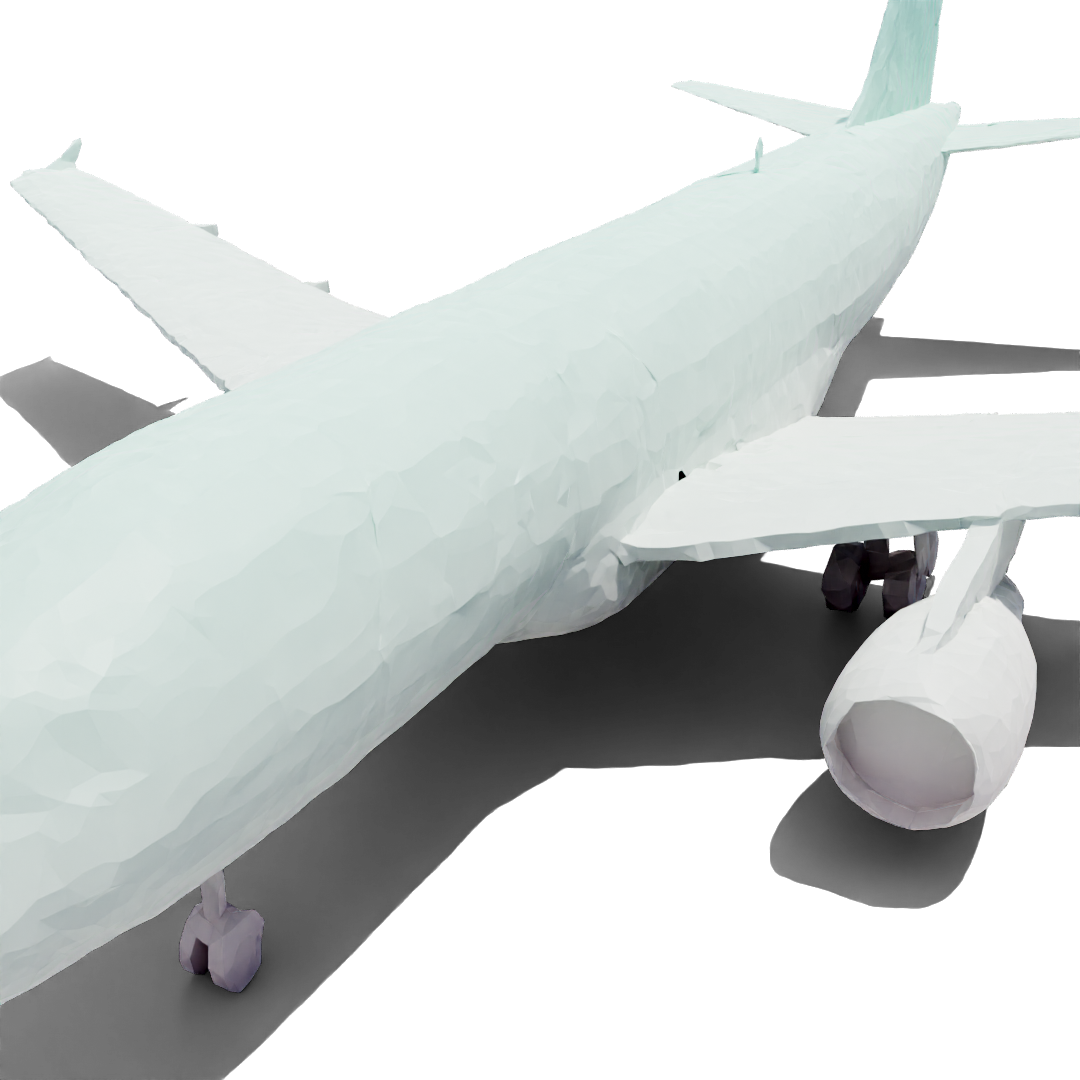}
\hspace{0.01\linewidth}
\includegraphics[width=0.30\linewidth, trim={9cm 5cm 1cm 5cm}, clip]
{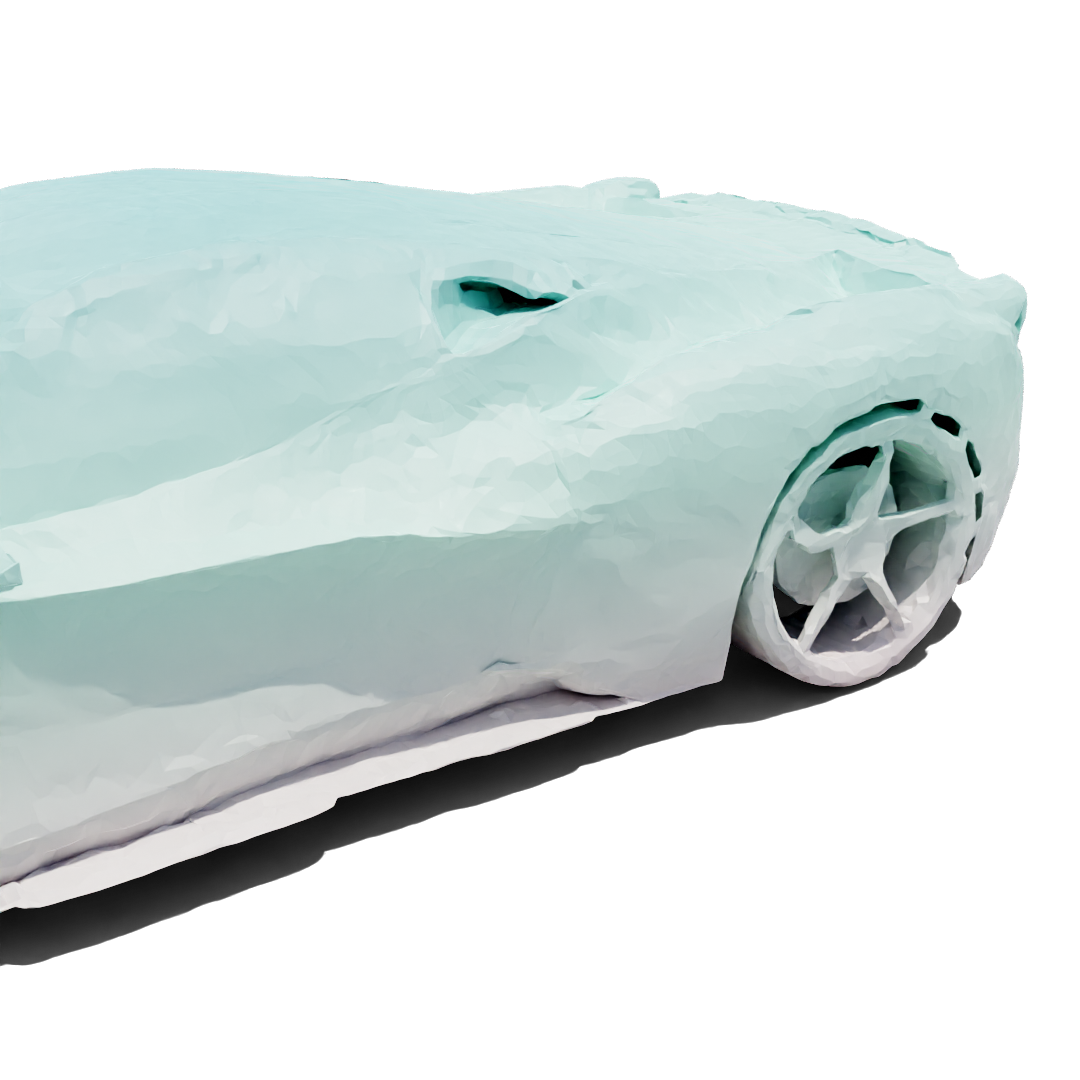}
\hspace{0.01\linewidth}
\includegraphics[width=0.30\linewidth]
{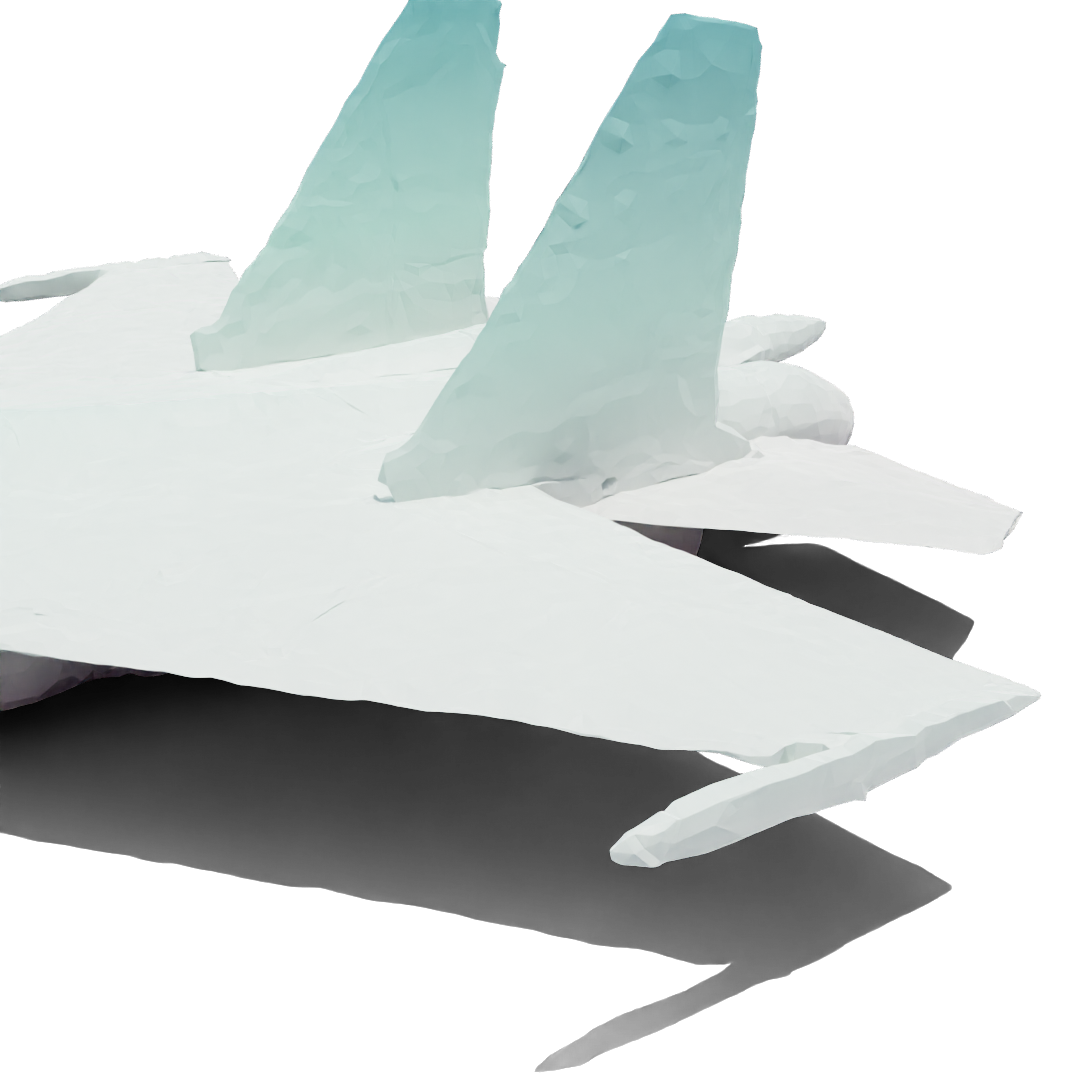}
\caption{\textbf{Details of generated meshes in high resolution.} \emph{TetraDiffusion} is able to output highly realistic shapes in a matter of seconds.}\label{fig:zoomedin}
\end{figure*}

Our efficient architecture is capable of generating meshes in unprecedented resolution, capturing intricate details such as chains, fine rims, brake disks and brake handles. Importantly, these details are represented in the geometry of the mesh and not as texture maps. Diffusing per-vertex texture information enforces the strong geometry. This capability is demonstrated in \Cref{fig:zoomedin}, where we showcase the model's skill in reproducing realistic and fine-grained elements, contributing to the overall realism and quality of the generated shapes.

\begin{figure*}[!ht]
\centering
\includegraphics[trim={1cm 5cm 1cm 5cm}, clip,width=0.48\linewidth]{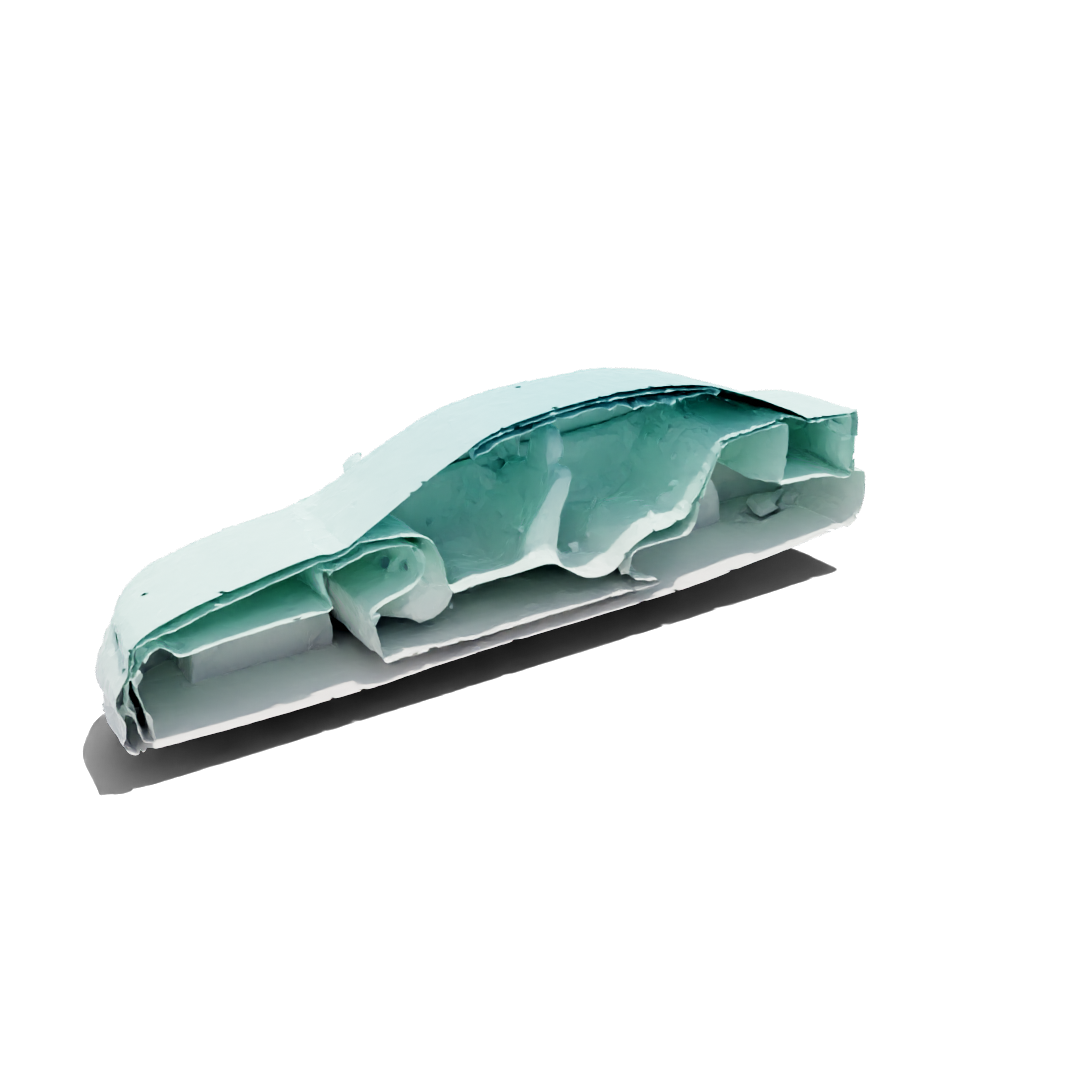}
\includegraphics[trim={1cm 5cm 1cm 5cm}, clip,width=0.48\linewidth]{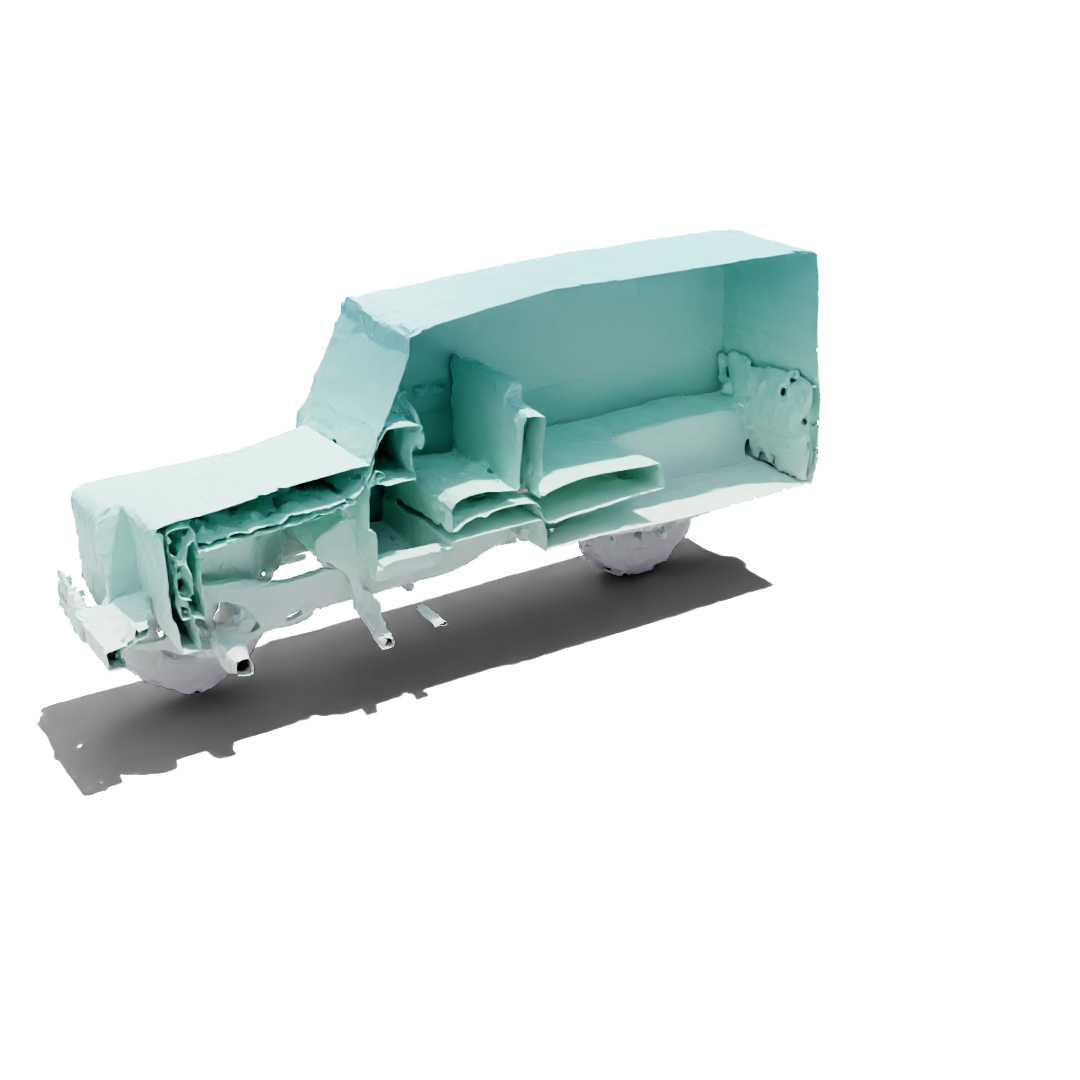}
\includegraphics[trim={1cm 5cm 1cm 5cm}, clip,width=0.48\linewidth]{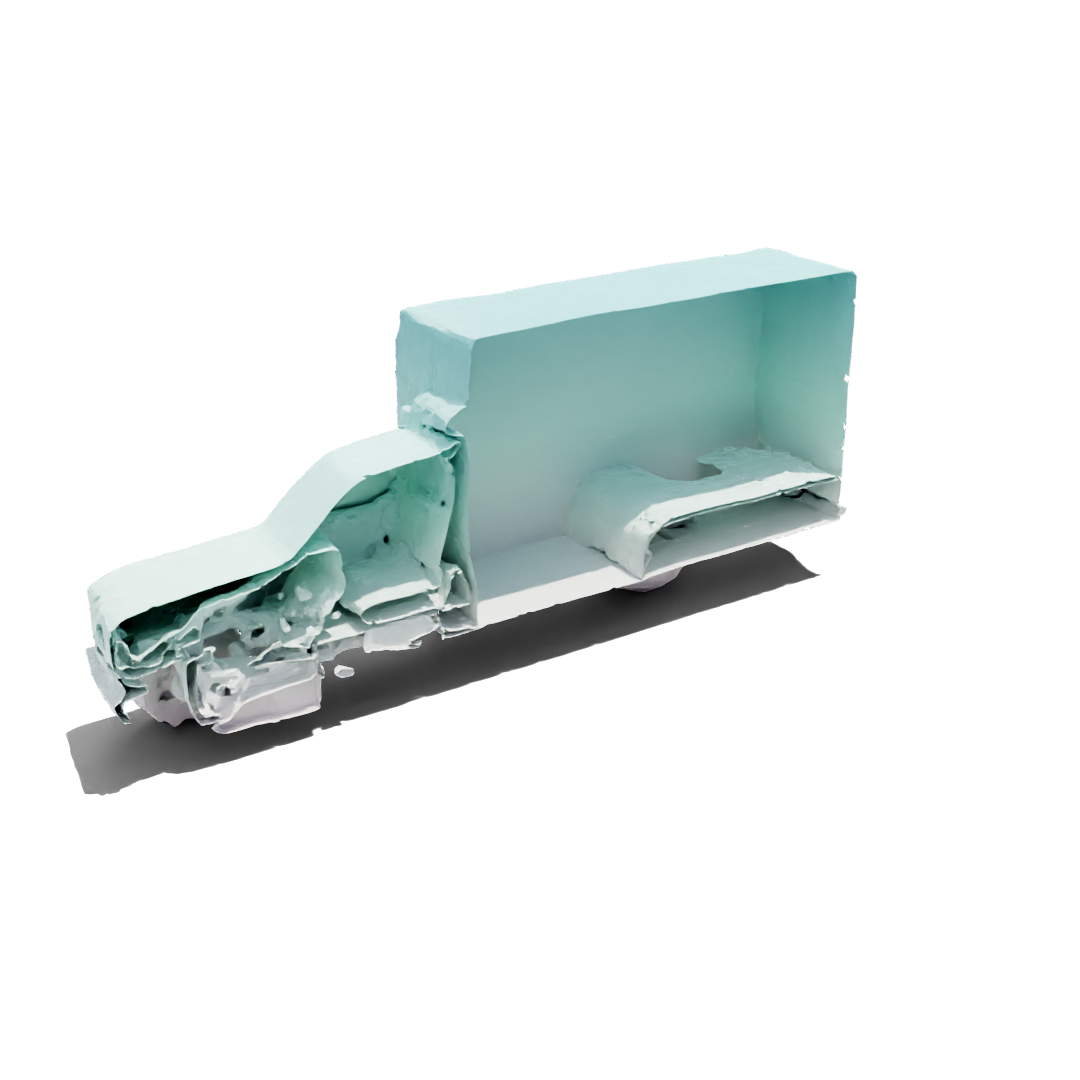}
\includegraphics[trim={1cm 5cm 1cm 5cm}, clip,width=0.48\linewidth]{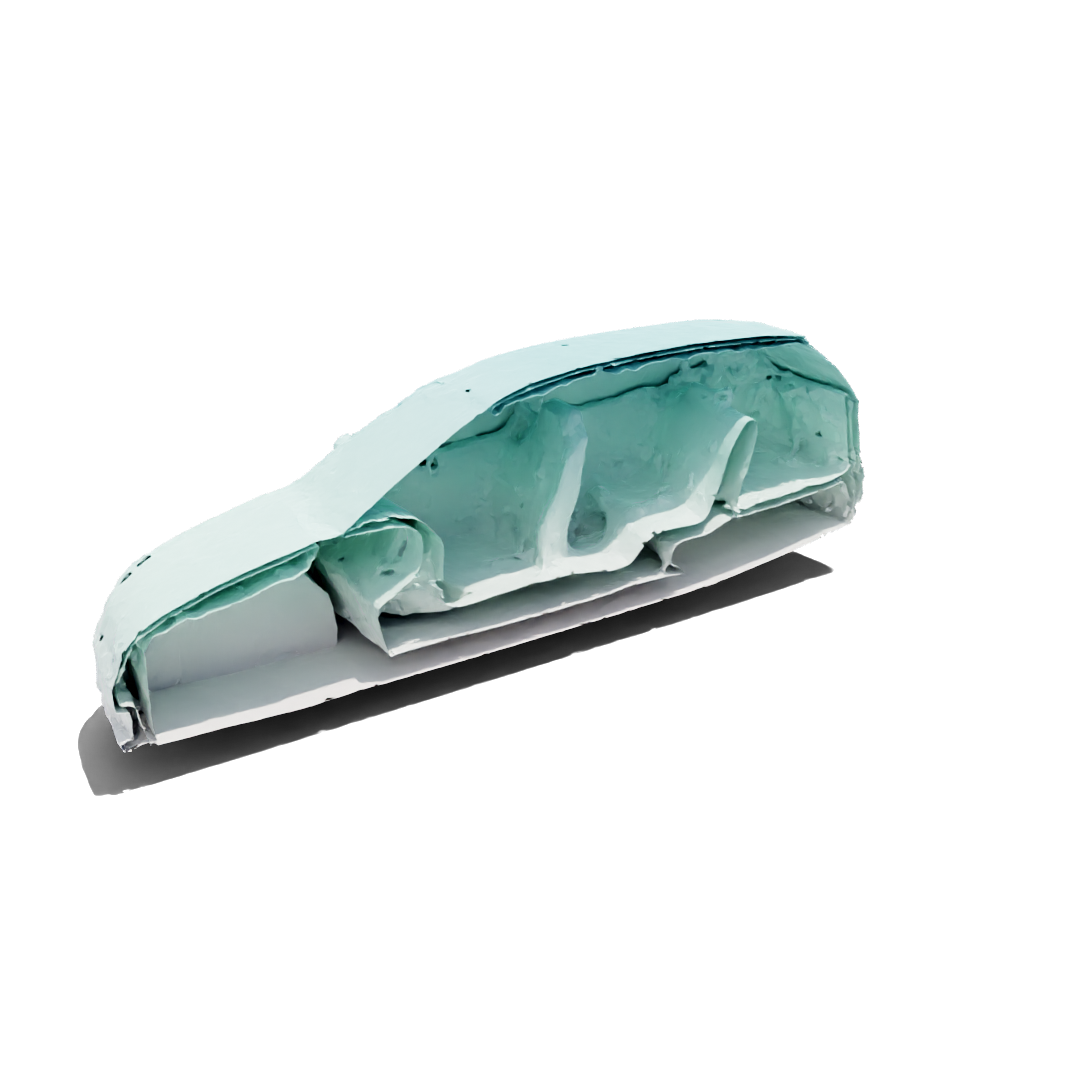}
\caption{\textbf{Generated inner structure.} Our ground truth preprocessing retains the inner structures of the original meshes, and \emph{TetraDiffusion} is able to reproduce such structures.}\label{fig:innerstructure}
\end{figure*}

Additionally, our ground truth contains inner structure, as explained in \cref{sec:dataprocessing}. The tetrahedral model is able to generate consistent inner structure, \eg seats, steering wheels or engines. Examples are displayed in \Cref{fig:innerstructure}. 

\clearpage
\newpage

\subsection{Unconditional generation}

\begin{figure*}[!ht]
\centering
\vspace{-0.4cm}
\includegraphics[width=0.16666666666666666\linewidth, trim={0 0cm 0 3cm}, clip]{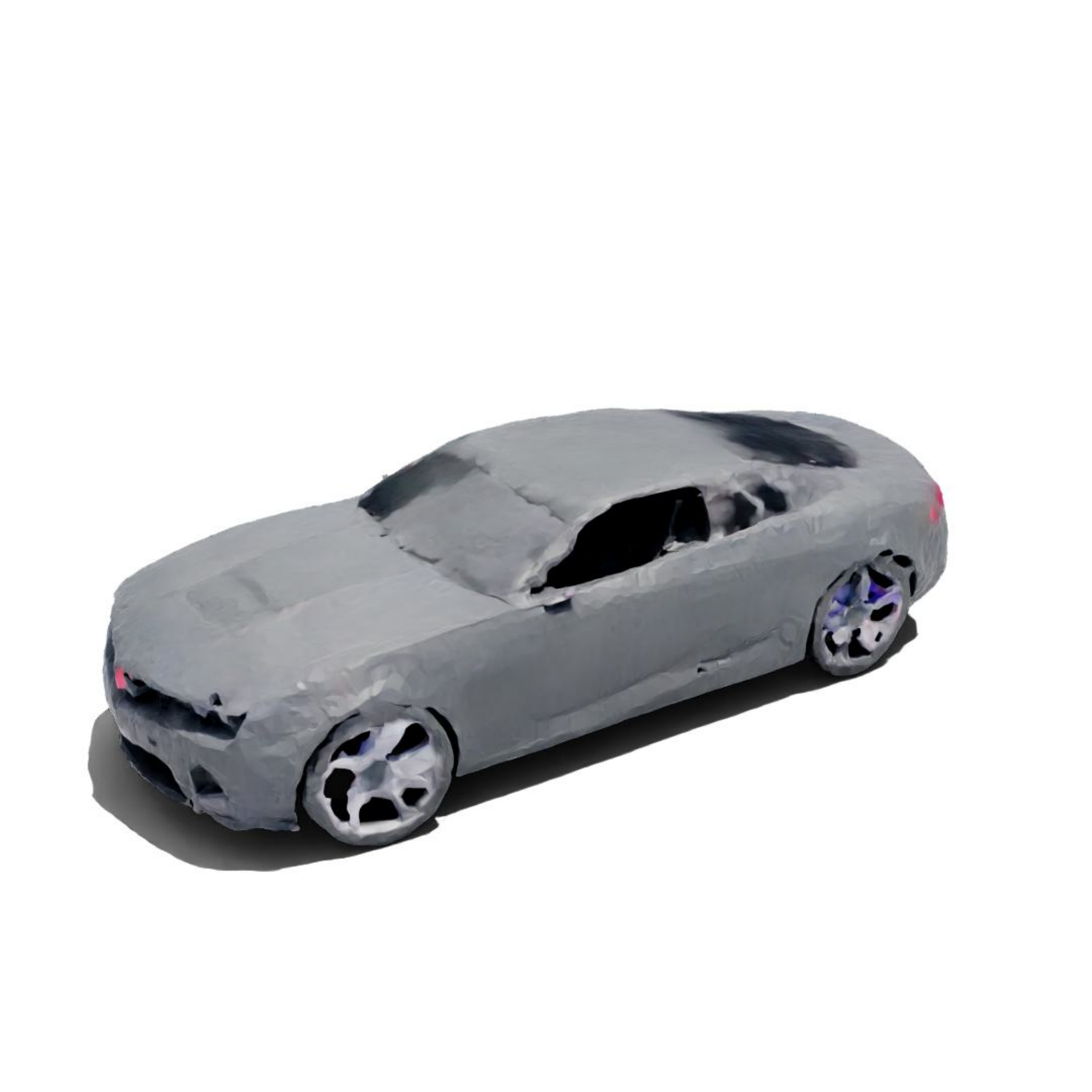}\includegraphics[width=0.16666666666666666\linewidth, trim={0 0cm 0 3cm}, clip]{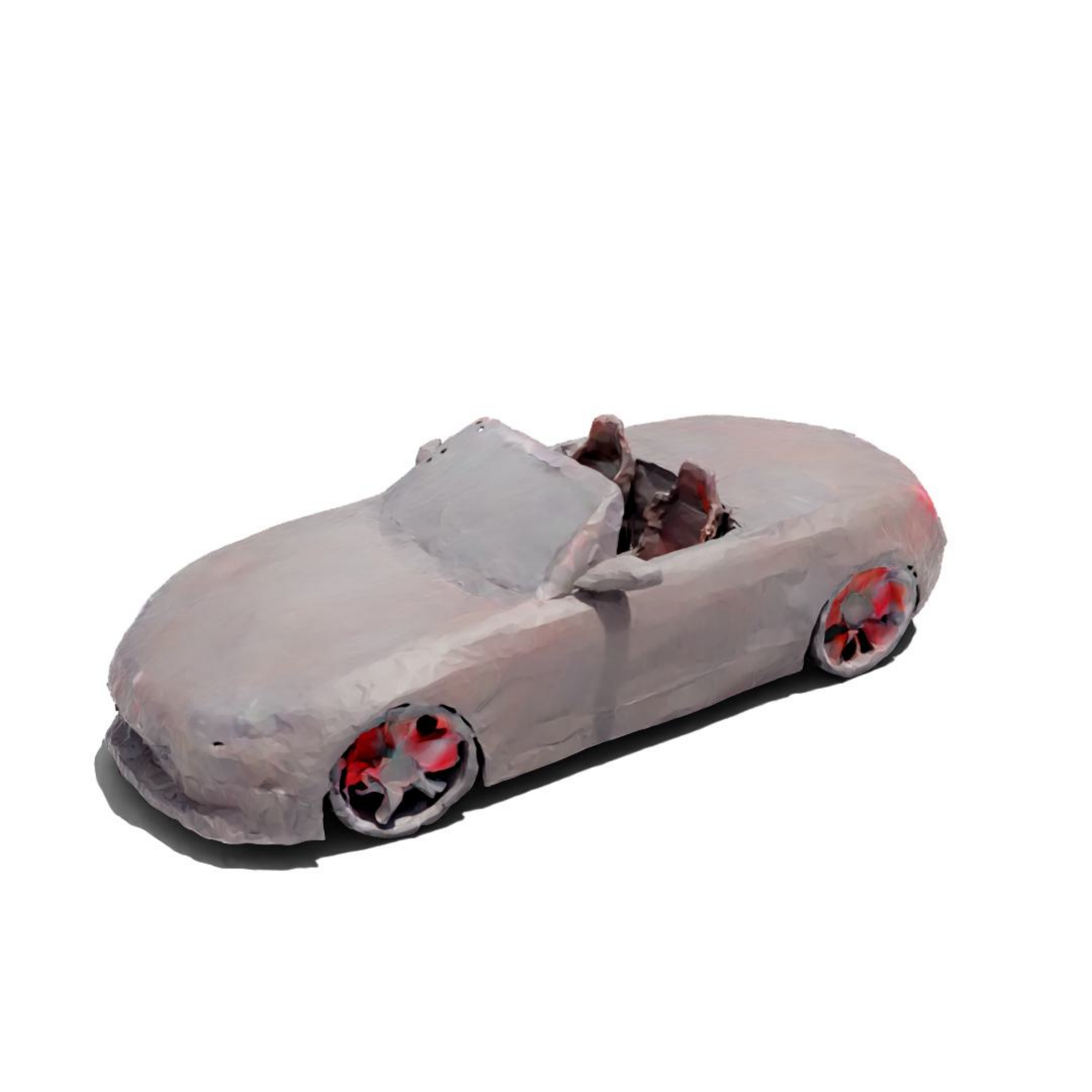}\includegraphics[width=0.16666666666666666\linewidth, trim={0 0cm 0 3cm}, clip]{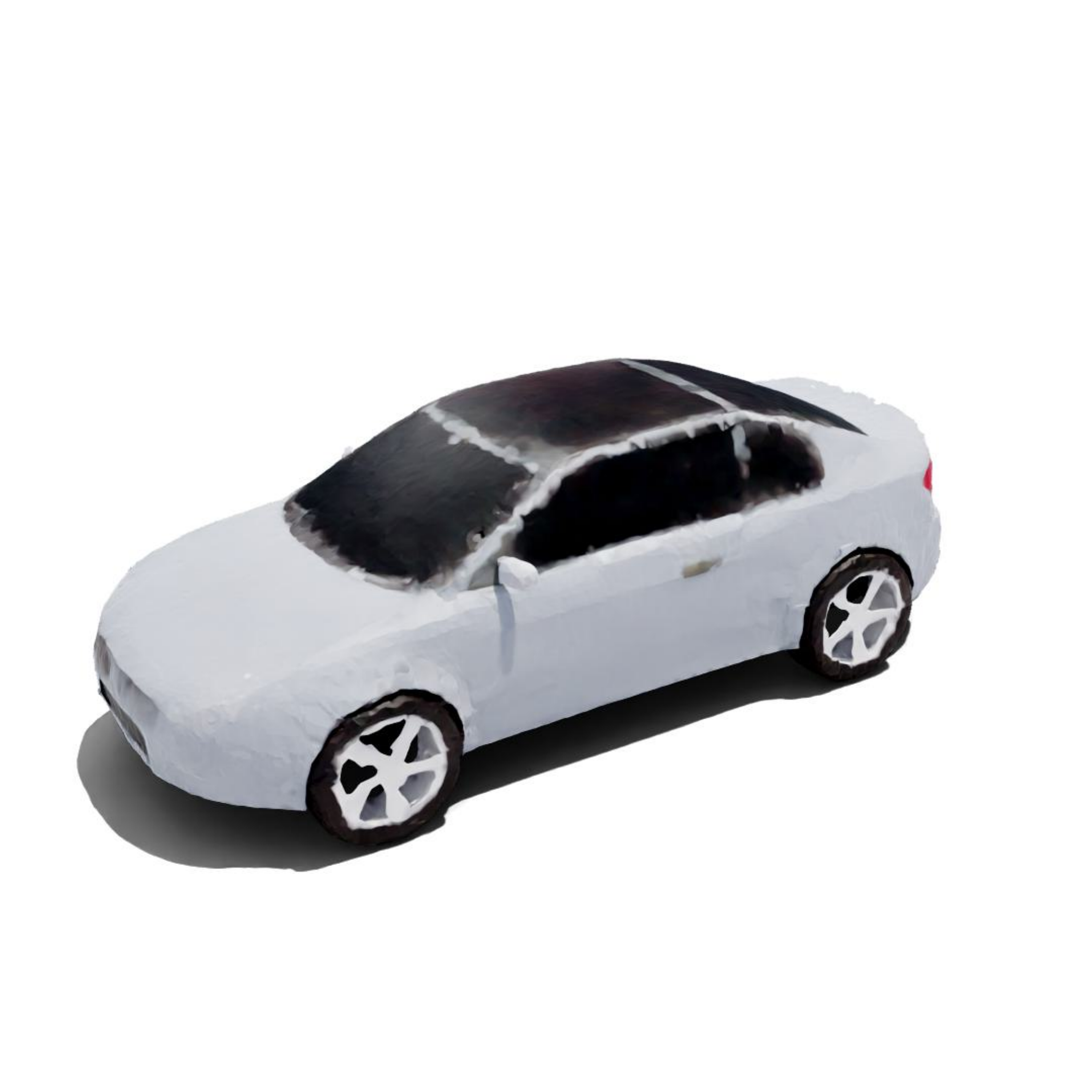}\includegraphics[width=0.16666666666666666\linewidth, trim={0 0cm 0 3cm}, clip]{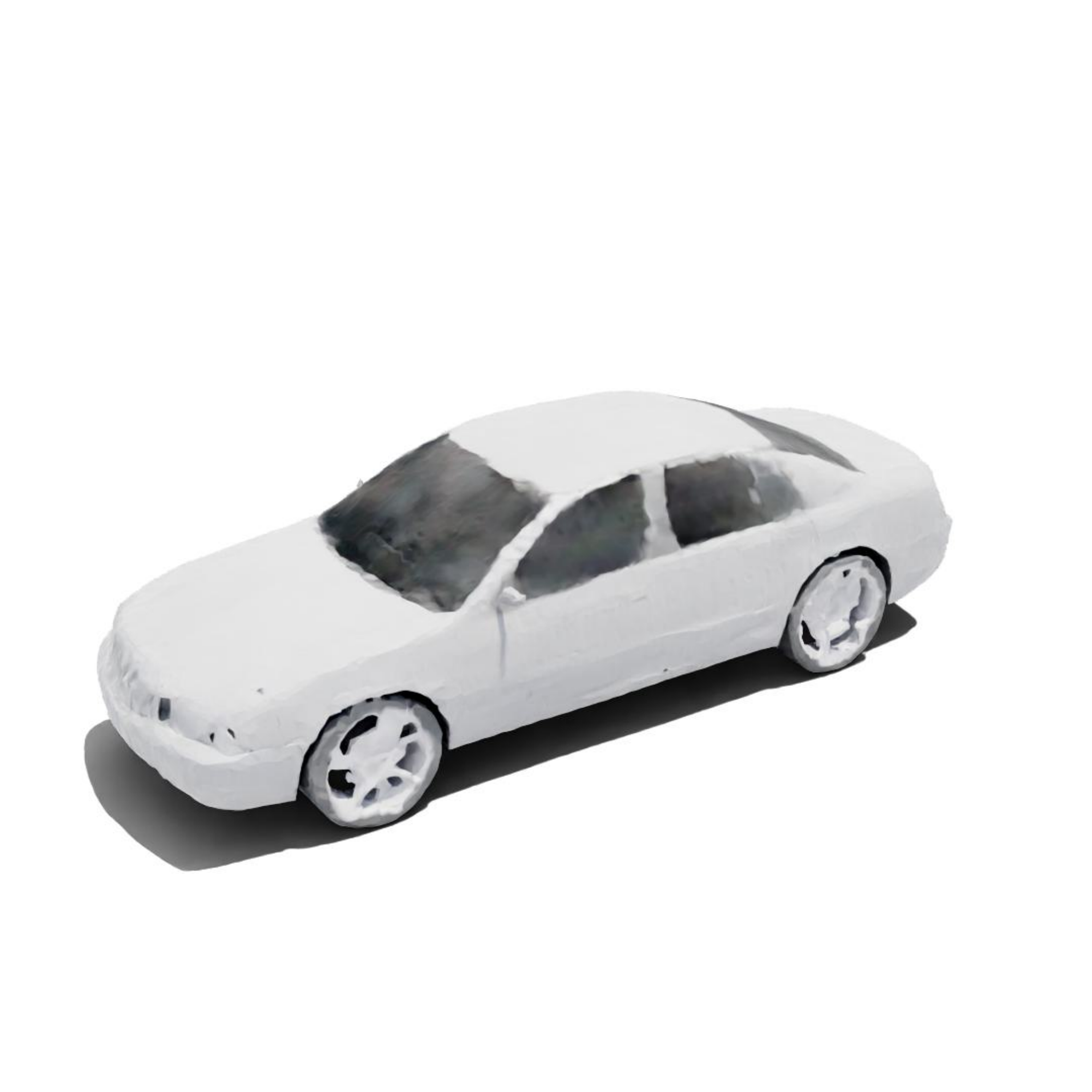}\includegraphics[width=0.16666666666666666\linewidth, trim={0 0cm 0 3cm}, clip]{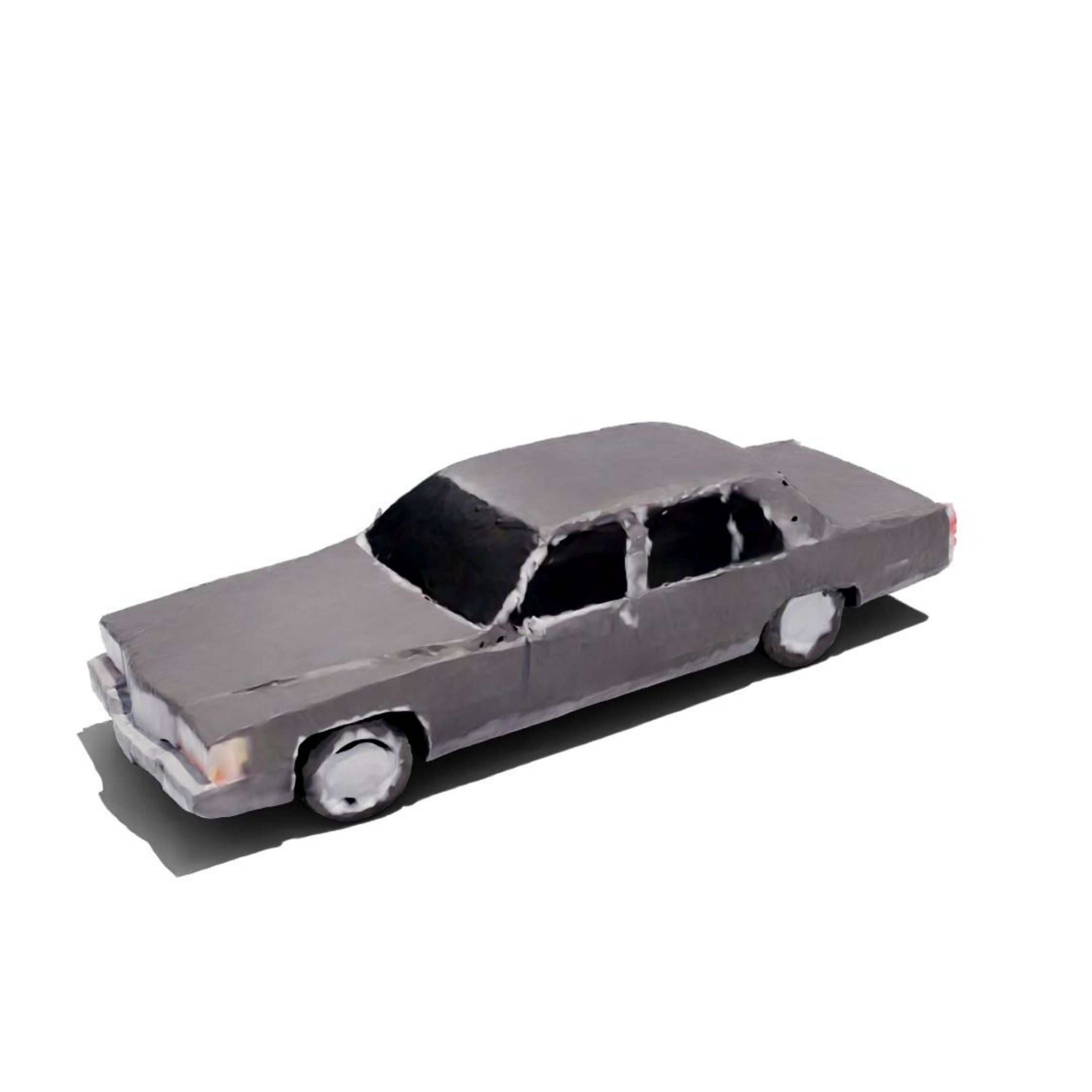}\includegraphics[width=0.16666666666666666\linewidth, trim={0 0cm 0 3cm}, clip]{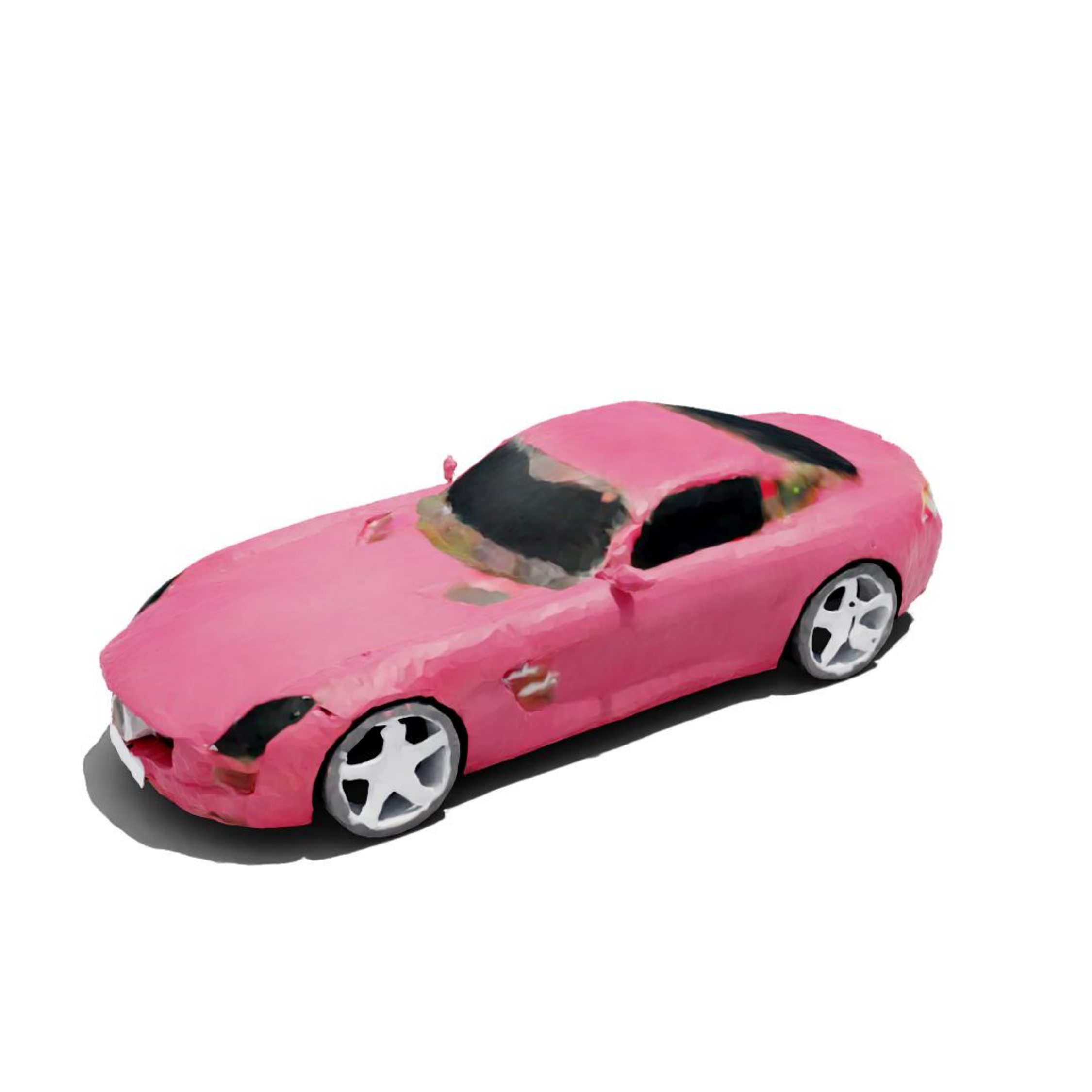}\\

\vspace{-0.43cm}
\includegraphics[width=0.16666666666666666\linewidth, trim={0 0cm 0 3cm}, clip]{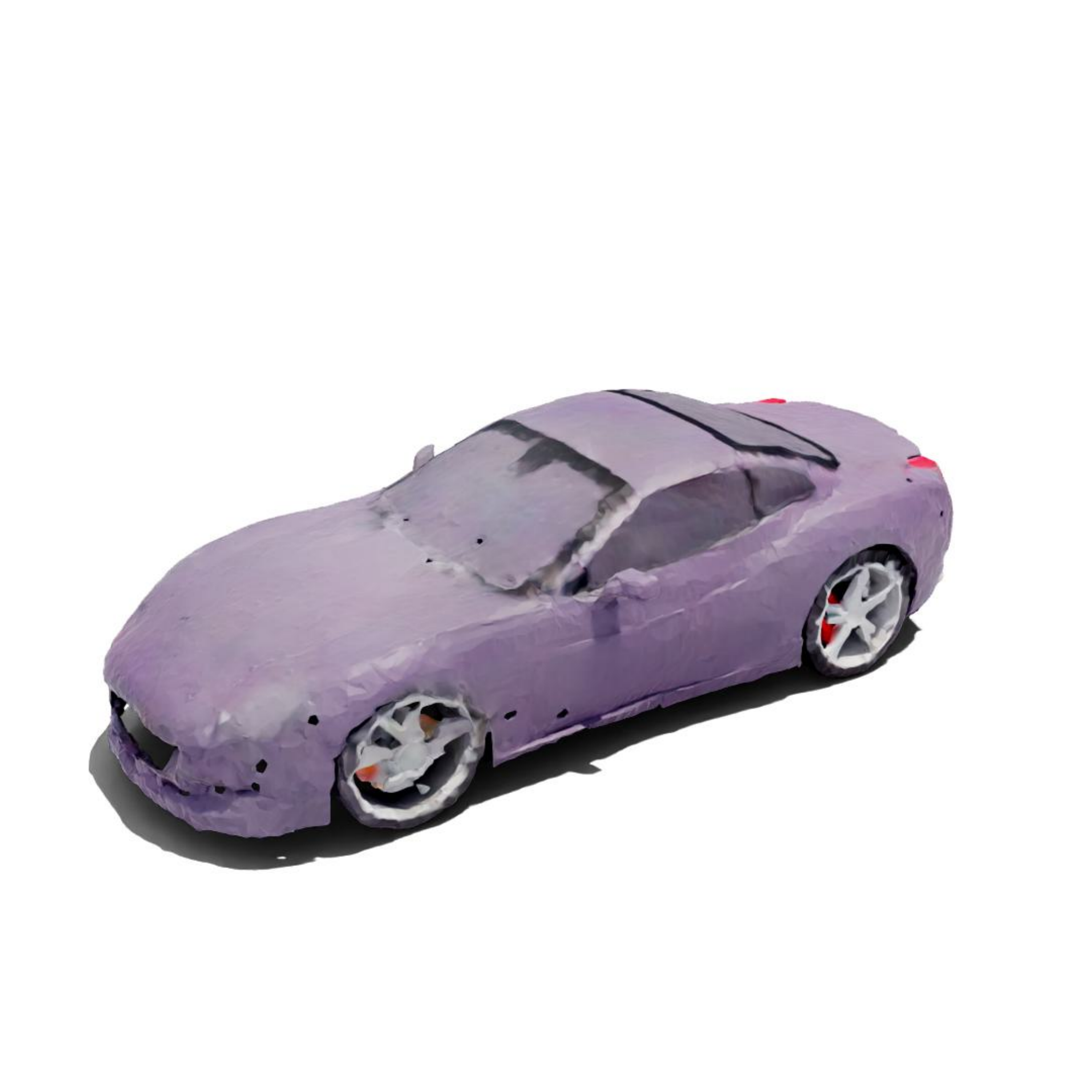}\includegraphics[width=0.16666666666666666\linewidth, trim={0 0cm 0 3cm}, clip]{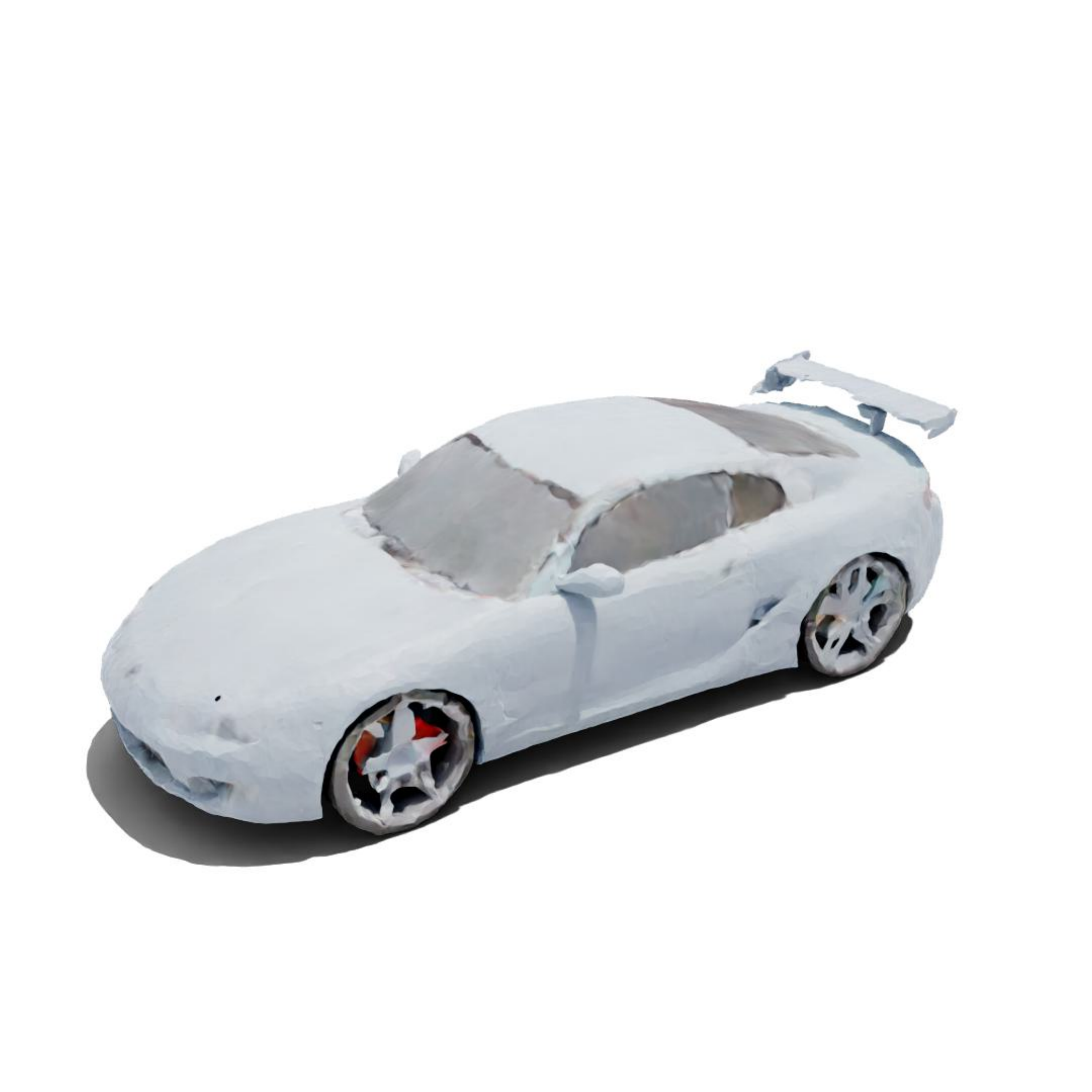}\includegraphics[width=0.16666666666666666\linewidth, trim={0 0cm 0 3cm}, clip]{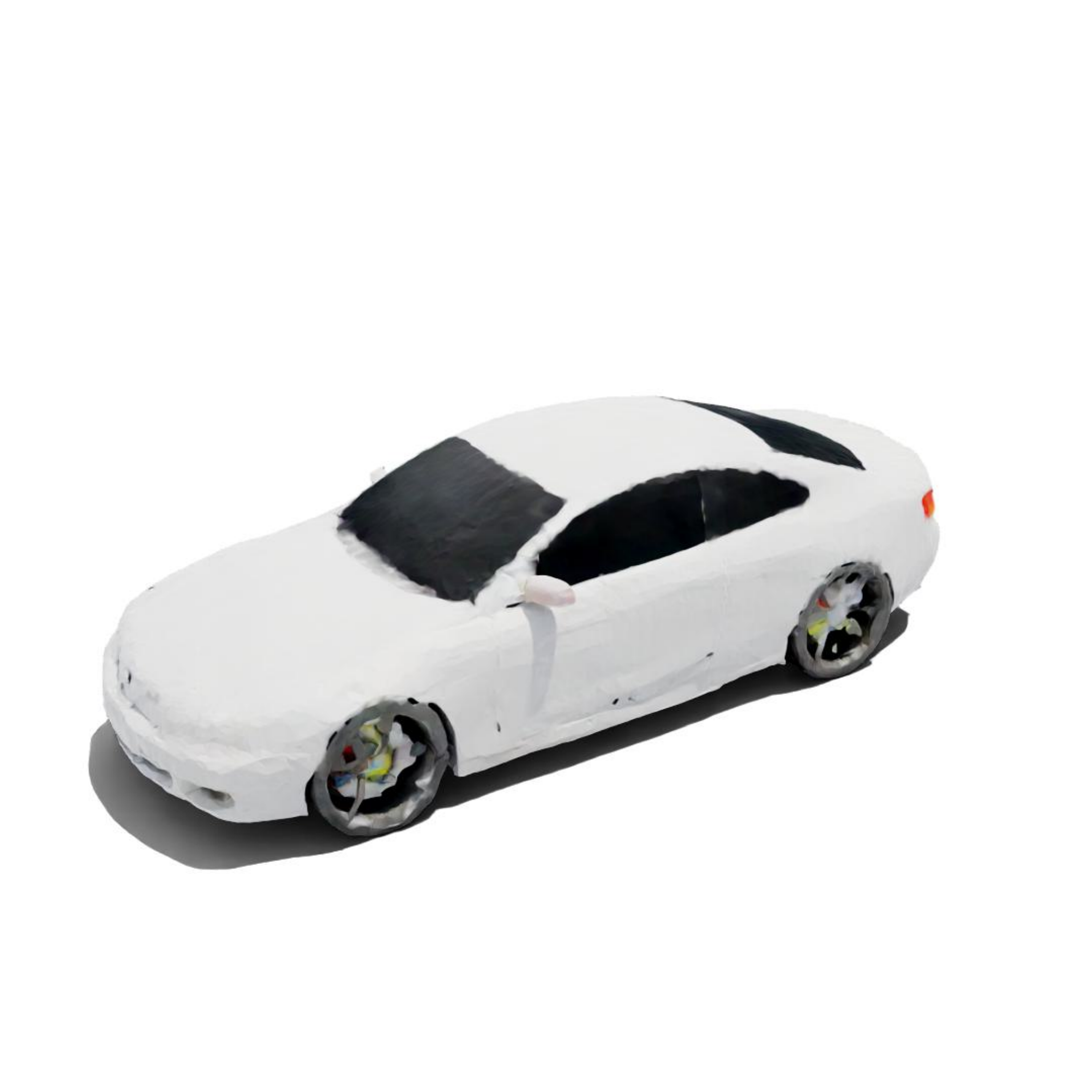}\includegraphics[width=0.16666666666666666\linewidth, trim={0 0cm 0 3cm}, clip]{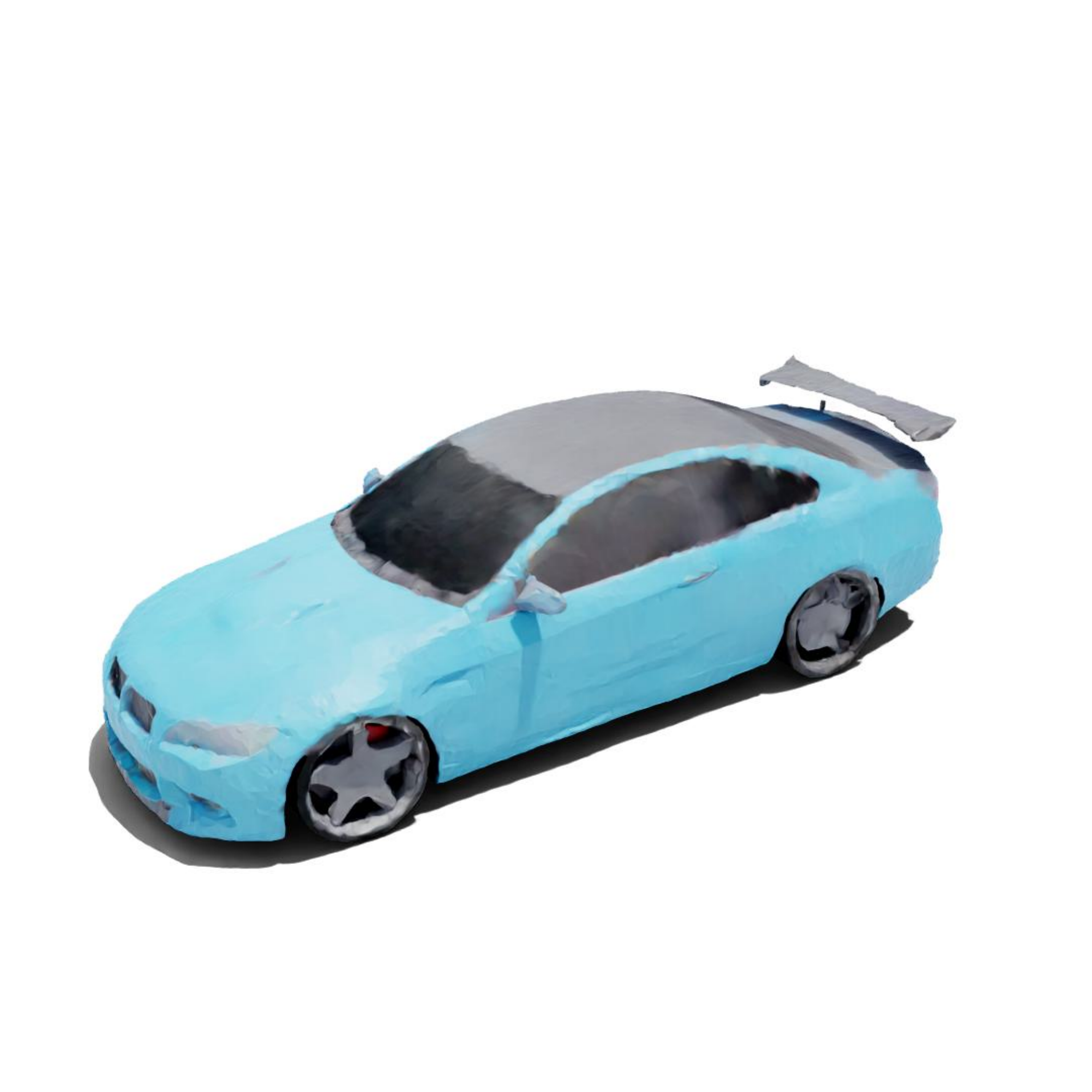}\includegraphics[width=0.16666666666666666\linewidth, trim={0 0cm 0 3cm}, clip]{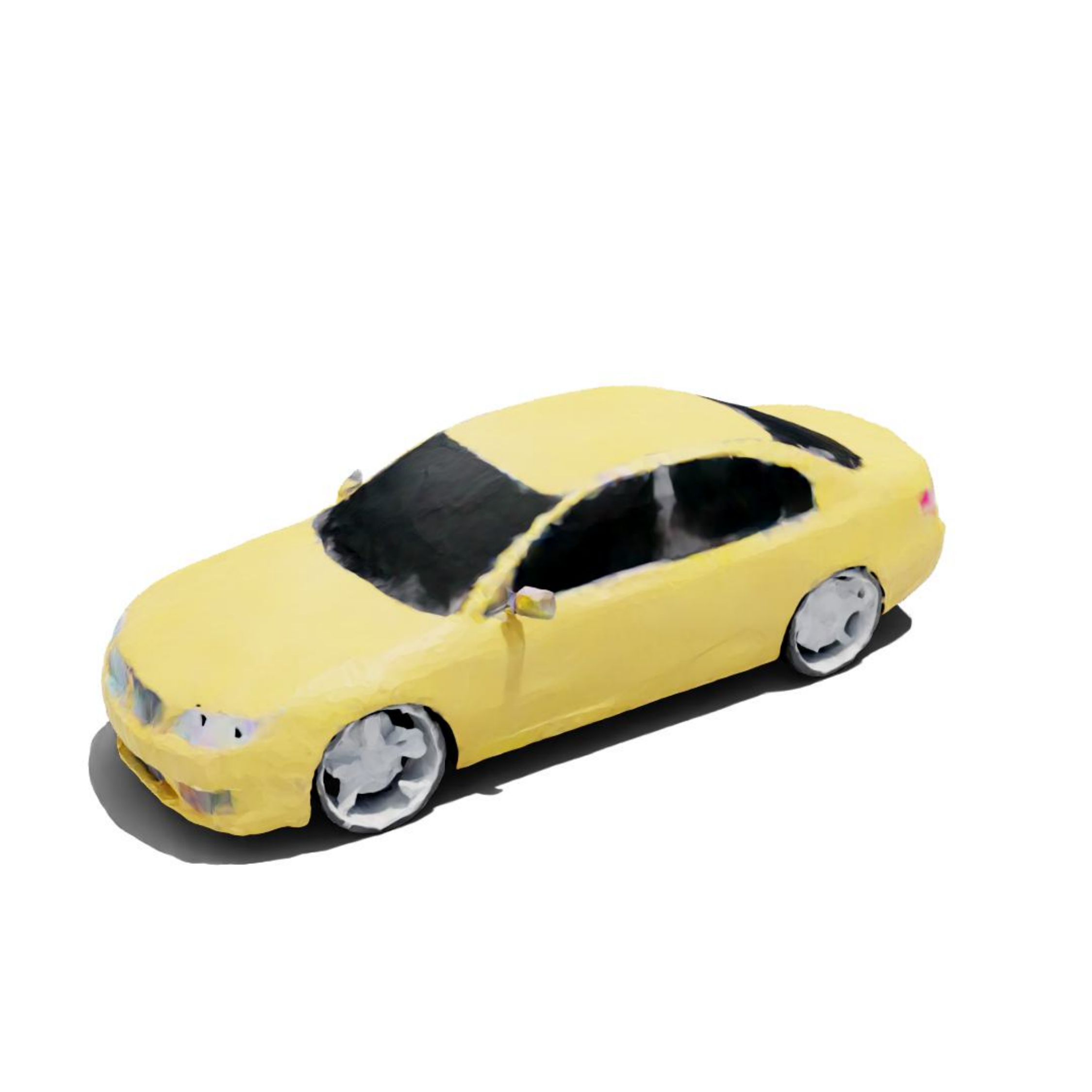}\includegraphics[width=0.16666666666666666\linewidth, trim={0 0cm 0 3cm}, clip]{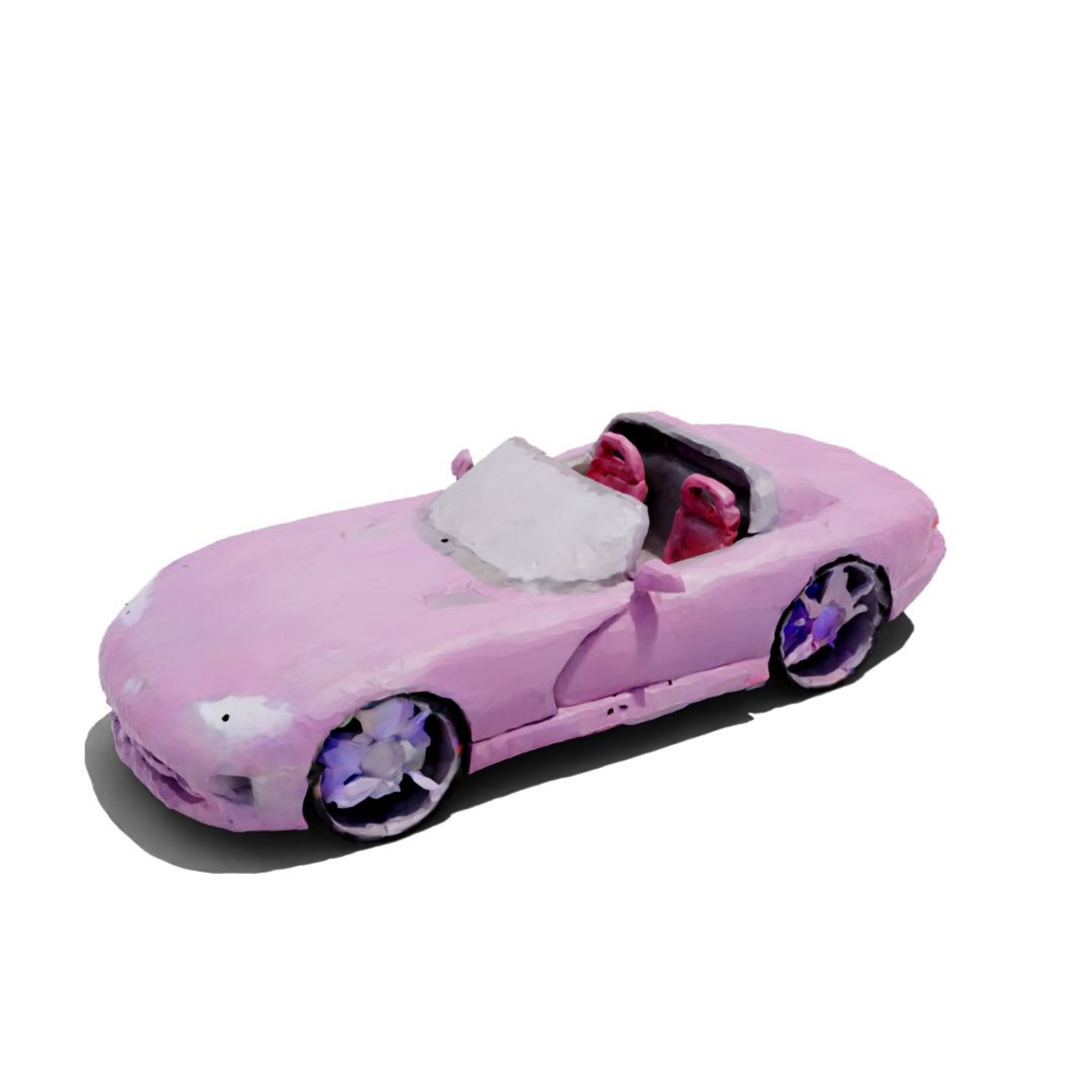}\\

\vspace{-0.43cm}
\includegraphics[width=0.16666666666666666\linewidth, trim={0 0cm 0 3cm}, clip]{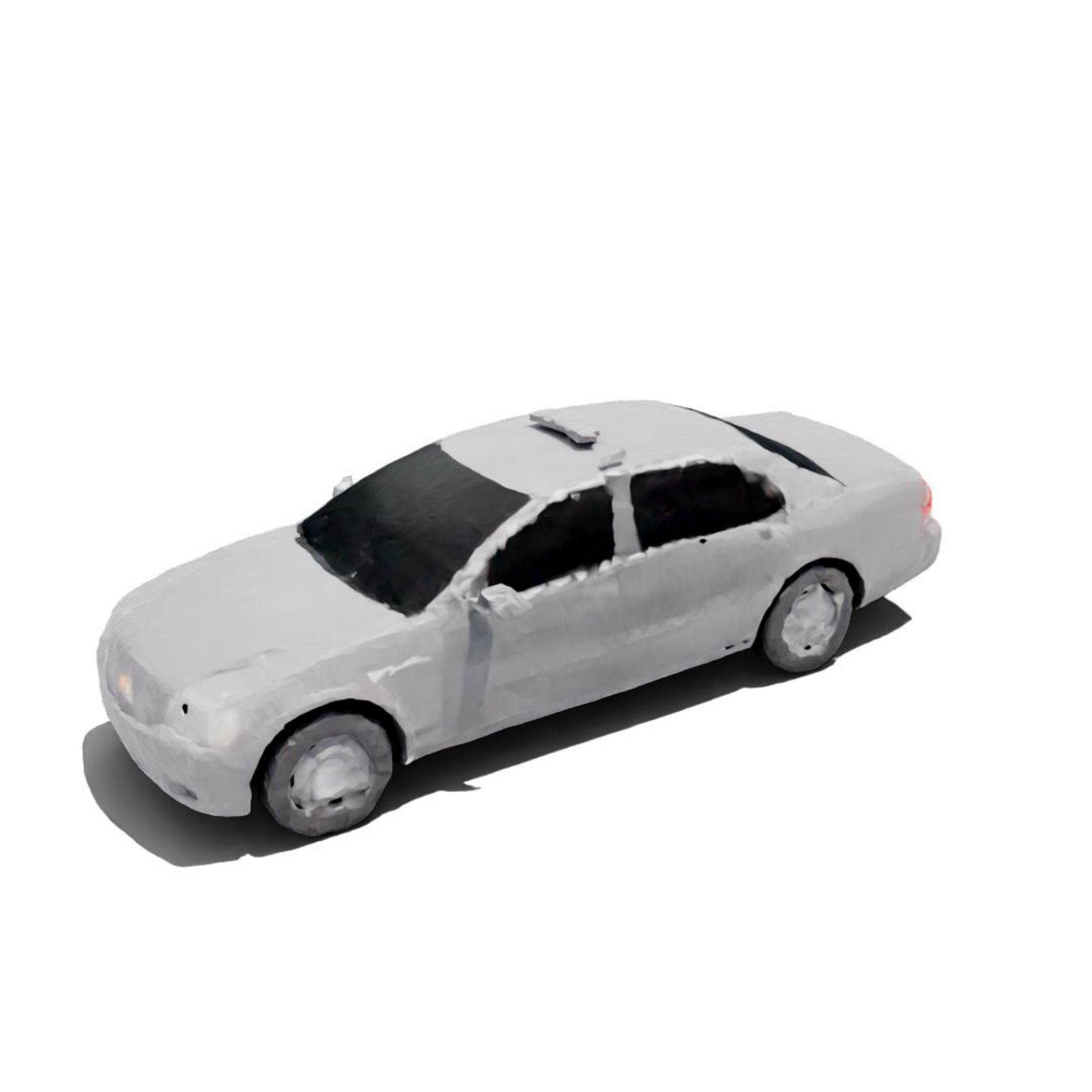}\includegraphics[width=0.16666666666666666\linewidth, trim={0 0cm 0 3cm}, clip]{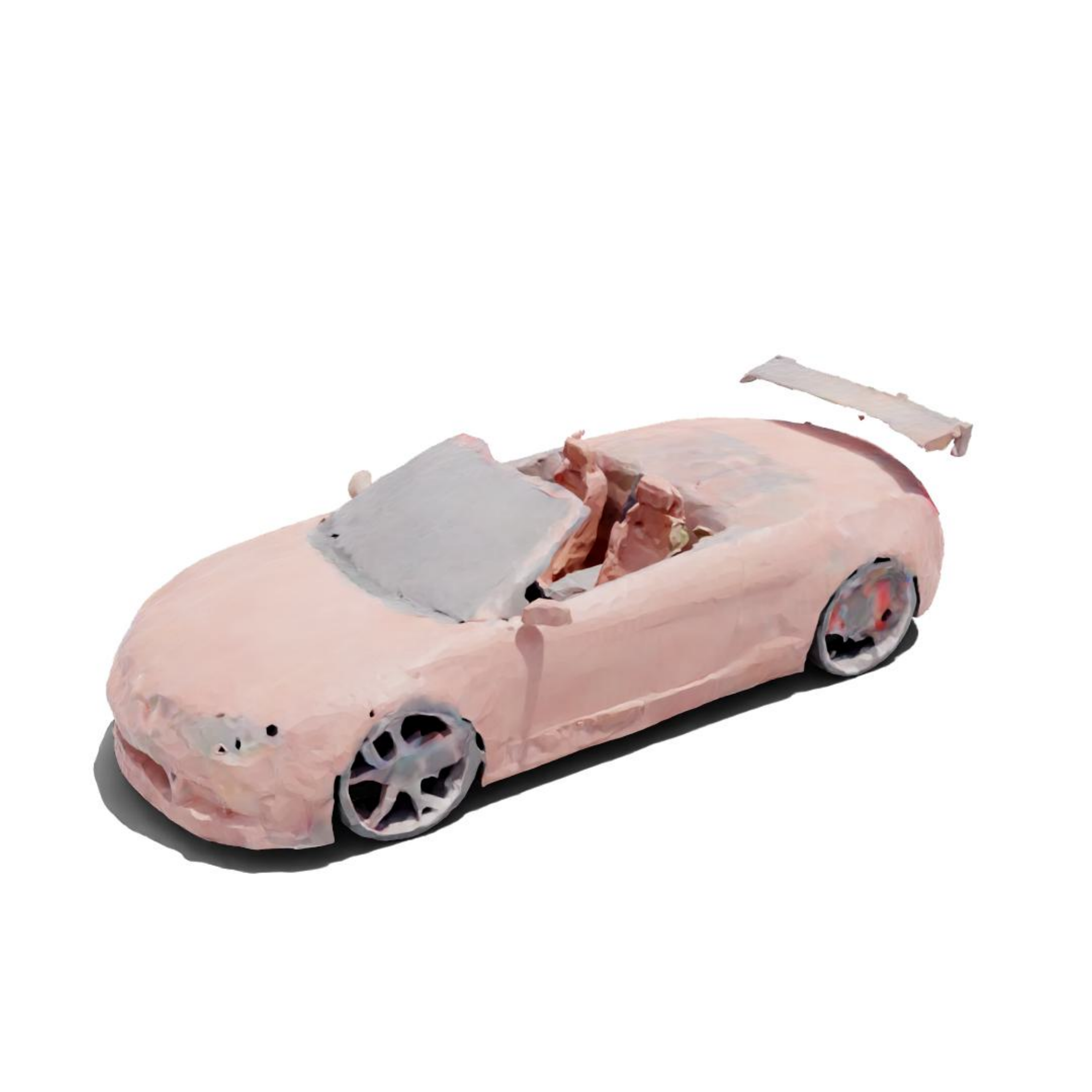}\includegraphics[width=0.16666666666666666\linewidth, trim={0 0cm 0 3cm}, clip]{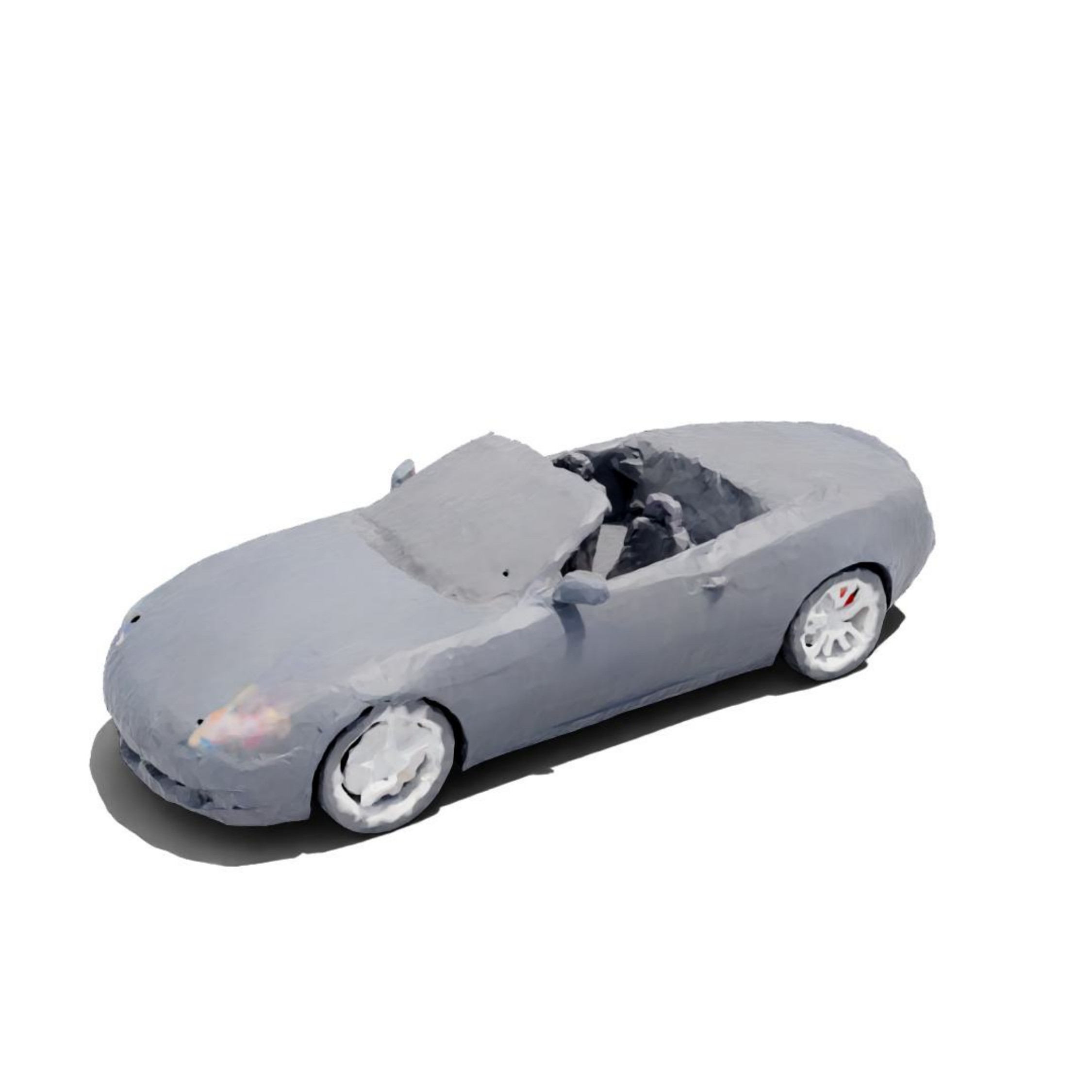}\includegraphics[width=0.16666666666666666\linewidth, trim={0 0cm 0 3cm}, clip]{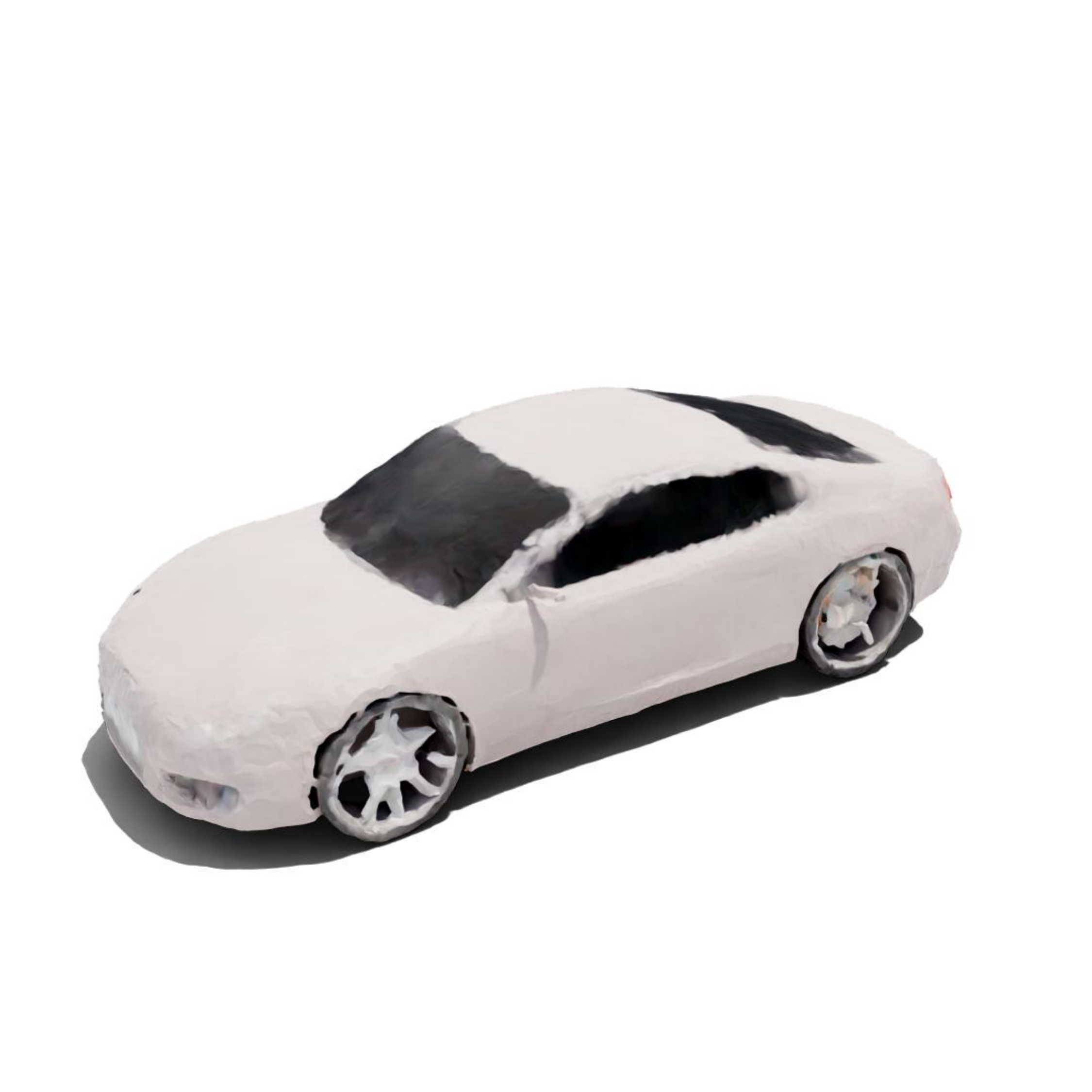}\includegraphics[width=0.16666666666666666\linewidth, trim={0 0cm 0 3cm}, clip]{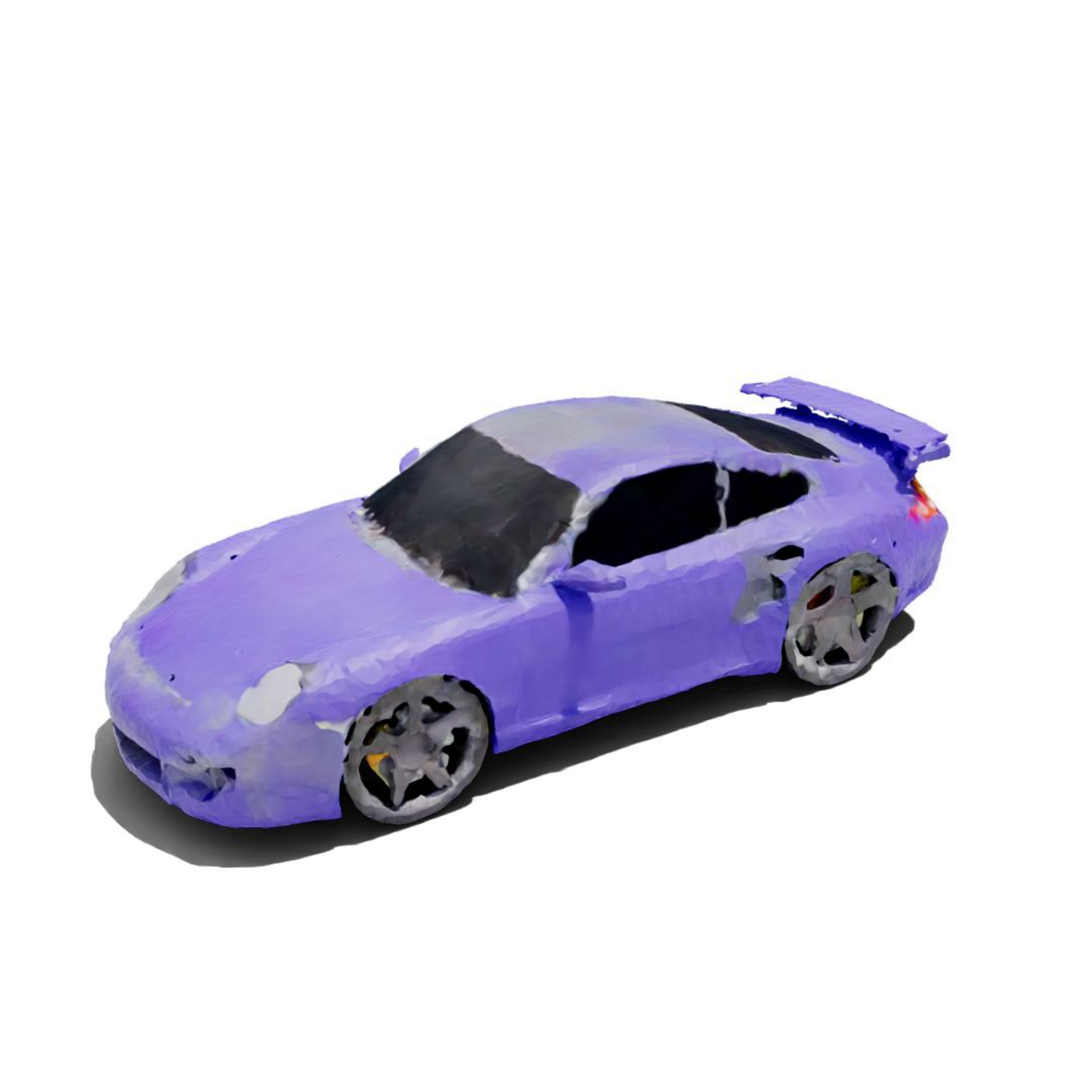}\includegraphics[width=0.16666666666666666\linewidth, trim={0 0cm 0 3cm}, clip]{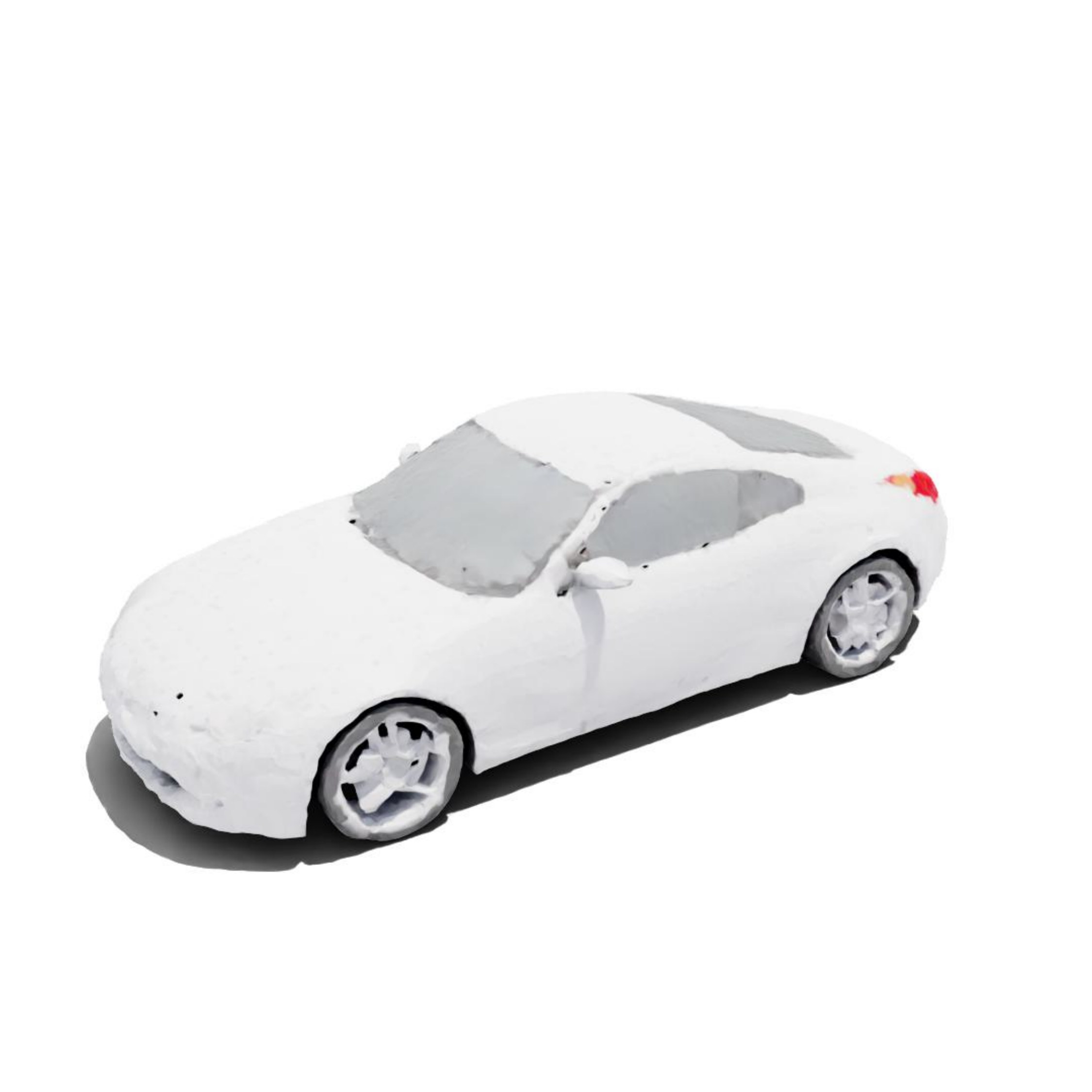}\\

\vspace{-0.43cm}
\includegraphics[width=0.16666666666666666\linewidth, trim={0 0cm 0 3cm}, clip]{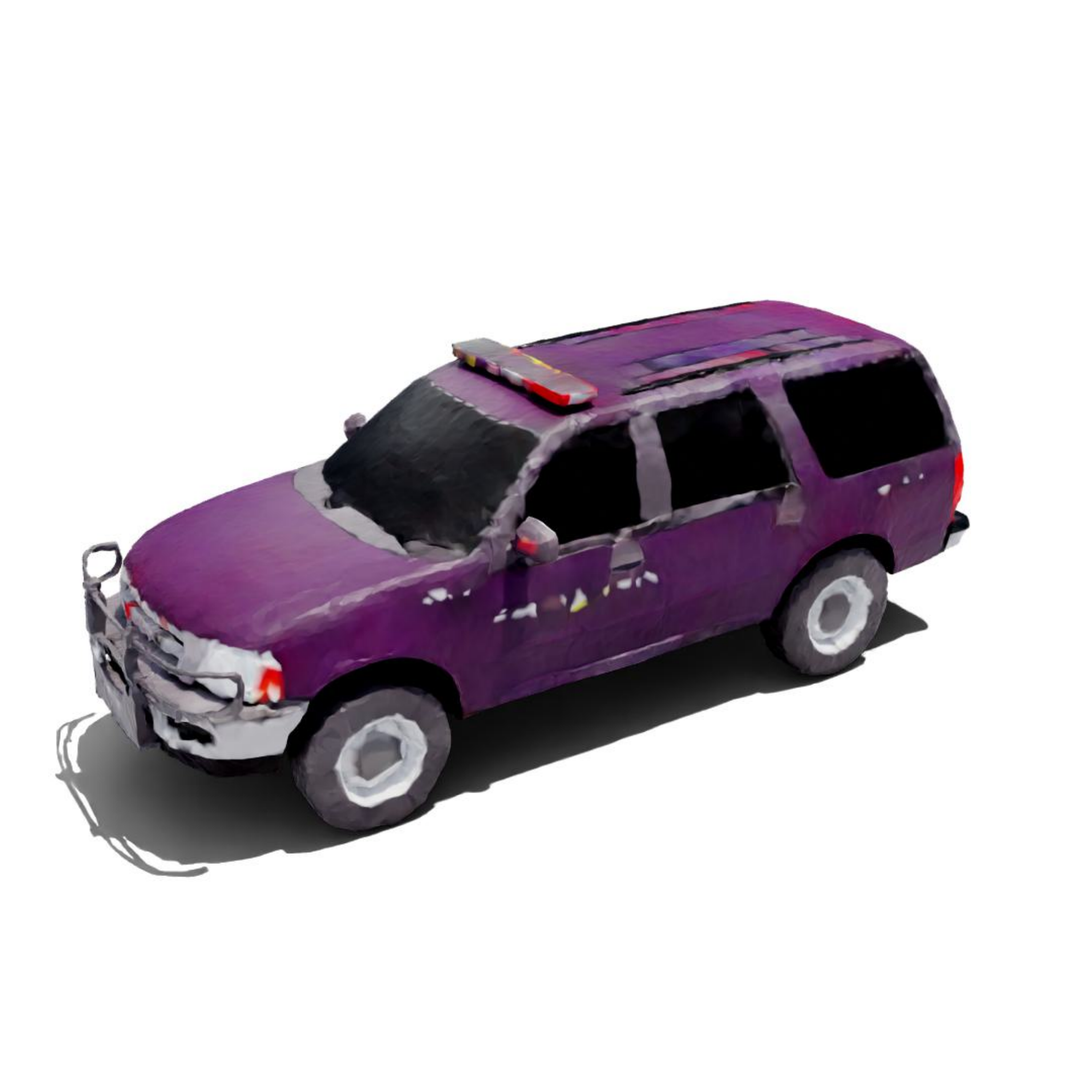}\includegraphics[width=0.16666666666666666\linewidth, trim={0 0cm 0 3cm}, clip]{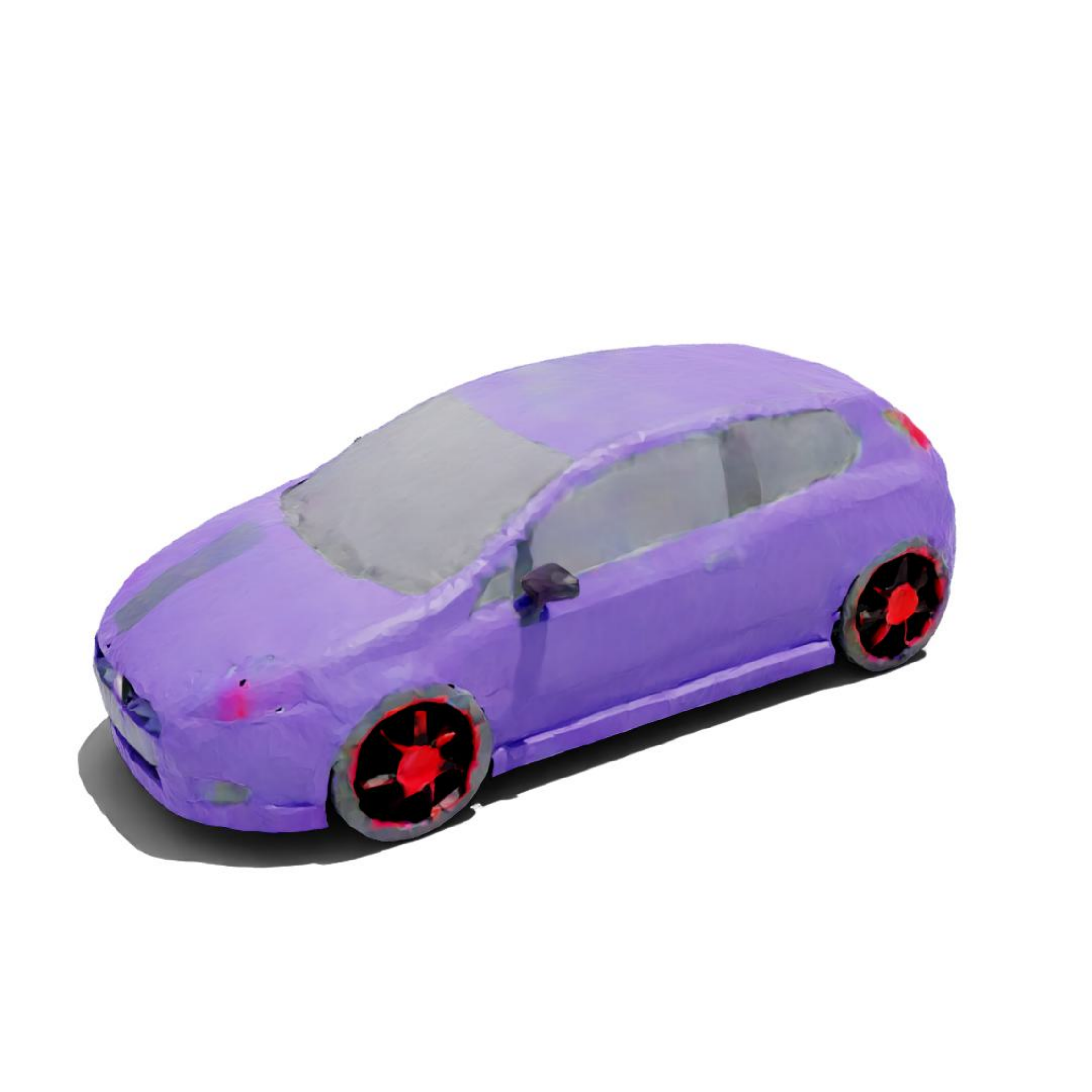}\includegraphics[width=0.16666666666666666\linewidth, trim={0 0cm 0 3cm}, clip]{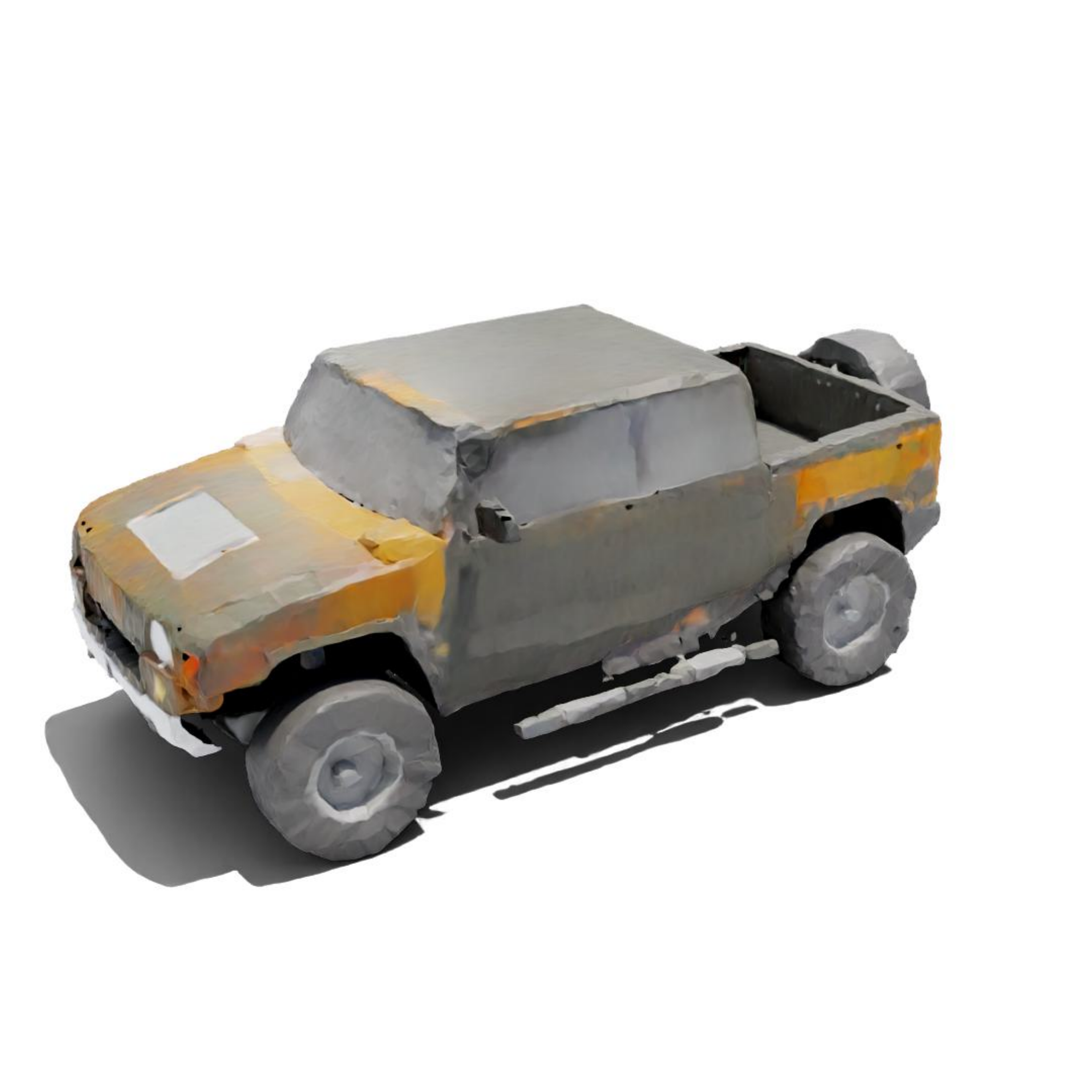}\includegraphics[width=0.16666666666666666\linewidth, trim={0 0cm 0 3cm}, clip]{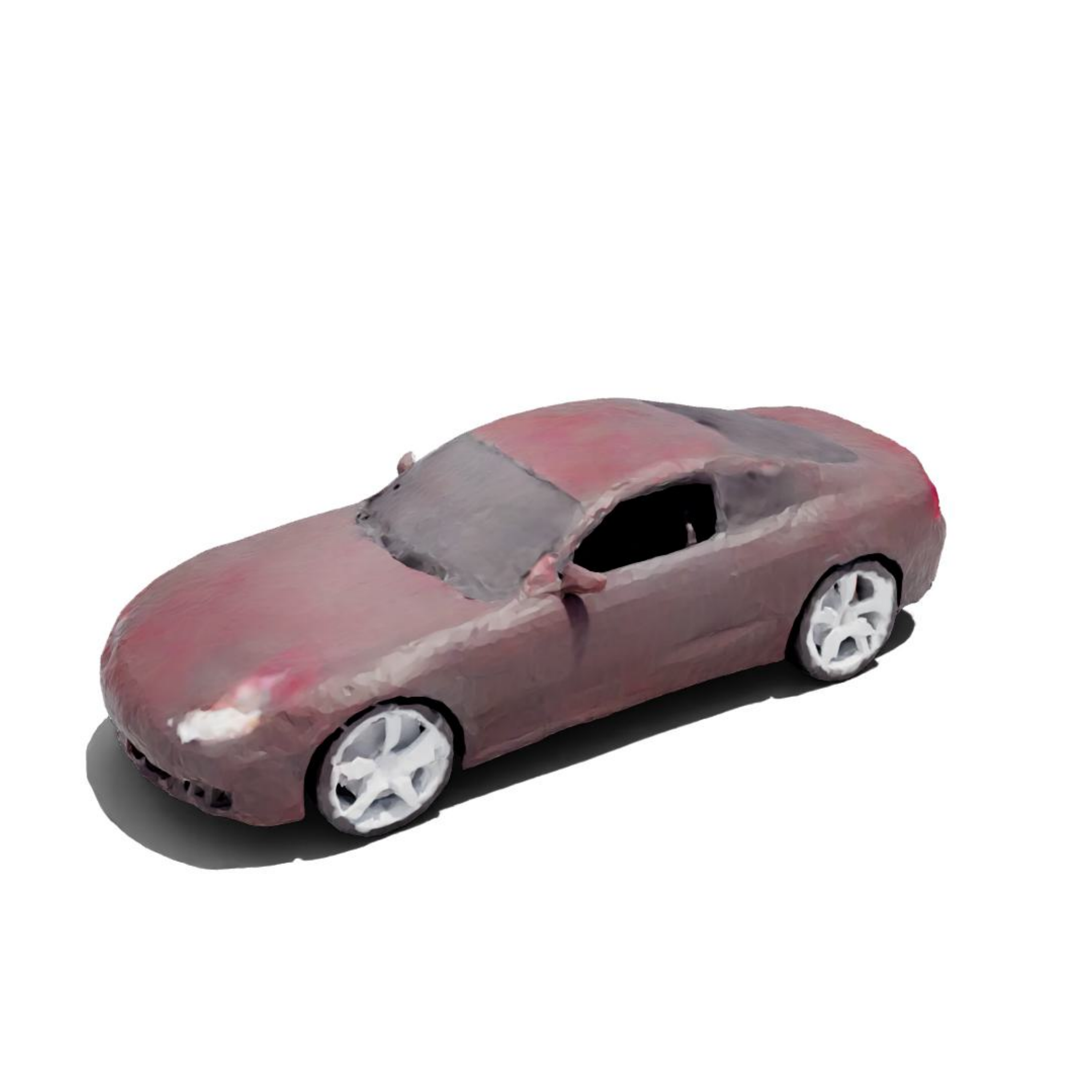}\includegraphics[width=0.16666666666666666\linewidth, trim={0 0cm 0 3cm}, clip]{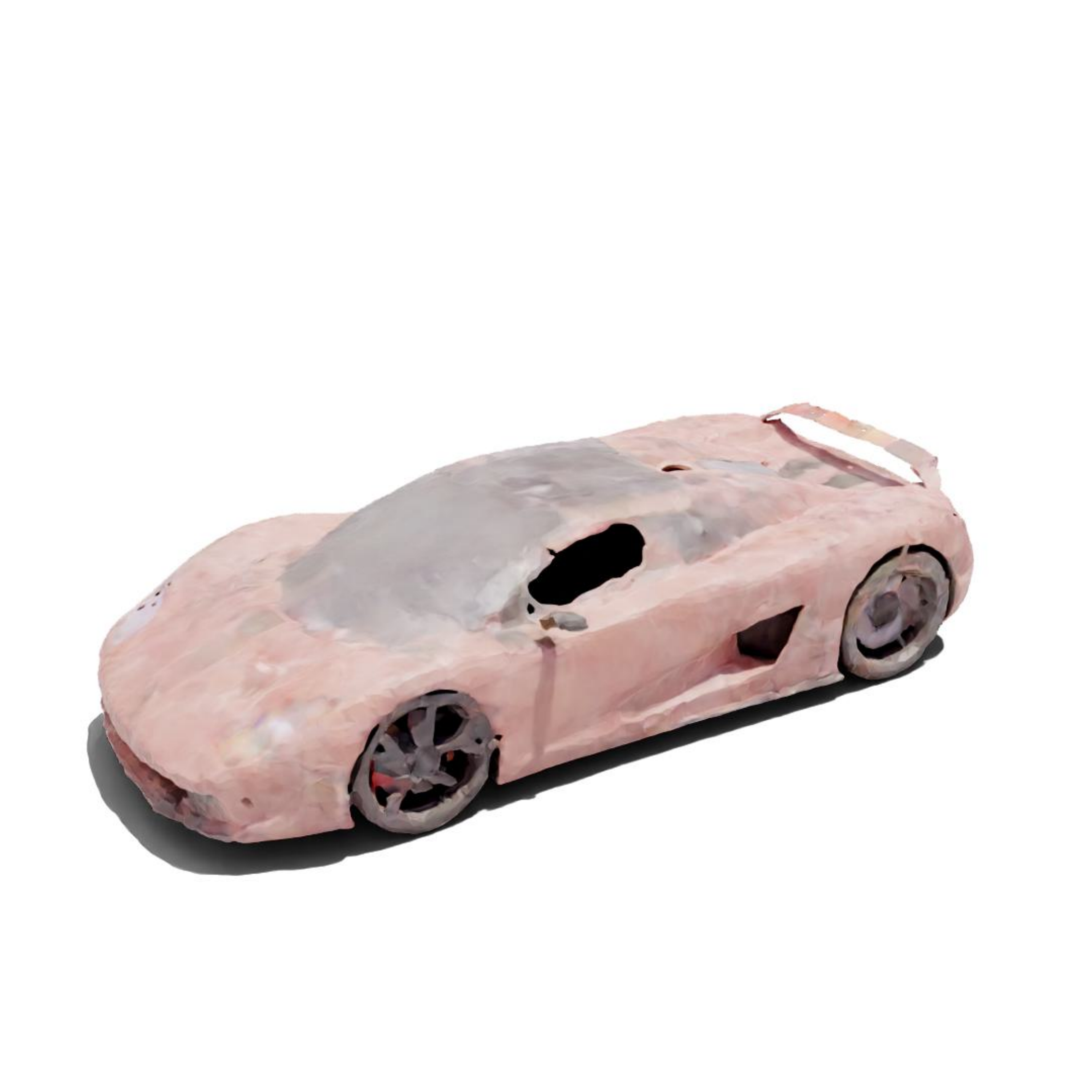}\includegraphics[width=0.16666666666666666\linewidth, trim={0 0cm 0 3cm}, clip]{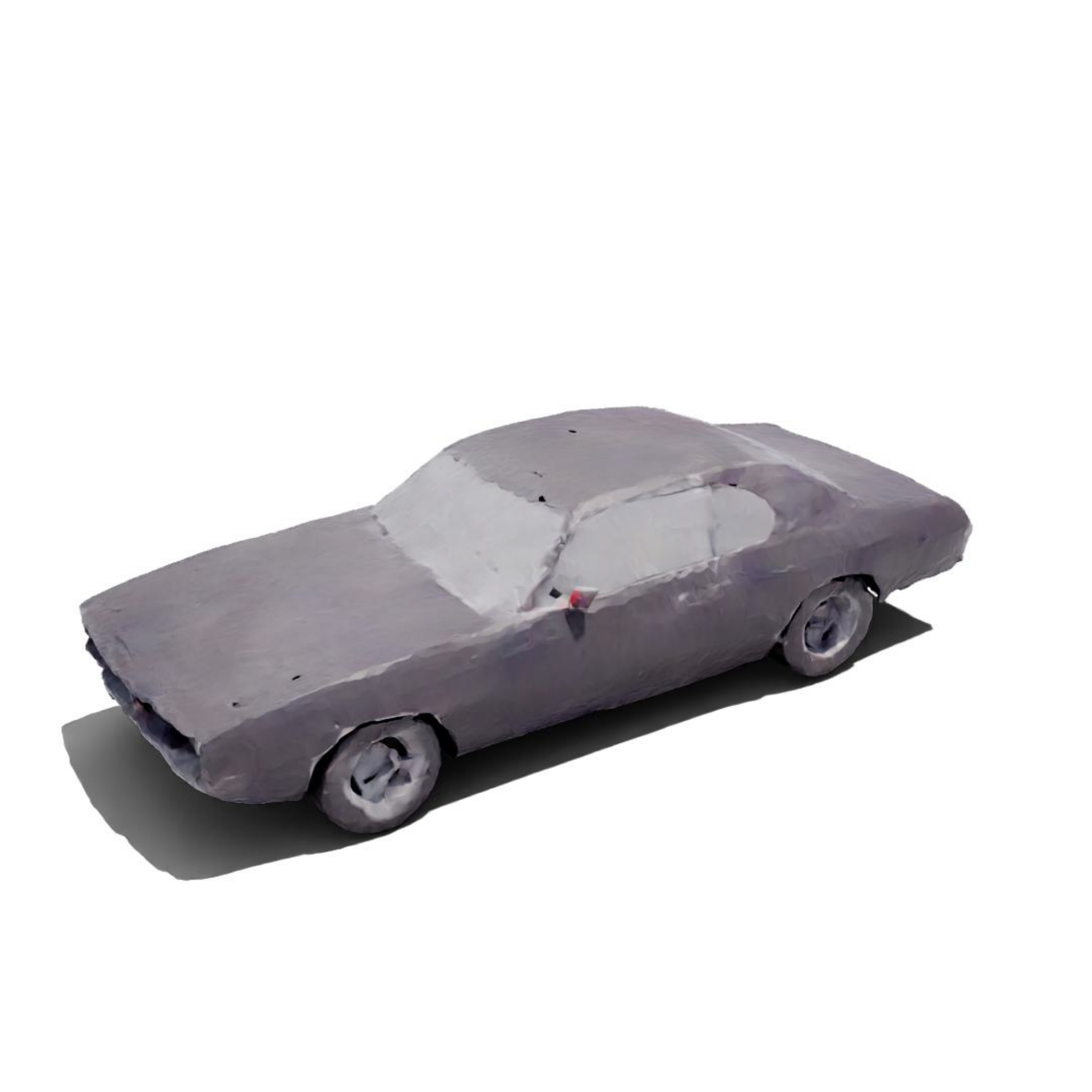}
\vspace{-0.7cm}
\caption{\textbf{Random selection of cars generated in high resolution.} Our high resolution model excels in generating meshes with a lot of geometric detail, including vertex-level texture.}
\label{fig:uncond:car:192}
\end{figure*}

\begin{figure*}[!ht]
\centering
\includegraphics[width=0.16666666666666666\linewidth, trim={0 0cm 0 4cm}, clip]{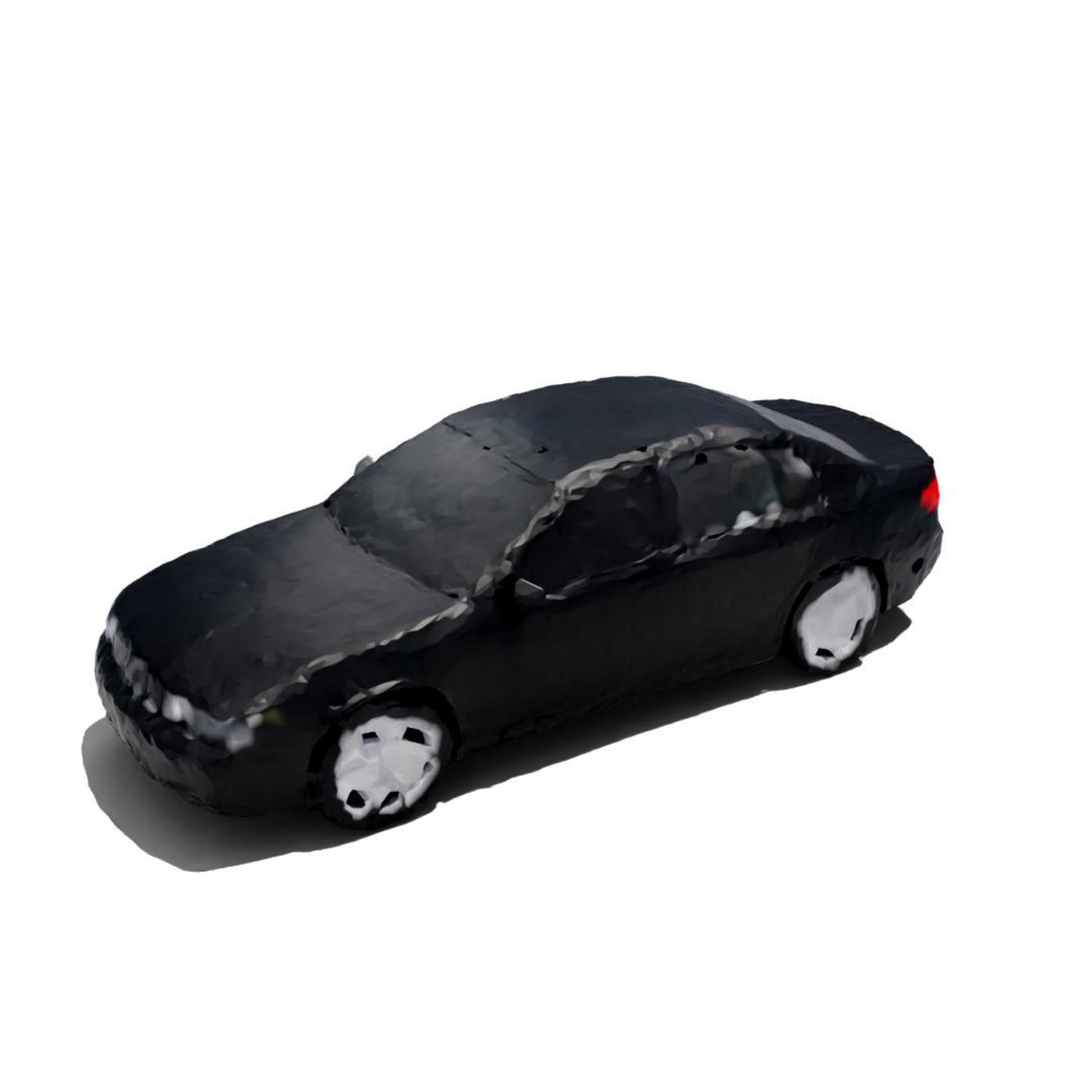}\includegraphics[width=0.16666666666666666\linewidth, trim={0 0cm 0 4cm}, clip]{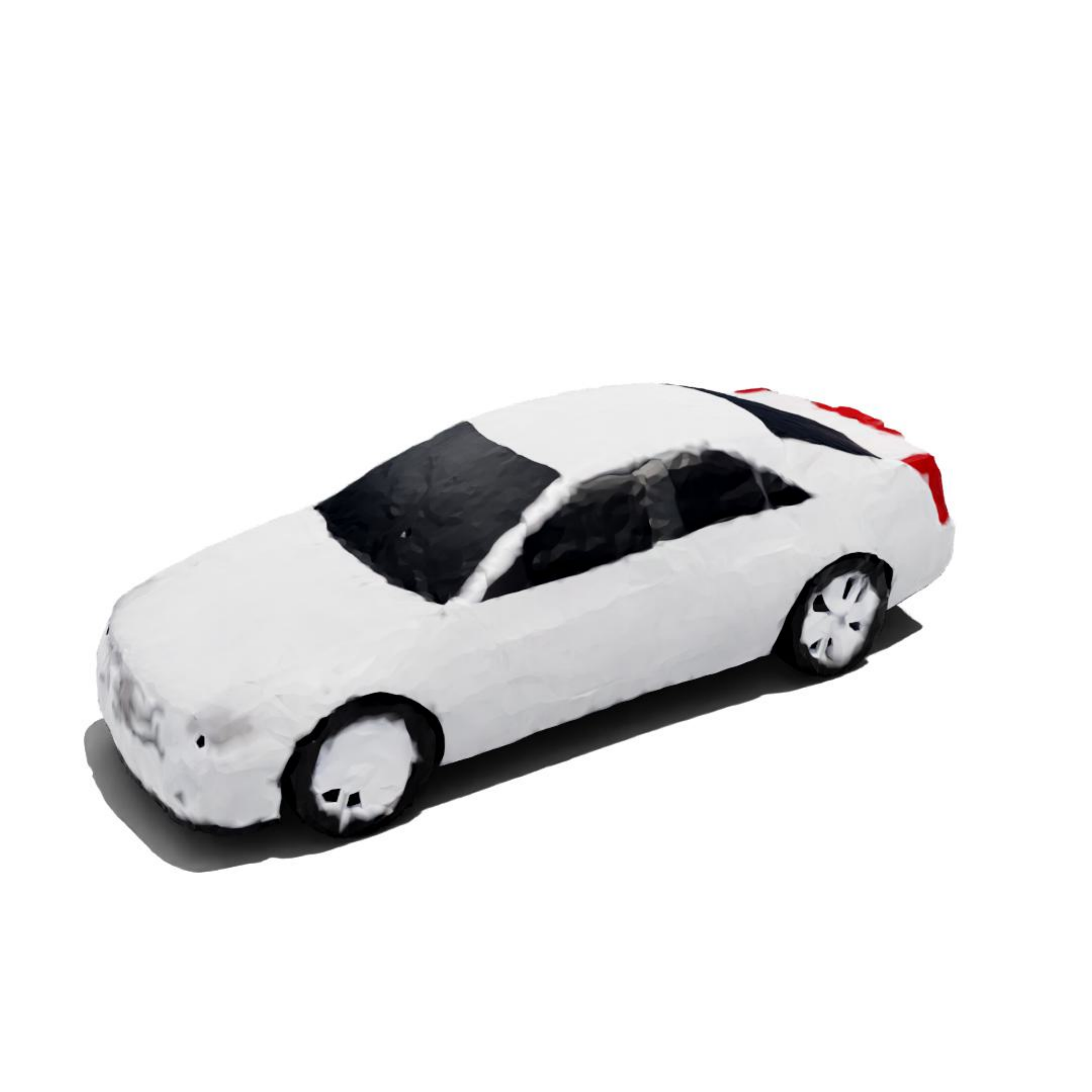}\includegraphics[width=0.16666666666666666\linewidth, trim={0 0cm 0 4cm}, clip]{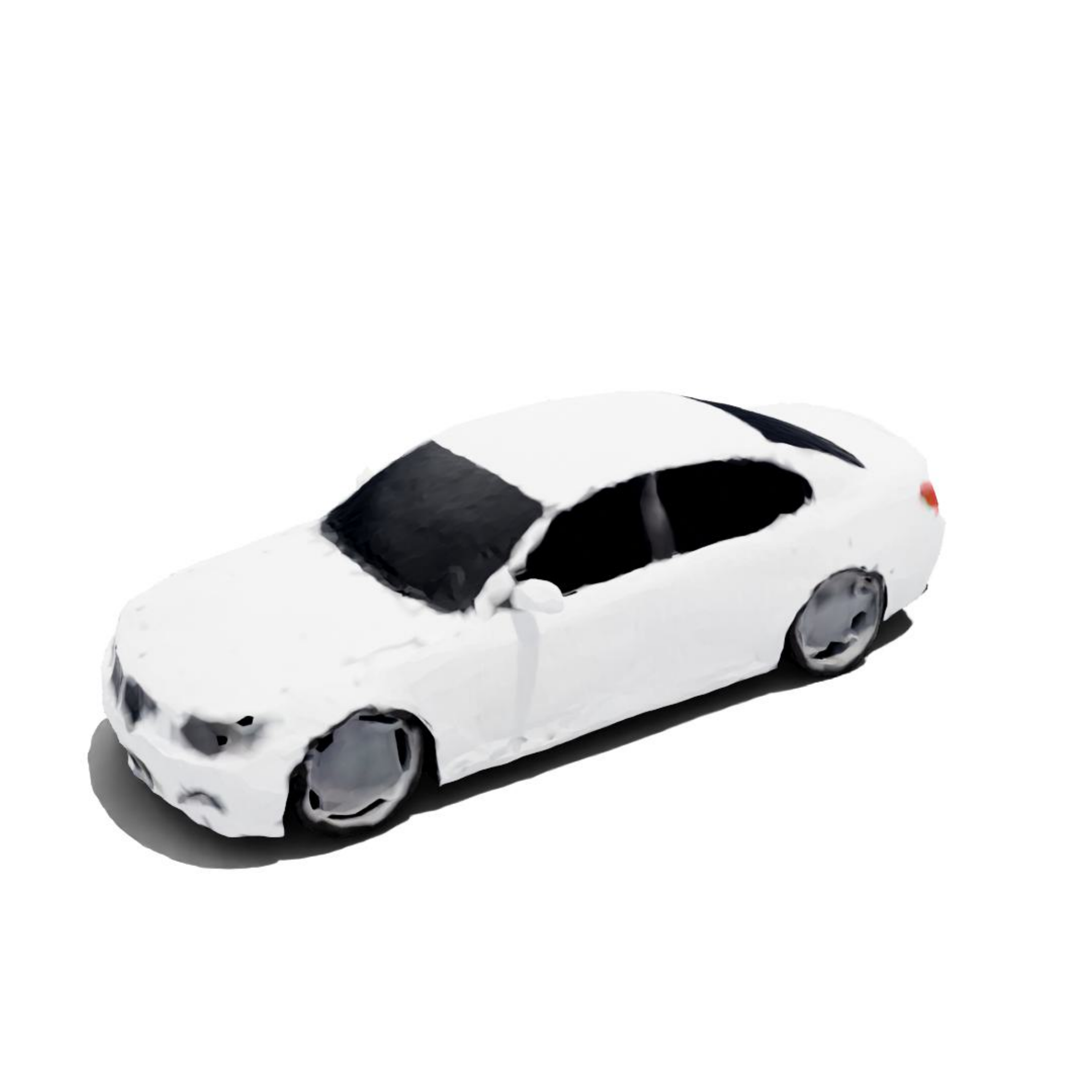}\includegraphics[width=0.16666666666666666\linewidth, trim={0 0cm 0 4cm}, clip]{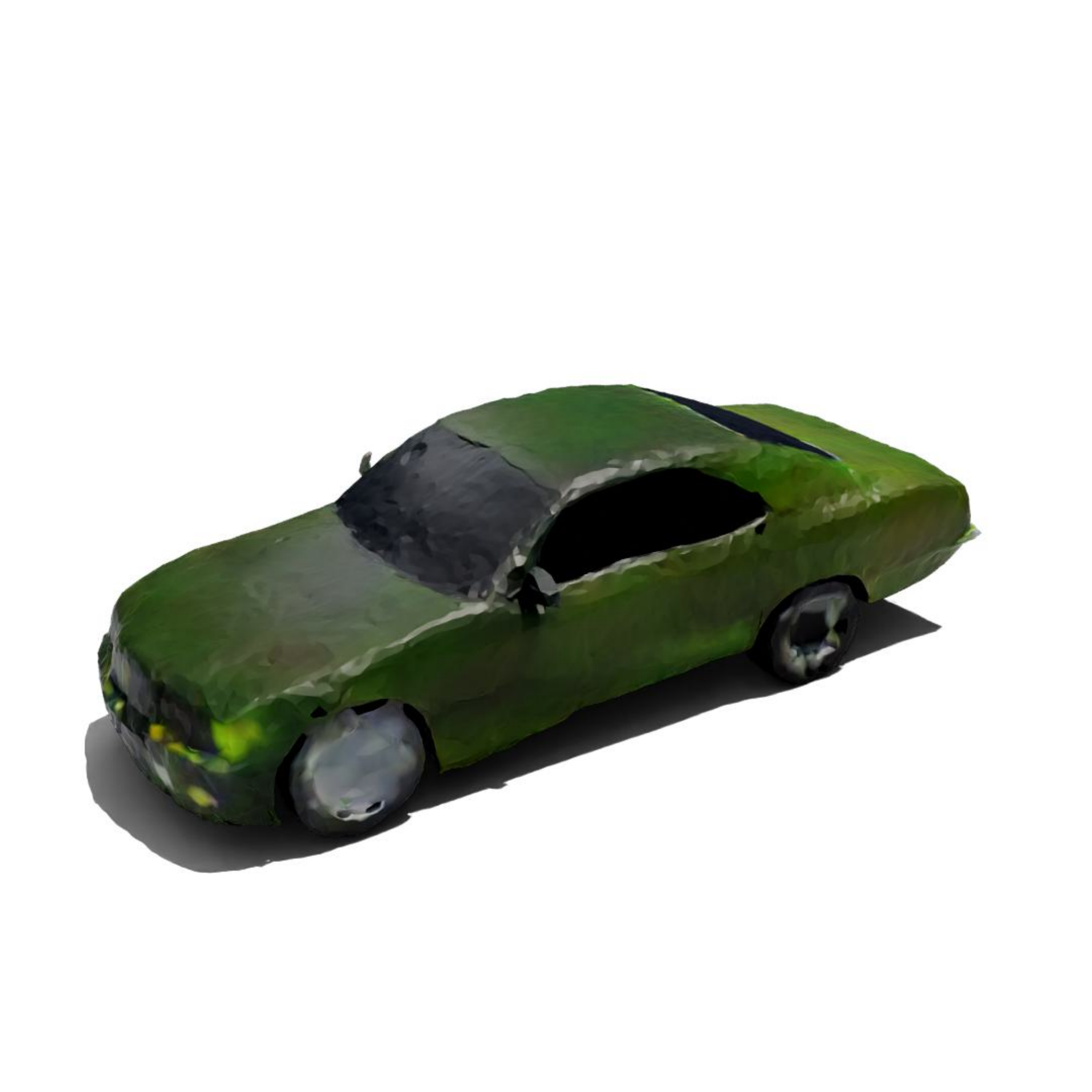}\includegraphics[width=0.16666666666666666\linewidth, trim={0 0cm 0 4cm}, clip]{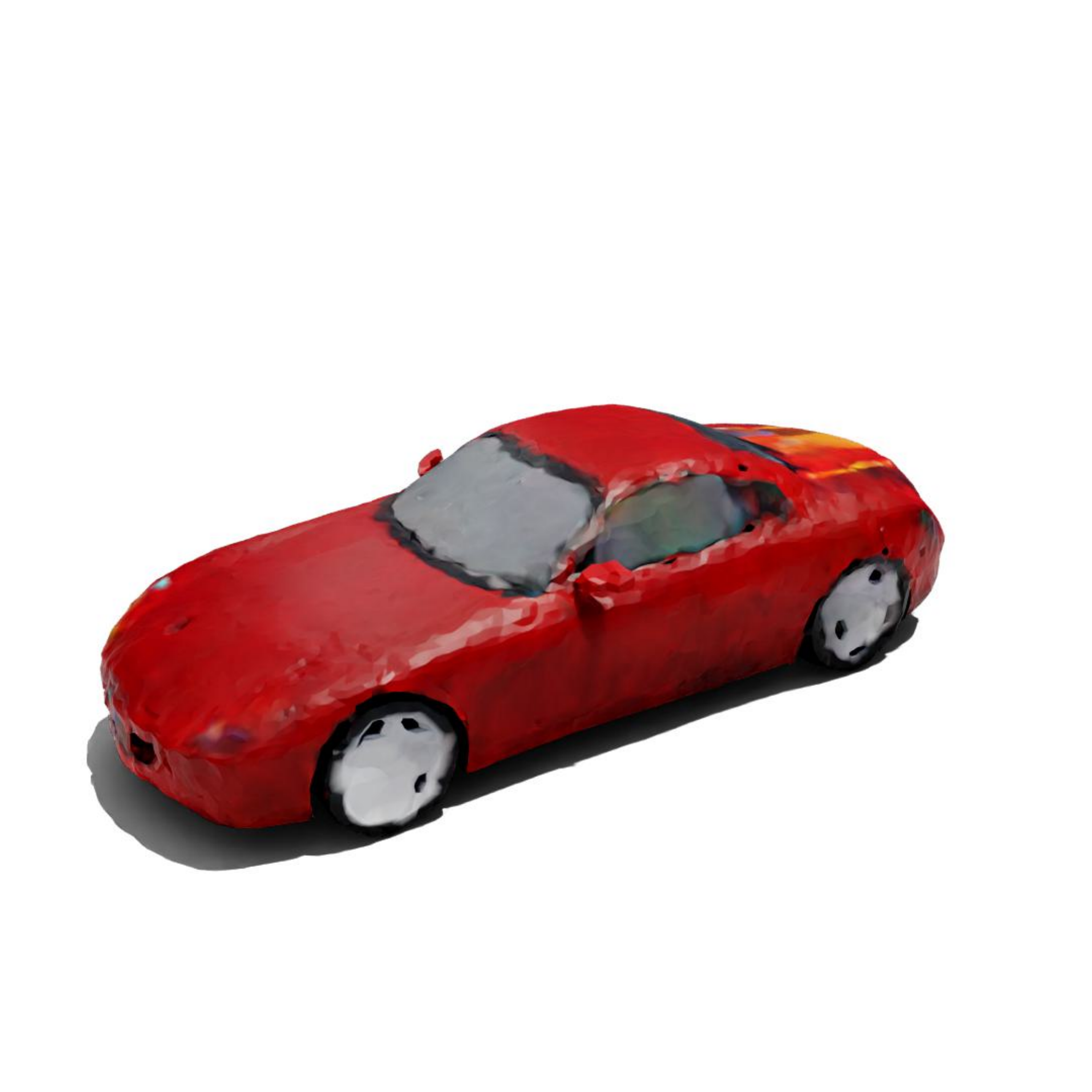}\includegraphics[width=0.16666666666666666\linewidth, trim={0 0cm 0 4cm}, clip]{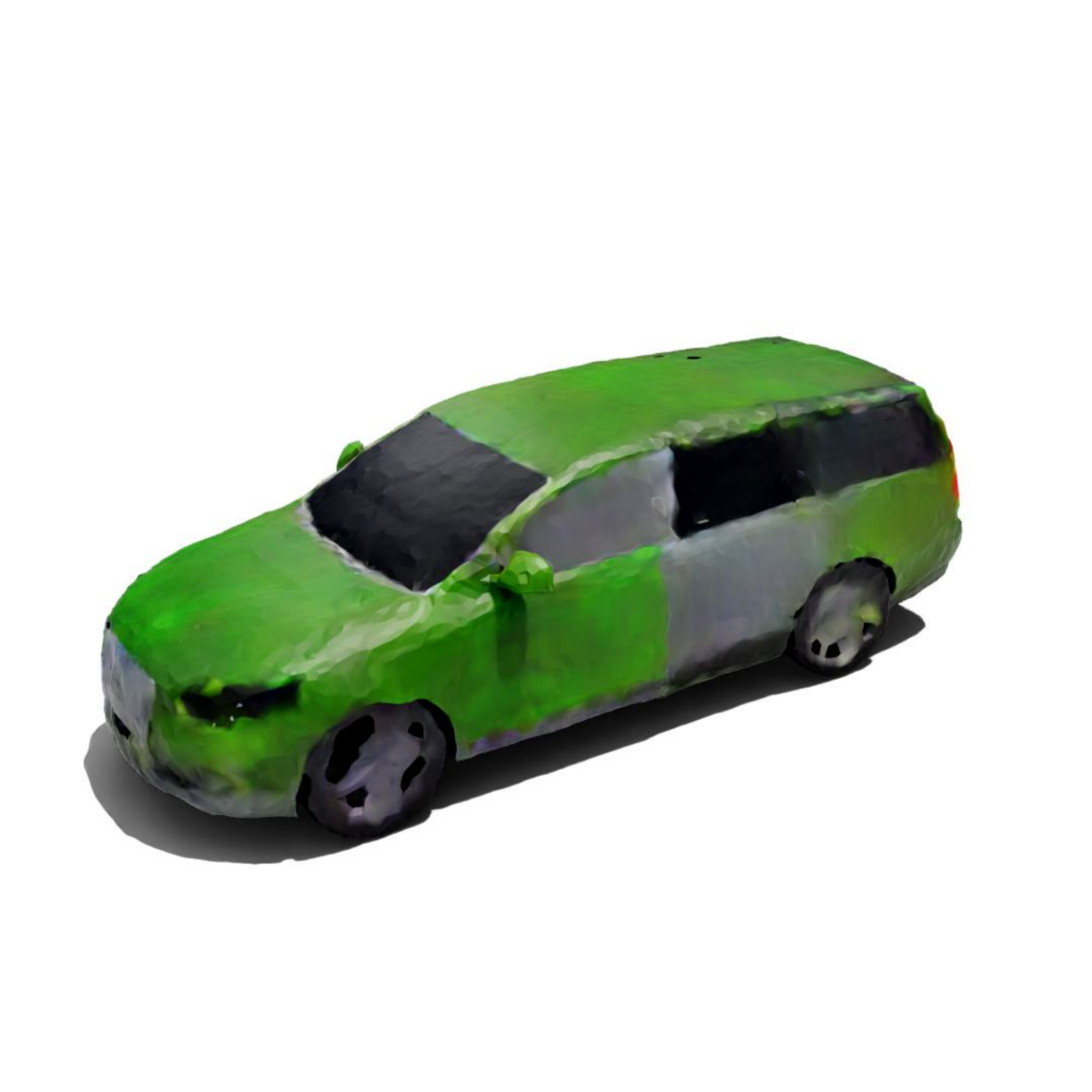}\\

\vspace{-0.43cm}
\includegraphics[width=0.16666666666666666\linewidth, trim={0 0cm 0 3cm}, clip]{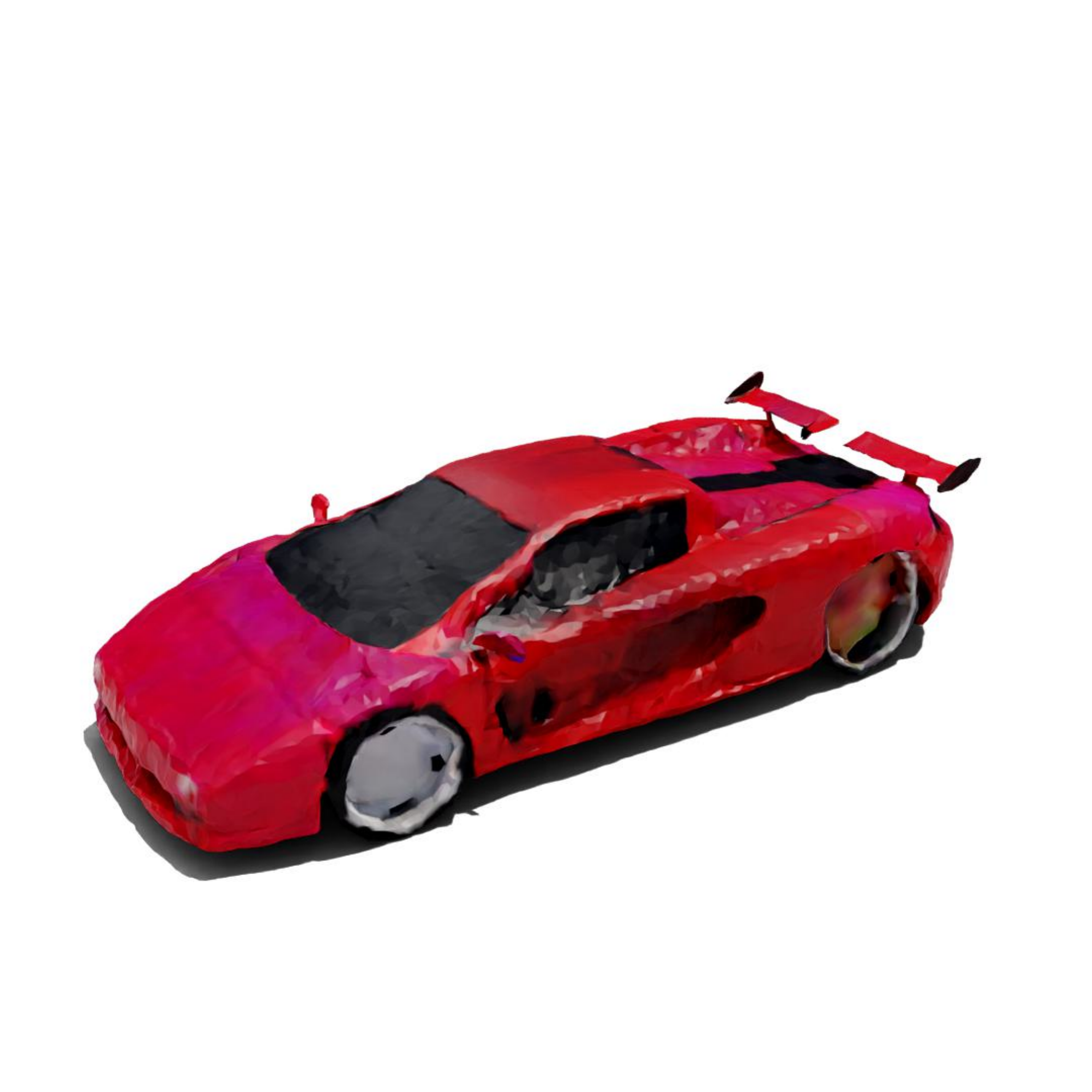}\includegraphics[width=0.16666666666666666\linewidth, trim={0 0cm 0 3cm}, clip]{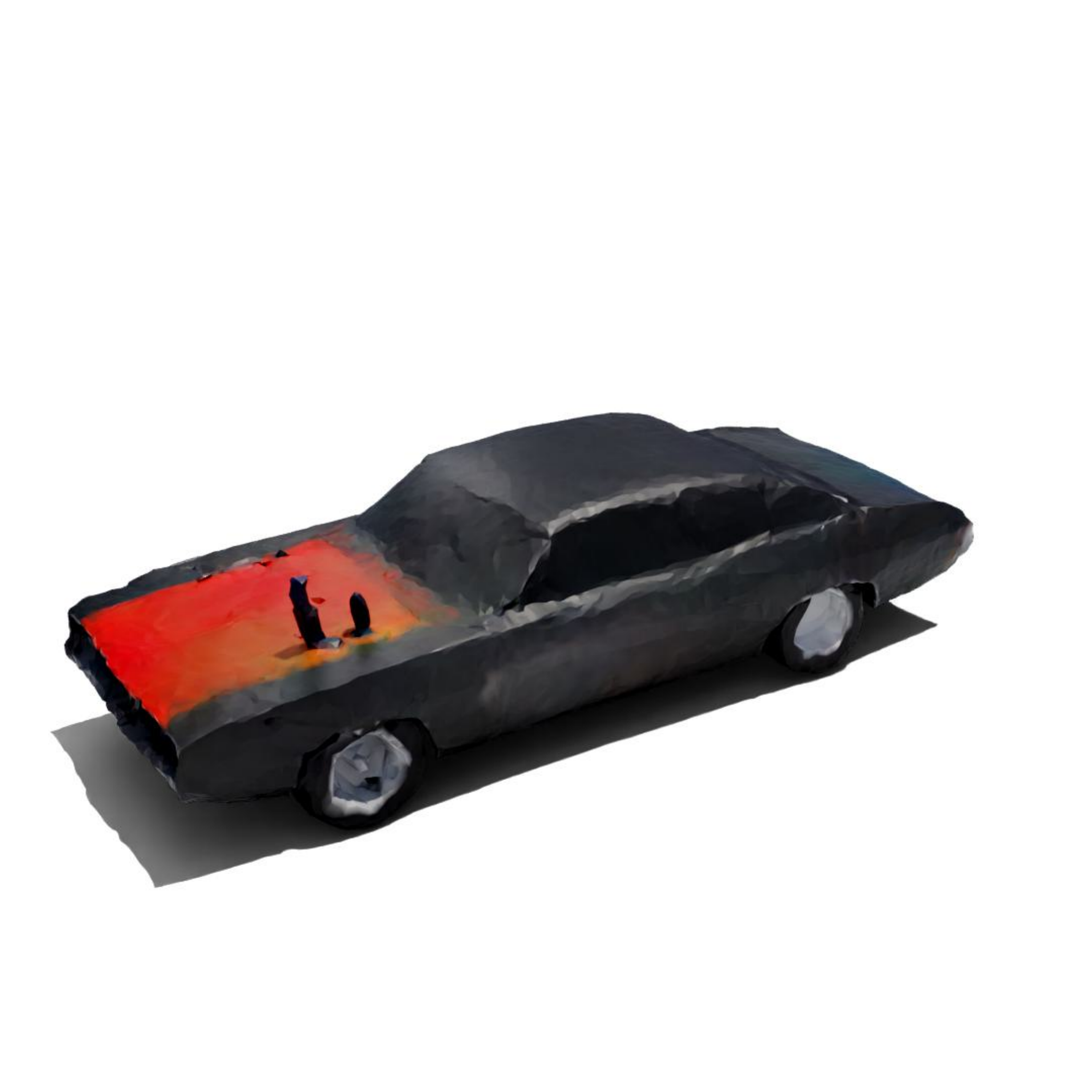}\includegraphics[width=0.16666666666666666\linewidth, trim={0 0cm 0 3cm}, clip]{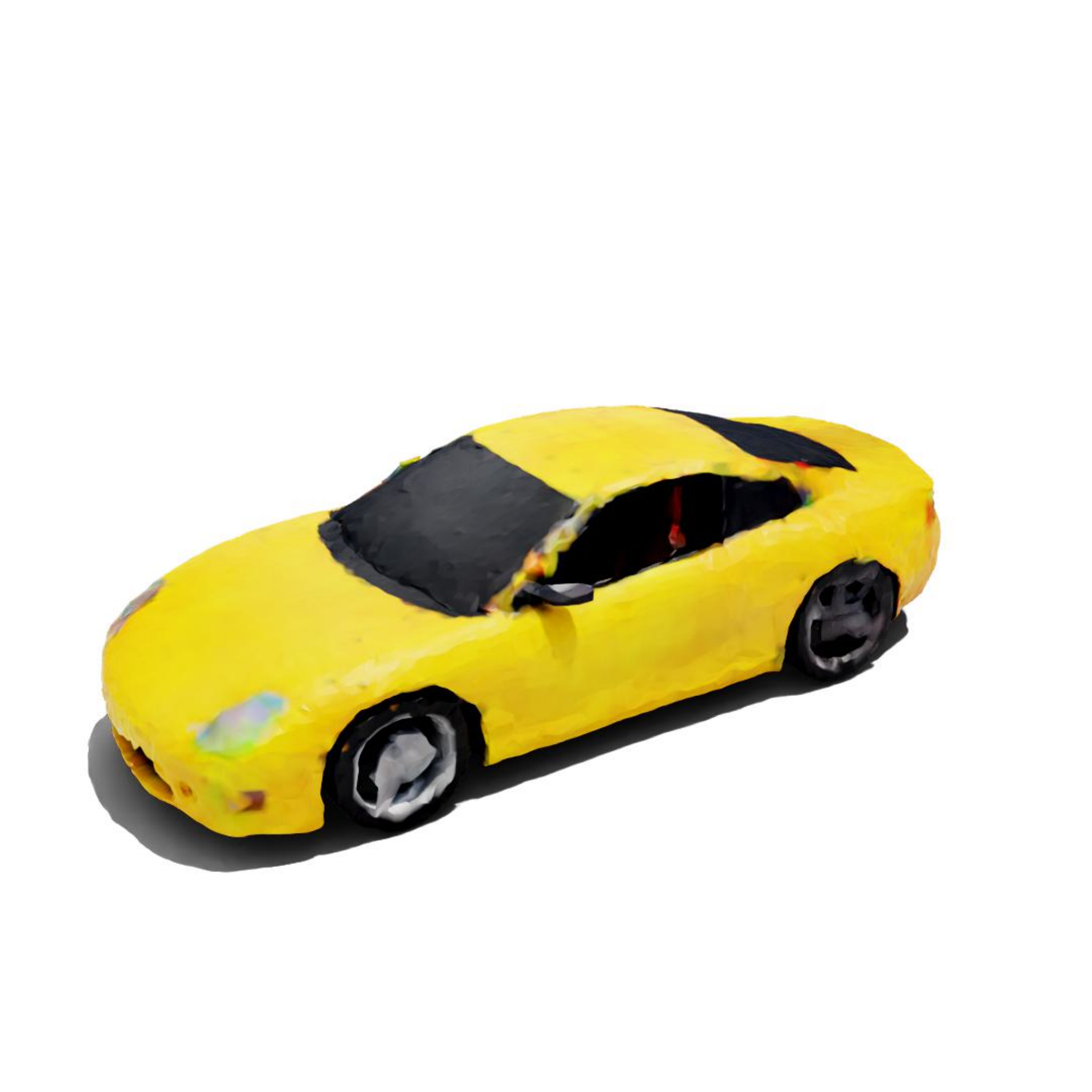}\includegraphics[width=0.16666666666666666\linewidth, trim={0 0cm 0 3cm}, clip]{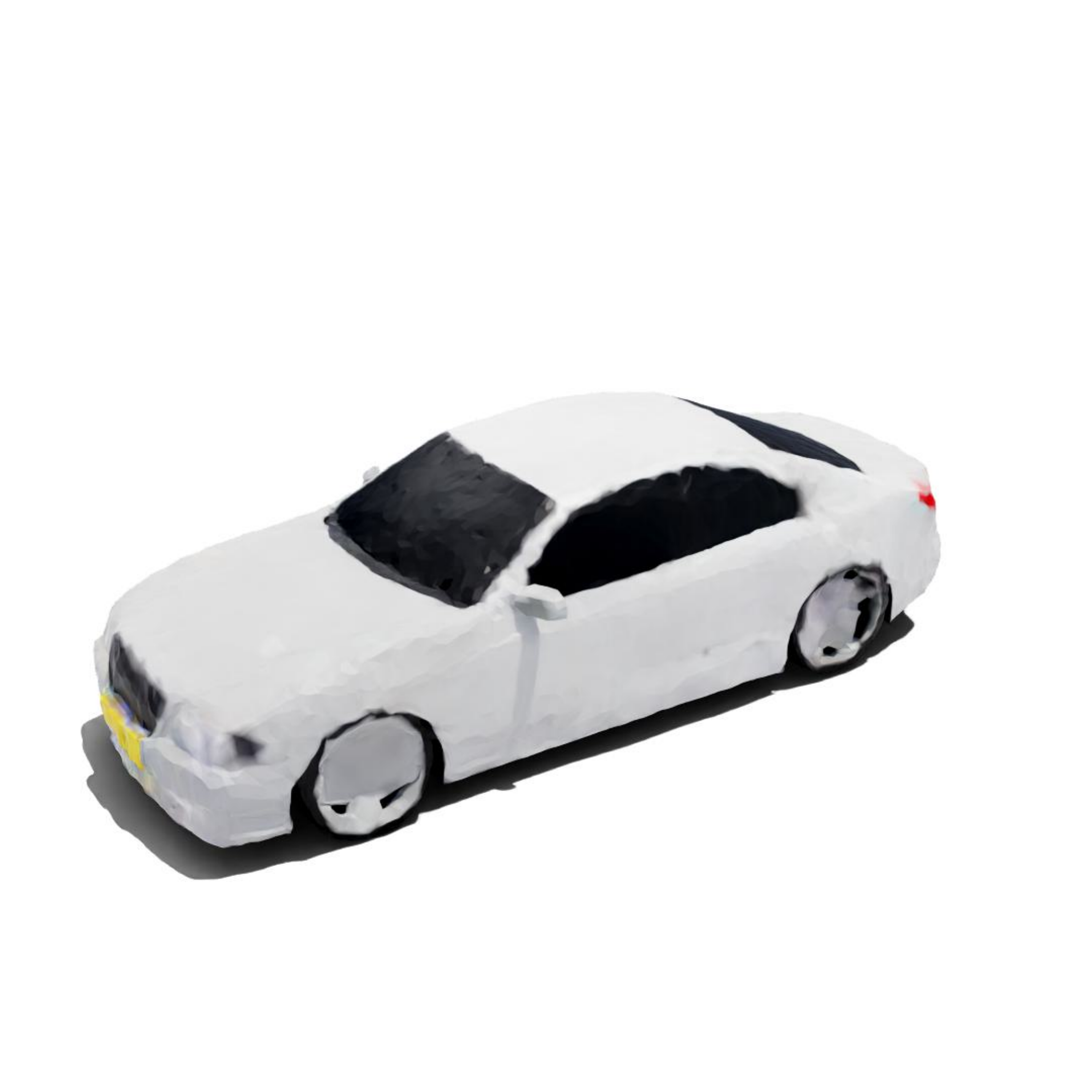}\includegraphics[width=0.16666666666666666\linewidth, trim={0 0cm 0 3cm}, clip]{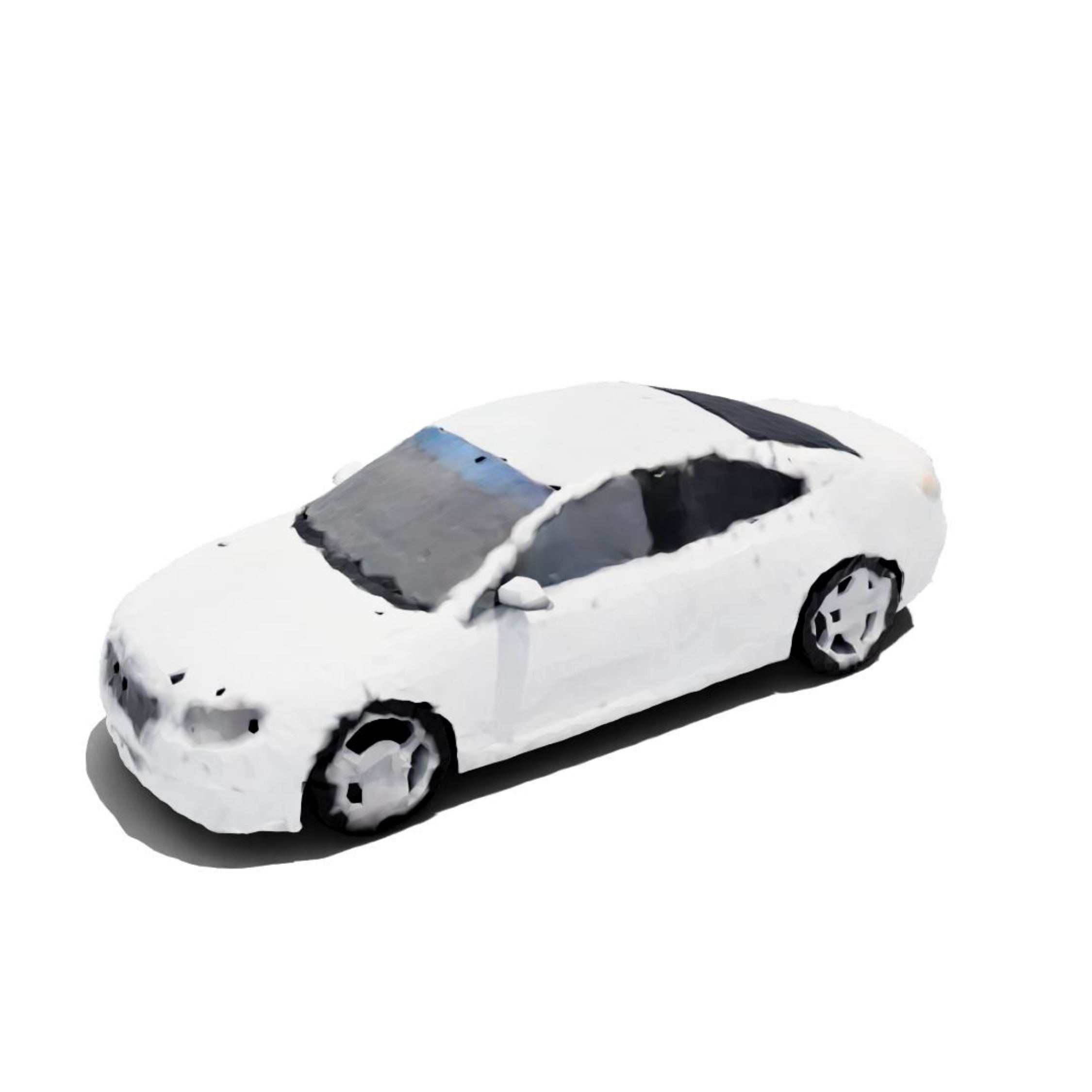}\includegraphics[width=0.16666666666666666\linewidth, trim={0 0cm 0 3cm}, clip]{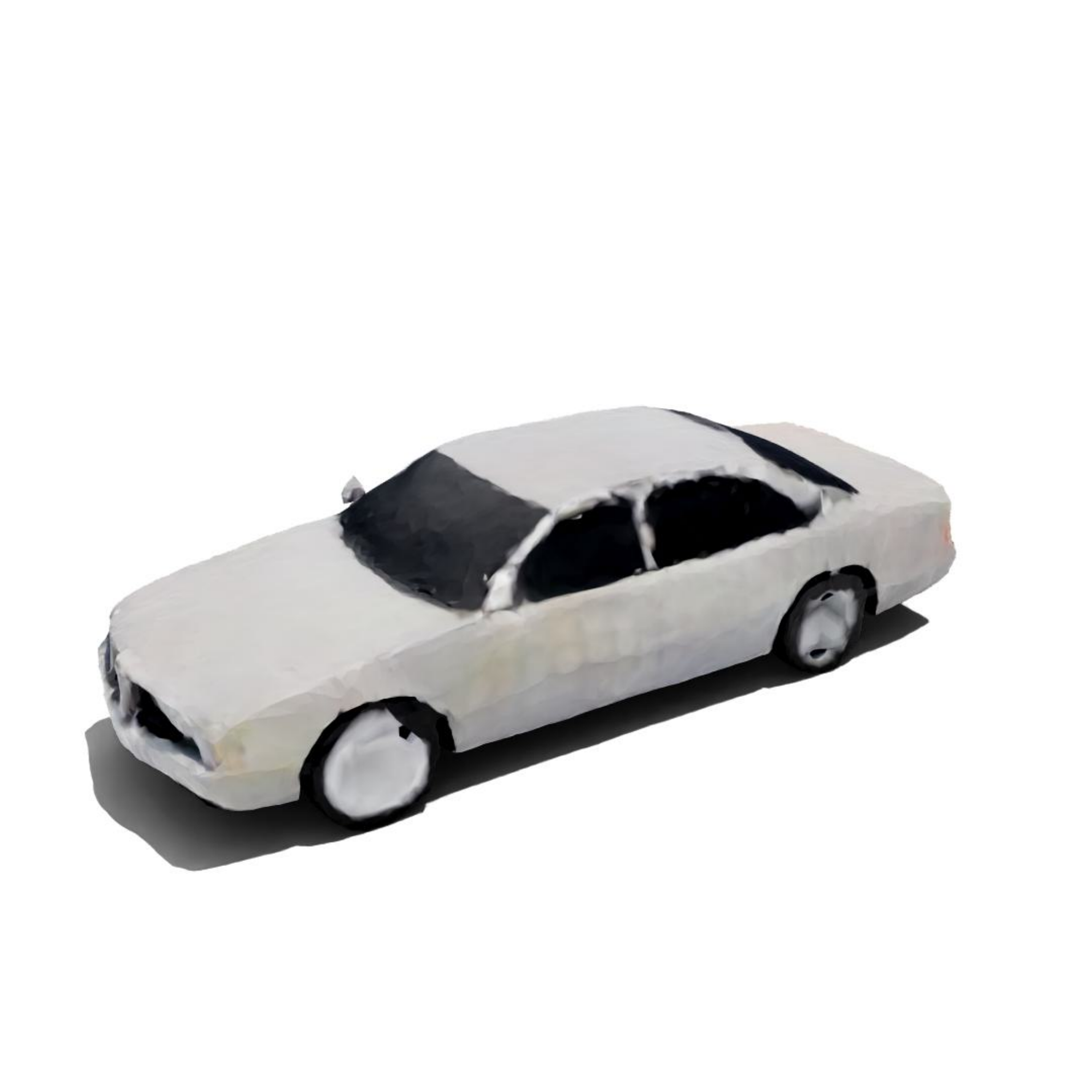}\\

\vspace{-0.43cm}
\includegraphics[width=0.16666666666666666\linewidth, trim={0 0cm 0 3cm}, clip]{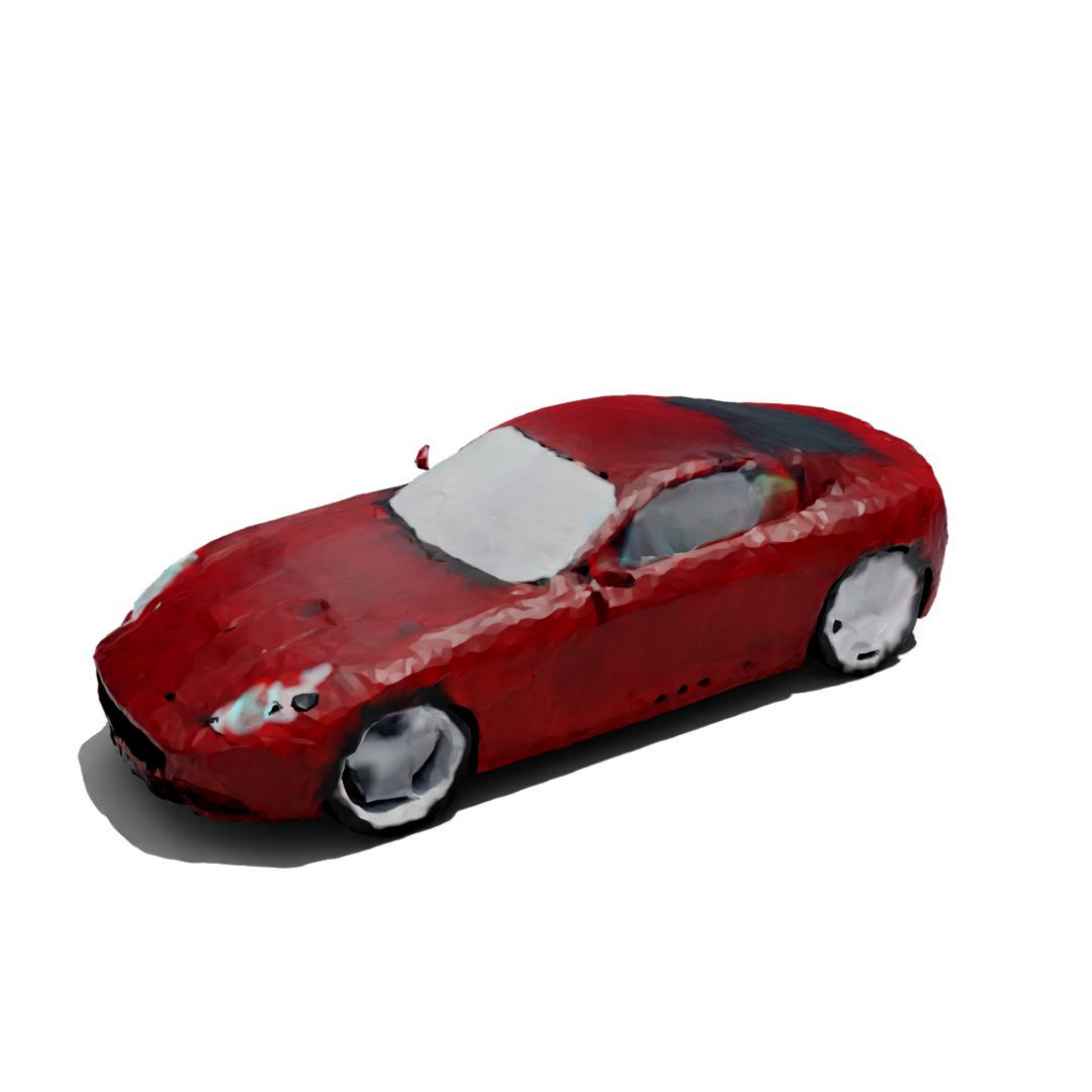}\includegraphics[width=0.16666666666666666\linewidth, trim={0 0cm 0 3cm}, clip]{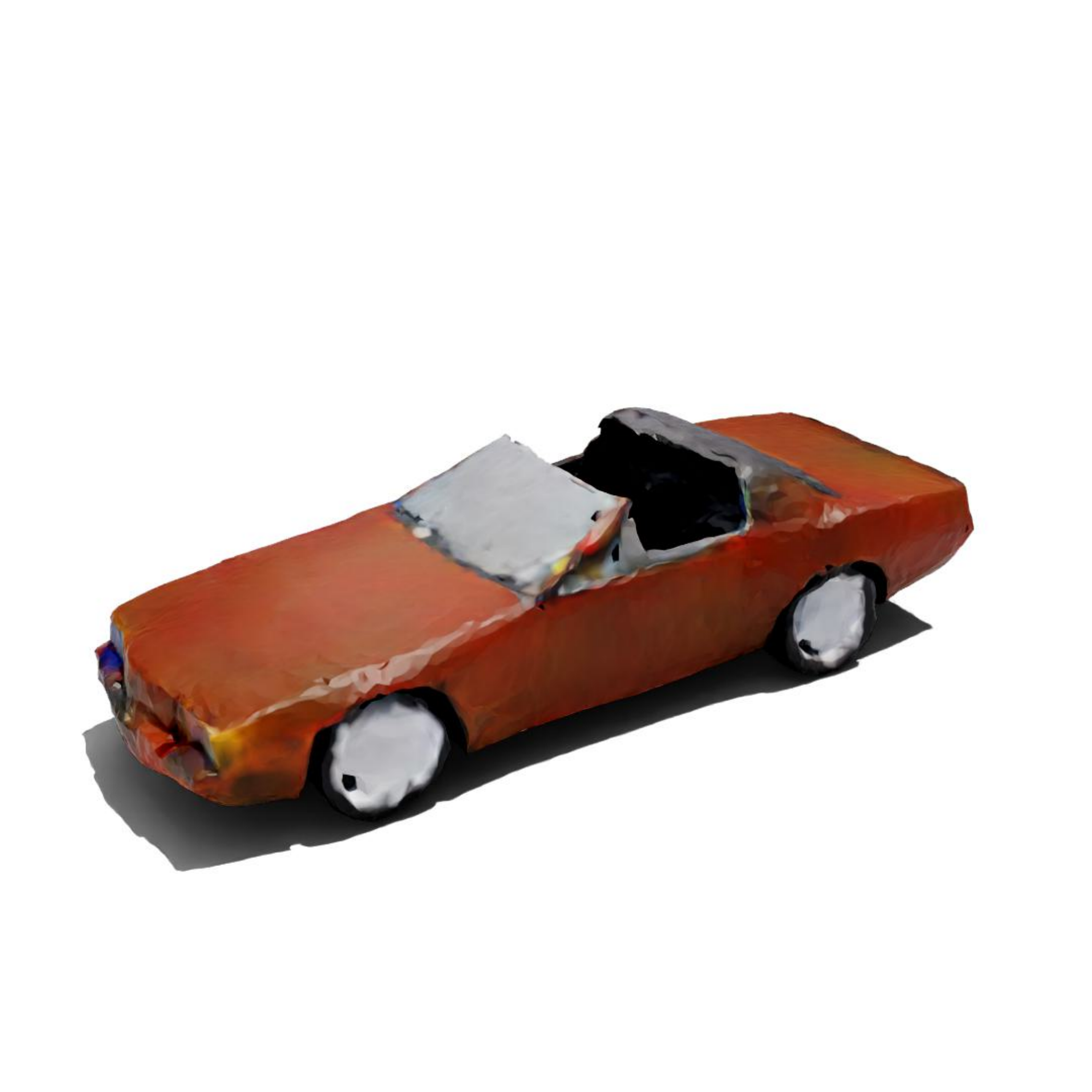}\includegraphics[width=0.16666666666666666\linewidth, trim={0 0cm 0 3cm}, clip]{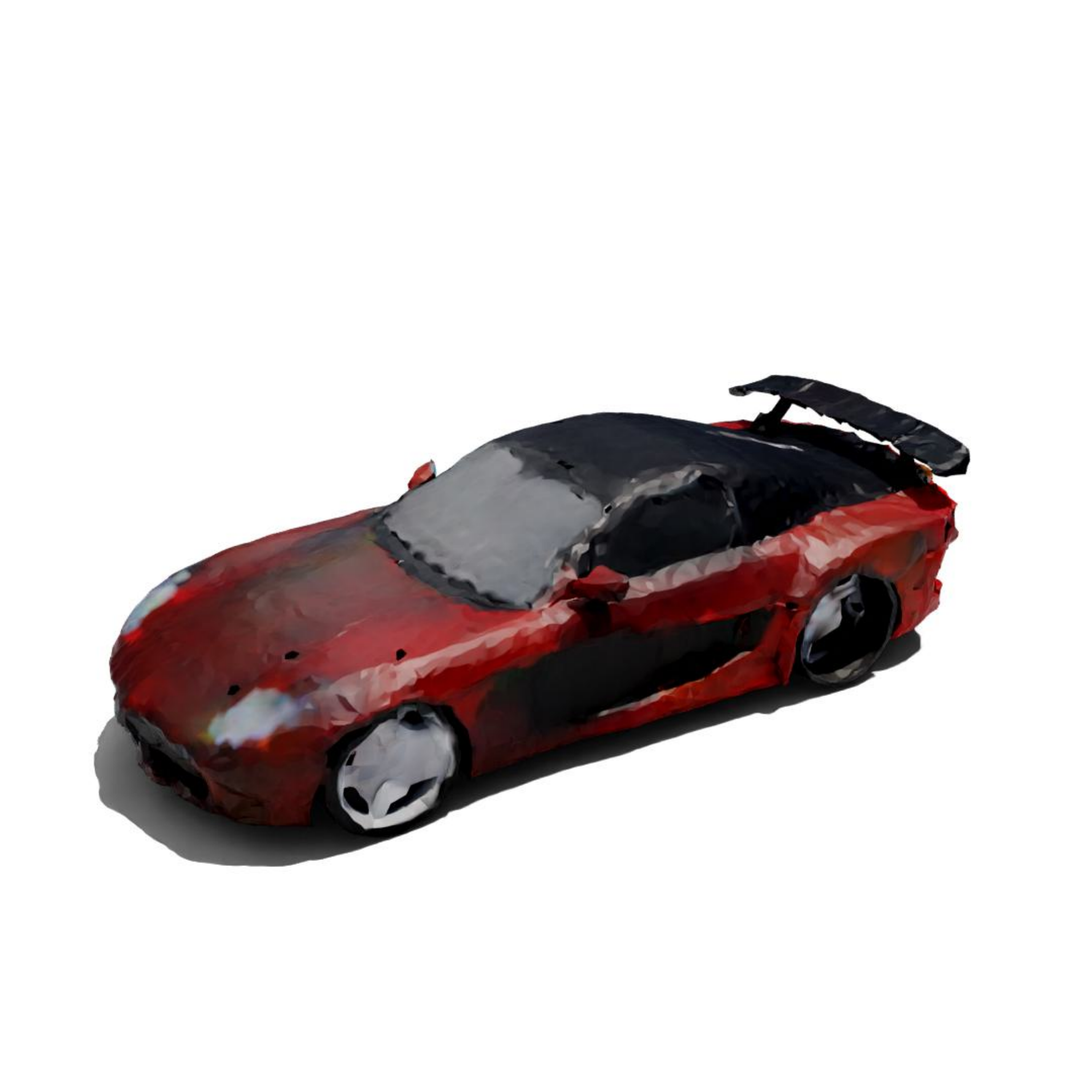}\includegraphics[width=0.16666666666666666\linewidth, trim={0 0cm 0 3cm}, clip]{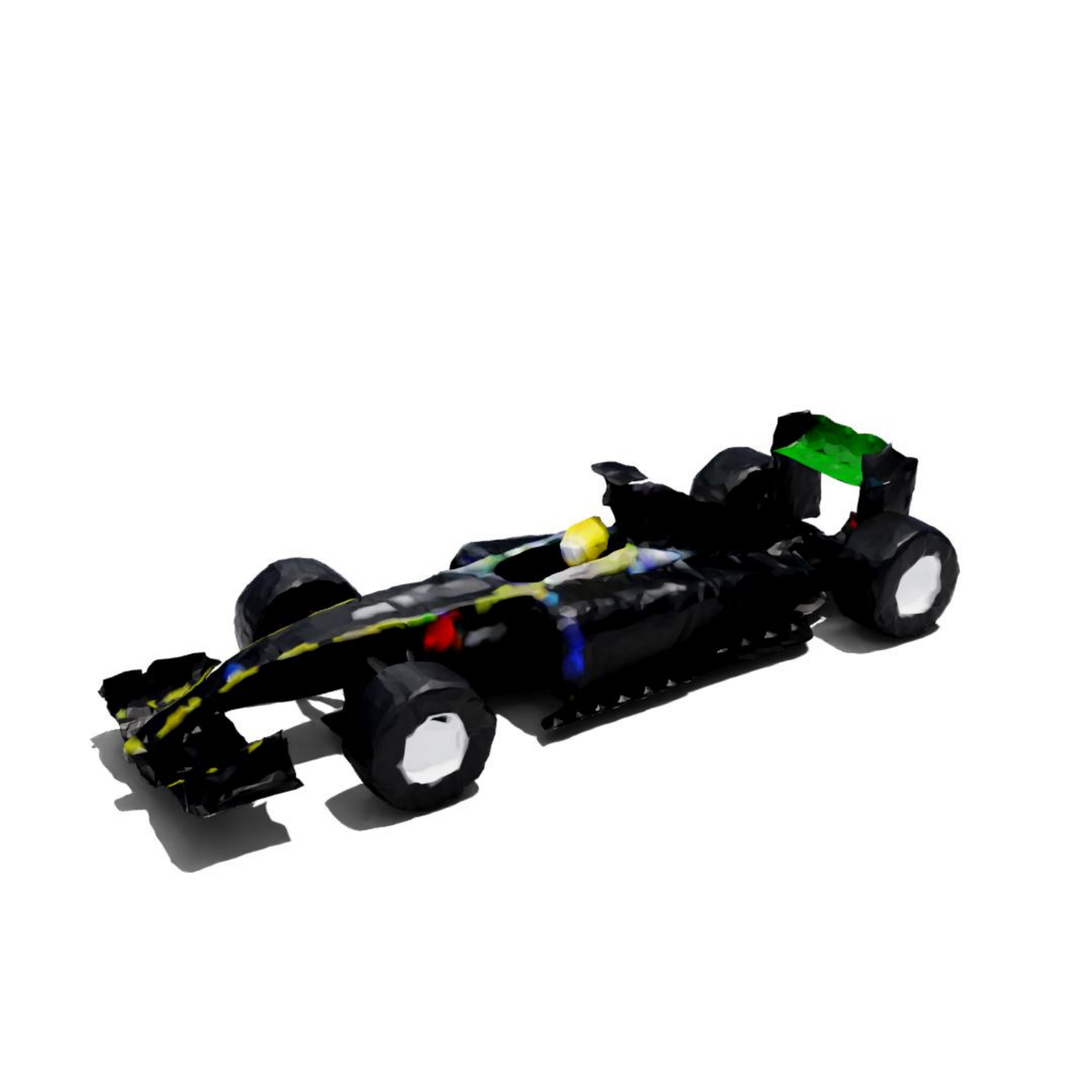}\includegraphics[width=0.16666666666666666\linewidth, trim={0 0cm 0 3cm}, clip]{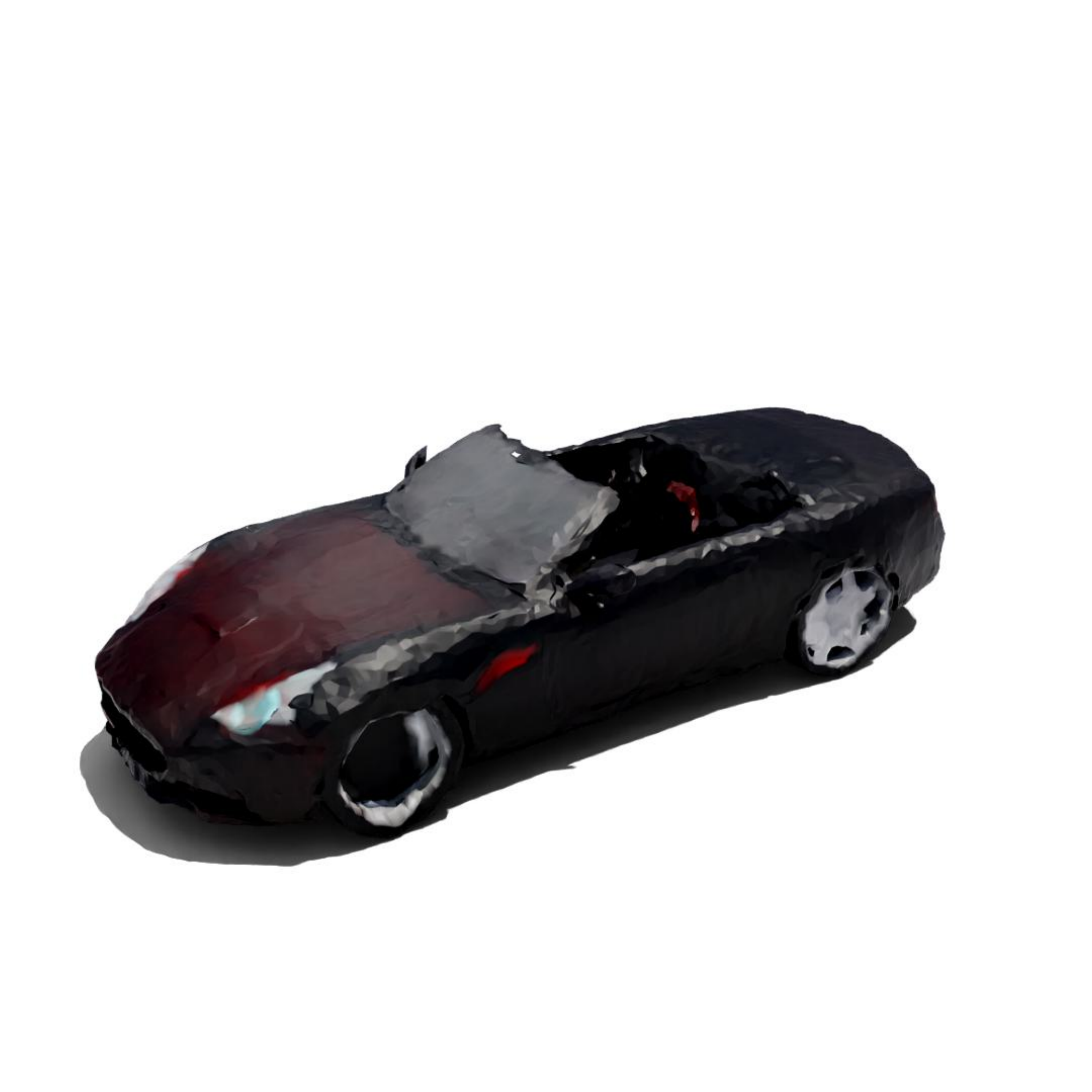}\includegraphics[width=0.16666666666666666\linewidth, trim={0 0cm 0 3cm}, clip]{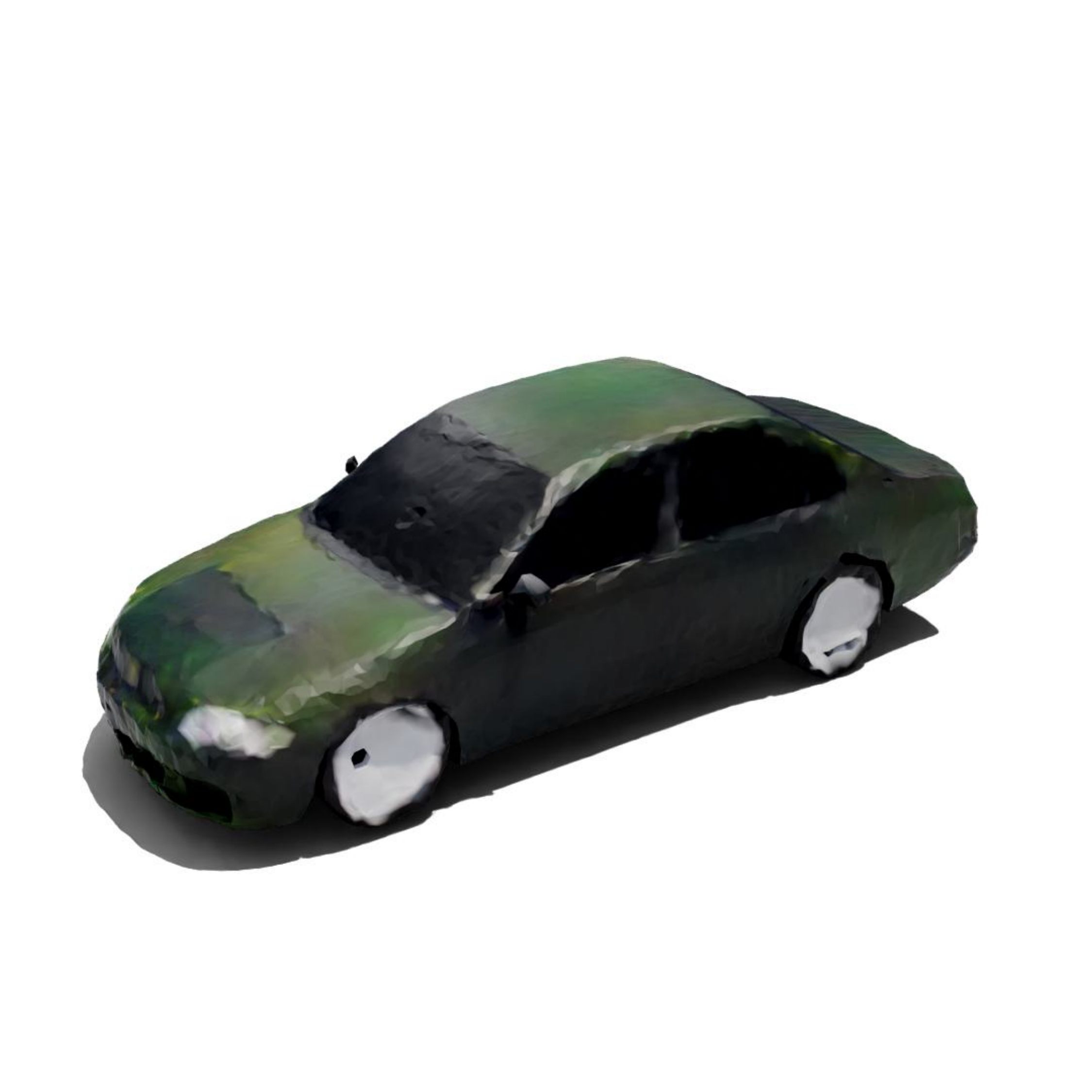}\\

\vspace{-0.43cm}
\includegraphics[width=0.16666666666666666\linewidth, trim={0 0cm 0 3cm}, clip]{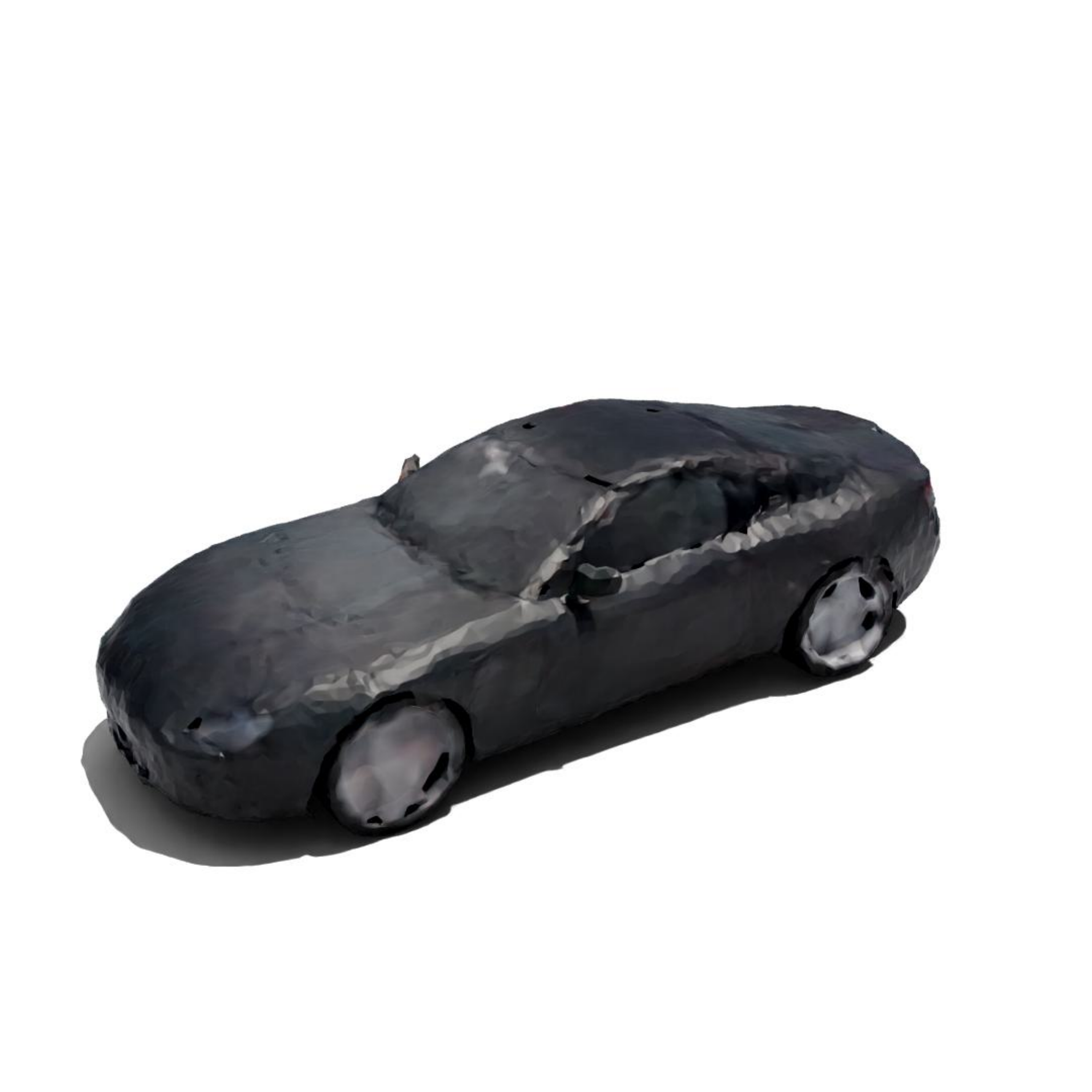}\includegraphics[width=0.16666666666666666\linewidth, trim={0 0cm 0 3cm}, clip]{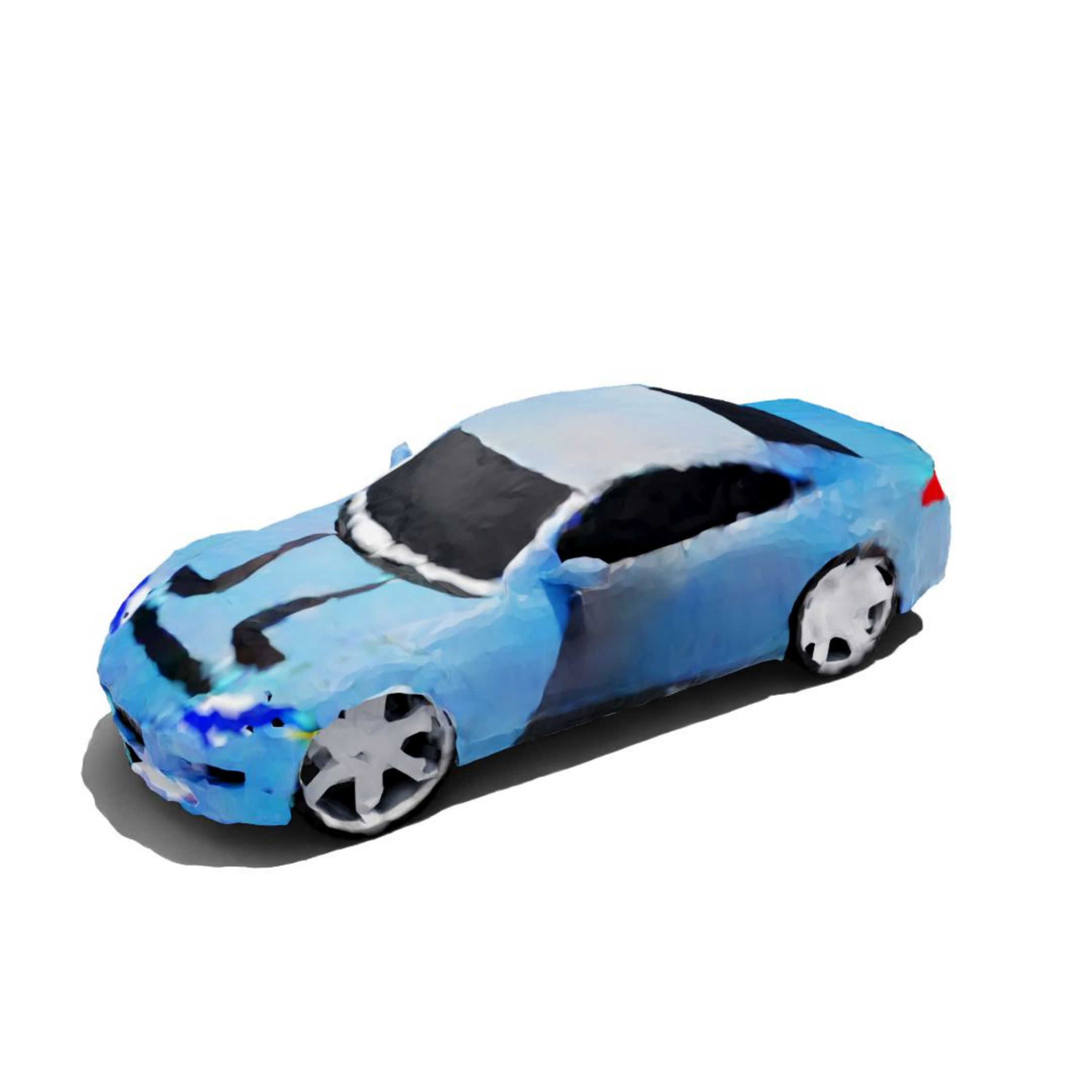}\includegraphics[width=0.16666666666666666\linewidth, trim={0 0cm 0 3cm}, clip]{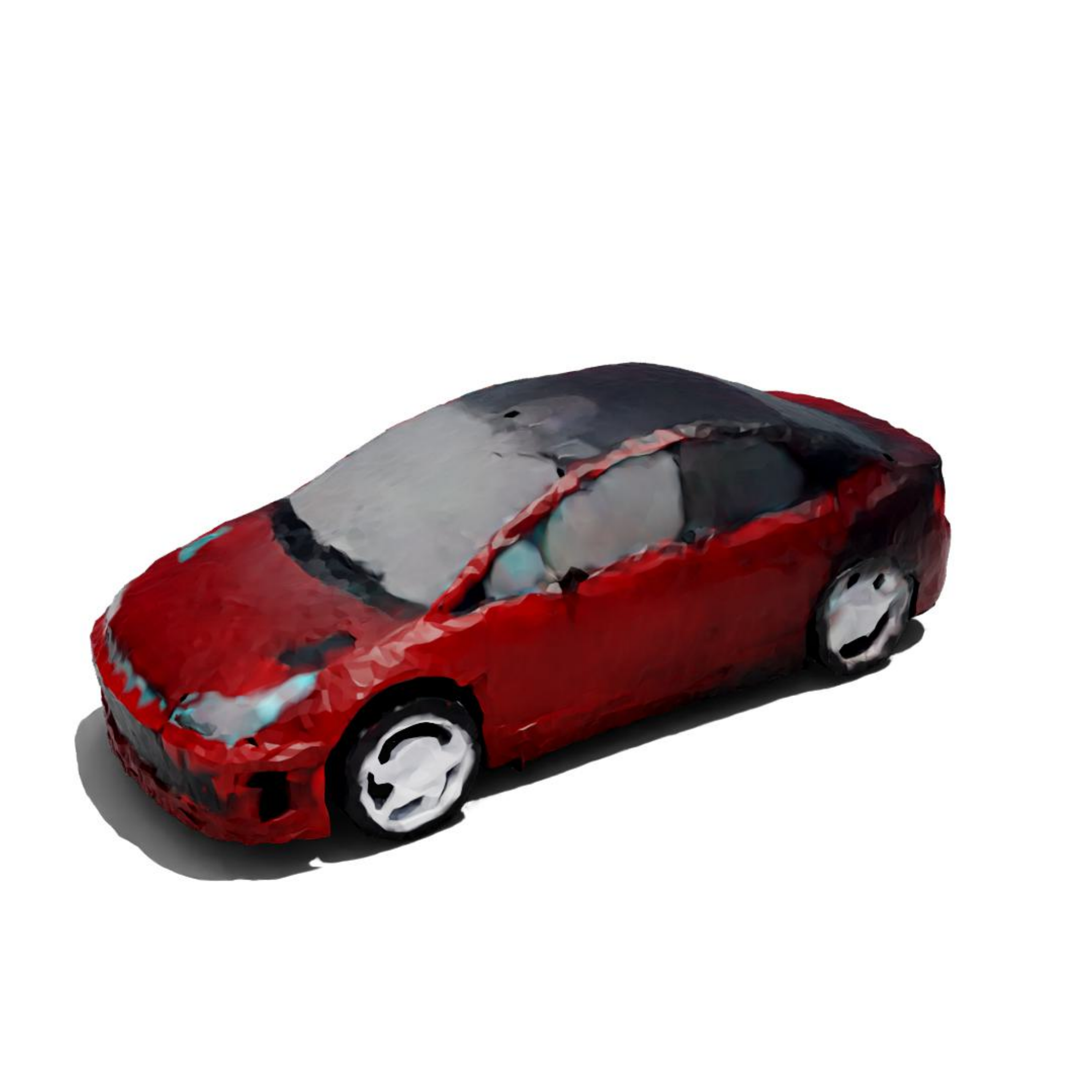}\includegraphics[width=0.16666666666666666\linewidth, trim={0 0cm 0 3cm}, clip]{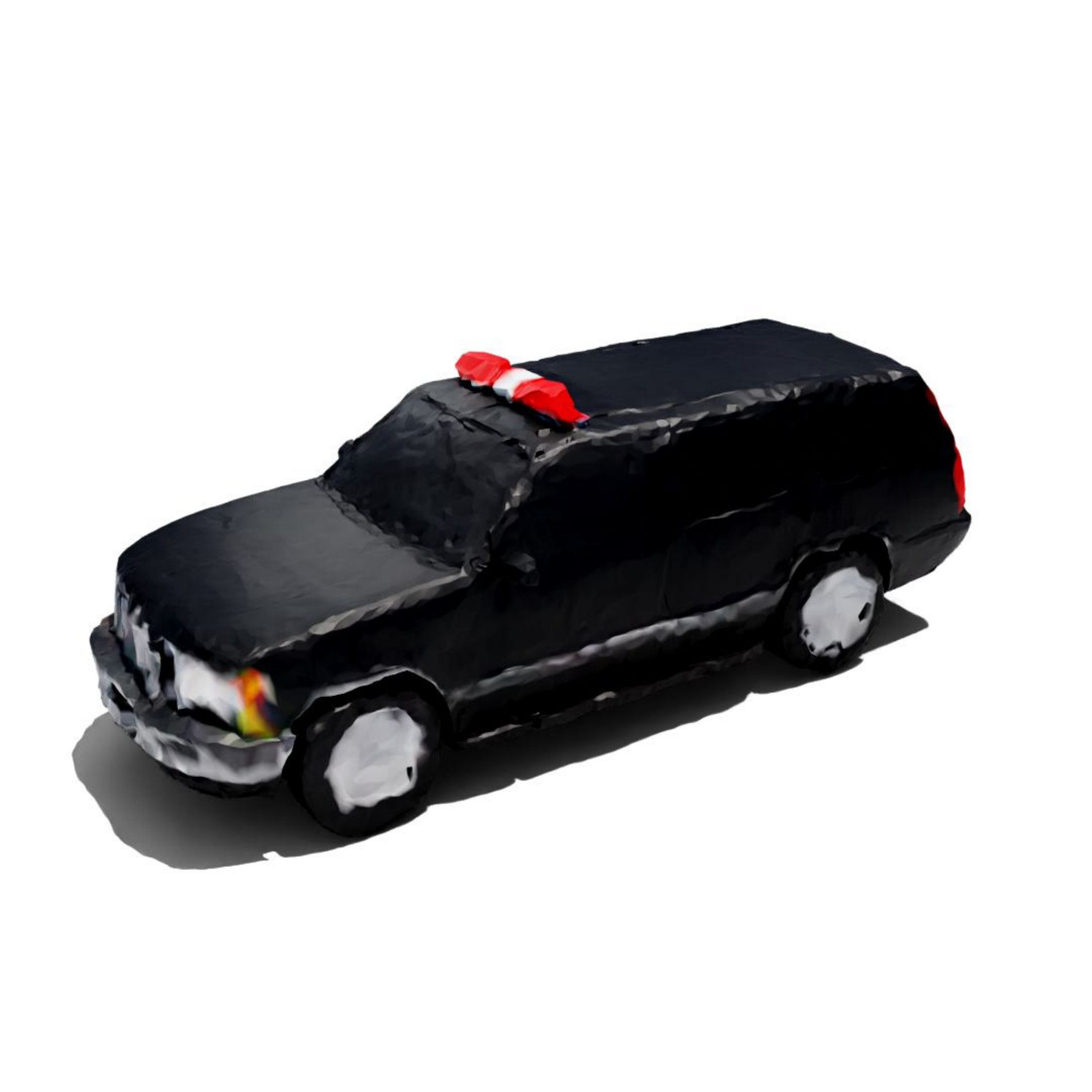}\includegraphics[width=0.16666666666666666\linewidth, trim={0 0cm 0 3cm}, clip]{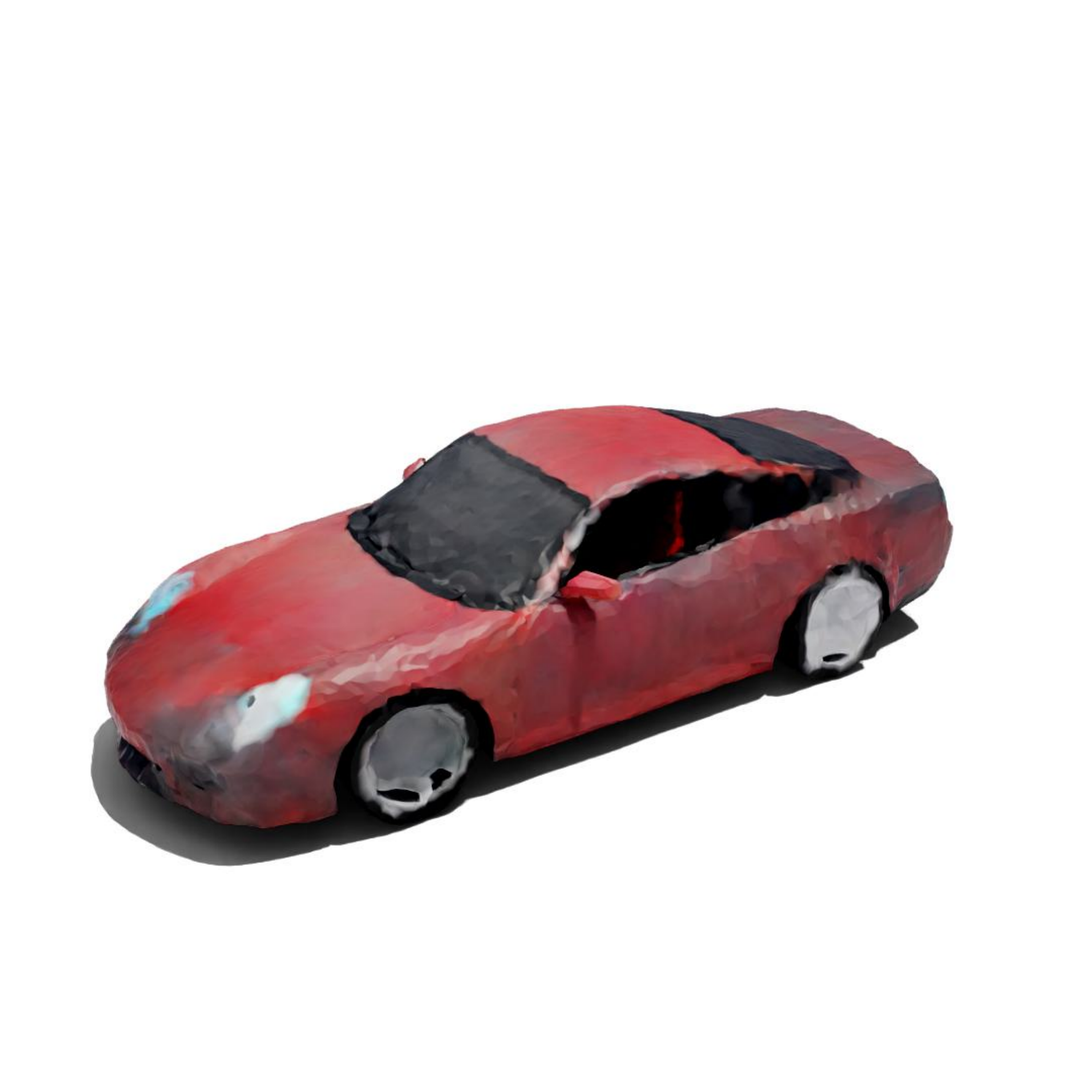}\includegraphics[width=0.16666666666666666\linewidth, trim={0 0cm 0 3cm}, clip]{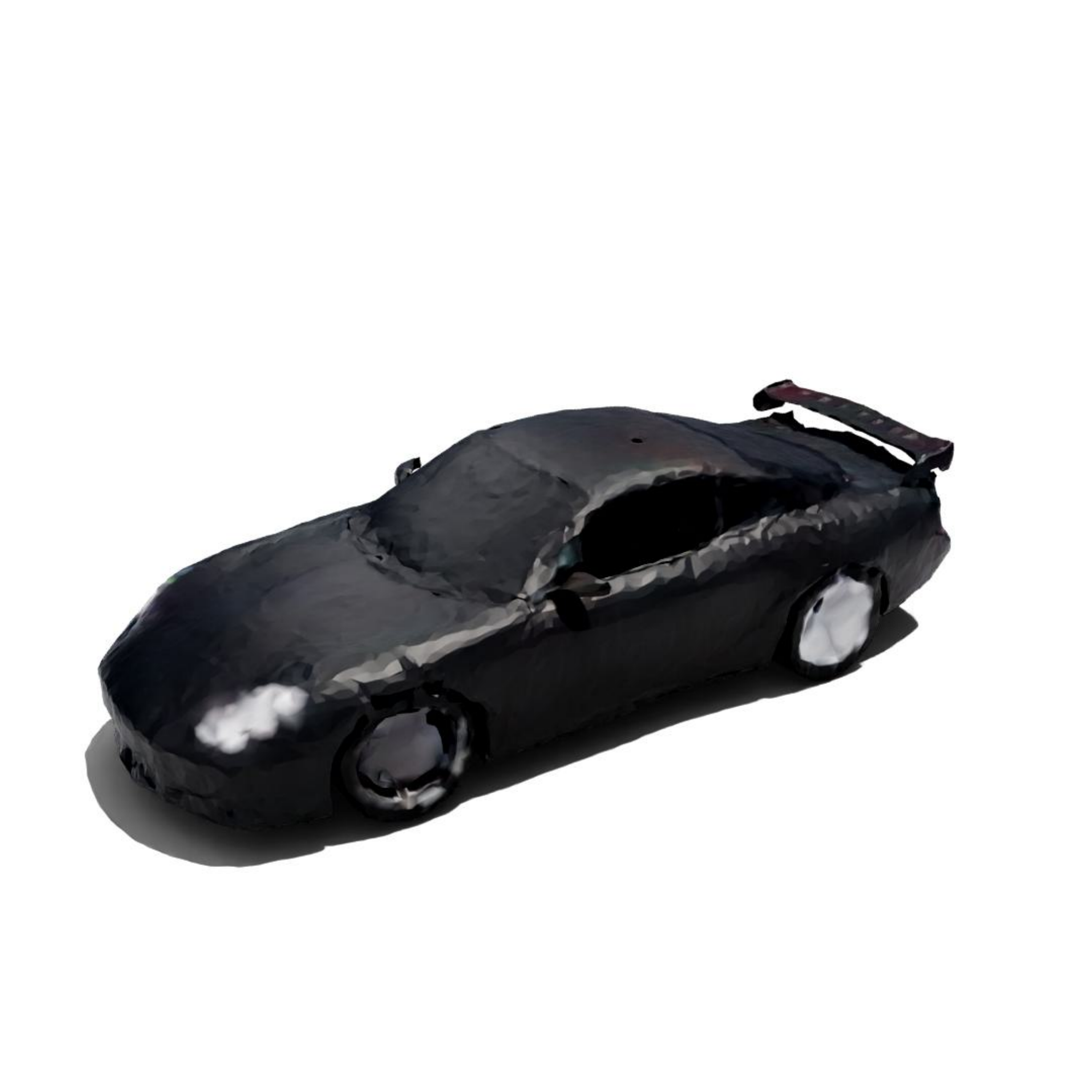}
\vspace{-0.7cm}
\caption{\textbf{Random selection of cars generated in standard resolution.} Our standard model already outputs diverse geometry, with fine structures and consistent textures.}
\label{fig:uncond:car:128}
\end{figure*}

\begin{figure*}[!ht]
\centering
\includegraphics[width=0.16666666666666666\linewidth, trim={0 0cm 0 3cm}, clip]{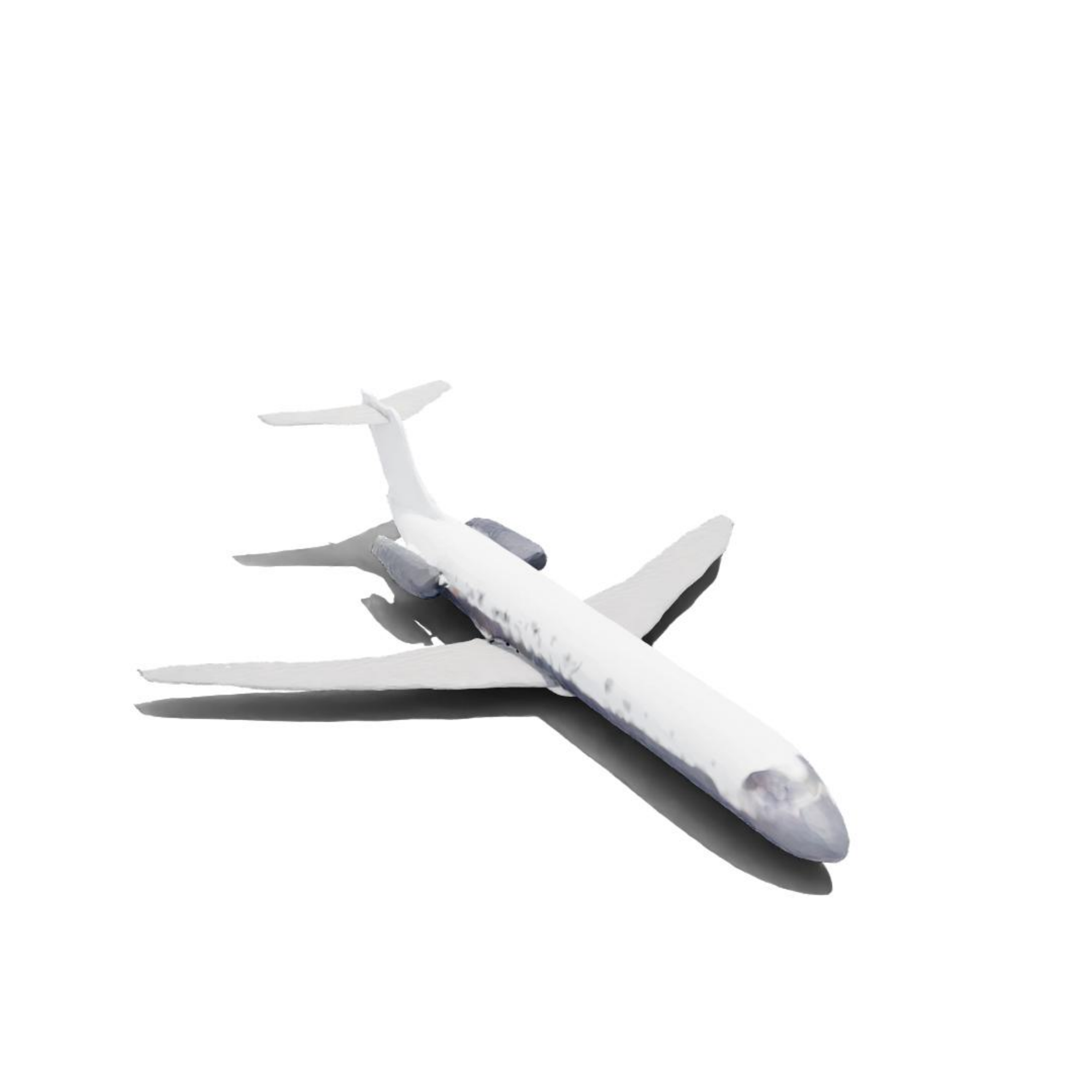}\includegraphics[width=0.16666666666666666\linewidth, trim={0 0cm 0 3cm}, clip]{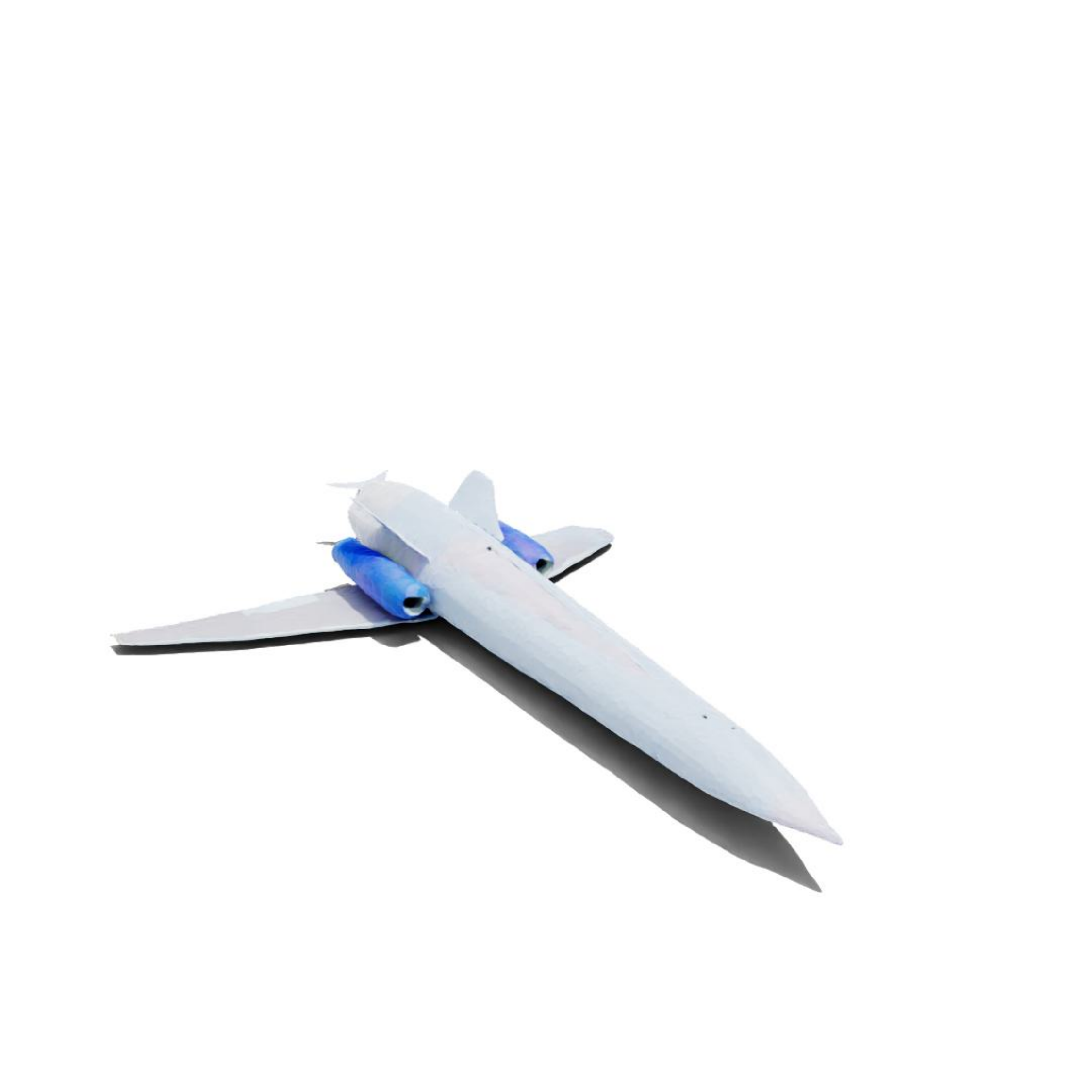}\includegraphics[width=0.16666666666666666\linewidth, trim={0 0cm 0 3cm}, clip]{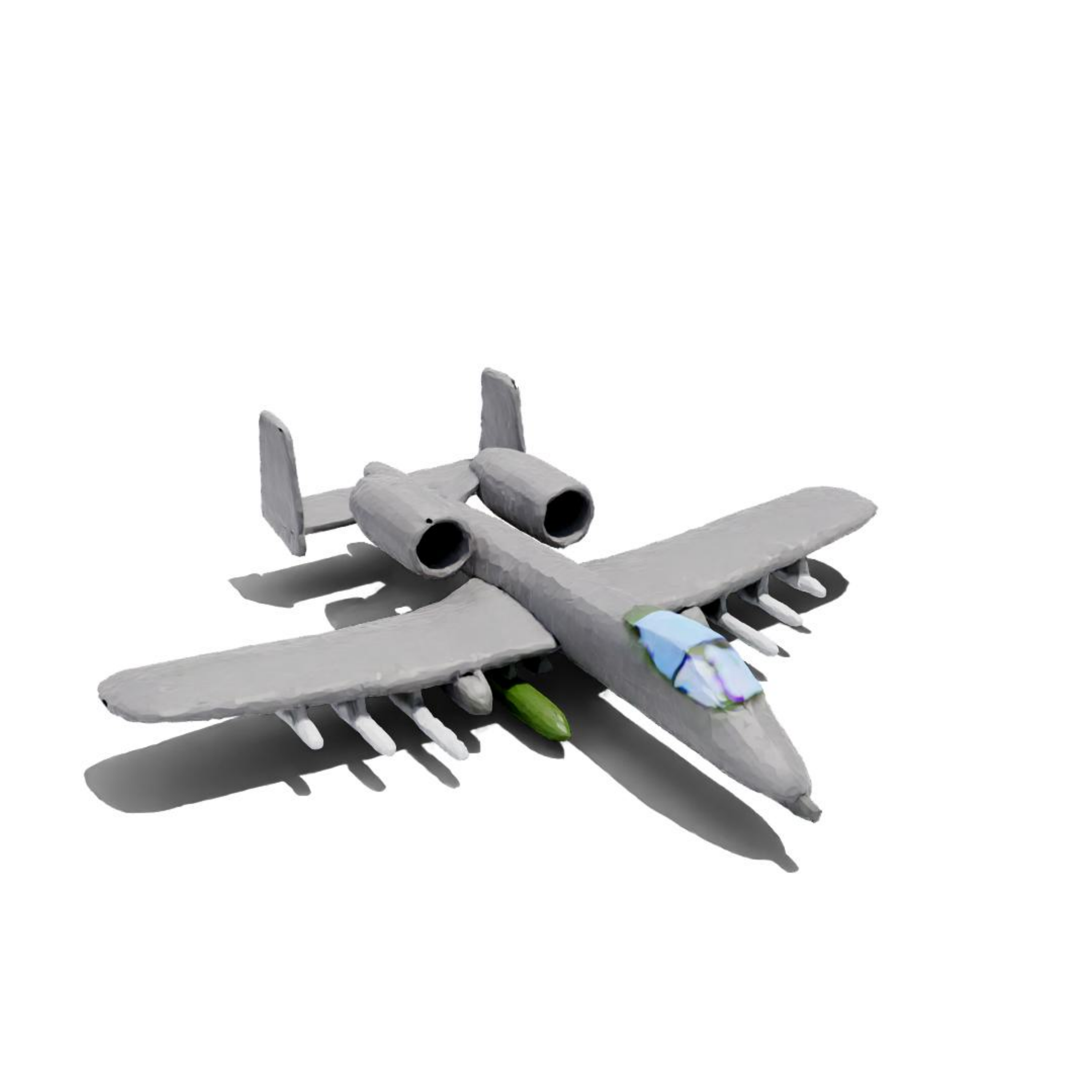}\includegraphics[width=0.16666666666666666\linewidth, trim={0 0cm 0 3cm}, clip]{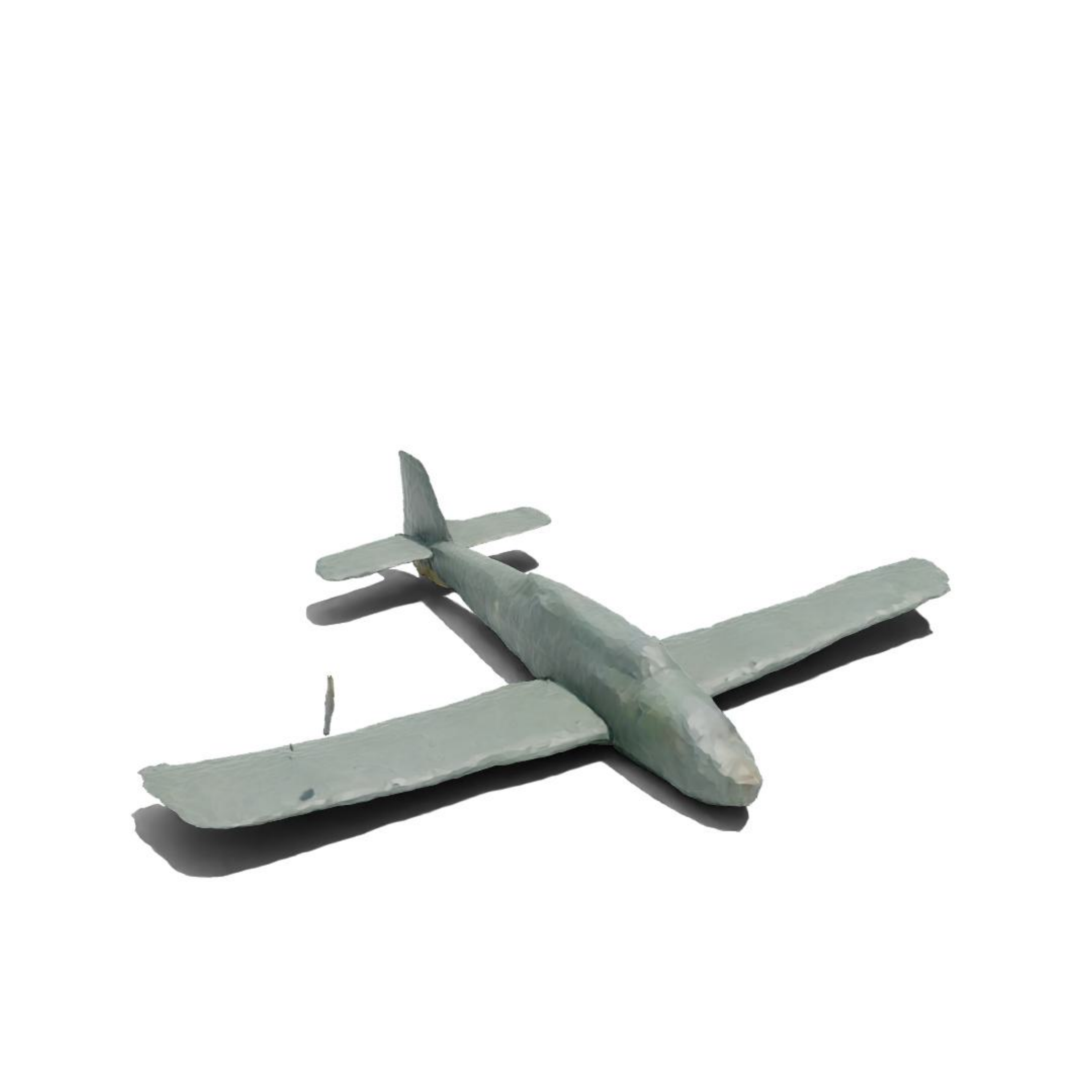}\includegraphics[width=0.16666666666666666\linewidth, trim={0 0cm 0 3cm}, clip]{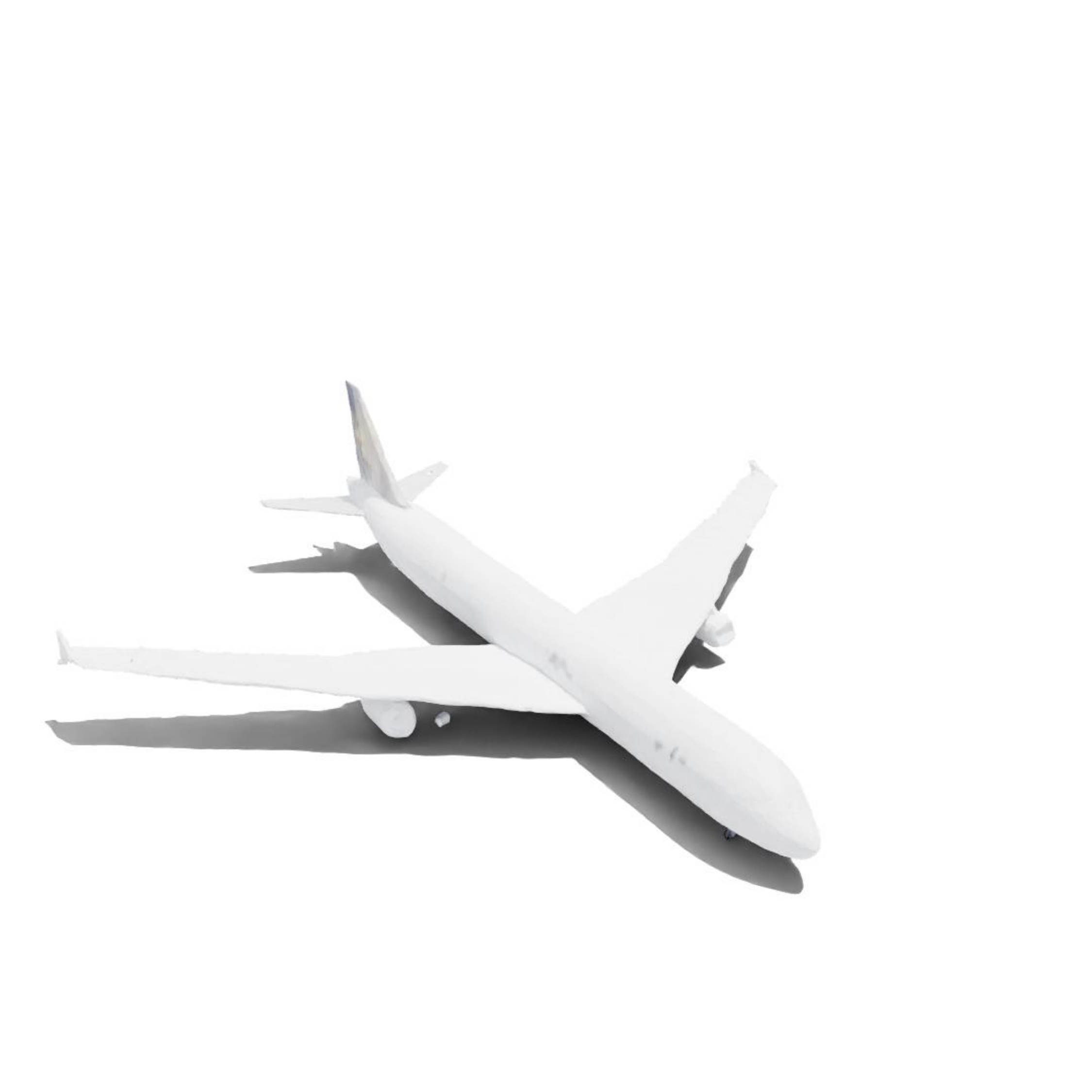}\includegraphics[width=0.16666666666666666\linewidth, trim={0 0cm 0 3cm}, clip]{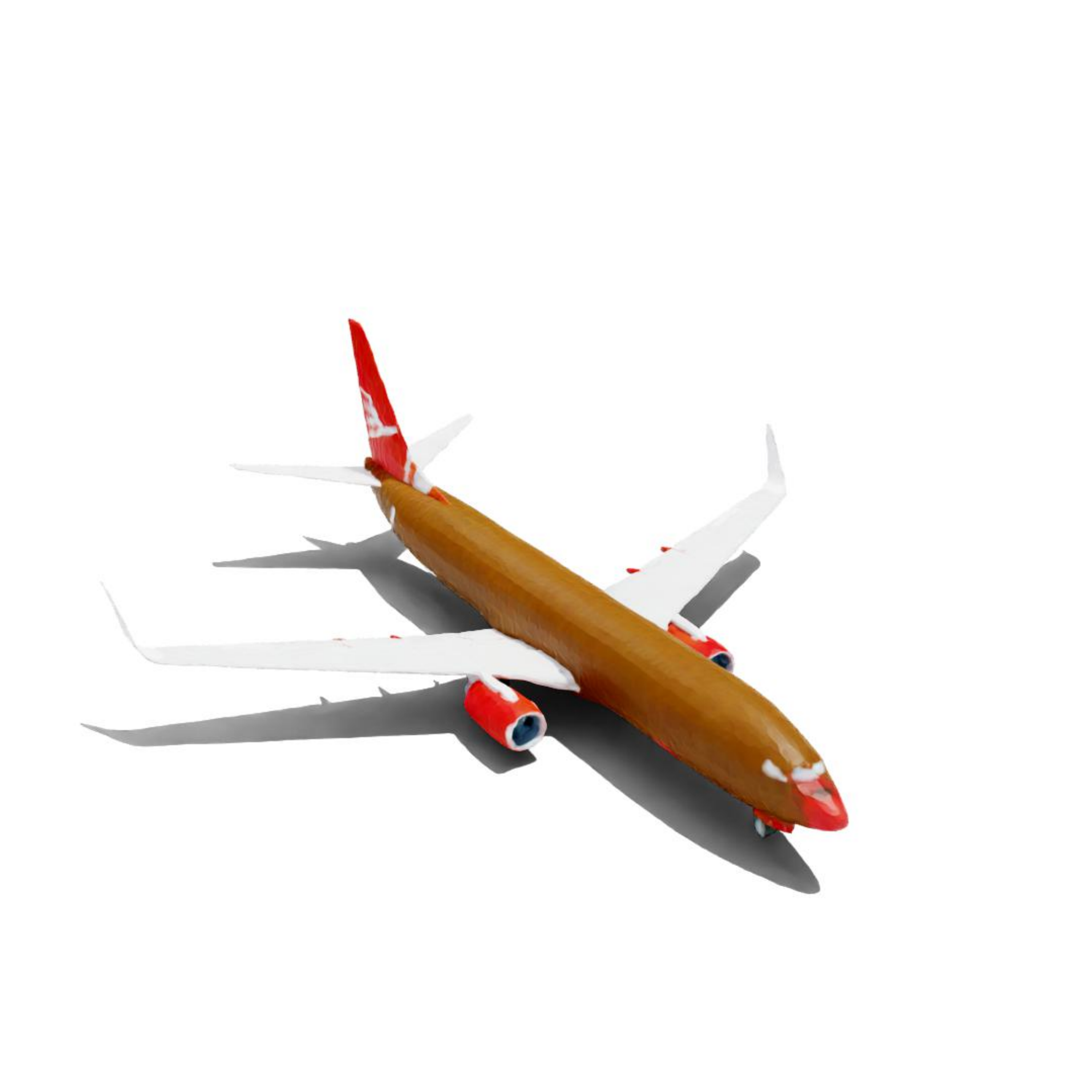}\\

\vspace{-0.43cm}
\includegraphics[width=0.16666666666666666\linewidth, trim={0 0cm 0 3cm}, clip]{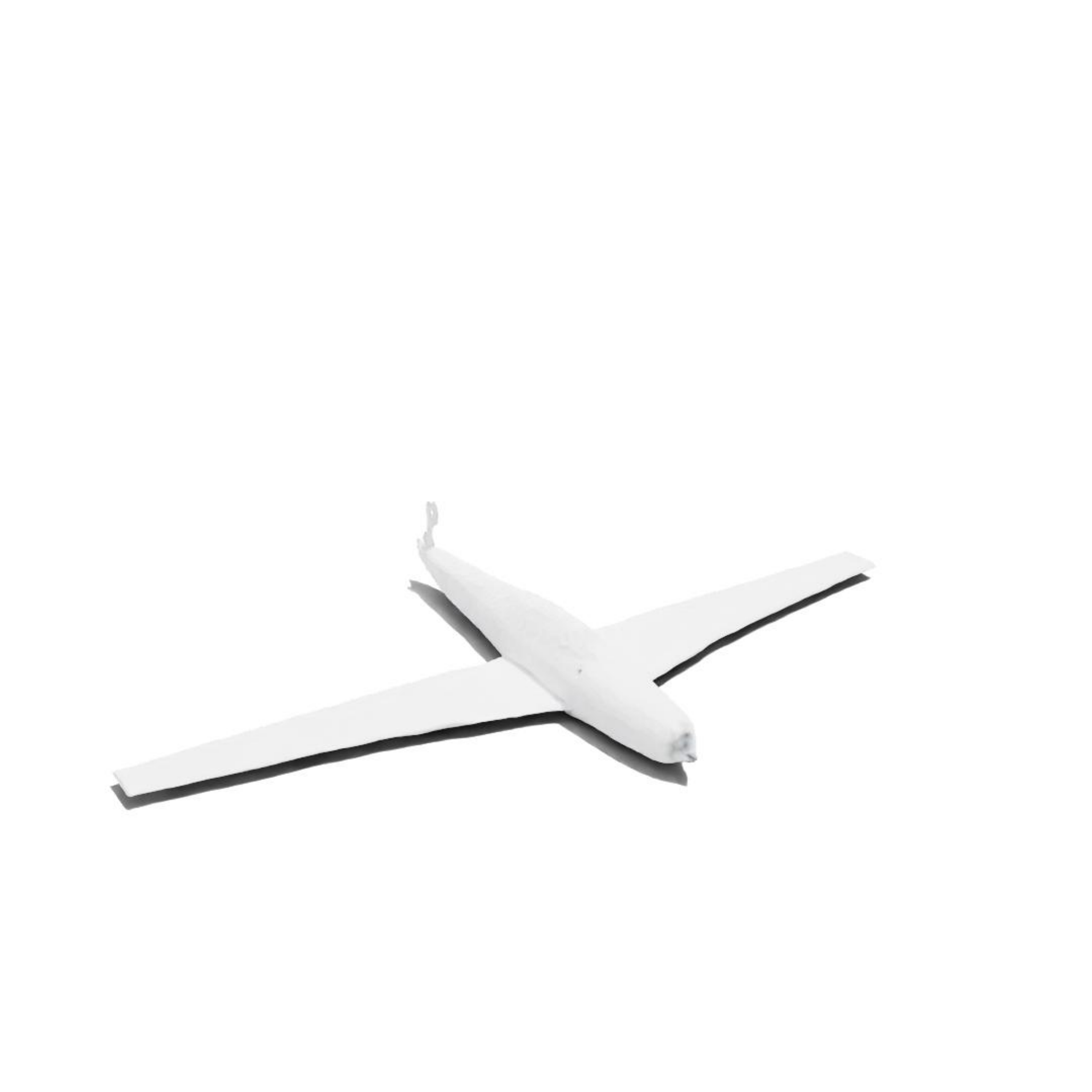}\includegraphics[width=0.16666666666666666\linewidth, trim={0 0cm 0 3cm}, clip]{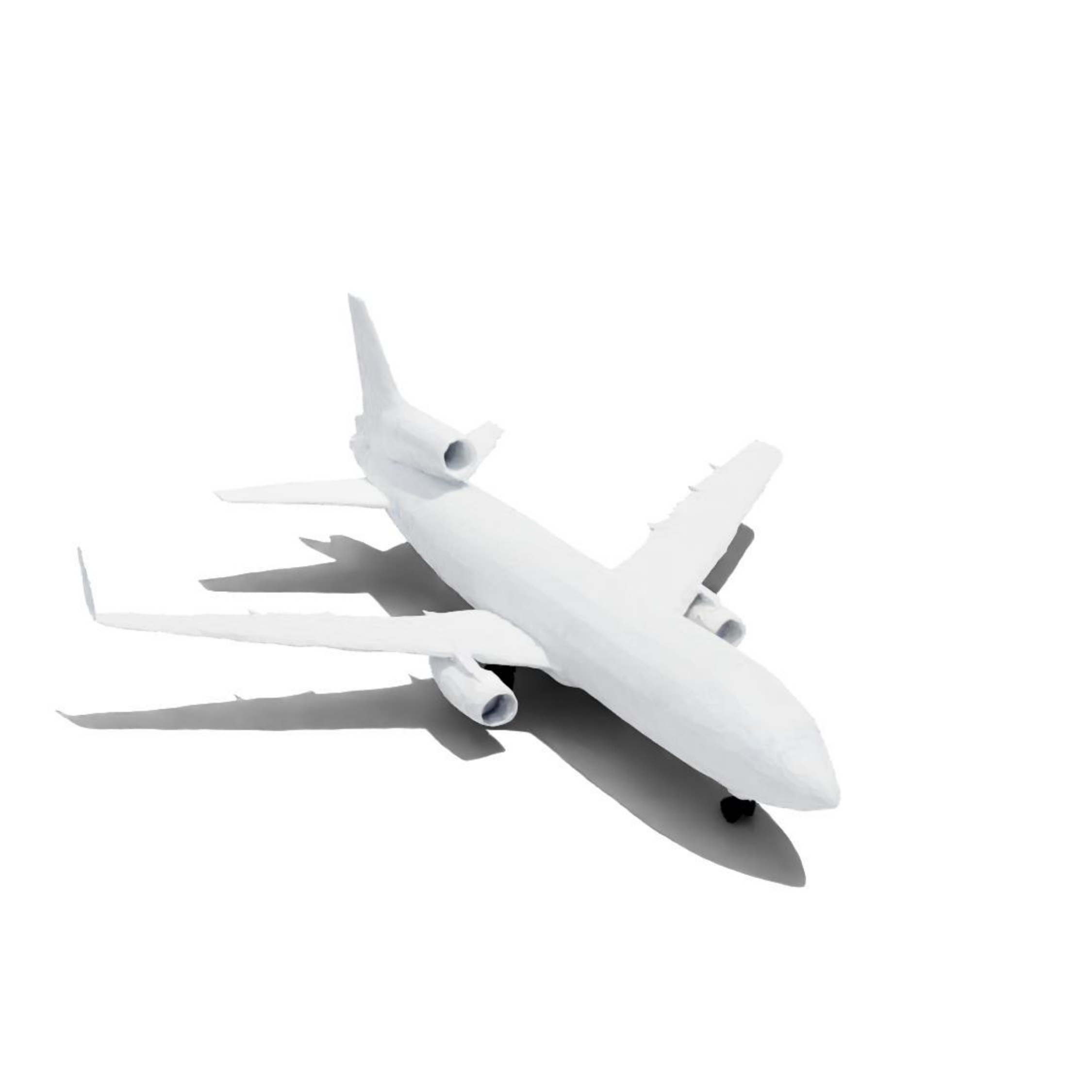}\includegraphics[width=0.16666666666666666\linewidth, trim={0 0cm 0 3cm}, clip]{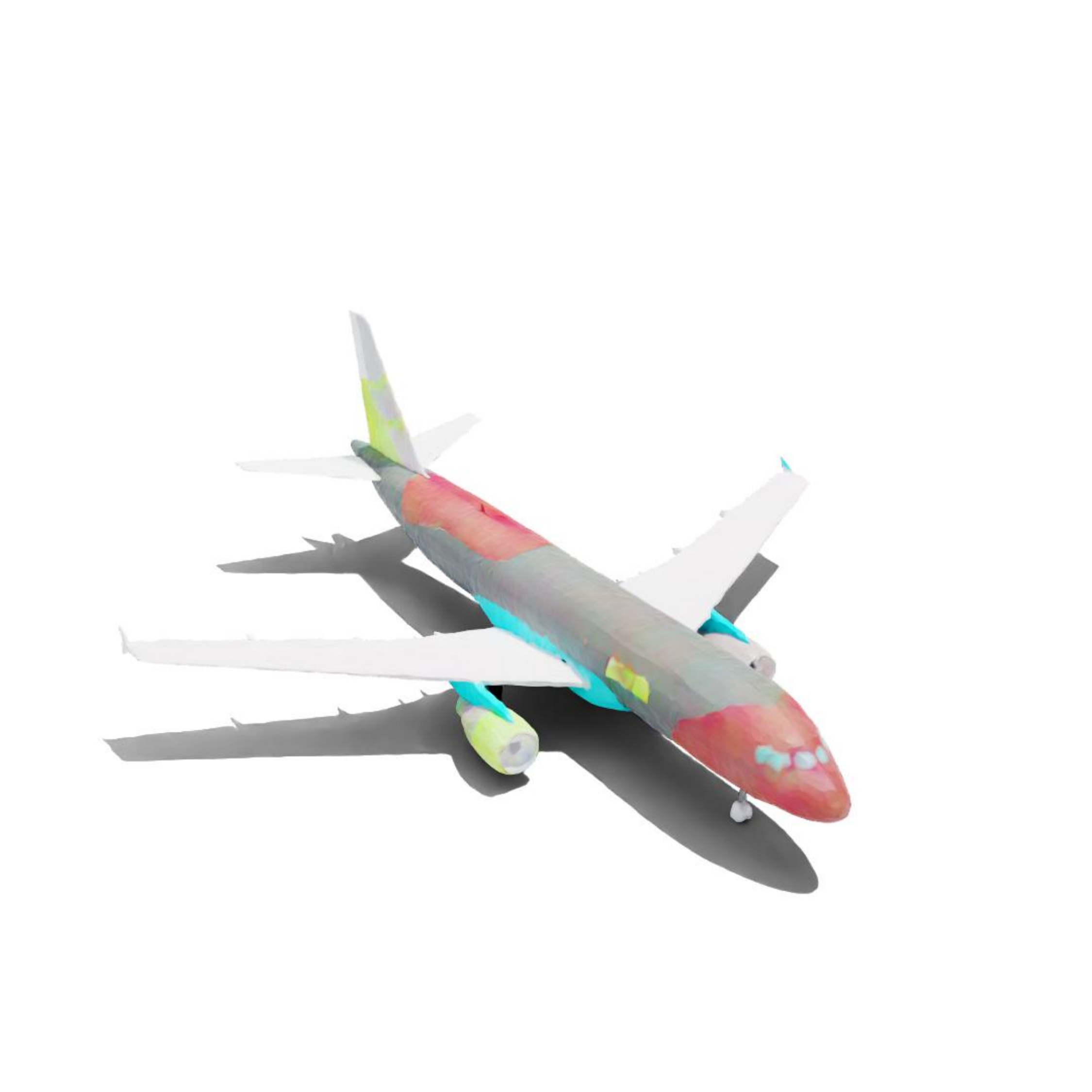}\includegraphics[width=0.16666666666666666\linewidth, trim={0 0cm 0 3cm}, clip]{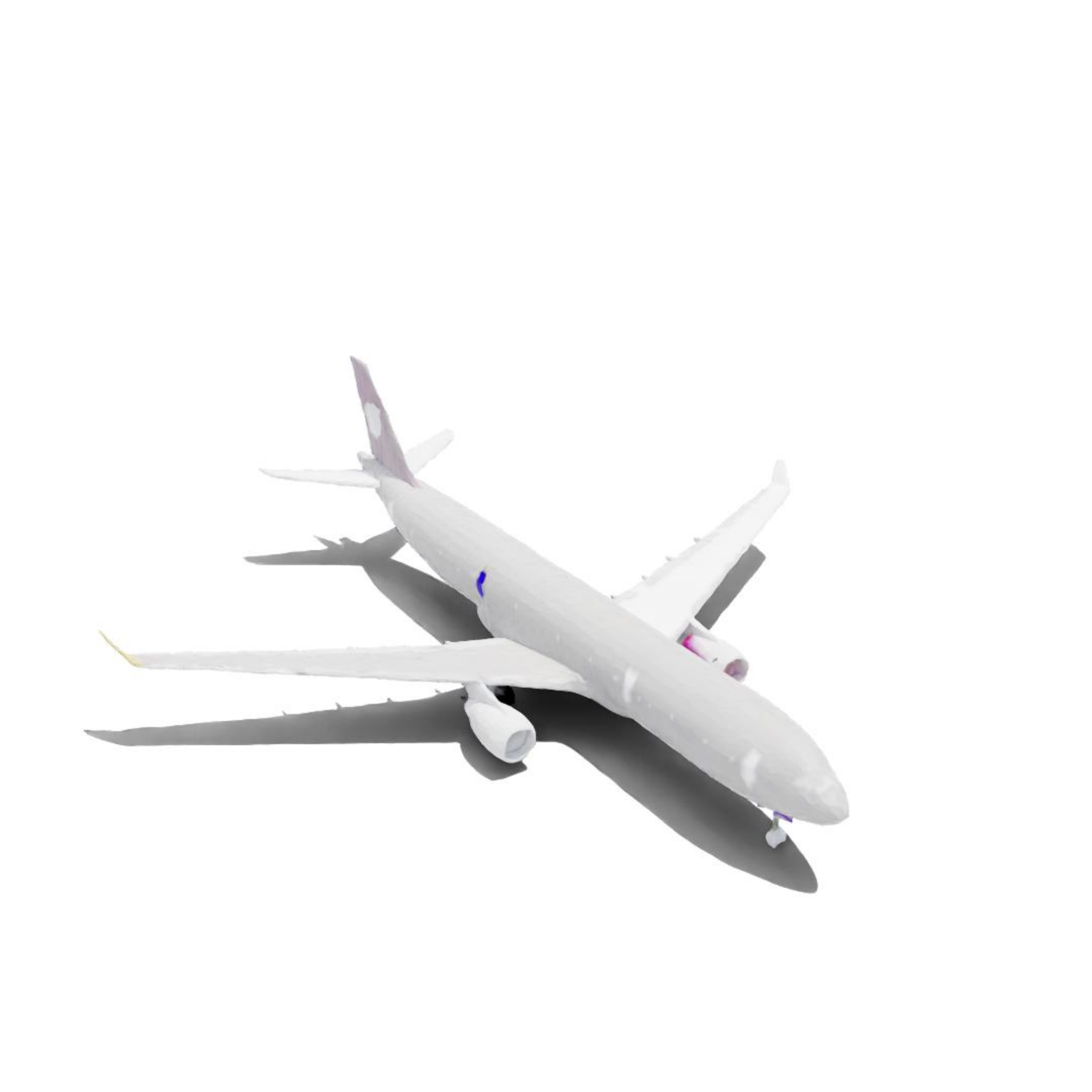}\includegraphics[width=0.16666666666666666\linewidth, trim={0 0cm 0 3cm}, clip]{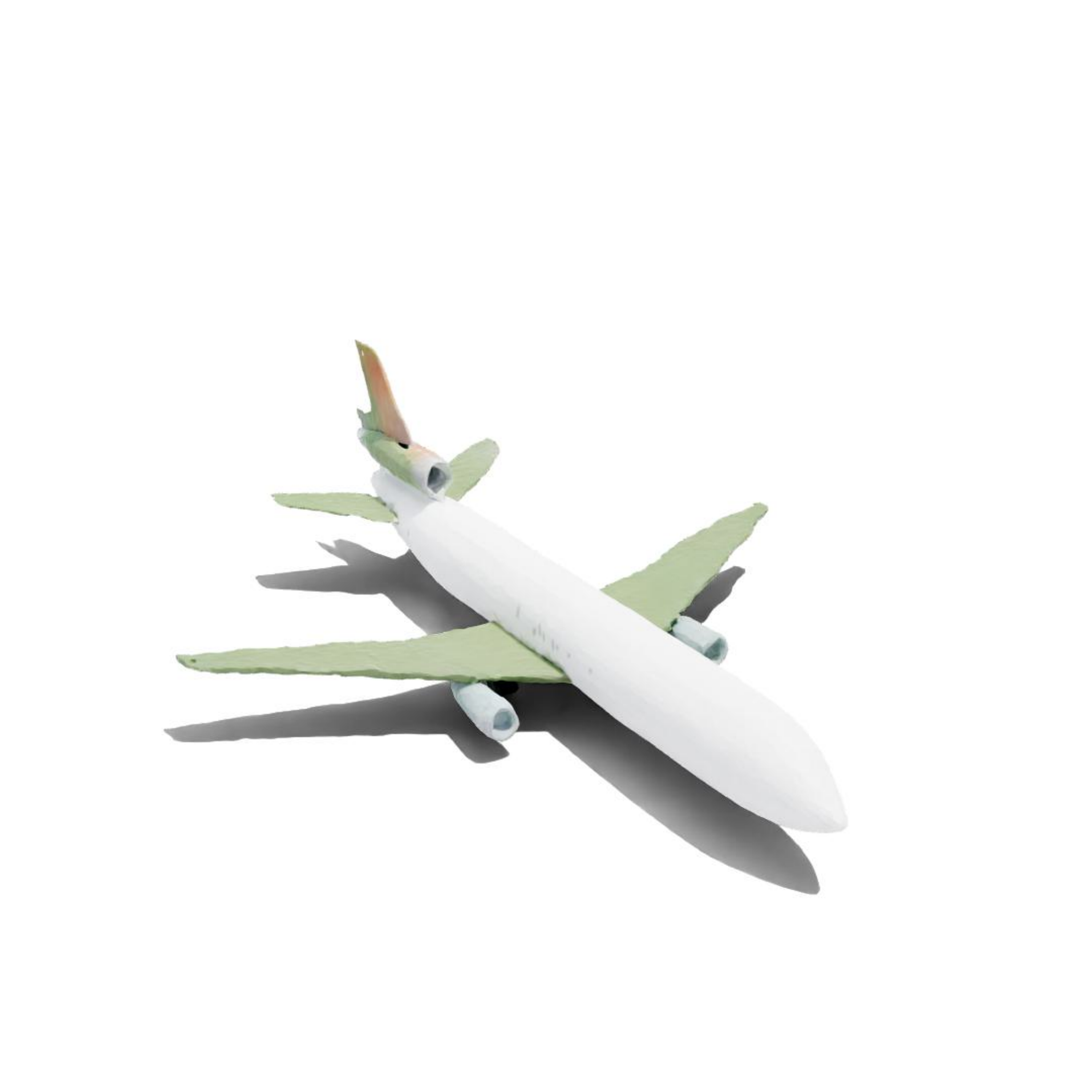}\includegraphics[width=0.16666666666666666\linewidth, trim={0 0cm 0 3cm}, clip]{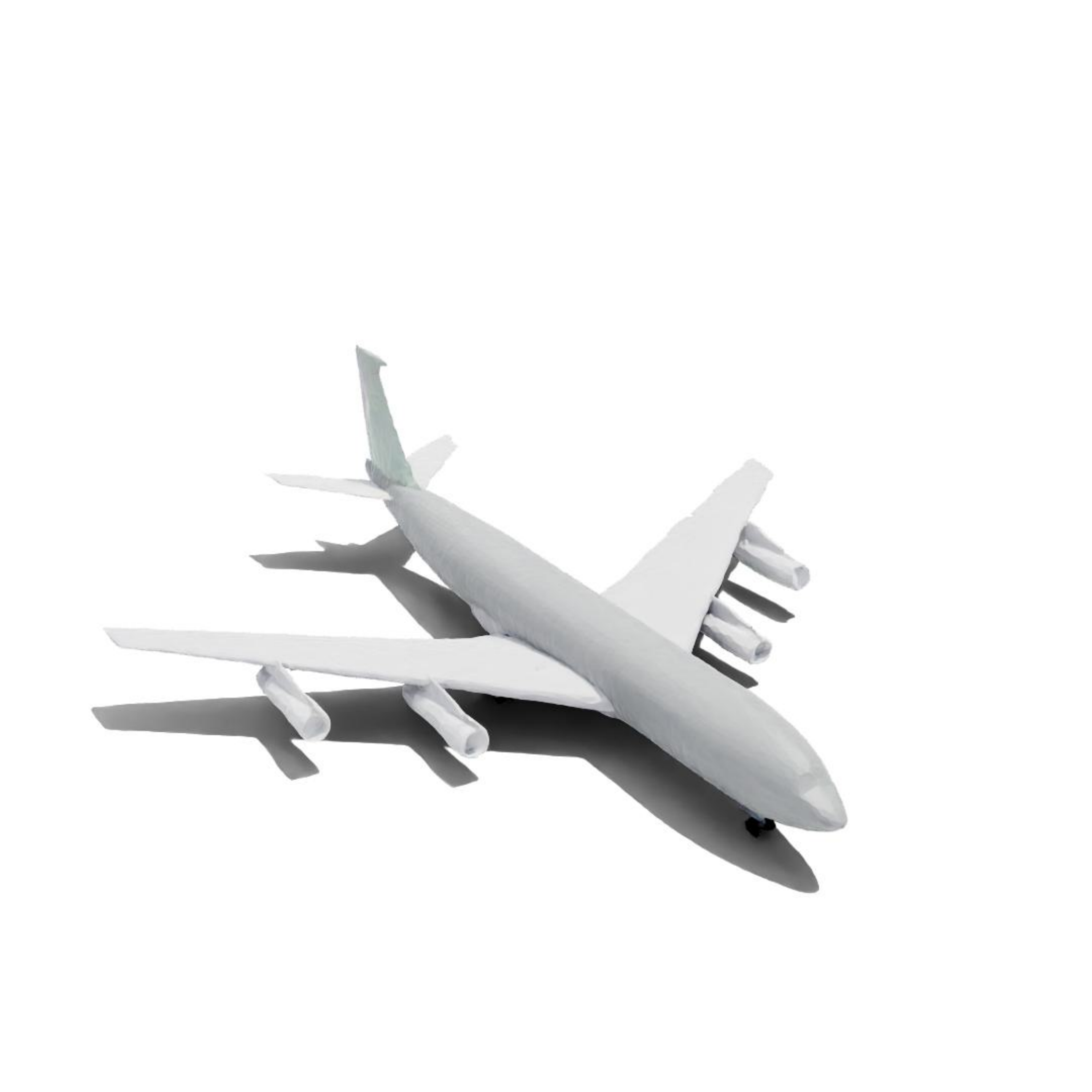}\\

\vspace{-0.43cm}
\includegraphics[width=0.16666666666666666\linewidth, trim={0 0cm 0 3cm}, clip]{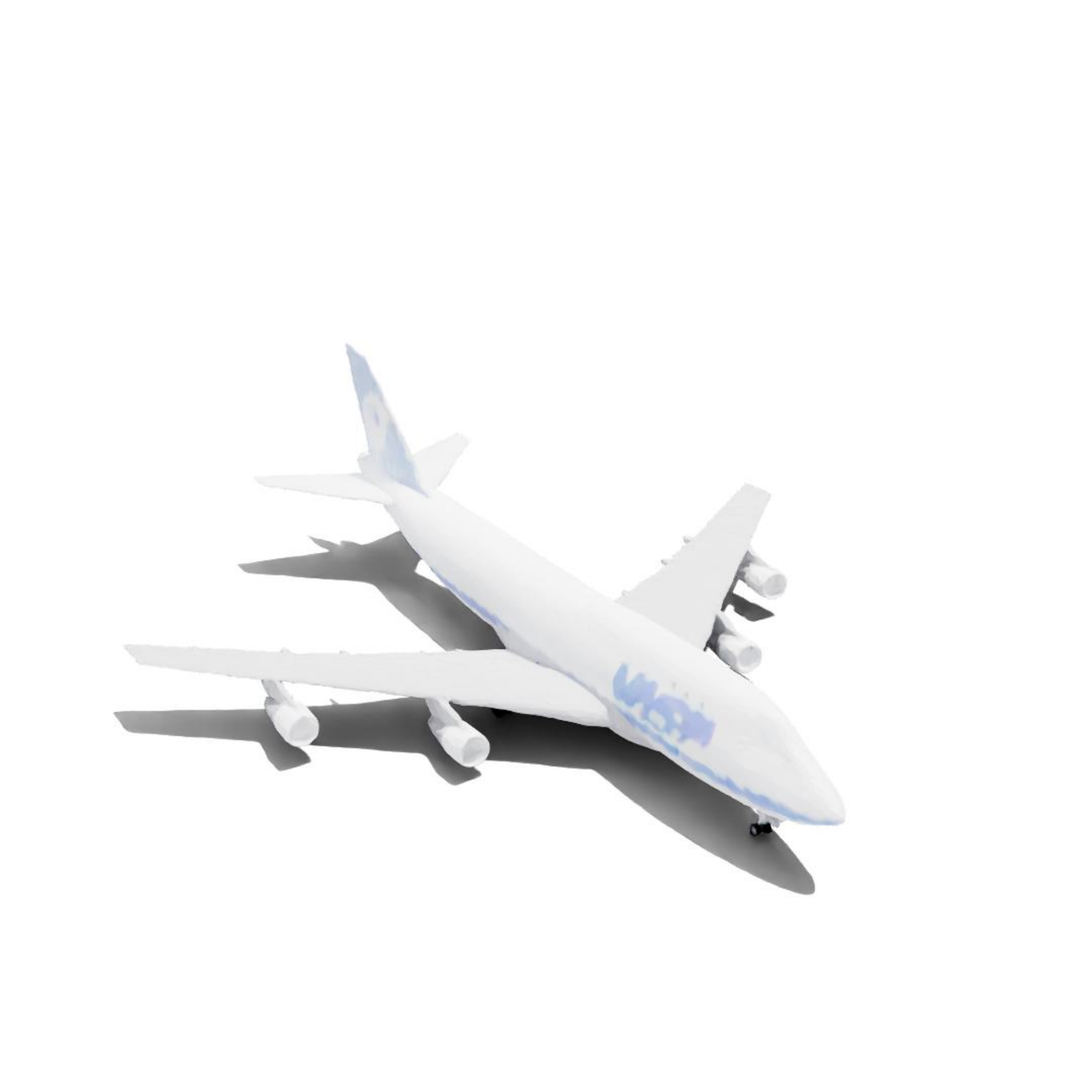}\includegraphics[width=0.16666666666666666\linewidth, trim={0 0cm 0 3cm}, clip]{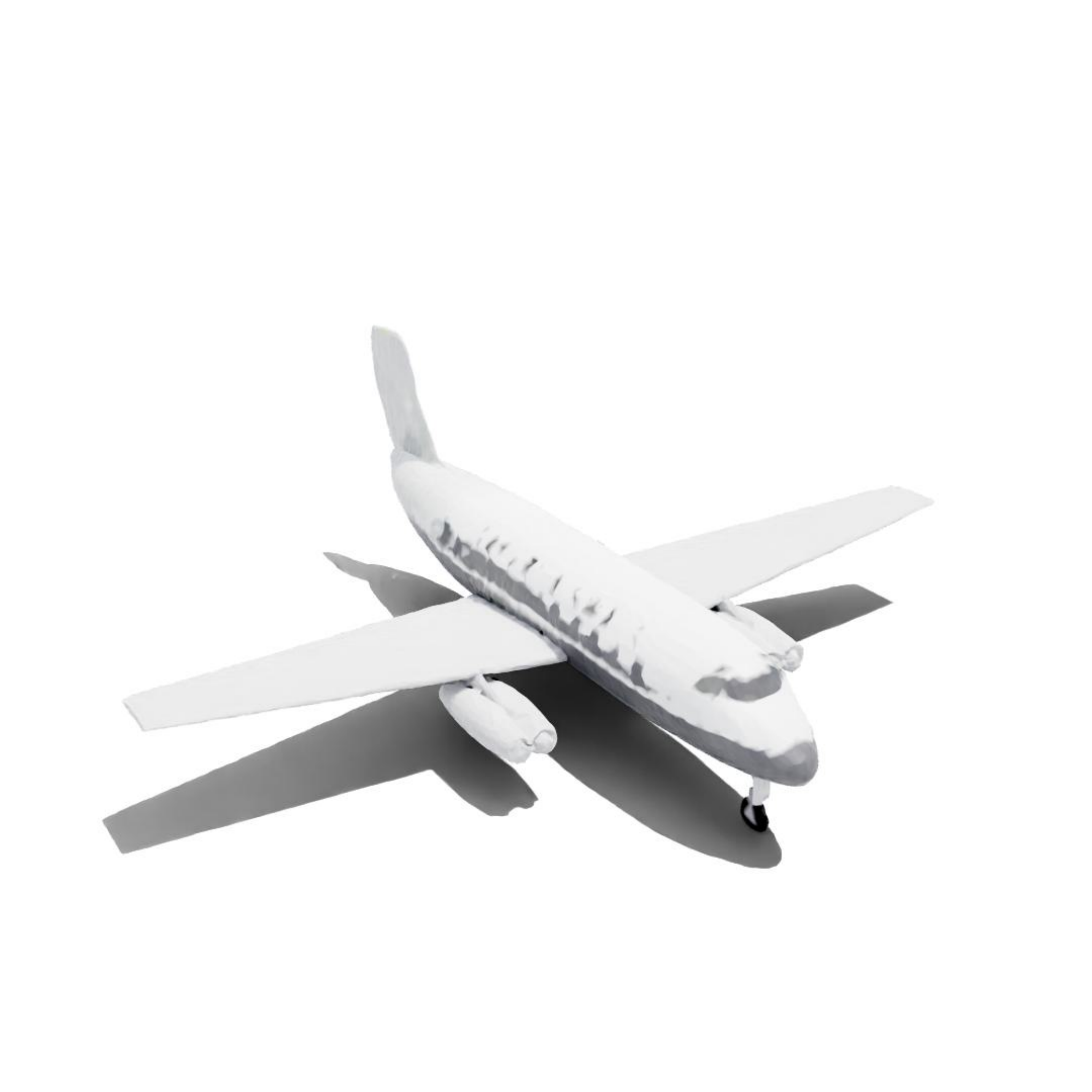}\includegraphics[width=0.16666666666666666\linewidth, trim={0 0cm 0 3cm}, clip]{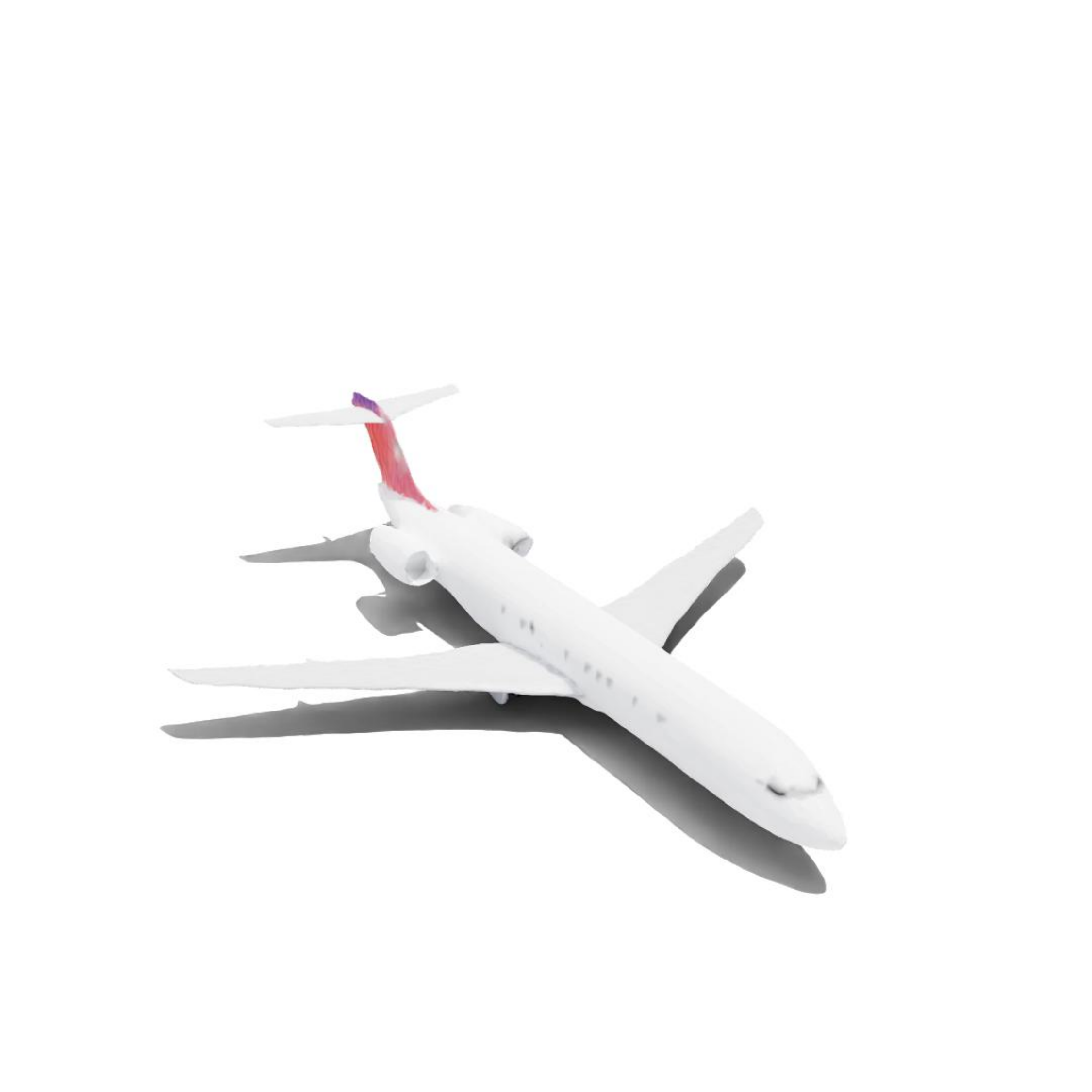}\includegraphics[width=0.16666666666666666\linewidth, trim={0 0cm 0 3cm}, clip]{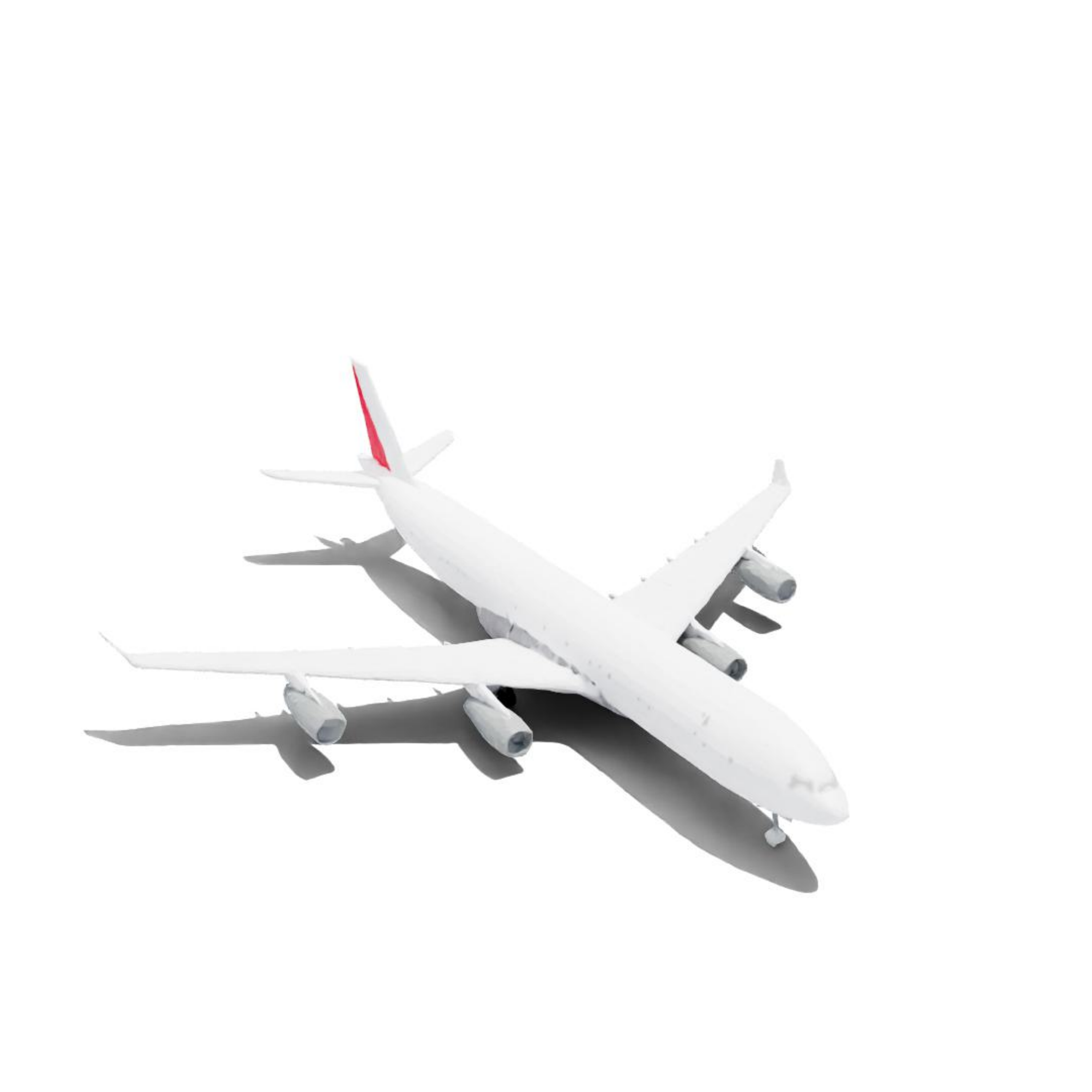}\includegraphics[width=0.16666666666666666\linewidth, trim={0 0cm 0 3cm}, clip]{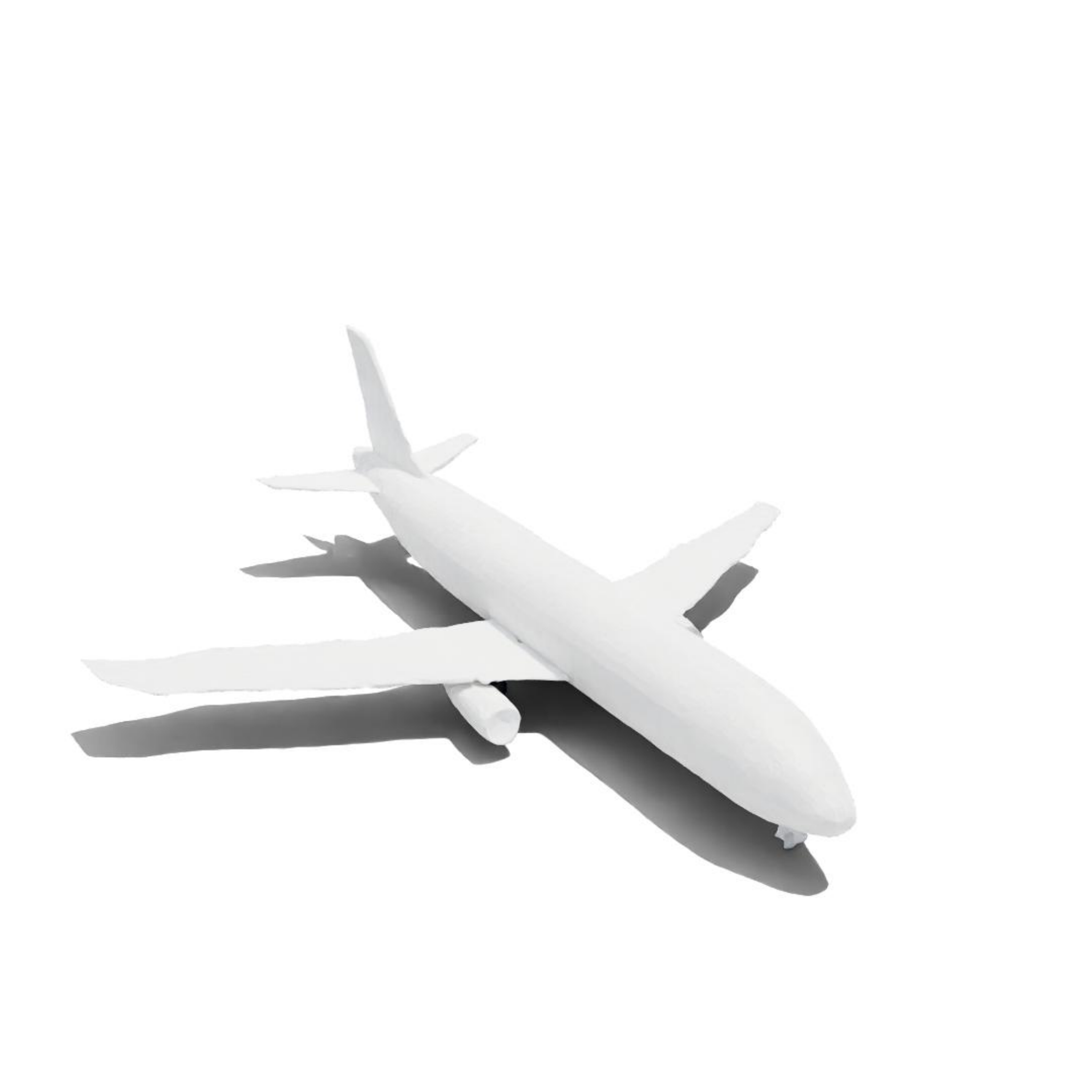}\includegraphics[width=0.16666666666666666\linewidth, trim={0 0cm 0 3cm}, clip]{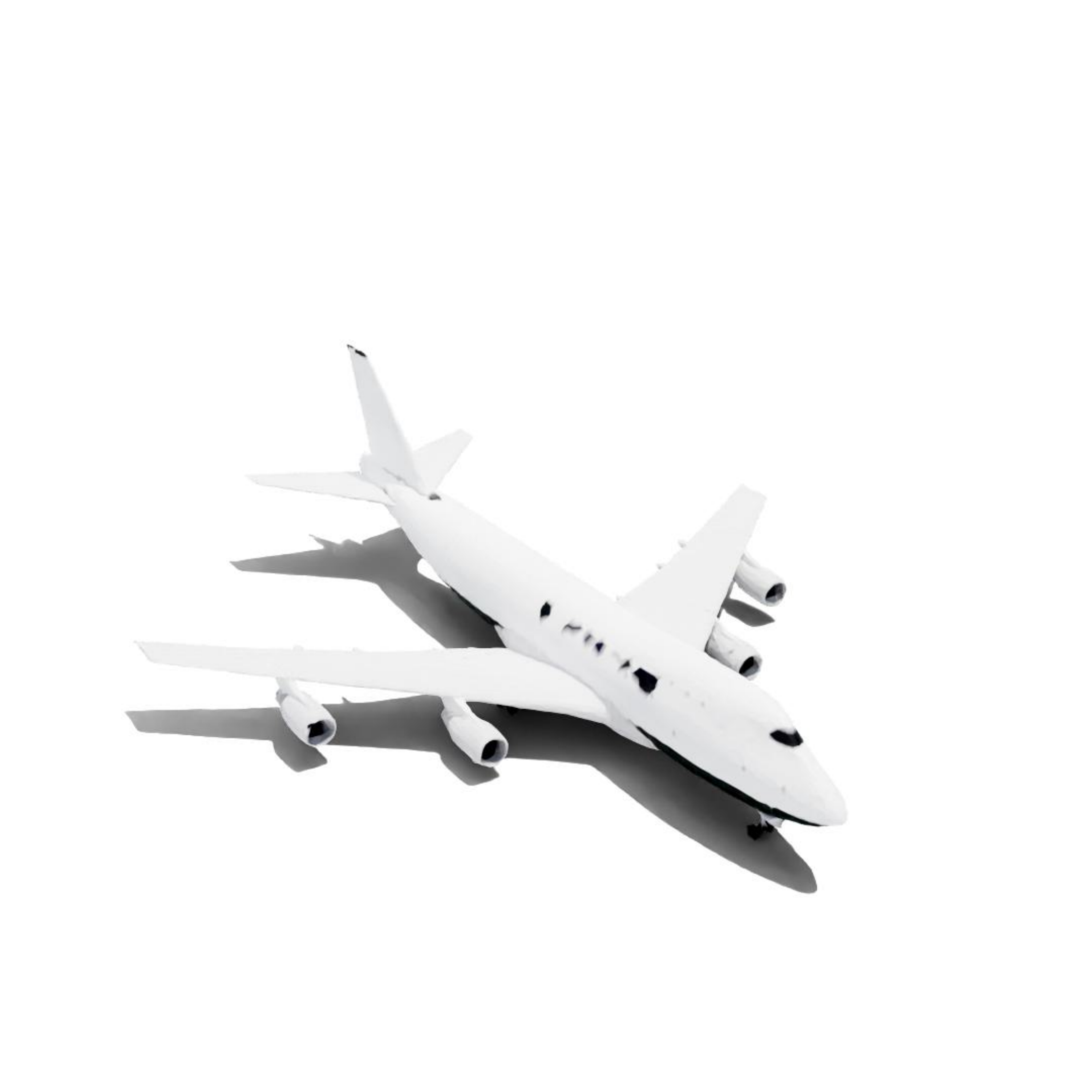}\\

\vspace{-0.43cm}
\includegraphics[width=0.16666666666666666\linewidth, trim={0 0cm 0 3cm}, clip]{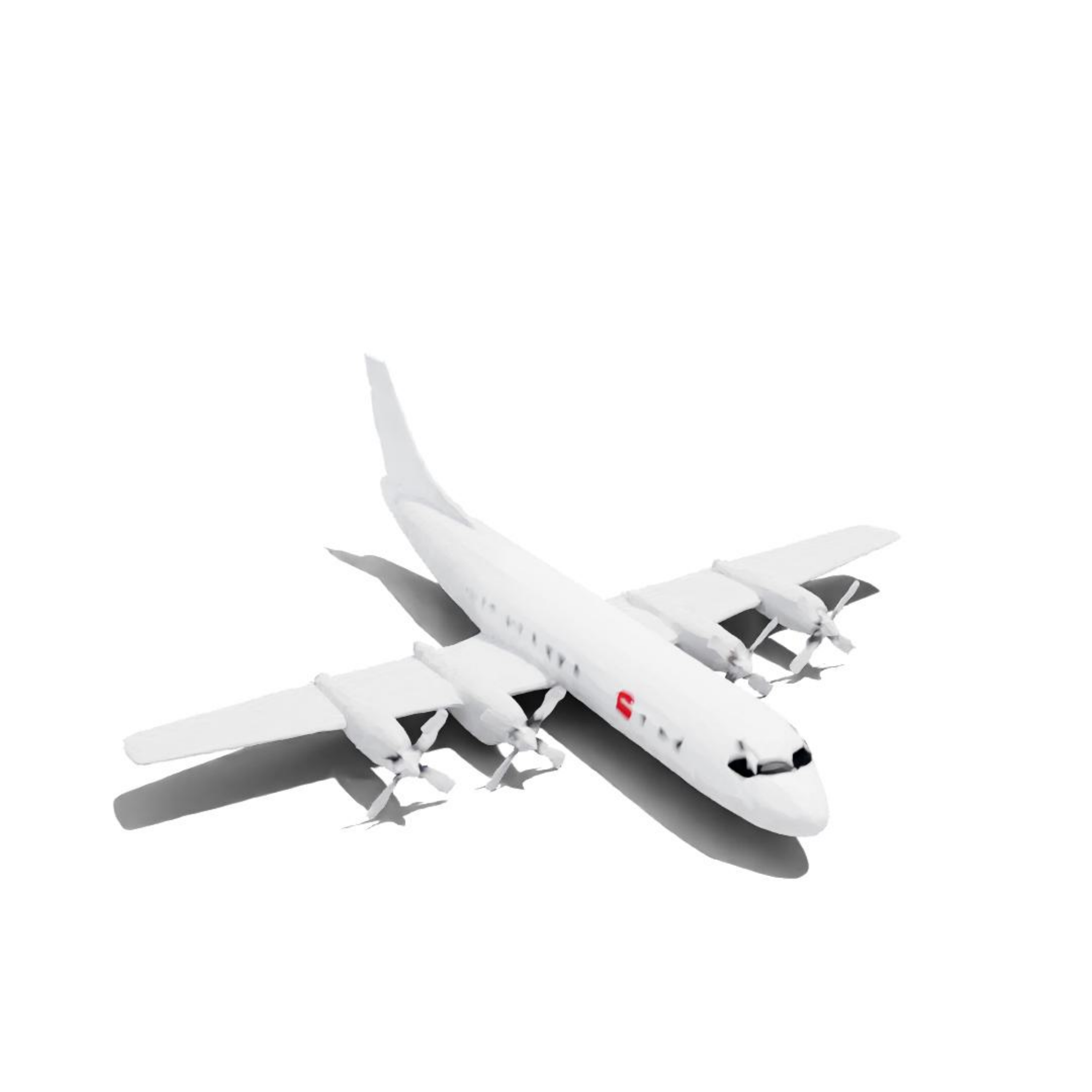}\includegraphics[width=0.16666666666666666\linewidth, trim={0 0cm 0 3cm}, clip]{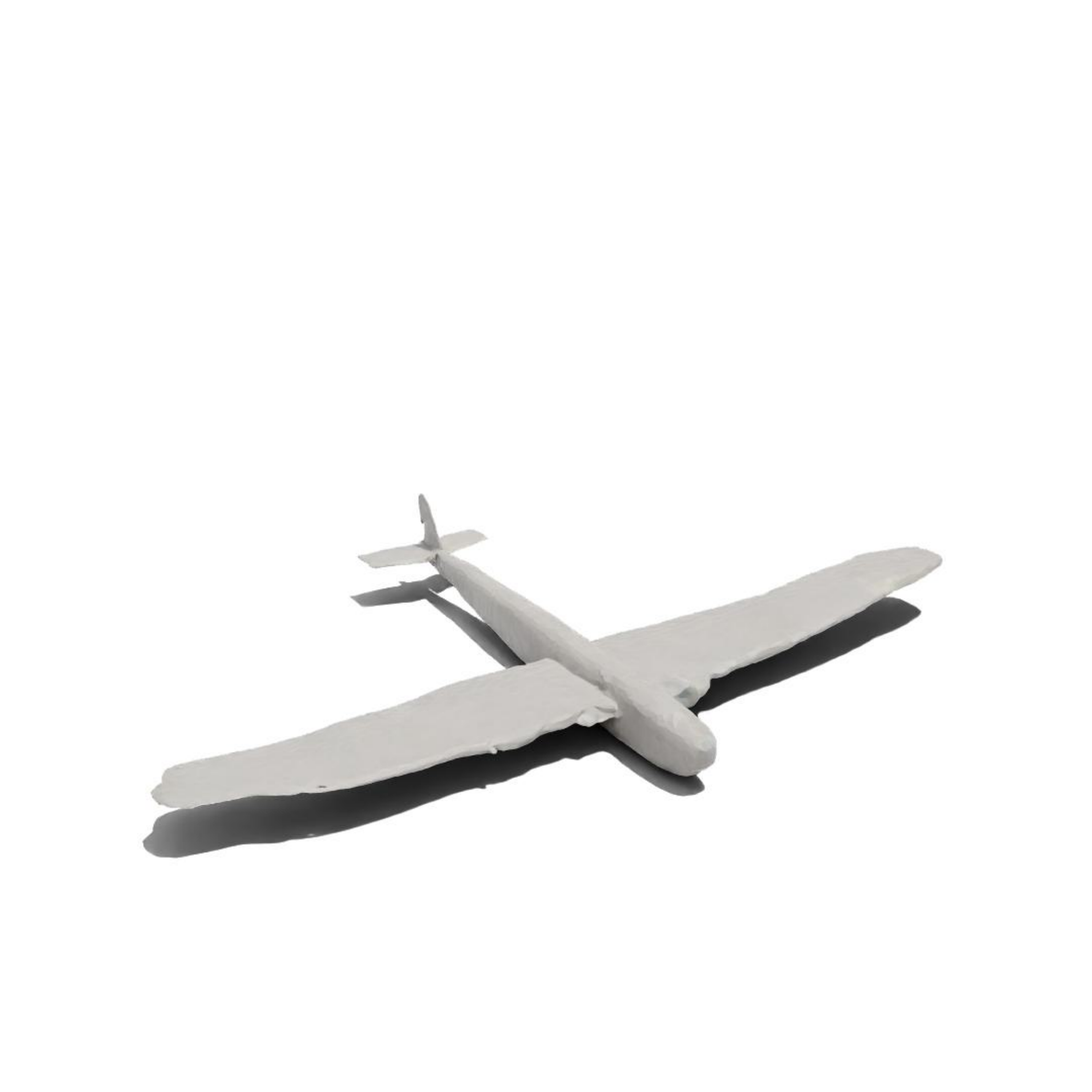}\includegraphics[width=0.16666666666666666\linewidth, trim={0 0cm 0 3cm}, clip]{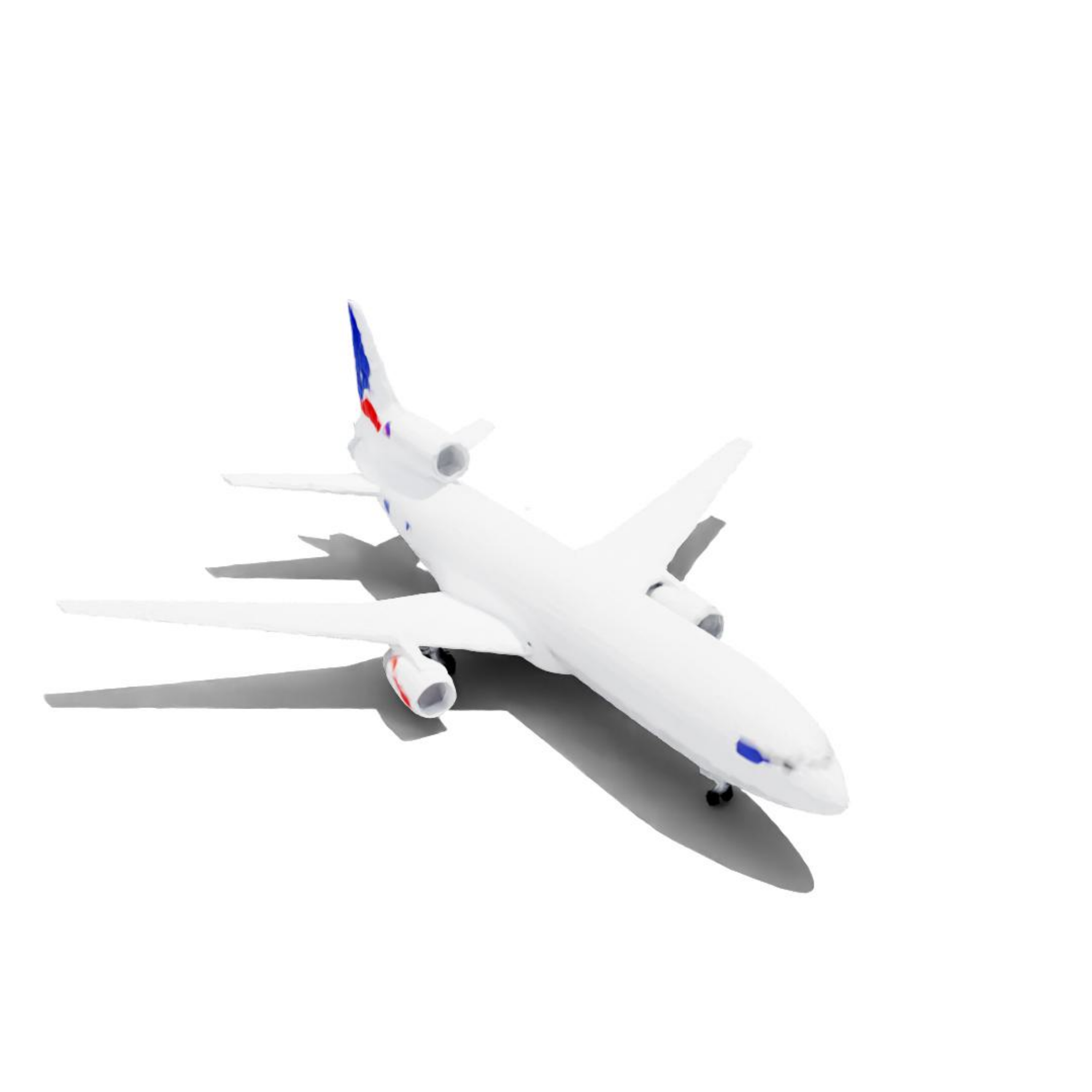}\includegraphics[width=0.16666666666666666\linewidth, trim={0 0cm 0 3cm}, clip]{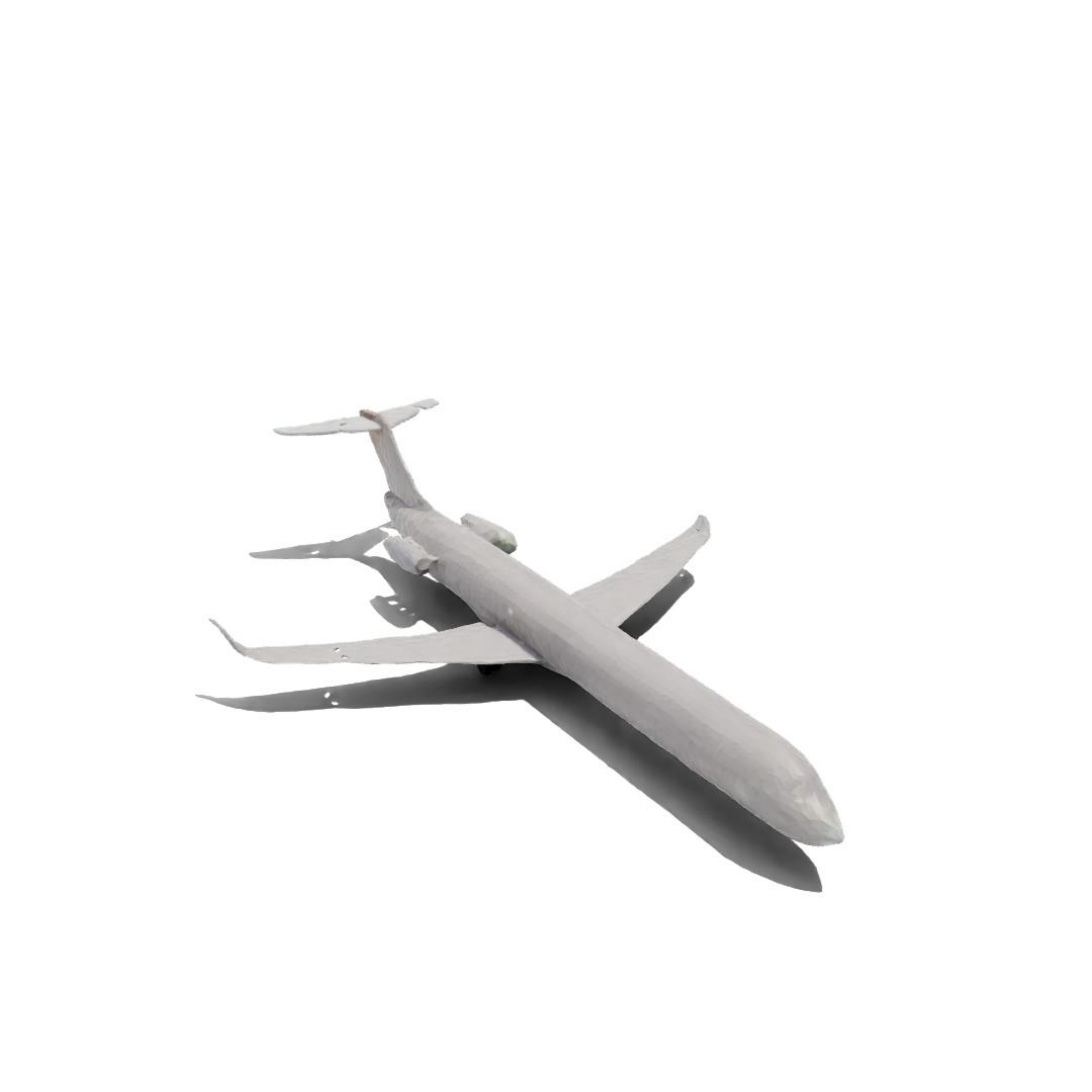}\includegraphics[width=0.16666666666666666\linewidth, trim={0 0cm 0 3cm}, clip]{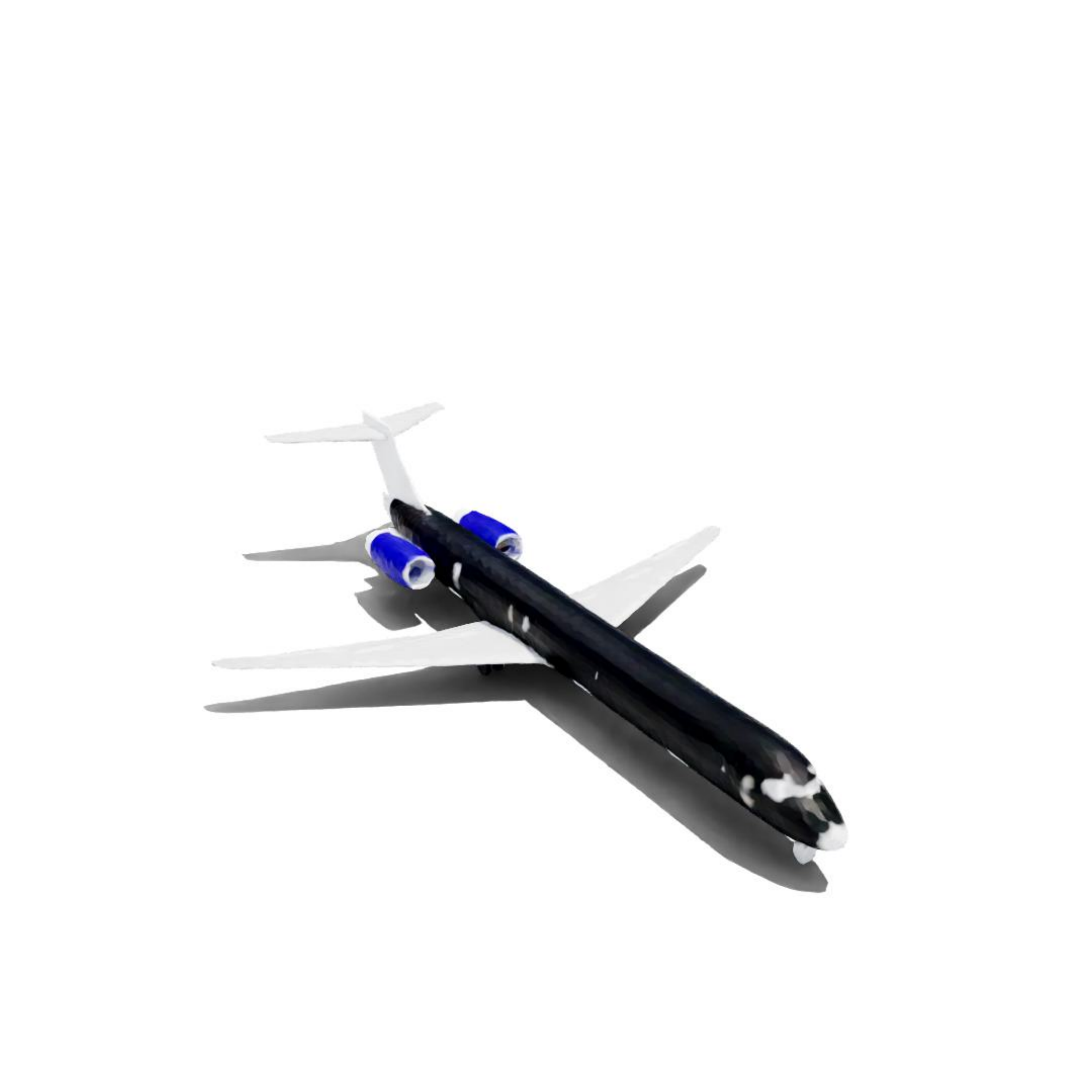}\includegraphics[width=0.16666666666666666\linewidth, trim={0 0cm 0 3cm}, clip]{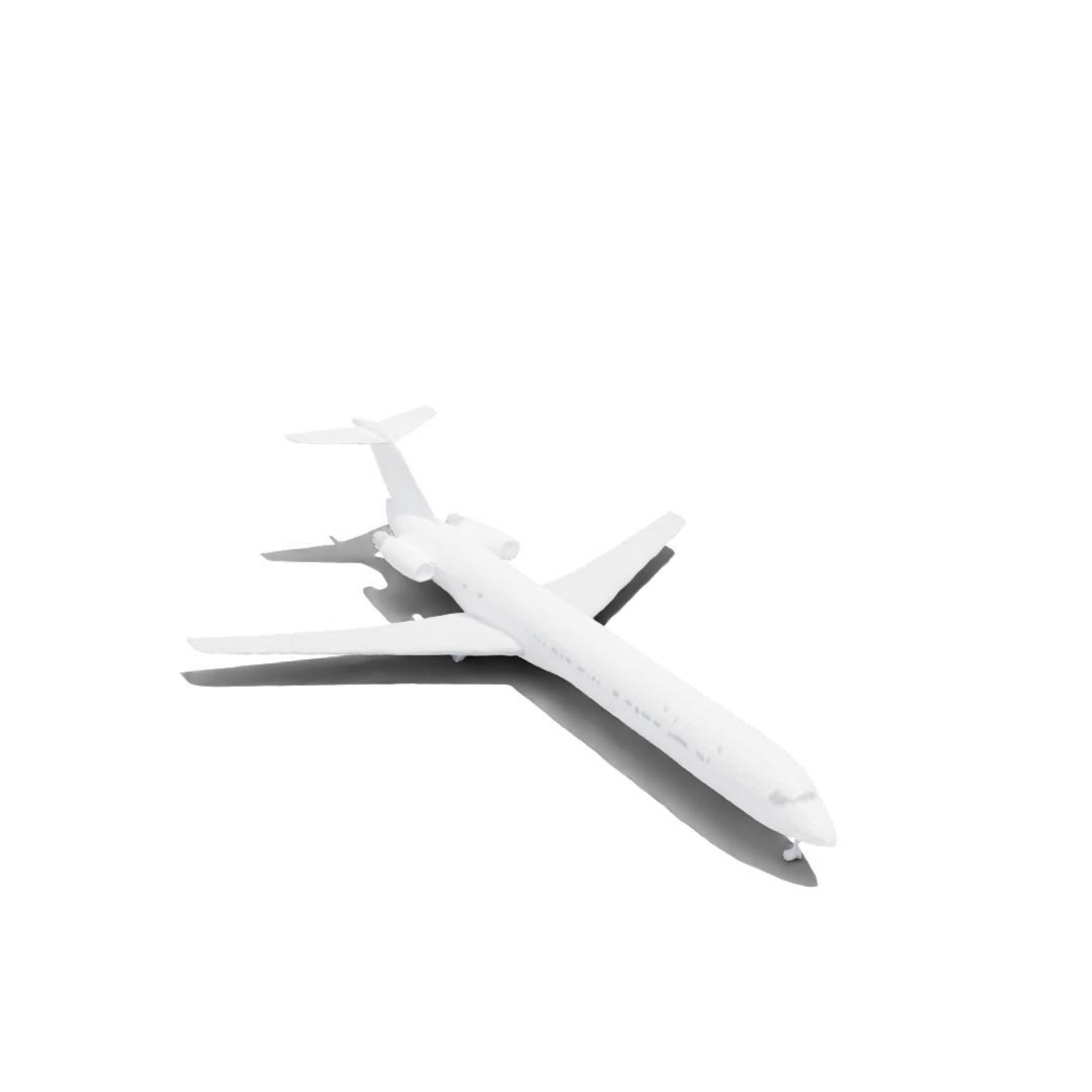}
\caption{\textbf{Random selection of airplanes generated in high resolution.}}
\label{fig:uncond:airplane:192}
\end{figure*}

\begin{figure*}[!ht]
\centering
\includegraphics[width=0.16666666666666666\linewidth, trim={0 0cm 0 3cm}, clip]{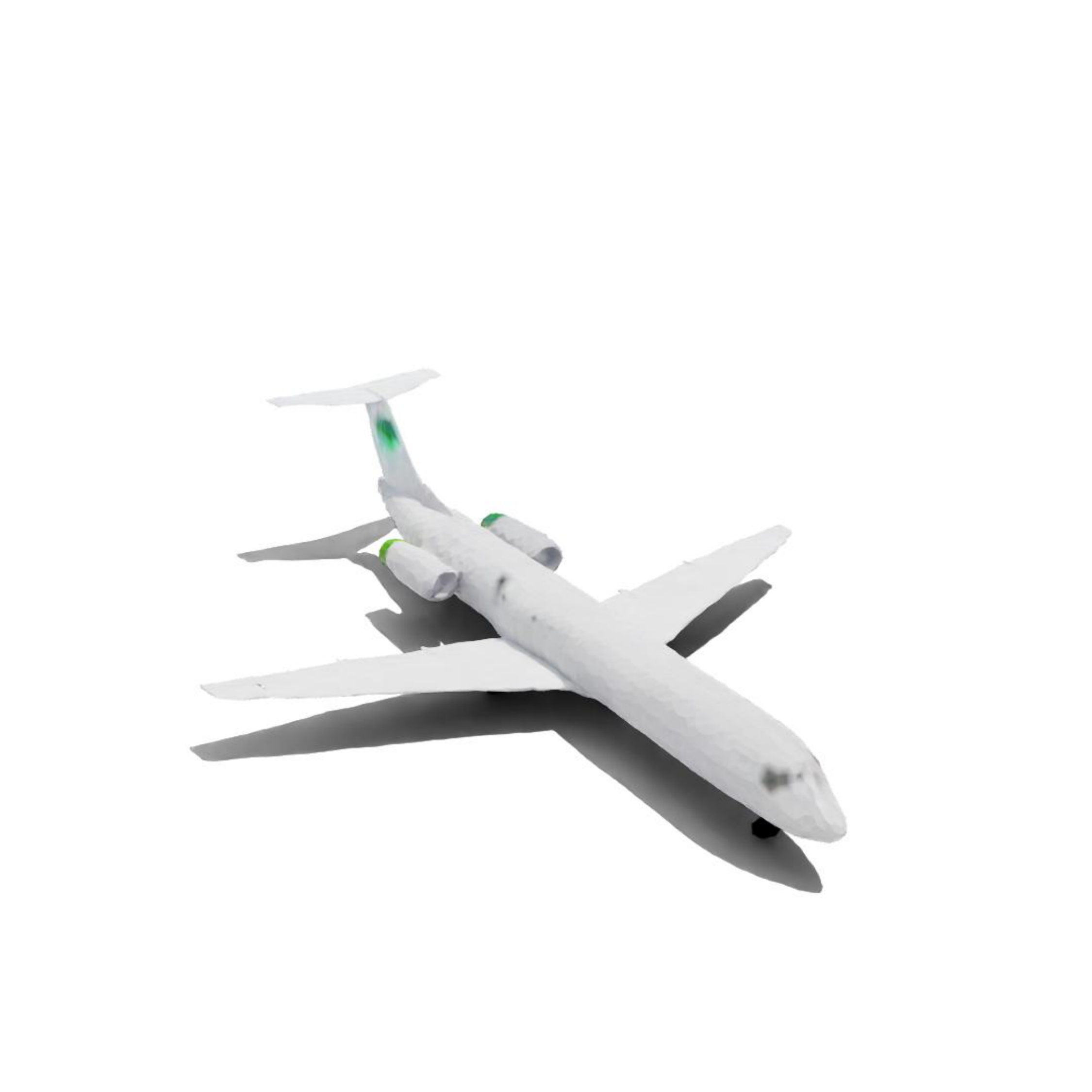}\includegraphics[width=0.16666666666666666\linewidth, trim={0 0cm 0 3cm}, clip]{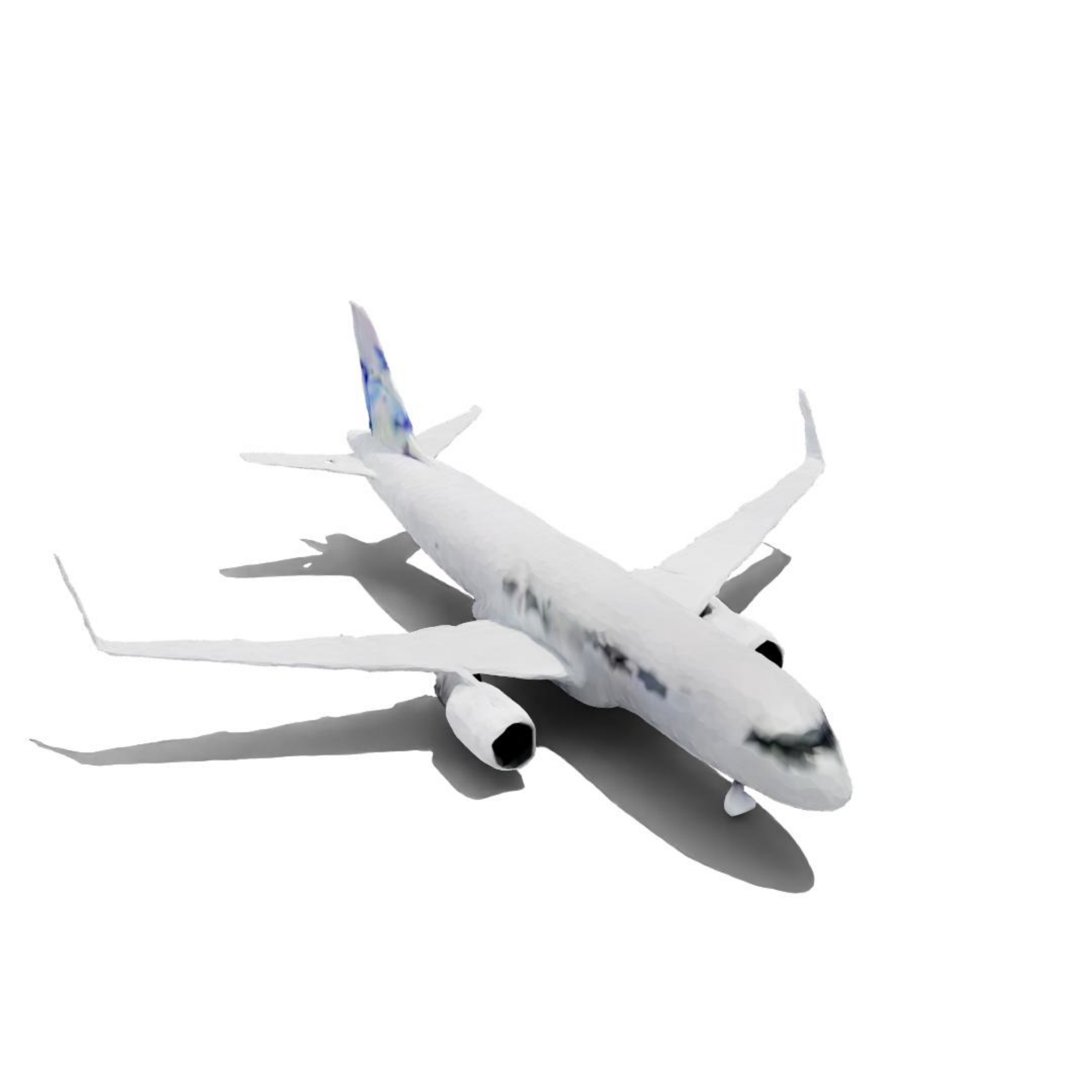}\includegraphics[width=0.16666666666666666\linewidth, trim={0 0cm 0 3cm}, clip]{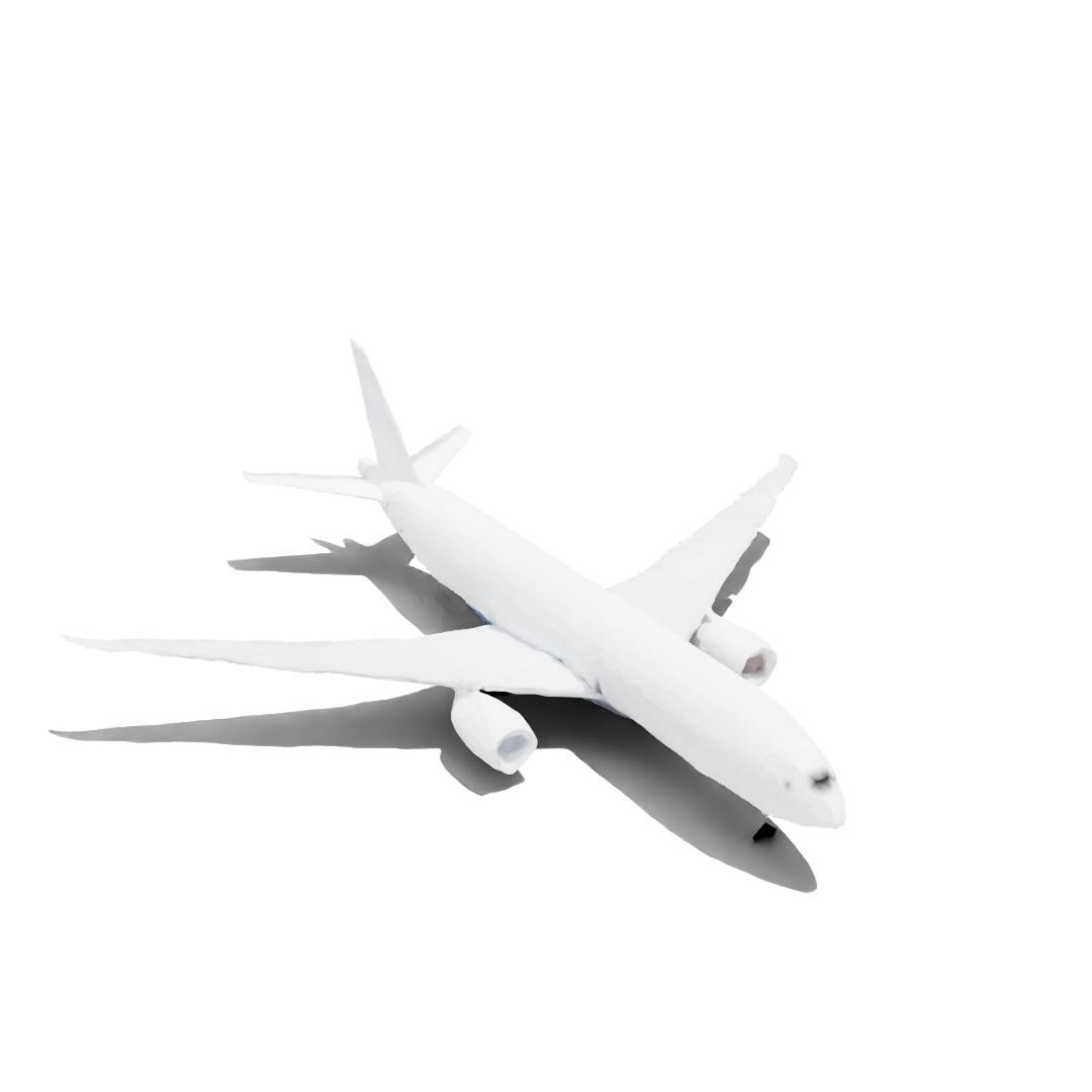}\includegraphics[width=0.16666666666666666\linewidth, trim={0 0cm 0 3cm}, clip]{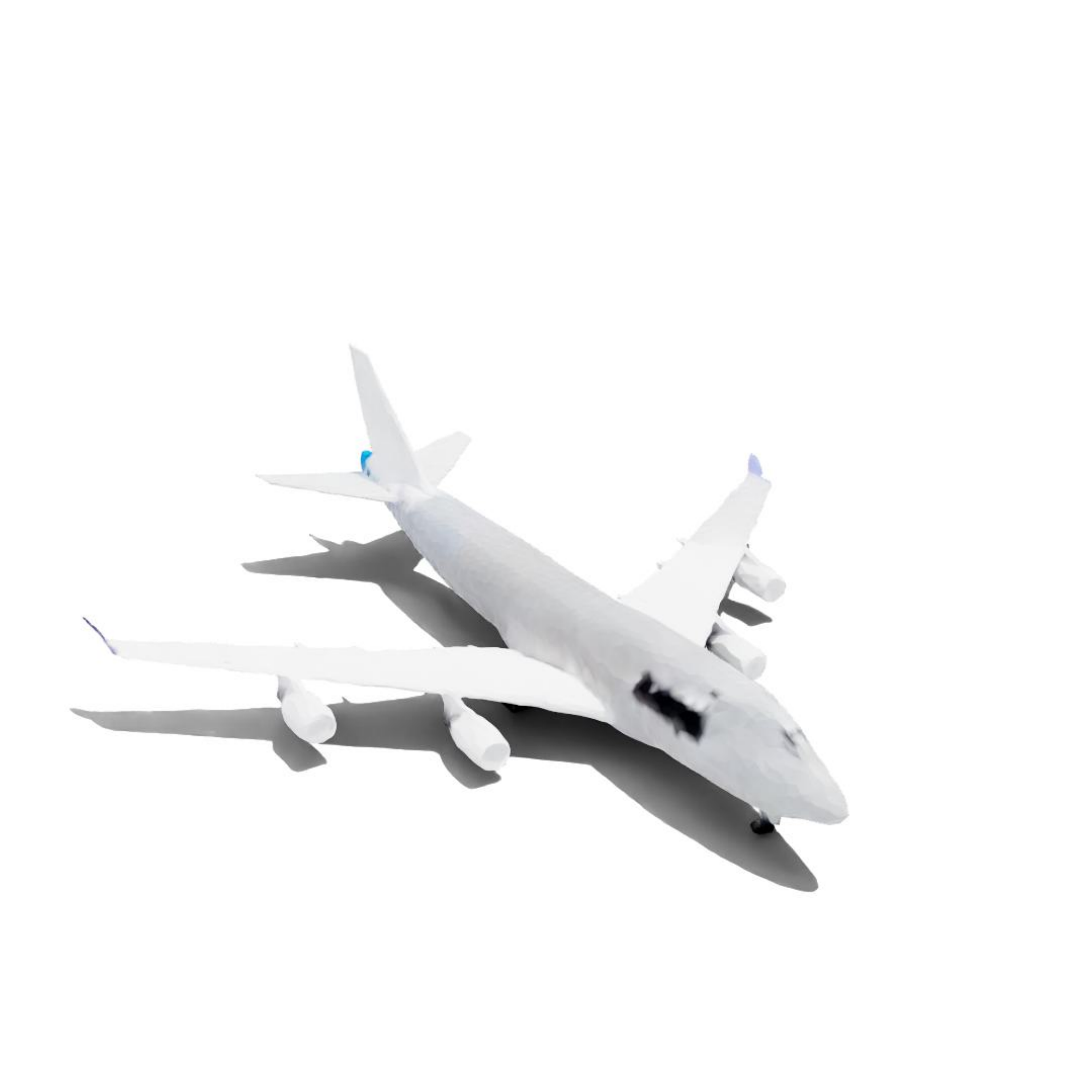}\includegraphics[width=0.16666666666666666\linewidth, trim={0 0cm 0 3cm}, clip]{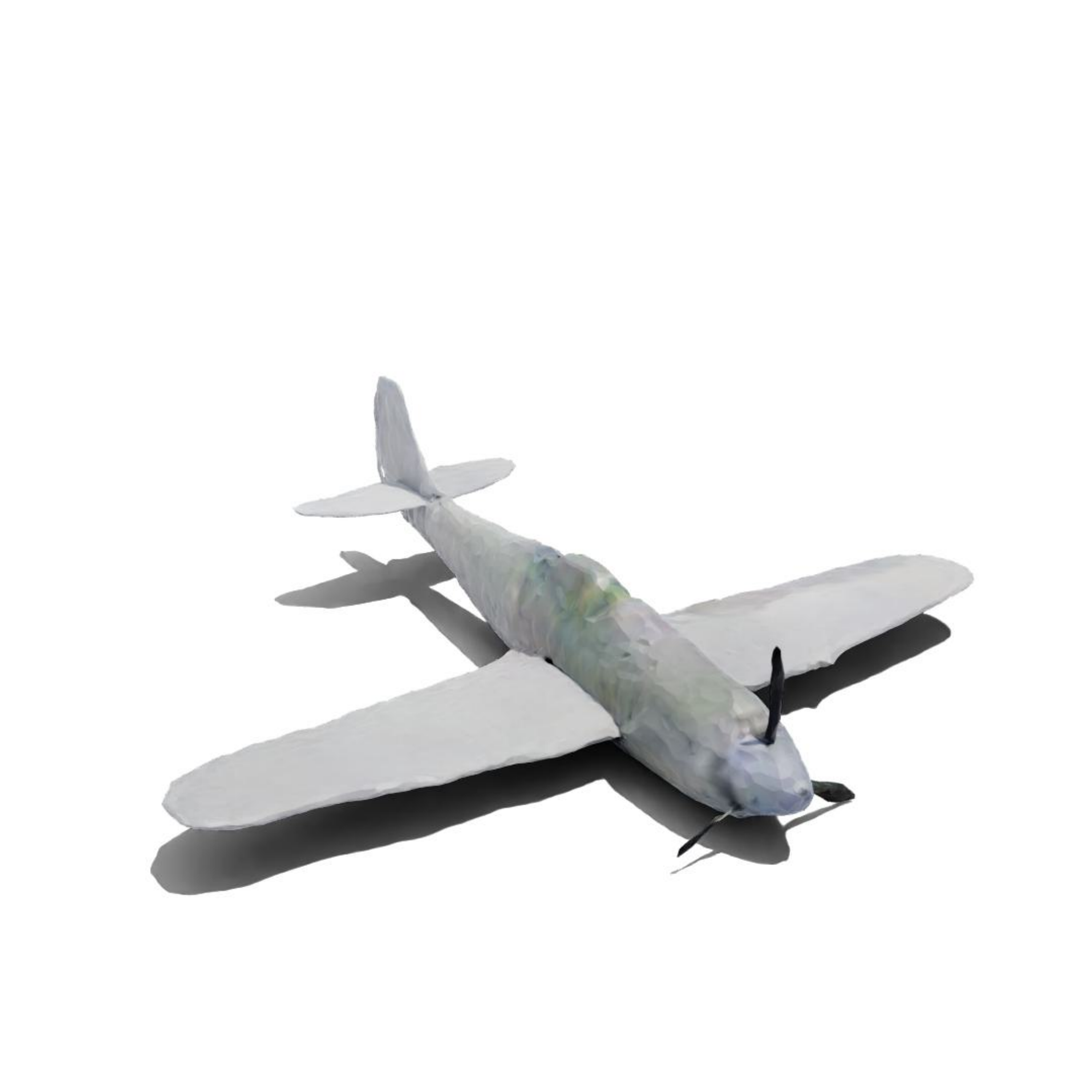}\includegraphics[width=0.16666666666666666\linewidth, trim={0 0cm 0 3cm}, clip]{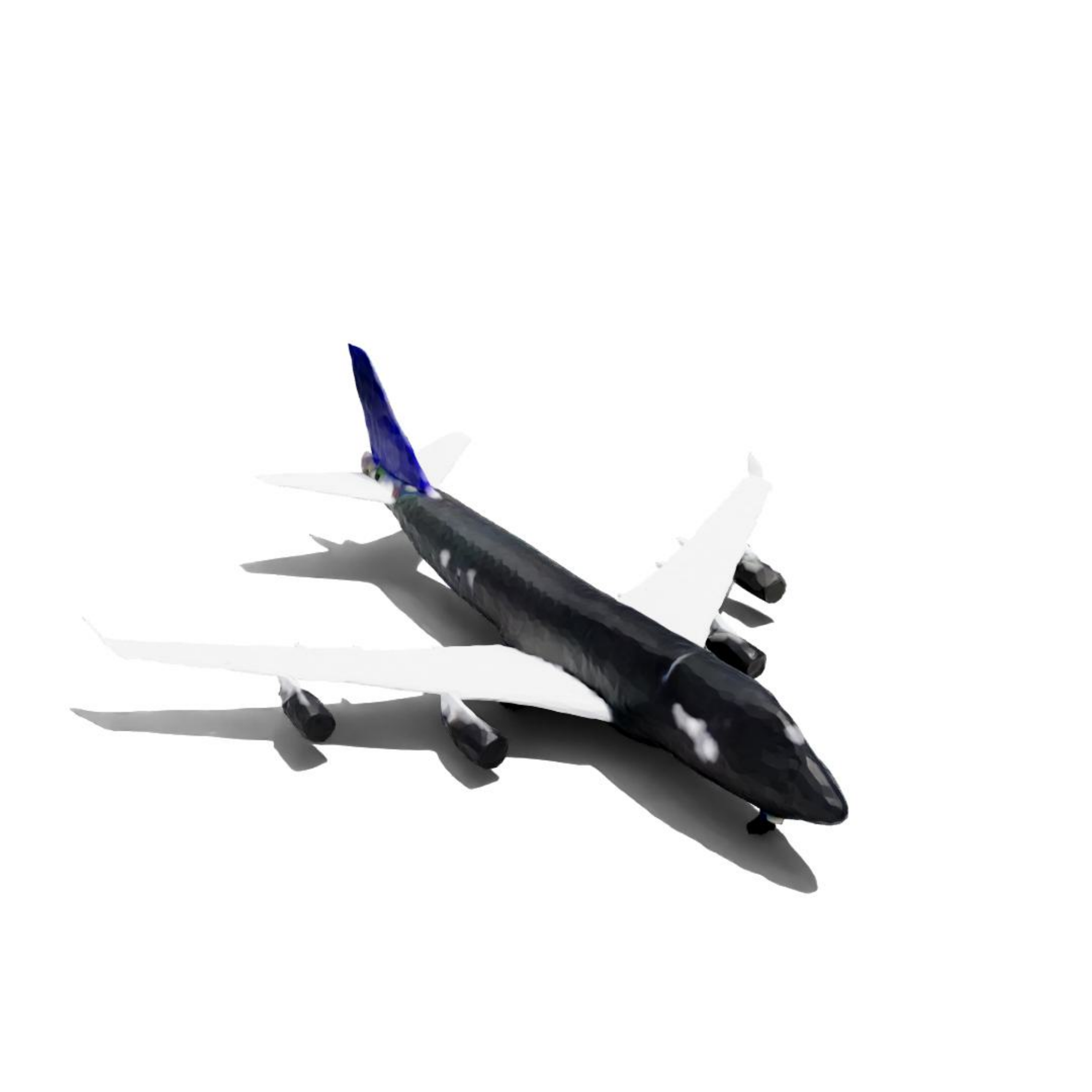}\\

\vspace{-0.43cm}
\includegraphics[width=0.16666666666666666\linewidth, trim={0 0cm 0 3cm}, clip]{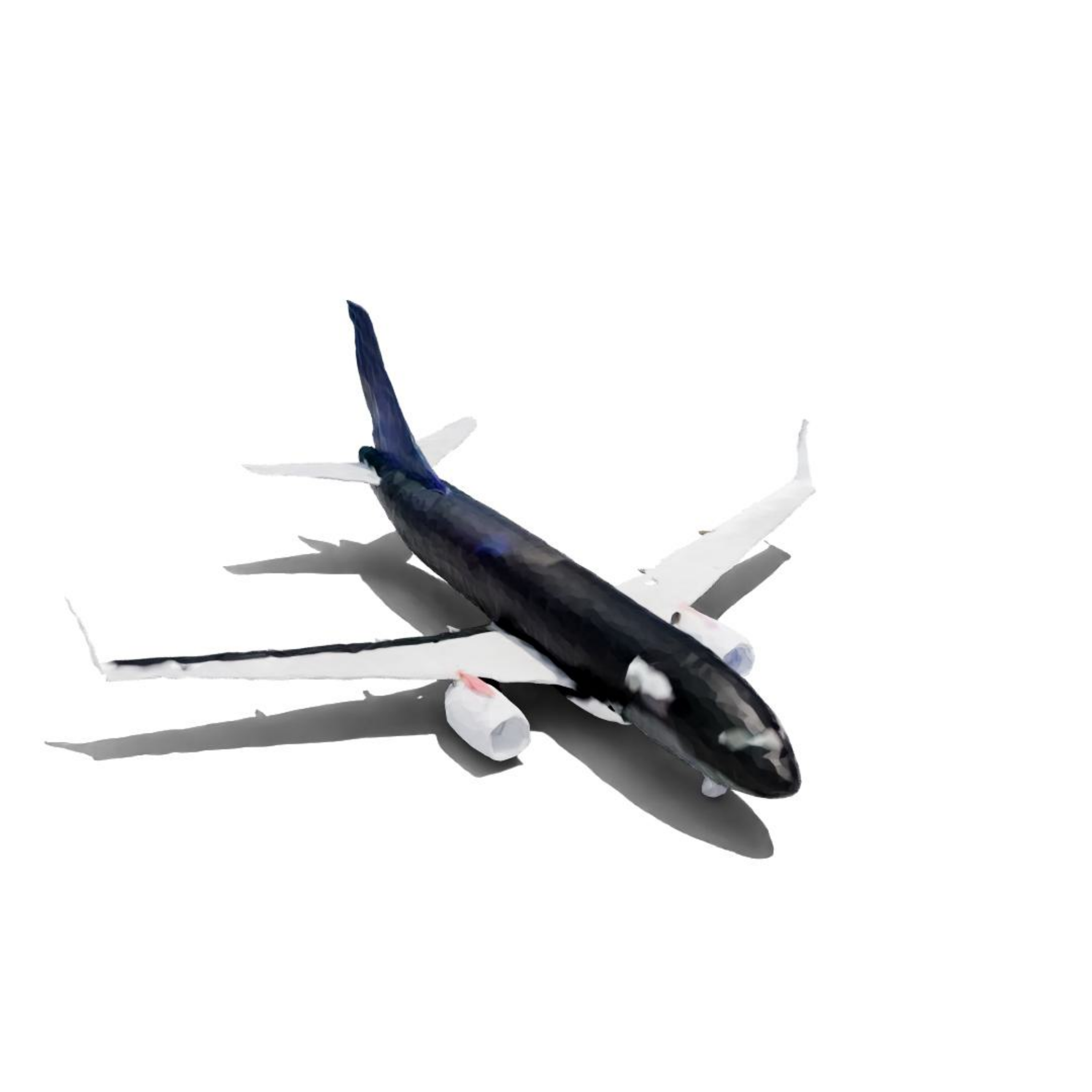}\includegraphics[width=0.16666666666666666\linewidth, trim={0 0cm 0 3cm}, clip]{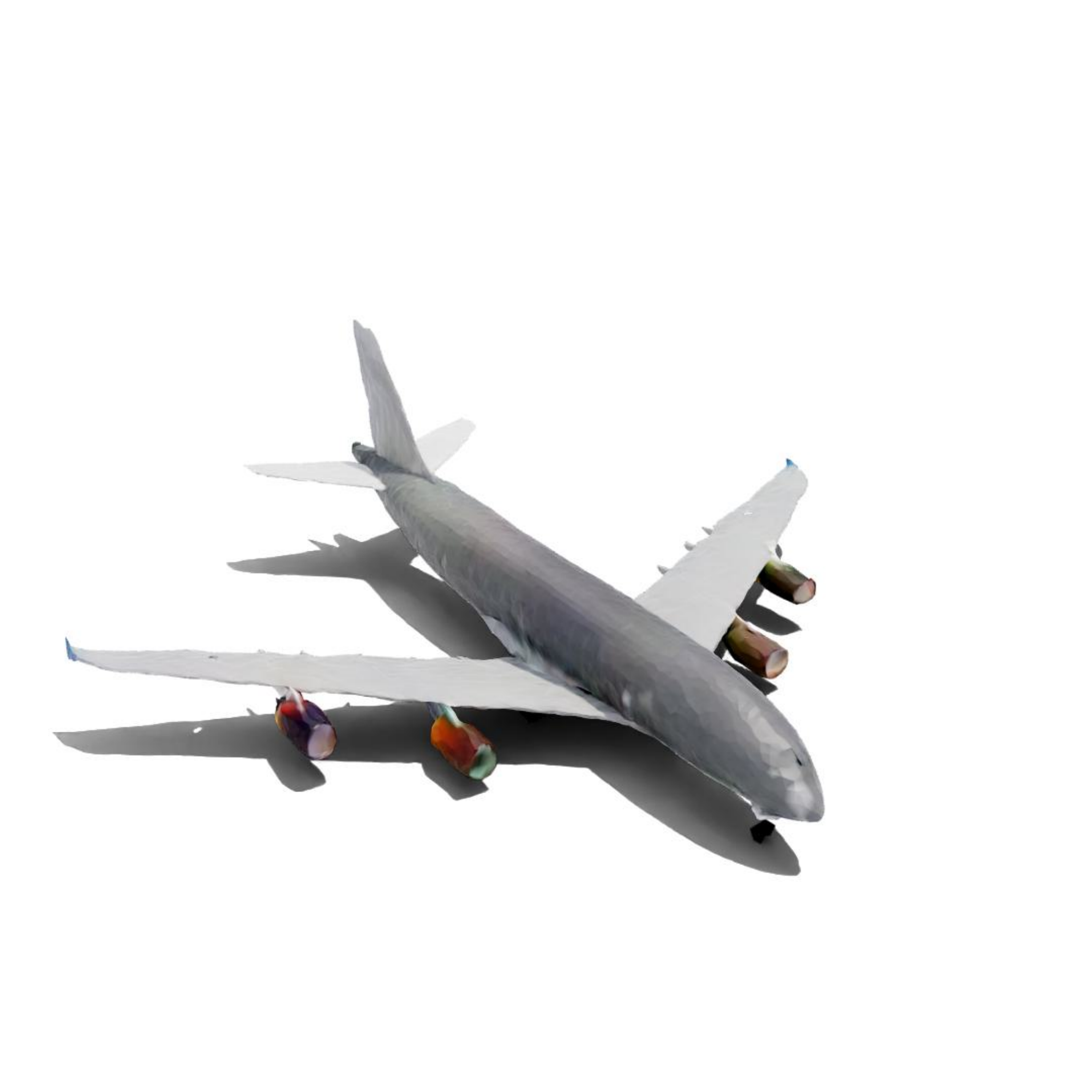}\includegraphics[width=0.16666666666666666\linewidth, trim={0 0cm 0 3cm}, clip]{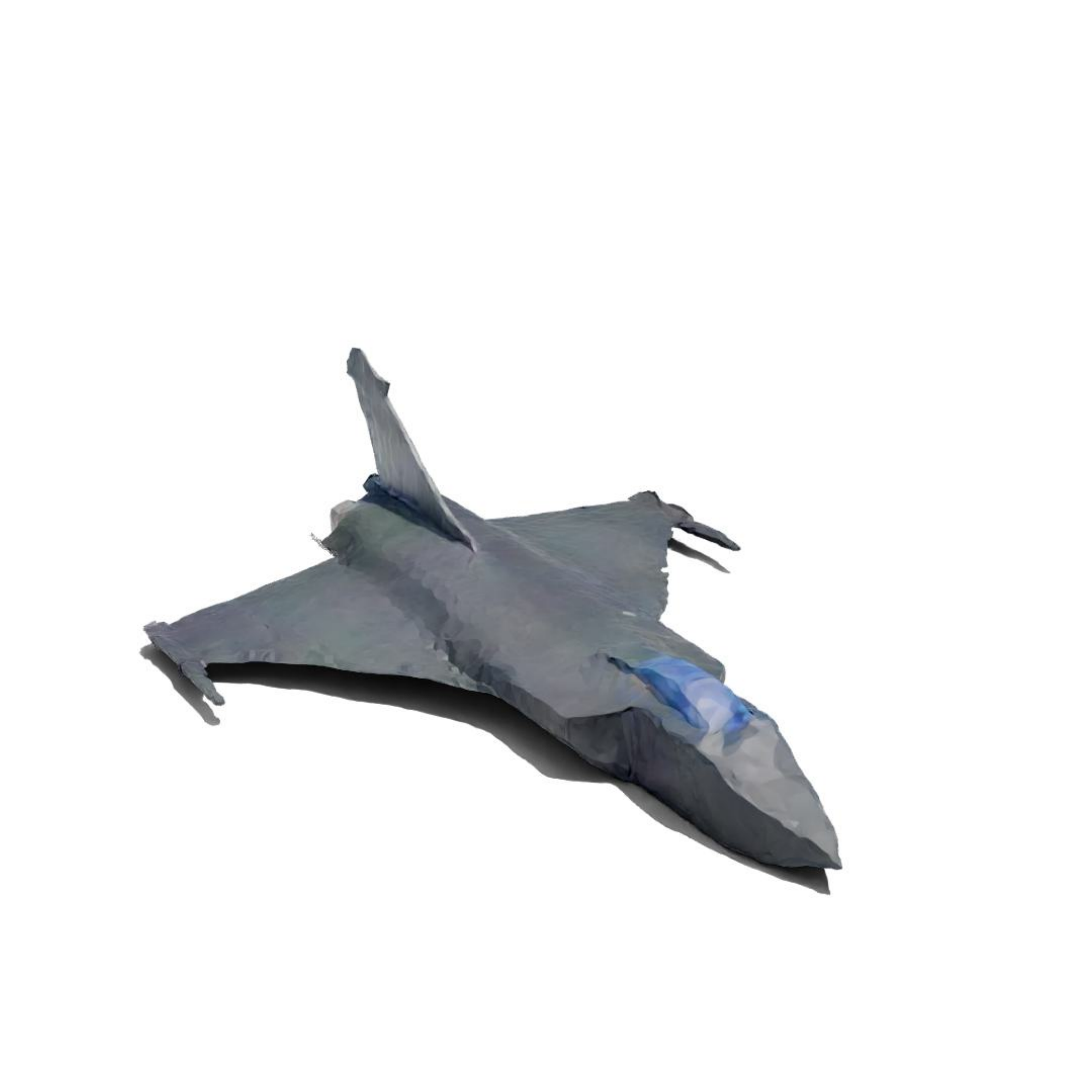}\includegraphics[width=0.16666666666666666\linewidth, trim={0 0cm 0 3cm}, clip]{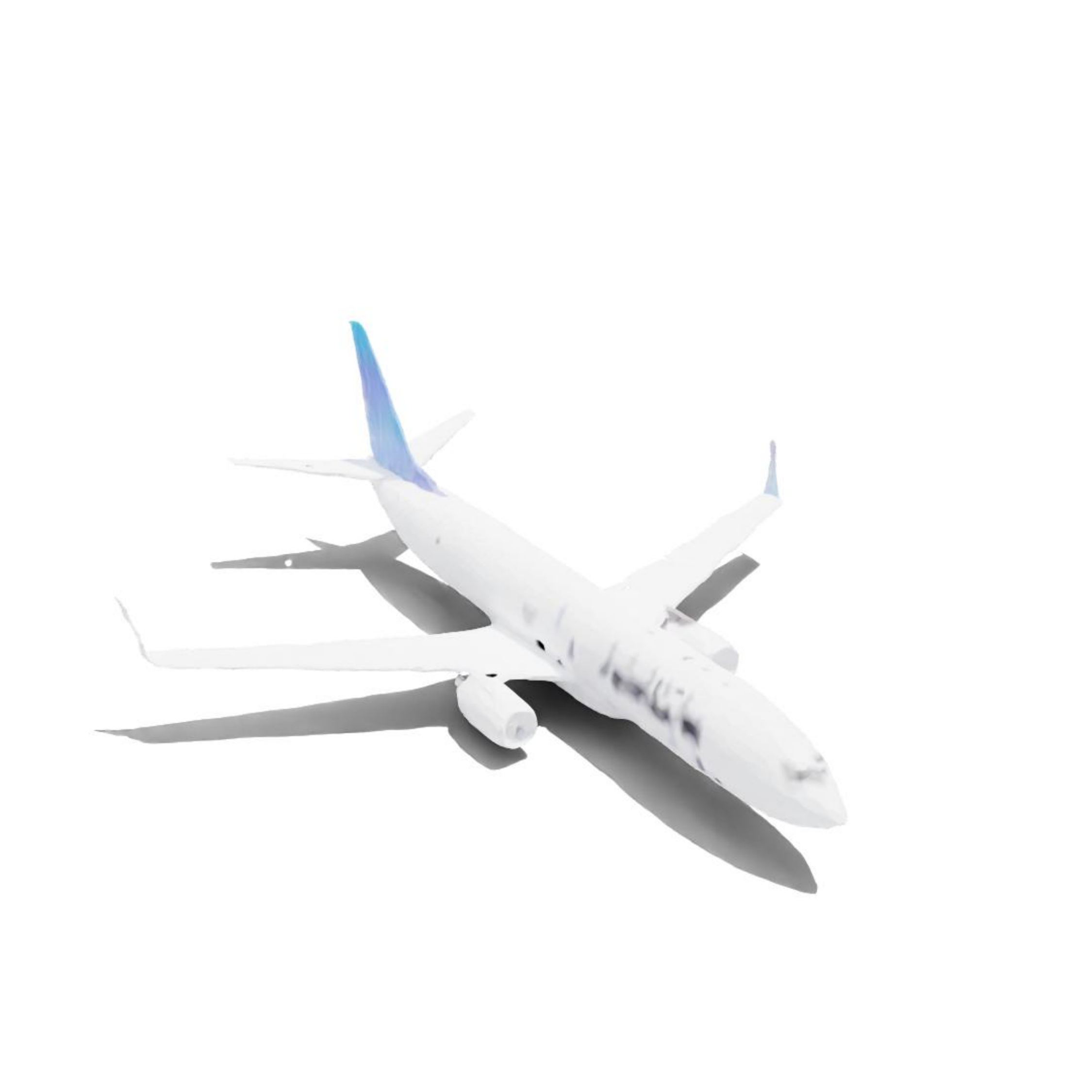}\includegraphics[width=0.16666666666666666\linewidth, trim={0 0cm 0 3cm}, clip]{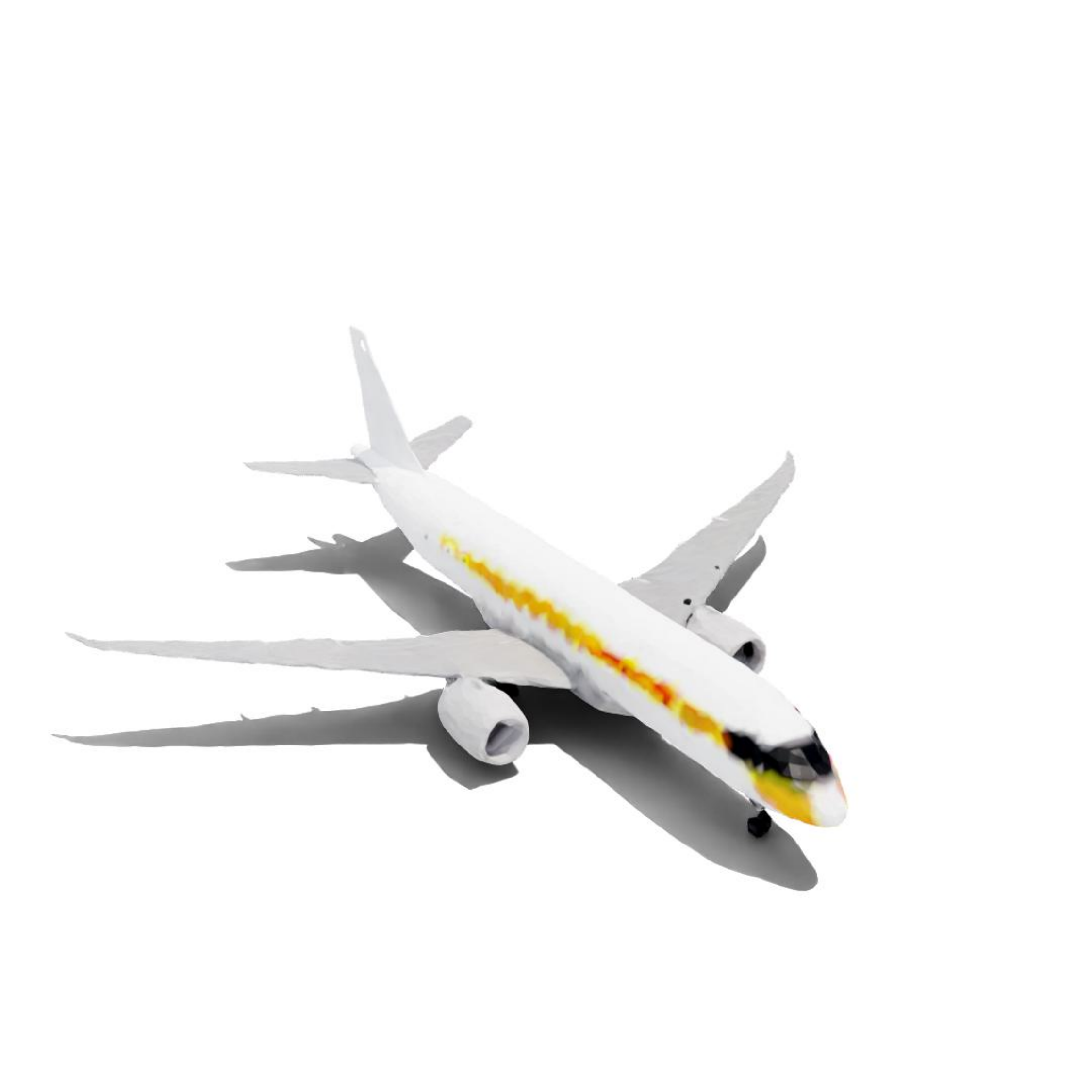}\includegraphics[width=0.16666666666666666\linewidth, trim={0 0cm 0 3cm}, clip]{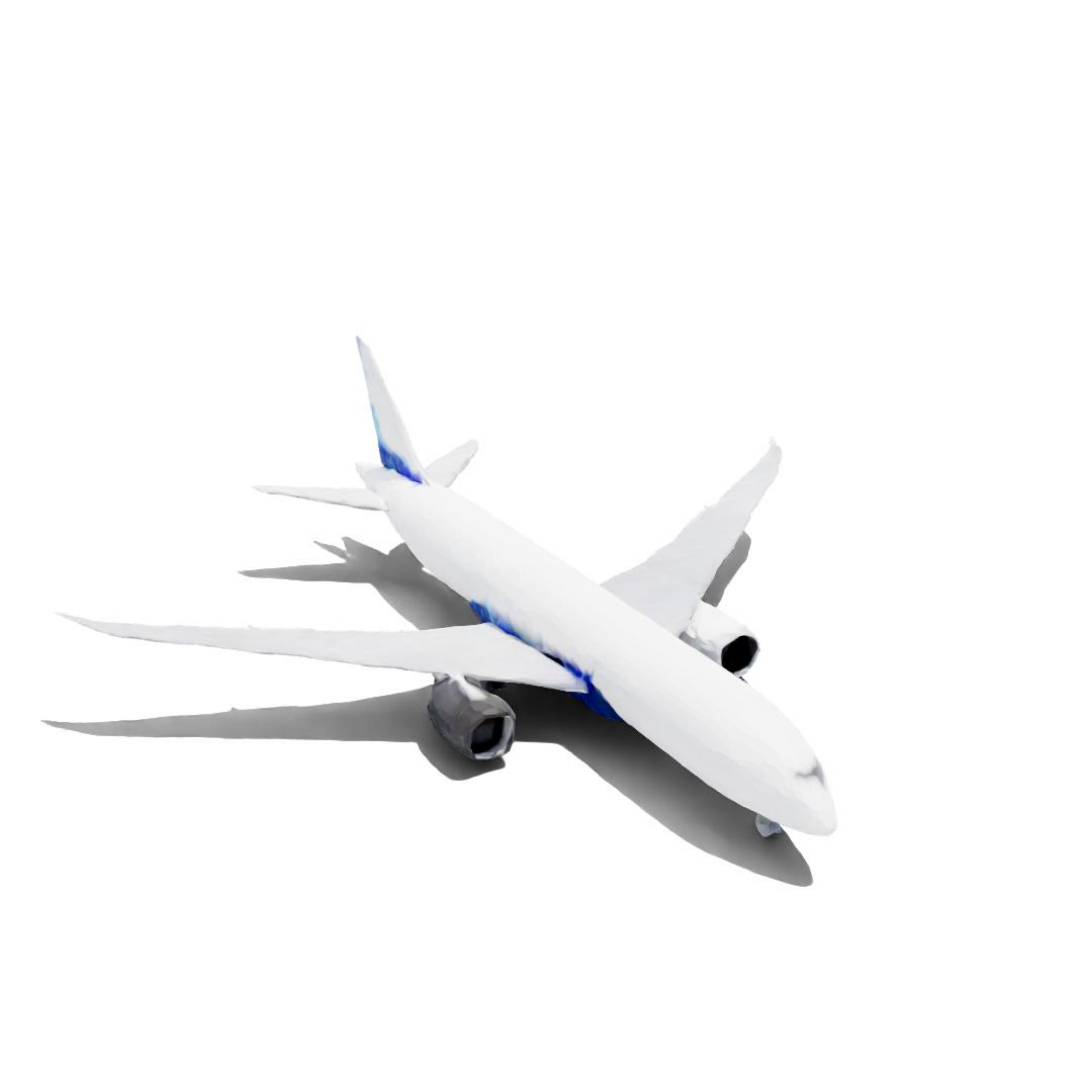}\\

\vspace{-0.43cm}
\includegraphics[width=0.16666666666666666\linewidth, trim={0 0cm 0 3cm}, clip]{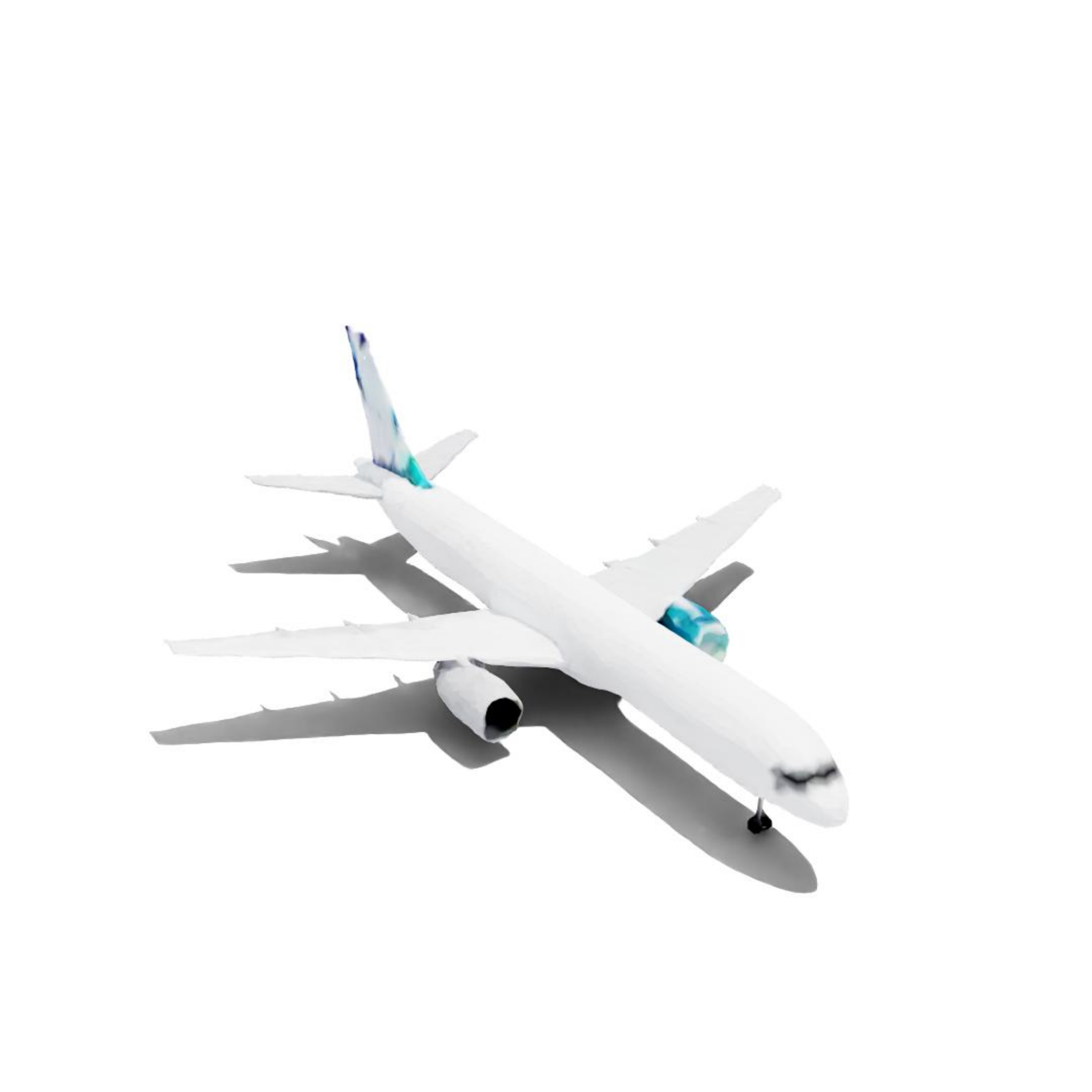}\includegraphics[width=0.16666666666666666\linewidth, trim={0 0cm 0 3cm}, clip]{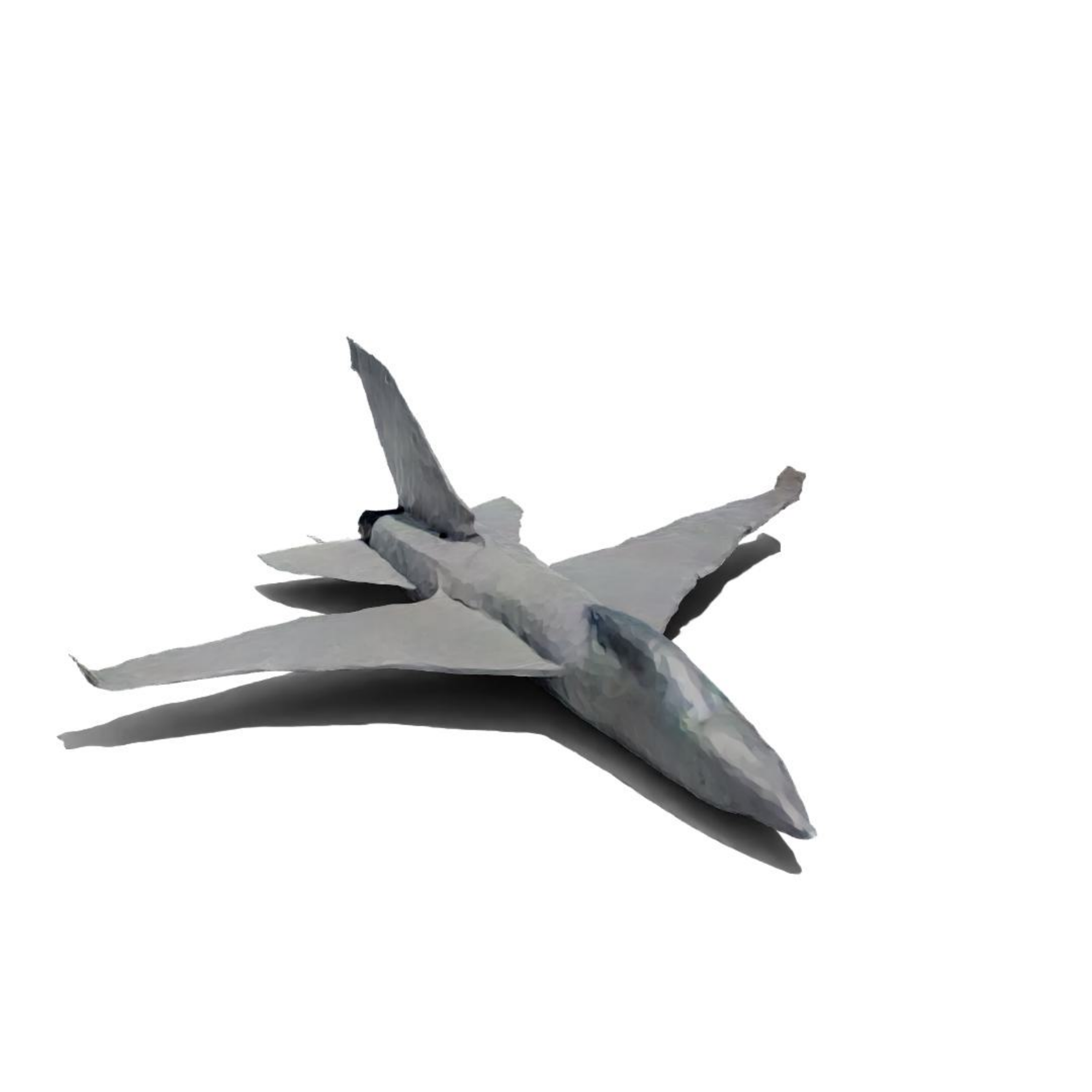}\includegraphics[width=0.16666666666666666\linewidth, trim={0 0cm 0 3cm}, clip]{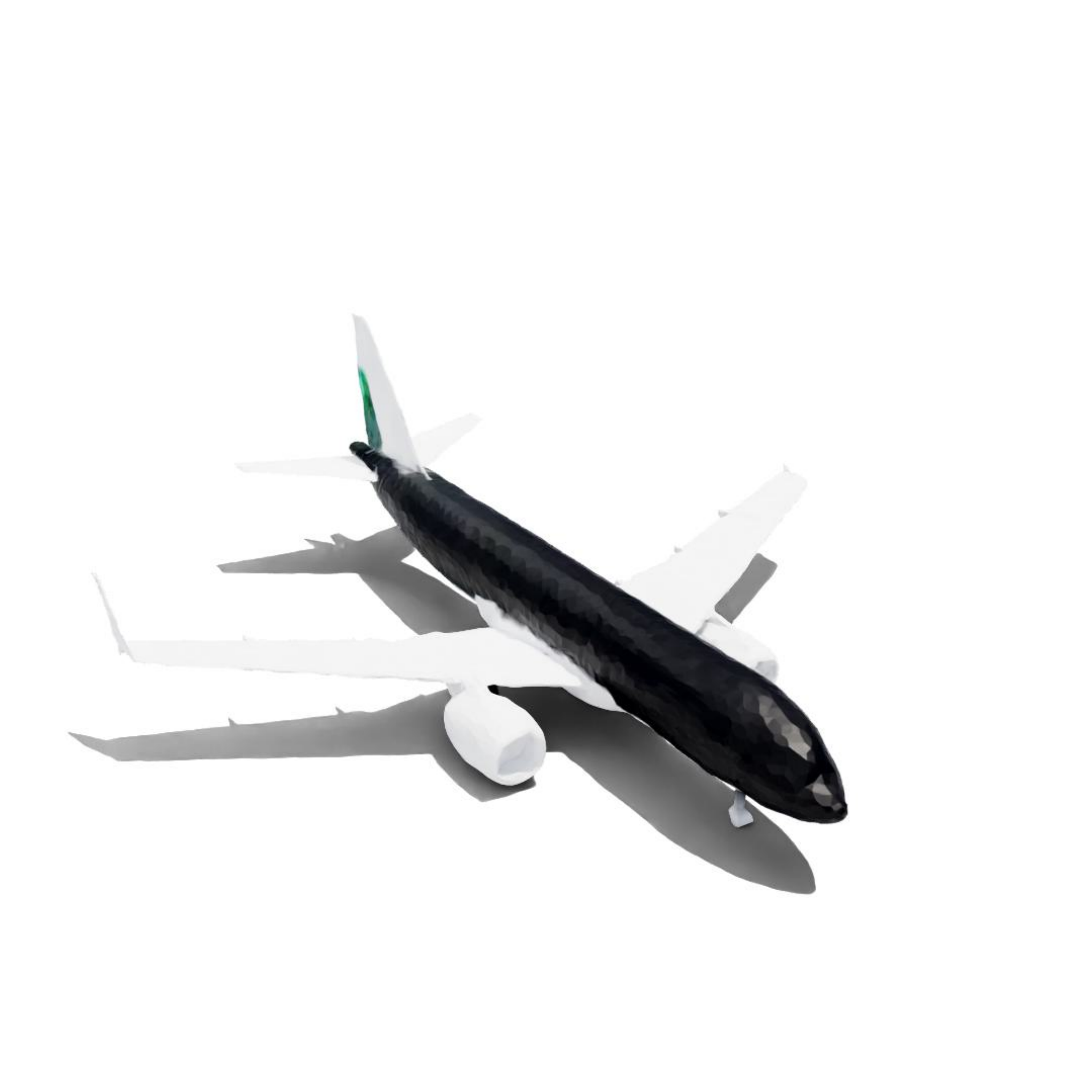}\includegraphics[width=0.16666666666666666\linewidth, trim={0 0cm 0 3cm}, clip]{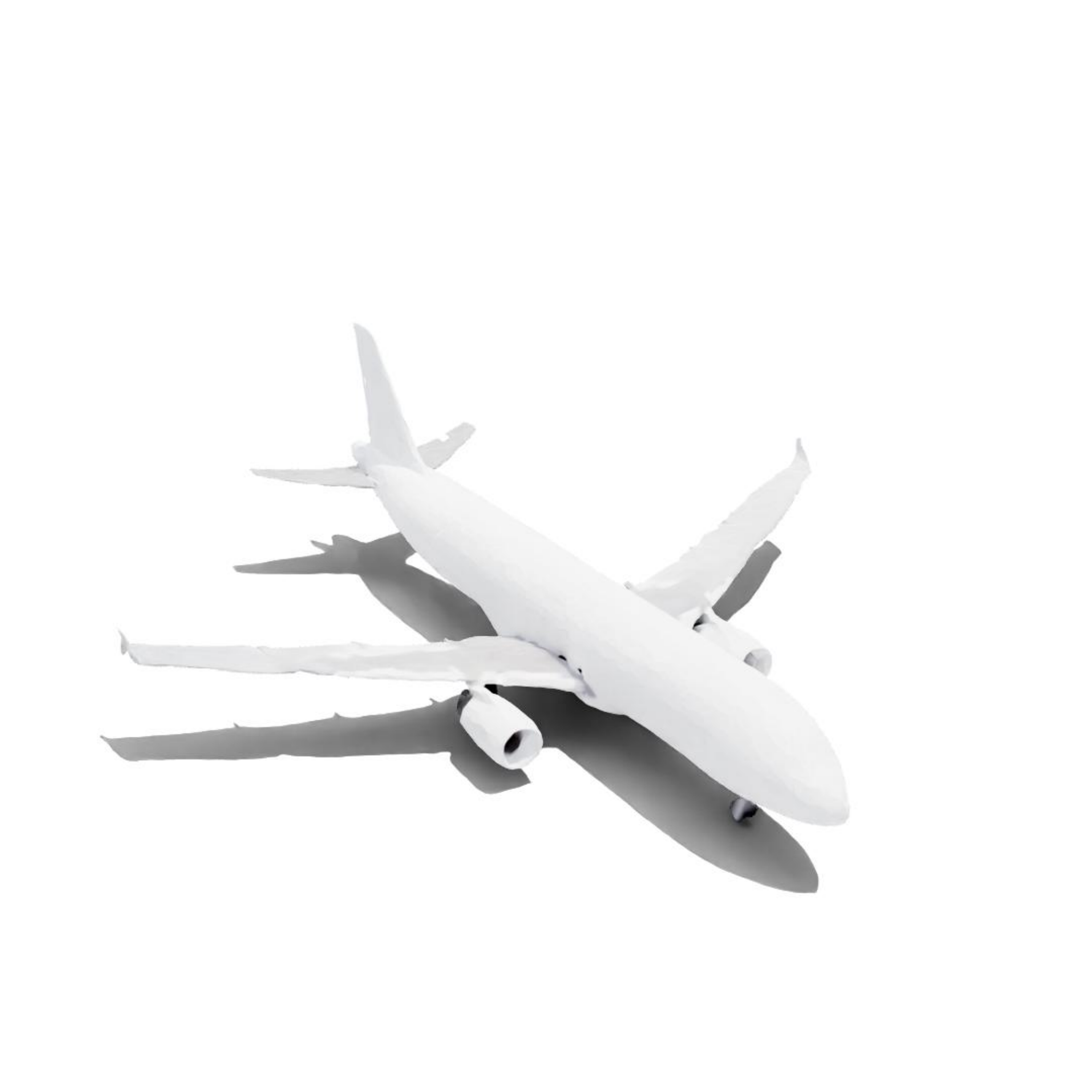}\includegraphics[width=0.16666666666666666\linewidth, trim={0 0cm 0 3cm}, clip]{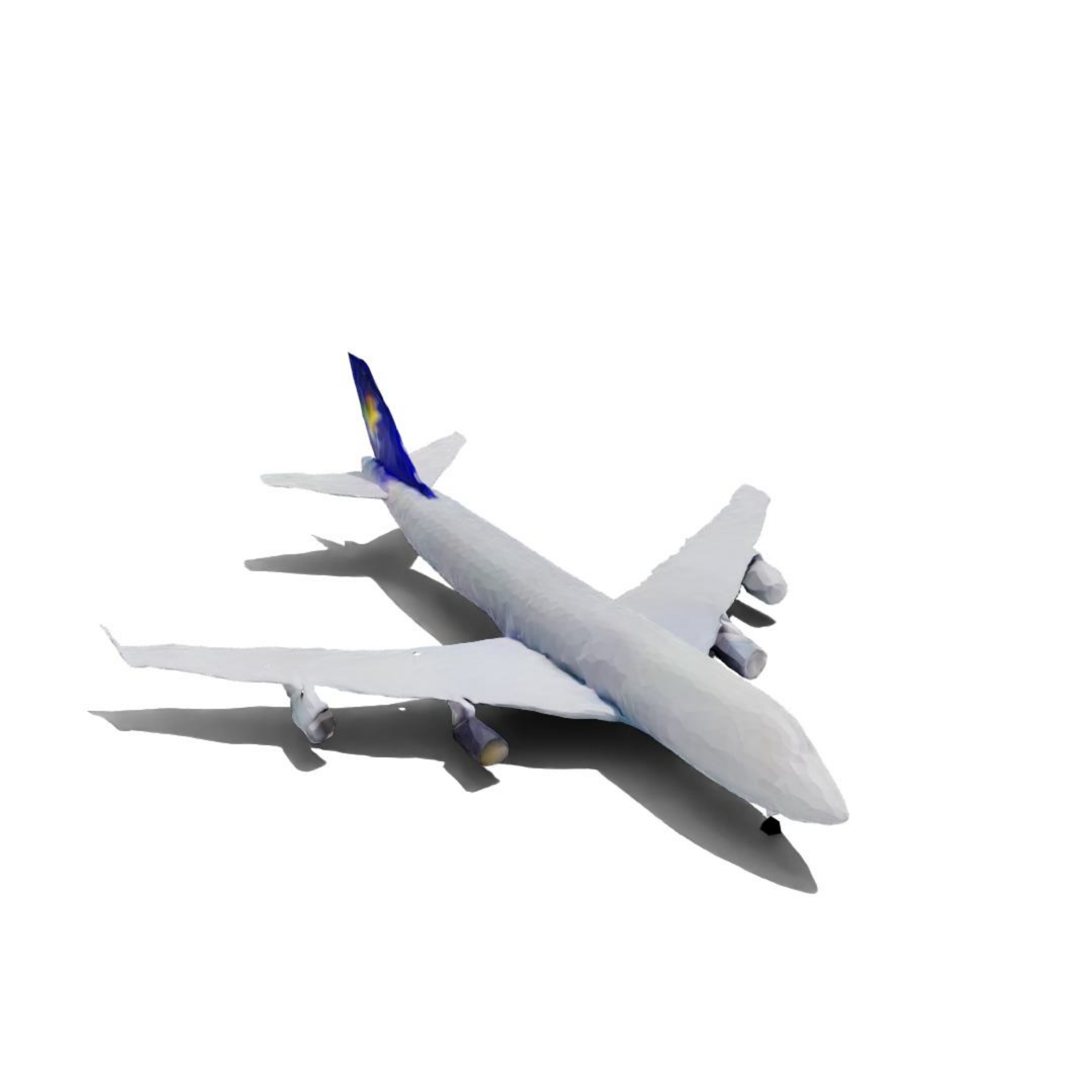}\includegraphics[width=0.16666666666666666\linewidth, trim={0 0cm 0 3cm}, clip]{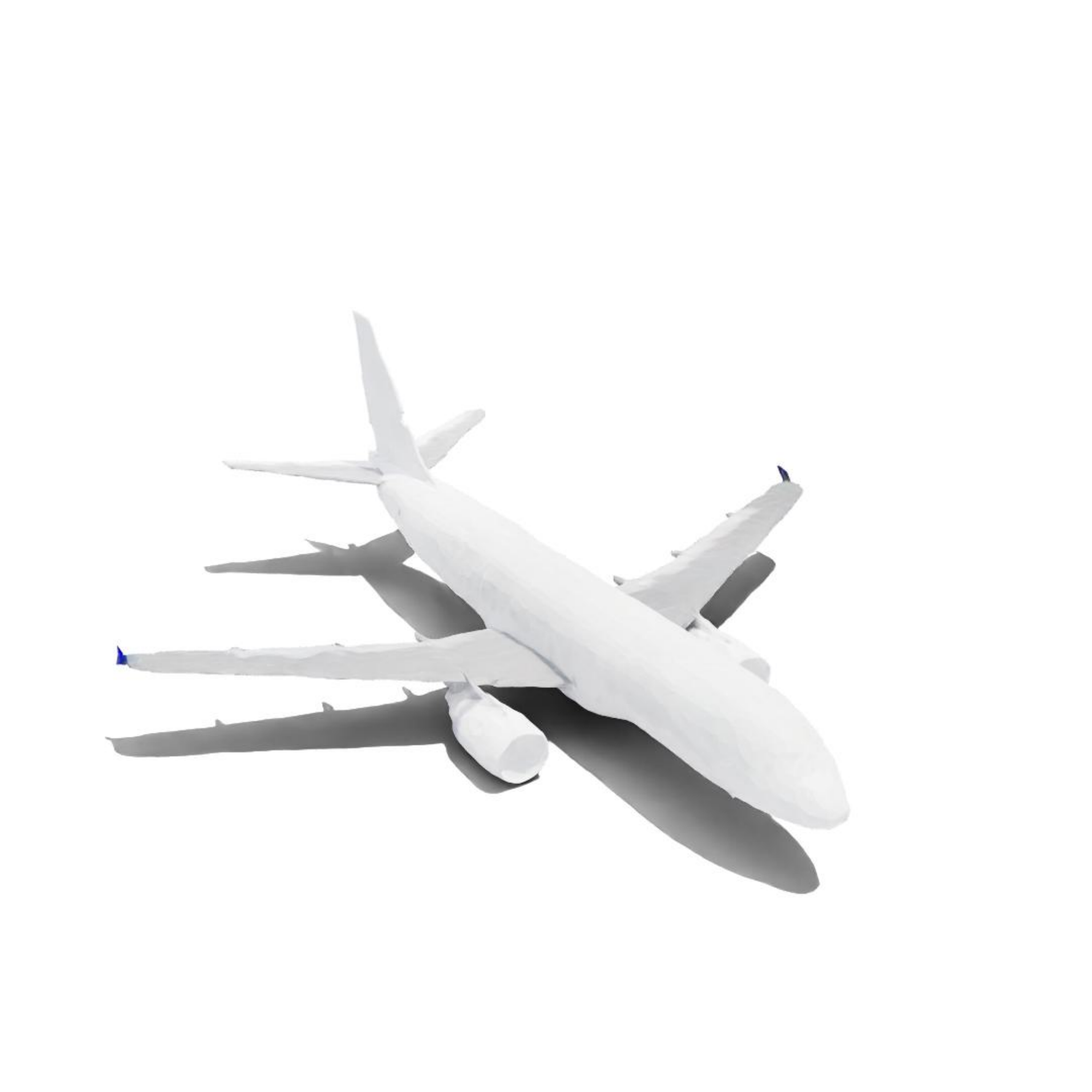}\\

\vspace{-0.43cm}
\includegraphics[width=0.16666666666666666\linewidth, trim={0 0cm 0 3cm}, clip]{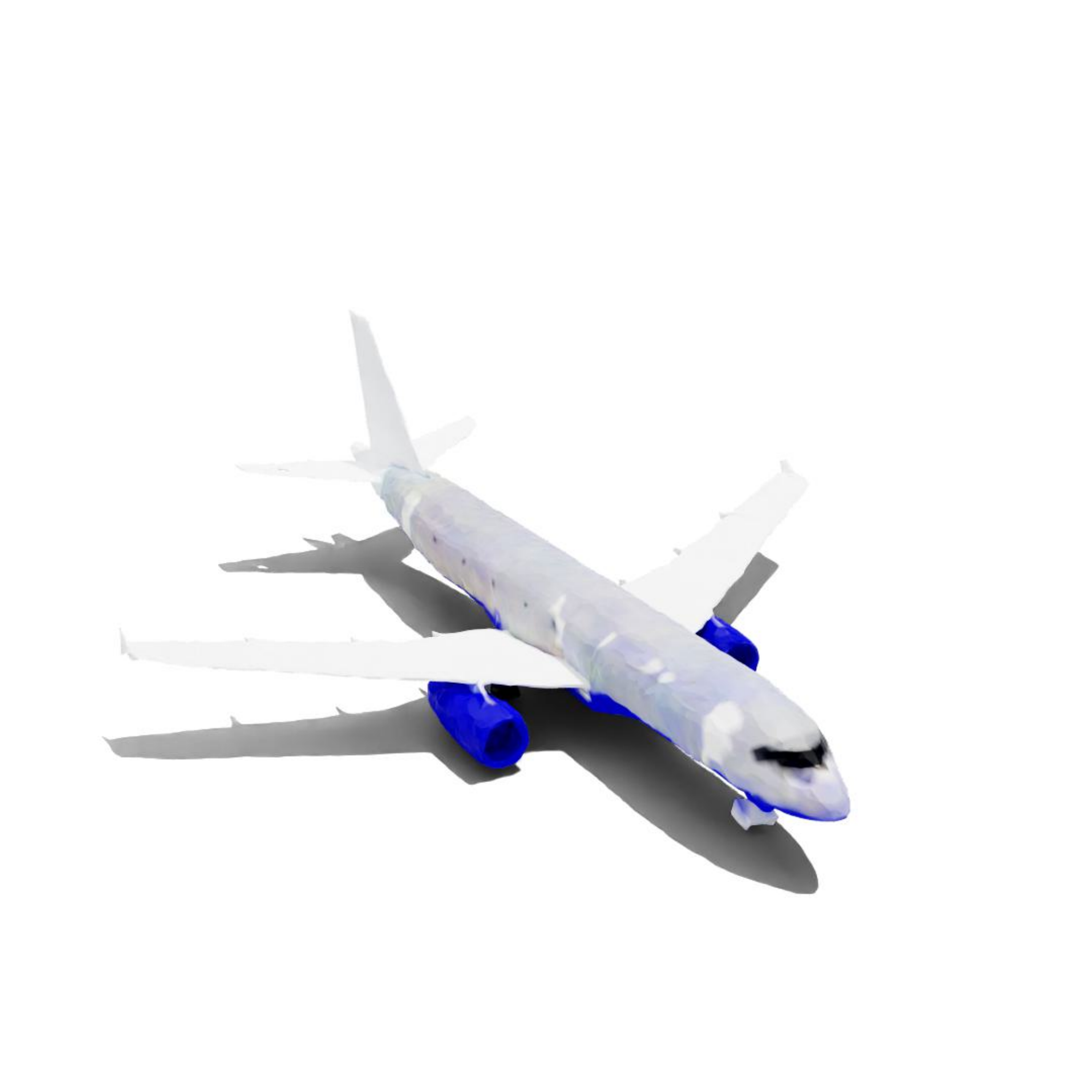}\includegraphics[width=0.16666666666666666\linewidth, trim={0 0cm 0 3cm}, clip]{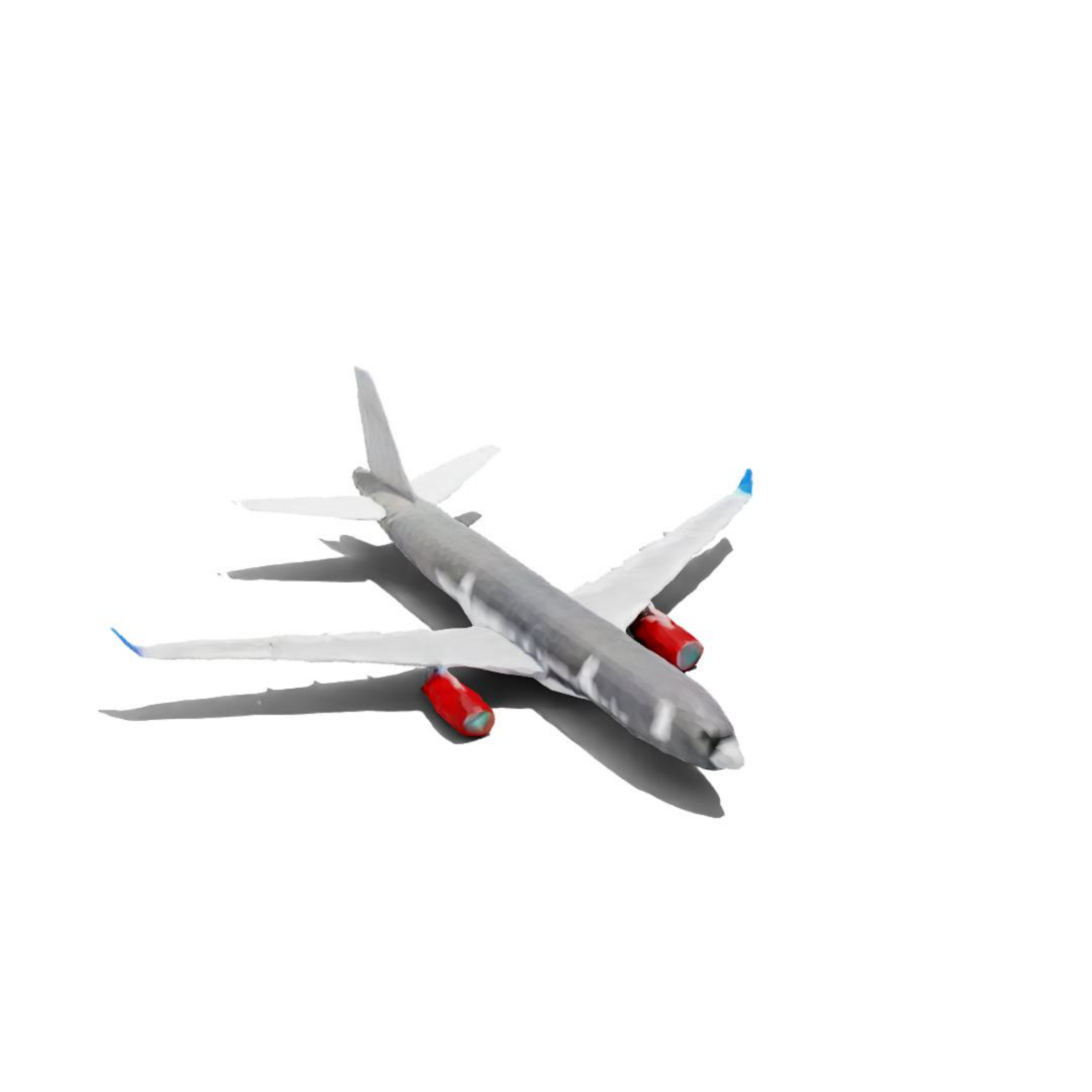}\includegraphics[width=0.16666666666666666\linewidth, trim={0 0cm 0 3cm}, clip]{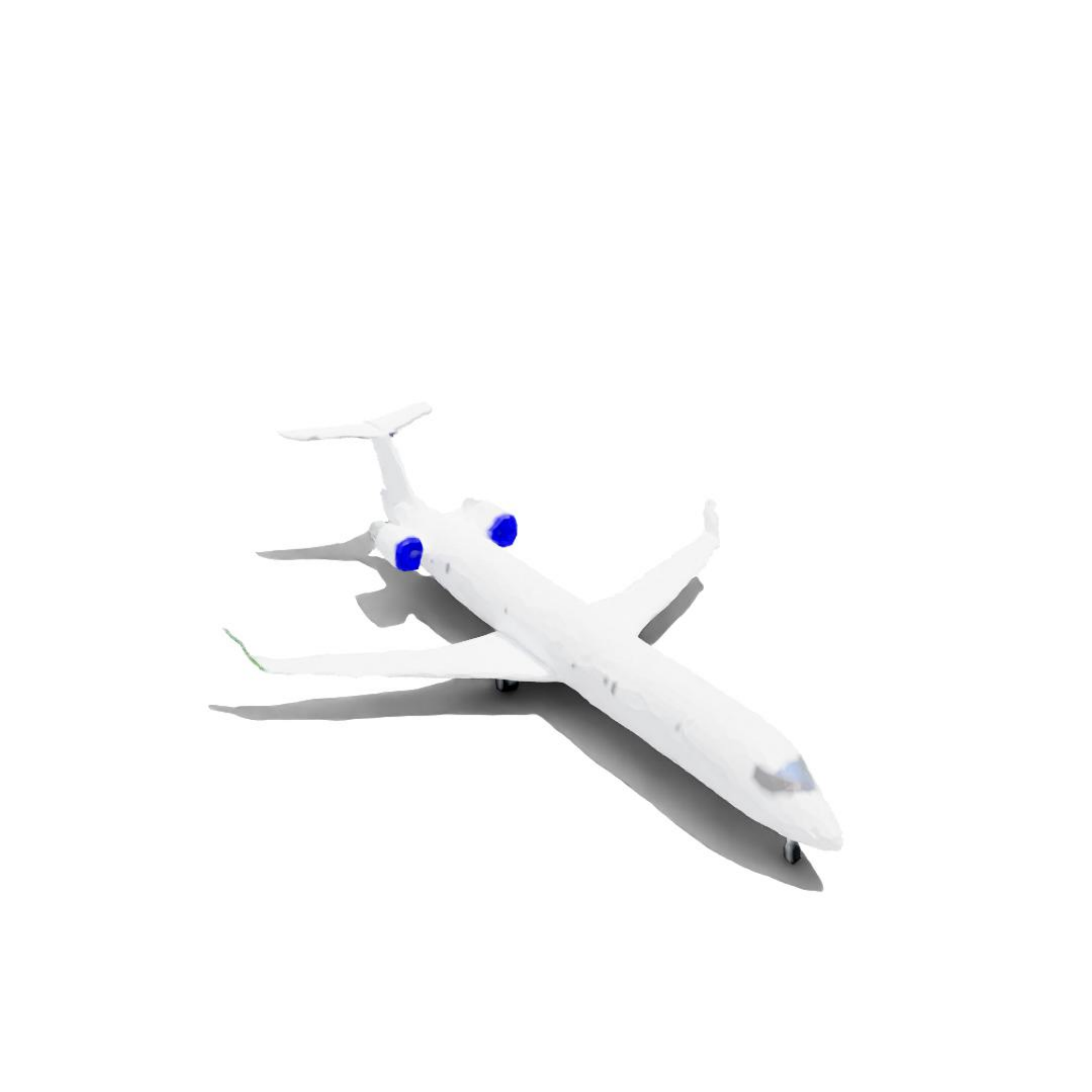}\includegraphics[width=0.16666666666666666\linewidth, trim={0 0cm 0 3cm}, clip]{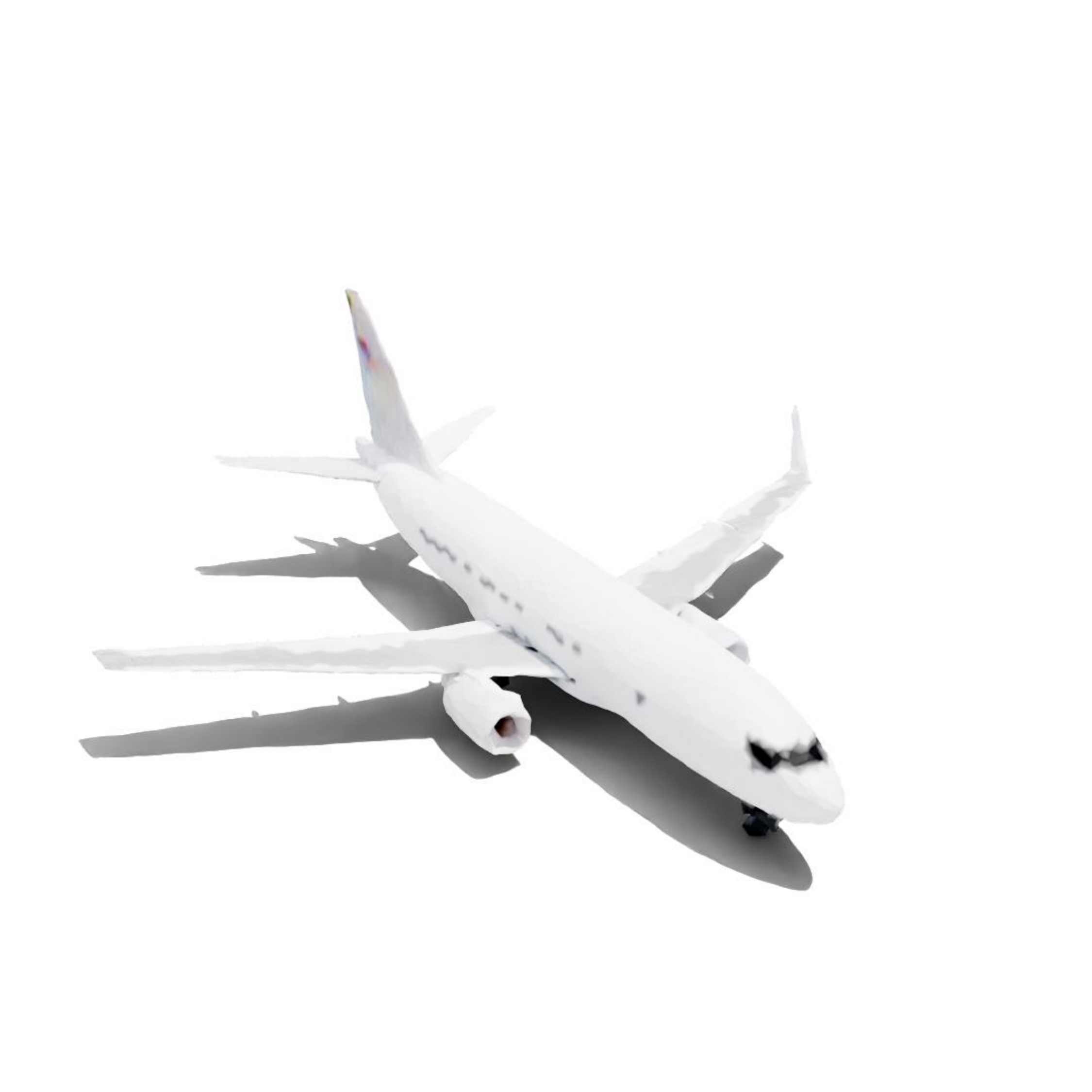}\includegraphics[width=0.16666666666666666\linewidth, trim={0 0cm 0 3cm}, clip]{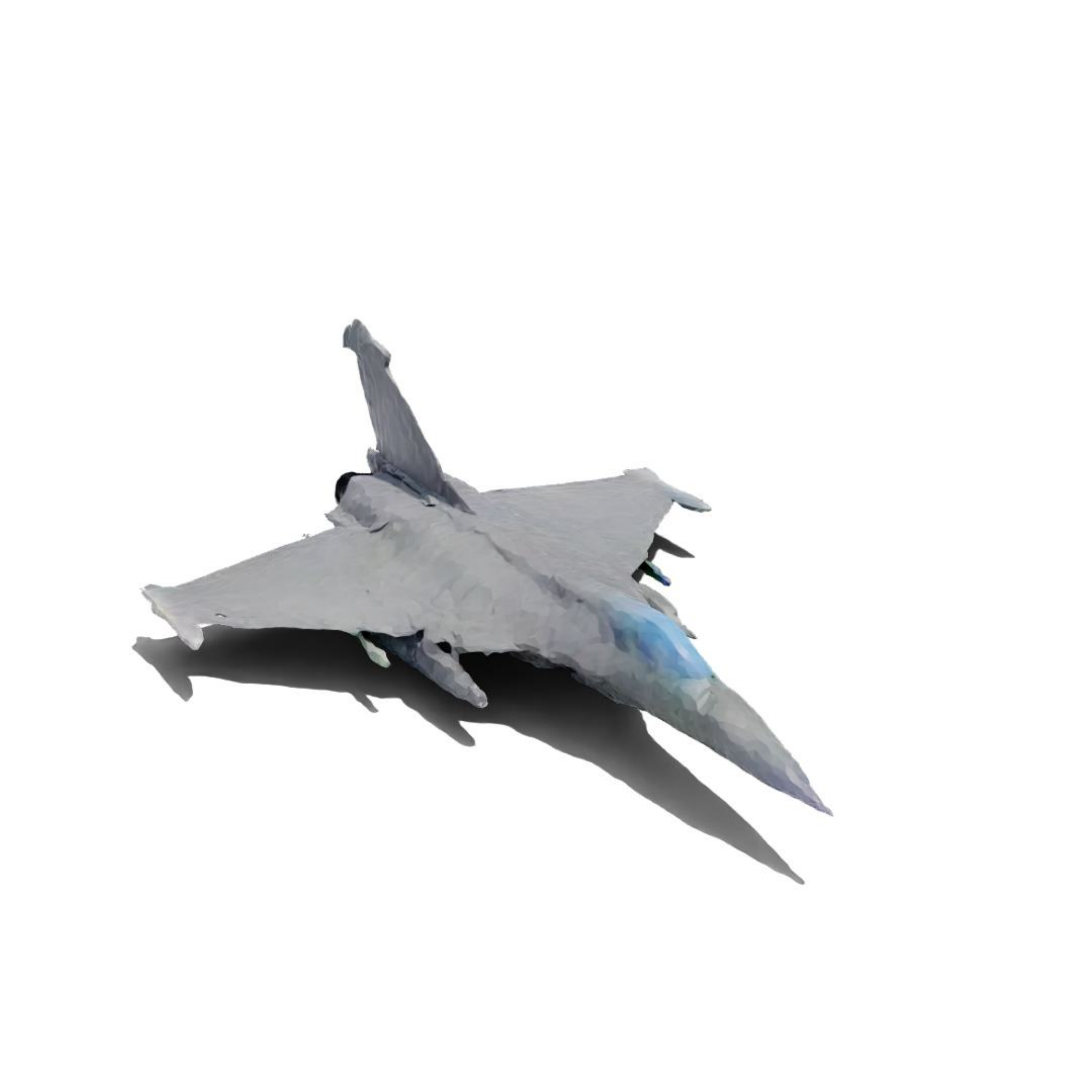}\includegraphics[width=0.16666666666666666\linewidth, trim={0 0cm 0 3cm}, clip]{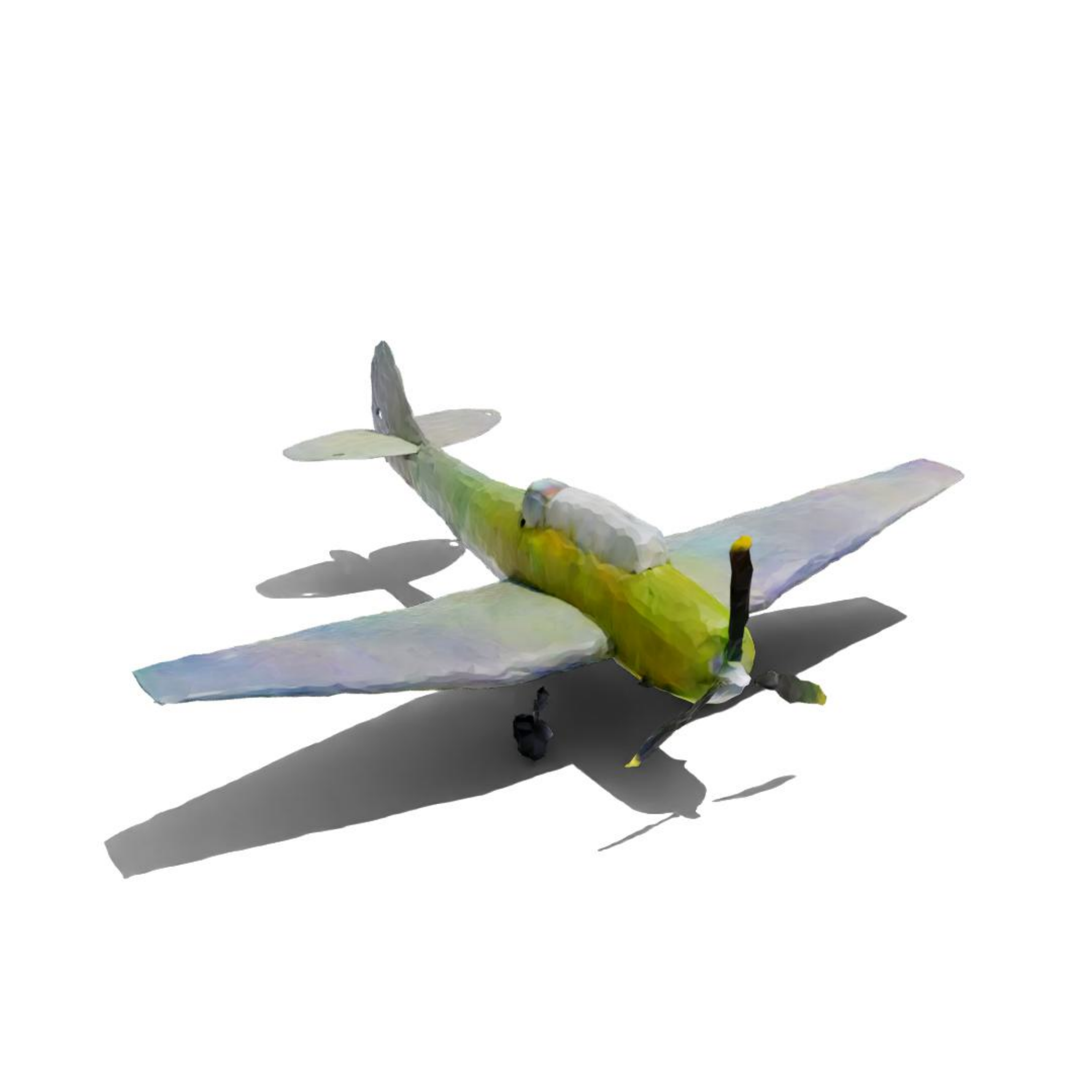}
\caption{\textbf{Random selection of airplanes generated in standard resolution.}}
\label{fig:uncond:airplane:128}
\end{figure*}

\begin{figure*}[!ht]
\centering
\includegraphics[width=0.16666666666666666\linewidth]{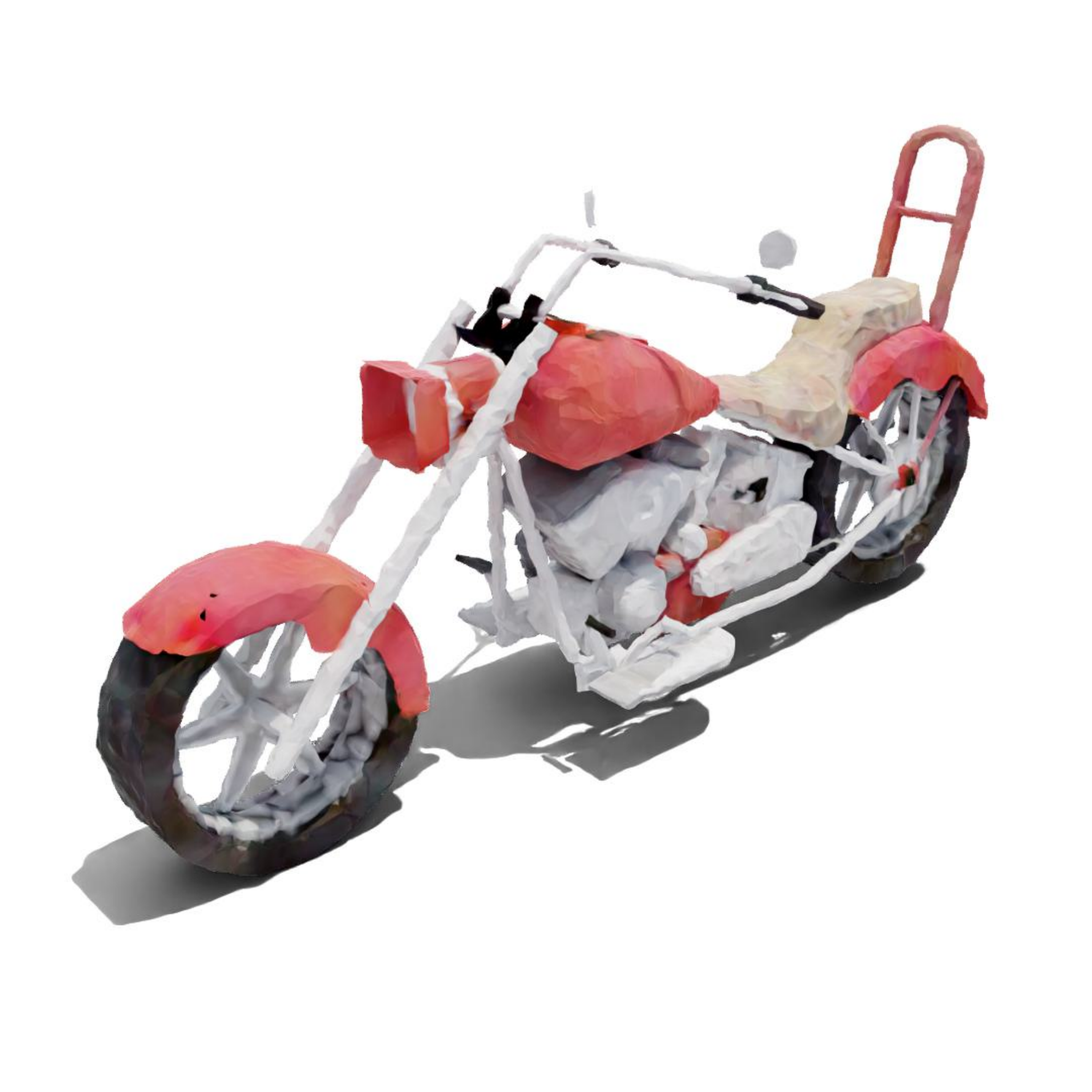}\includegraphics[width=0.16666666666666666\linewidth]{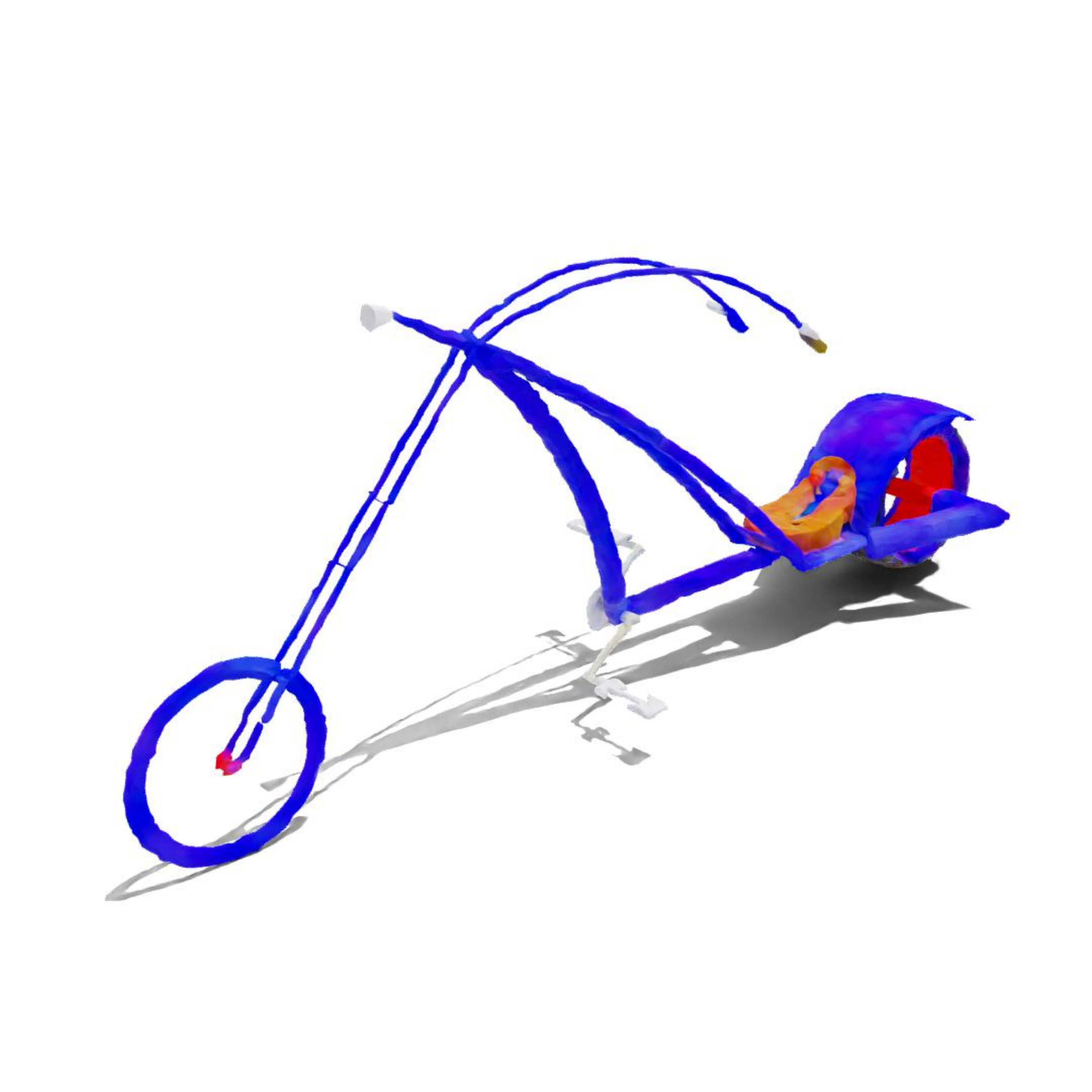}\includegraphics[width=0.16666666666666666\linewidth]{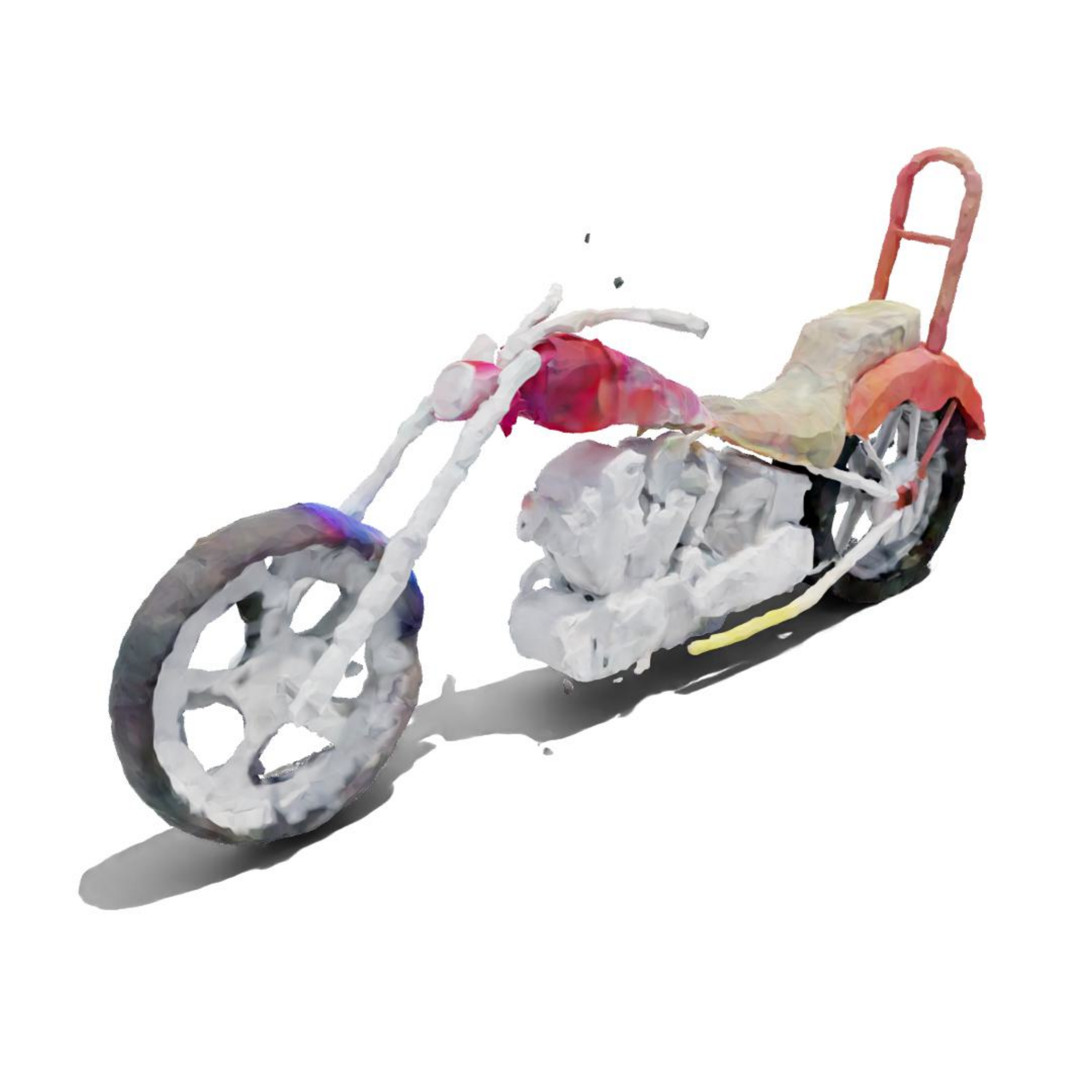}\includegraphics[width=0.16666666666666666\linewidth]{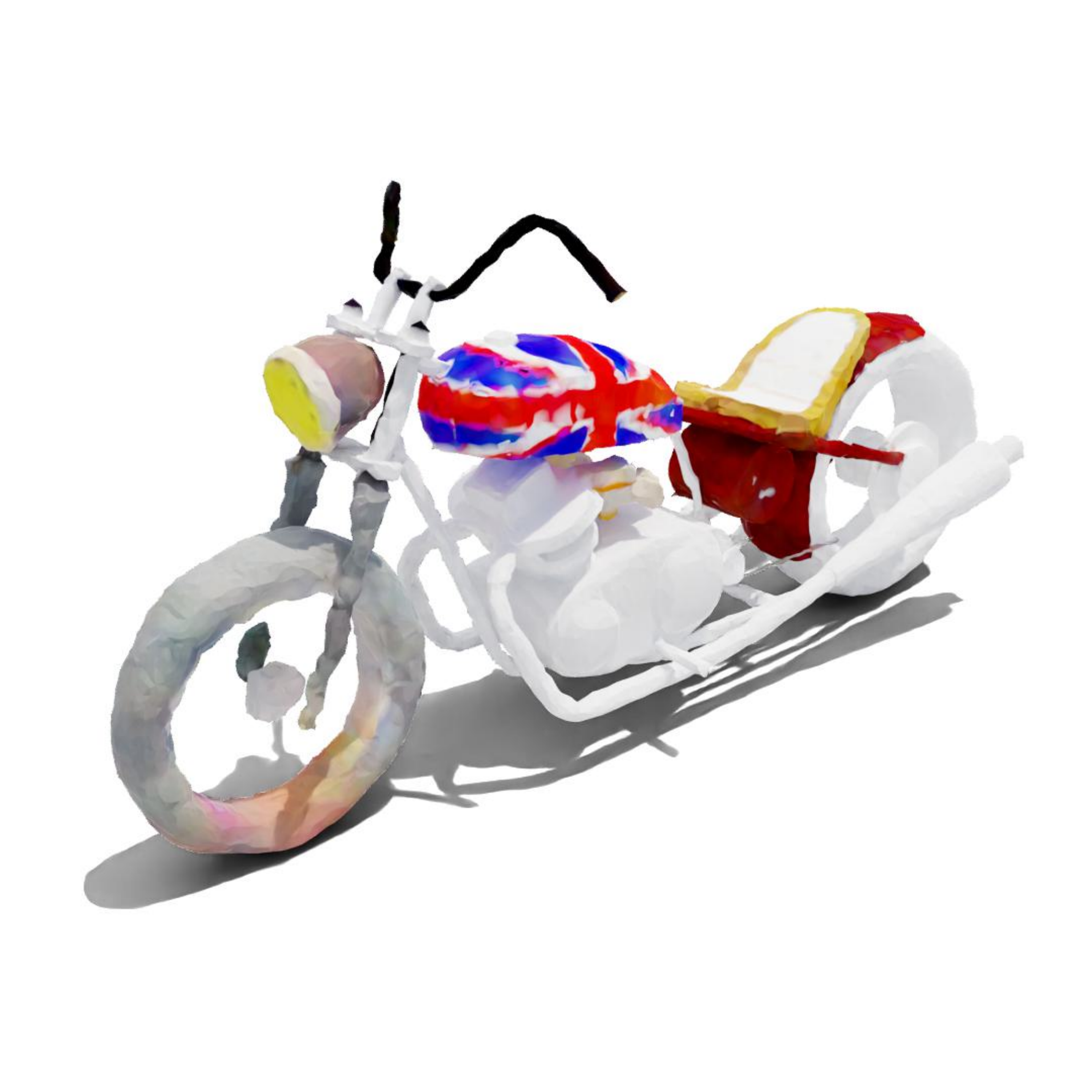}\includegraphics[width=0.16666666666666666\linewidth]{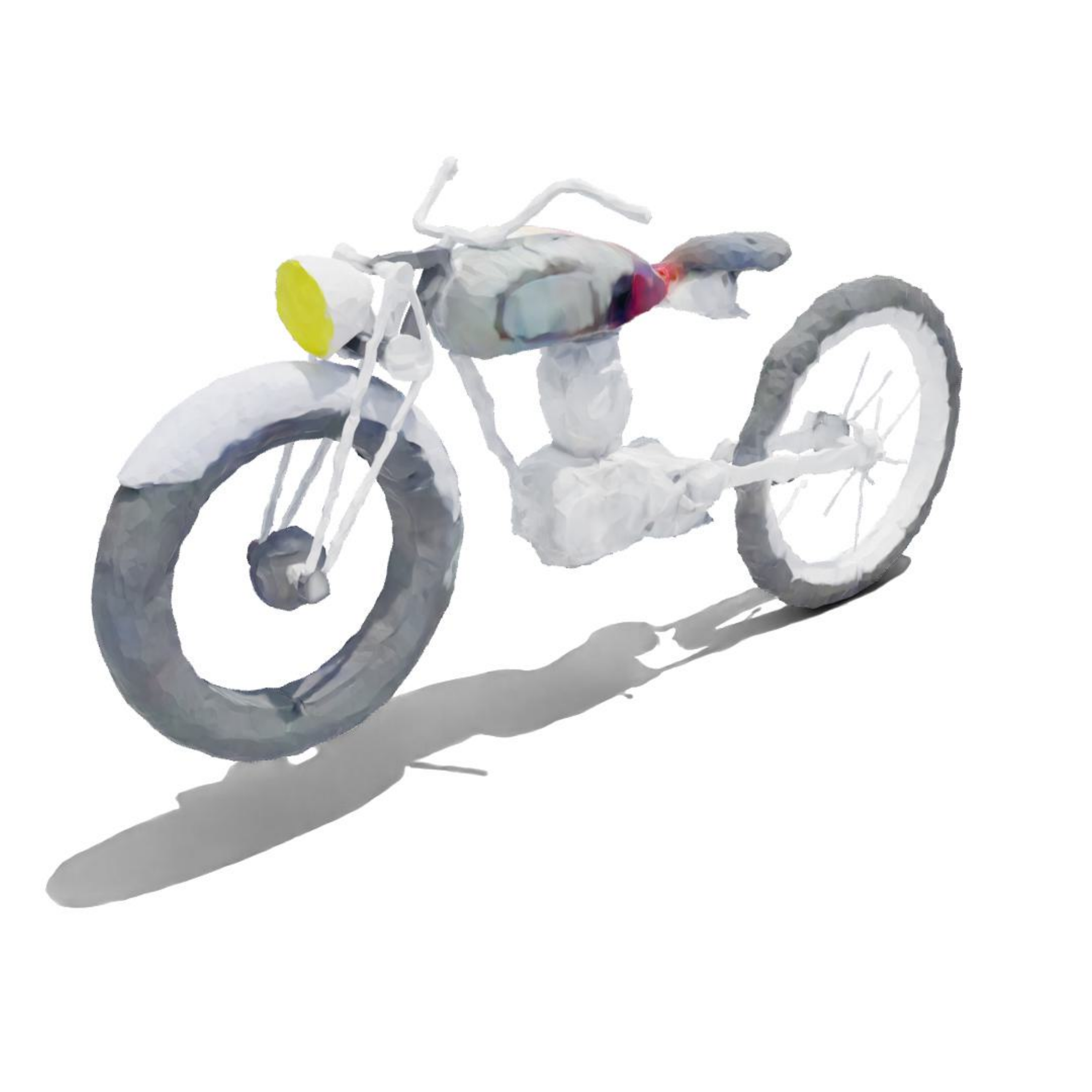}\includegraphics[width=0.16666666666666666\linewidth]{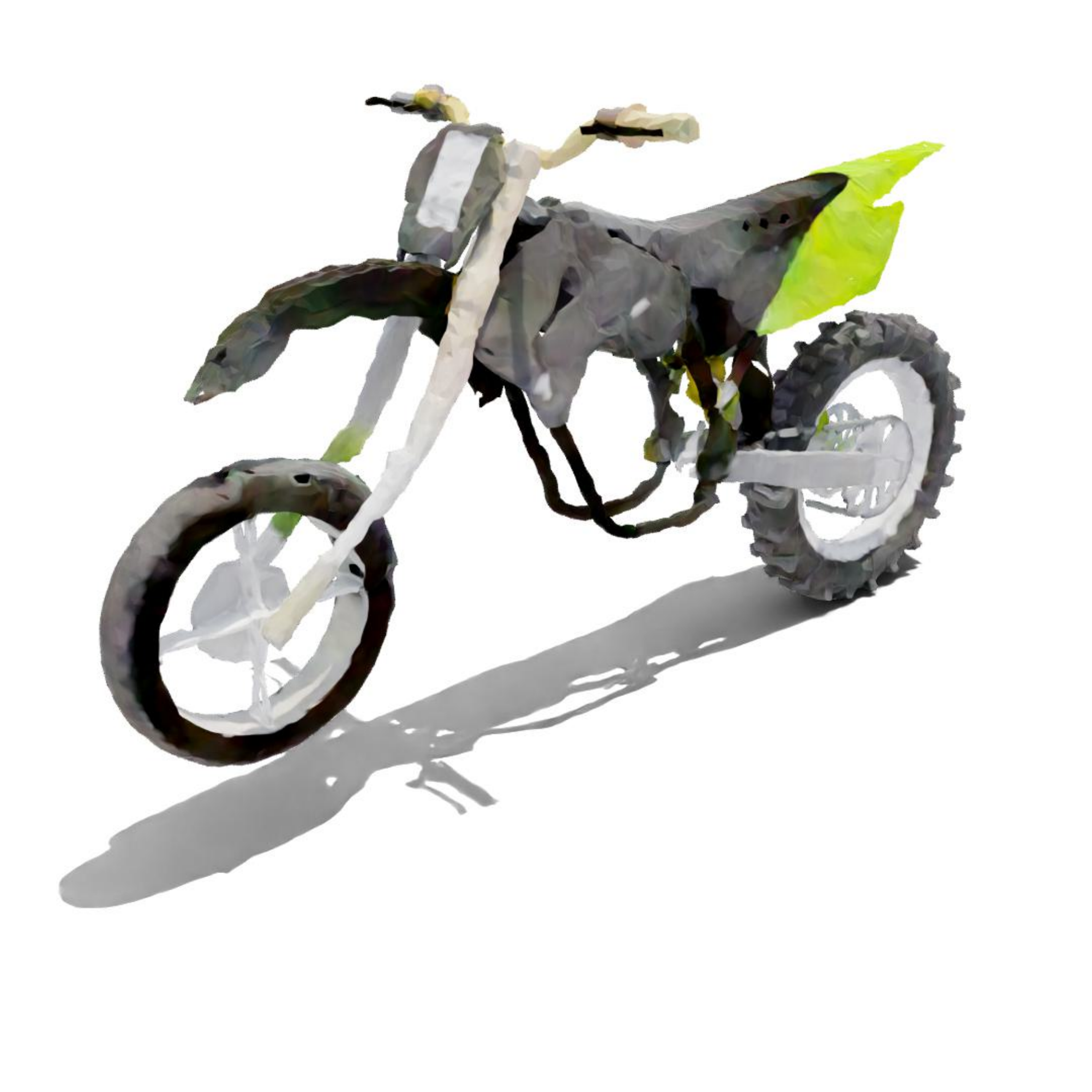}\\

\vspace{-0.4cm}
\includegraphics[width=0.16666666666666666\linewidth]{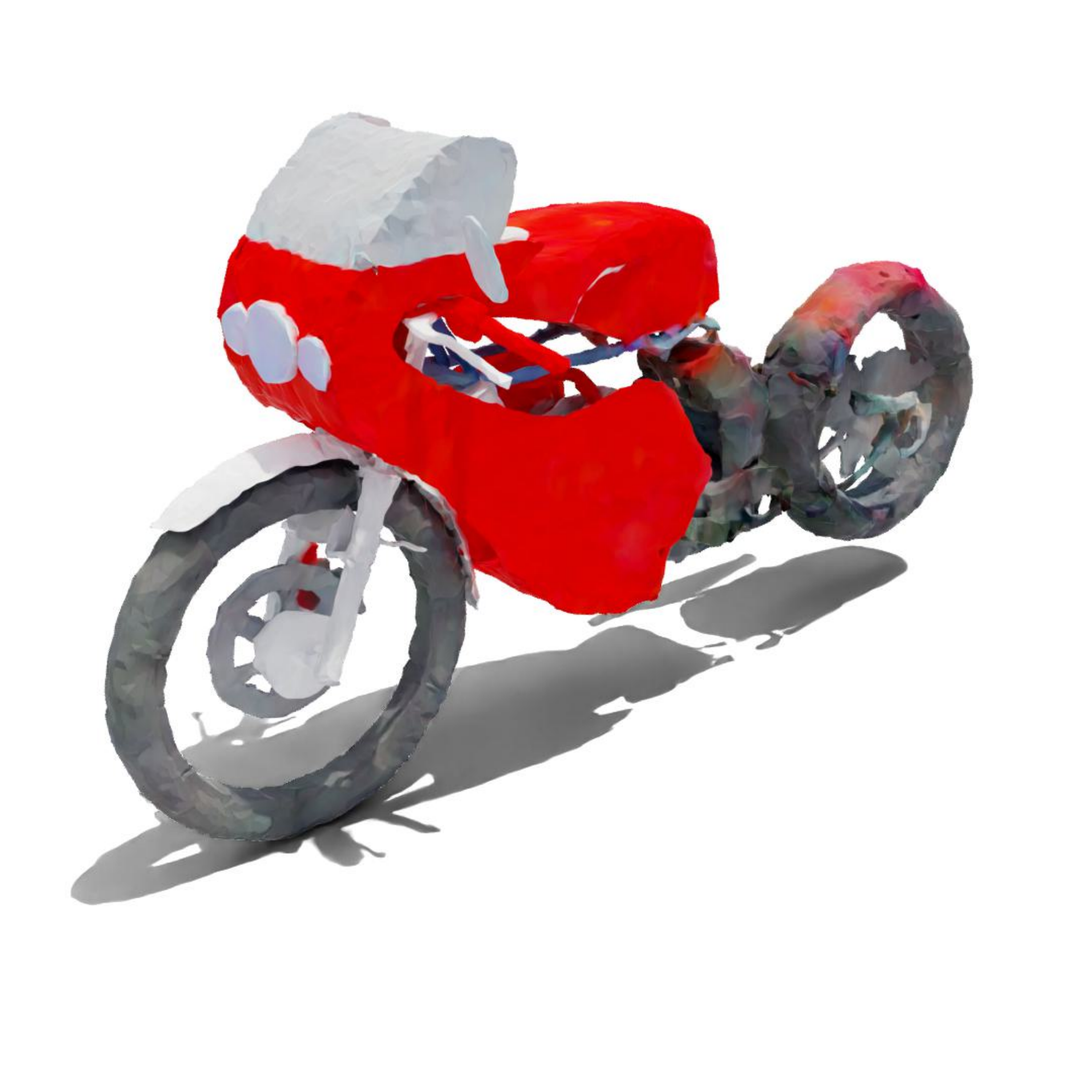}\includegraphics[width=0.16666666666666666\linewidth]{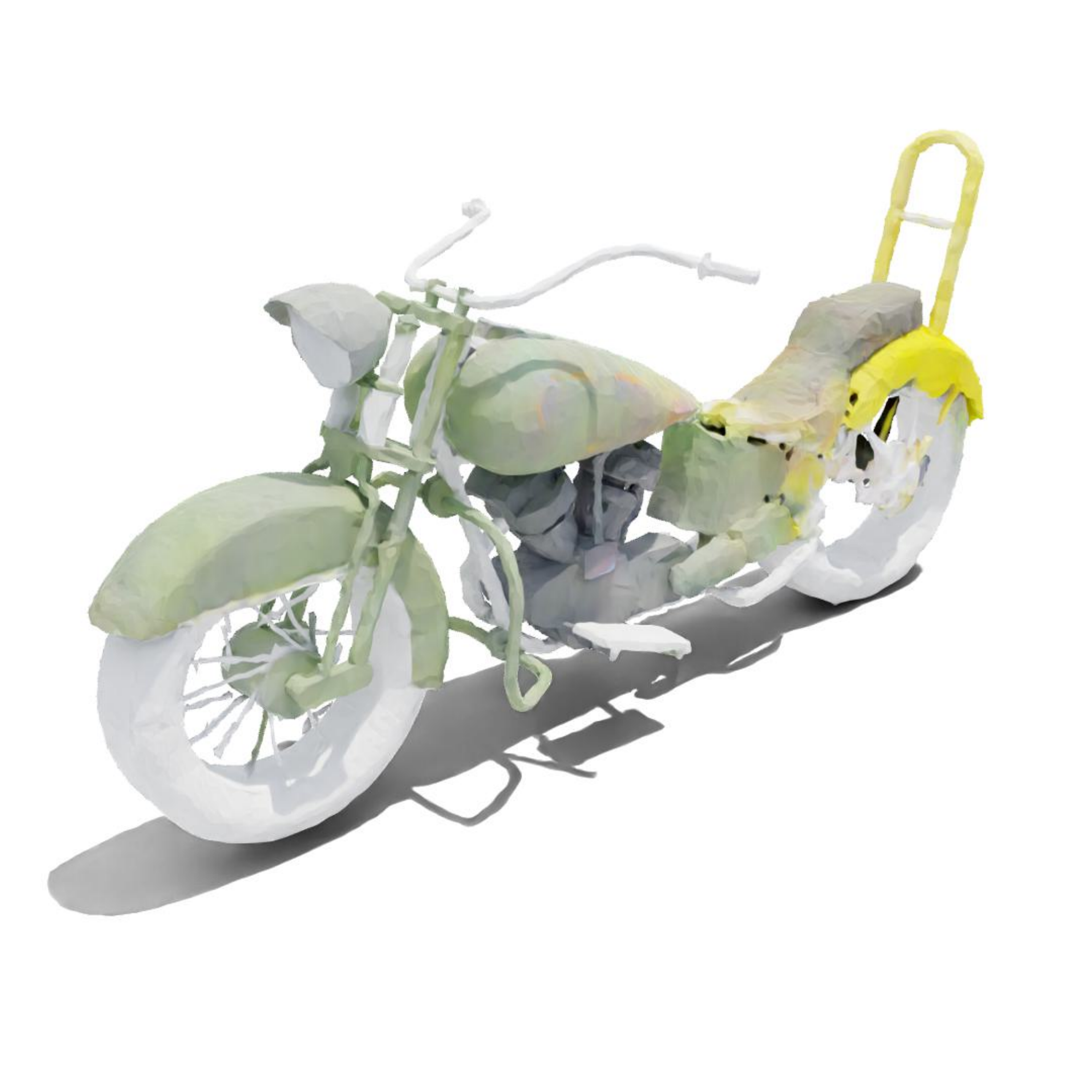}\includegraphics[width=0.16666666666666666\linewidth]{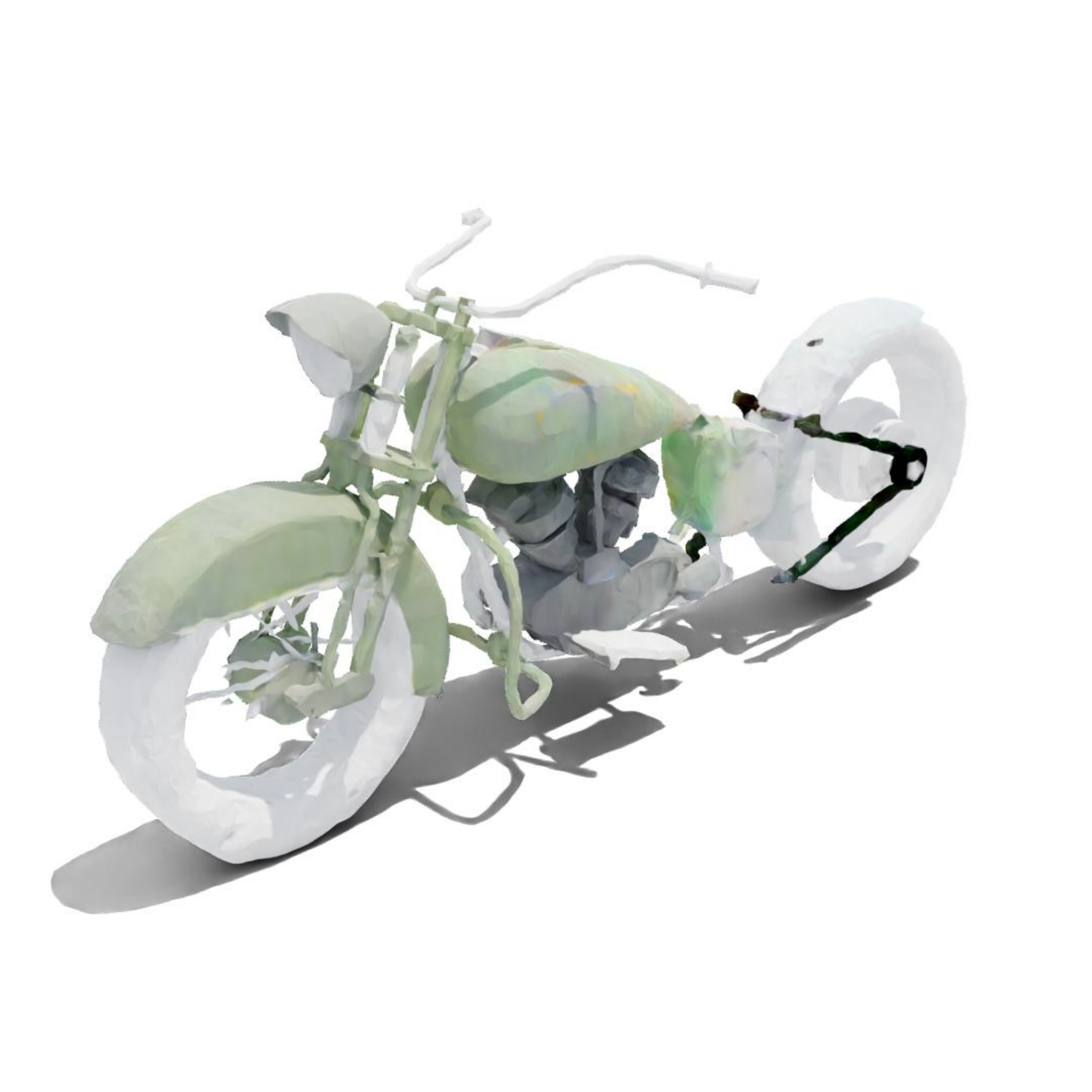}\includegraphics[width=0.16666666666666666\linewidth]{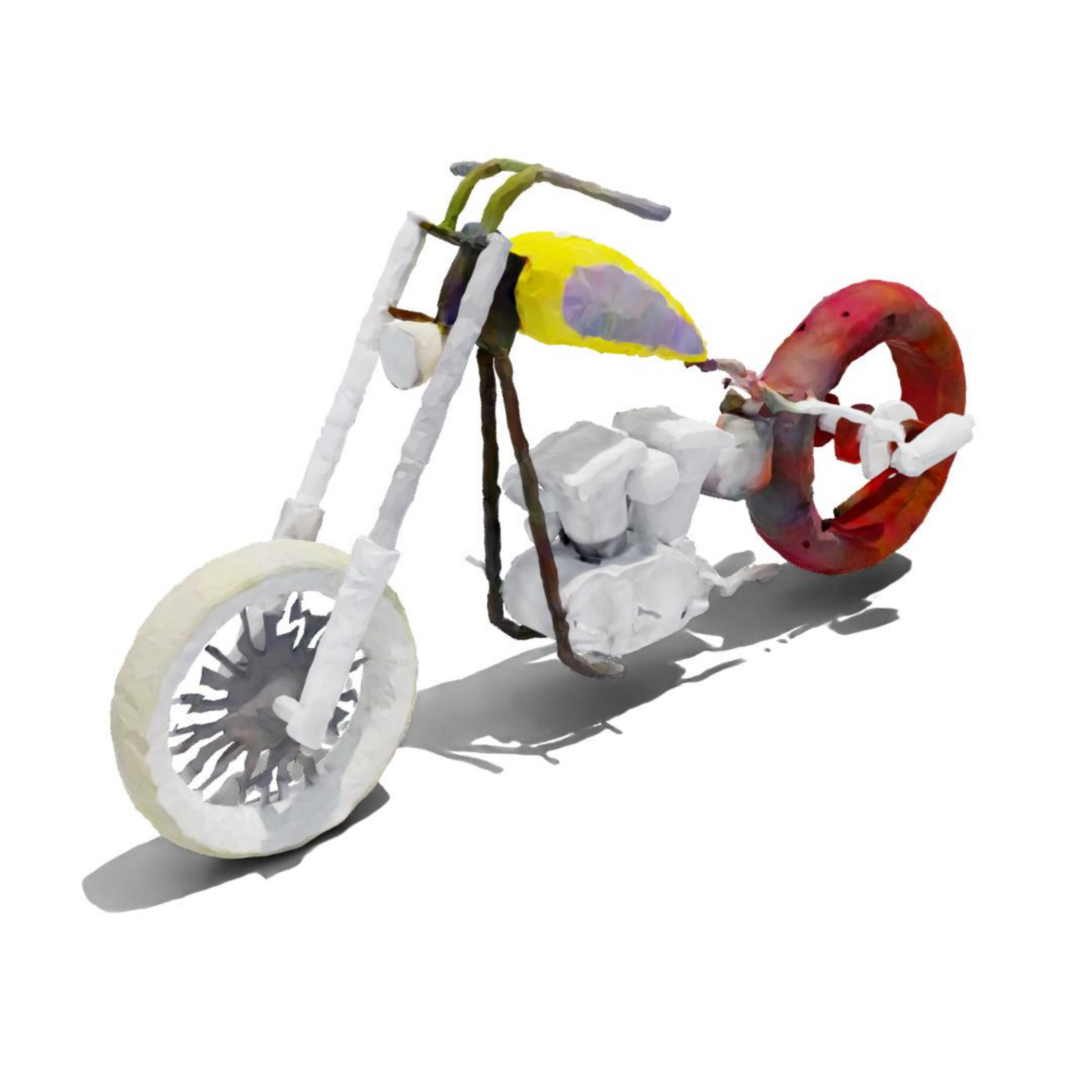}\includegraphics[width=0.16666666666666666\linewidth]{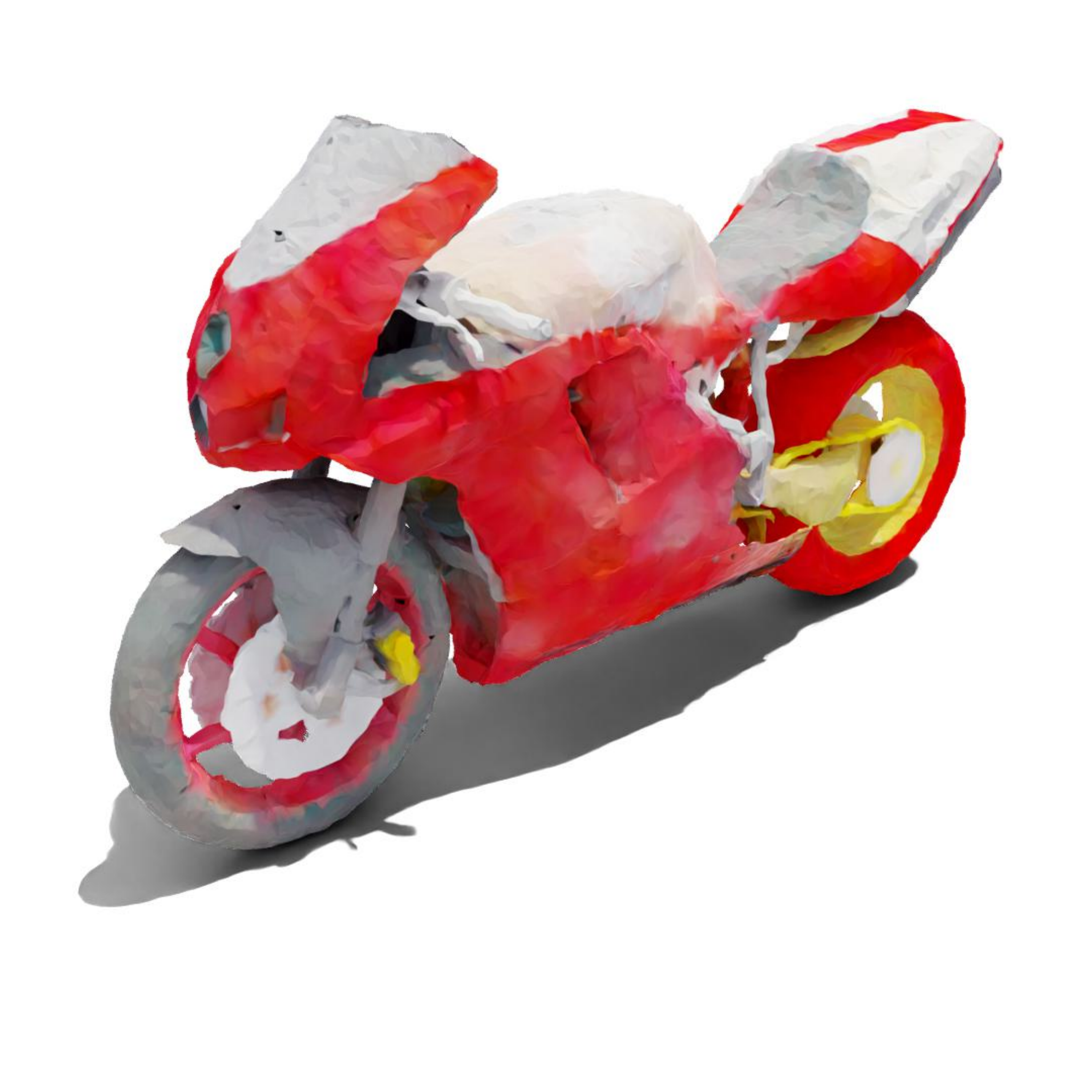}\includegraphics[width=0.16666666666666666\linewidth]{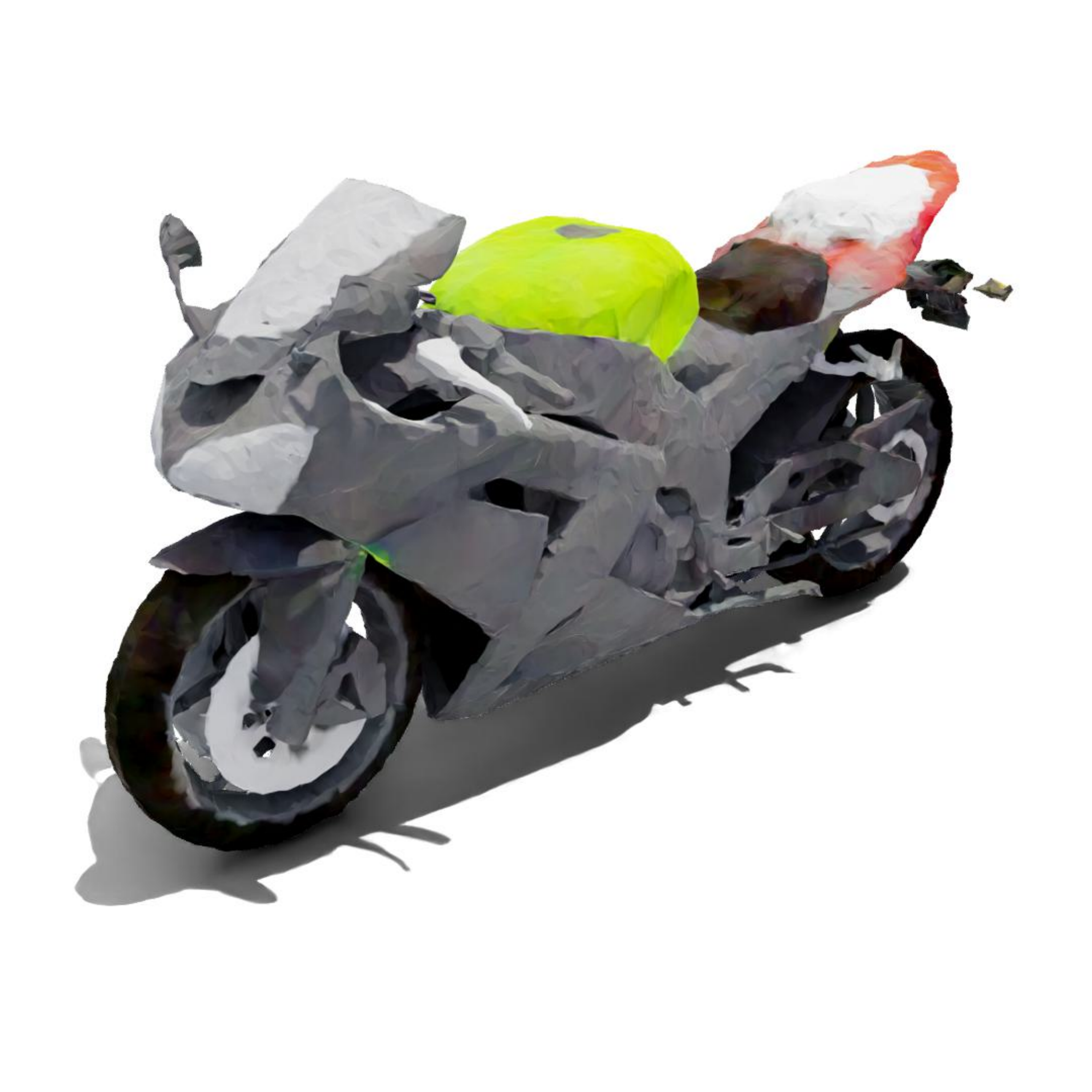}\\

\vspace{-0.4cm}
\includegraphics[width=0.16666666666666666\linewidth]{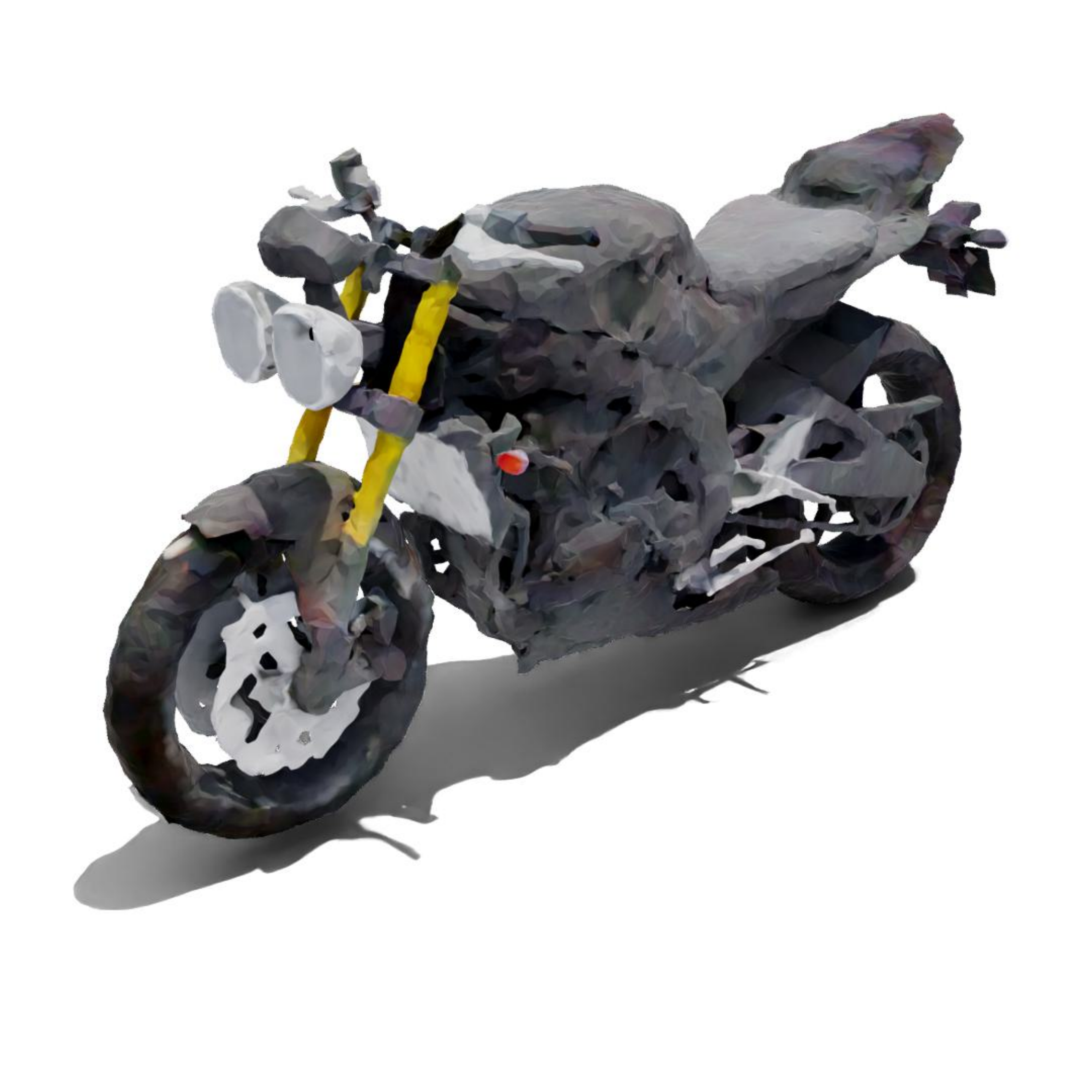}\includegraphics[width=0.16666666666666666\linewidth]{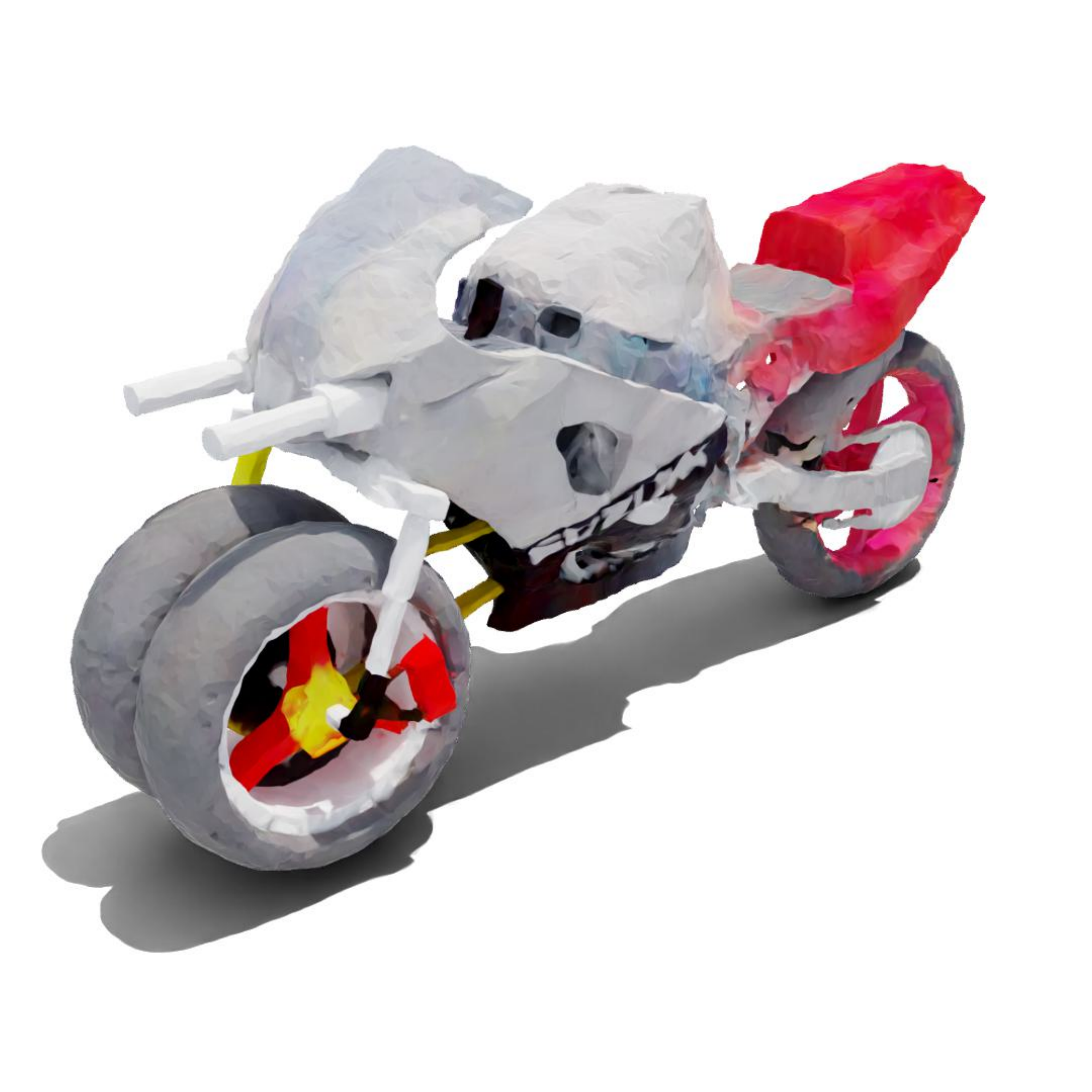}\includegraphics[width=0.16666666666666666\linewidth]{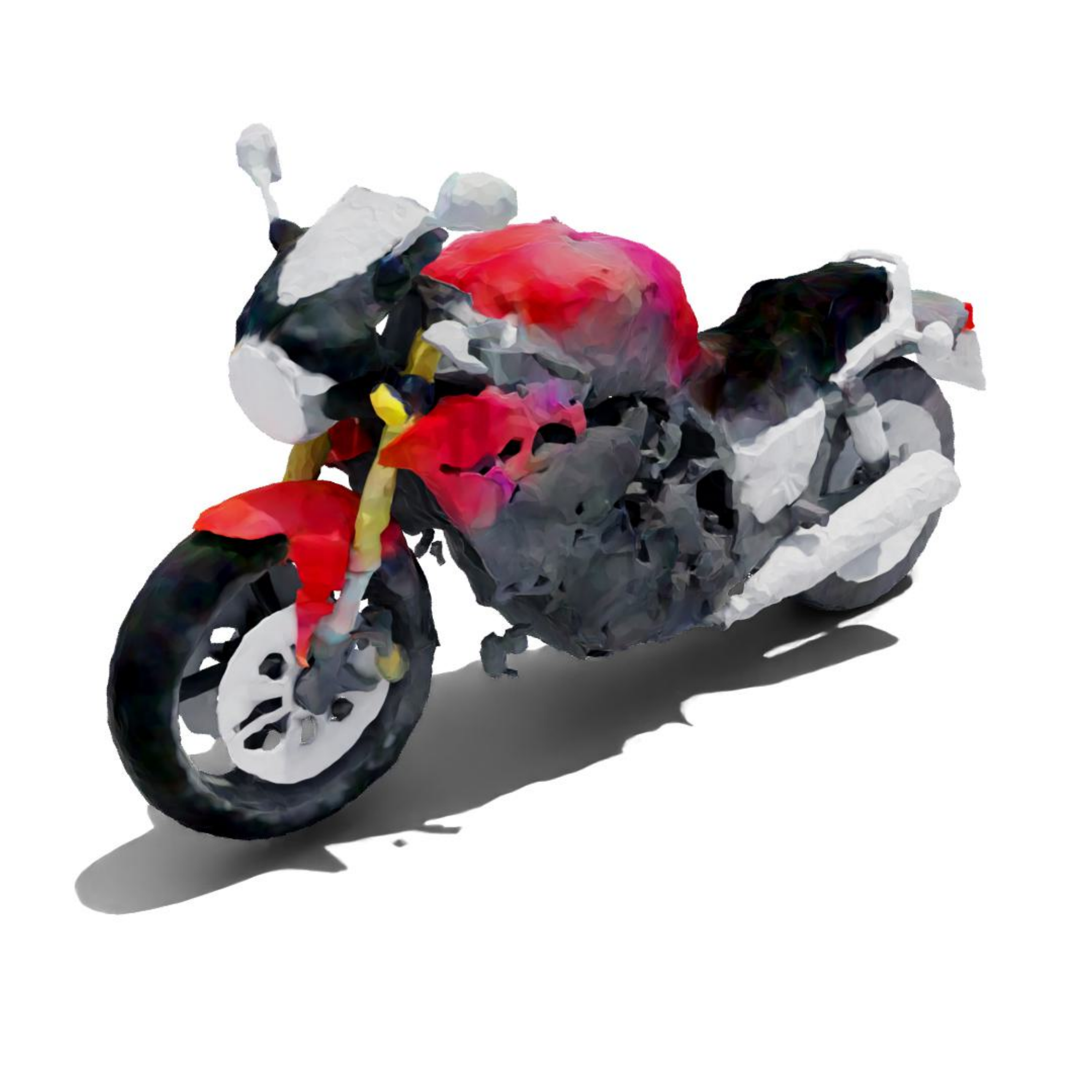}\includegraphics[width=0.16666666666666666\linewidth]{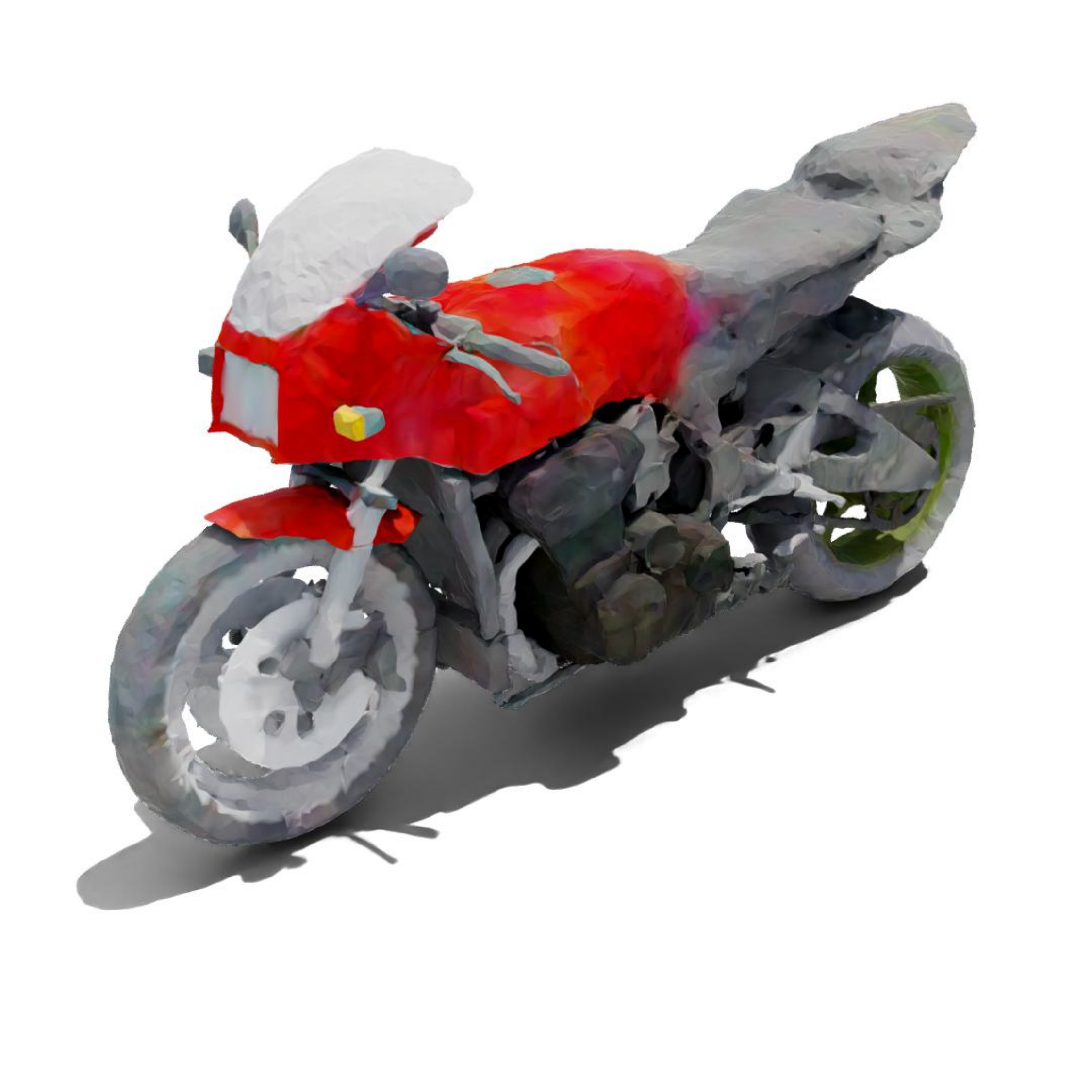}\includegraphics[width=0.16666666666666666\linewidth]{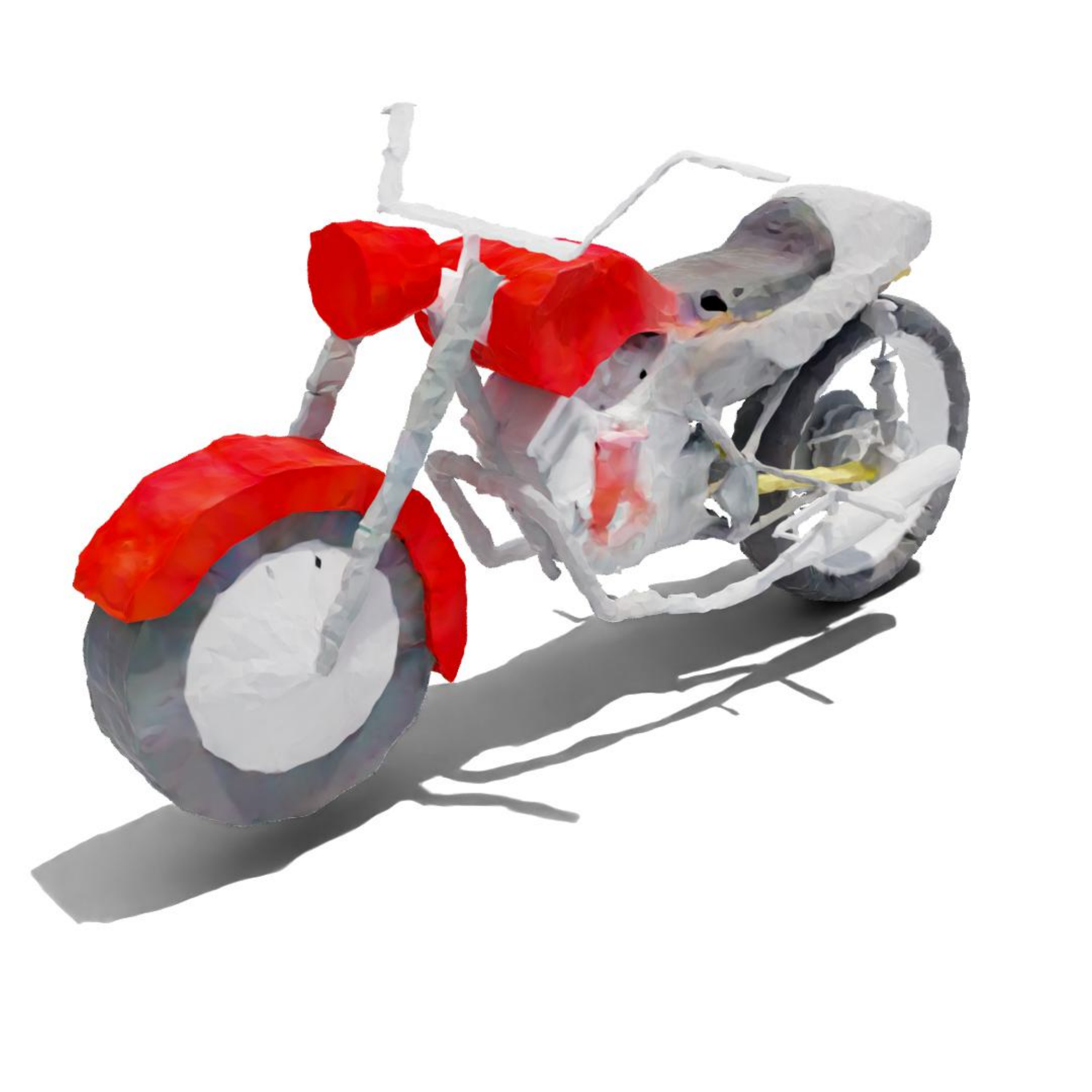}\includegraphics[width=0.16666666666666666\linewidth]{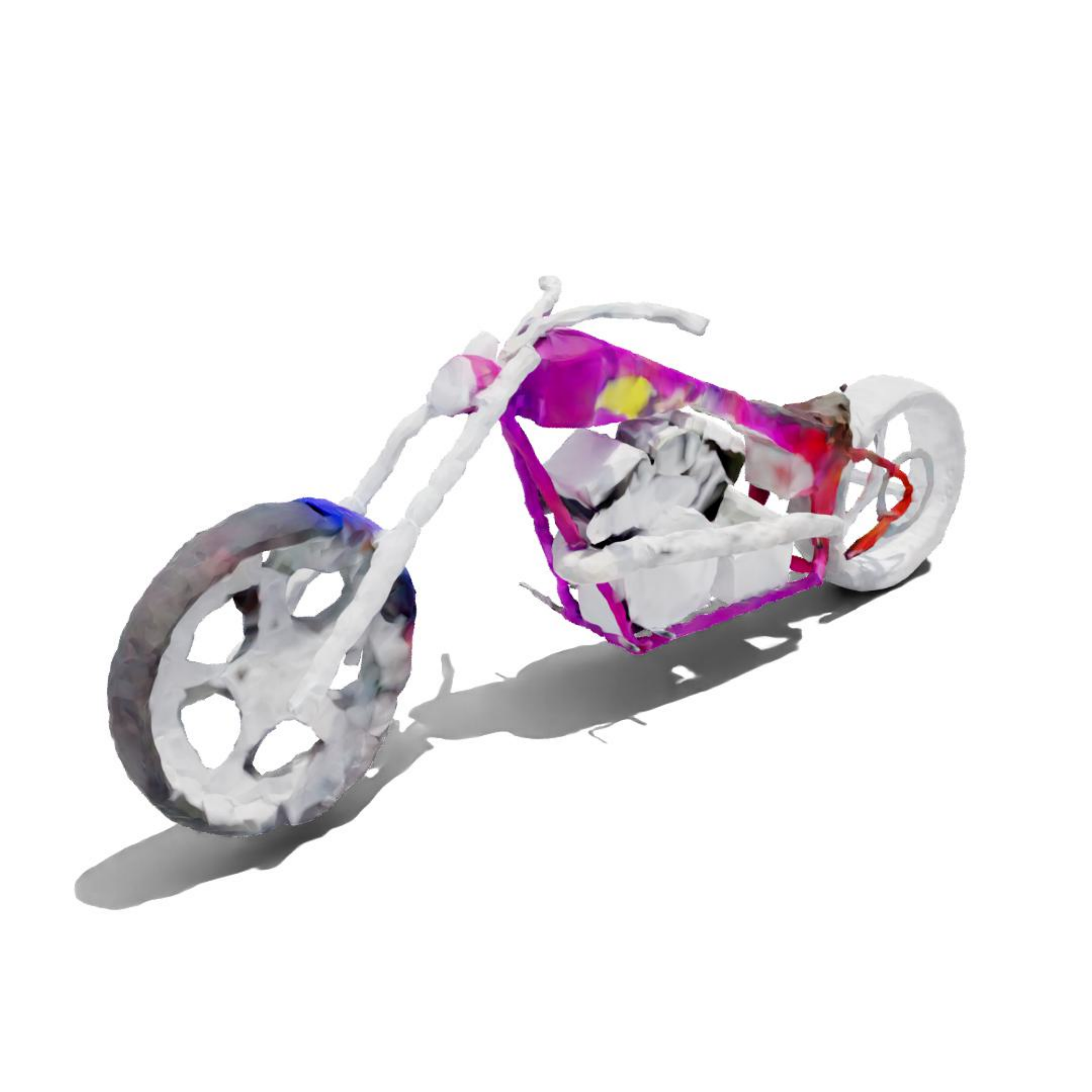}\\

\vspace{-0.4cm}
\includegraphics[width=0.16666666666666666\linewidth]{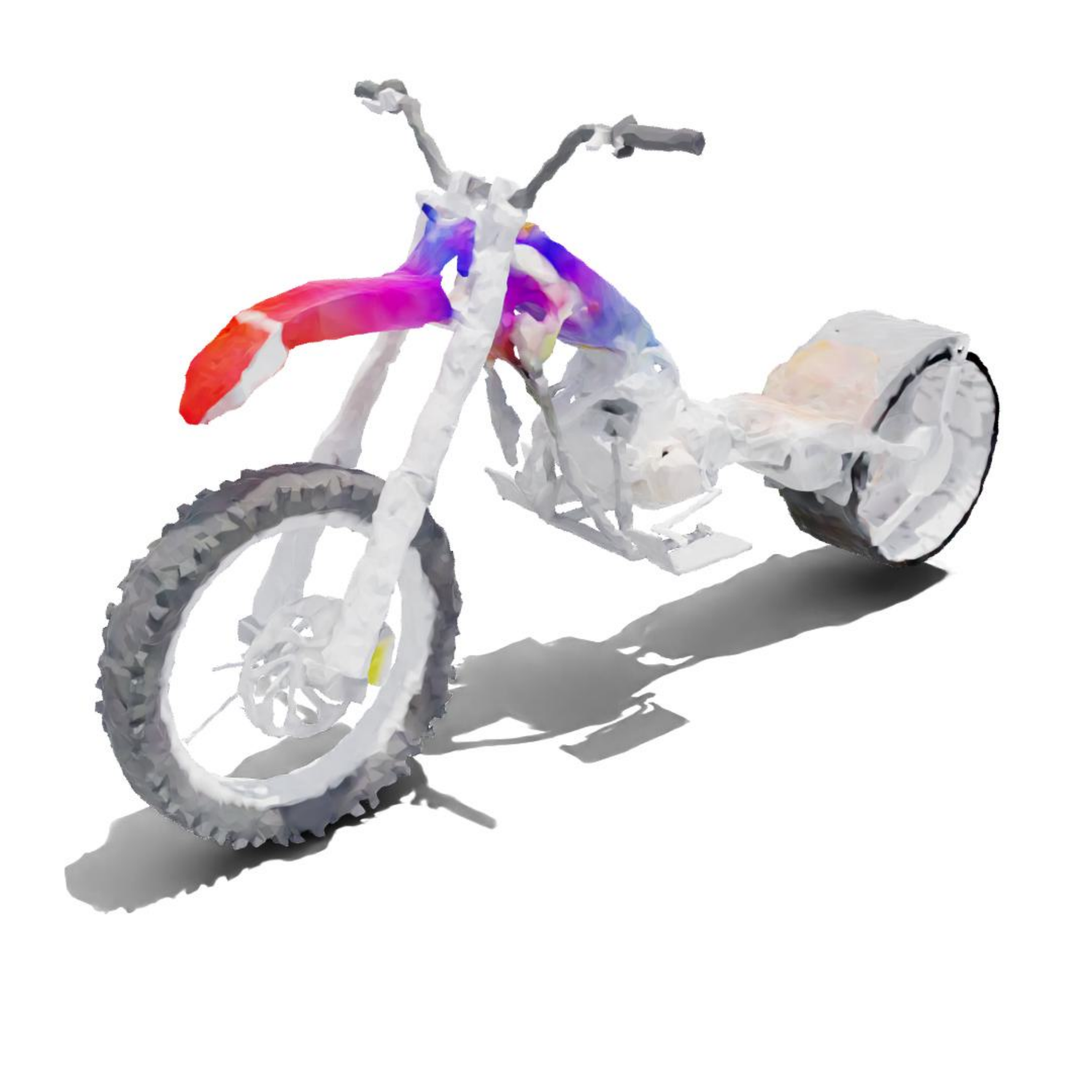}\includegraphics[width=0.16666666666666666\linewidth]{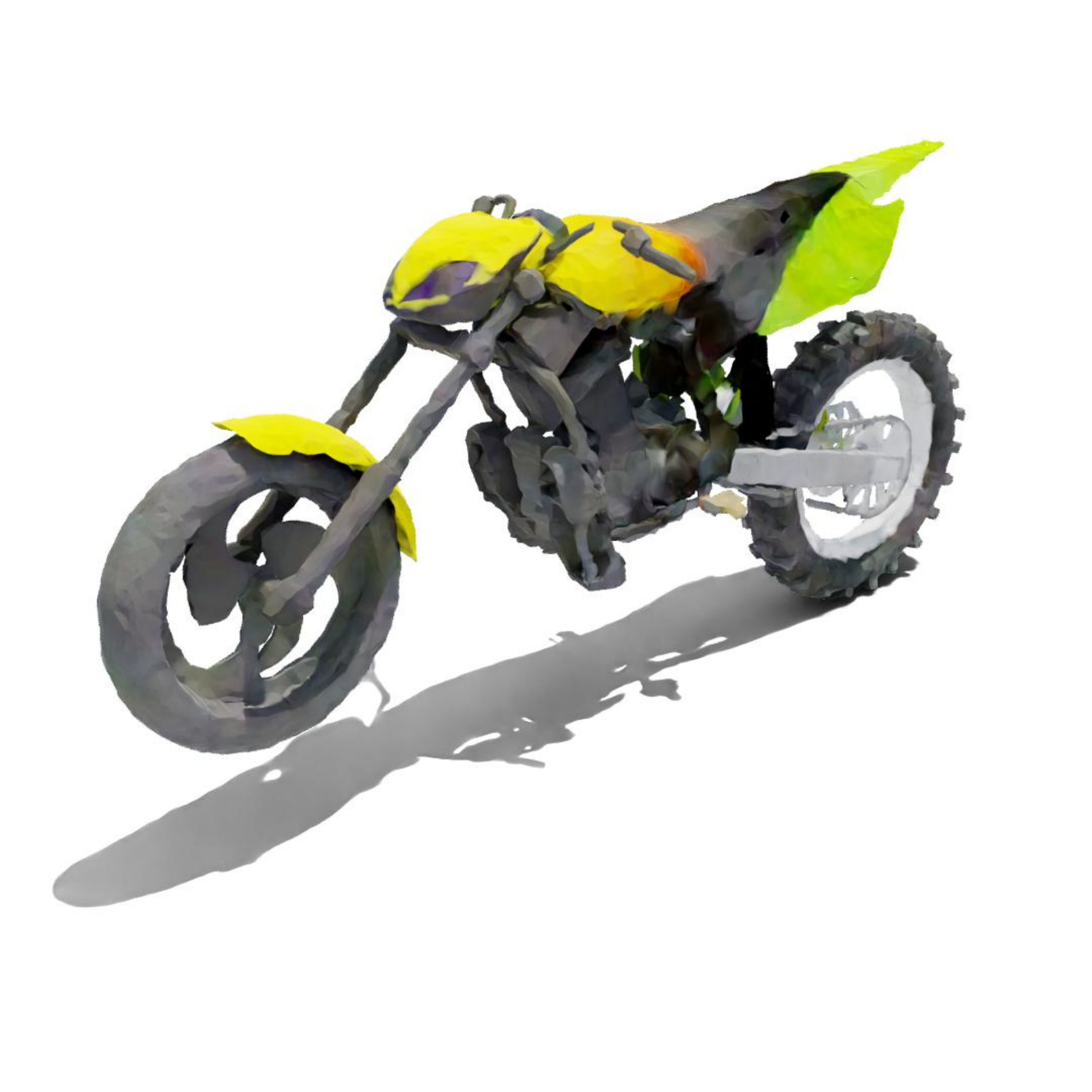}\includegraphics[width=0.16666666666666666\linewidth]{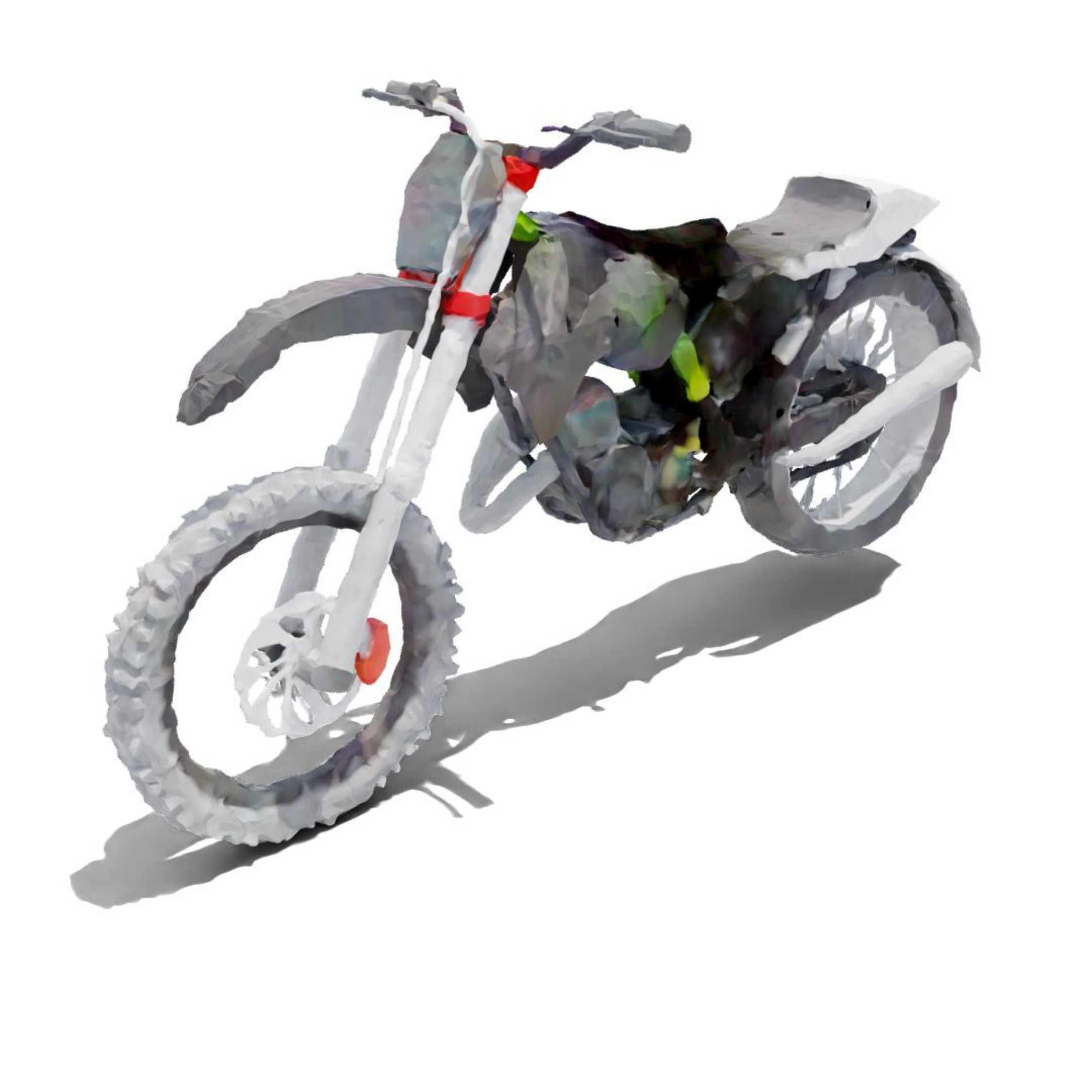}\includegraphics[width=0.16666666666666666\linewidth]{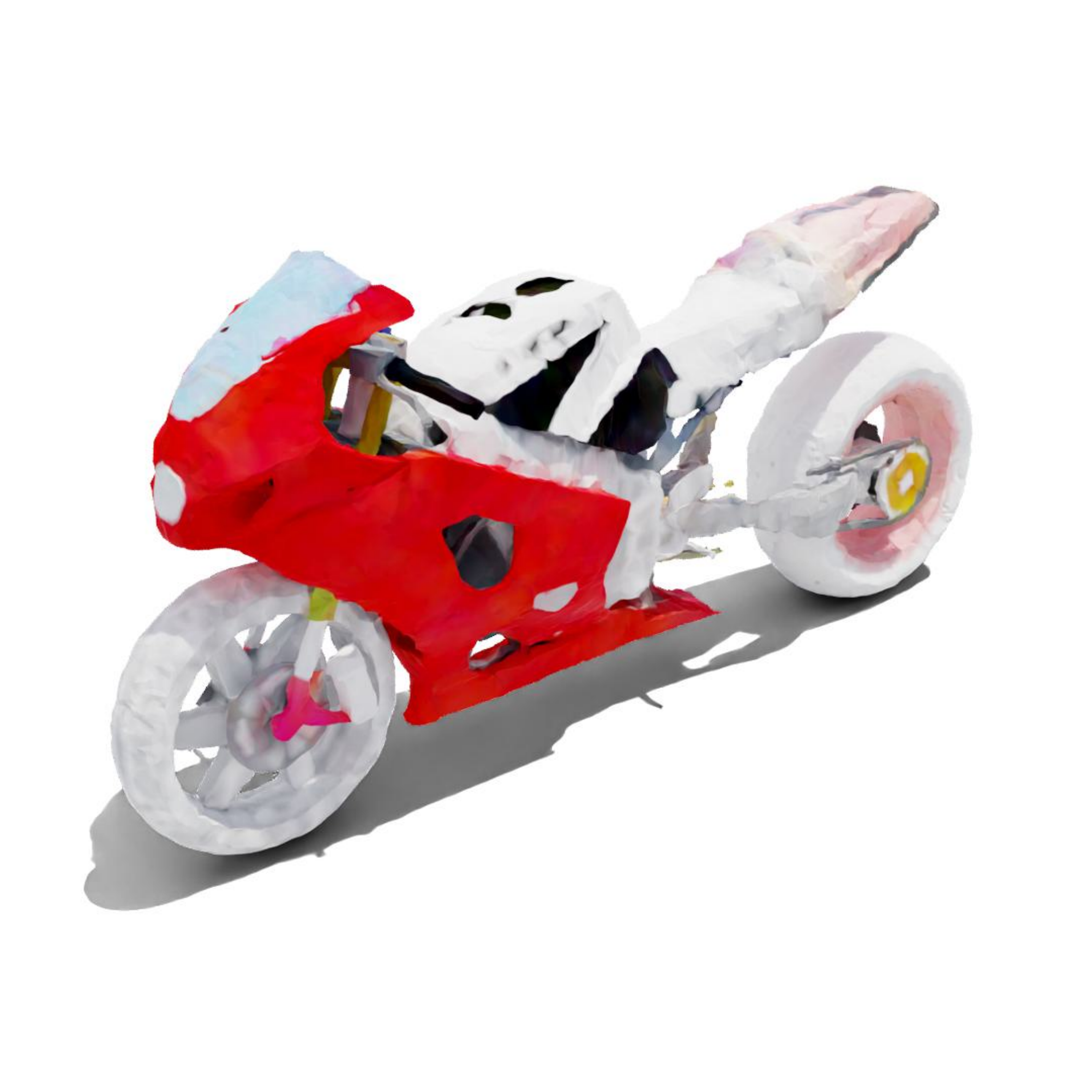}\includegraphics[width=0.16666666666666666\linewidth]{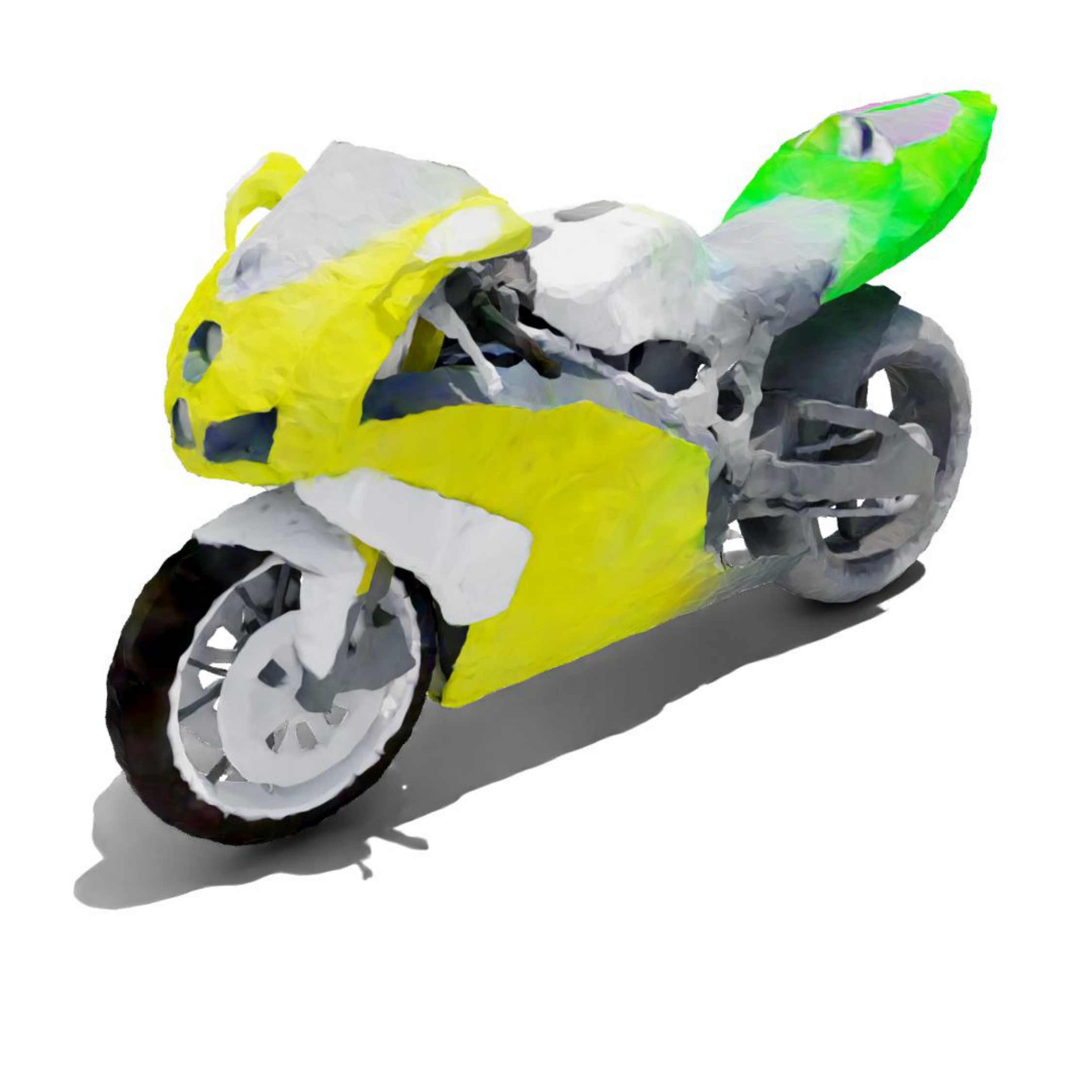}\includegraphics[width=0.16666666666666666\linewidth]{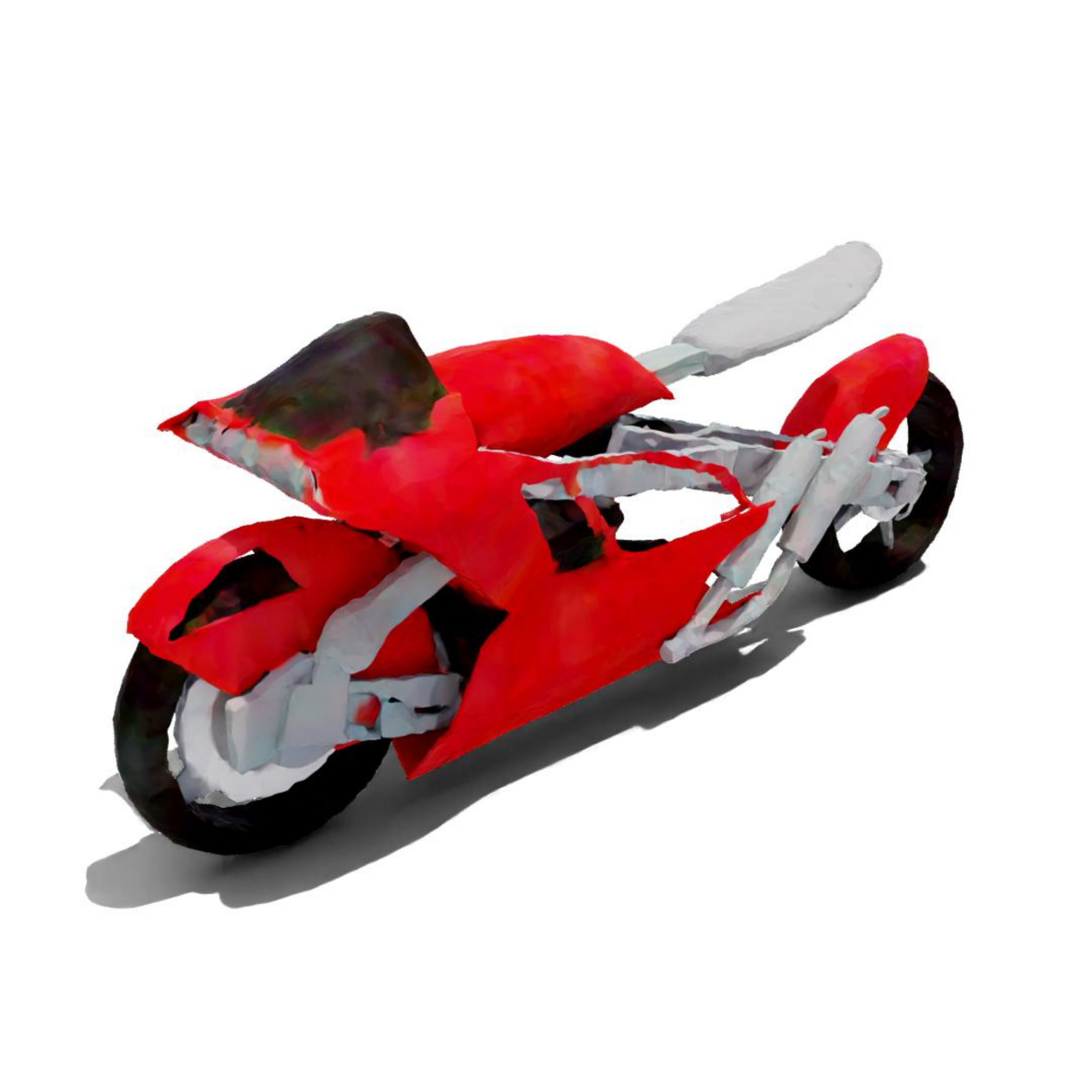}
\caption{\textbf{Random selection of motorbikes generated in high resolution.}}
\label{fig:uncond:bike:192}
\end{figure*}

\begin{figure*}[!ht]
\centering
\includegraphics[width=0.16666666666666666\linewidth]{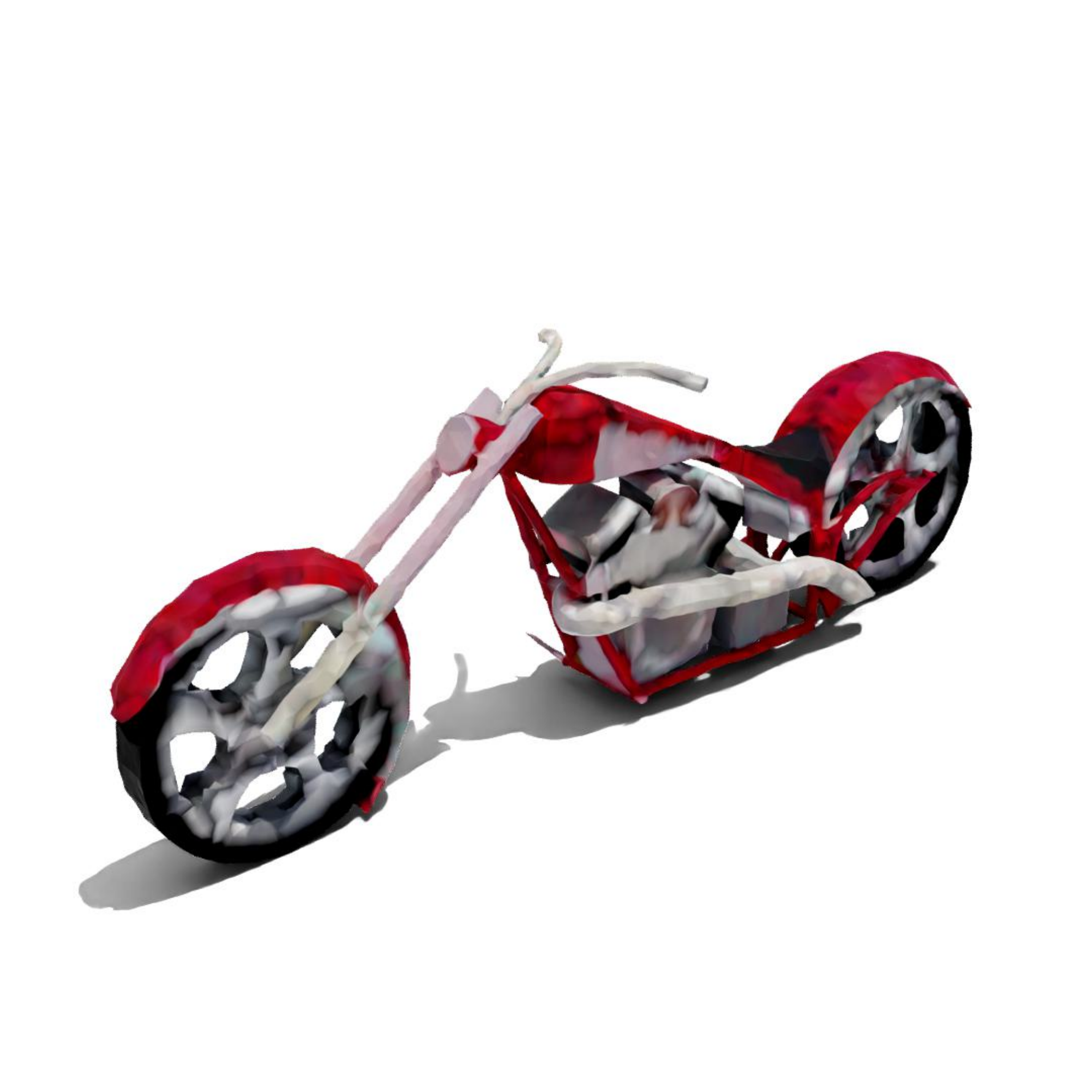}\includegraphics[width=0.16666666666666666\linewidth]{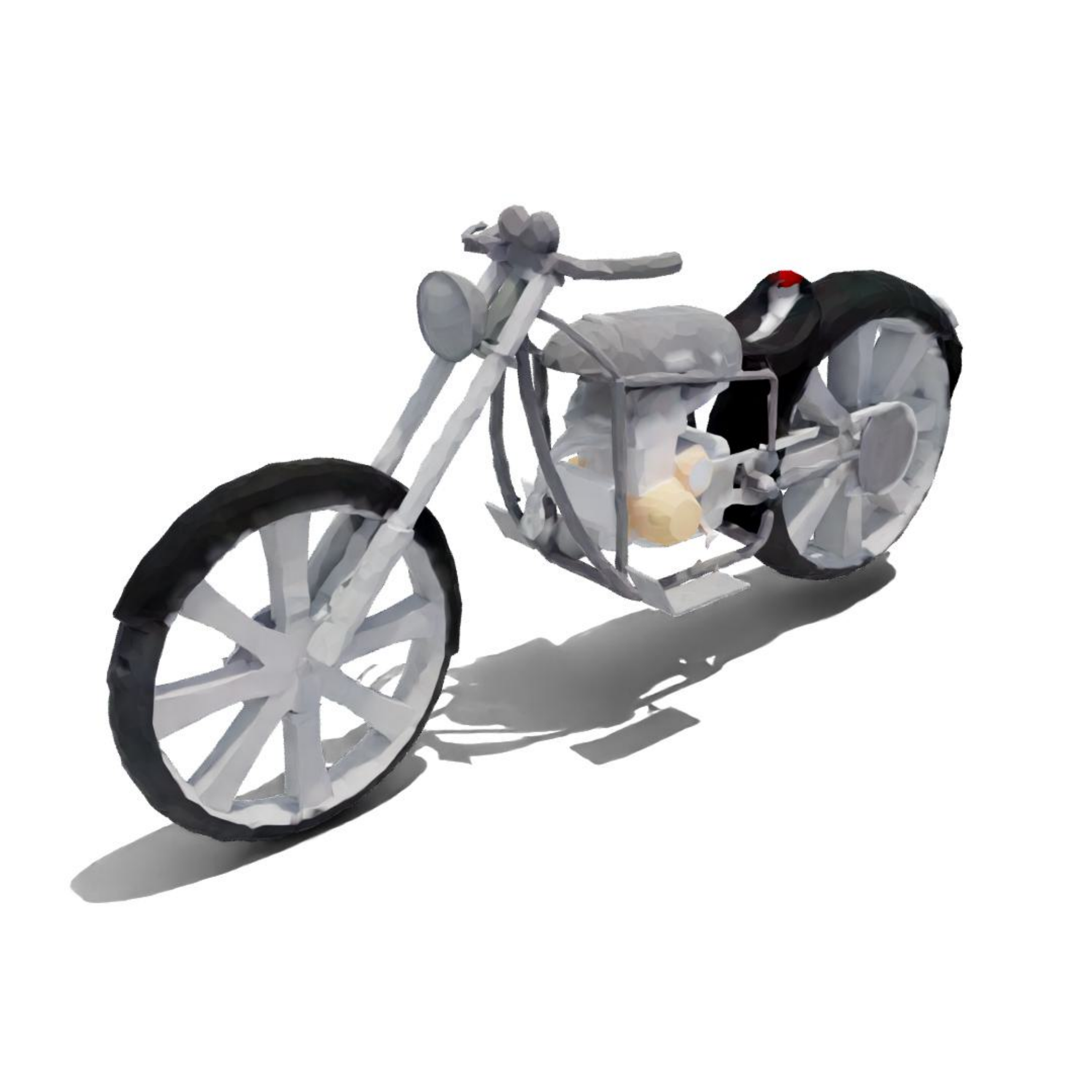}\includegraphics[width=0.16666666666666666\linewidth]{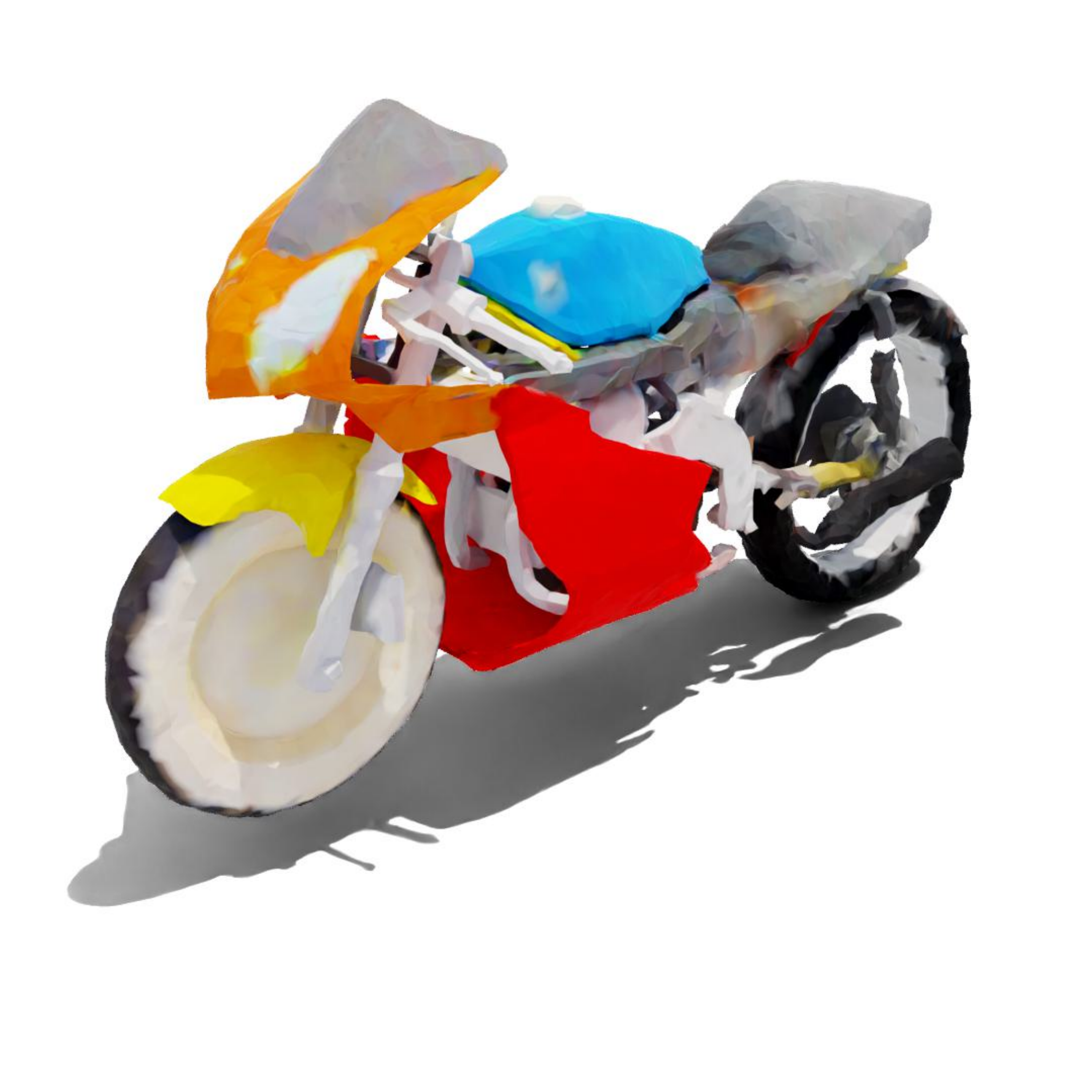}\includegraphics[width=0.16666666666666666\linewidth]{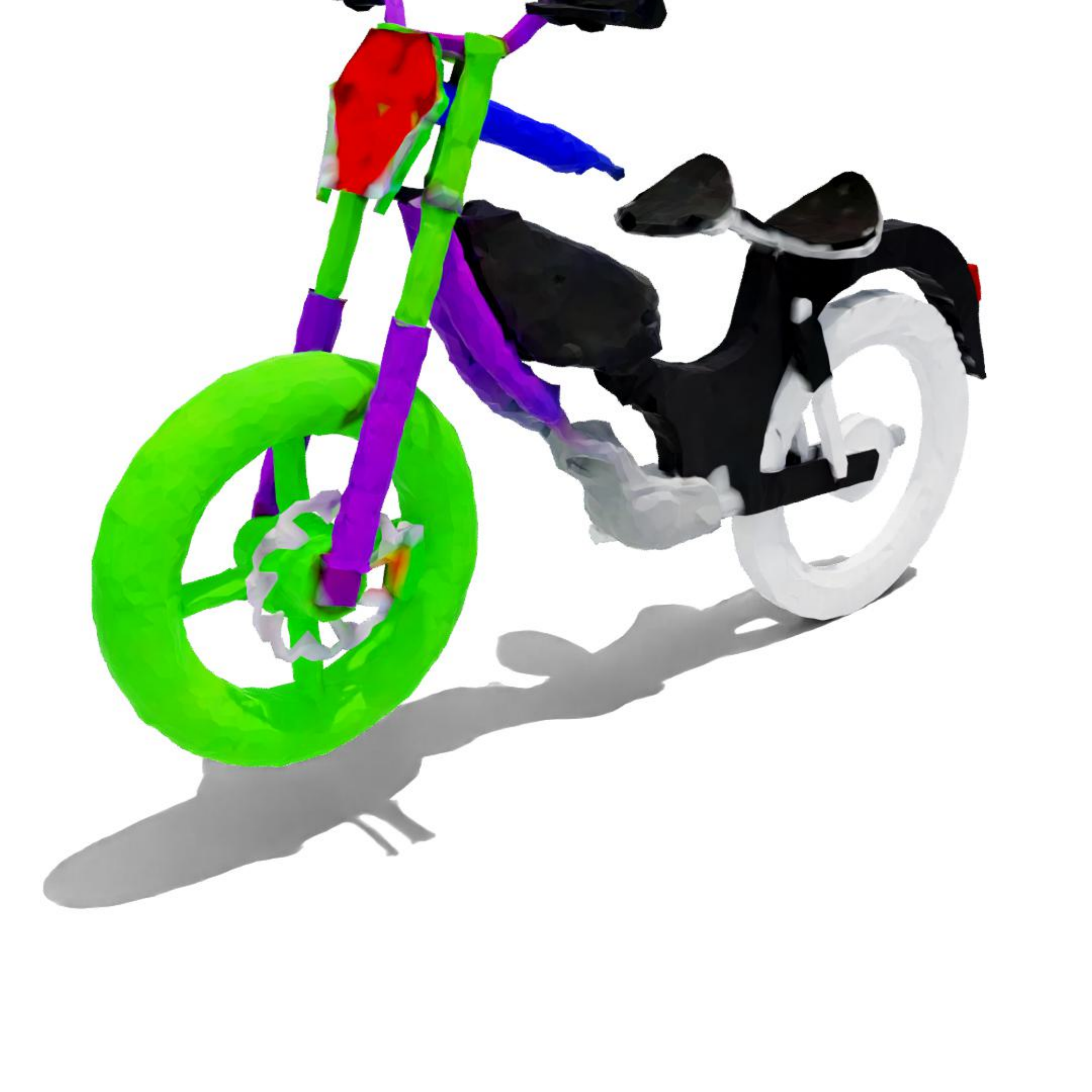}\includegraphics[width=0.16666666666666666\linewidth]{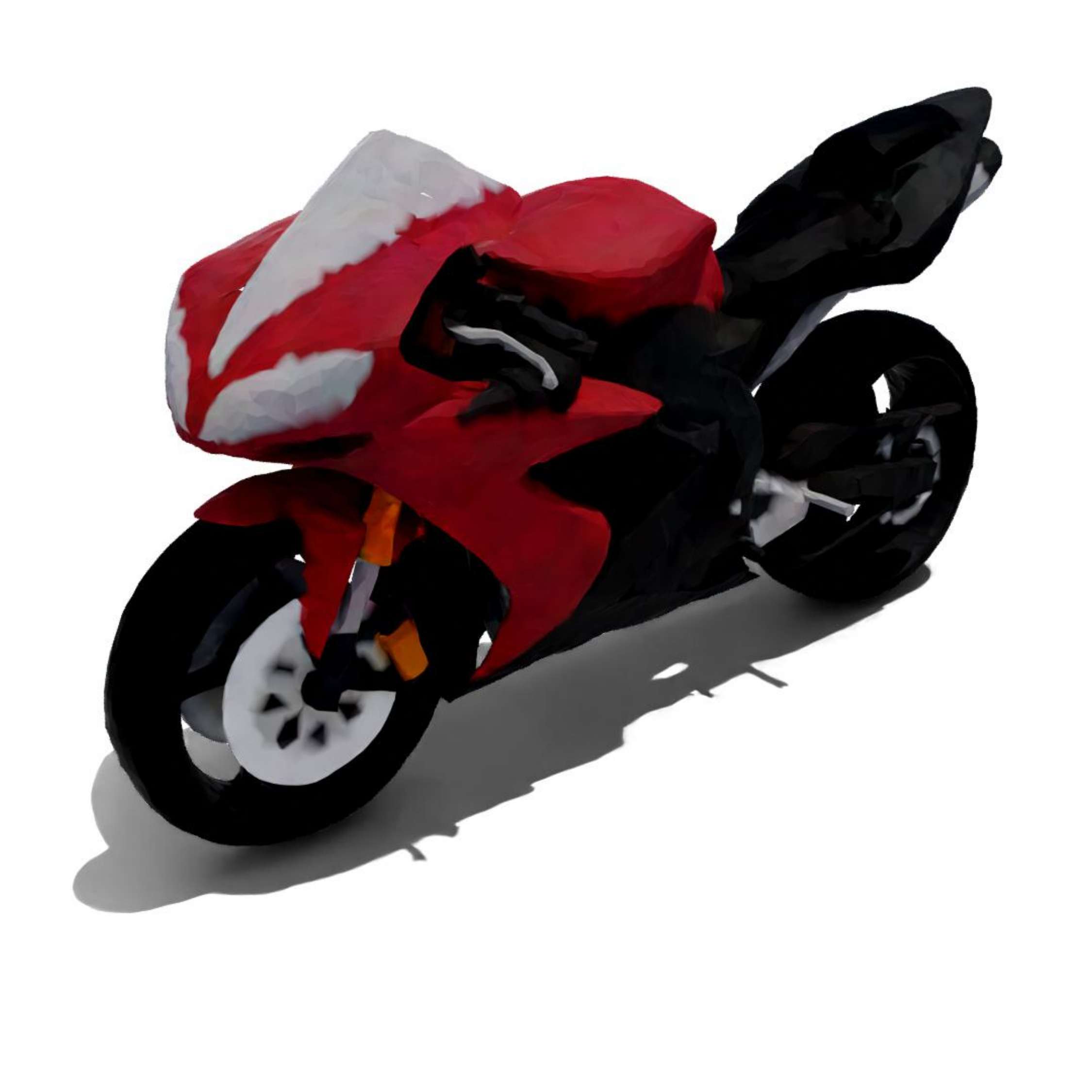}\includegraphics[width=0.16666666666666666\linewidth]{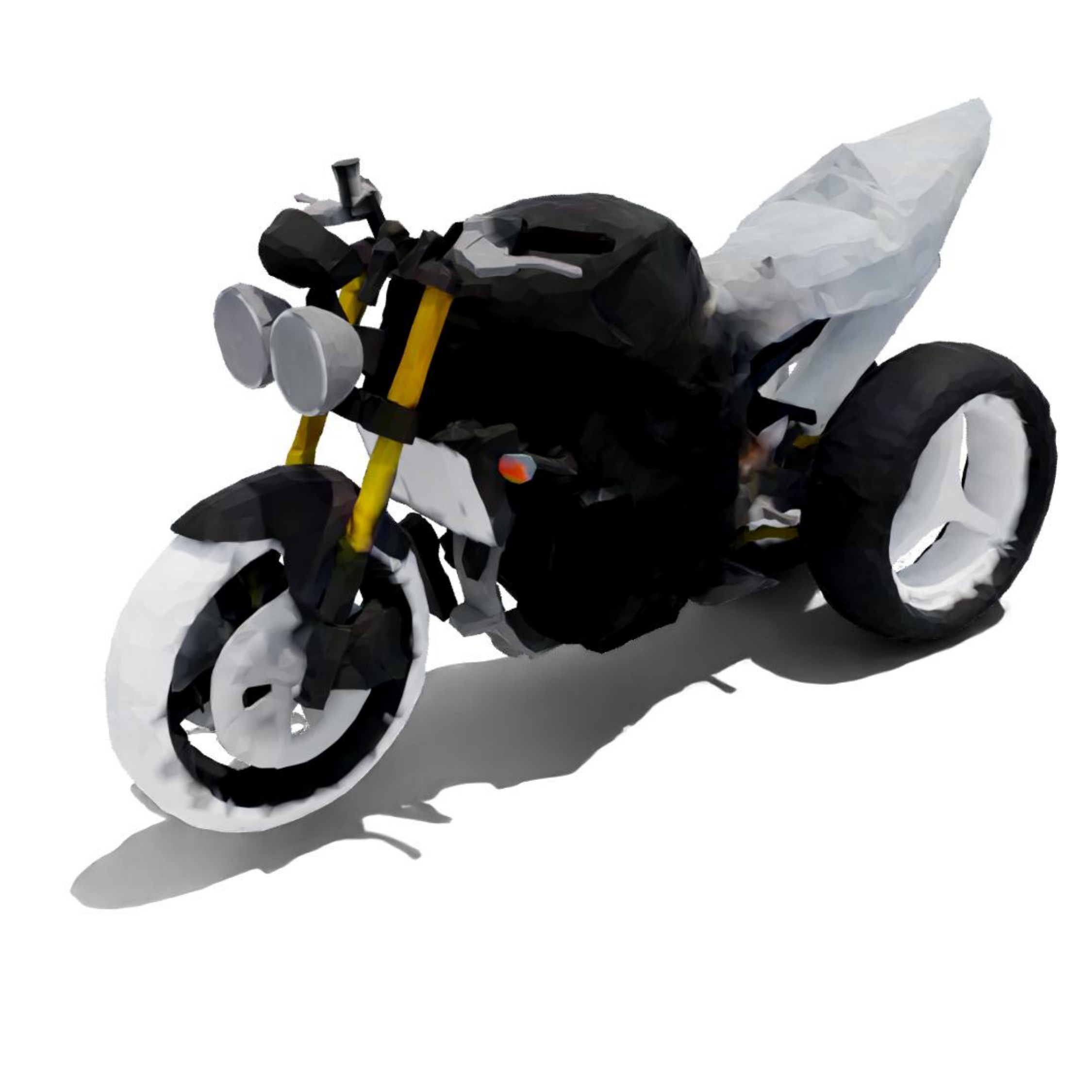}\\

\vspace{-0.4cm}
\includegraphics[width=0.16666666666666666\linewidth]{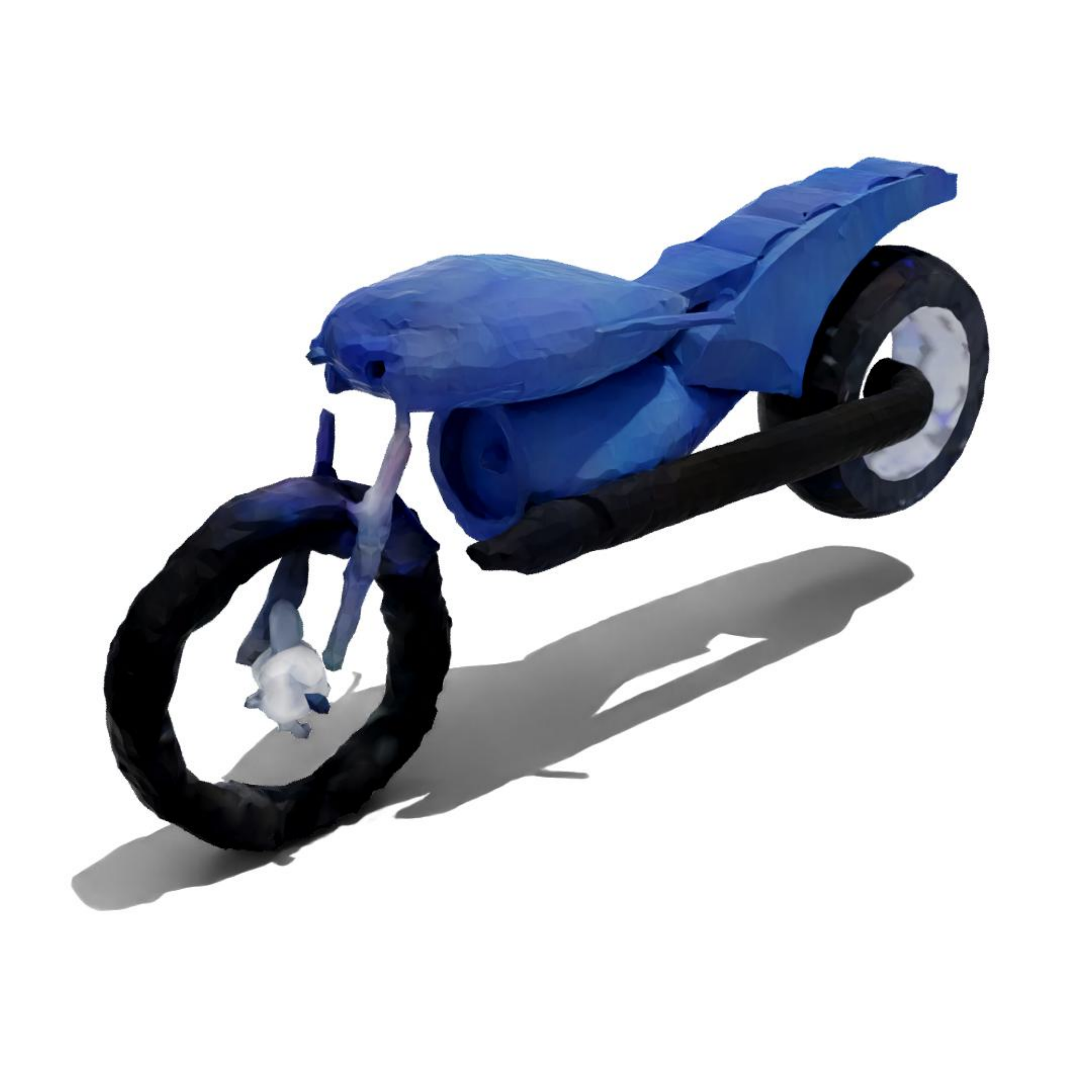}\includegraphics[width=0.16666666666666666\linewidth]{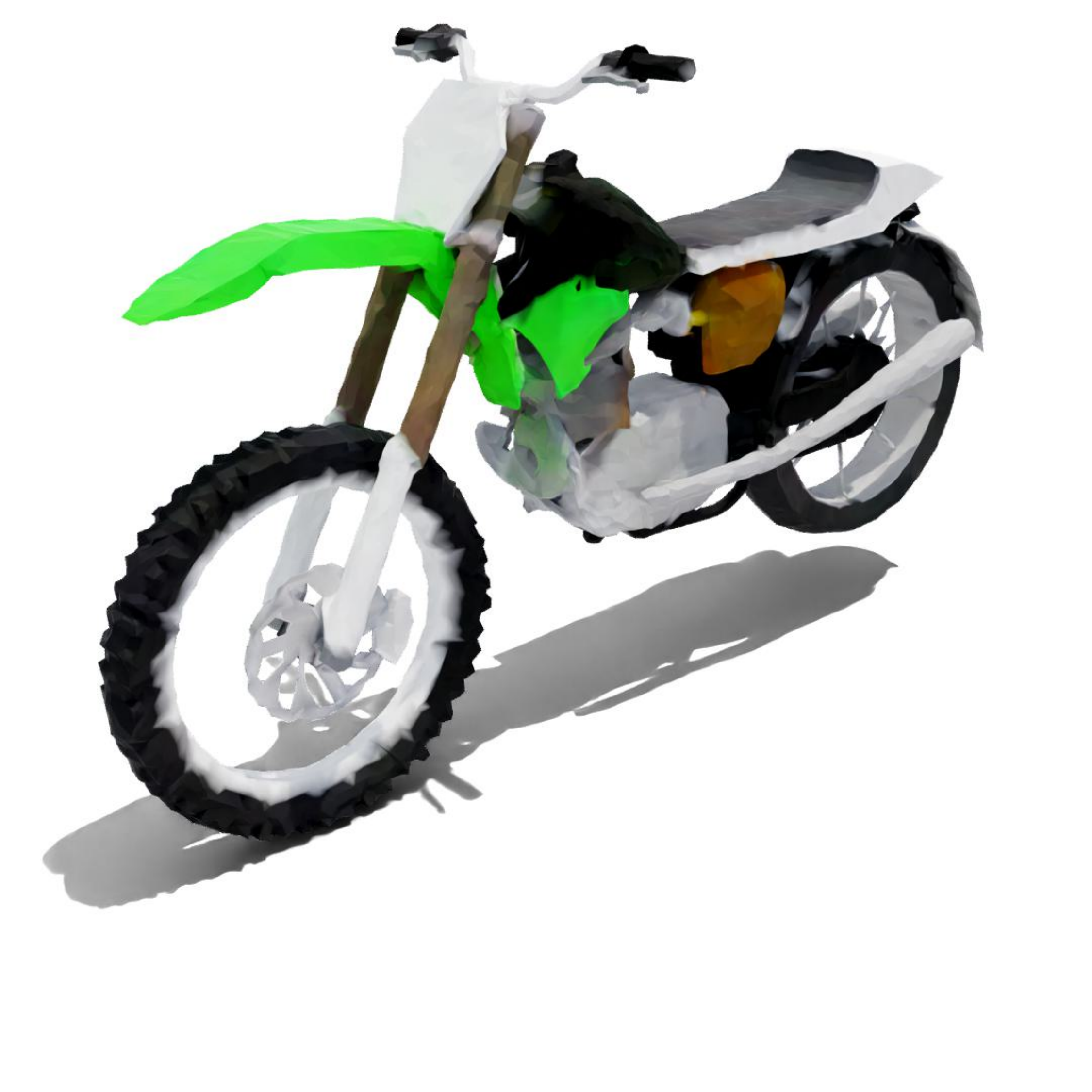}\includegraphics[width=0.16666666666666666\linewidth]{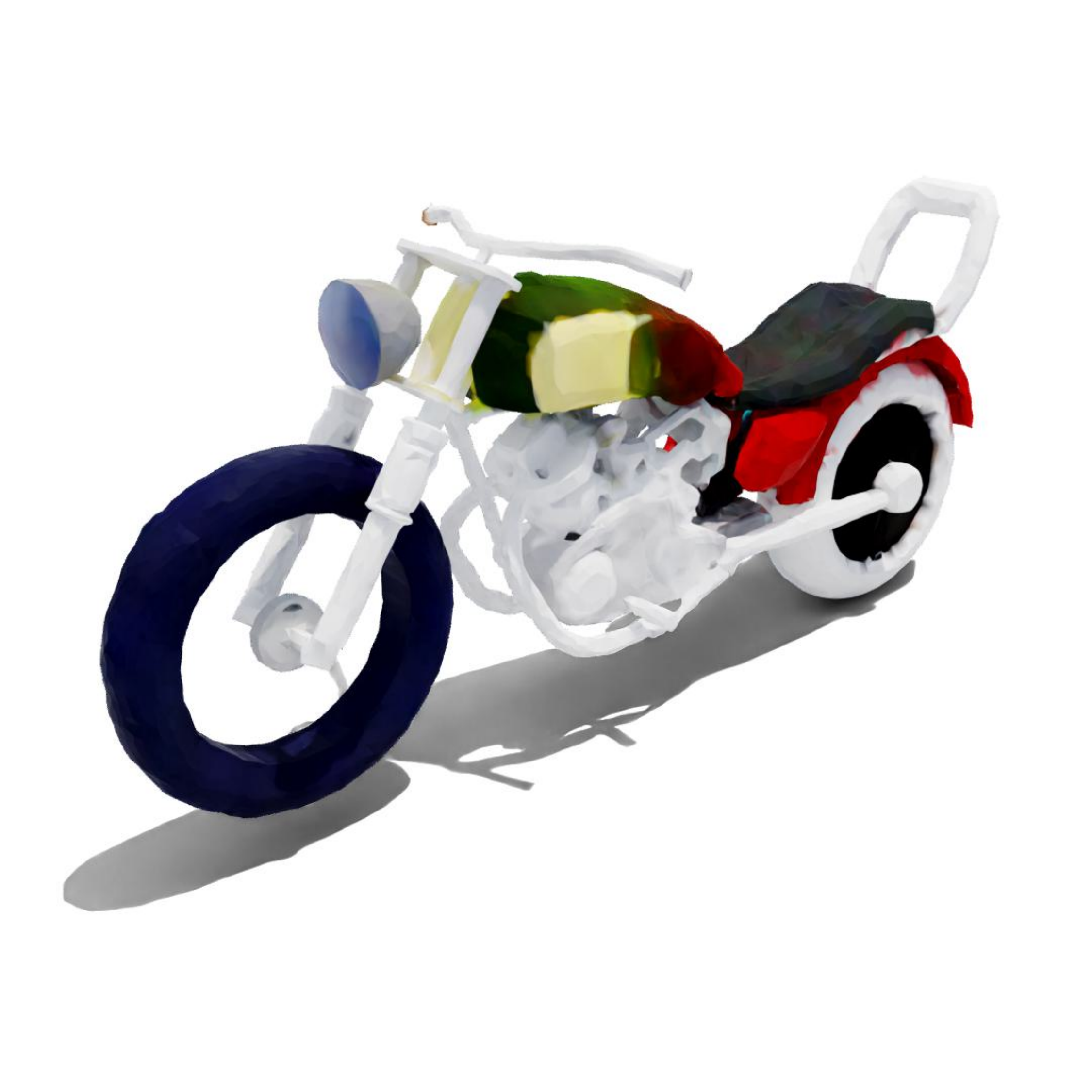}\includegraphics[width=0.16666666666666666\linewidth]{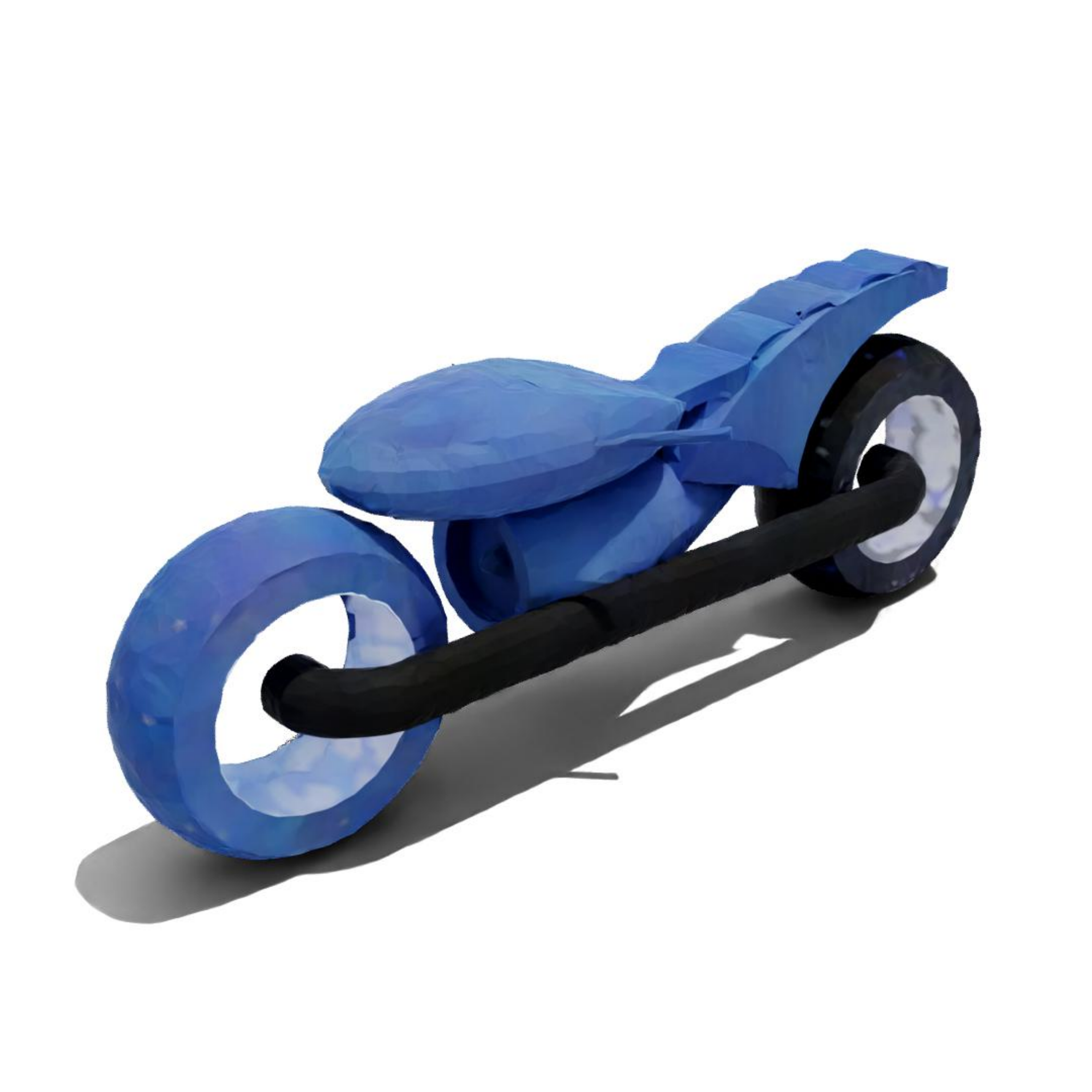}\includegraphics[width=0.16666666666666666\linewidth]{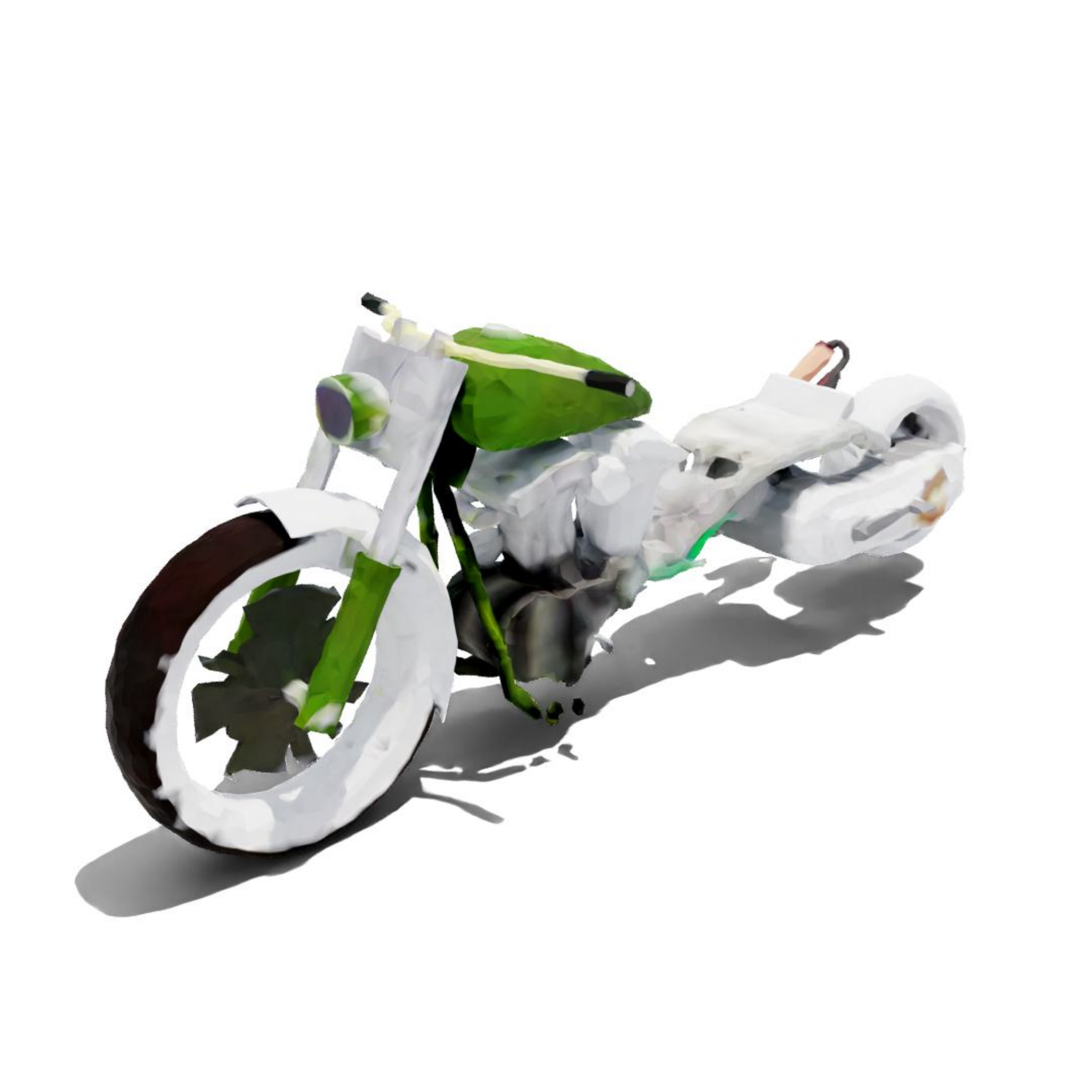}\includegraphics[width=0.16666666666666666\linewidth]{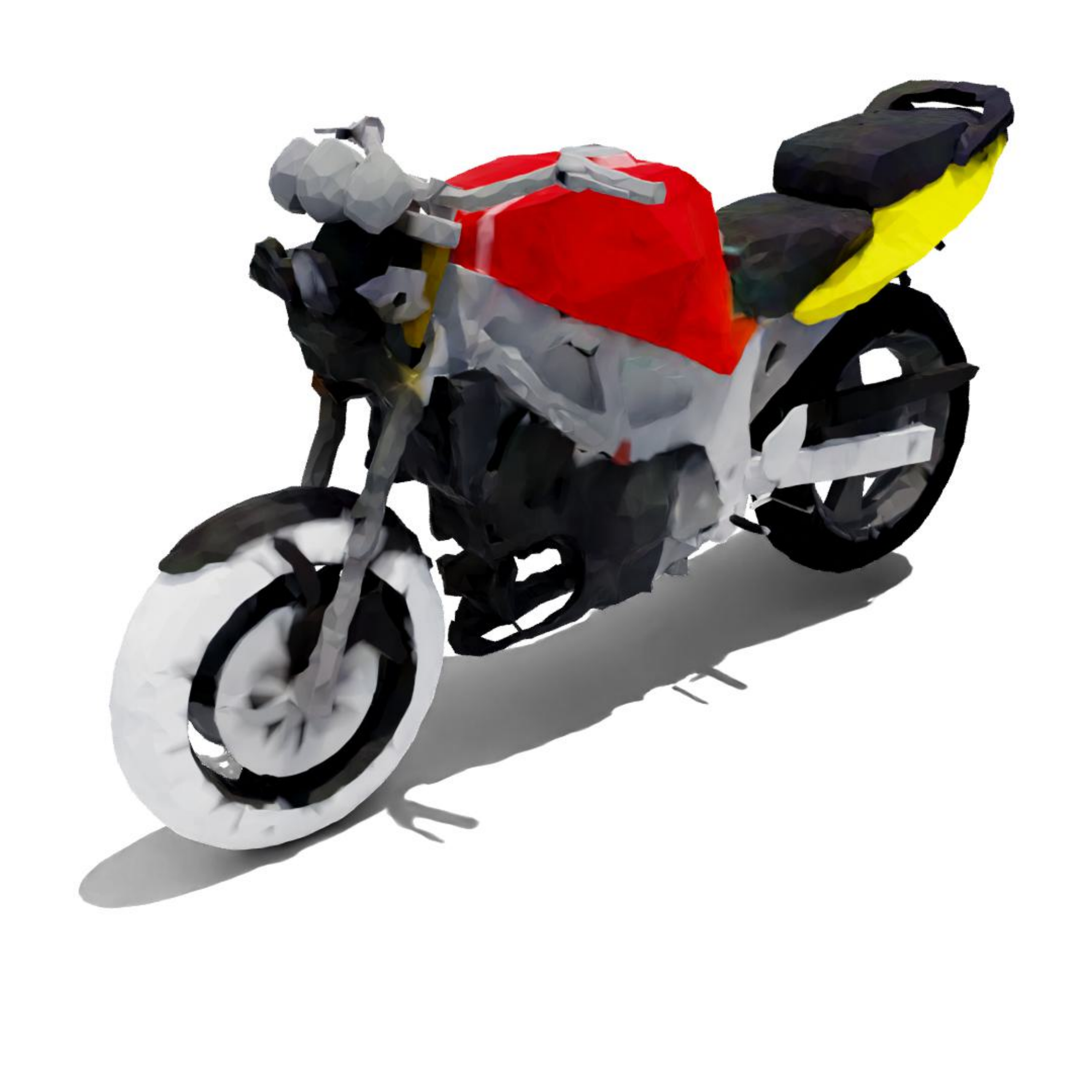}\\

\vspace{-0.4cm}
\includegraphics[width=0.16666666666666666\linewidth]{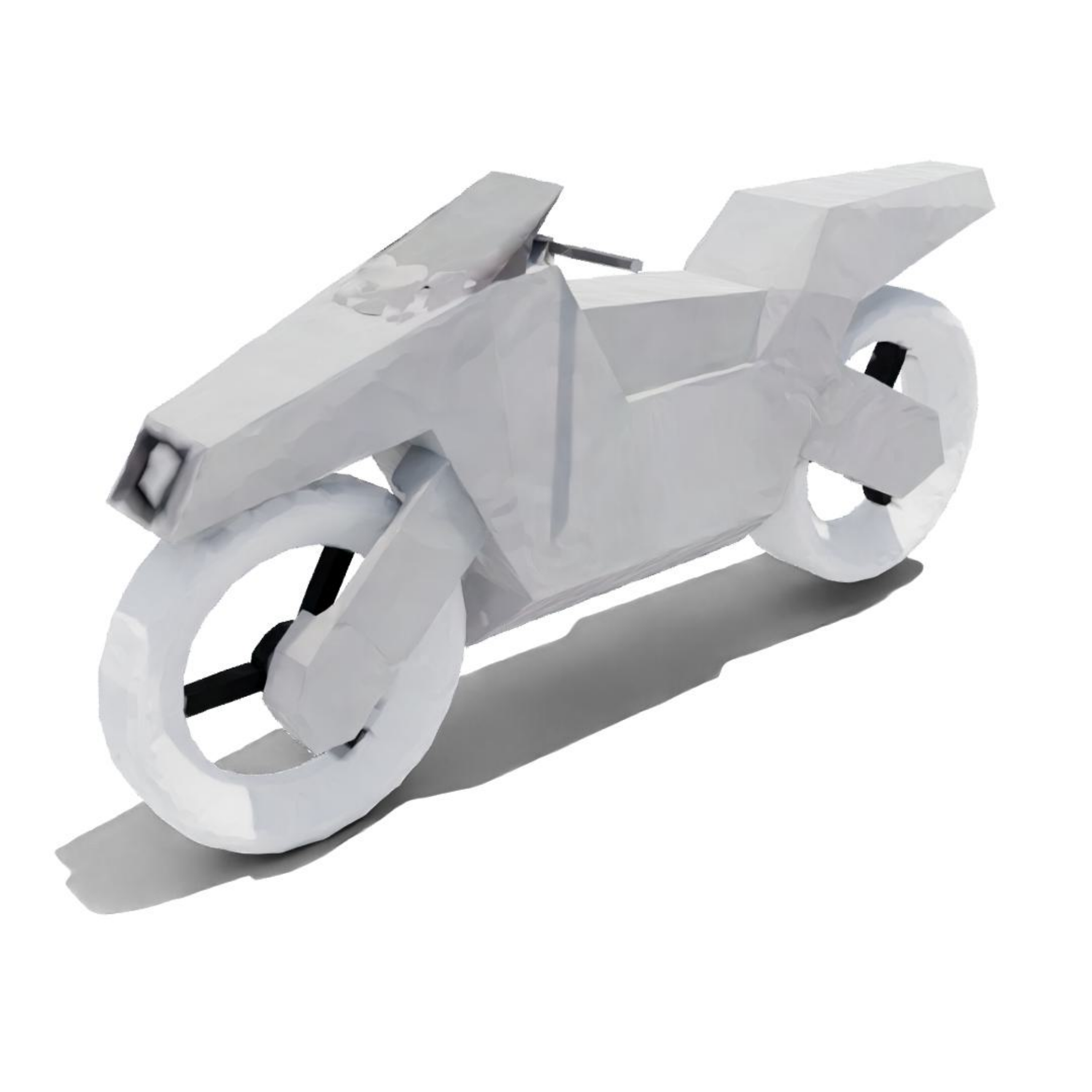}\includegraphics[width=0.16666666666666666\linewidth]{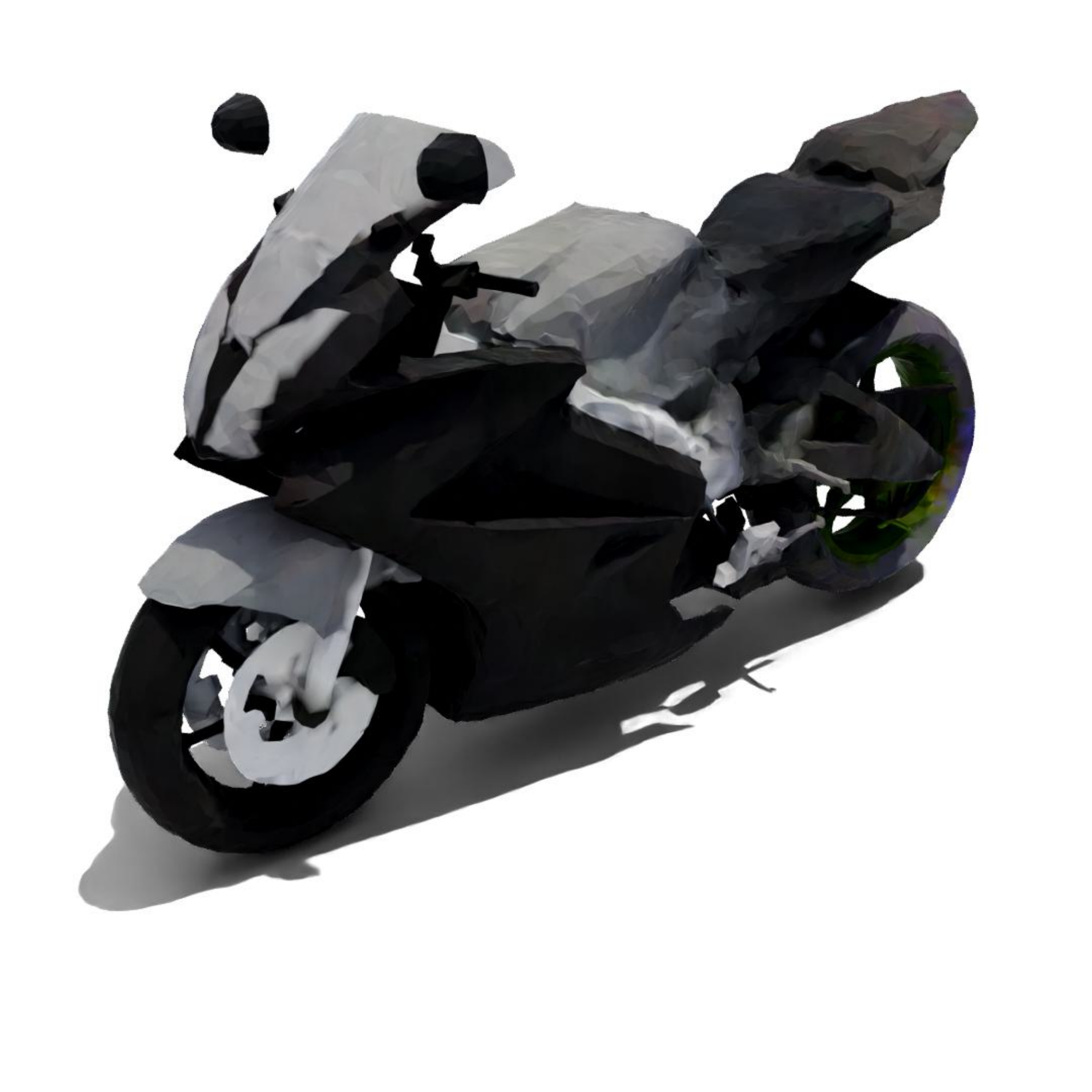}\includegraphics[width=0.16666666666666666\linewidth]{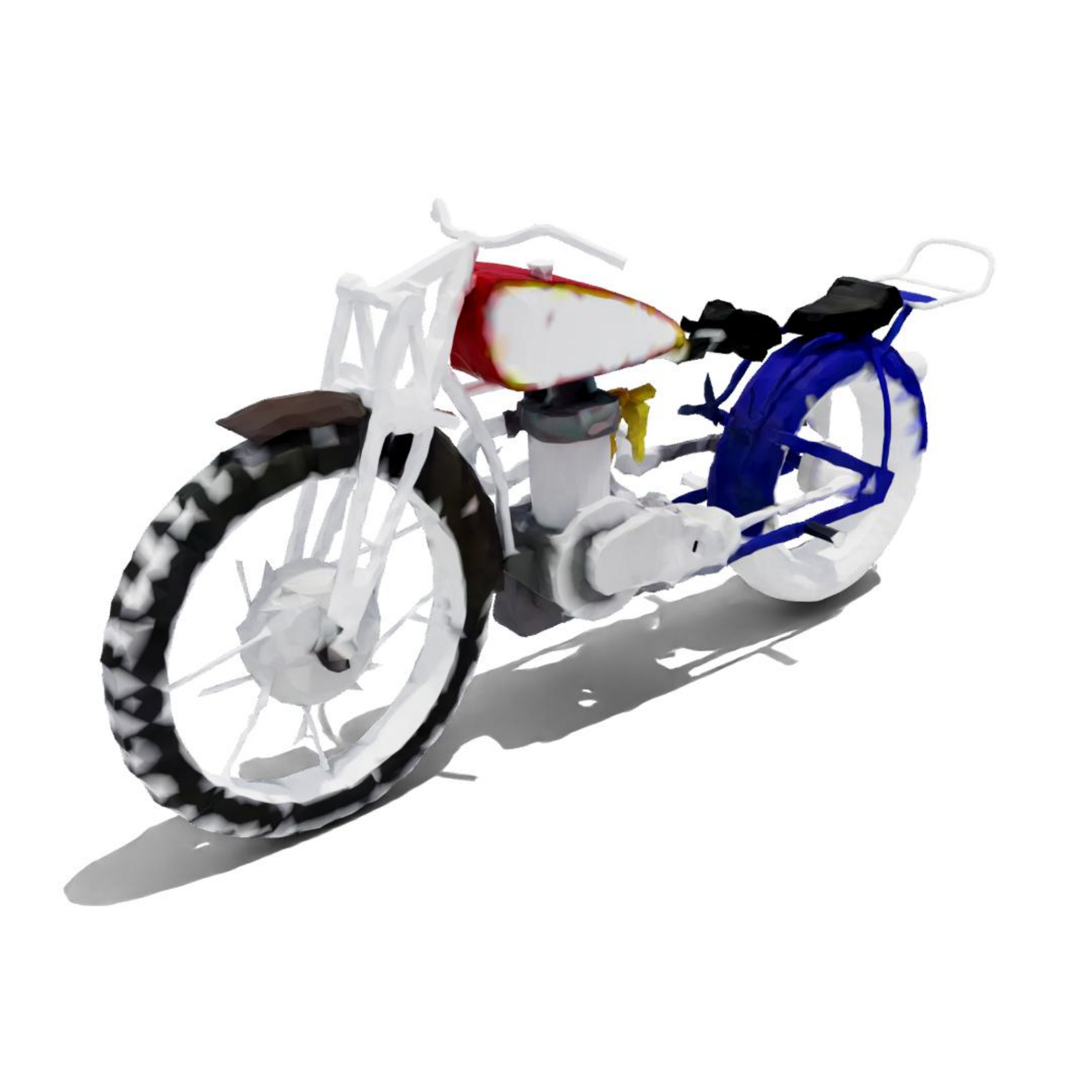}\includegraphics[width=0.16666666666666666\linewidth]{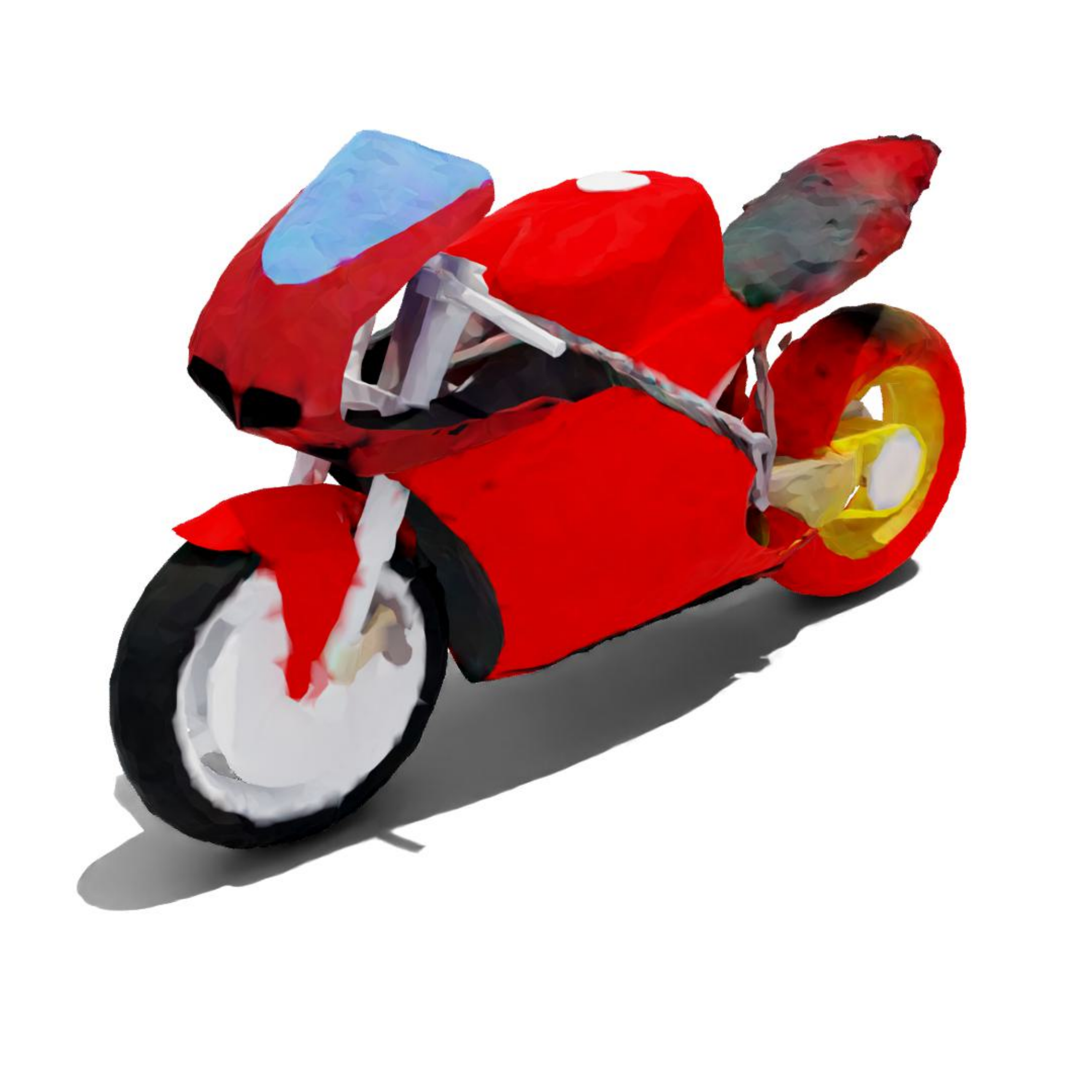}\includegraphics[width=0.16666666666666666\linewidth]{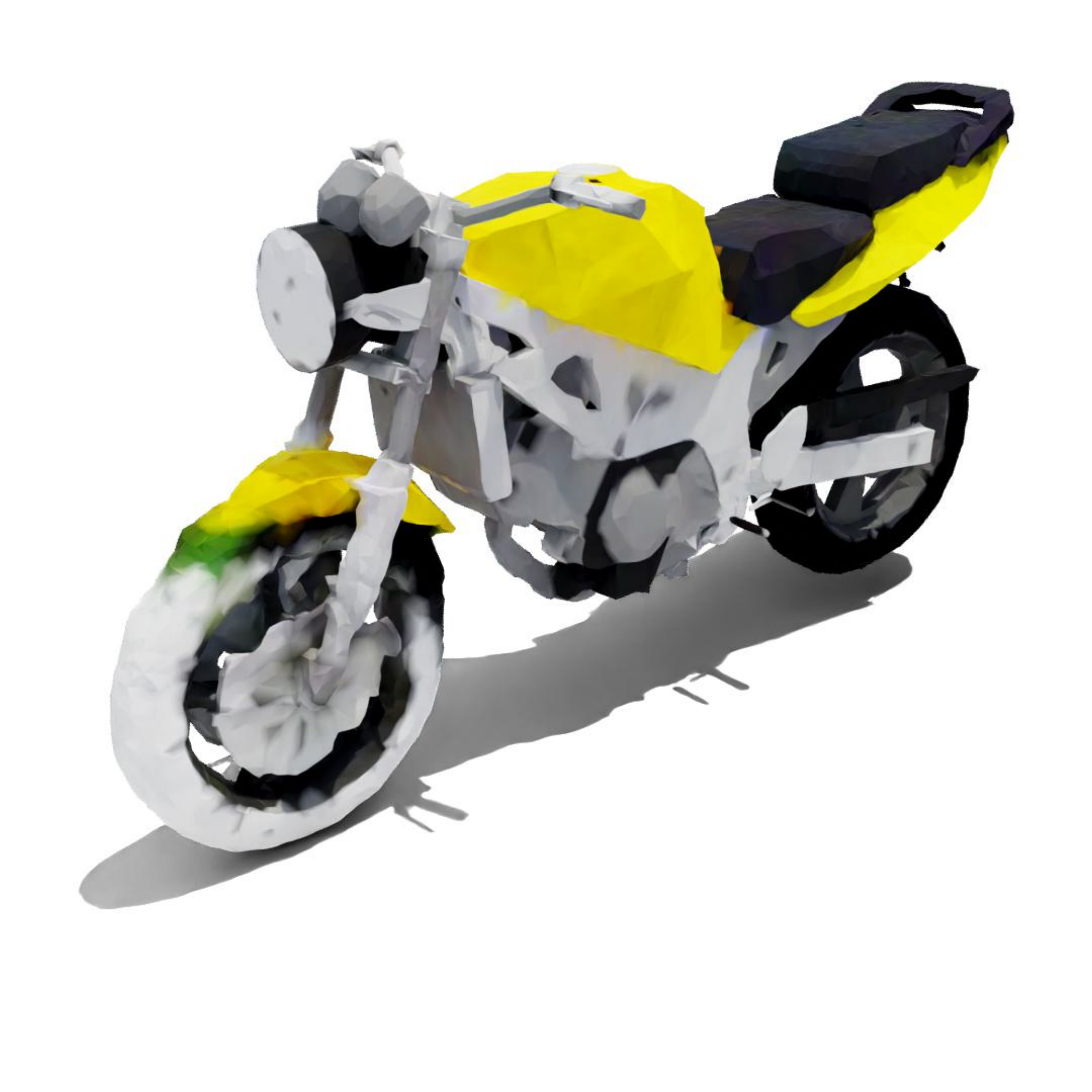}\includegraphics[width=0.16666666666666666\linewidth]{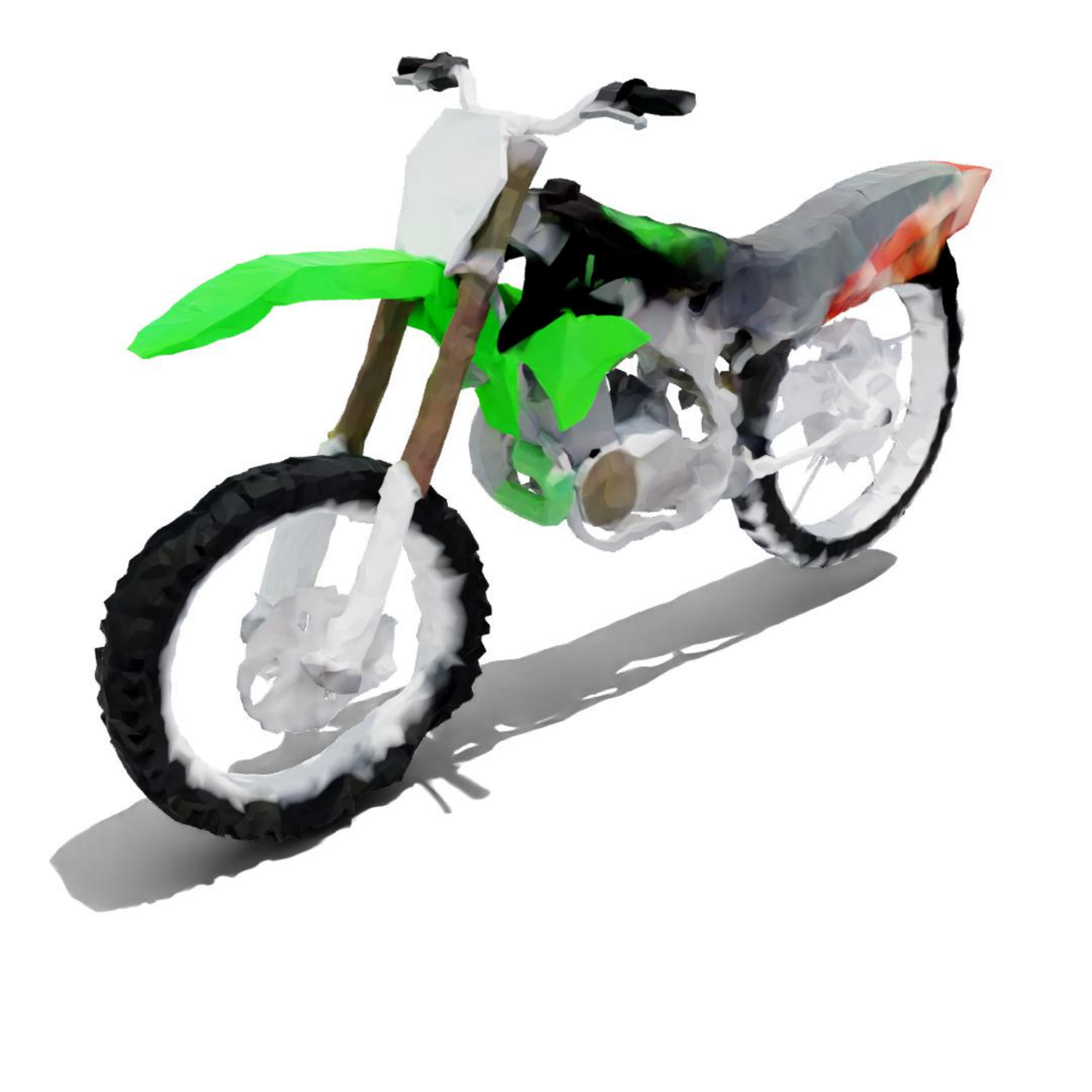}\\

\vspace{-0.4cm}
\includegraphics[width=0.16666666666666666\linewidth]{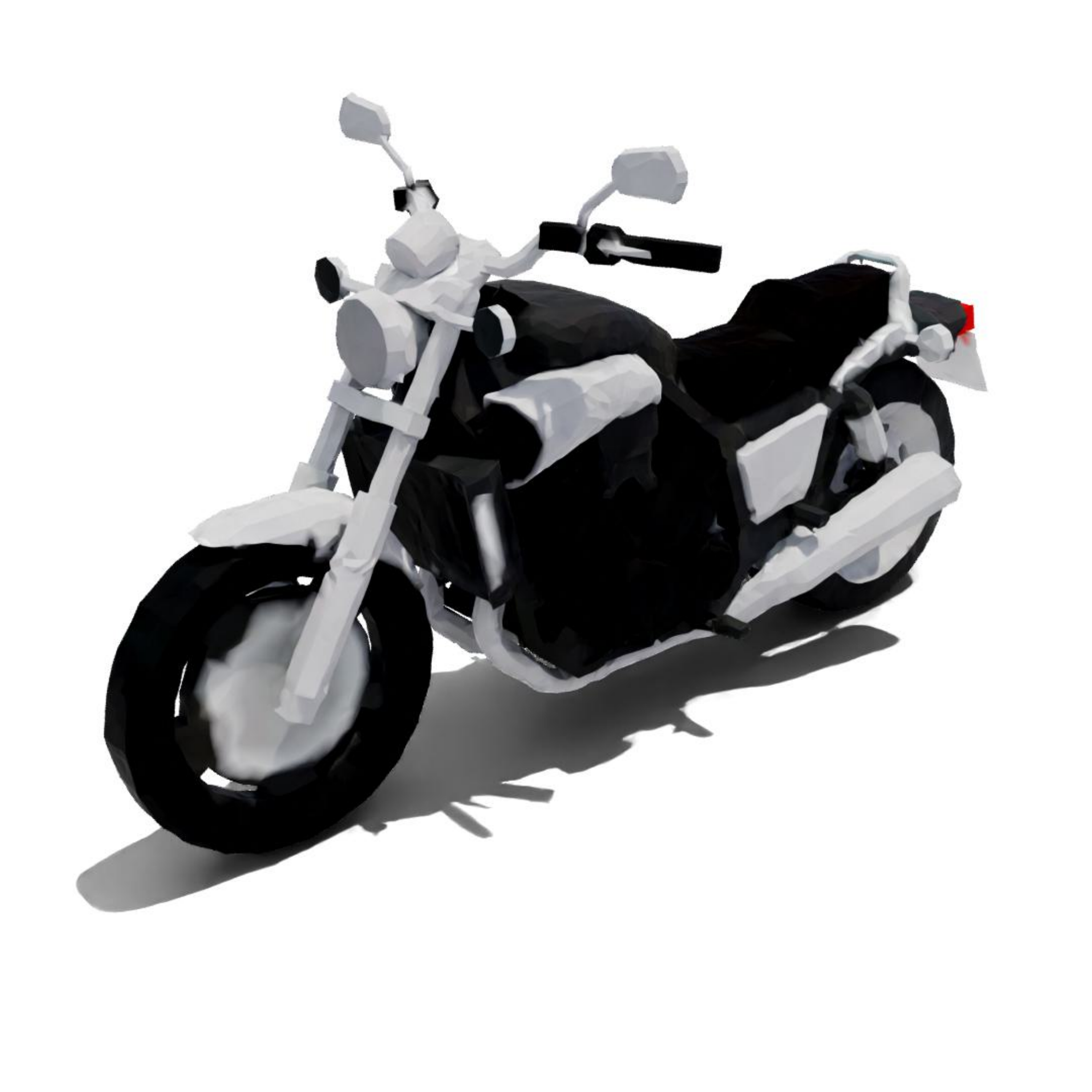}\includegraphics[width=0.16666666666666666\linewidth]{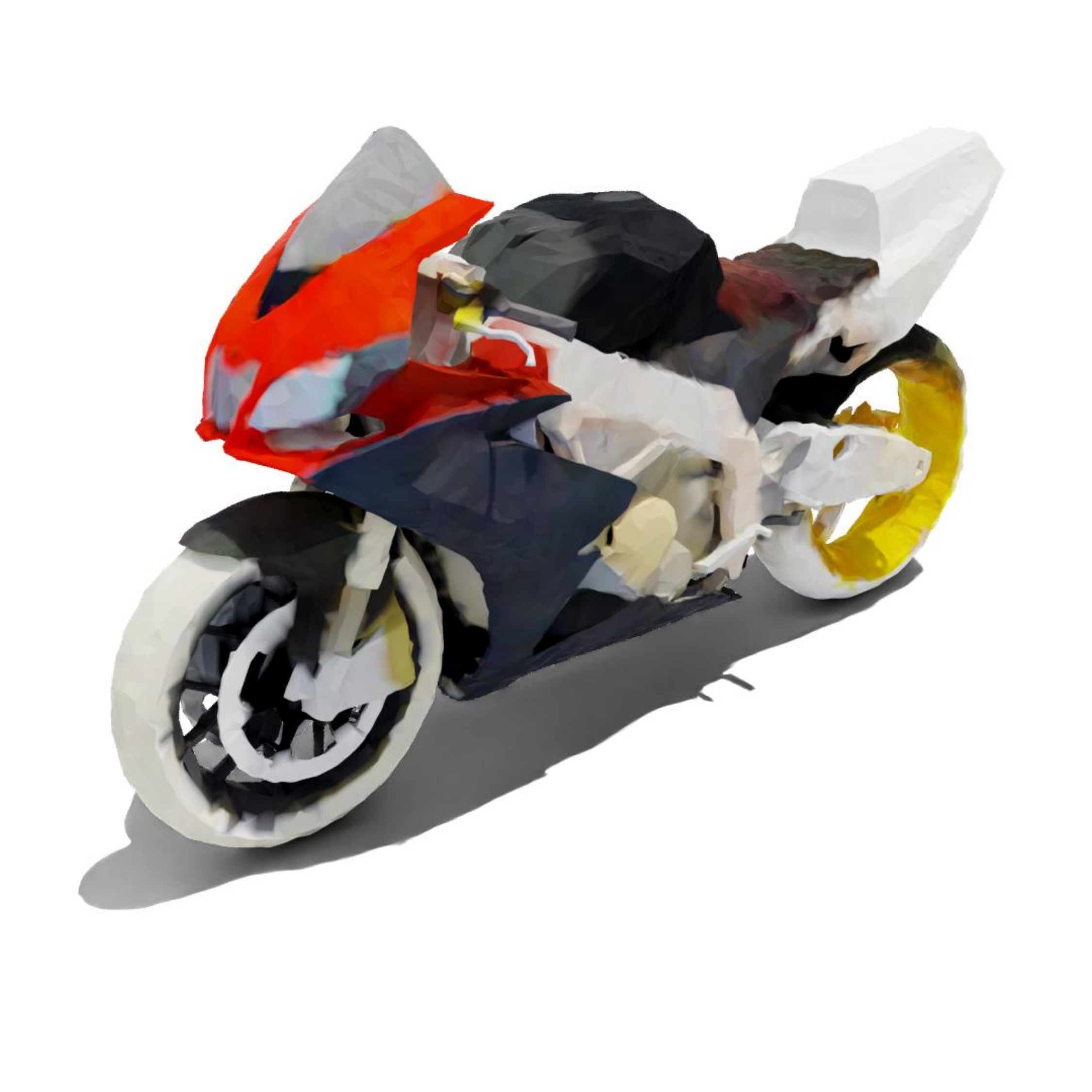}\includegraphics[width=0.16666666666666666\linewidth]{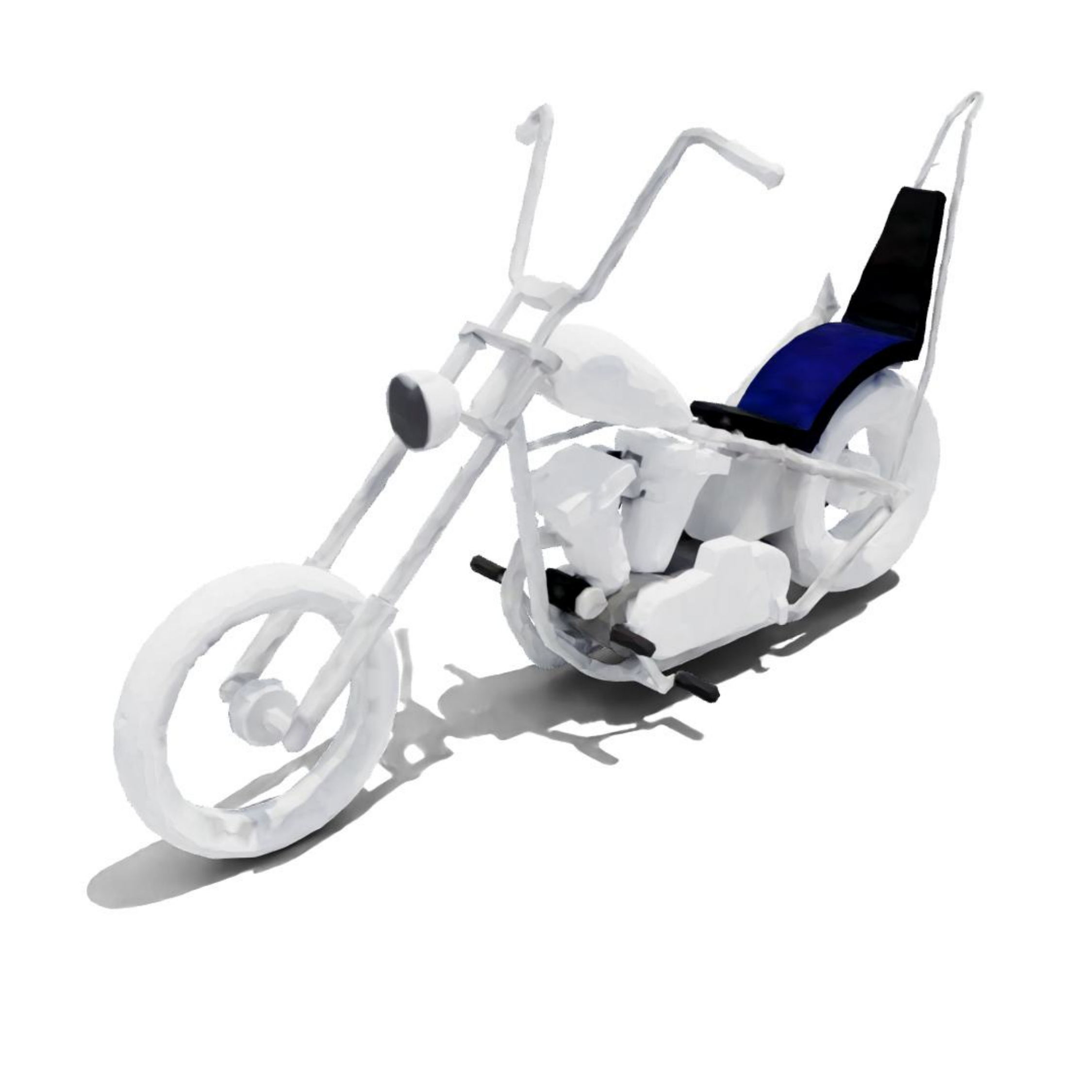}\includegraphics[width=0.16666666666666666\linewidth]{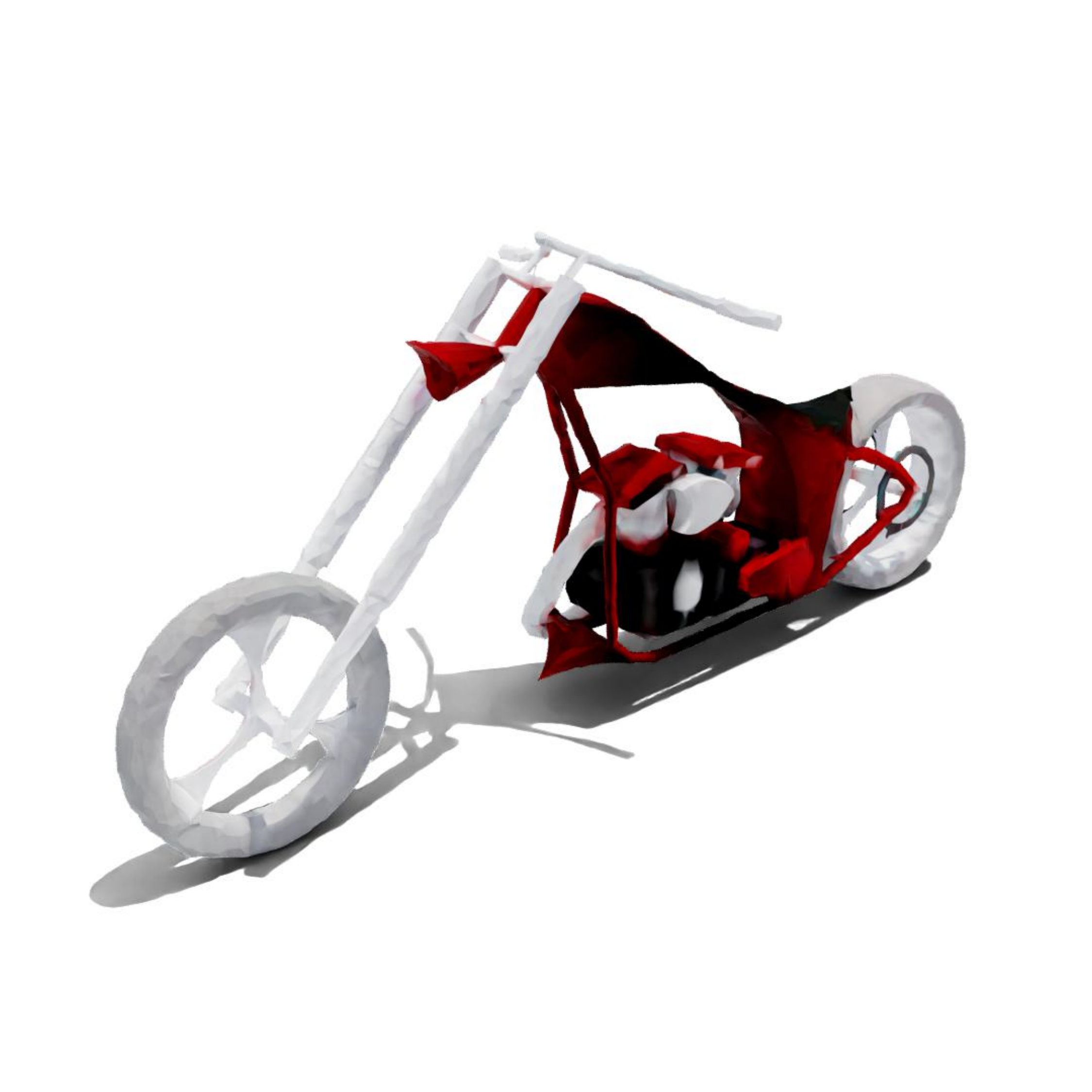}\includegraphics[width=0.16666666666666666\linewidth]{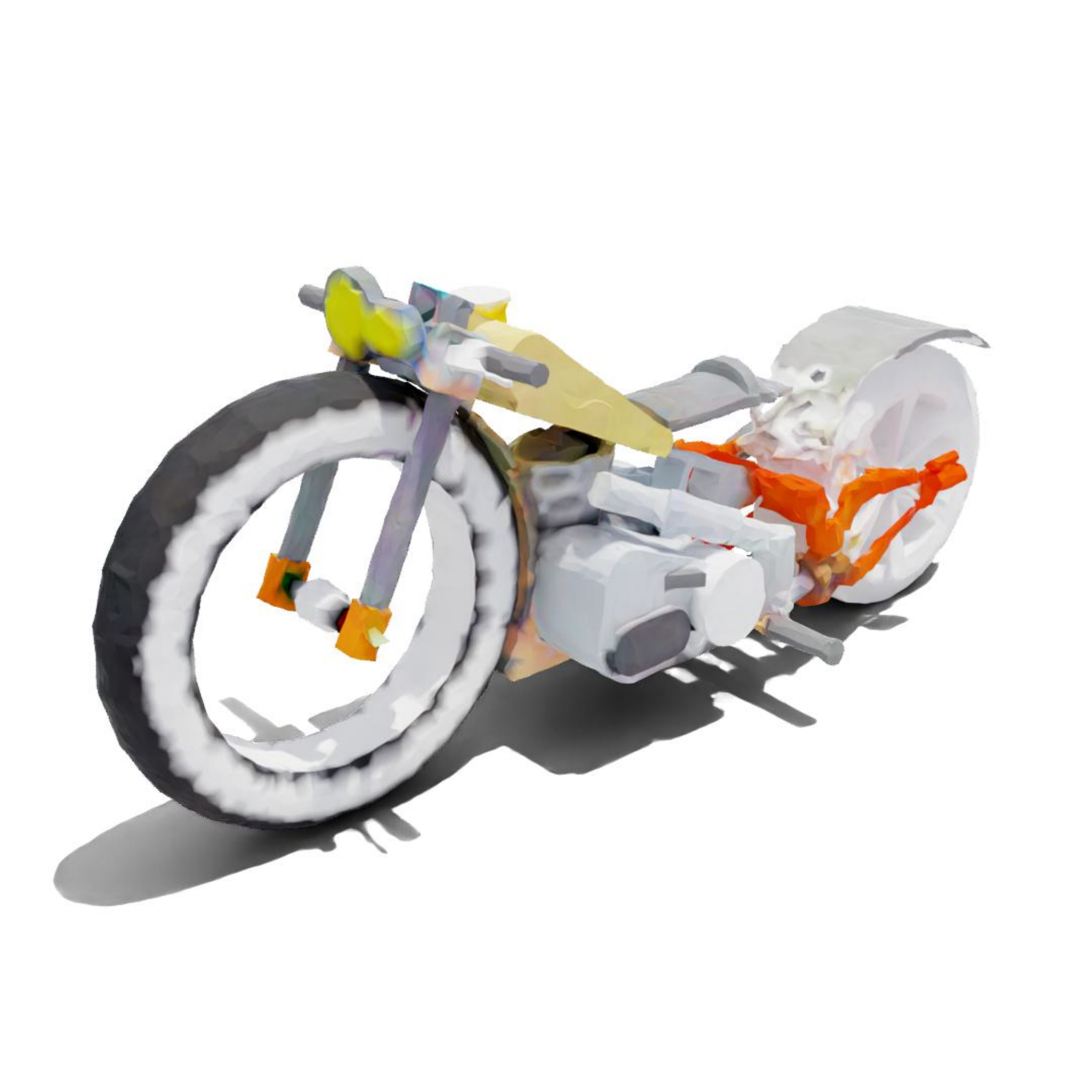}\includegraphics[width=0.16666666666666666\linewidth]{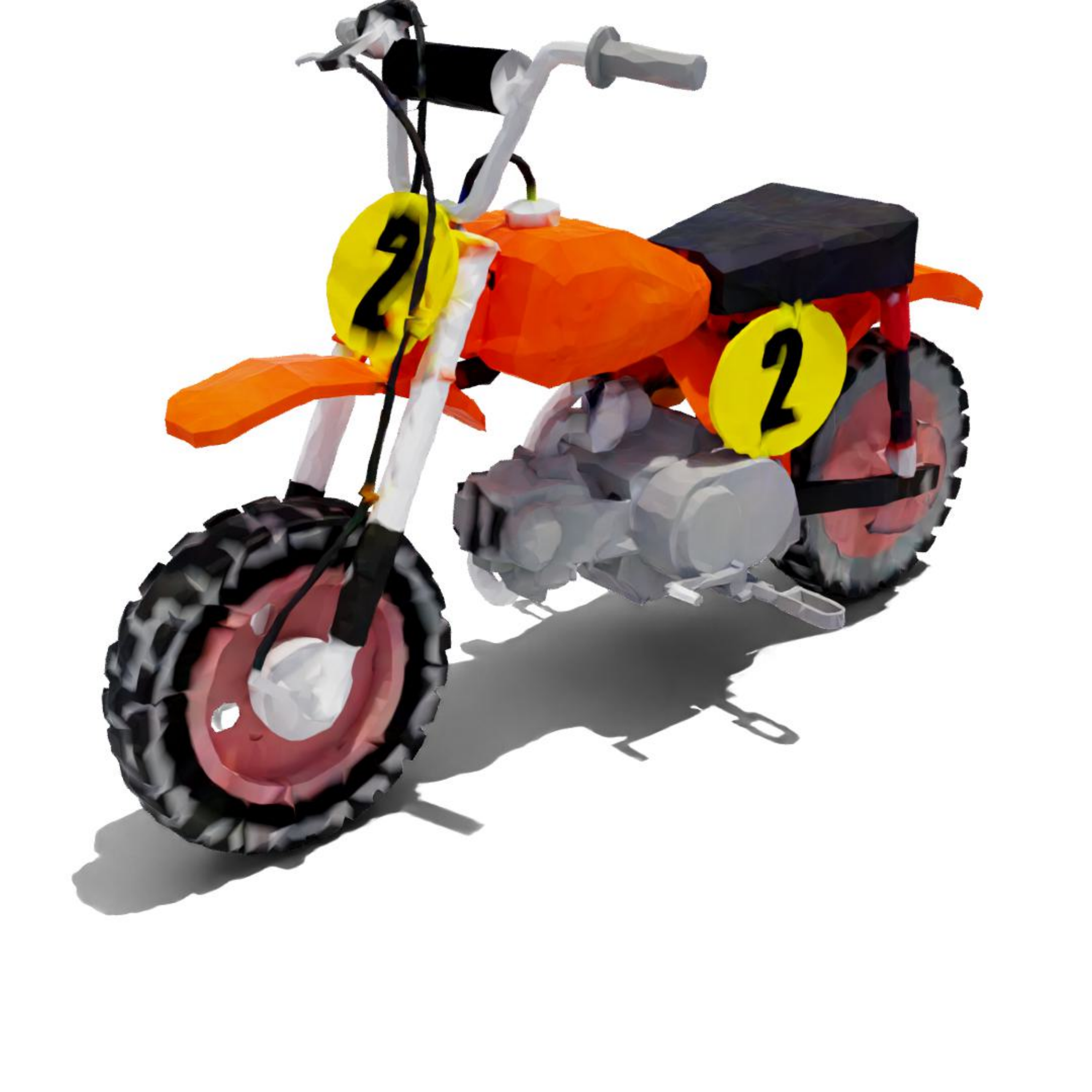}
\caption{\textbf{Random selection of motorbikes generated in standard resolution.}}
\label{fig:uncond:bike:128}
\end{figure*}

\begin{figure*}[!ht]
\centering
\includegraphics[width=0.16666666666666666\linewidth]{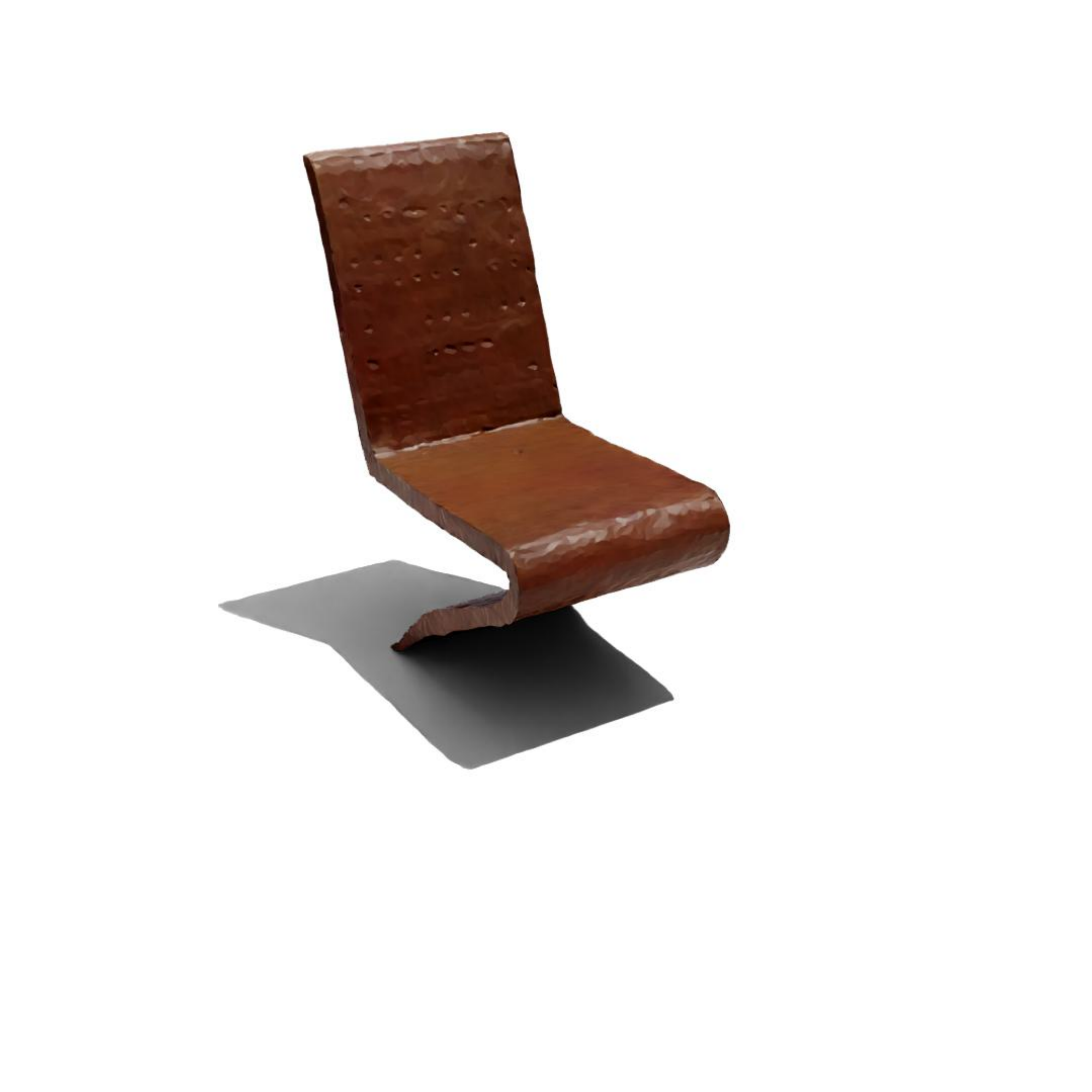}\includegraphics[width=0.16666666666666666\linewidth]{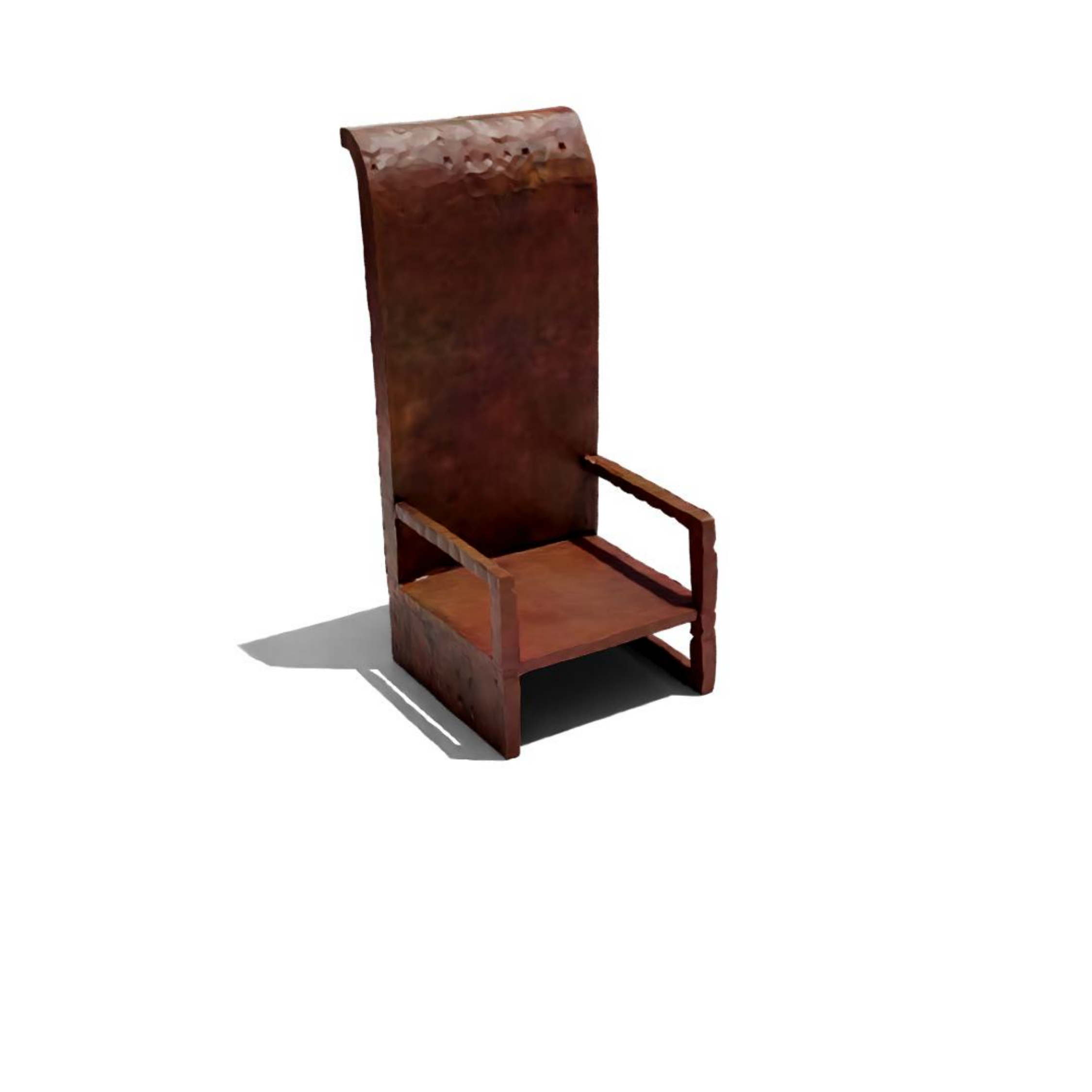}\includegraphics[width=0.16666666666666666\linewidth]{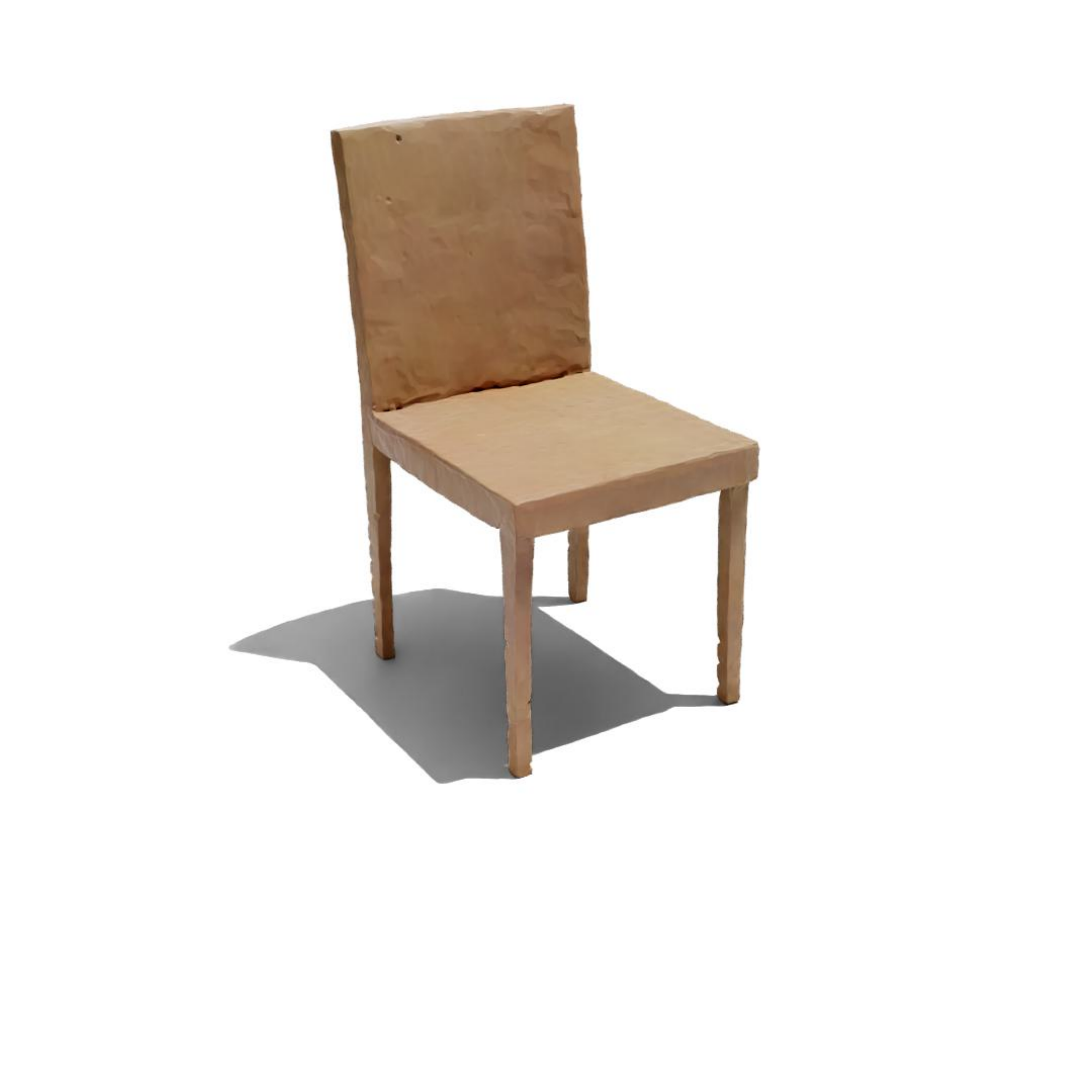}\includegraphics[width=0.16666666666666666\linewidth]{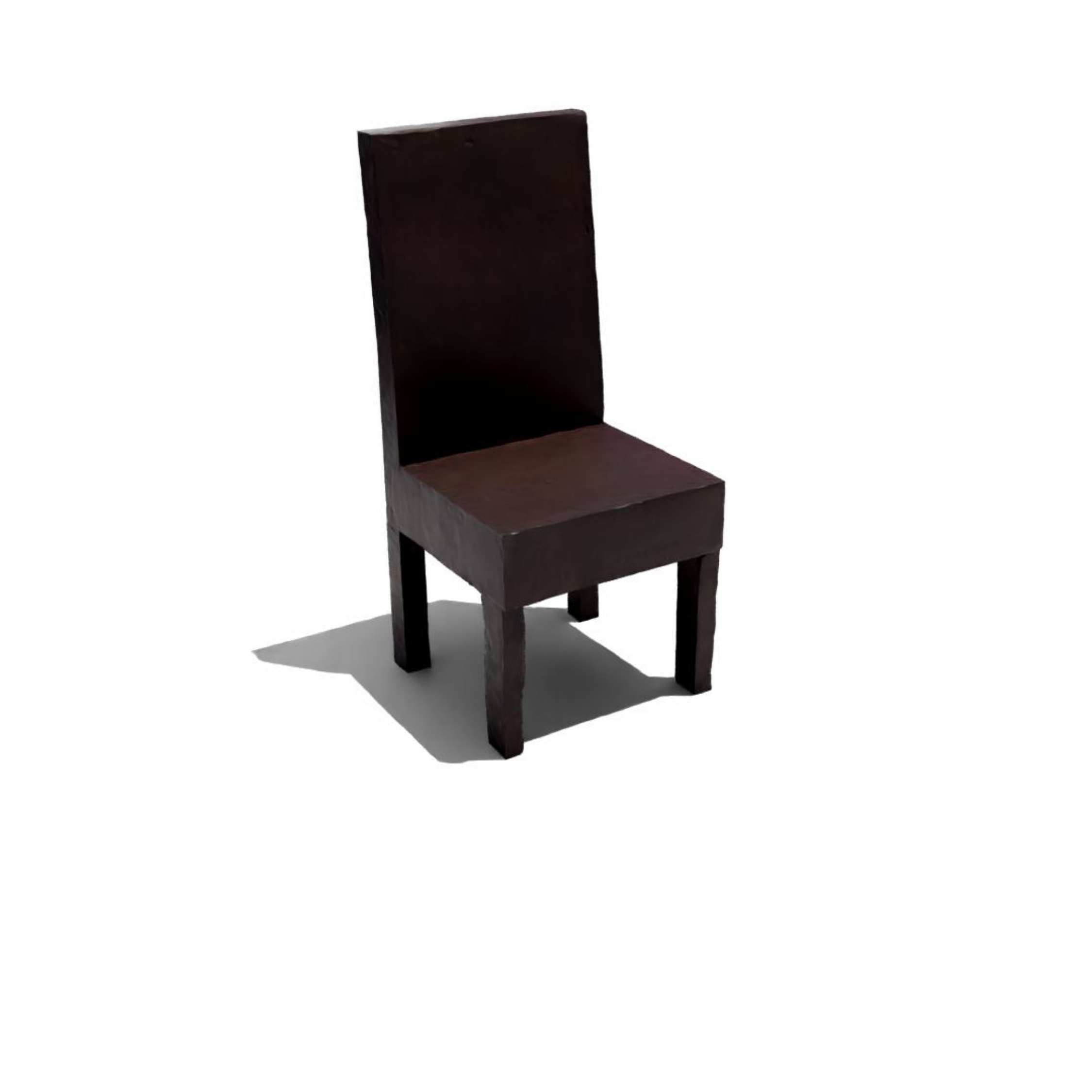}\includegraphics[width=0.16666666666666666\linewidth]{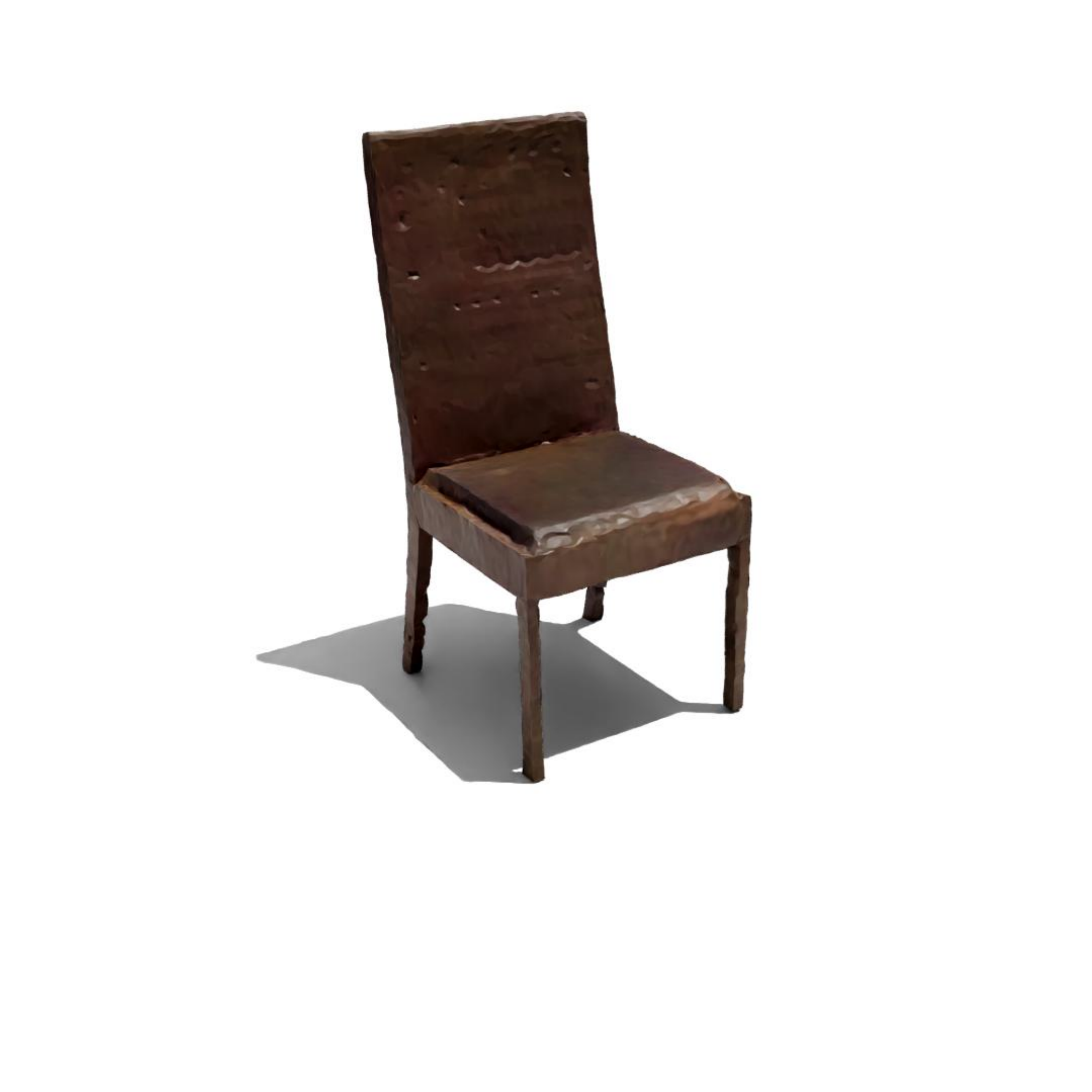}\includegraphics[width=0.16666666666666666\linewidth]{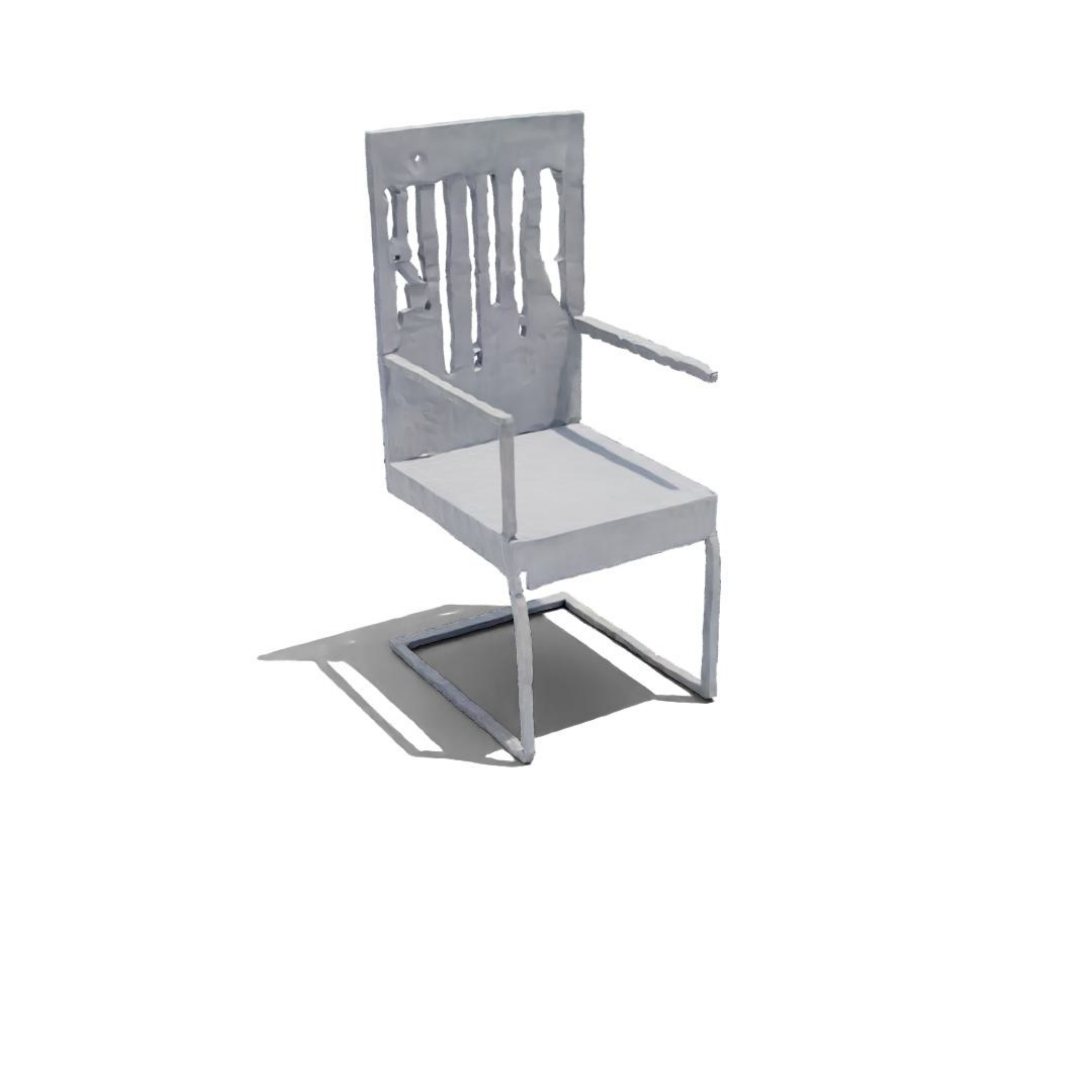}\\

\vspace{-0.3cm}
\includegraphics[width=0.15666666666666666\linewidth]{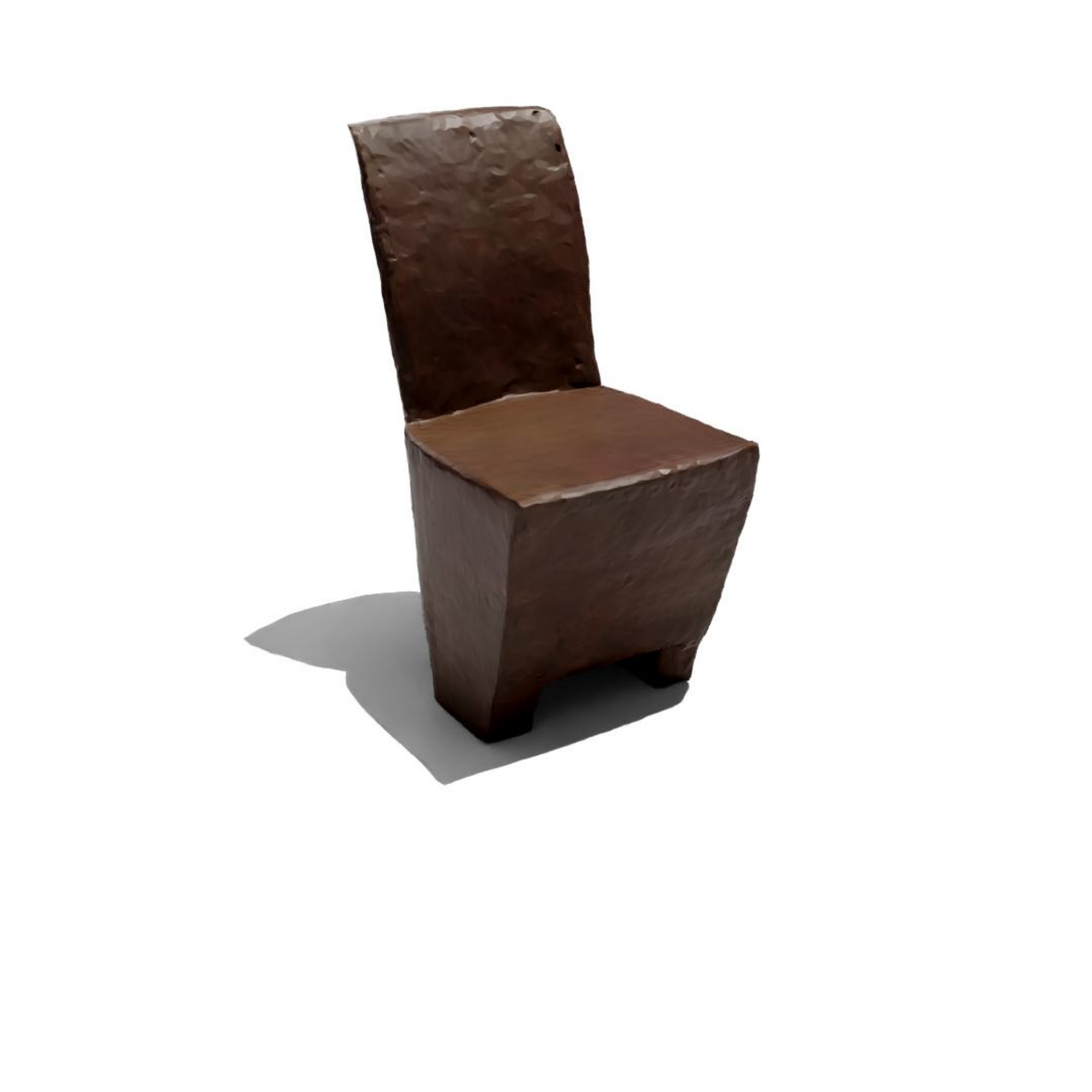}
\includegraphics[width=0.16666666666666666\linewidth]{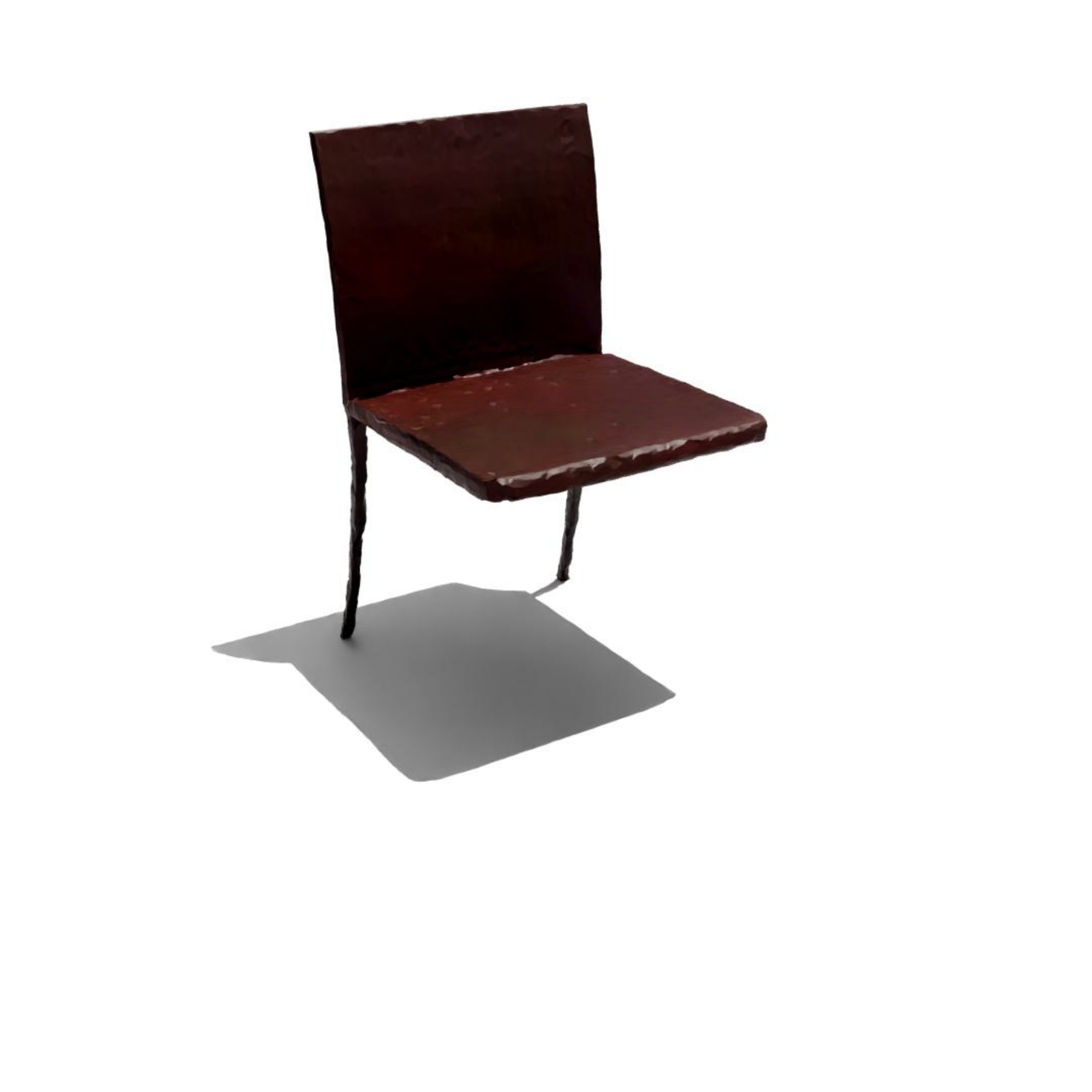}\includegraphics[width=0.16666666666666666\linewidth]{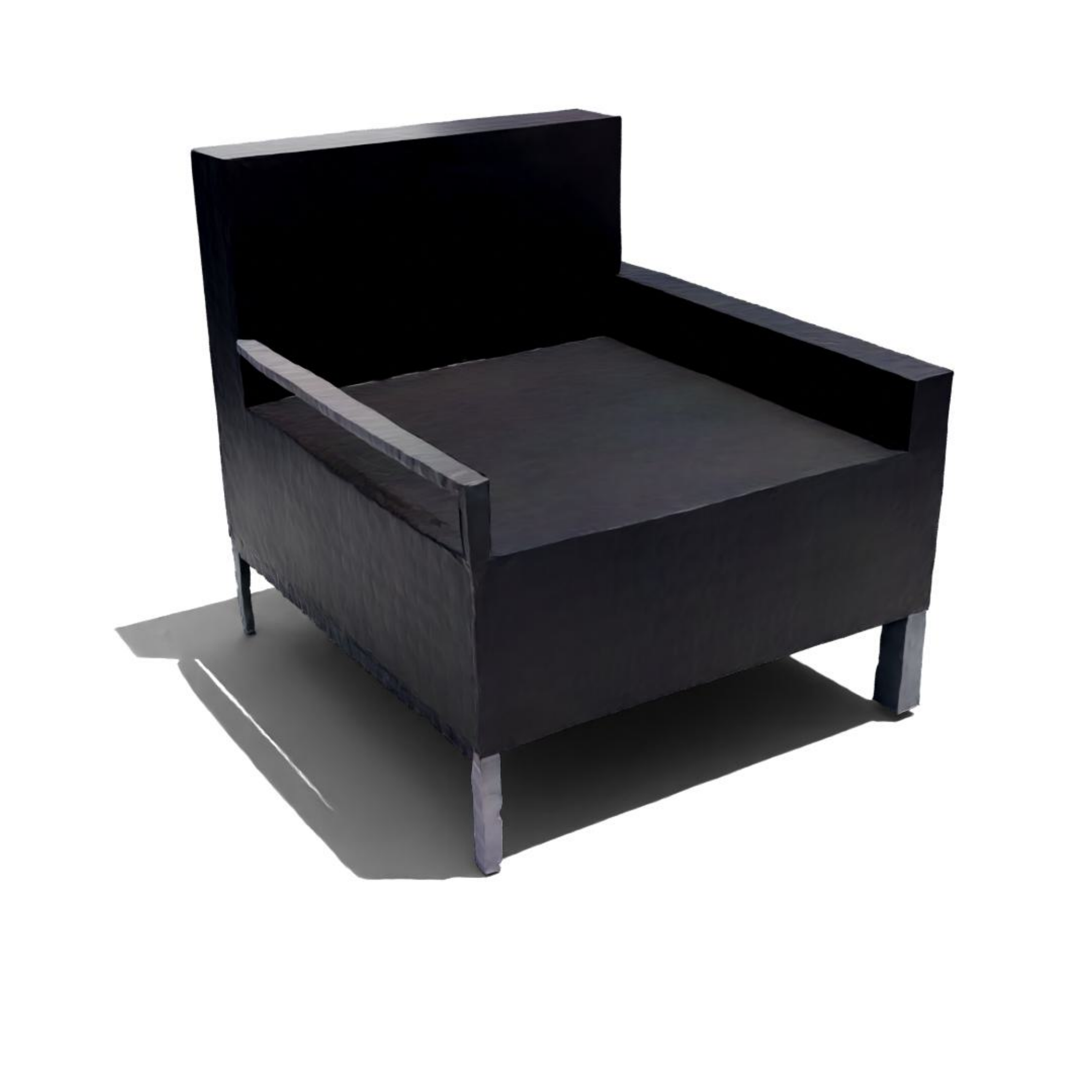}\includegraphics[width=0.16666666666666666\linewidth]{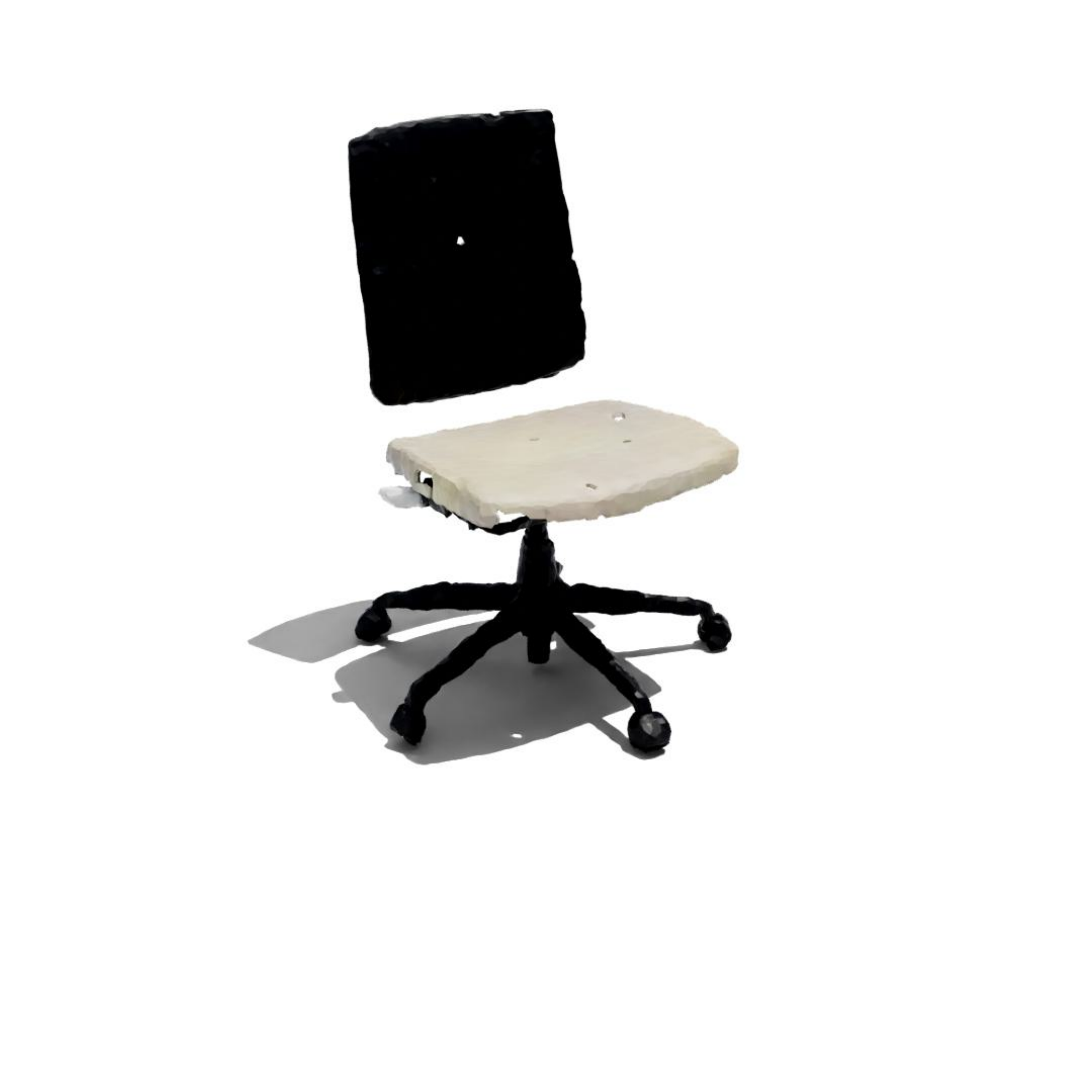}\includegraphics[width=0.16666666666666666\linewidth]{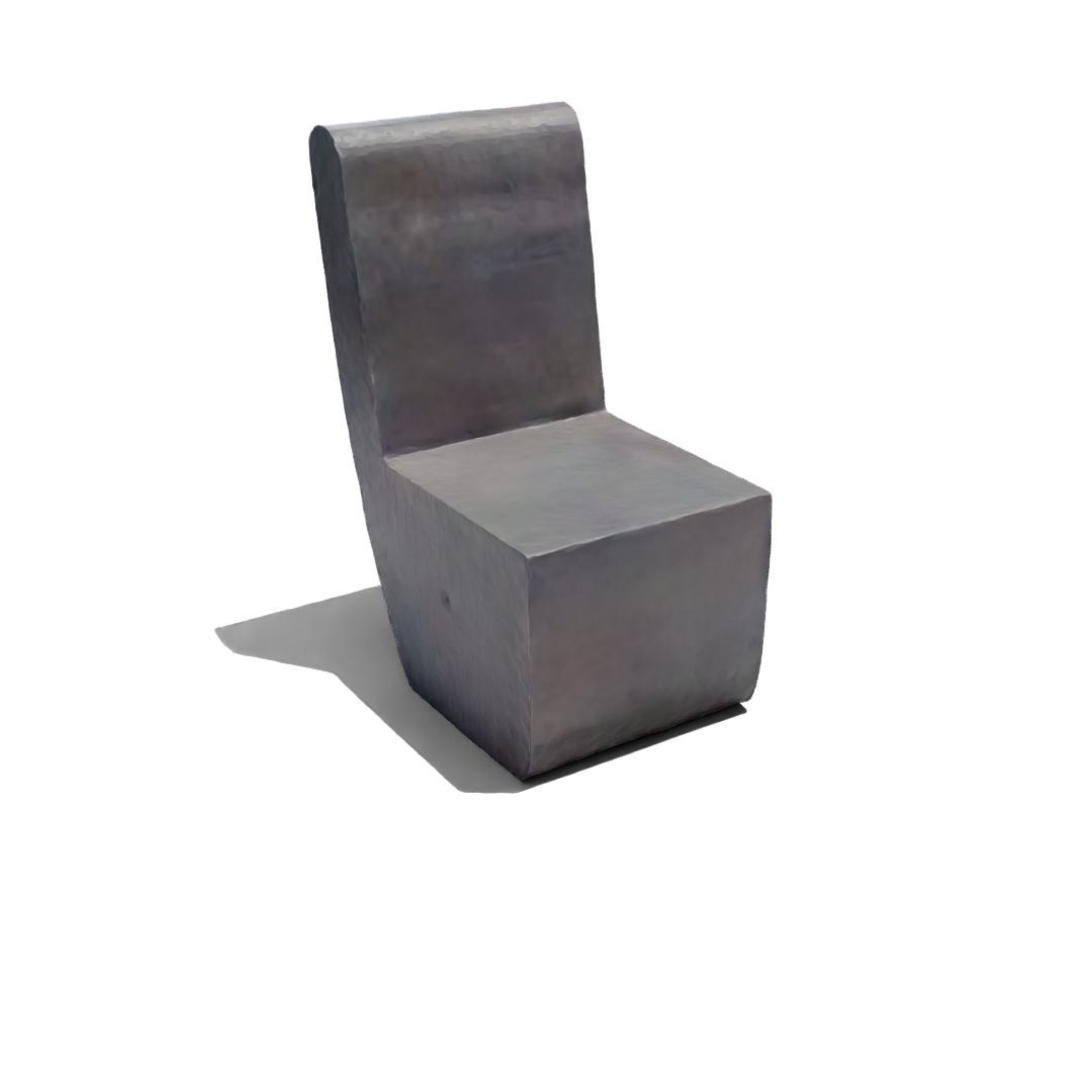}\includegraphics[width=0.16666666666666666\linewidth]{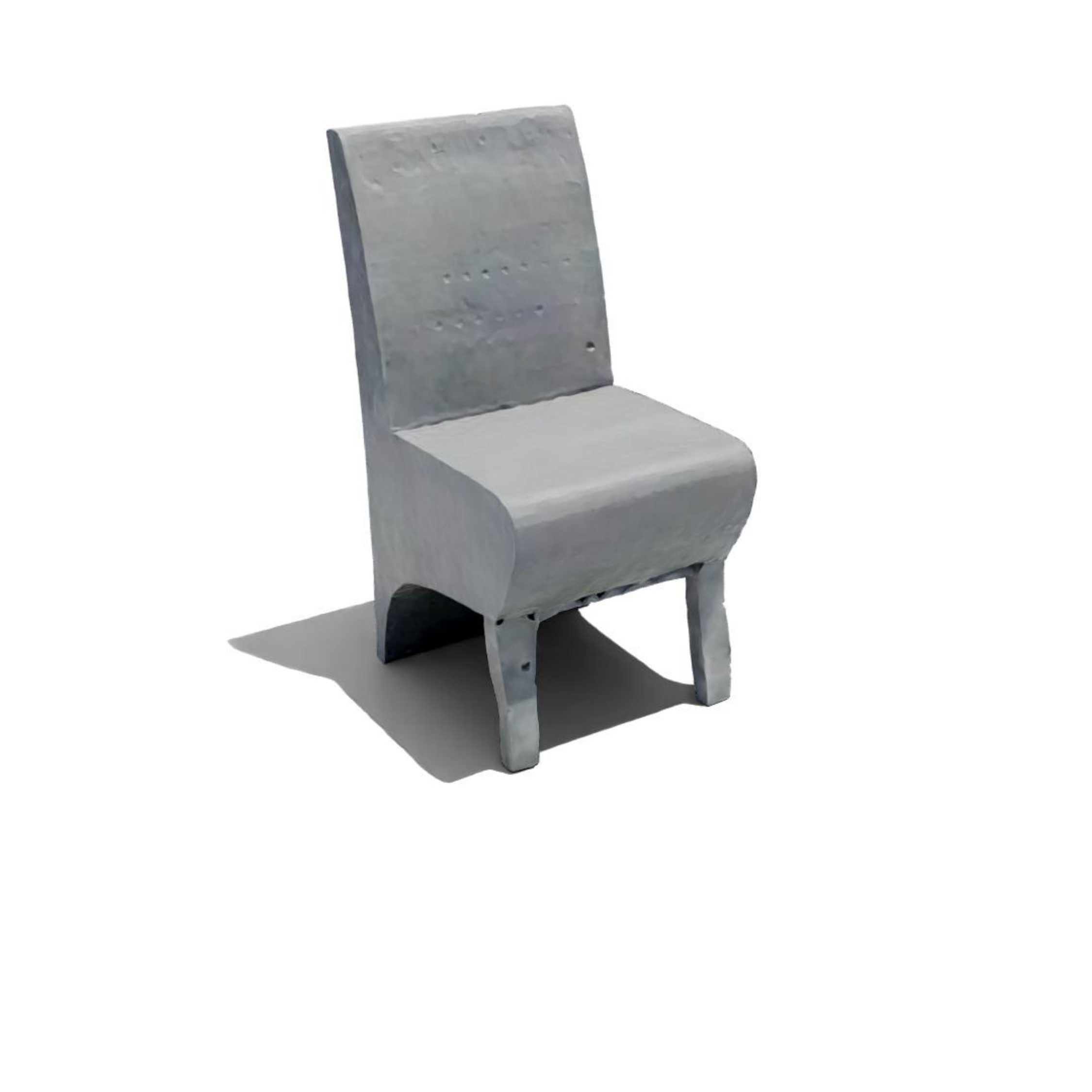}\\
\vspace{-0.3cm}
\includegraphics[width=0.16666666666666666\linewidth]{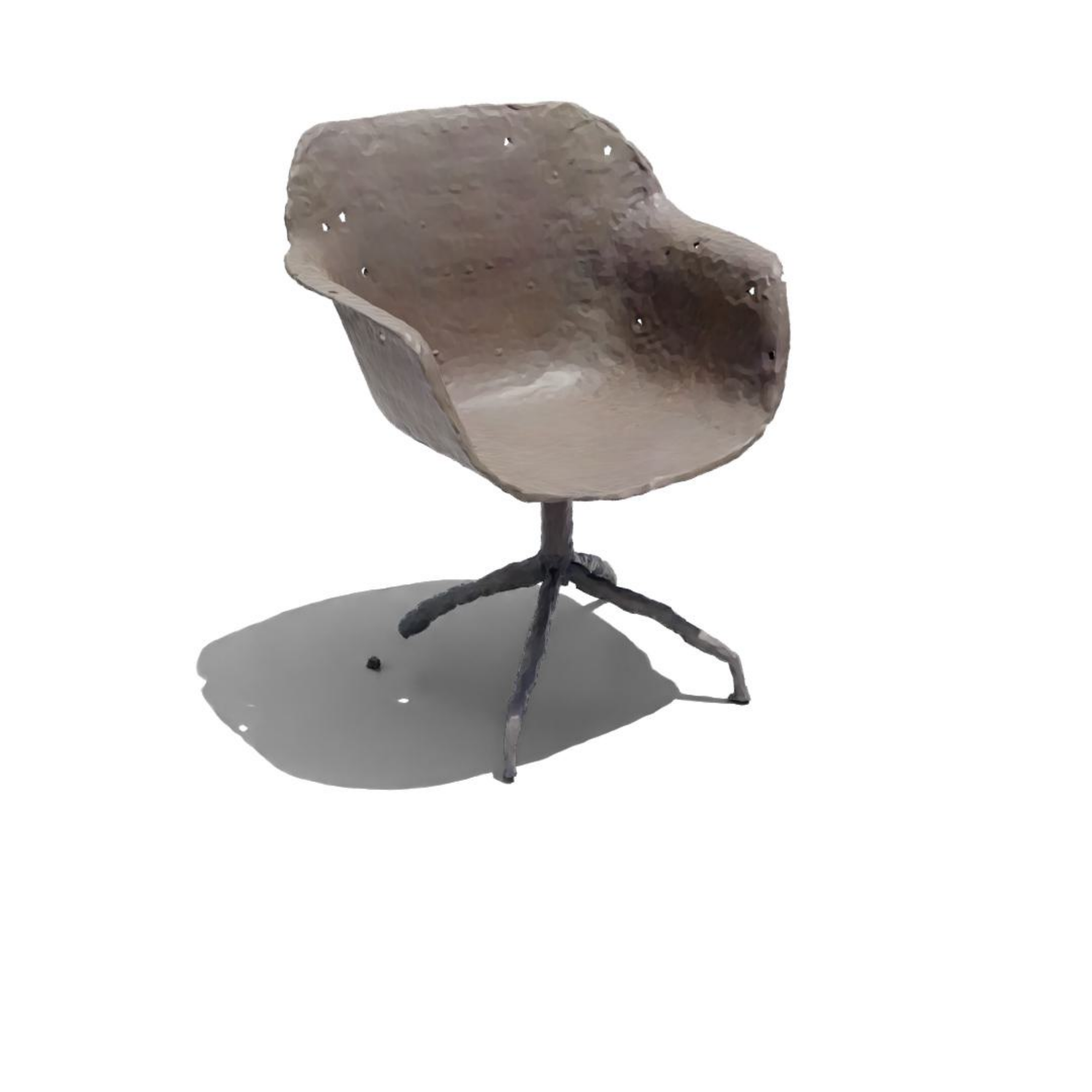}\includegraphics[width=0.16666666666666666\linewidth]{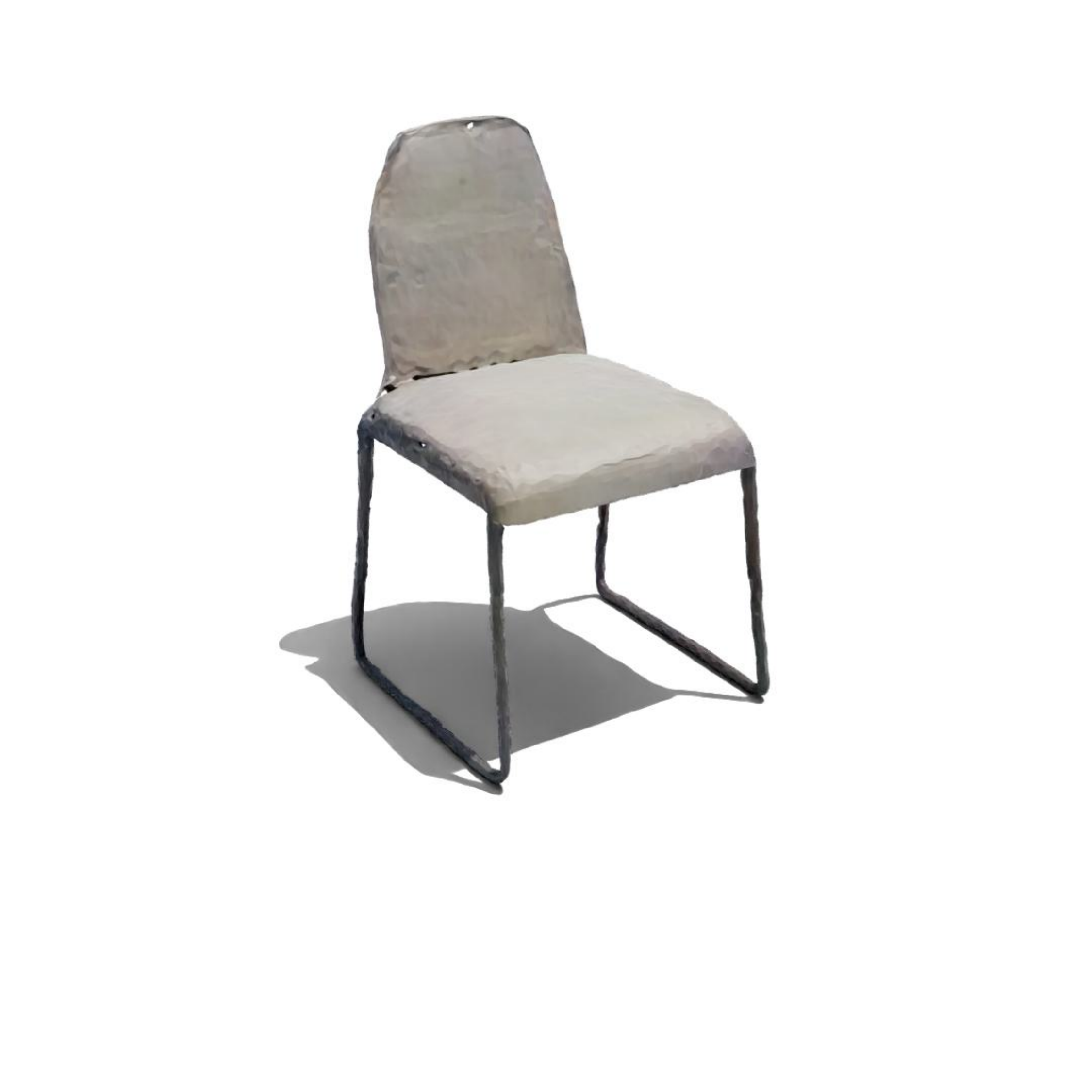}\includegraphics[width=0.16666666666666666\linewidth]{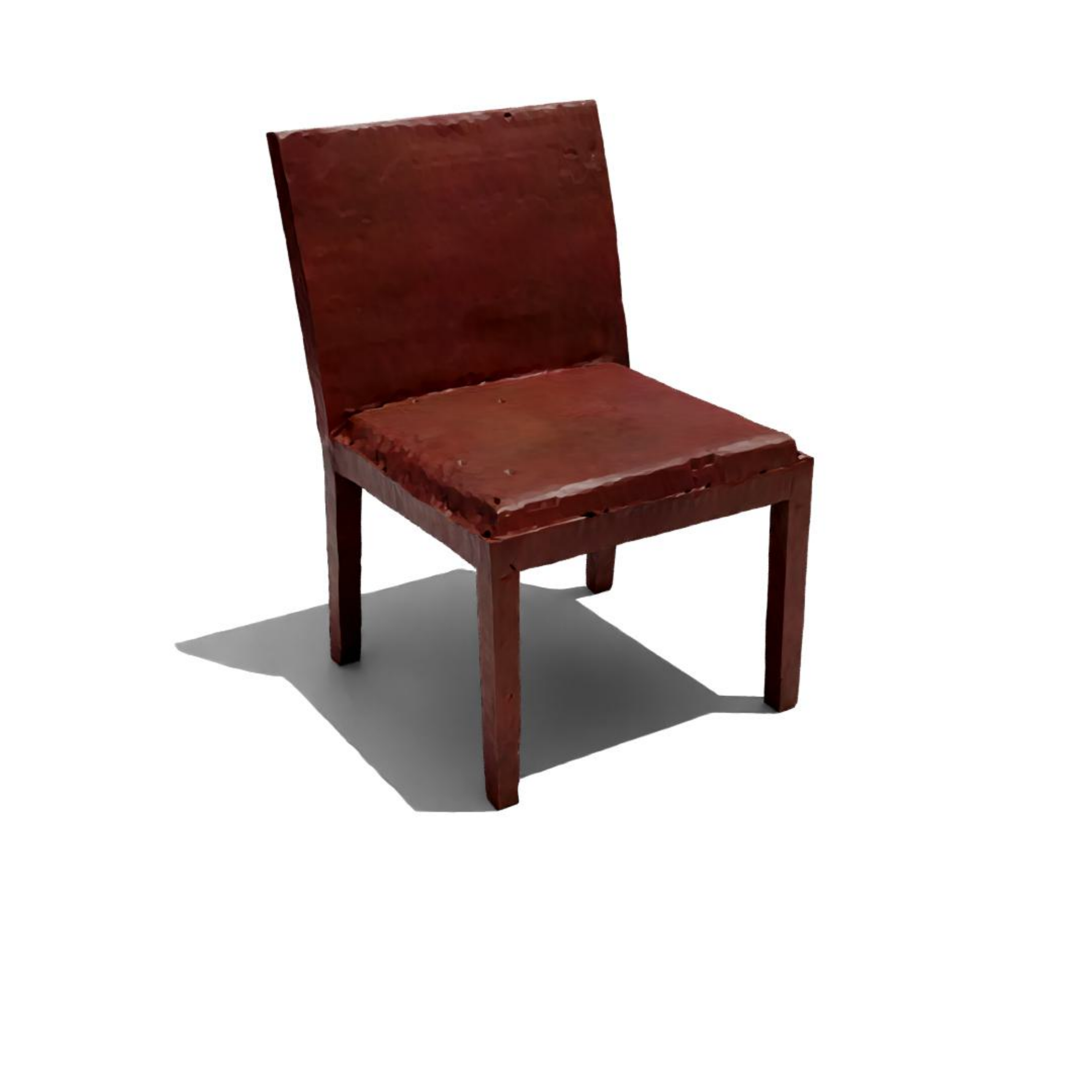}\includegraphics[width=0.16666666666666666\linewidth]{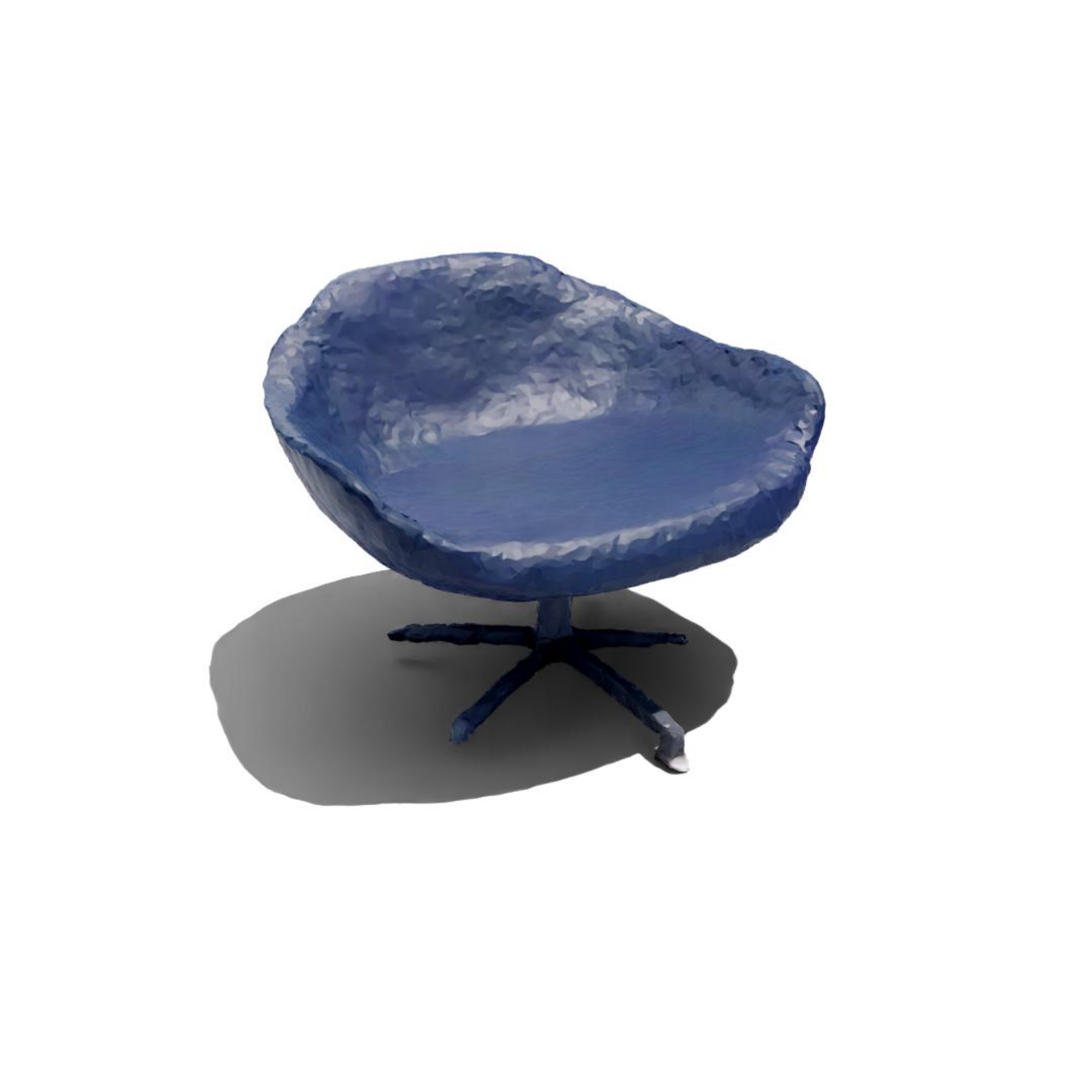}\includegraphics[width=0.16666666666666666\linewidth]{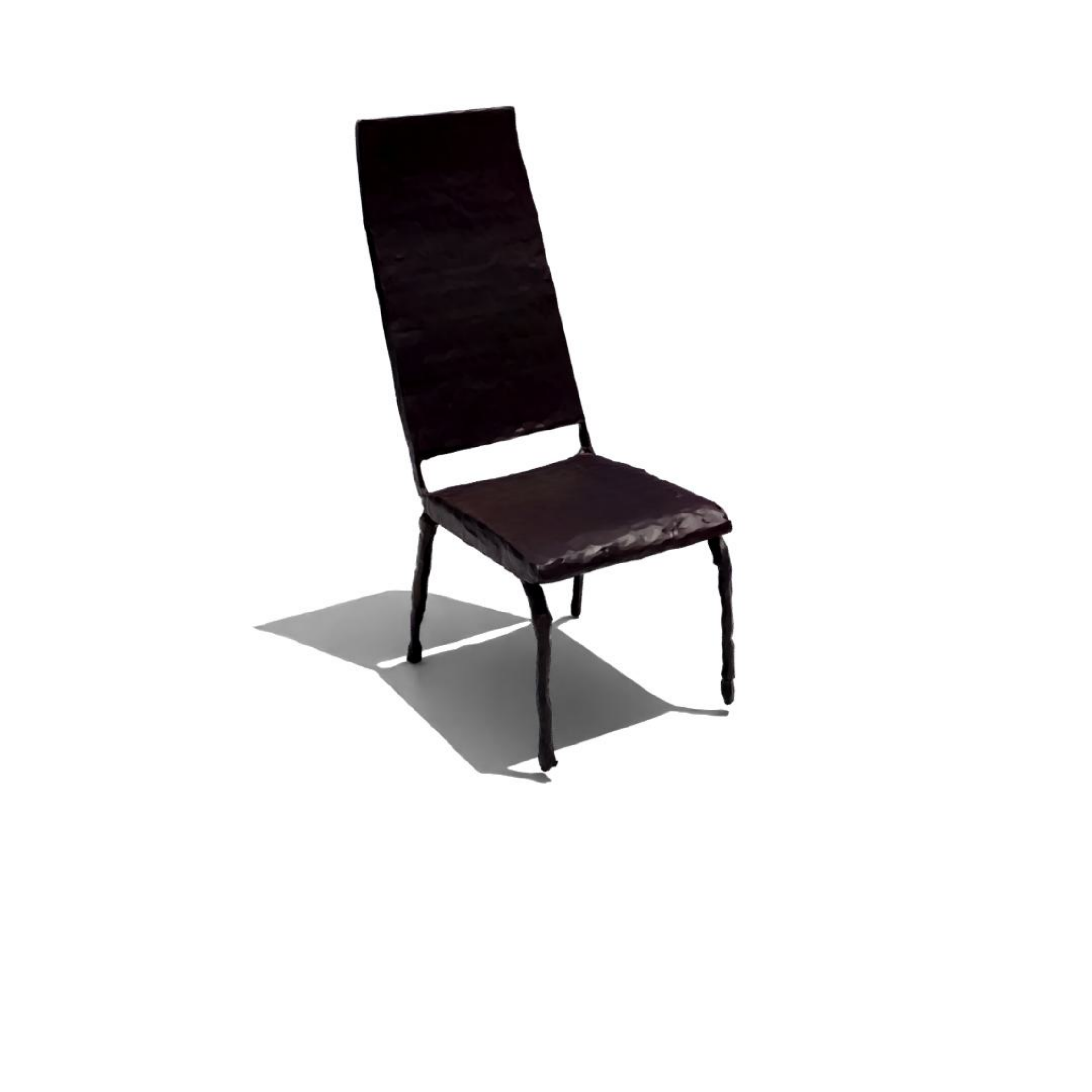}\includegraphics[width=0.16666666666666666\linewidth]{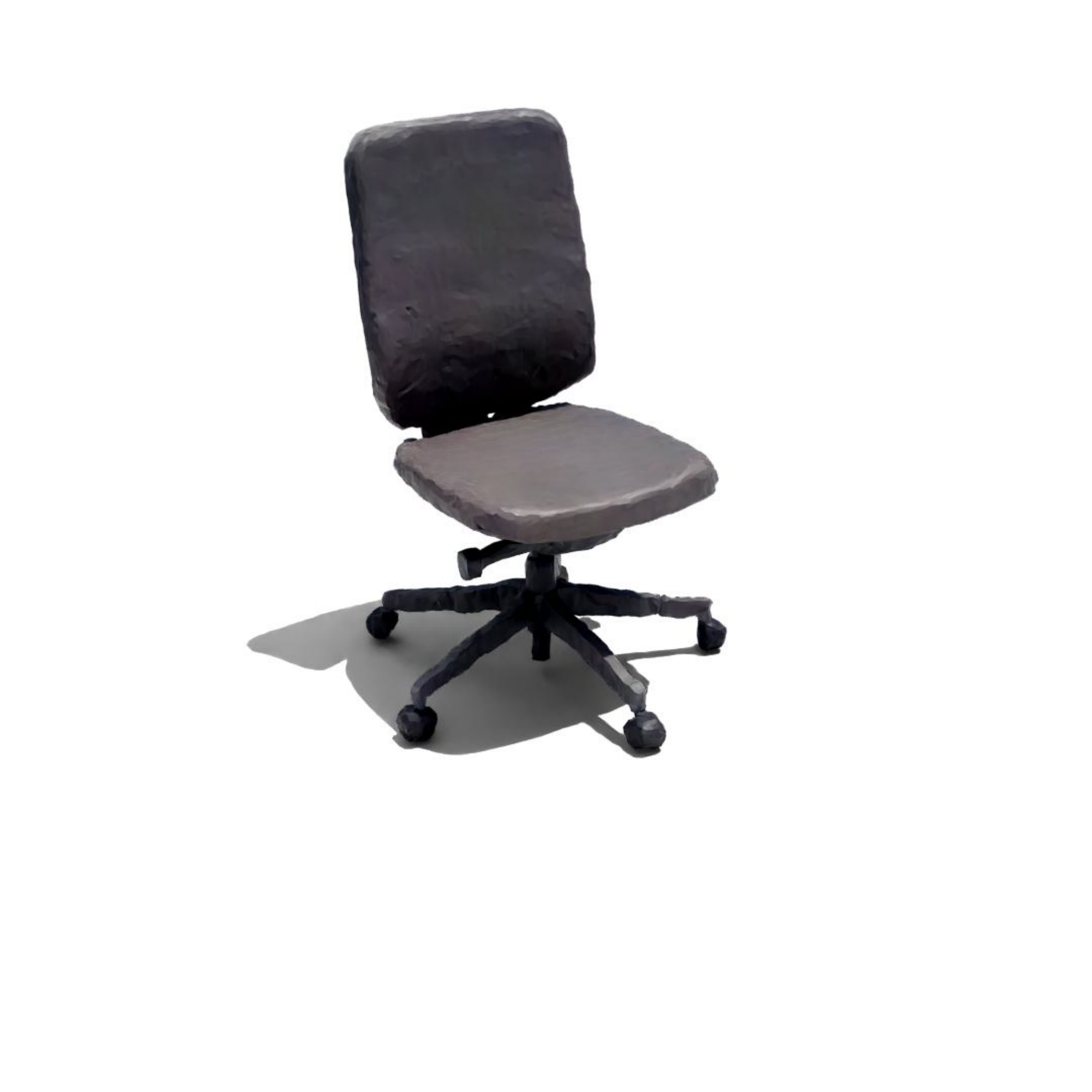}\\

\vspace{-0.3cm}
\includegraphics[width=0.16666666666666666\linewidth]{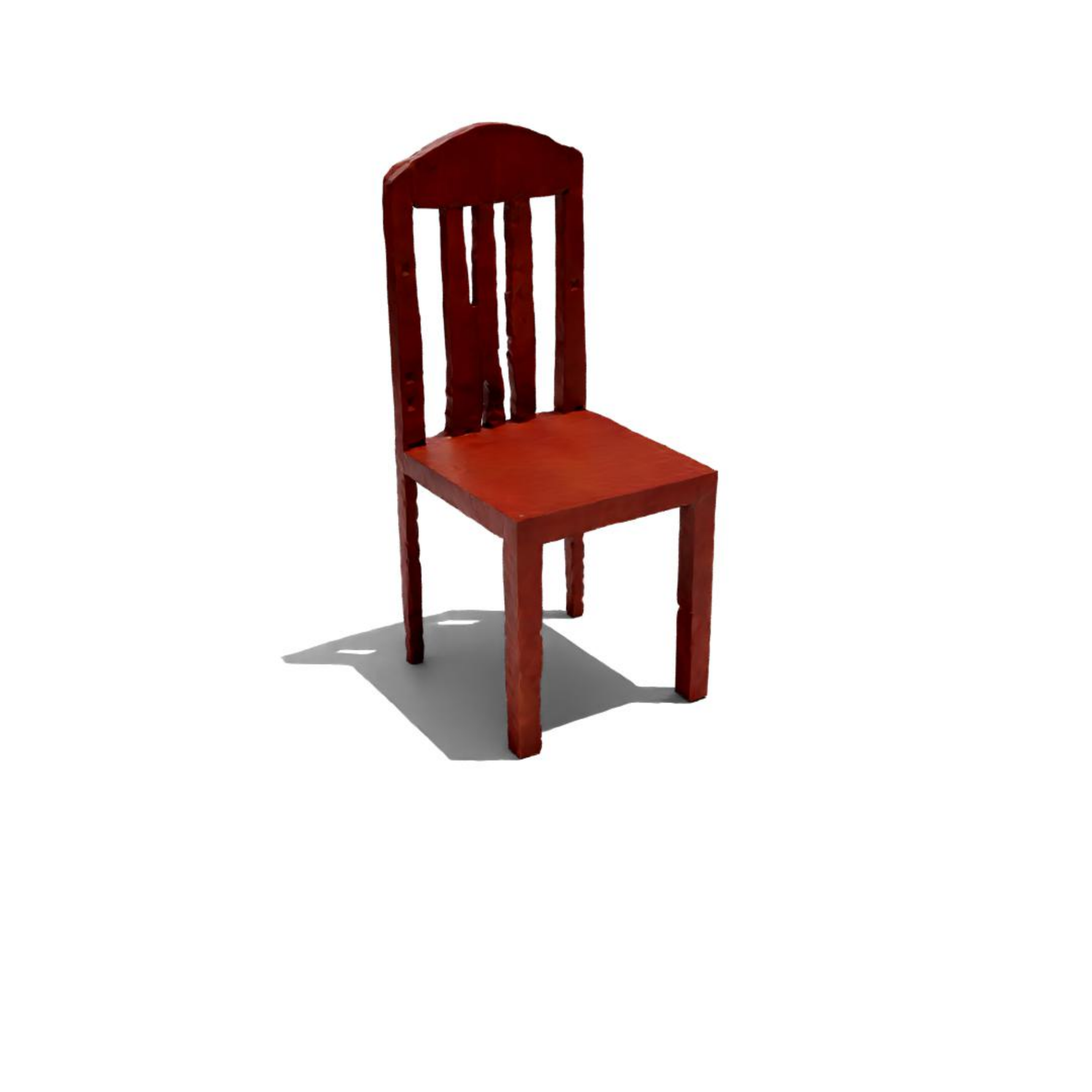}\includegraphics[width=0.16666666666666666\linewidth]{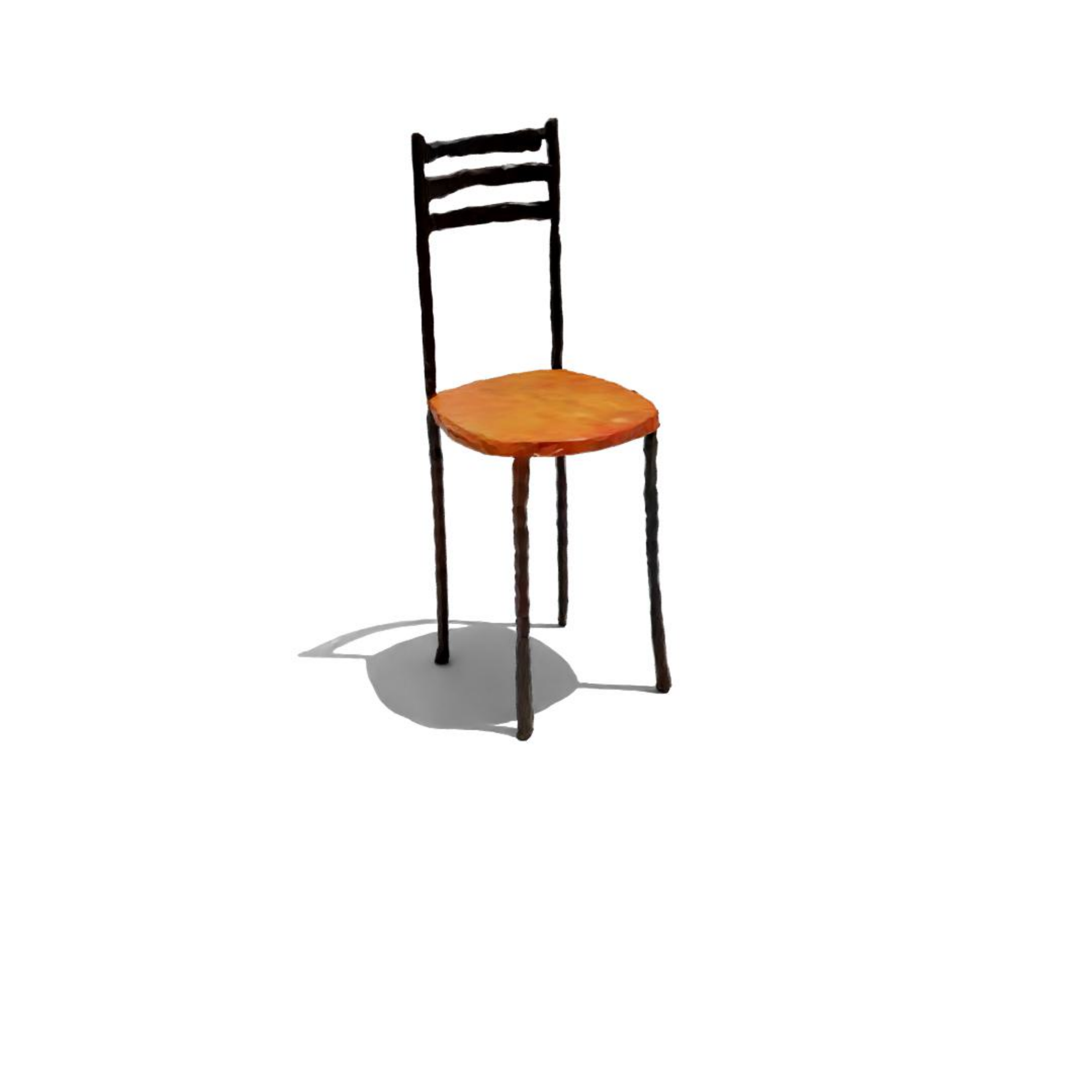}\includegraphics[width=0.16666666666666666\linewidth]{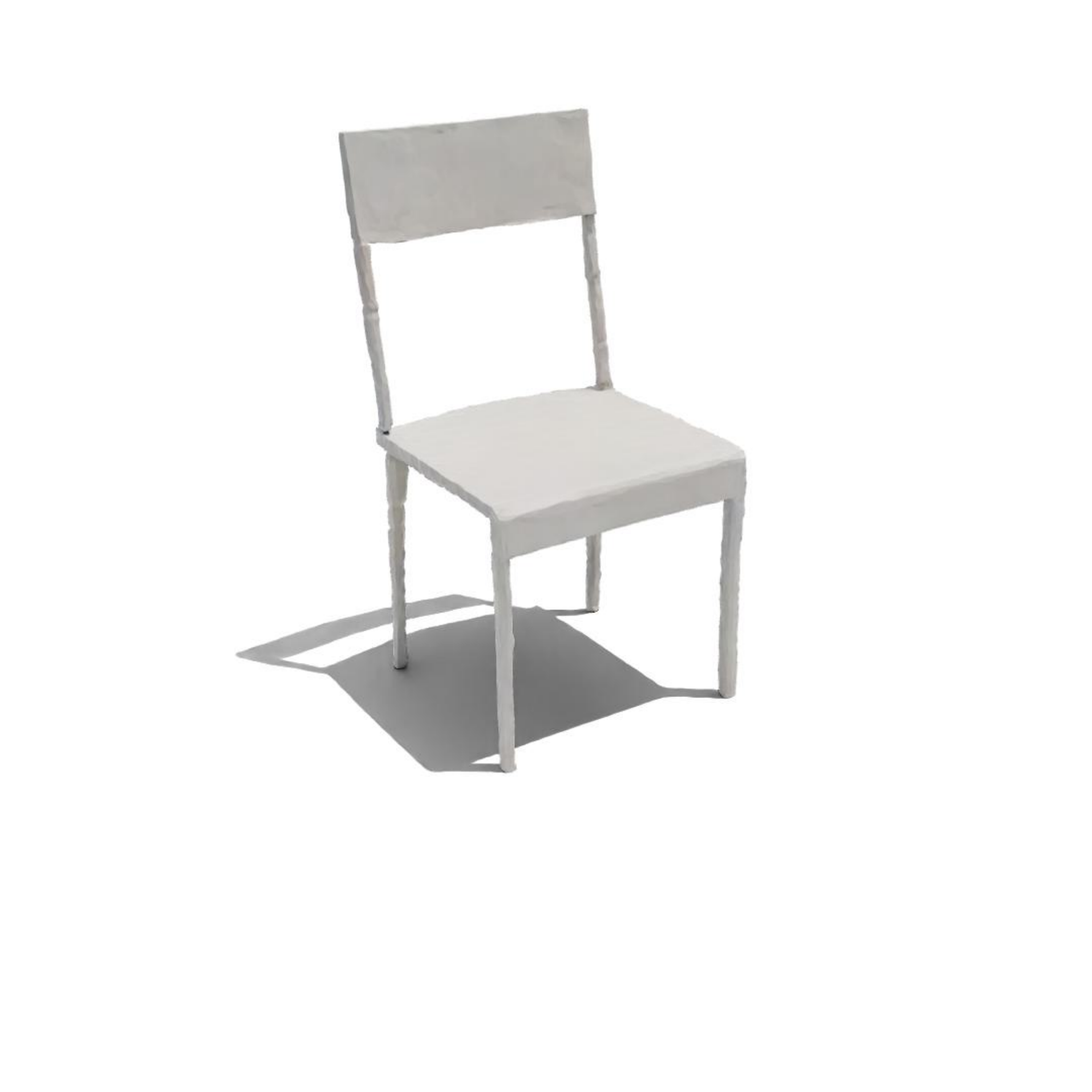}\includegraphics[width=0.16666666666666666\linewidth]{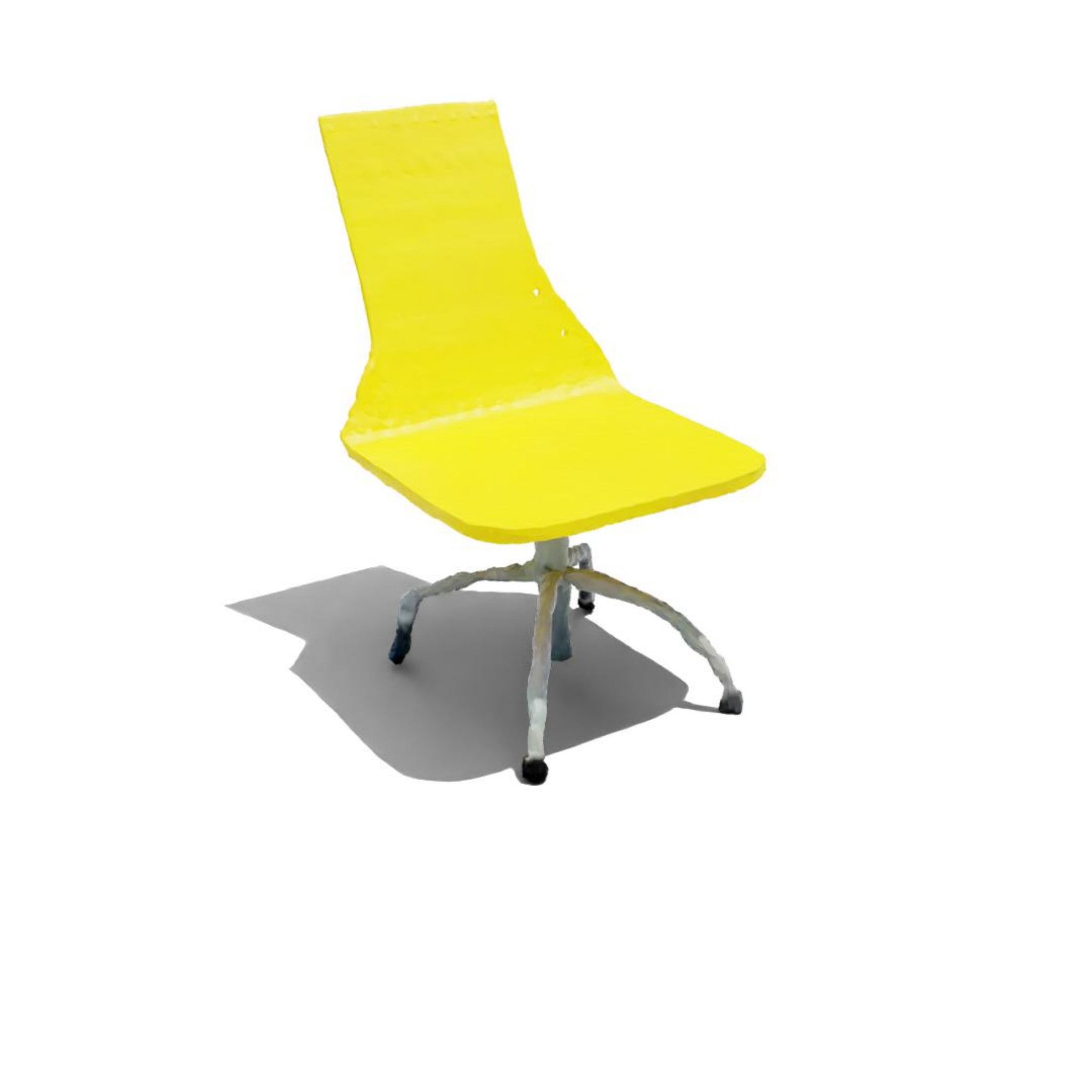}\includegraphics[width=0.16666666666666666\linewidth]{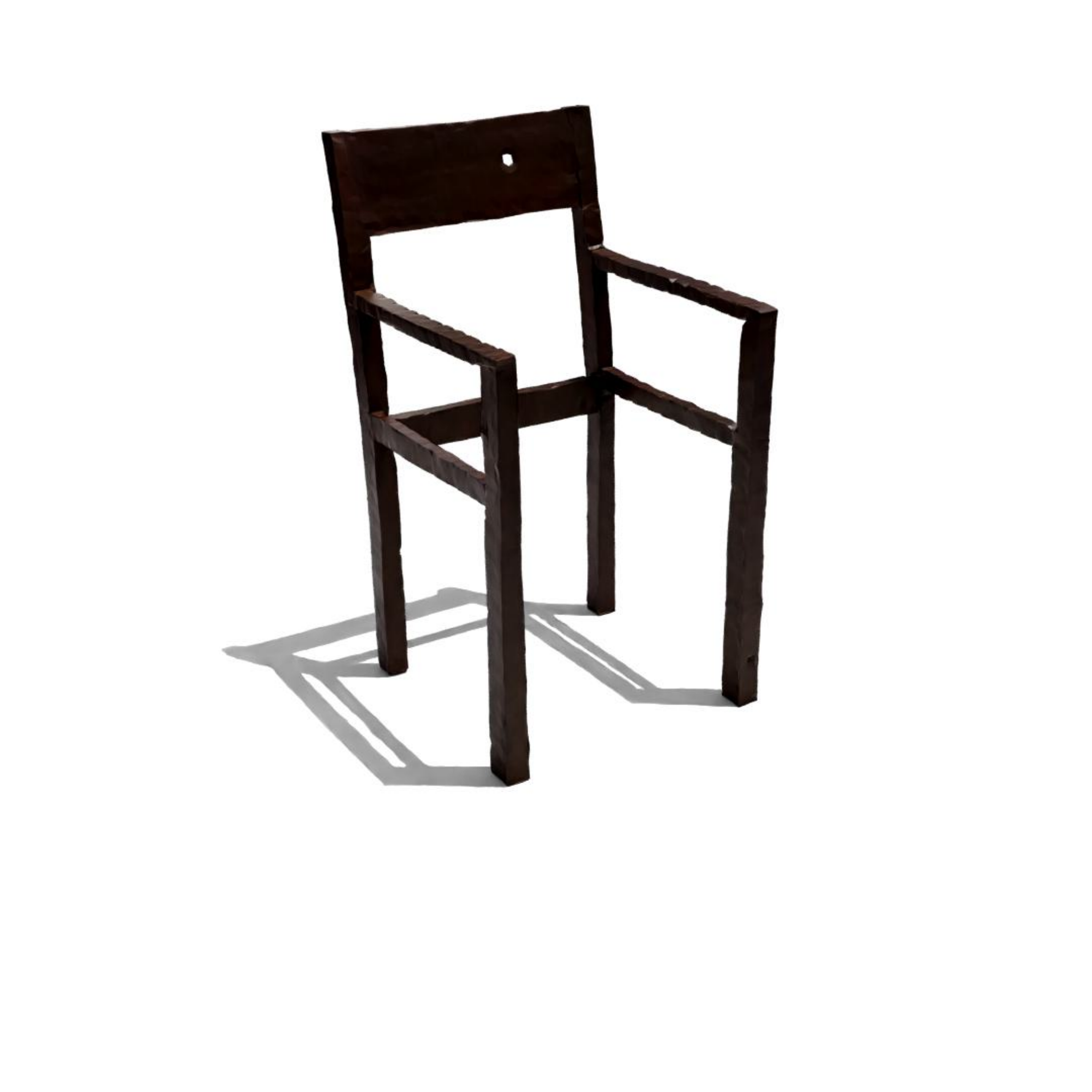}\includegraphics[width=0.16666666666666666\linewidth]{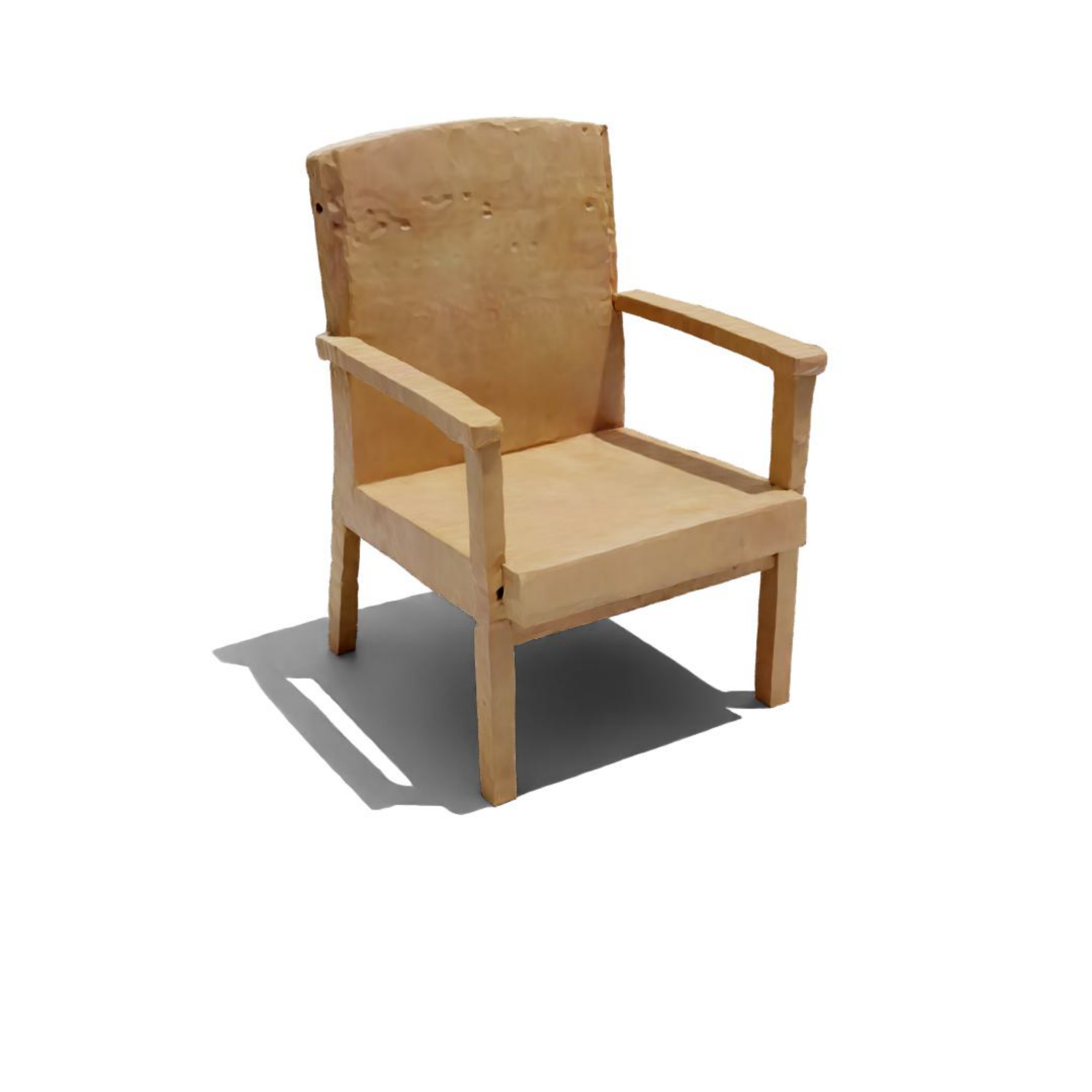}\\

\vspace{-0.3cm}
\includegraphics[width=0.16666666666666666\linewidth]{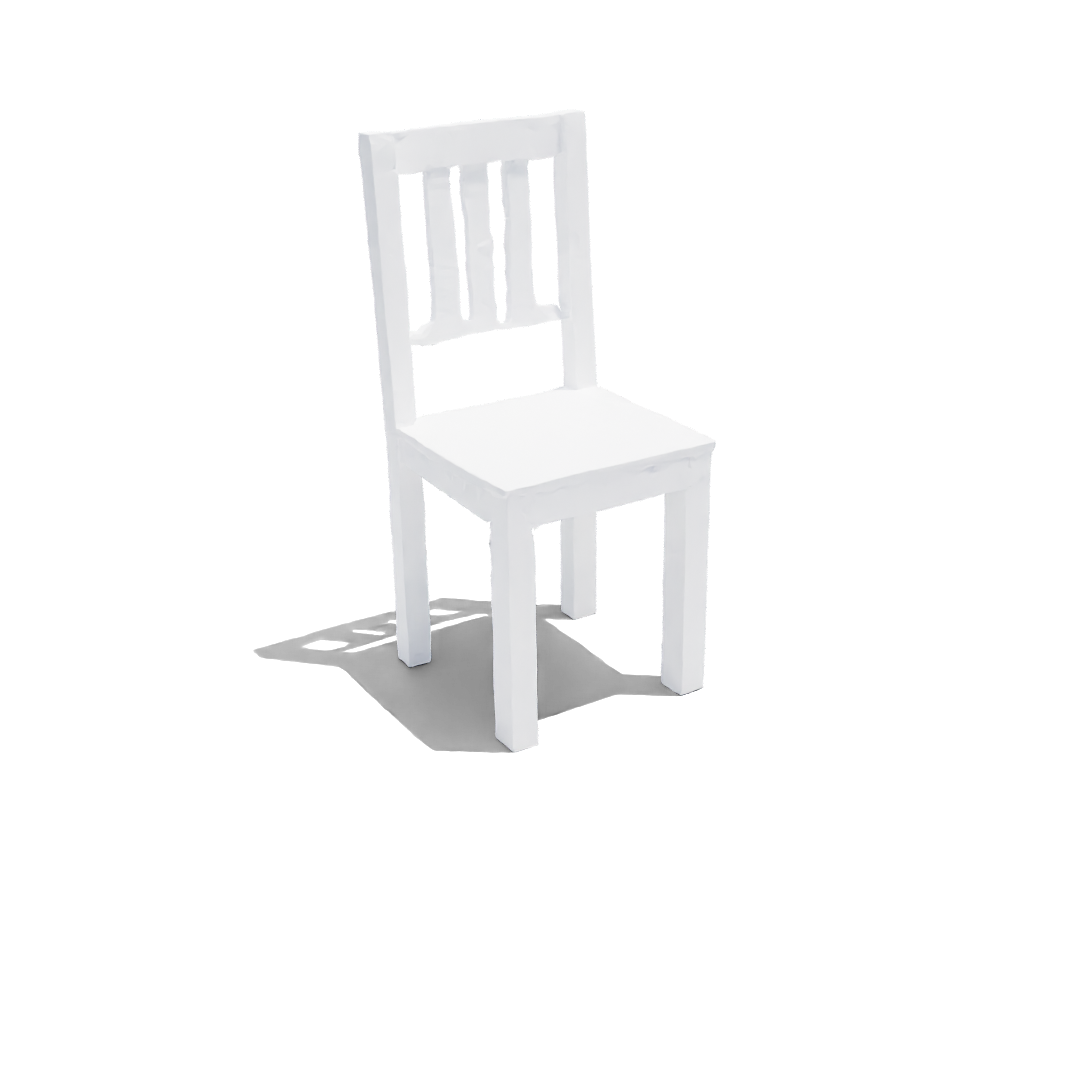}\includegraphics[width=0.16666666666666666\linewidth]{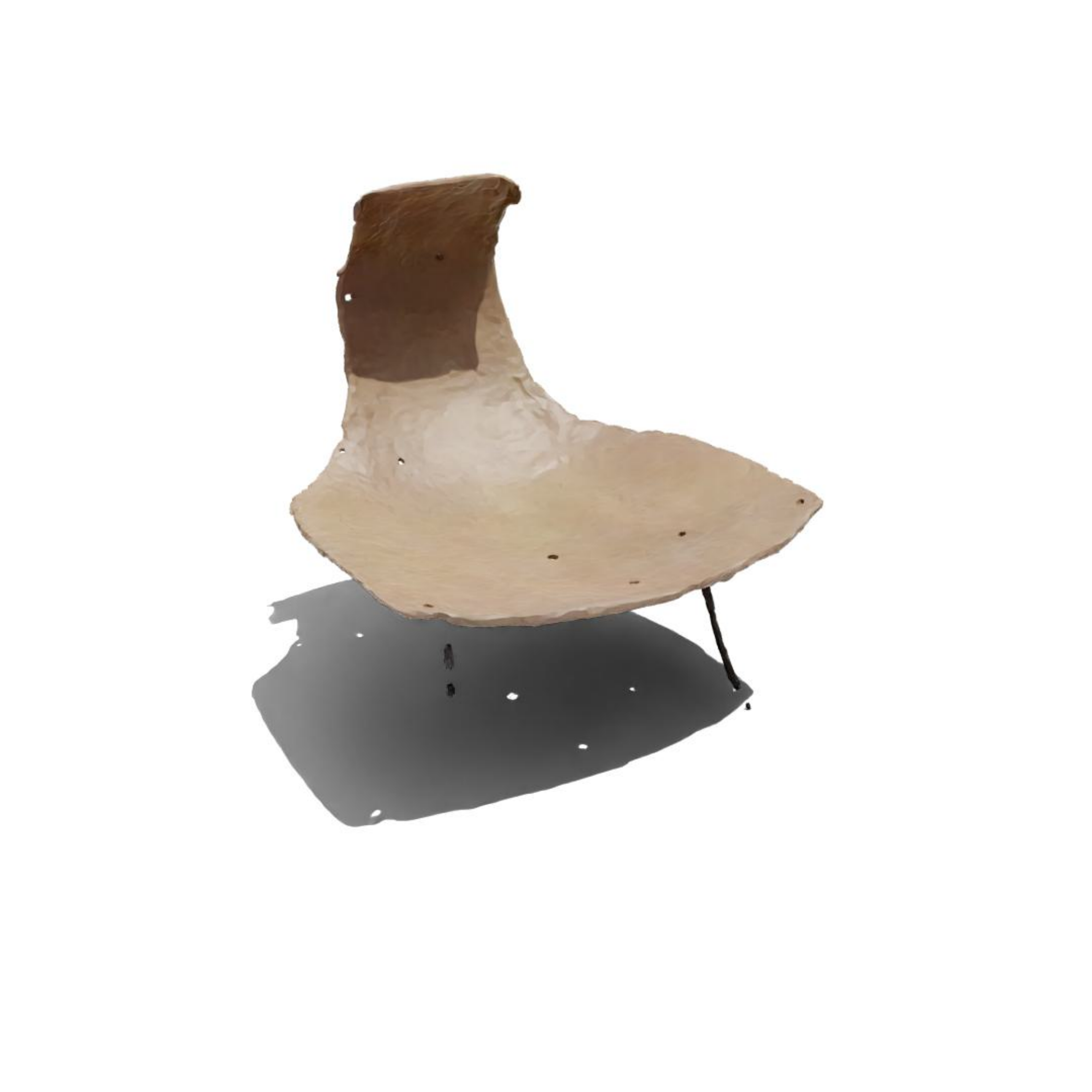}\includegraphics[width=0.16666666666666666\linewidth]{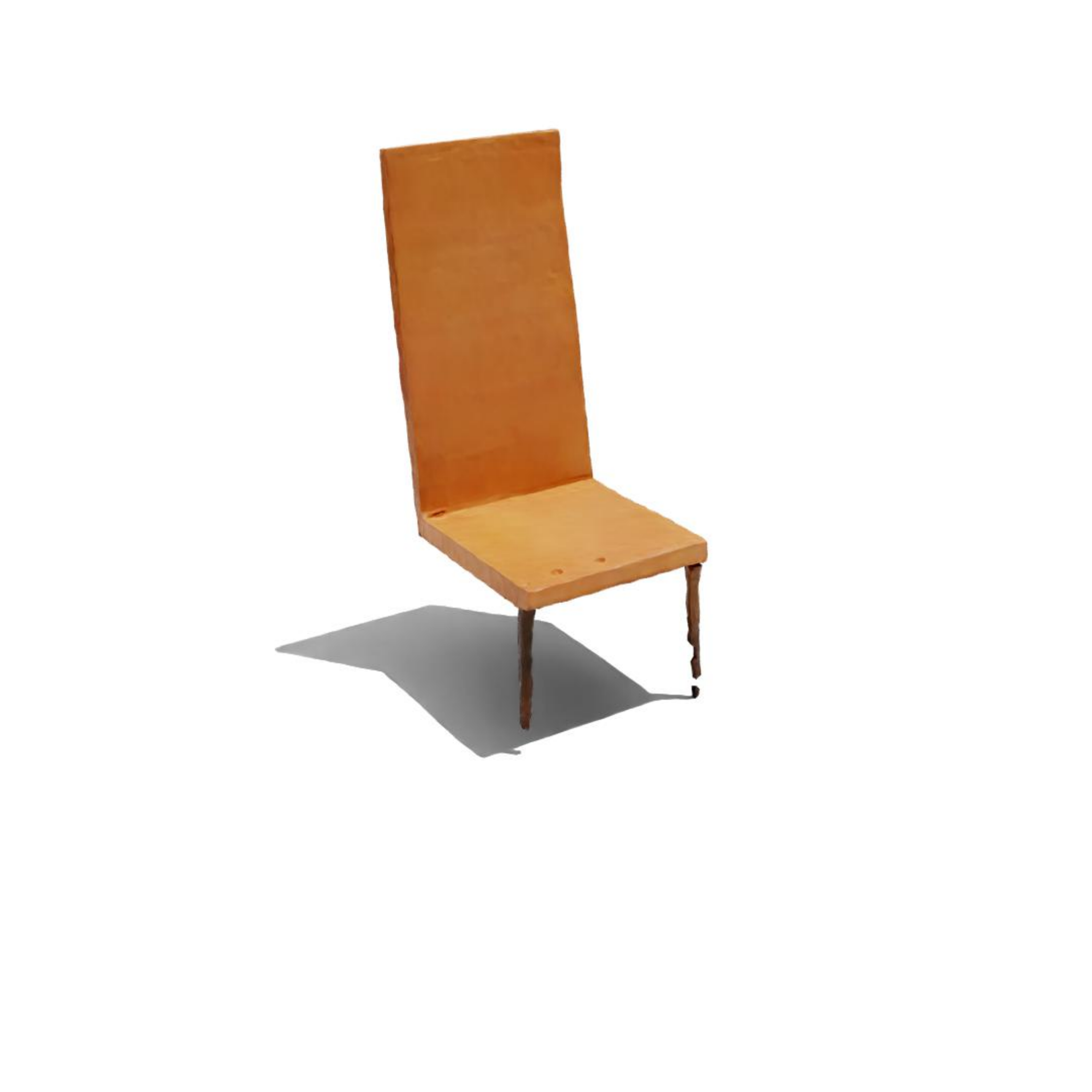}\includegraphics[width=0.16666666666666666\linewidth]{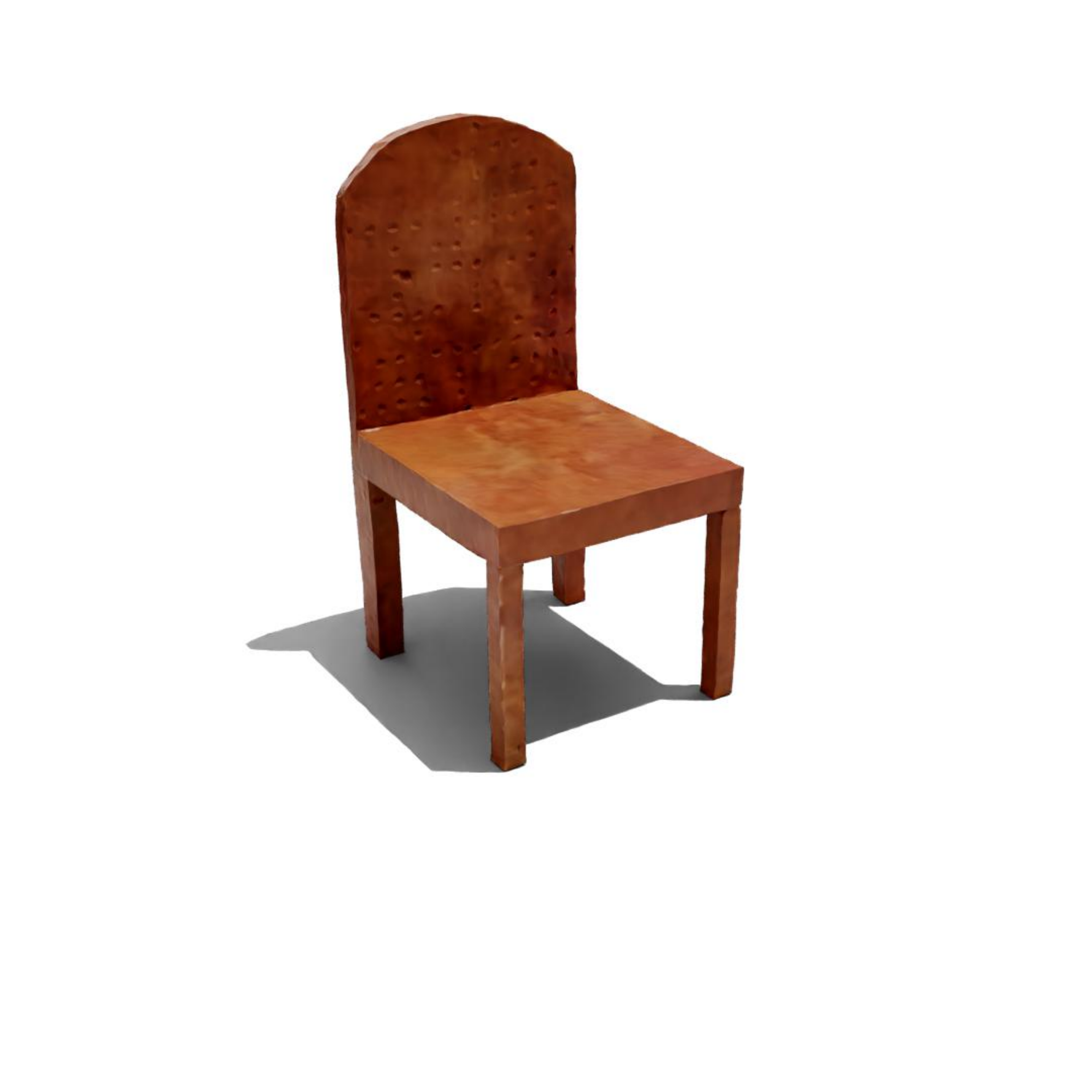}\includegraphics[width=0.16666666666666666\linewidth]{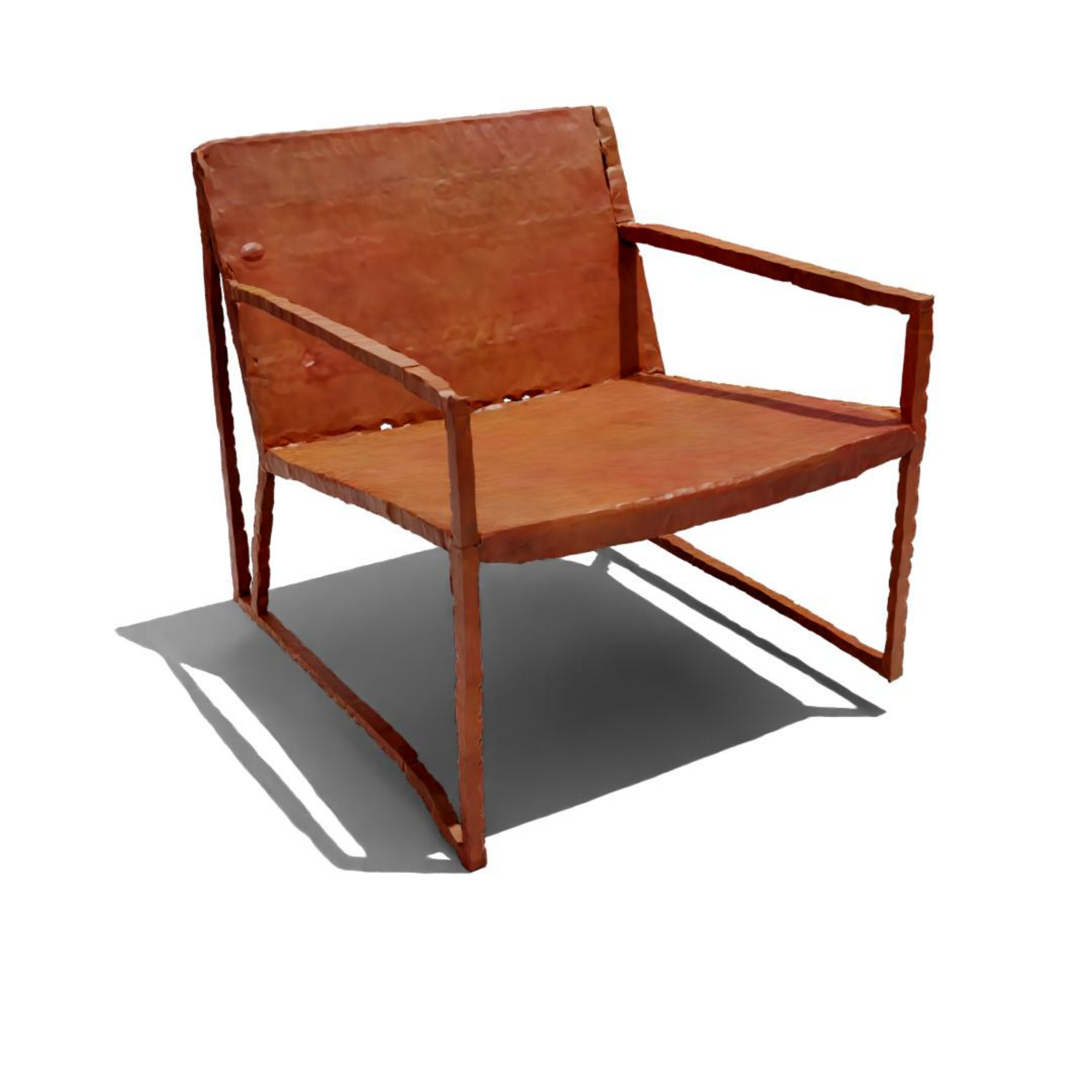}\includegraphics[width=0.16666666666666666\linewidth]{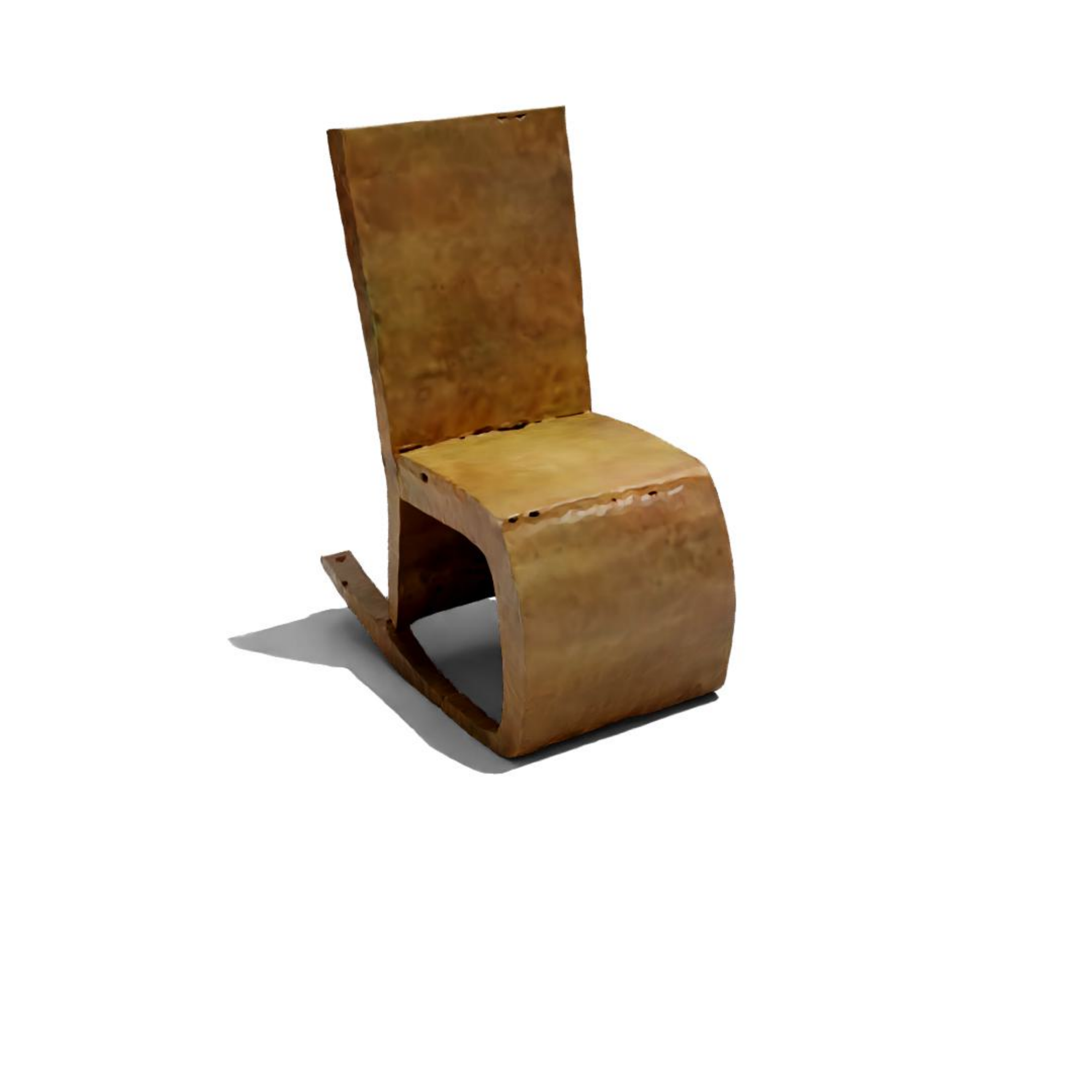}\\

\vspace{-0.3cm}
\includegraphics[width=0.16666666666666666\linewidth]{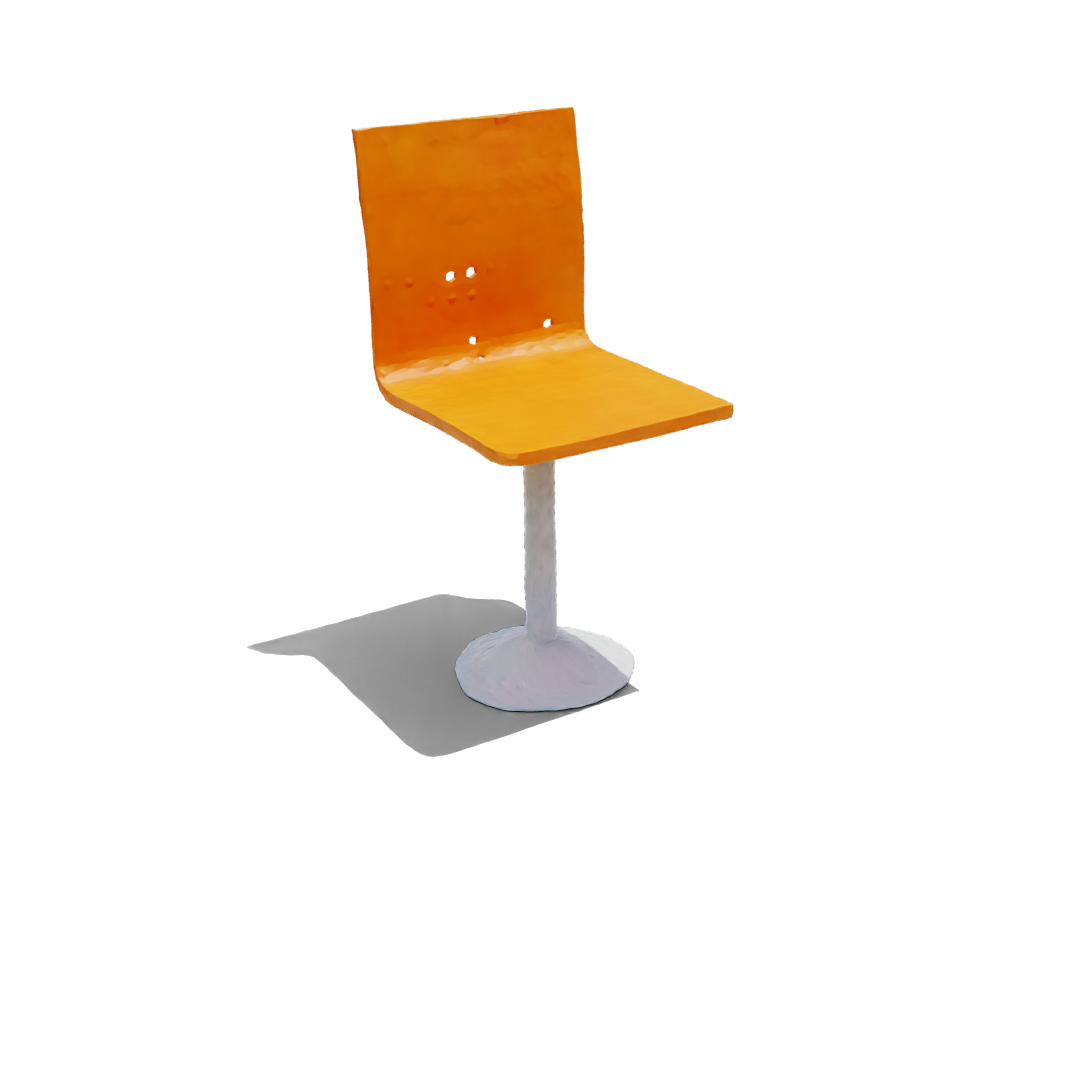}\includegraphics[width=0.16666666666666666\linewidth]{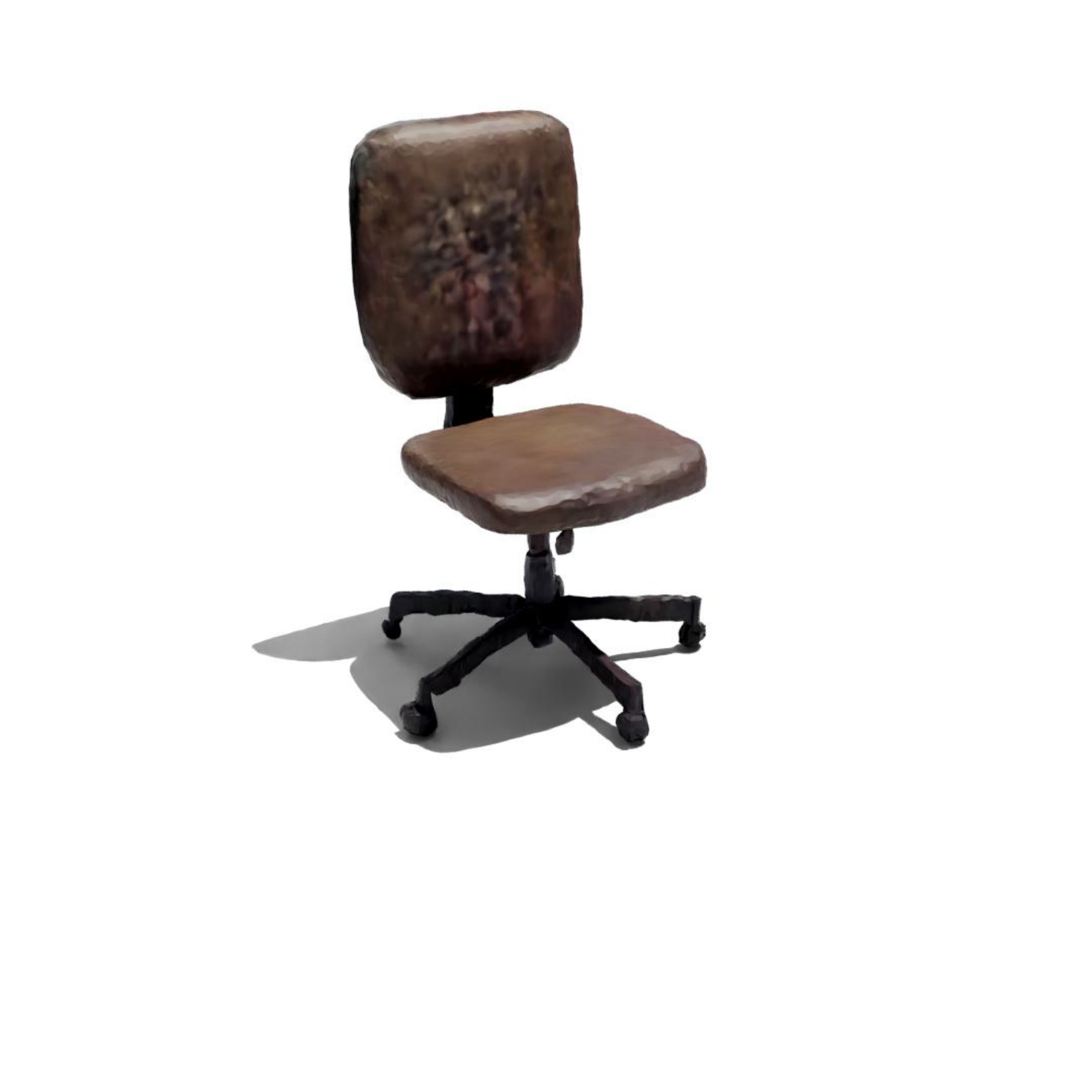}\includegraphics[width=0.16666666666666666\linewidth]{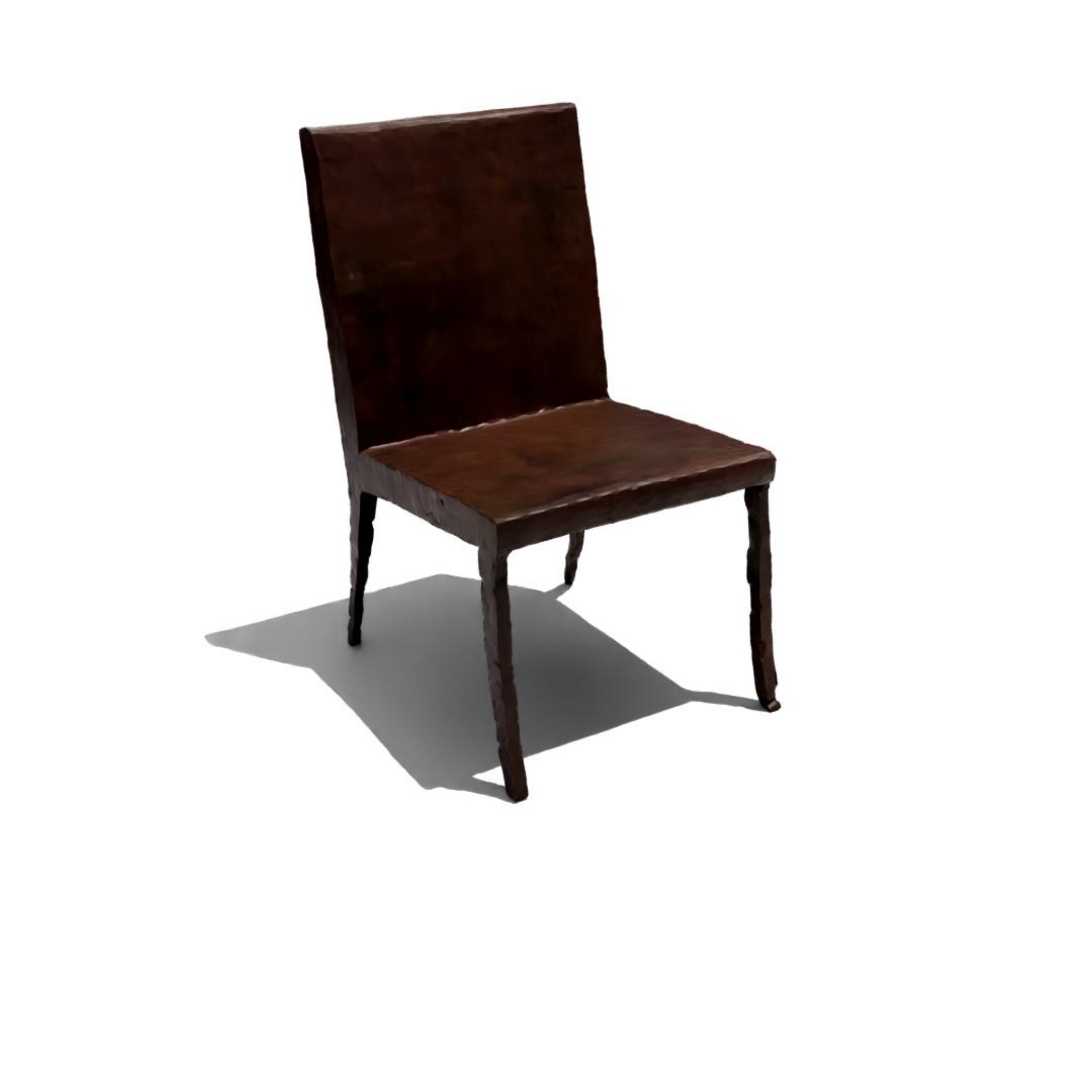}\includegraphics[width=0.16666666666666666\linewidth]{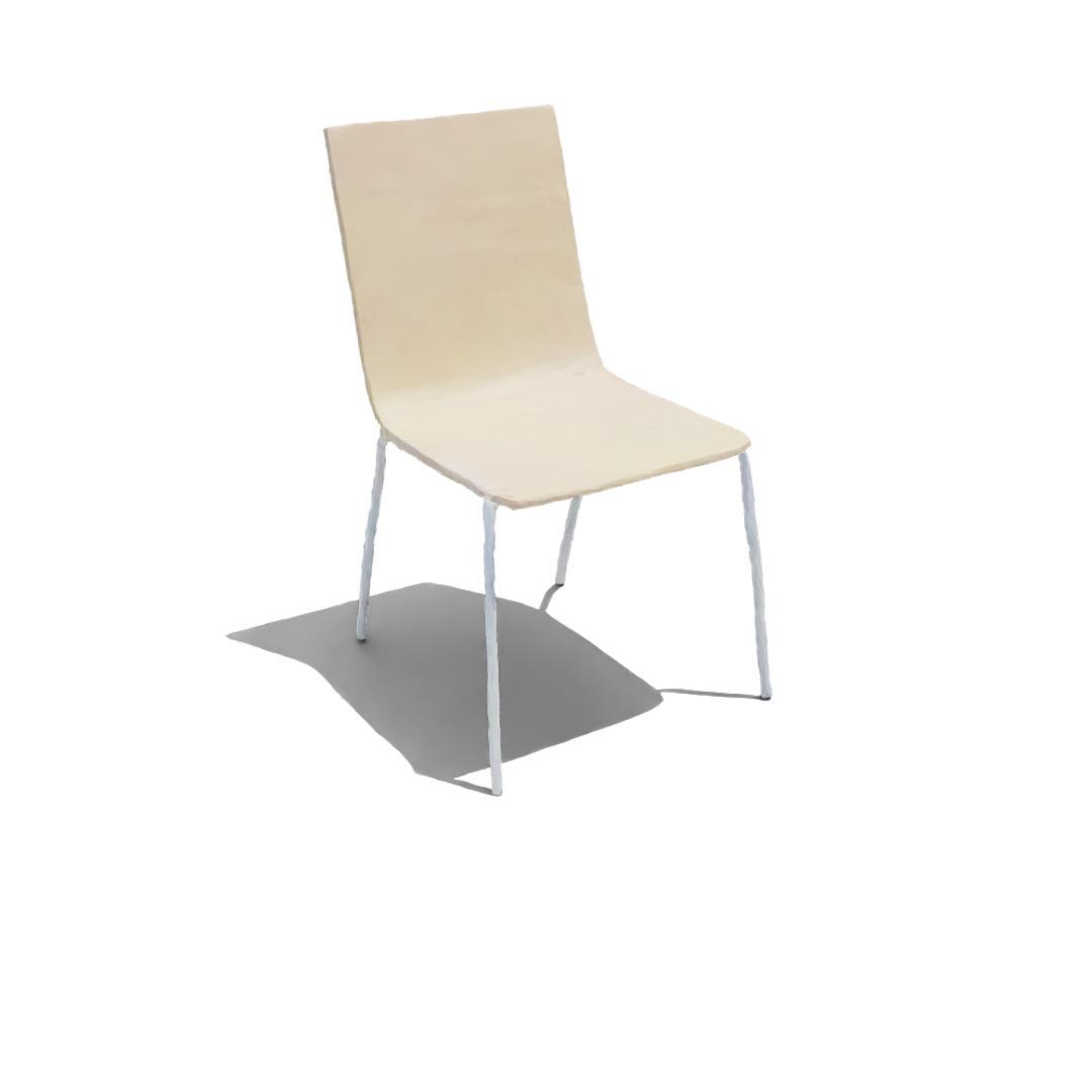}\includegraphics[width=0.16666666666666666\linewidth]{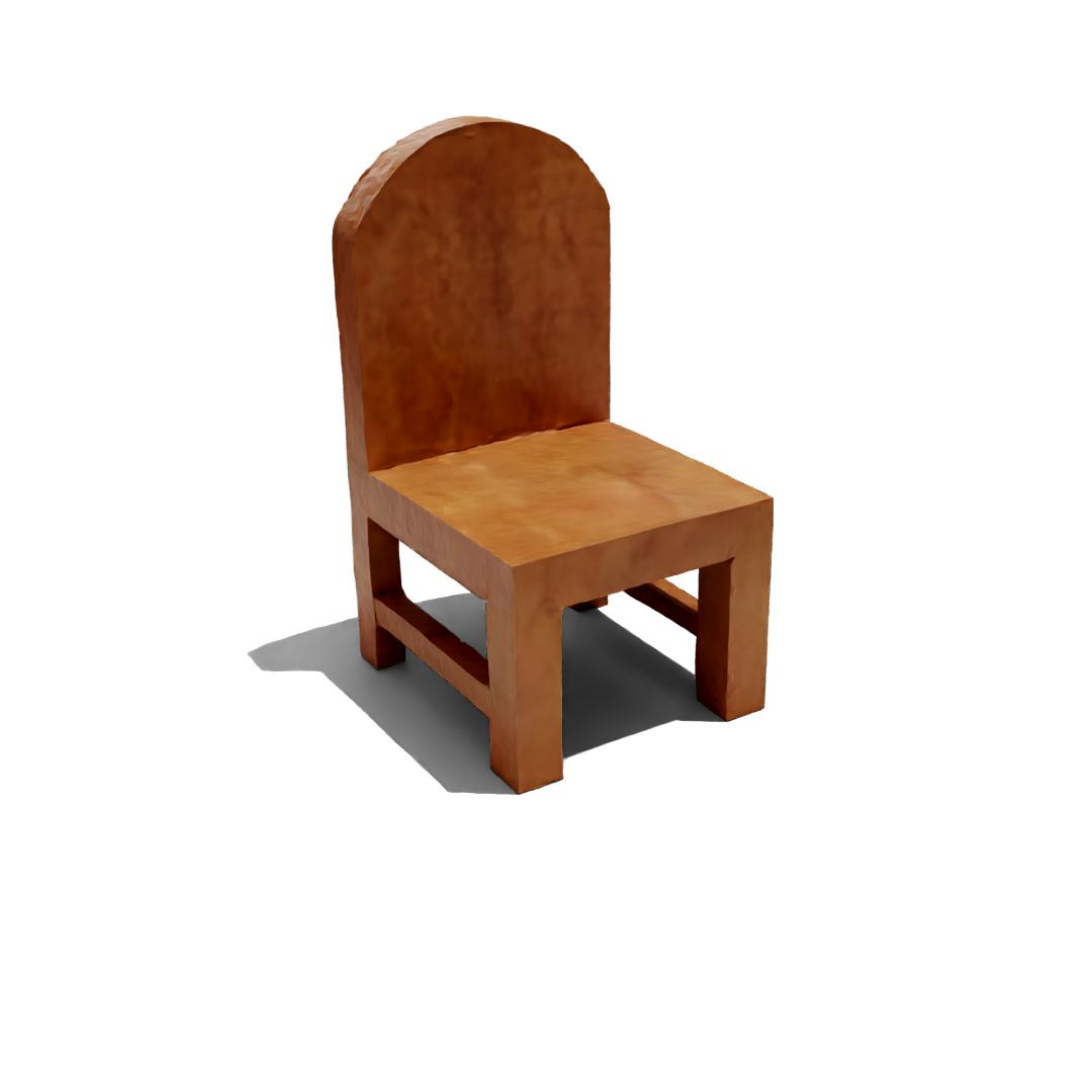}\includegraphics[width=0.16666666666666666\linewidth]{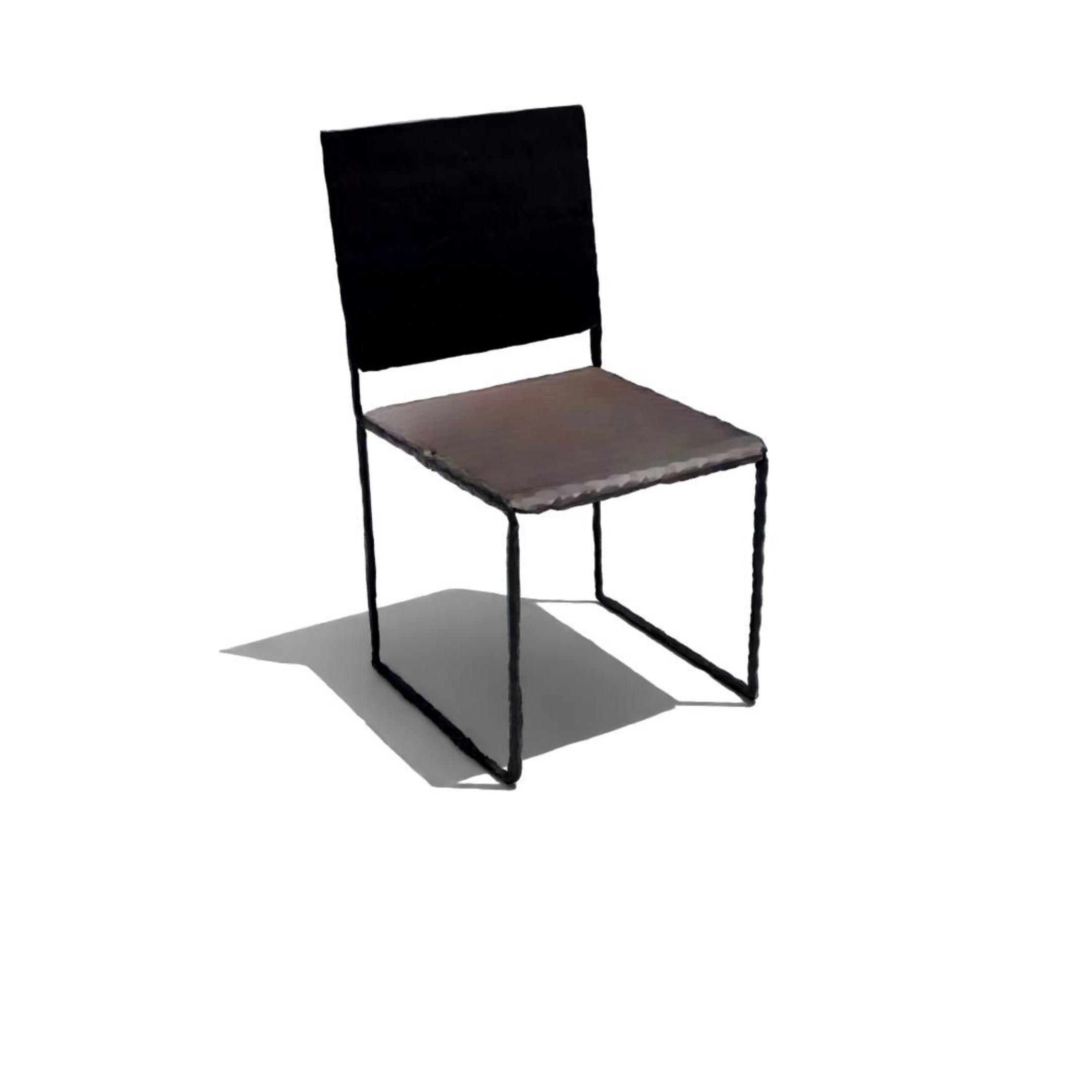}\\

\vspace{-0.3cm}
\includegraphics[width=0.16666666666666666\linewidth]{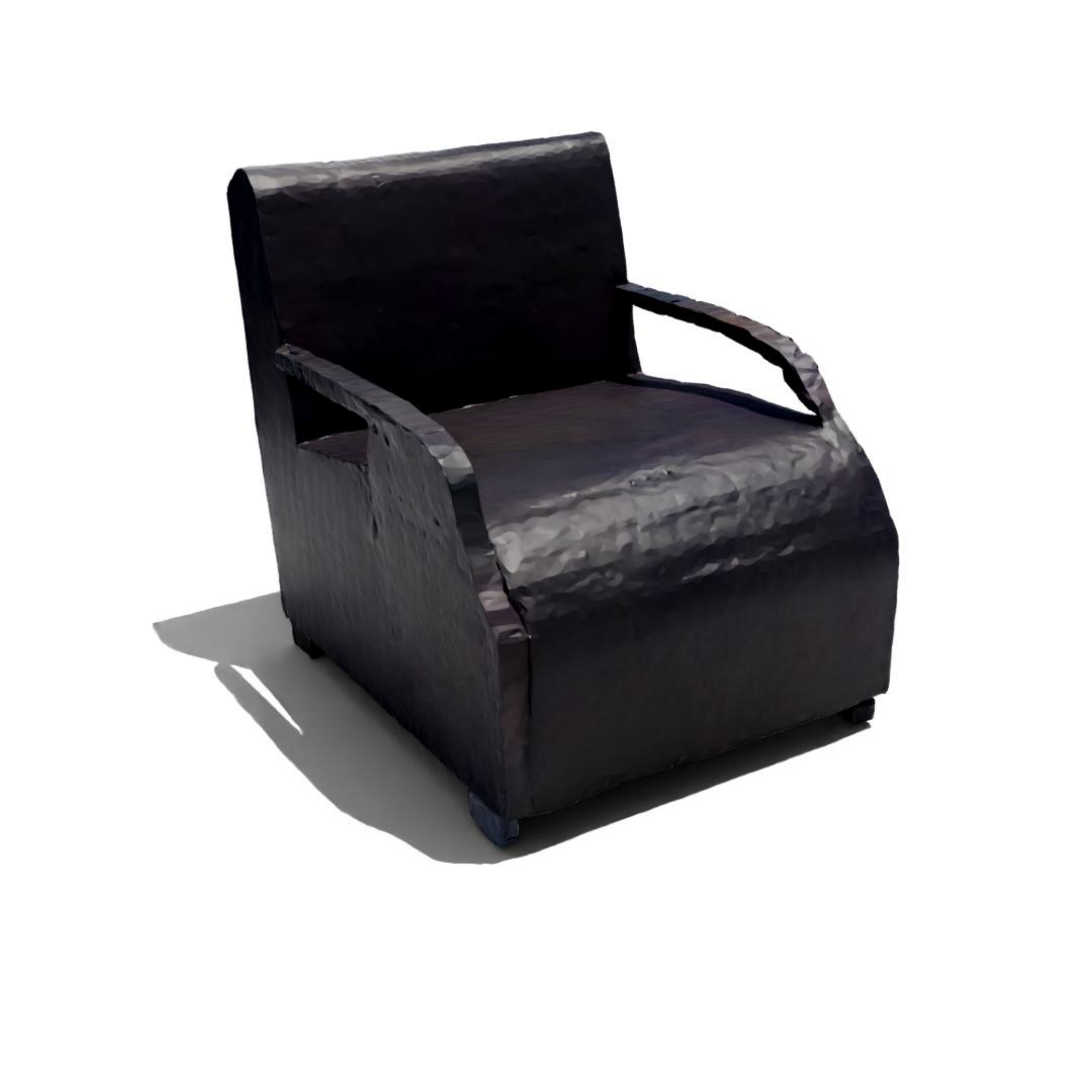}\includegraphics[width=0.16666666666666666\linewidth]{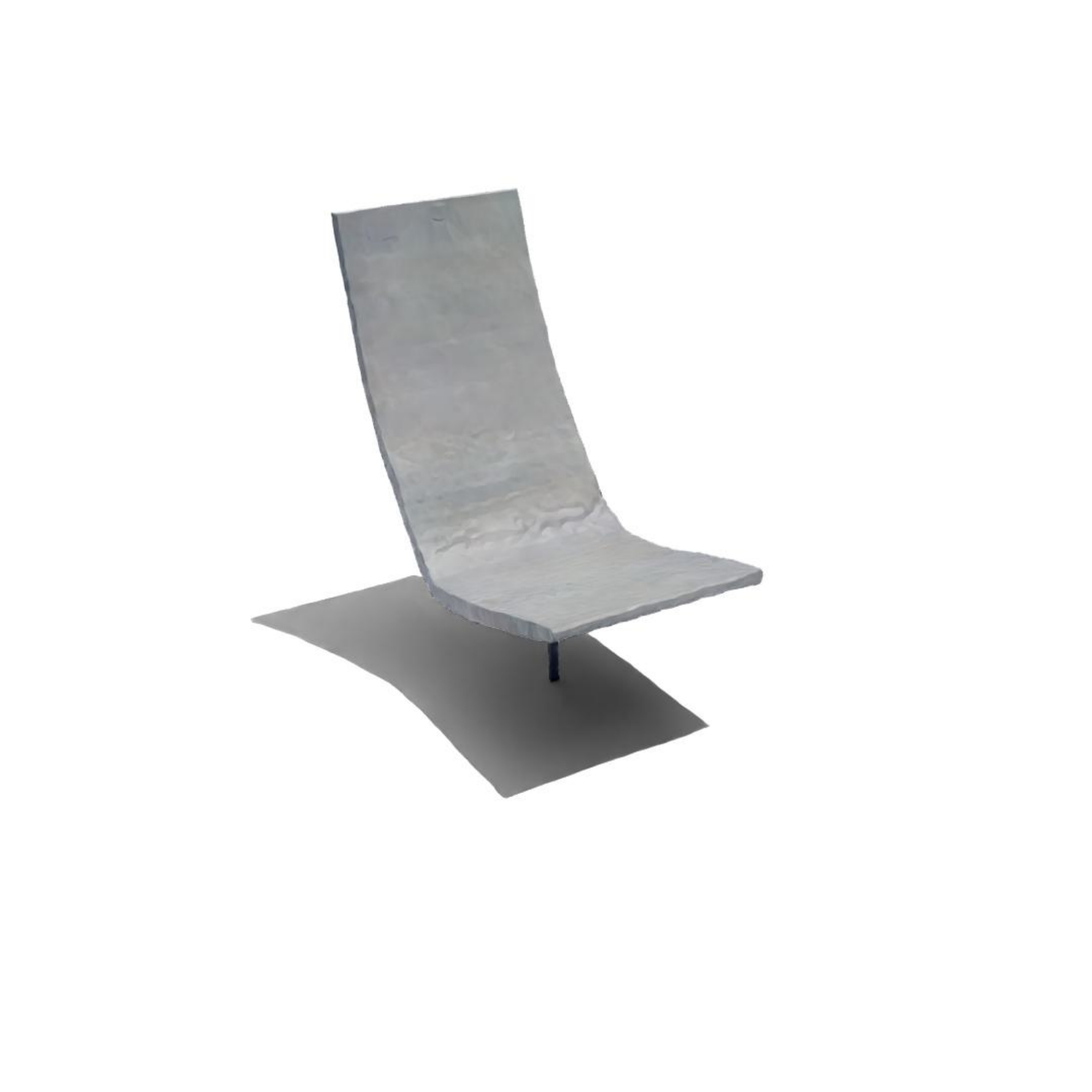}\includegraphics[width=0.16666666666666666\linewidth]{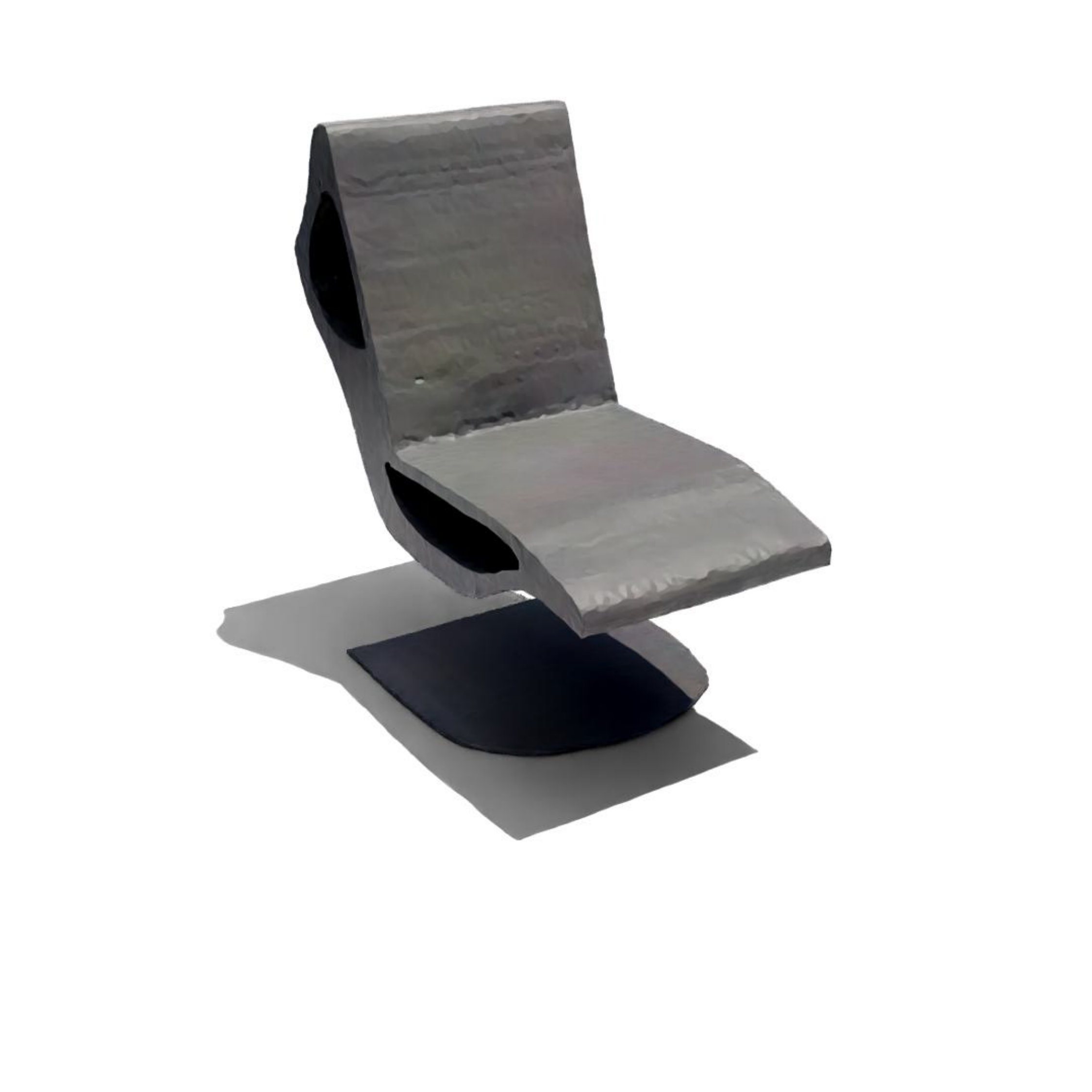}\includegraphics[width=0.16666666666666666\linewidth]{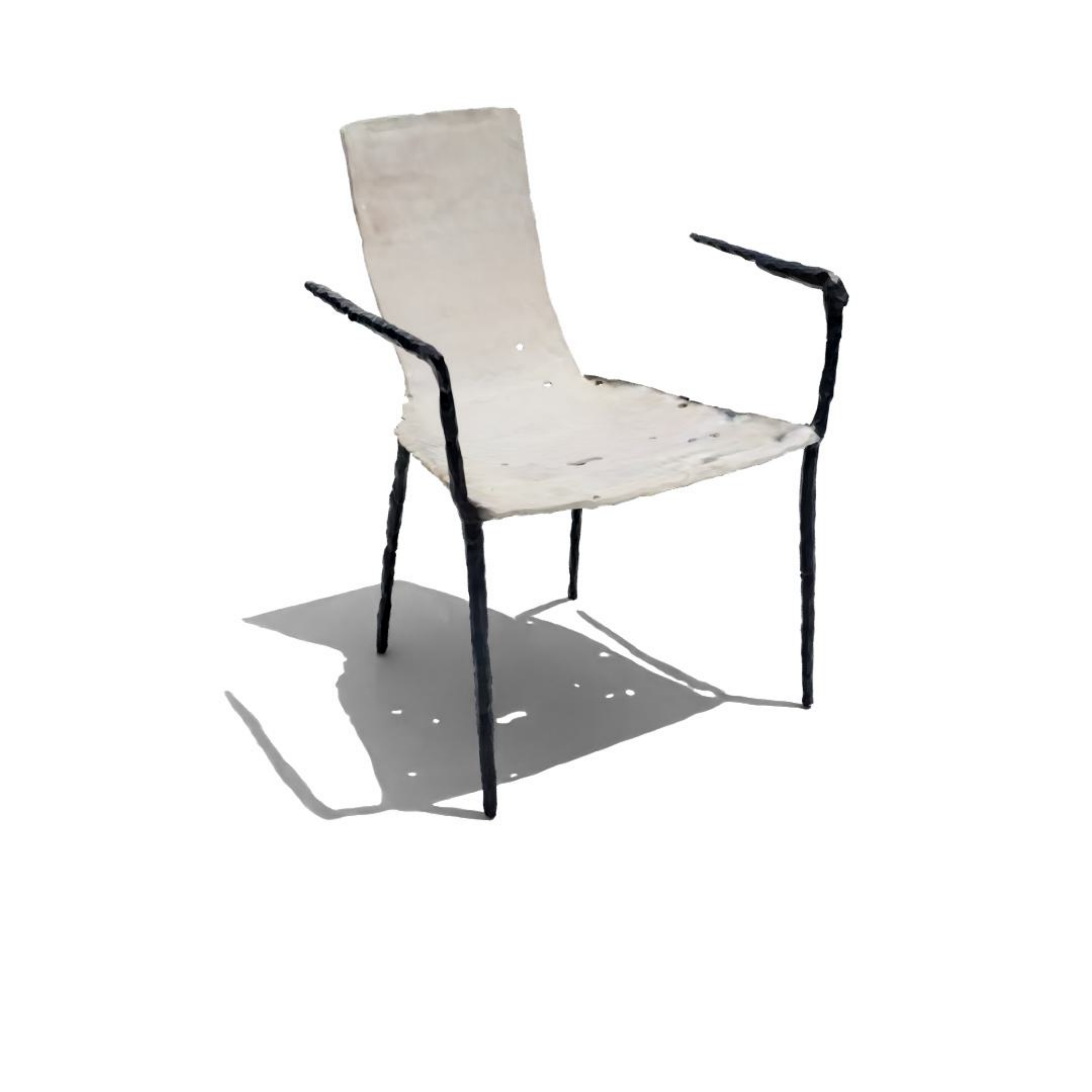}\includegraphics[width=0.16666666666666666\linewidth]{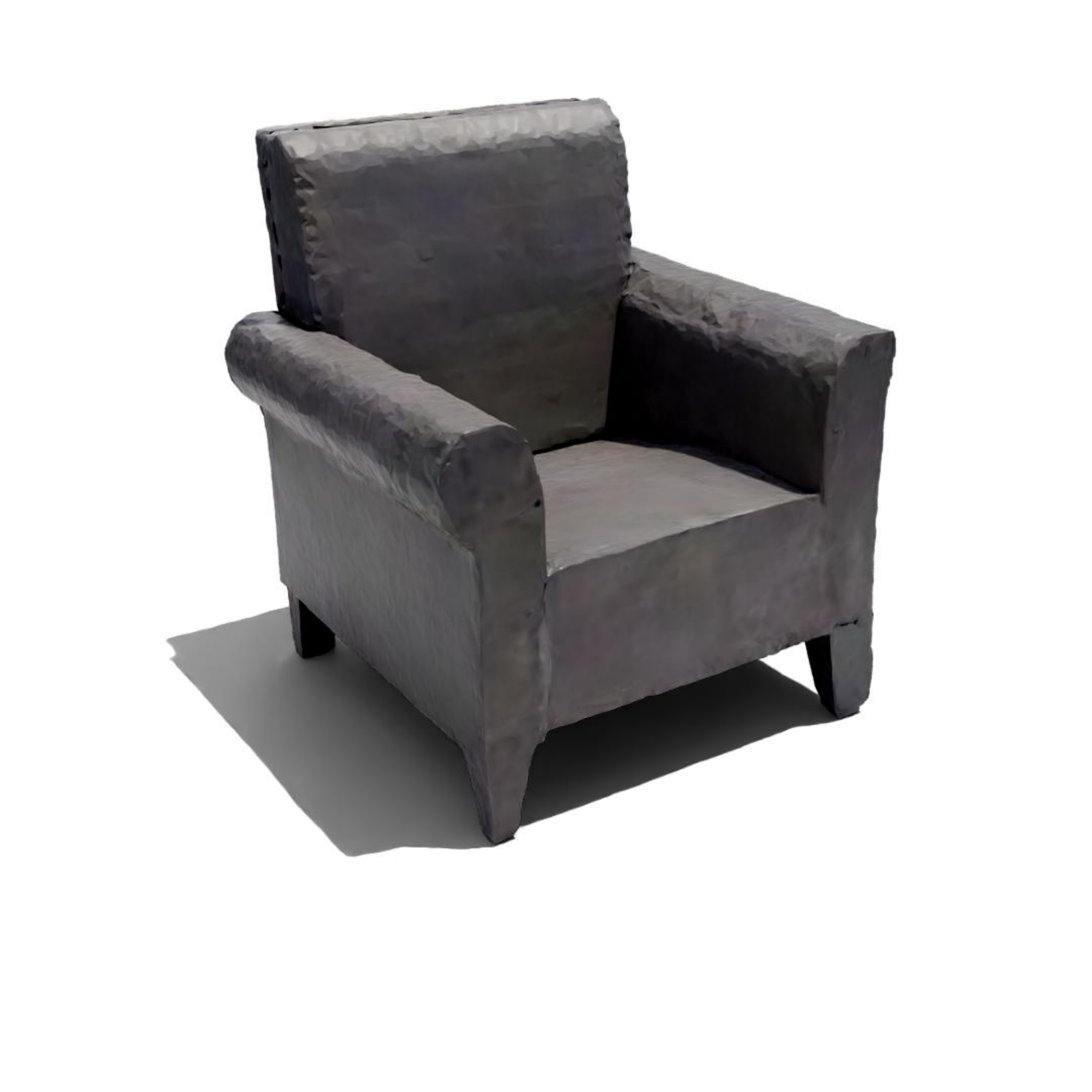}\includegraphics[width=0.16666666666666666\linewidth]{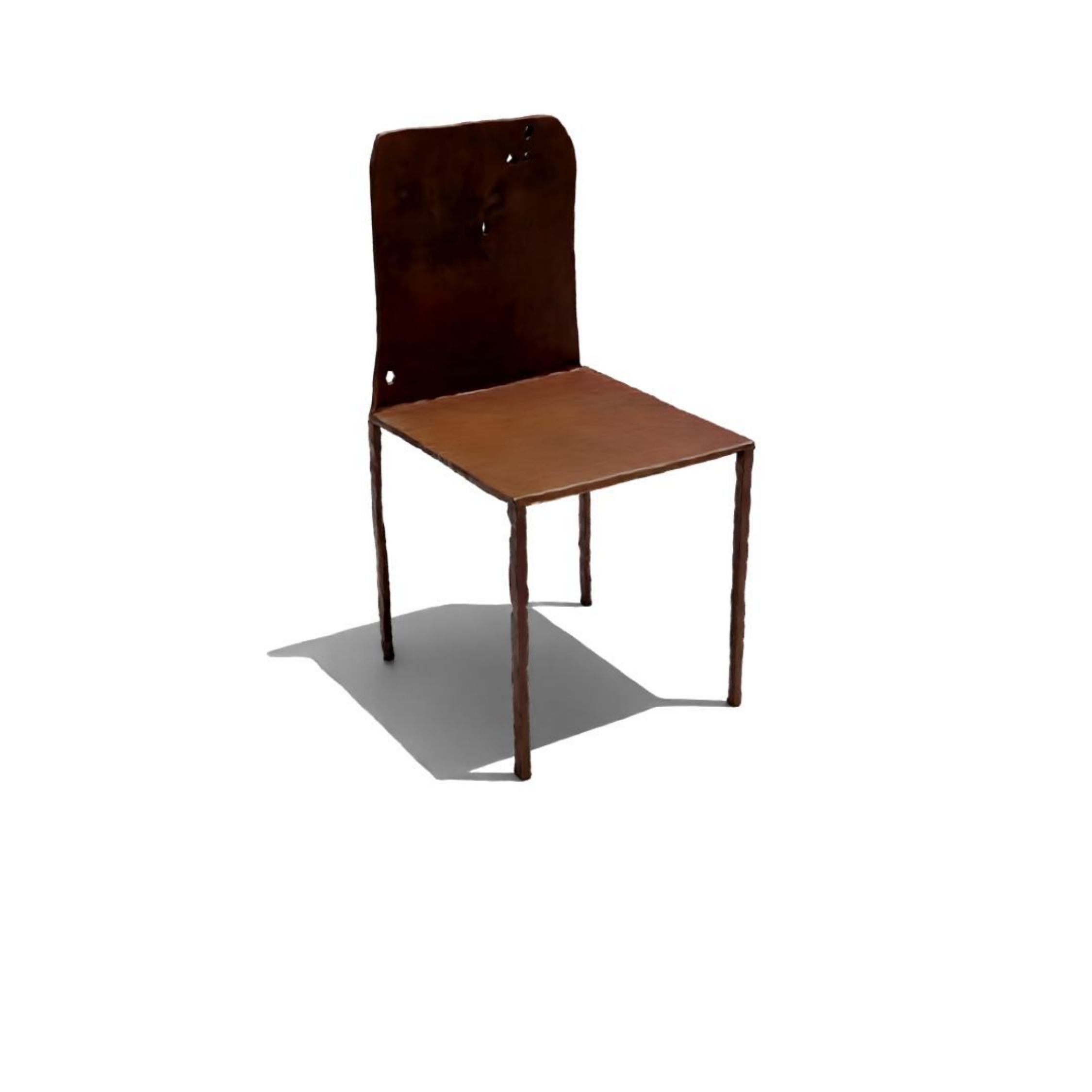}\\

\vspace{-0.3cm}
\includegraphics[width=0.16666666666666666\linewidth]{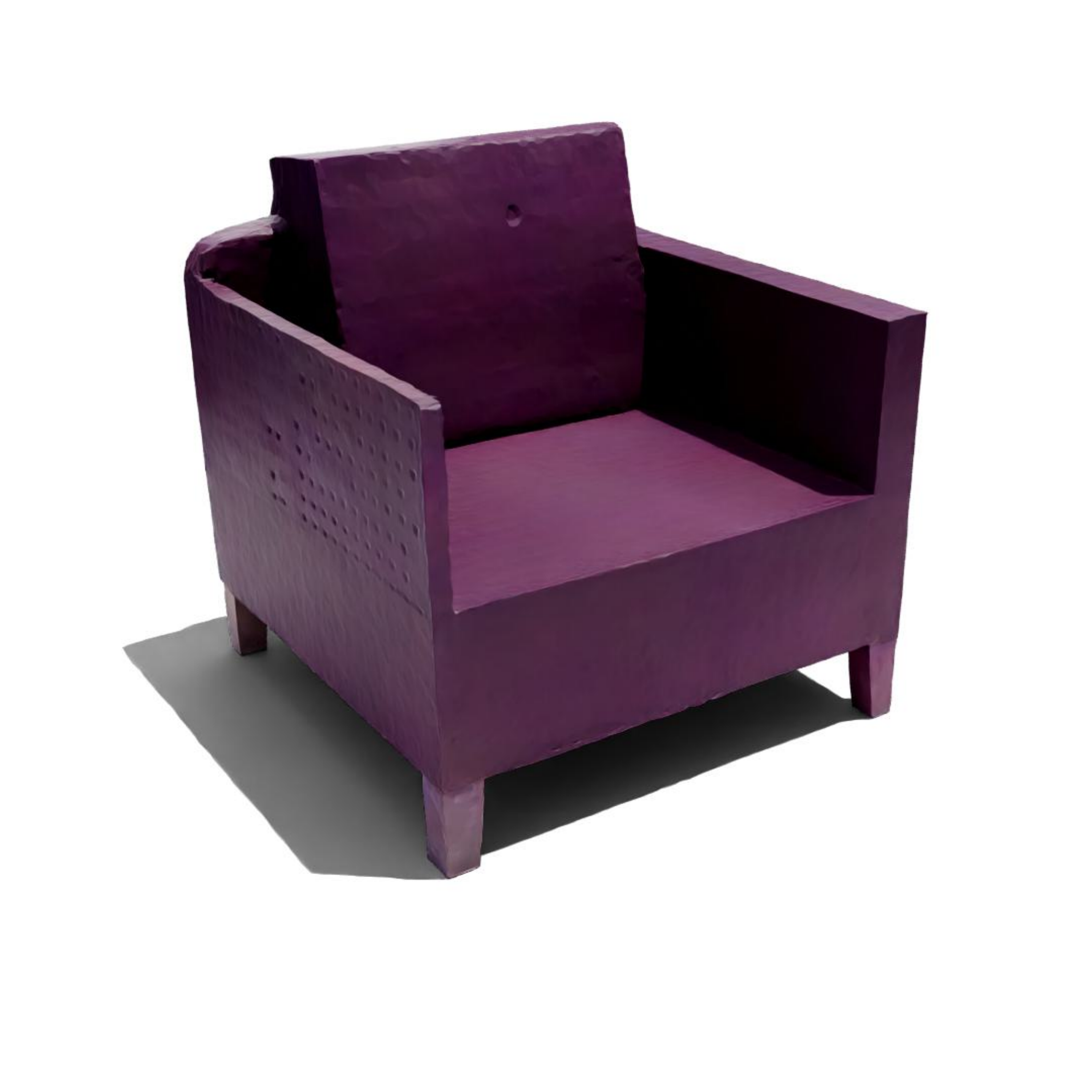}\includegraphics[width=0.16666666666666666\linewidth]{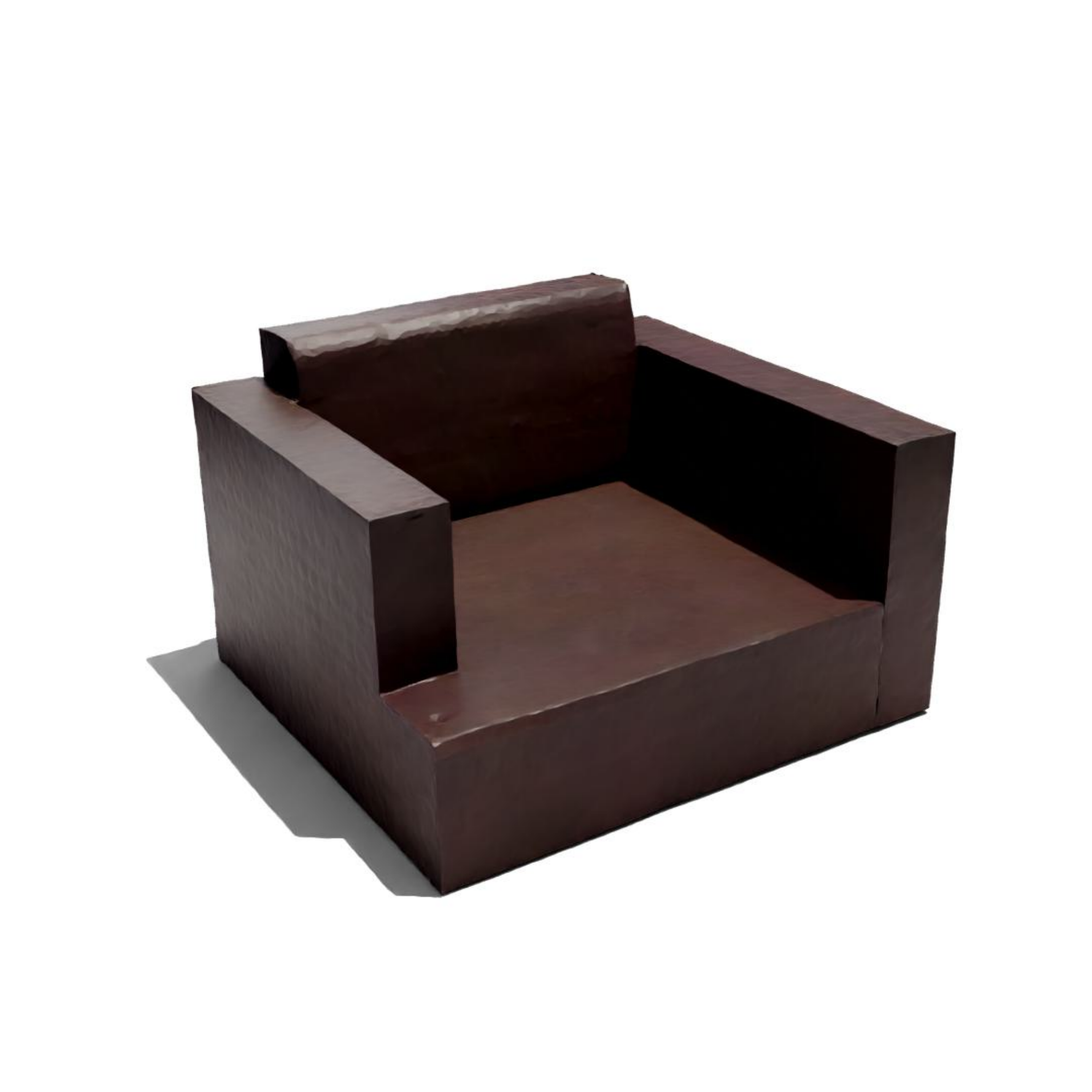}\includegraphics[width=0.16666666666666666\linewidth]{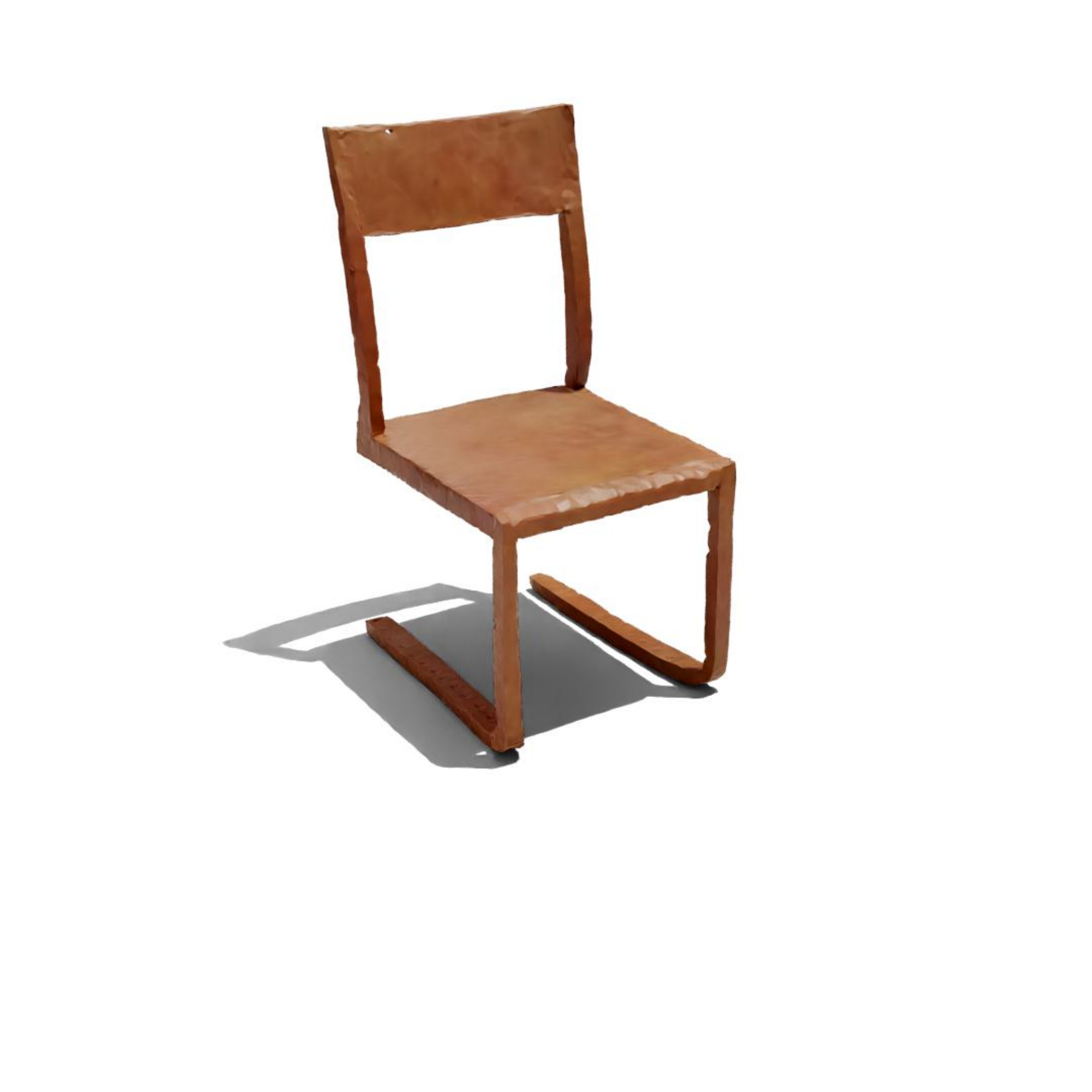}\includegraphics[width=0.16666666666666666\linewidth]{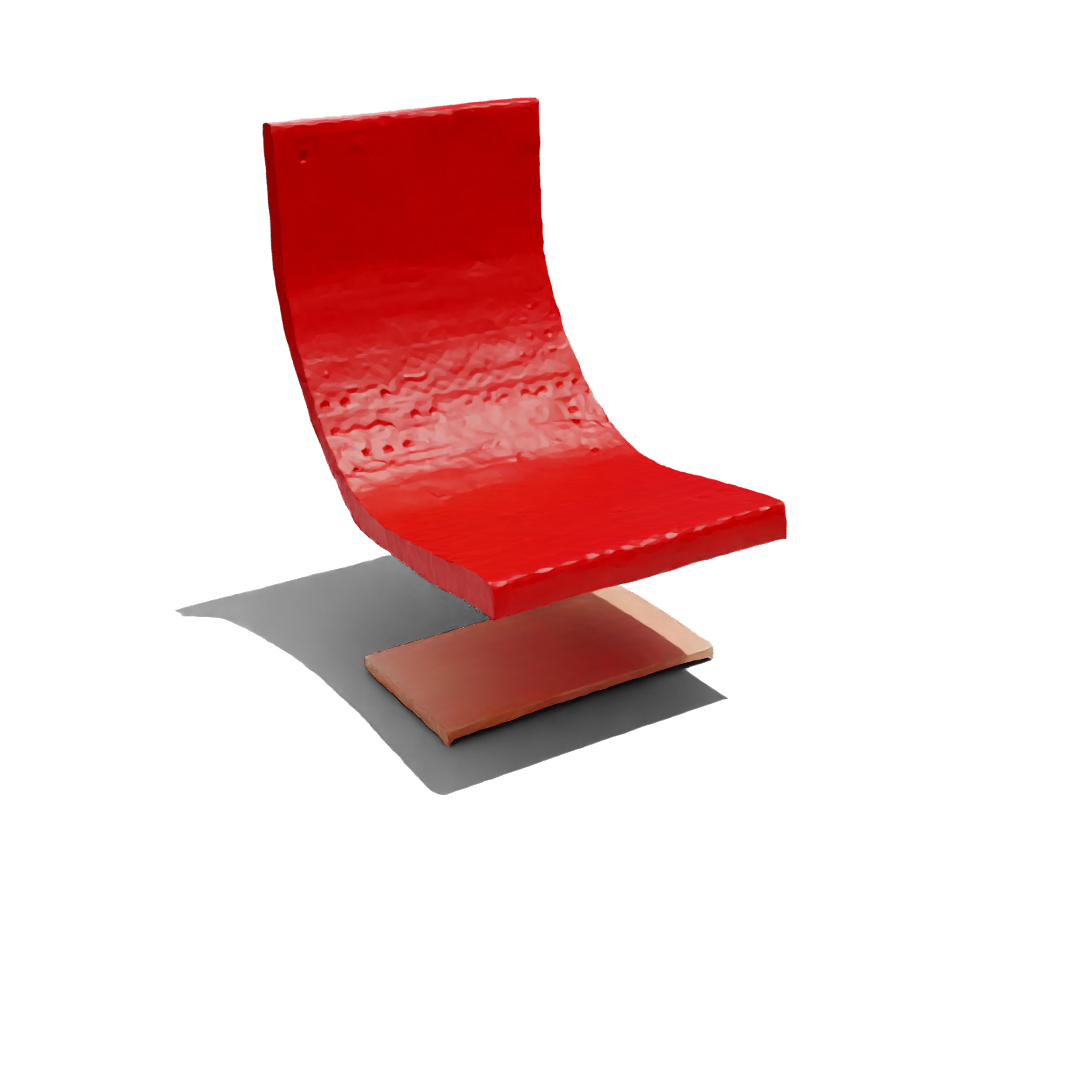}\includegraphics[width=0.16666666666666666\linewidth]{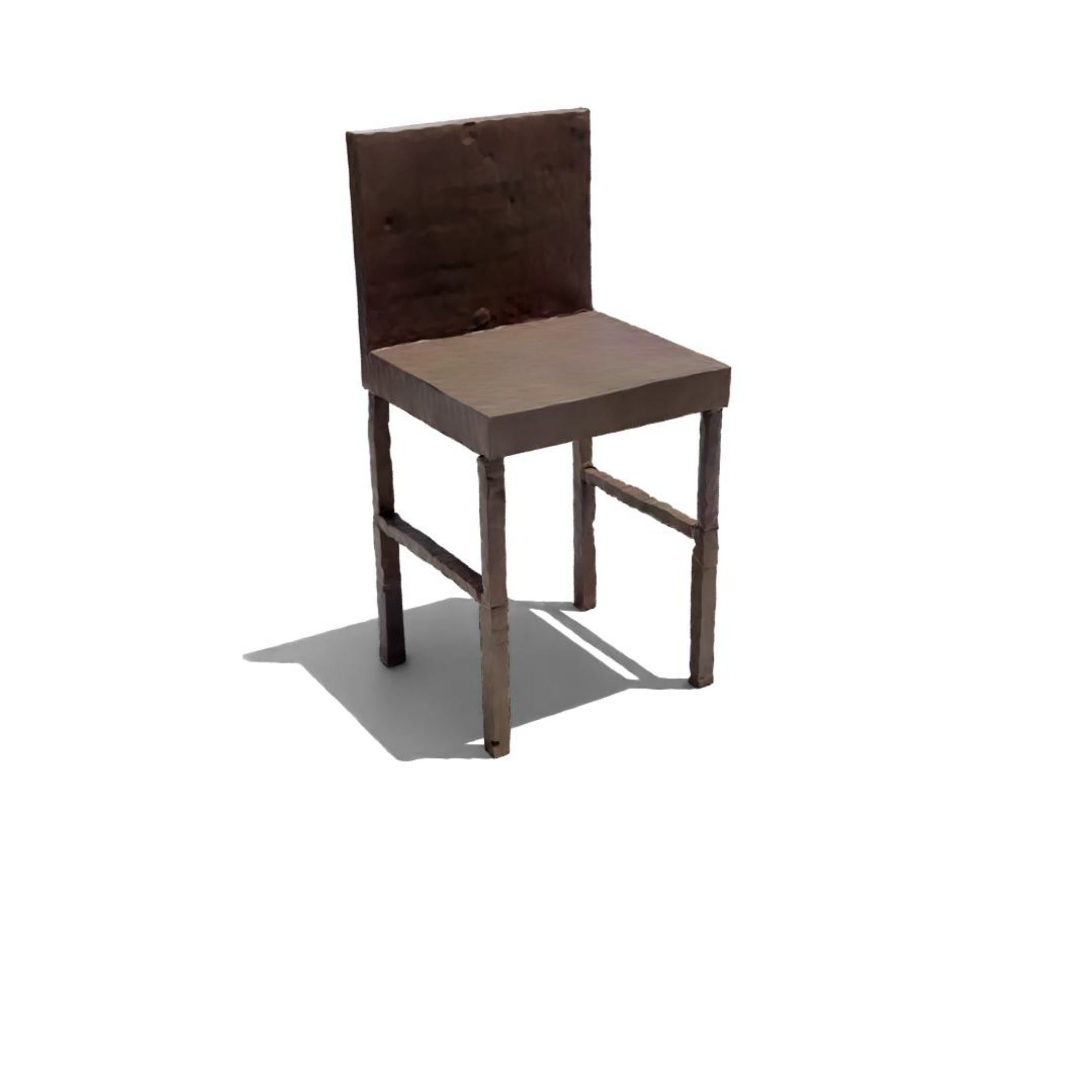}\includegraphics[width=0.16666666666666666\linewidth]{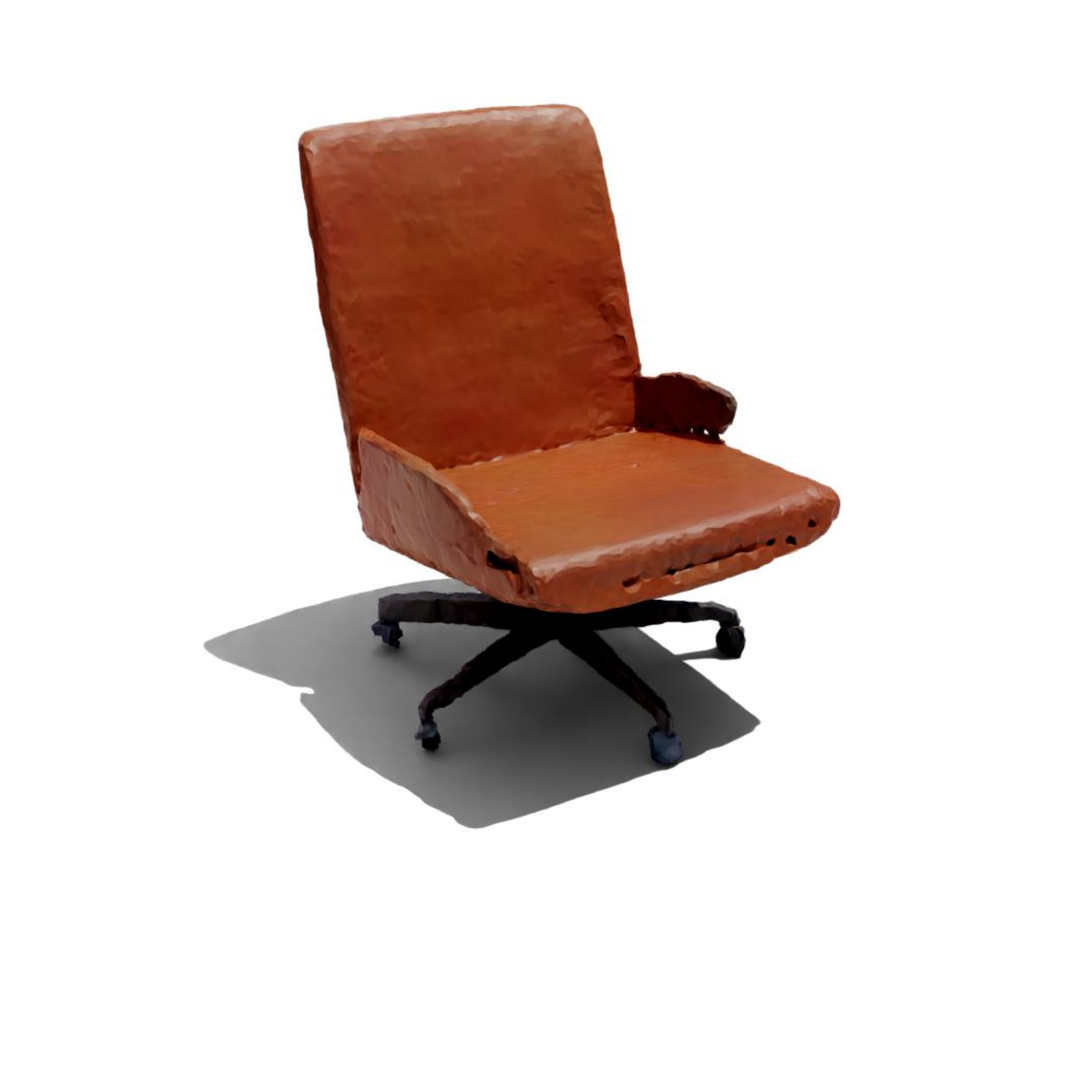}

\caption{\textbf{Random selection of chairs generated in standard resolution.}}\label{fig:uncond:chair:128}
\end{figure*}

Additional renderings of shapes generated with TetraDiffusion can be found in \cref{fig:uncond:airplane:128}, \cref{fig:uncond:bike:128}, \cref{fig:uncond:car:128}, \cref{fig:uncond:chair:128}, \cref{fig:uncond:airplane:192}, \cref{fig:uncond:bike:192} and \cref{fig:uncond:car:192}. TetraDiffusion consistently outputs high-quality results with substantial diversity. Notably, even in low data regimes such as the ShapeNet motorbike class, our model learns a meaningful manifold and reliably interpolates novel shapes.

\vspace{-0.7cm}

\clearpage
\newpage

\vspace{-0.1cm}

\clearpage
\newpage

\clearpage
\newpage

\clearpage
\newpage

\subsection{Interpolation}

Here, we perform shape interpolation, a common task in generative models. We smoothly transform one shape to another by interpolating along the noise vectors. Let the noise vector of shape $i$ be $z_{i,t}$ at time step $t$. In 2D diffusion it is common to linearly interpolate between two noise vectors $z_{1,t}$ and $z_{2,t}$. However, similar to \cite{zeng2022lion}, we find that a simple convex combination of the noise vectors leads to poor results in 3D, \eg holes, deformations or corrupted colors. The authors of \cite{zeng2022lion} provide an intuitive explanation: due to the high dimensionality of the input space, the noise samples almost certainly lie on a thin spherical shell according to the Gaussian annulus theorem. Therefore, they propose to use a square root-based interpolation, in order to stay within the training set learned by the model.  

Another, simpler reason is that the linear interpolation of two standard Gaussians is not a standard Gaussian anymore. For $i=1,2$ let $X_i \sim \mathcal{N}(0,I)$ be independent Gaussian random variables and $\lambda_1 = 1 - \lambda_2$, $\lambda_i \in [0, 1]$. Then $X \coloneqq \lambda_1 X_1 + \lambda_2 X_2 \sim \mathcal{N}(0, \sigma^2 \coloneqq \lambda_1^2 + \lambda_2^2)$, \ie in particular $\sigma^2 = 1 \iff \lambda_i = 1$ (see \cite{wegner2021lecture} For a more general proof). In particular, we end up with a non-isotropic Gaussian between the end values. To obtain isotropic Gaussian samples in between, one needs to take the square root of the interpolation weights, as proposed by \cite{zeng2022lion}.

\begin{figure*}[!ht]
\centering
\includegraphics[width=0.08333333333333333\linewidth]{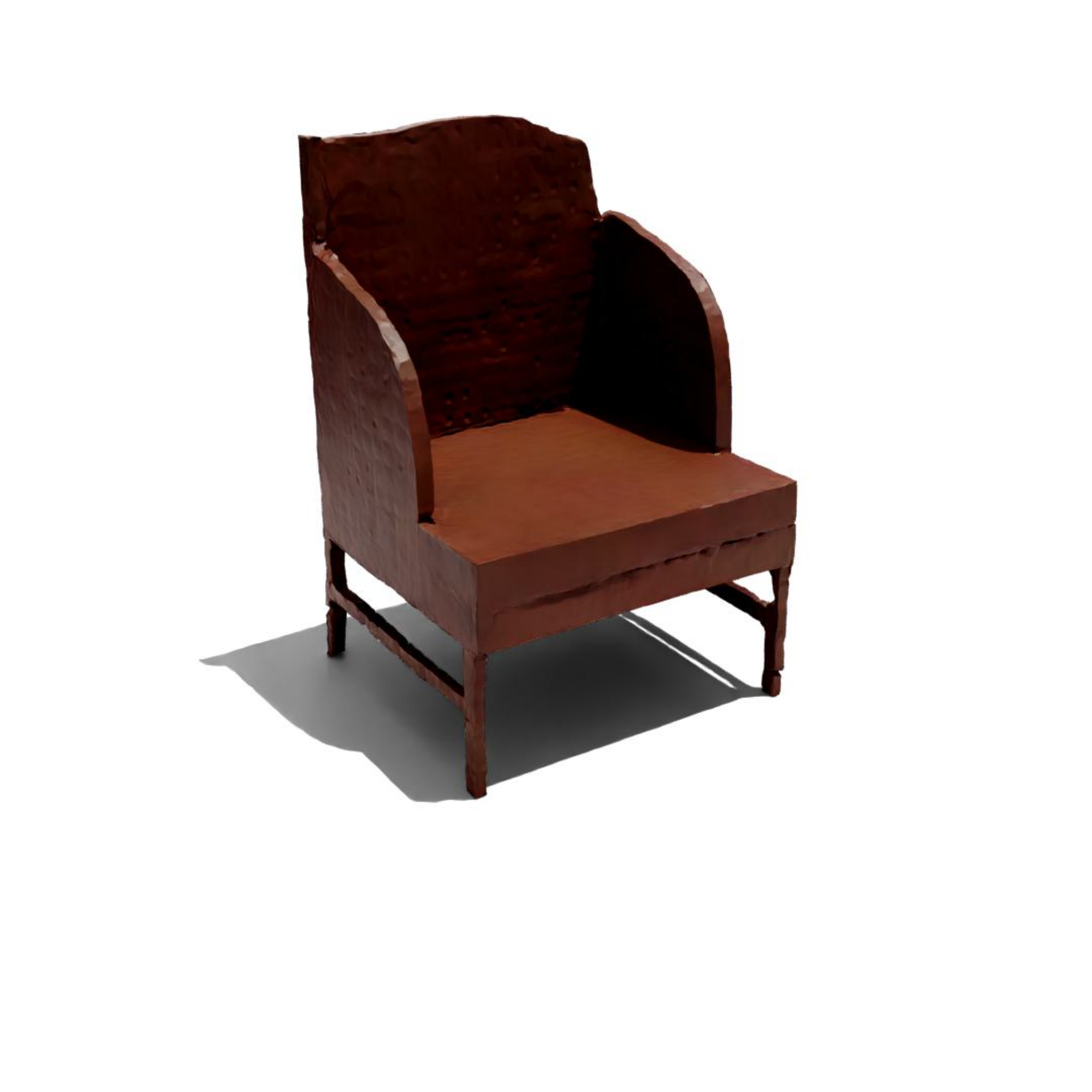}\includegraphics[width=0.08333333333333333\linewidth]{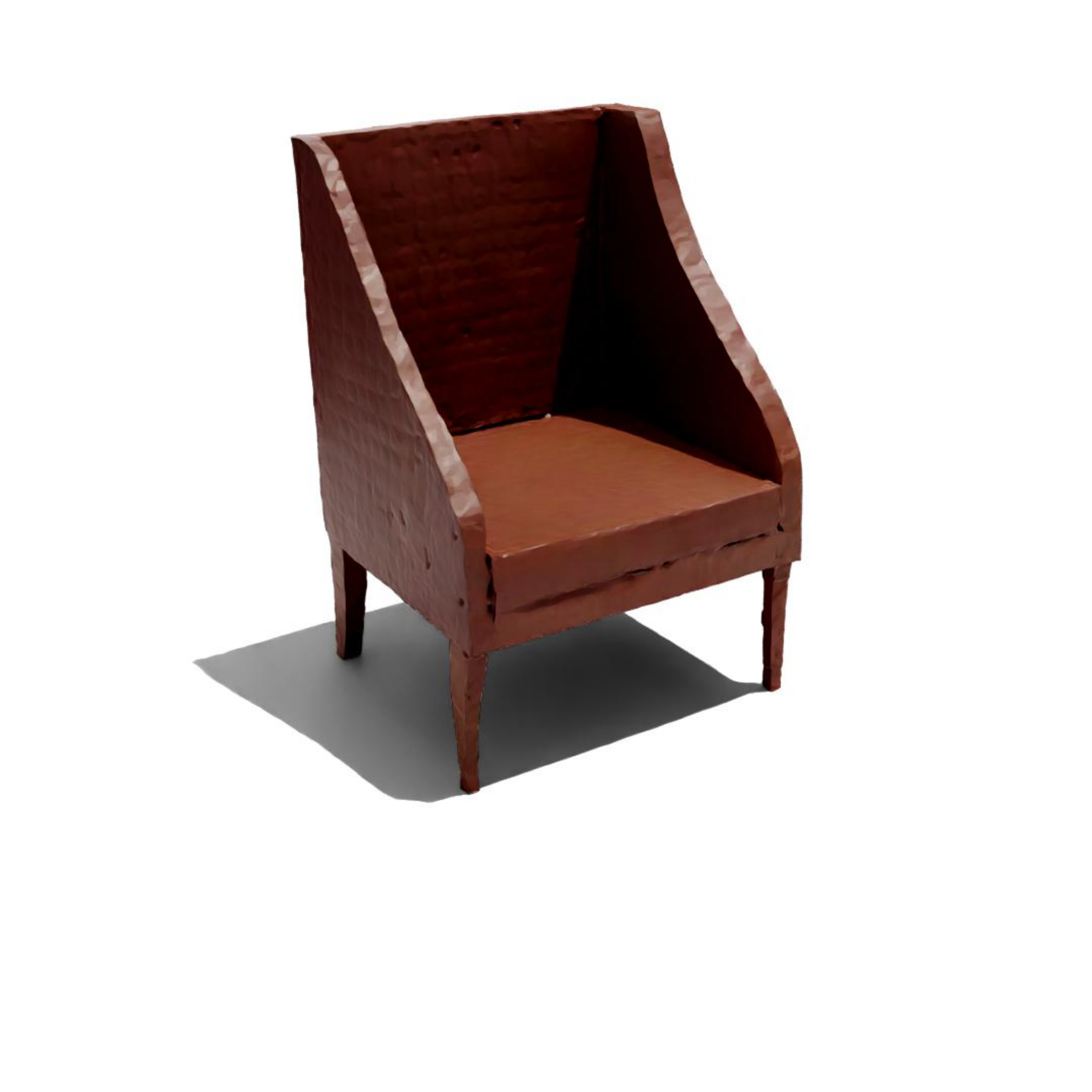}\includegraphics[width=0.08333333333333333\linewidth]{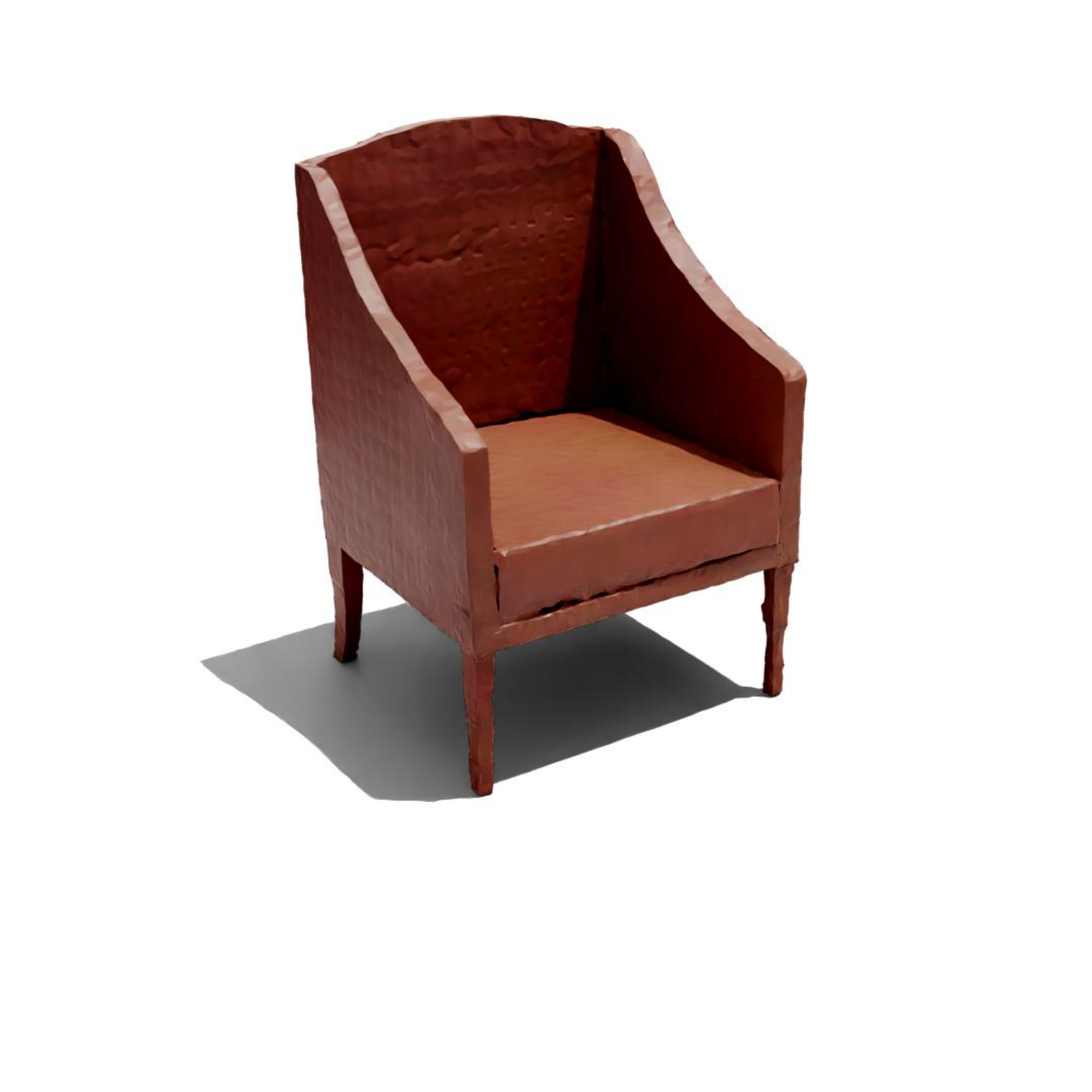}\includegraphics[width=0.08333333333333333\linewidth]{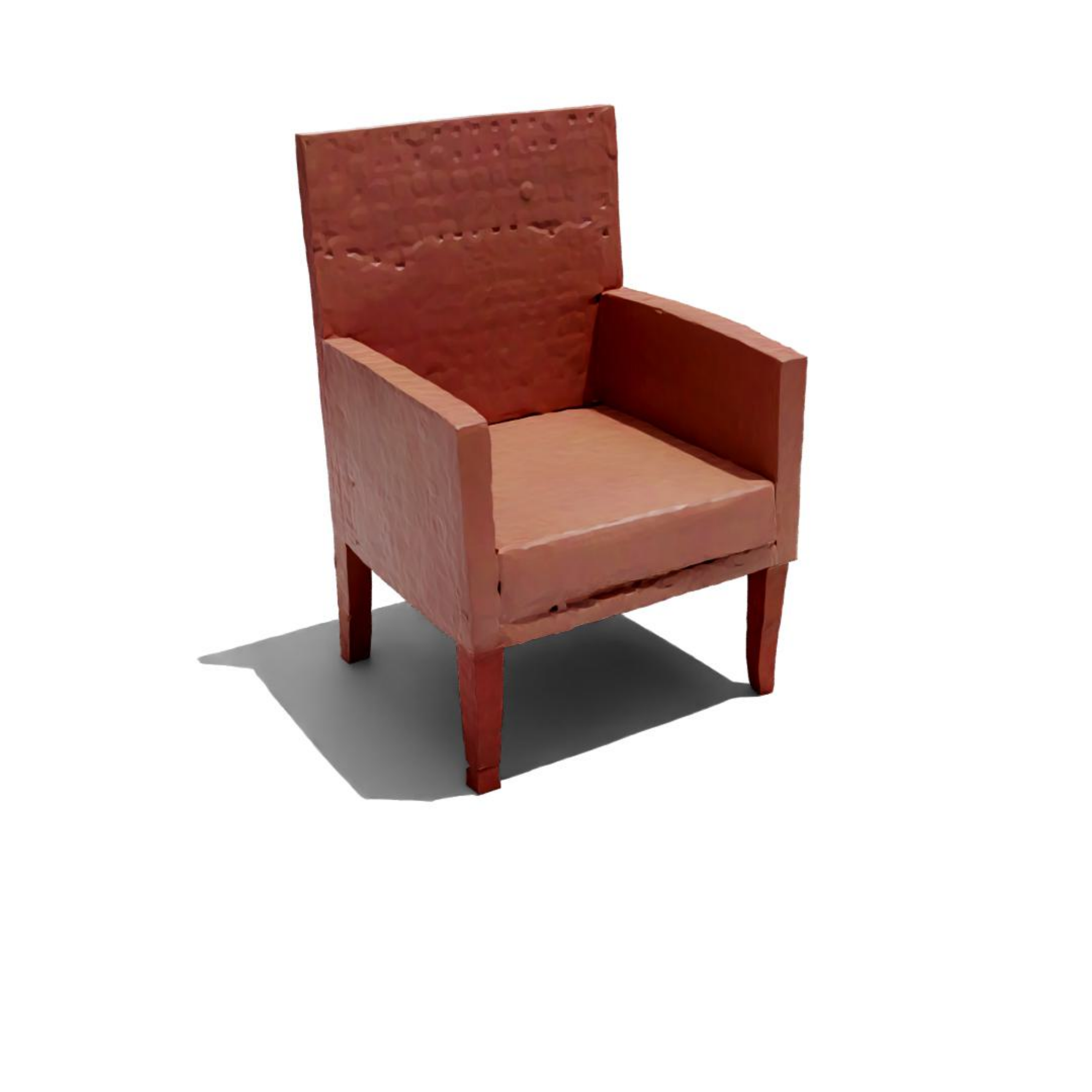}\includegraphics[width=0.08333333333333333\linewidth]{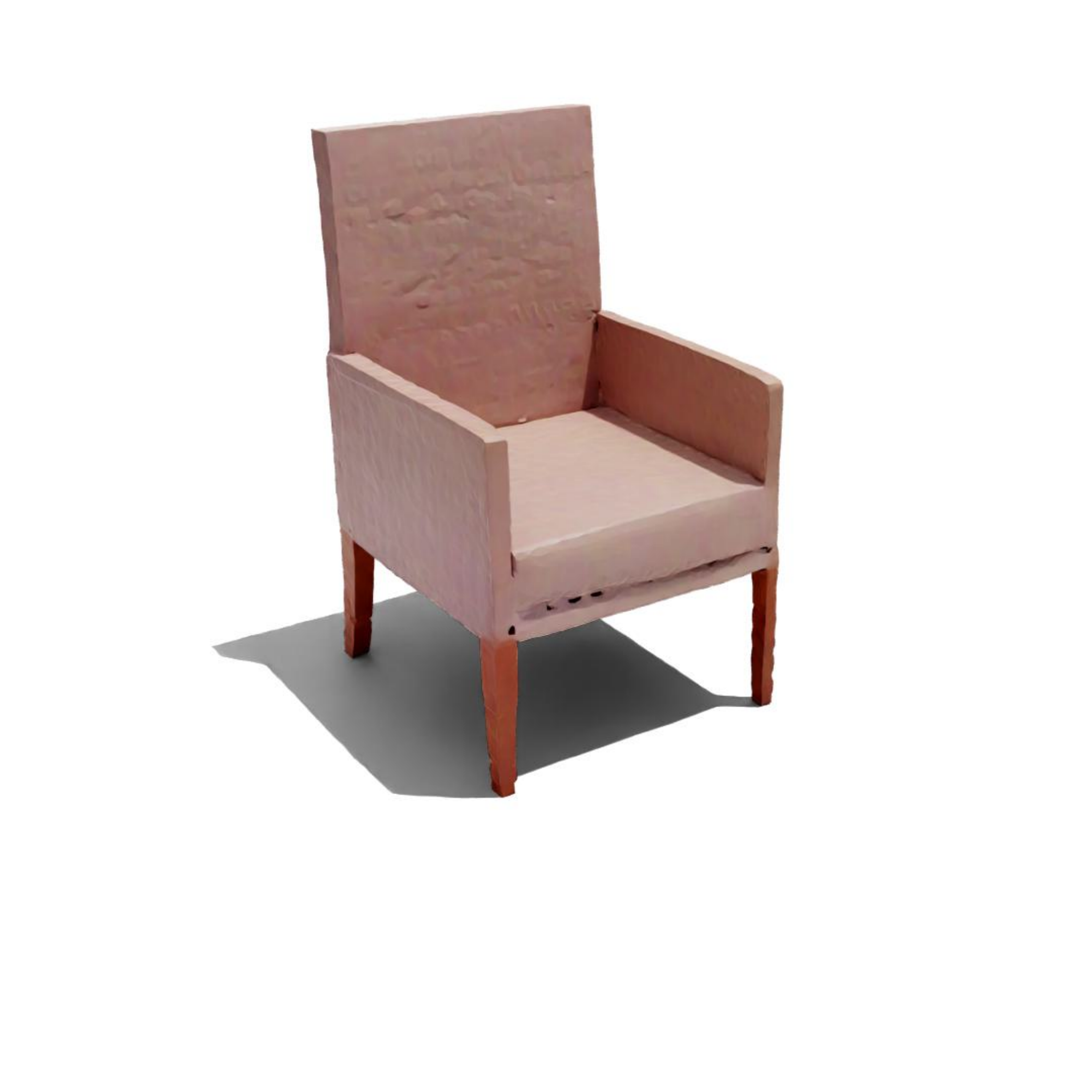}\includegraphics[width=0.08333333333333333\linewidth]{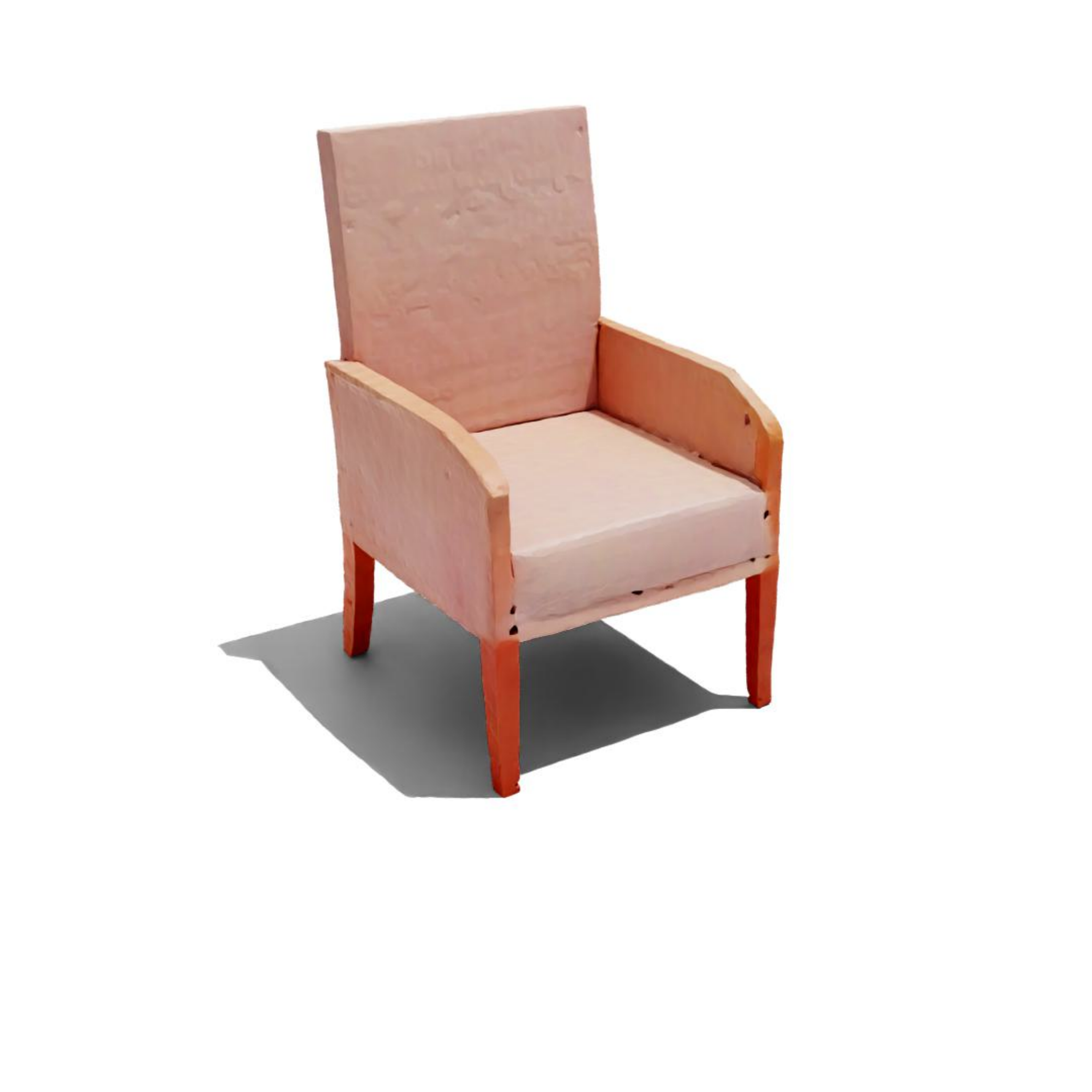}\includegraphics[width=0.08333333333333333\linewidth]{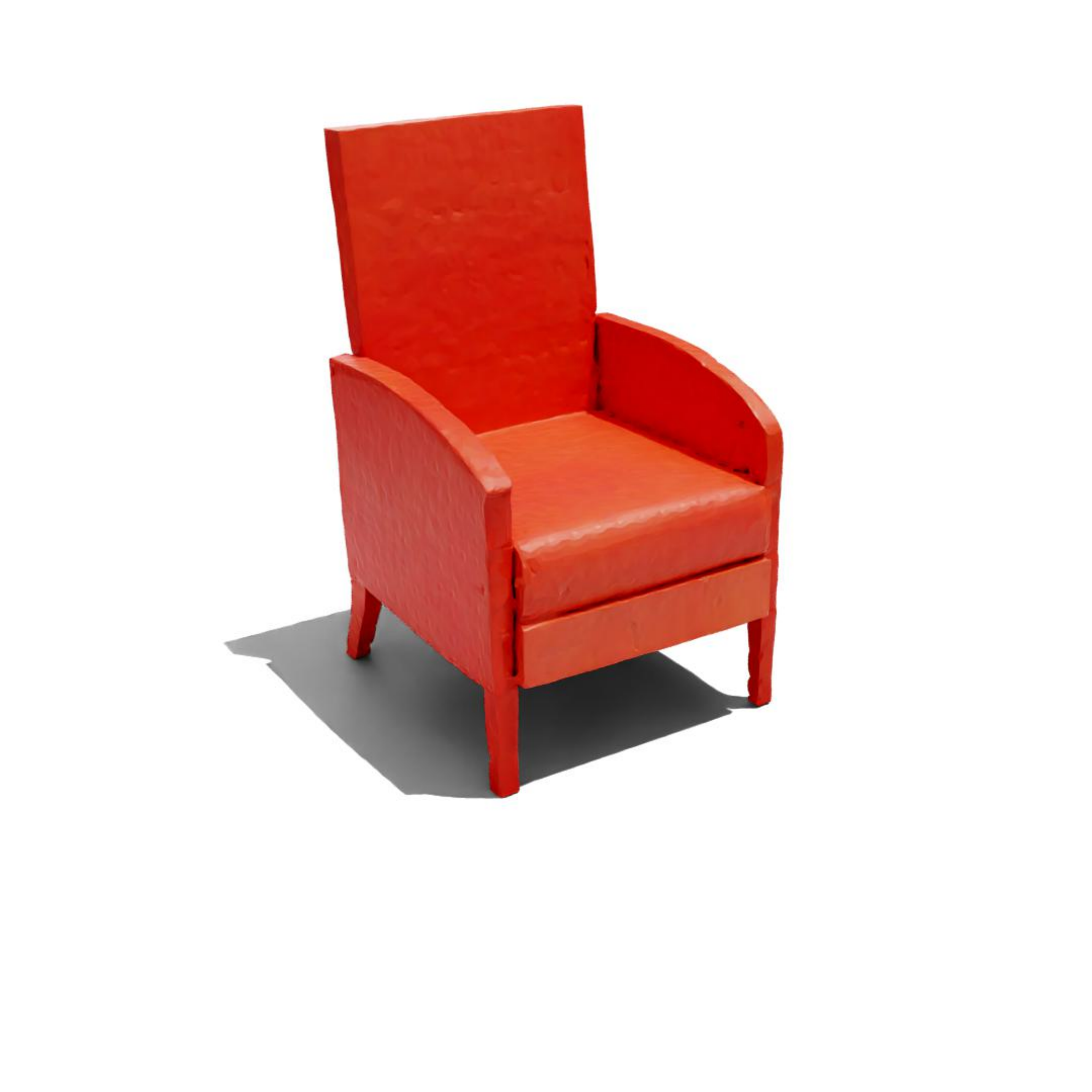}\includegraphics[width=0.08333333333333333\linewidth]{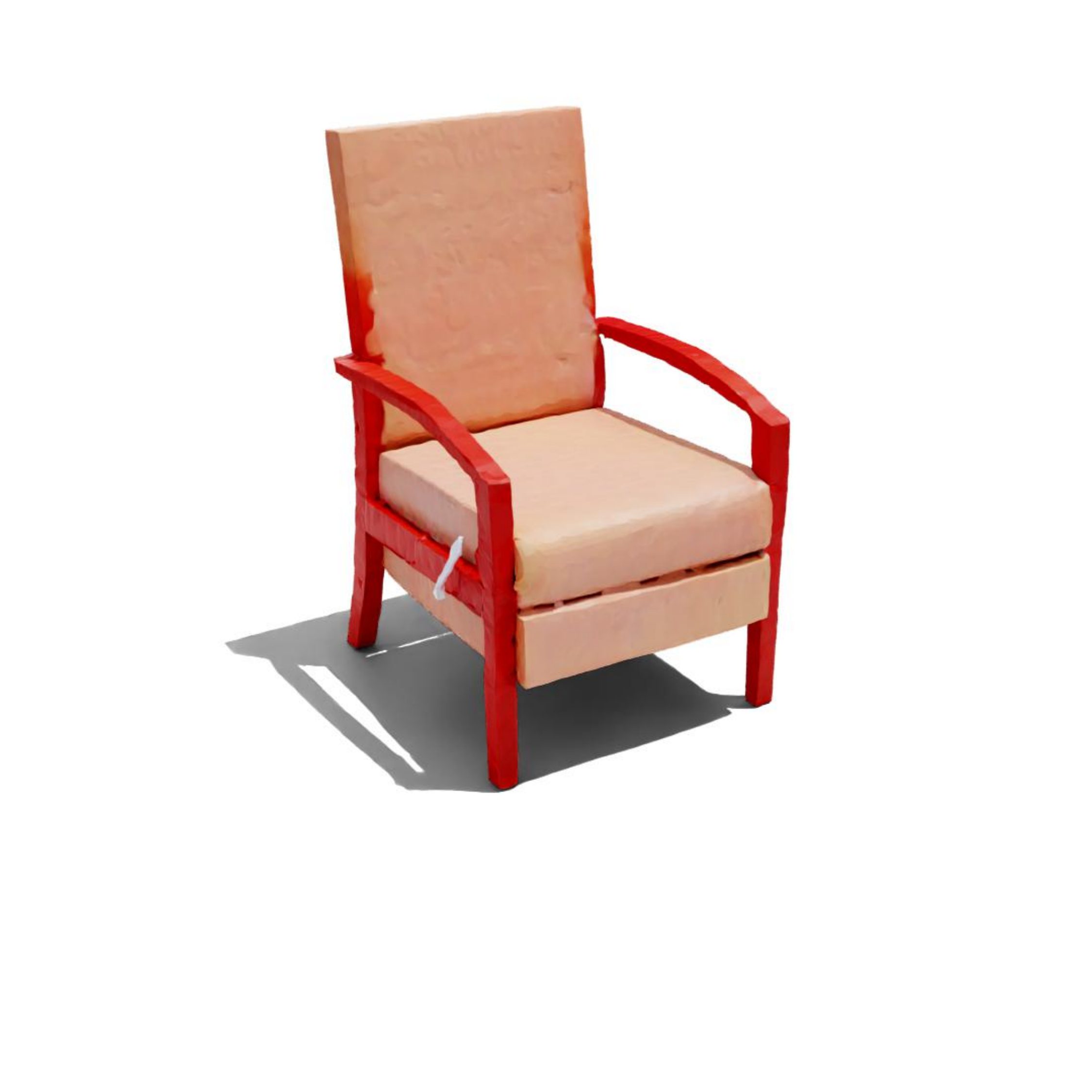}\includegraphics[width=0.08333333333333333\linewidth]{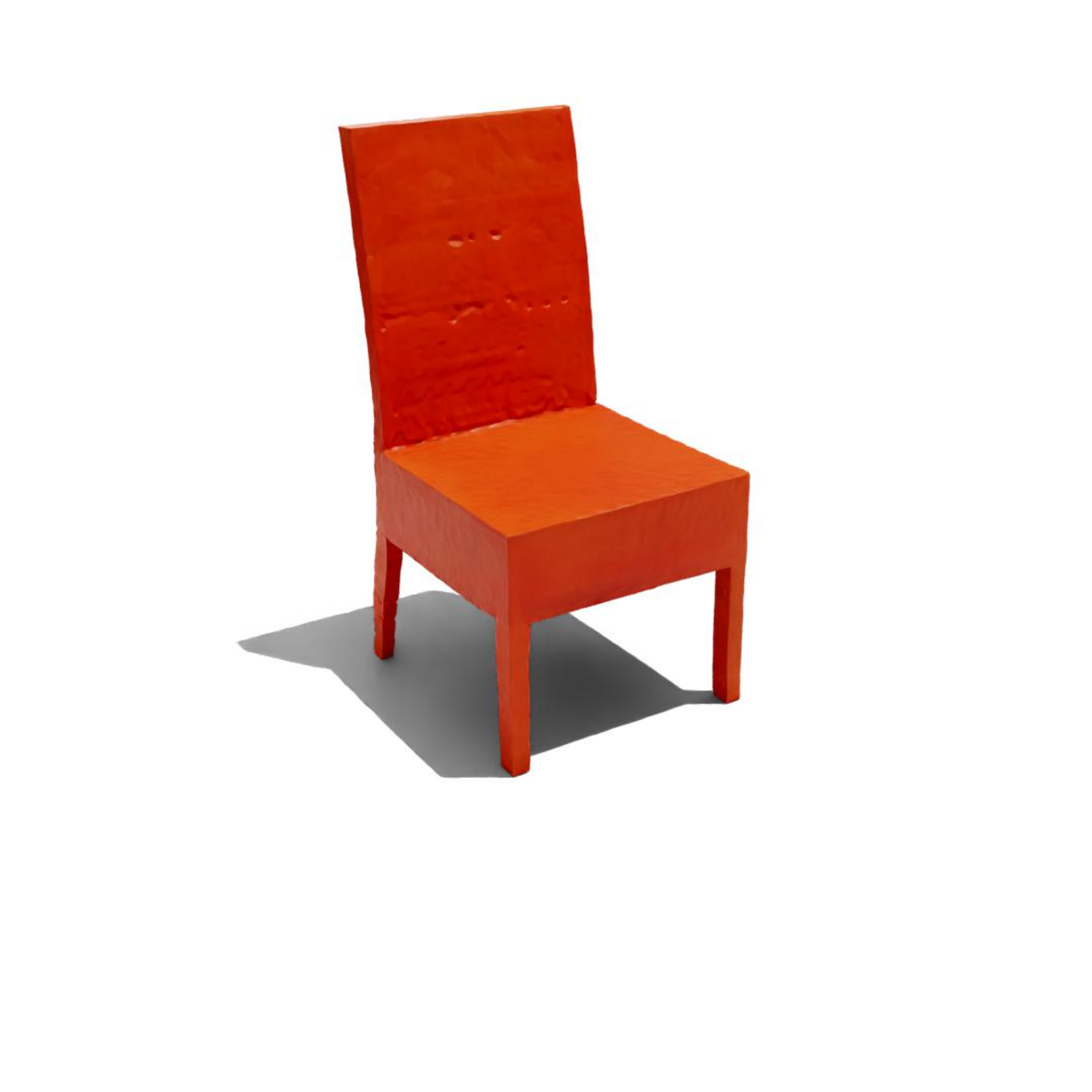}\includegraphics[width=0.08333333333333333\linewidth]{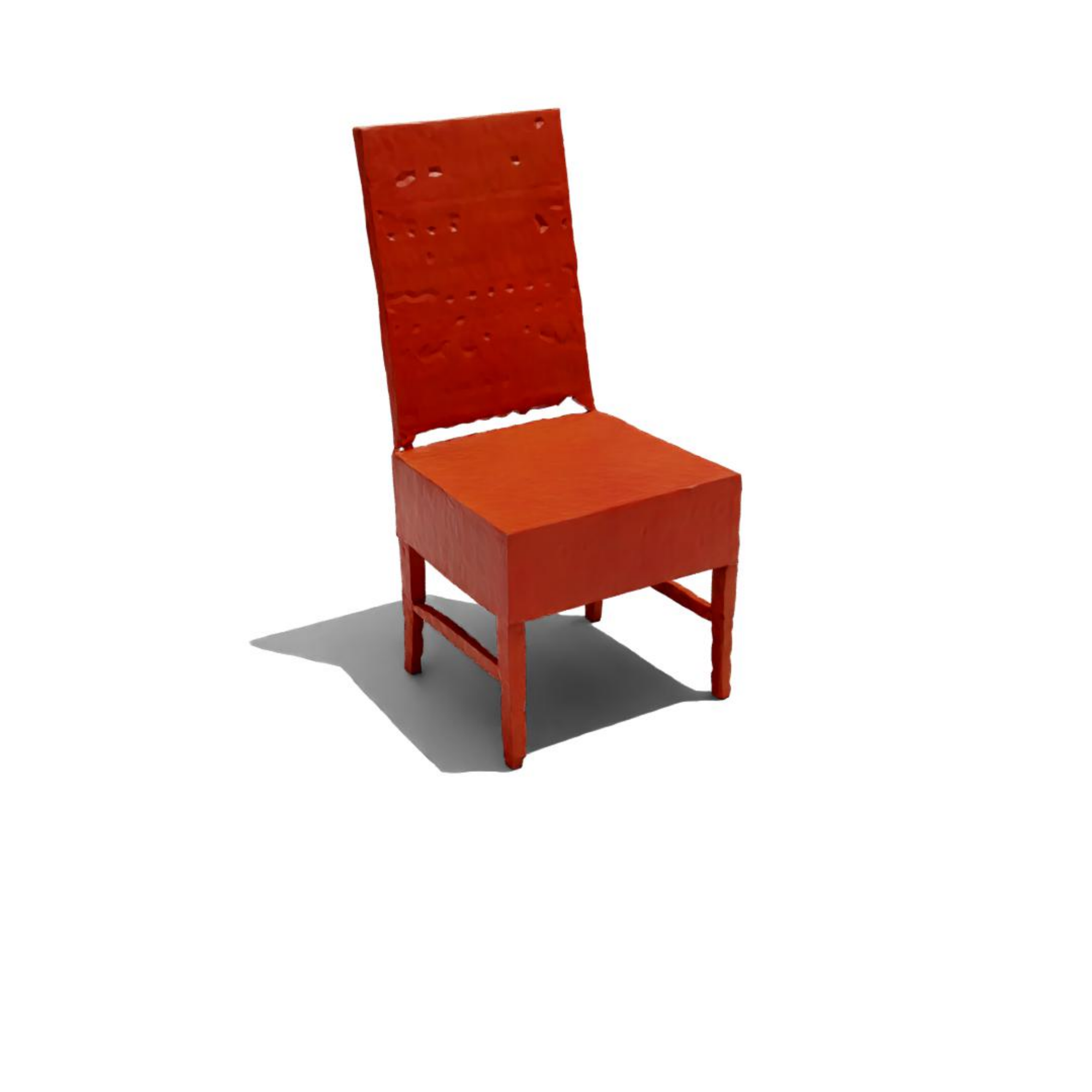}\includegraphics[width=0.08333333333333333\linewidth]{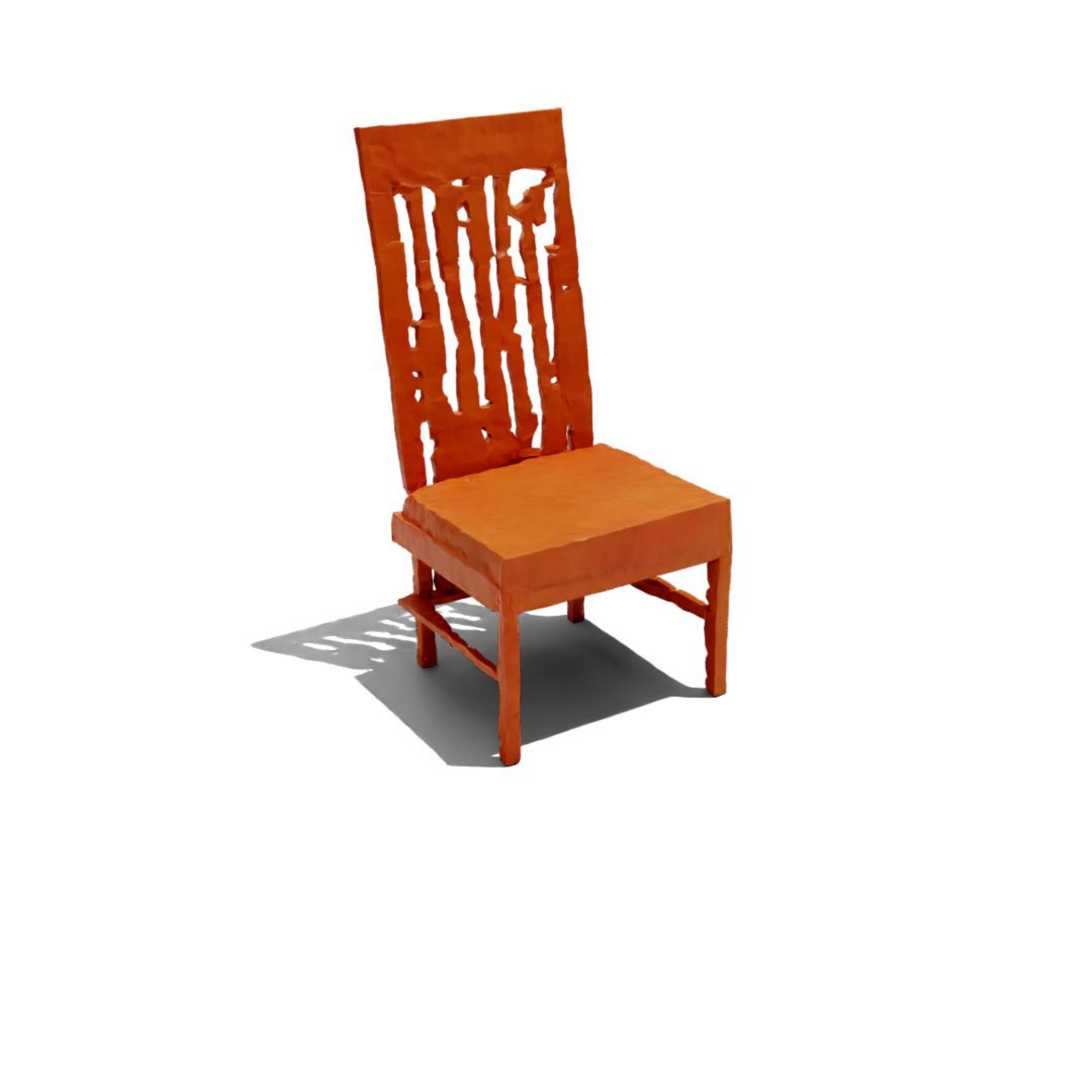}\includegraphics[width=0.08333333333333333\linewidth]{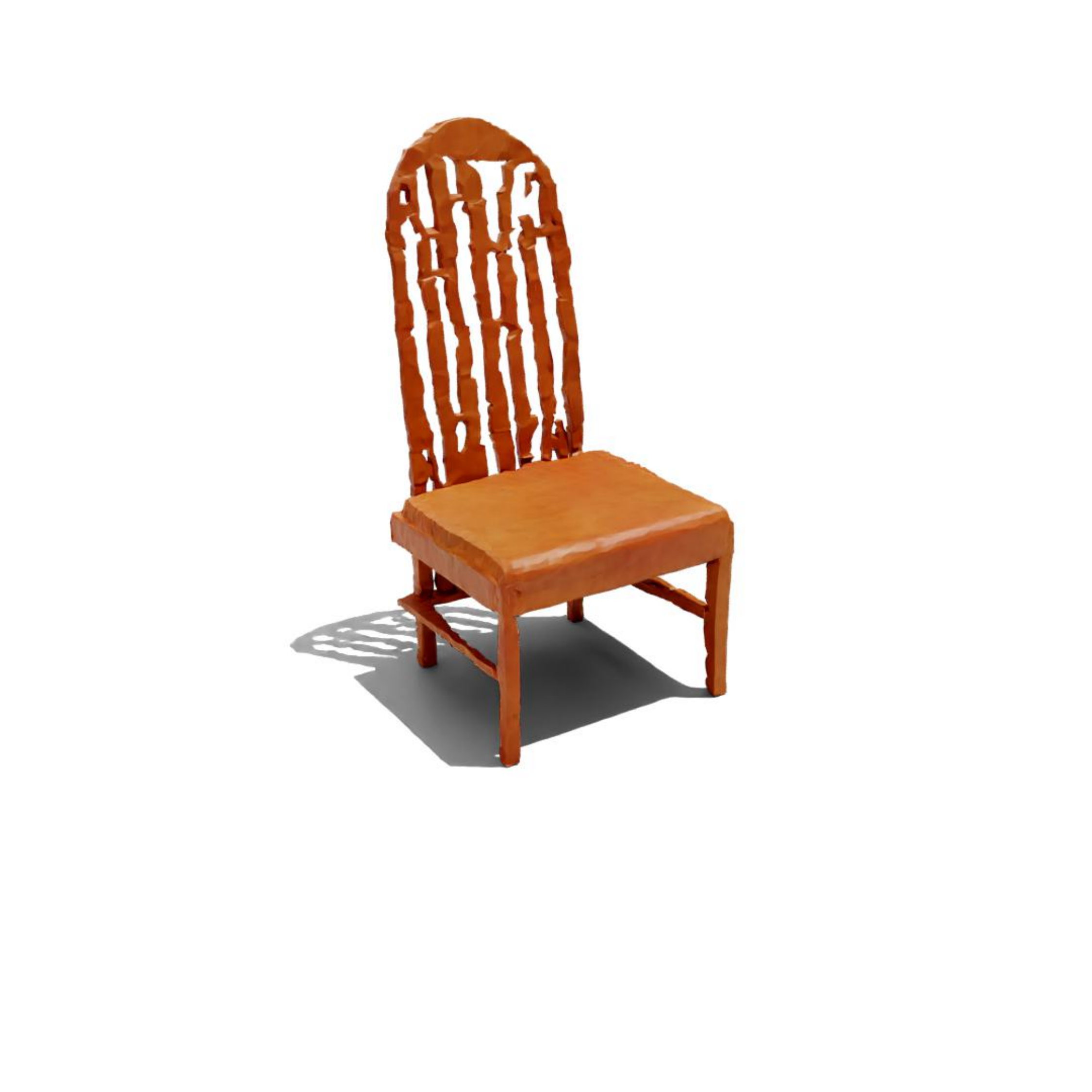}\\
\includegraphics[width=0.08333333333333333\linewidth]{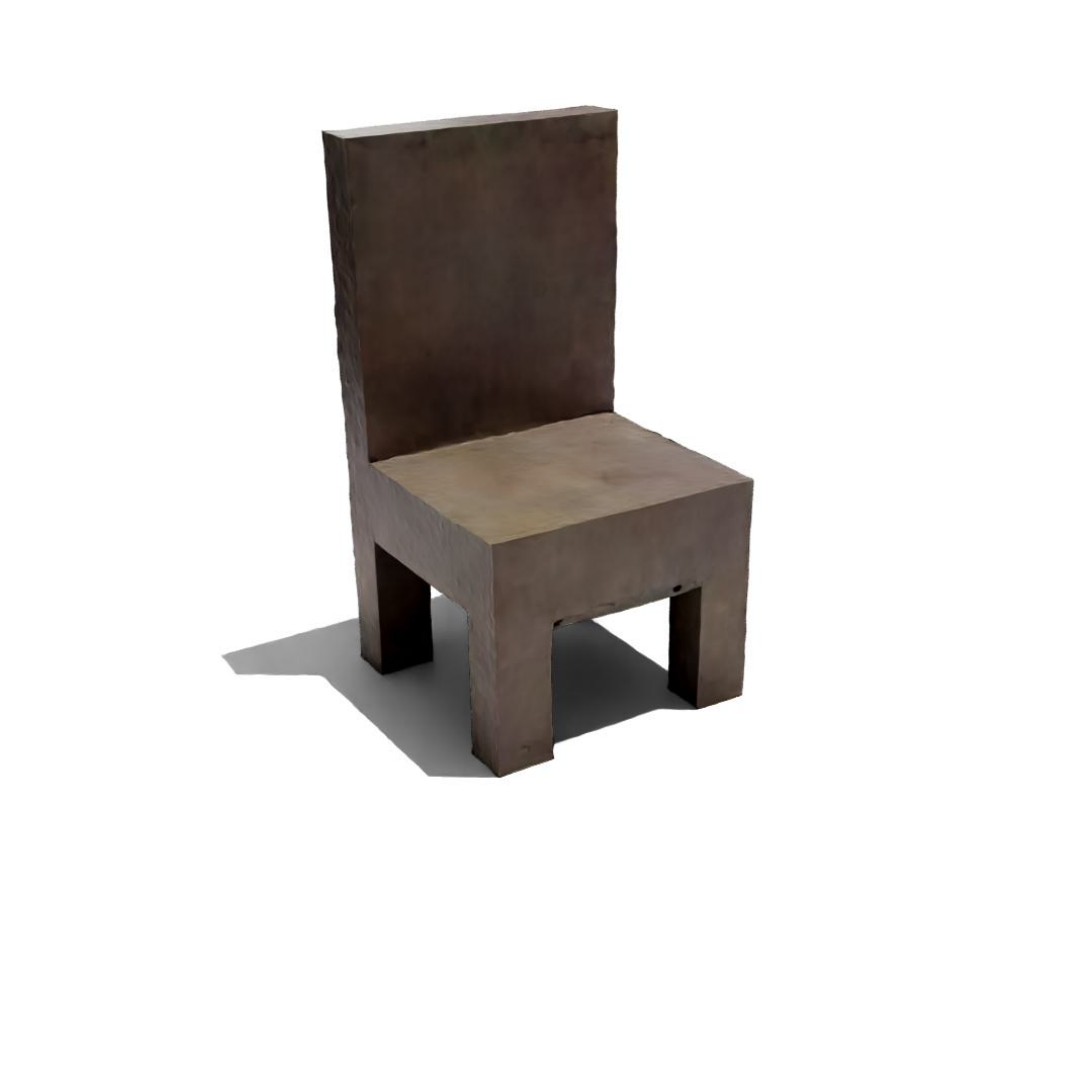}\includegraphics[width=0.08333333333333333\linewidth]{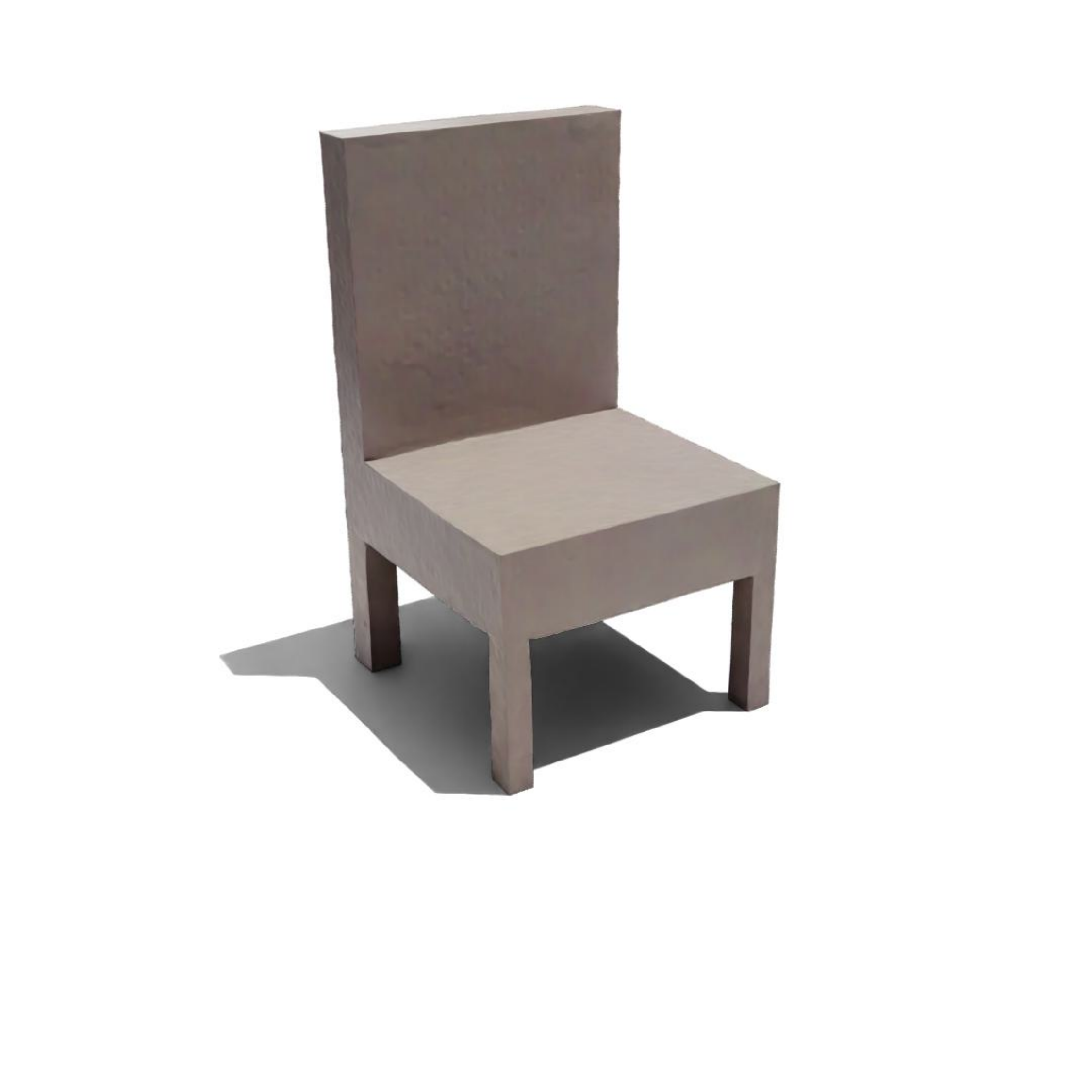}\includegraphics[width=0.08333333333333333\linewidth]{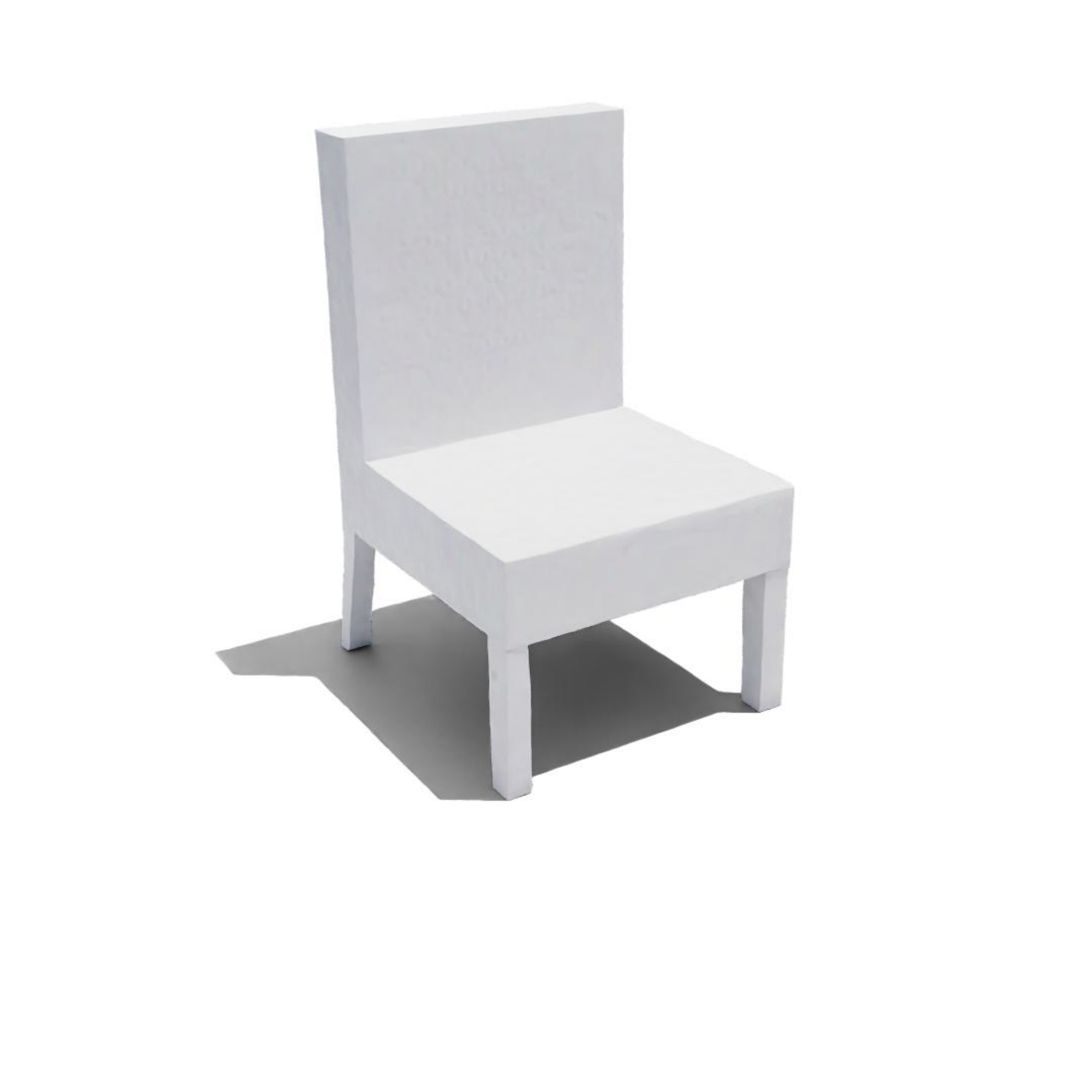}\includegraphics[width=0.08333333333333333\linewidth]{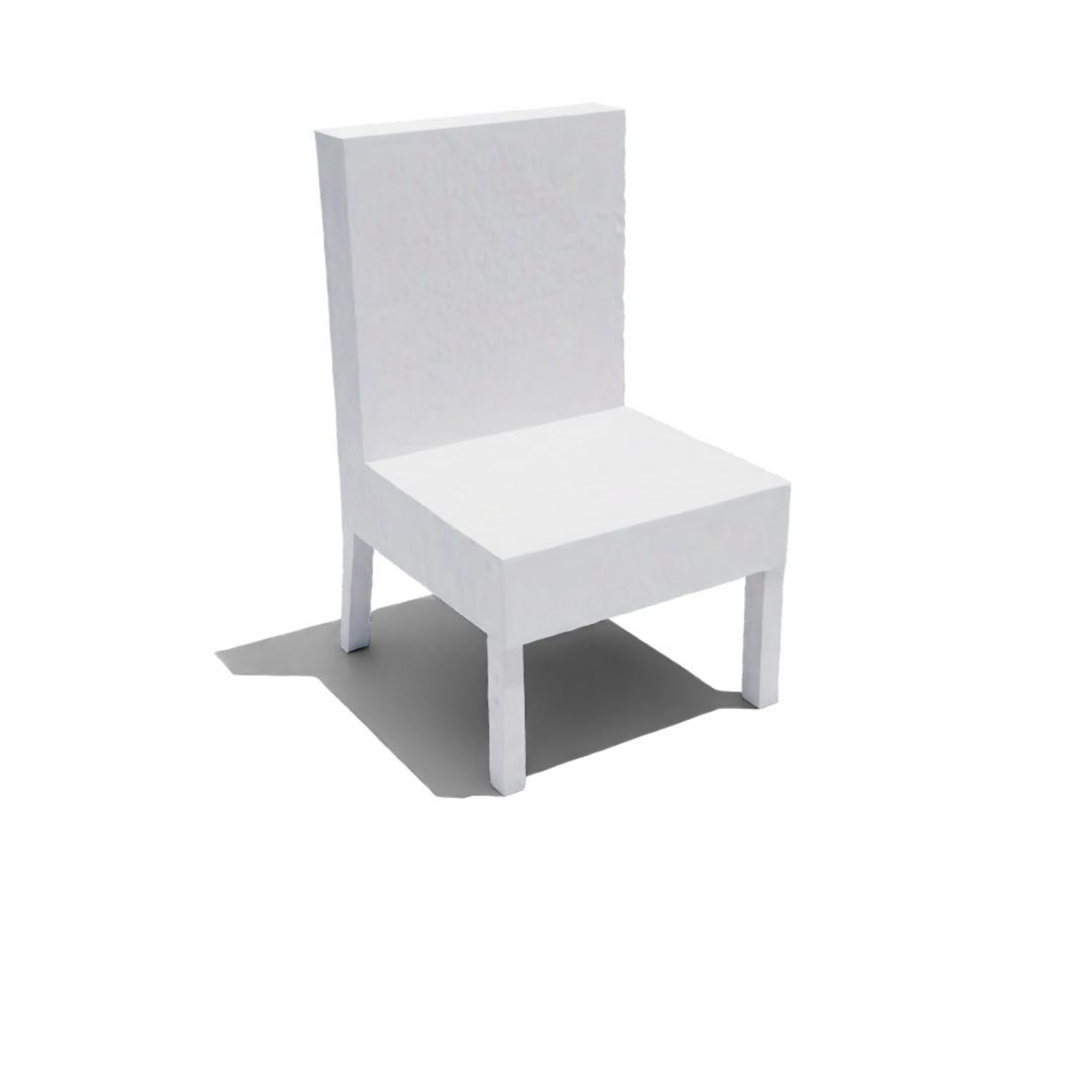}\includegraphics[width=0.08333333333333333\linewidth]{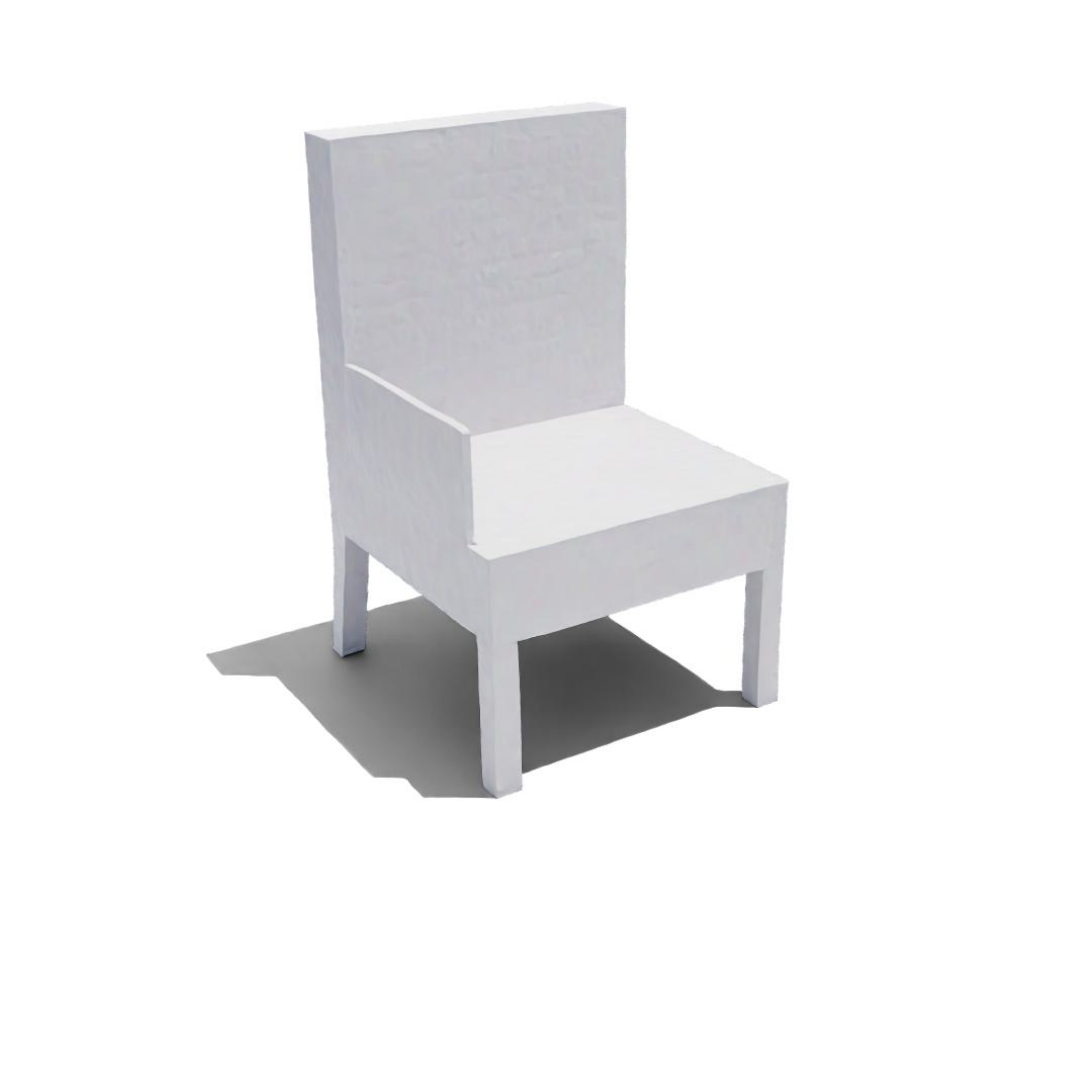}\includegraphics[width=0.08333333333333333\linewidth]{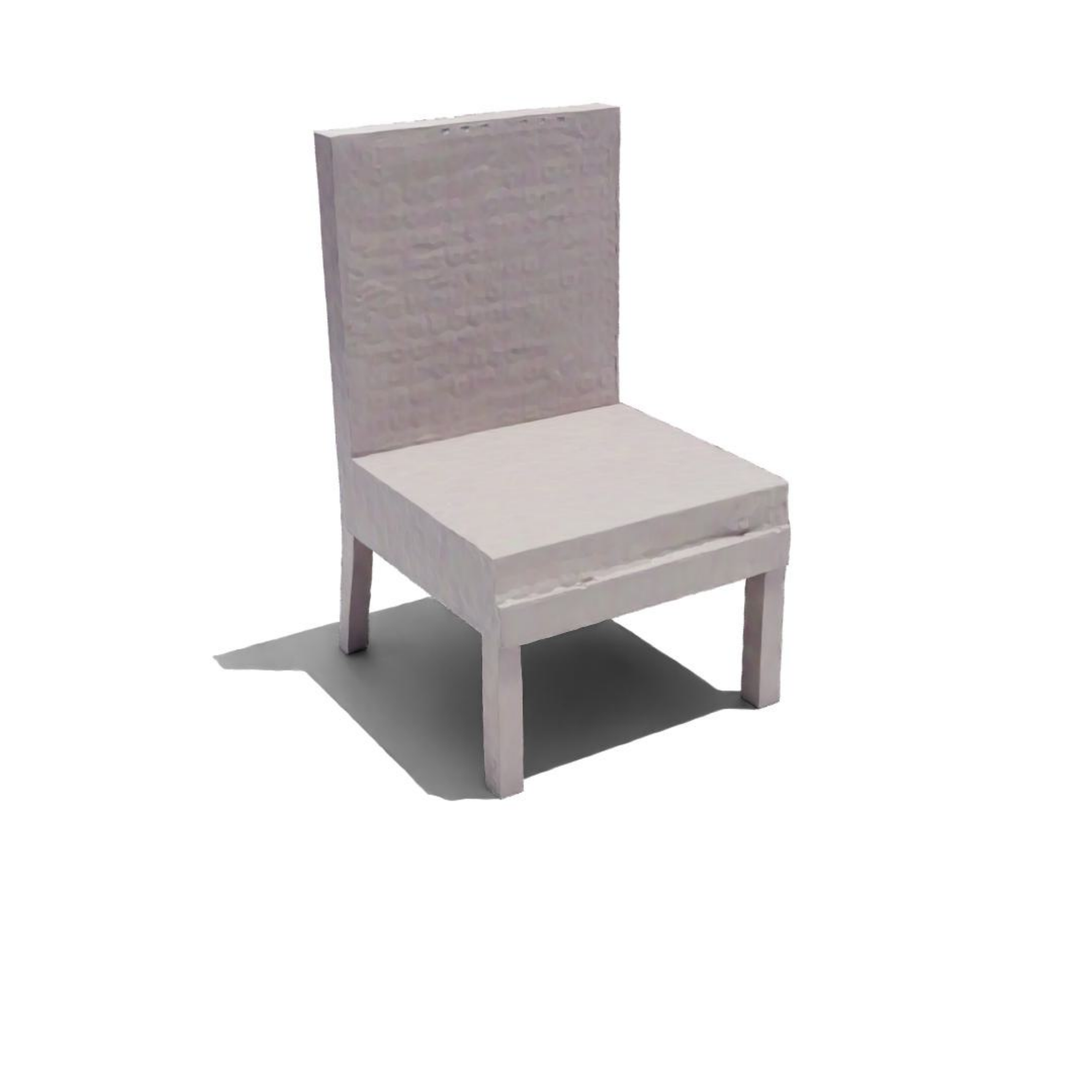}\includegraphics[width=0.08333333333333333\linewidth]{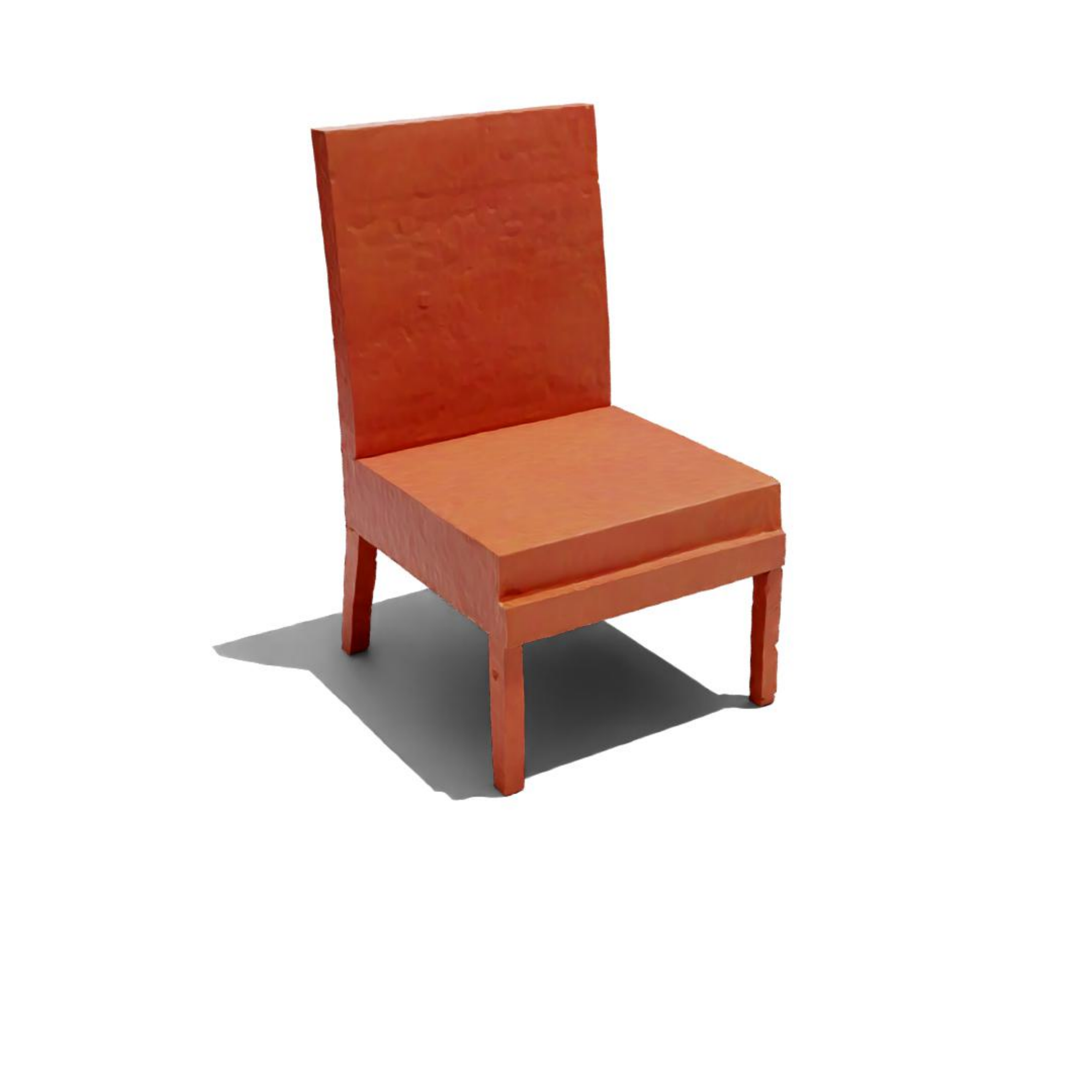}\includegraphics[width=0.08333333333333333\linewidth]{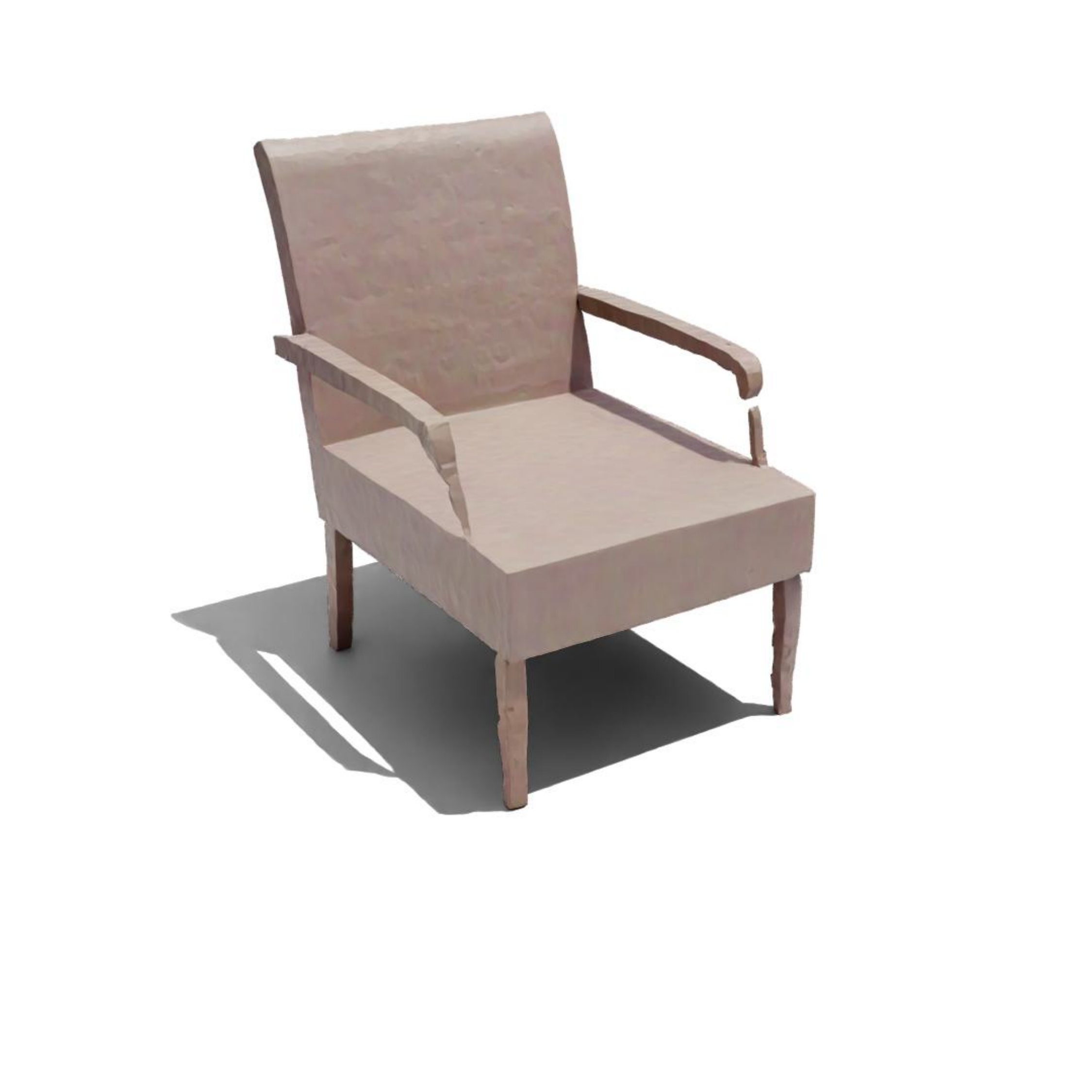}\includegraphics[width=0.08333333333333333\linewidth]{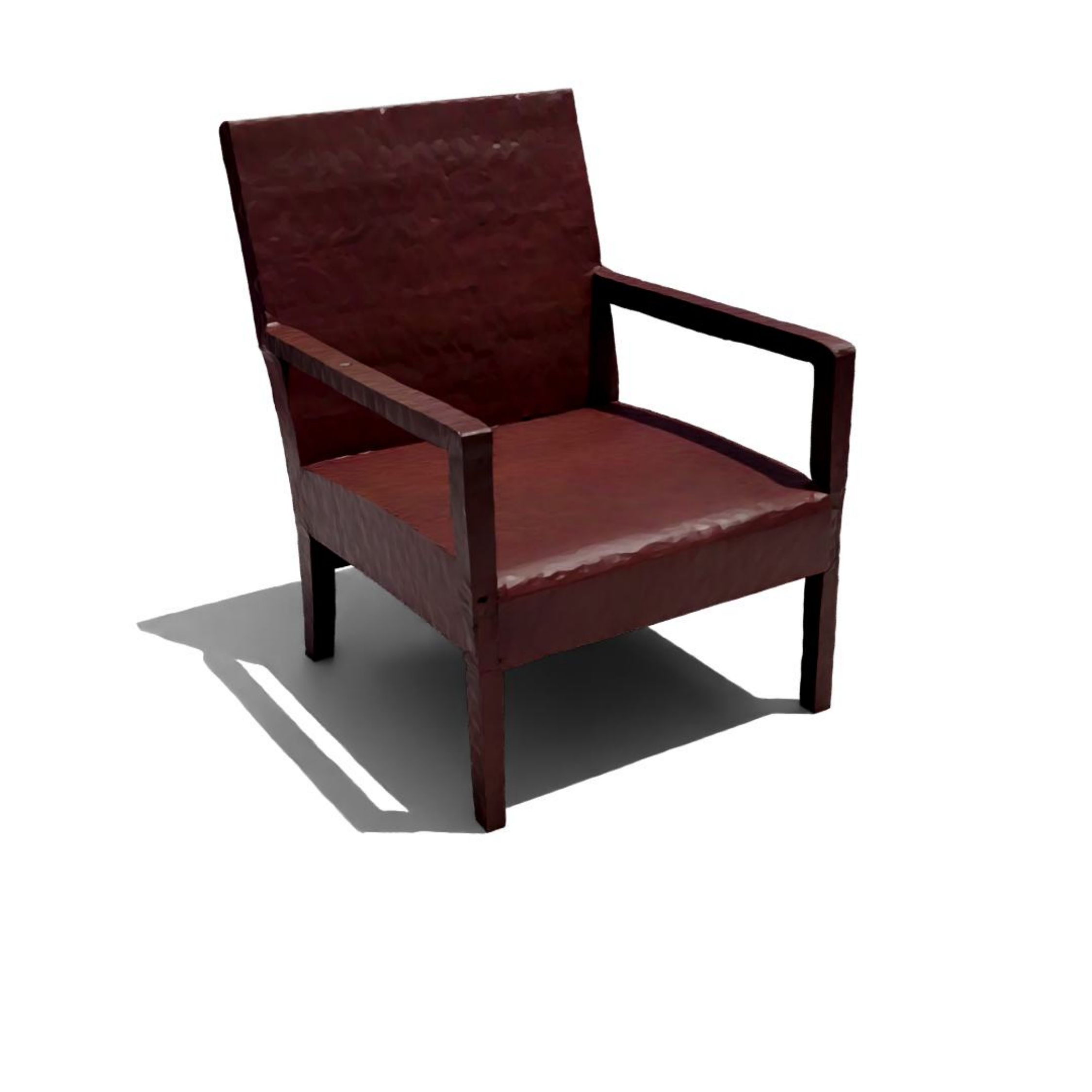}\includegraphics[width=0.08333333333333333\linewidth]{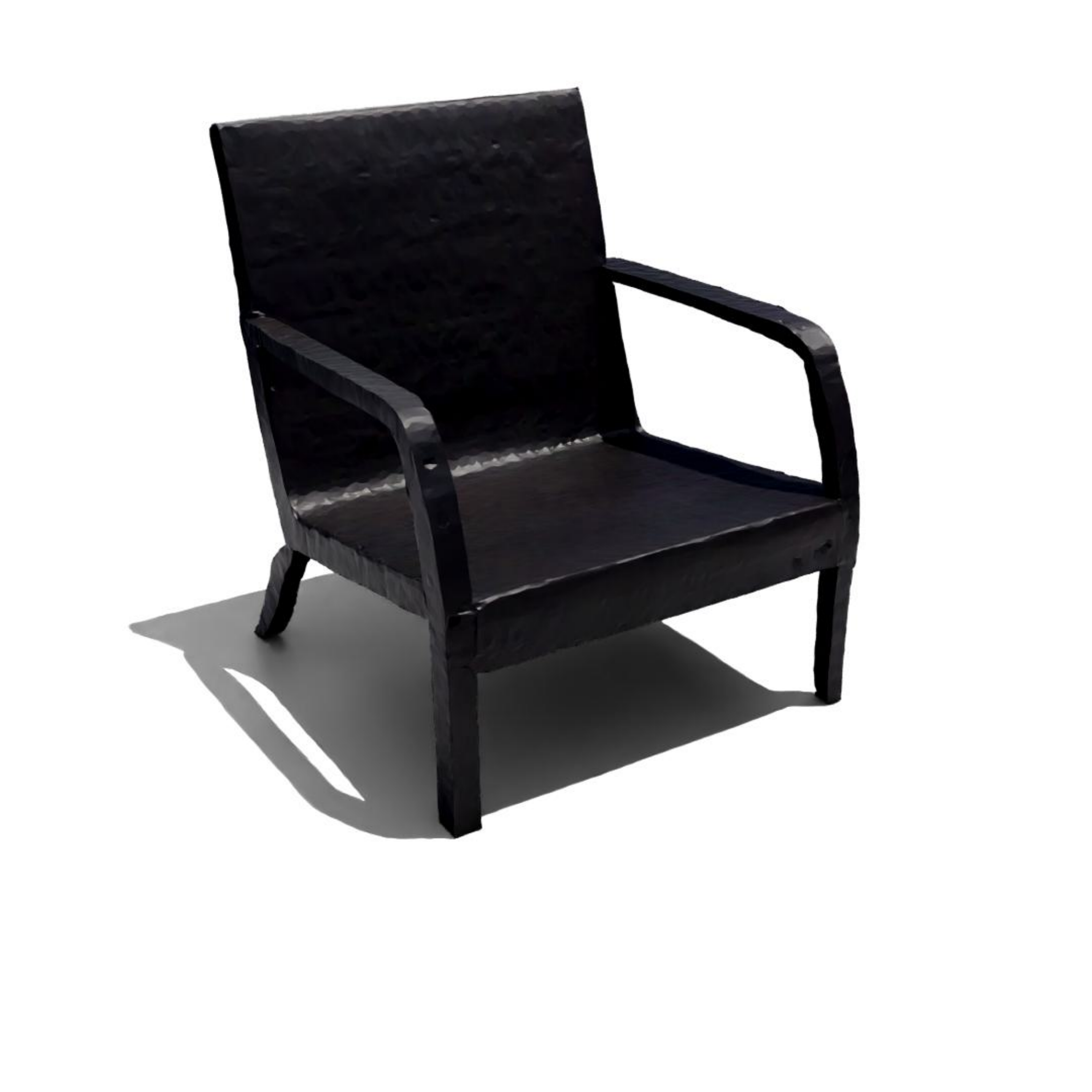}\includegraphics[width=0.08333333333333333\linewidth]{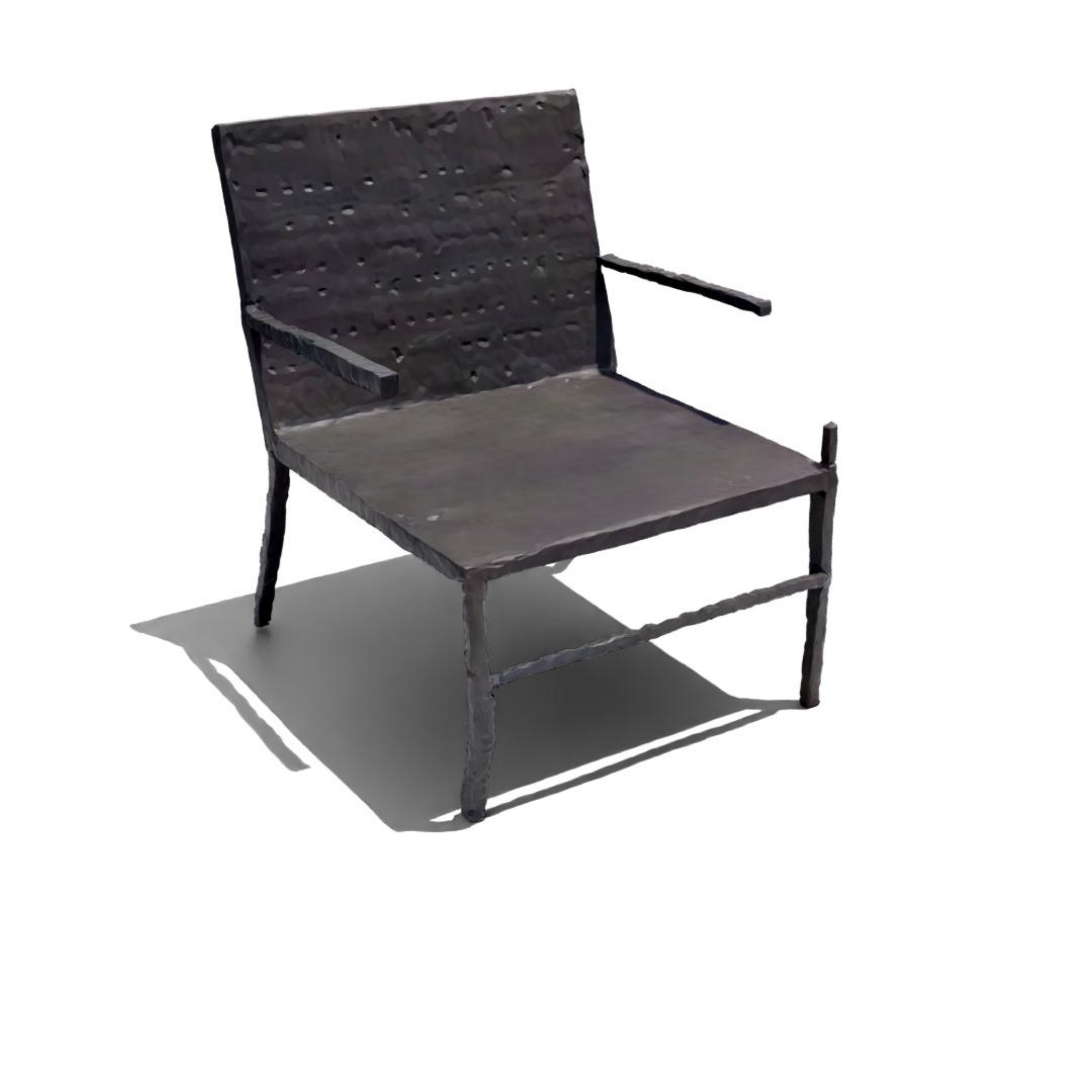}\includegraphics[width=0.08333333333333333\linewidth]{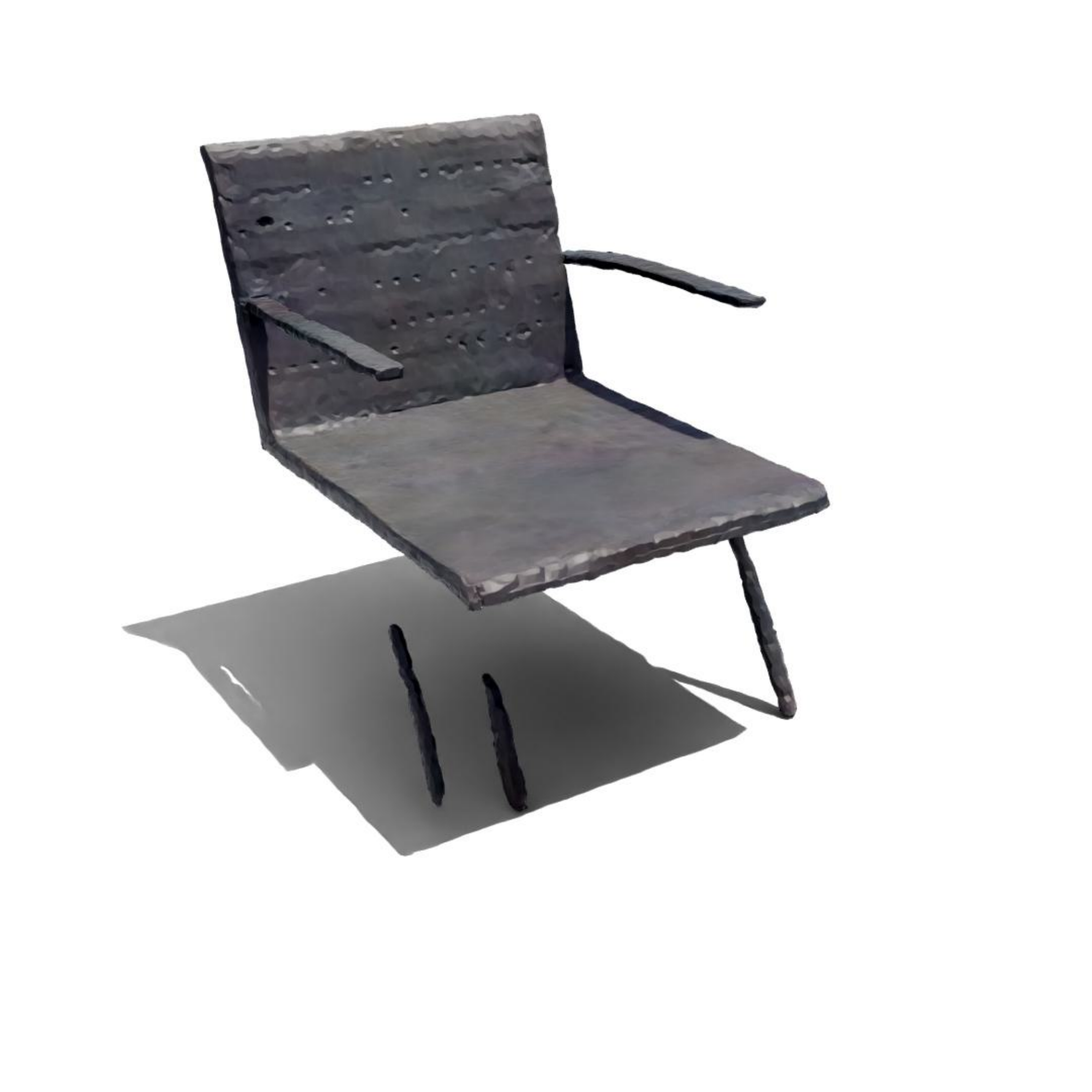}\\
\includegraphics[width=0.08333333333333333\linewidth]{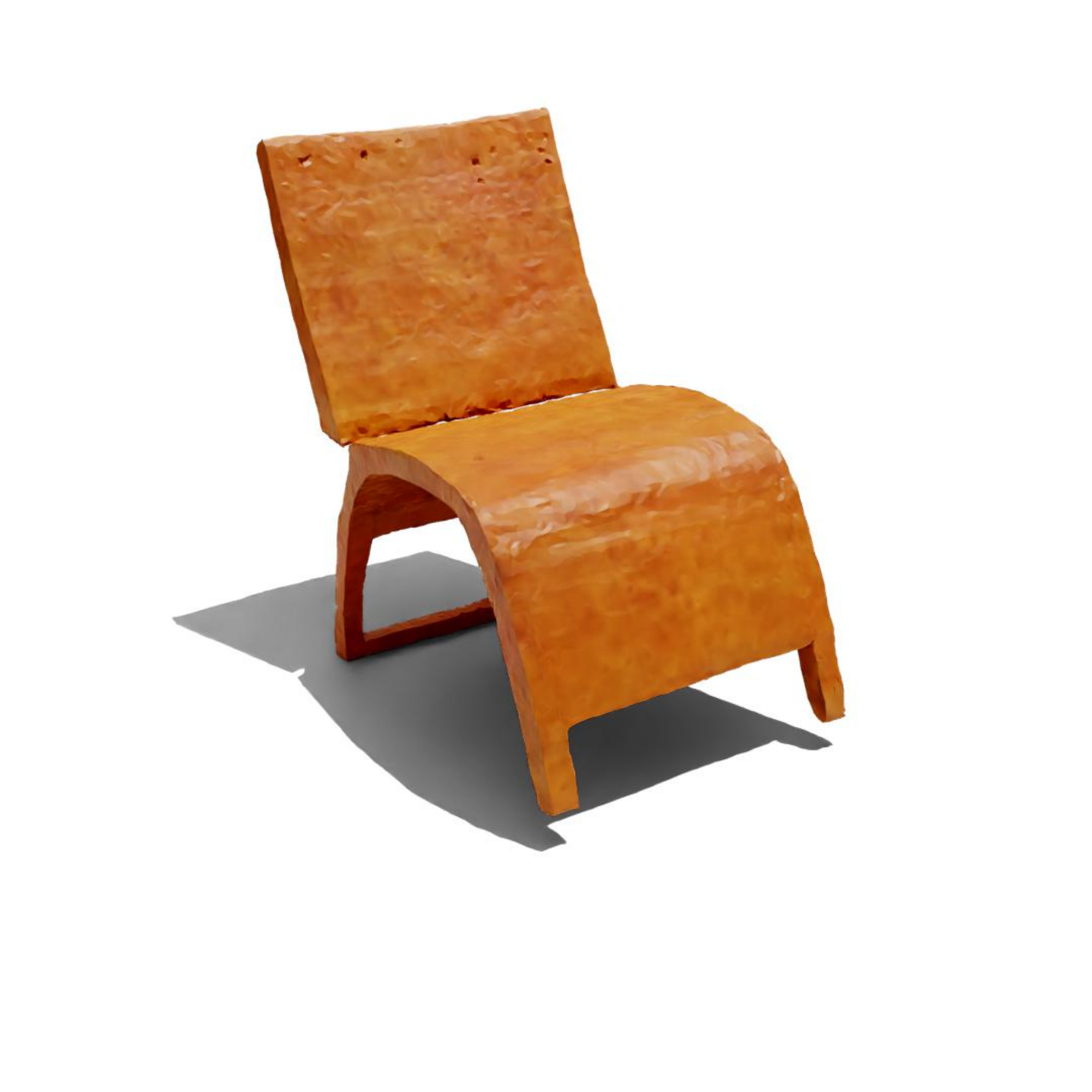}\includegraphics[width=0.08333333333333333\linewidth]{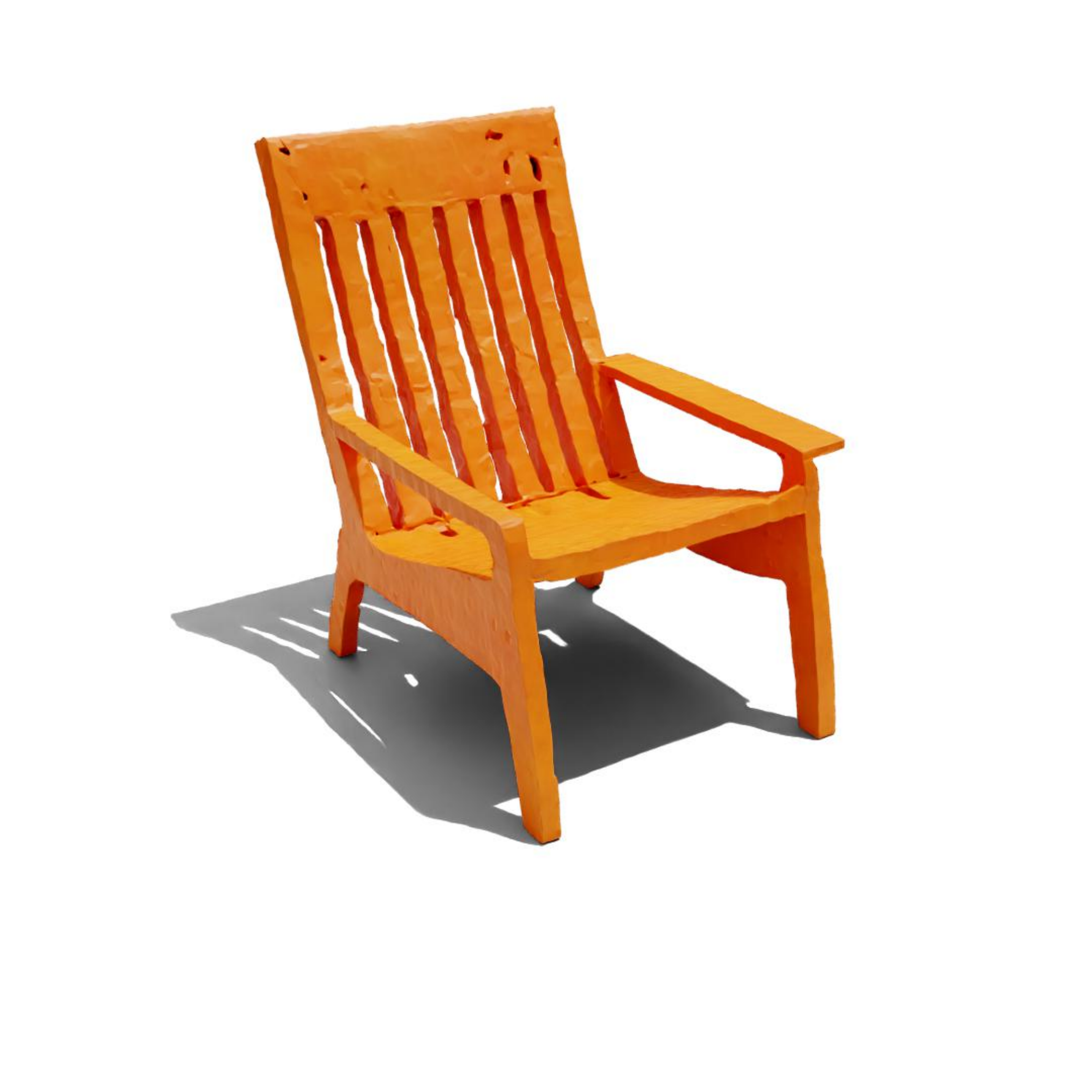}\includegraphics[width=0.08333333333333333\linewidth]{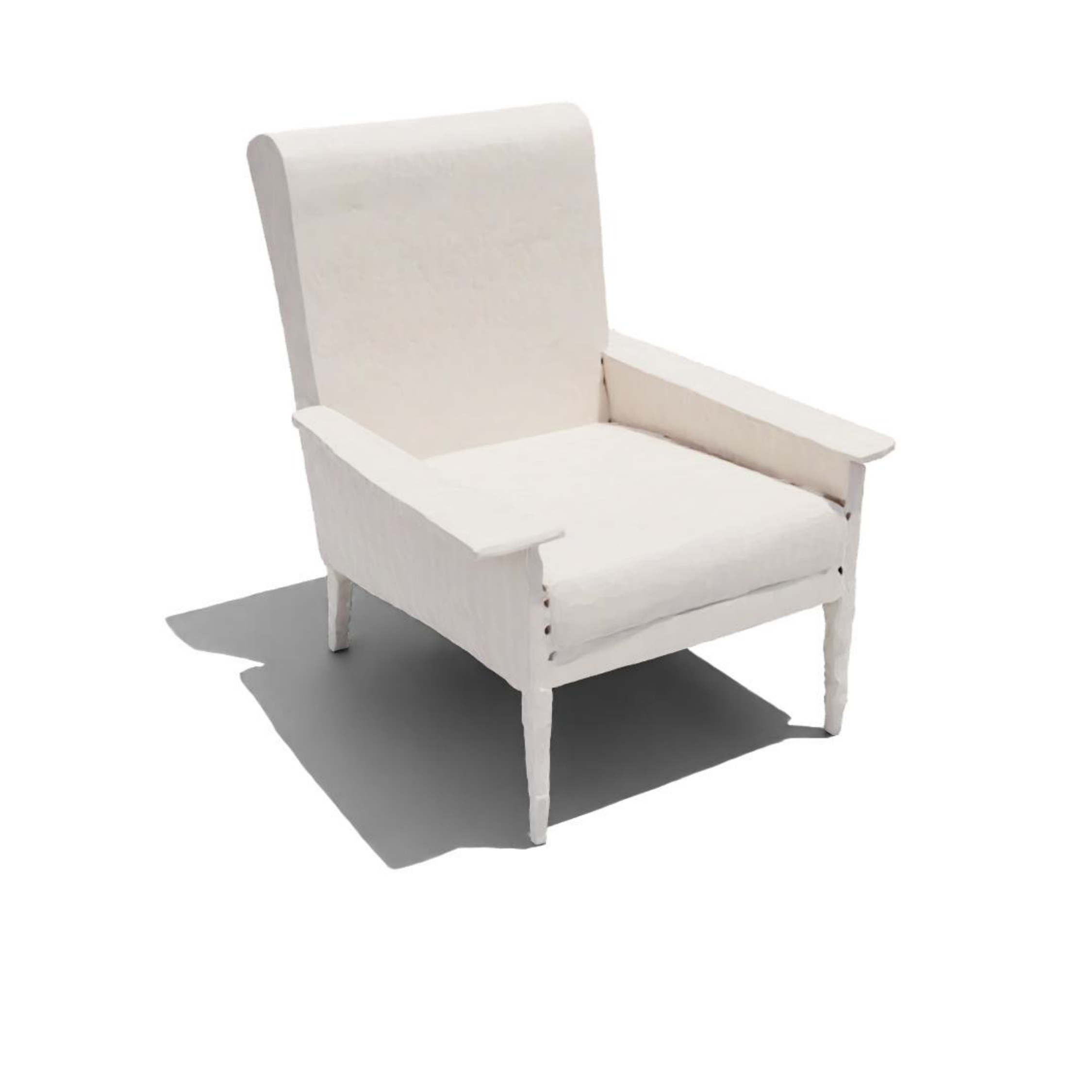}\includegraphics[width=0.08333333333333333\linewidth]{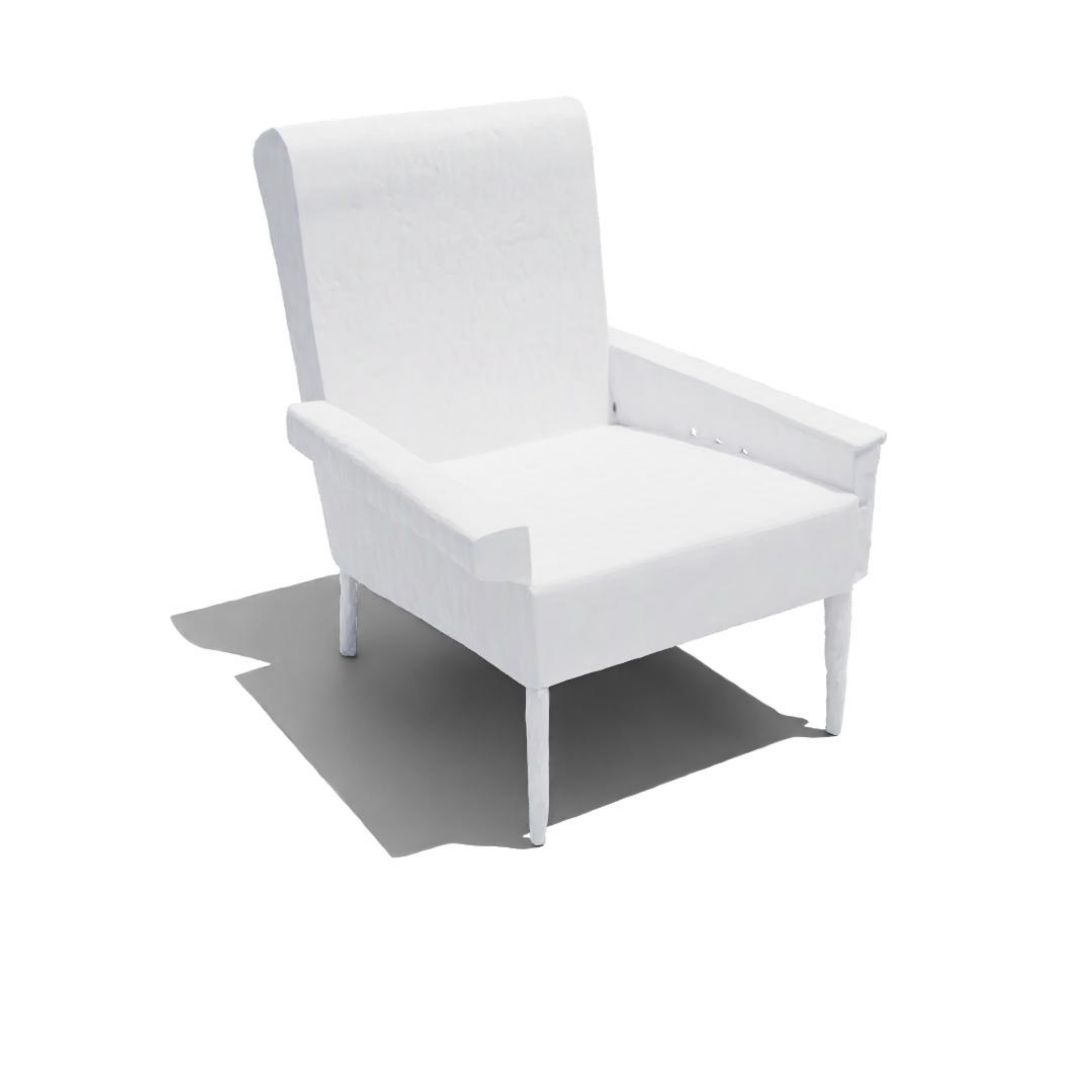}\includegraphics[width=0.08333333333333333\linewidth]{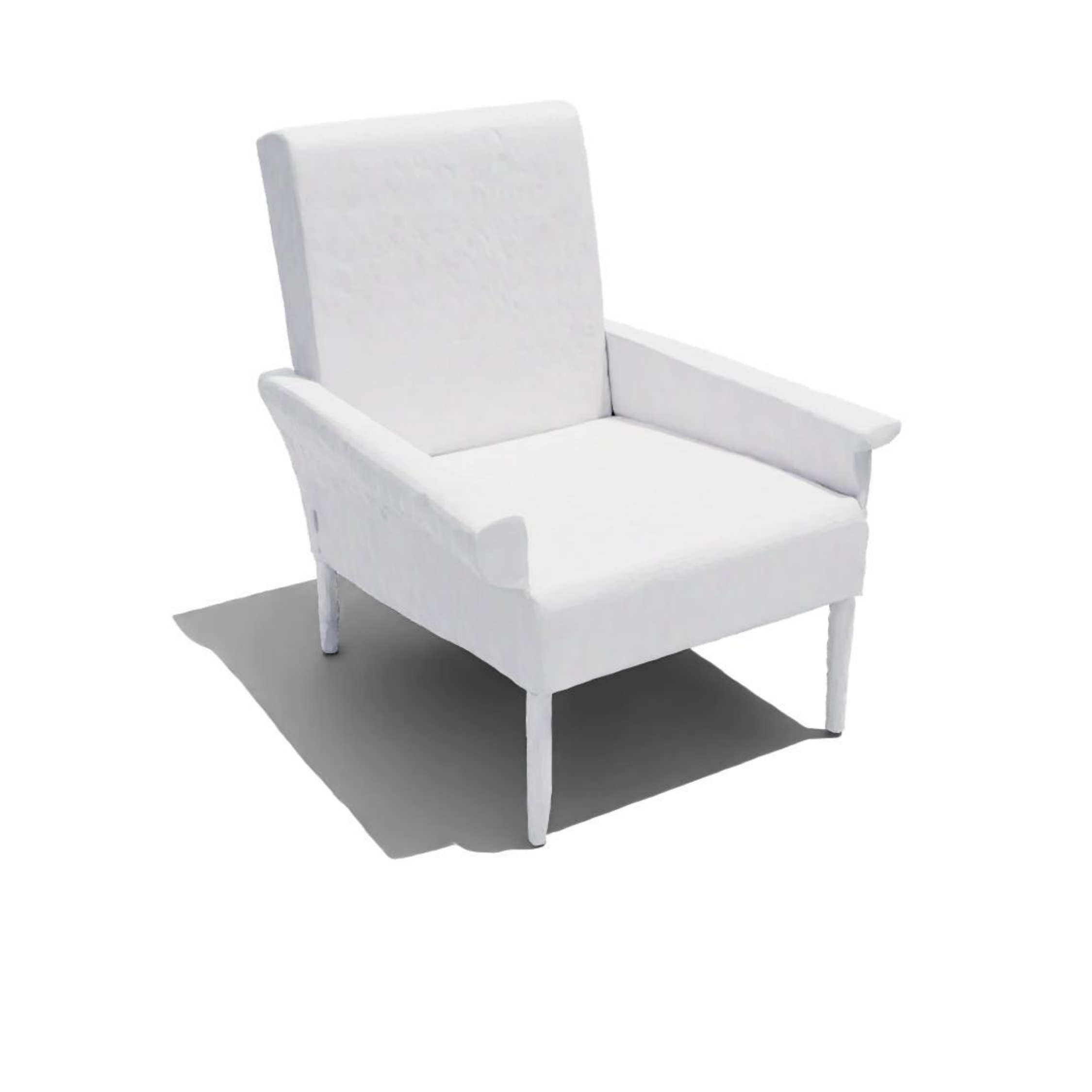}\includegraphics[width=0.08333333333333333\linewidth]{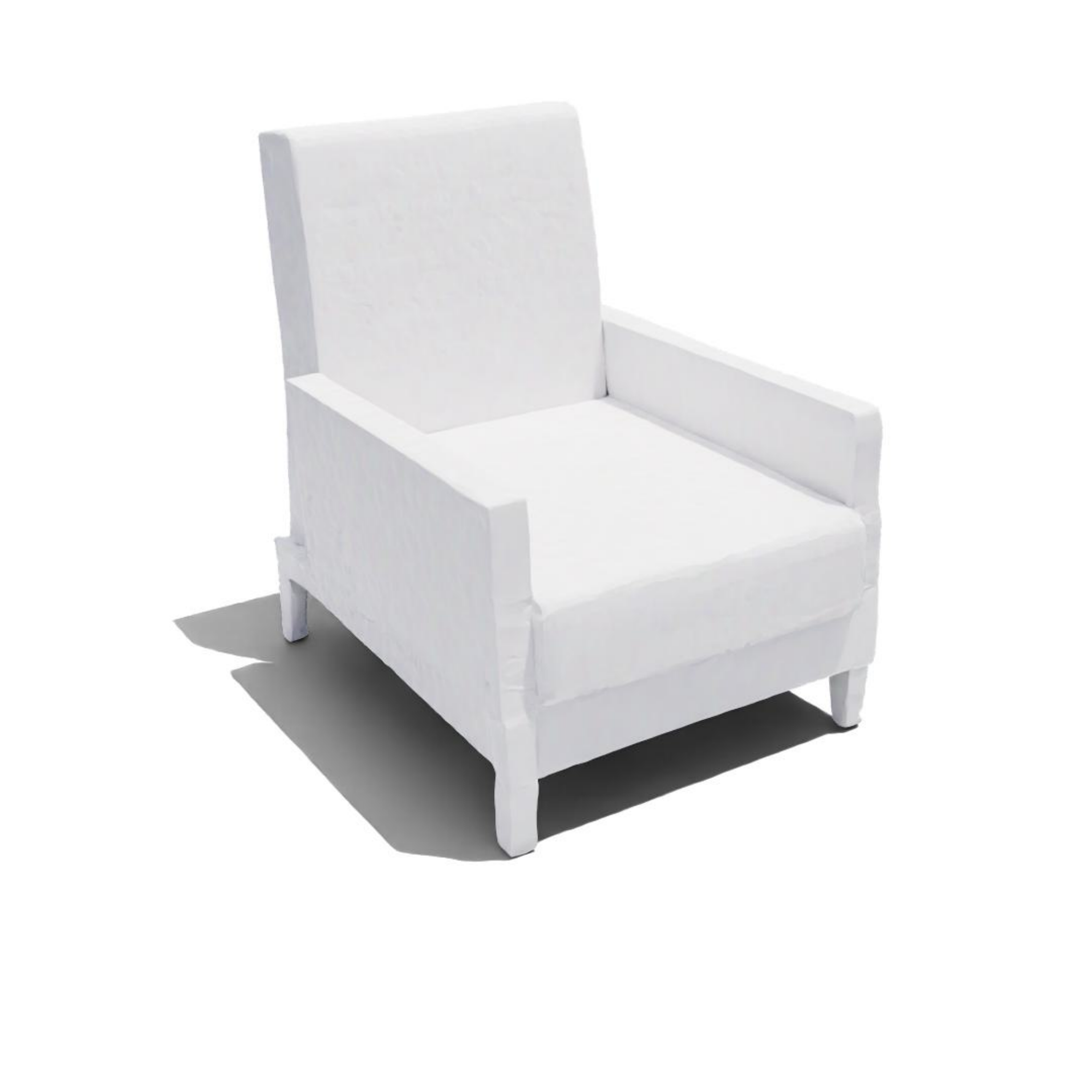}\includegraphics[width=0.08333333333333333\linewidth]{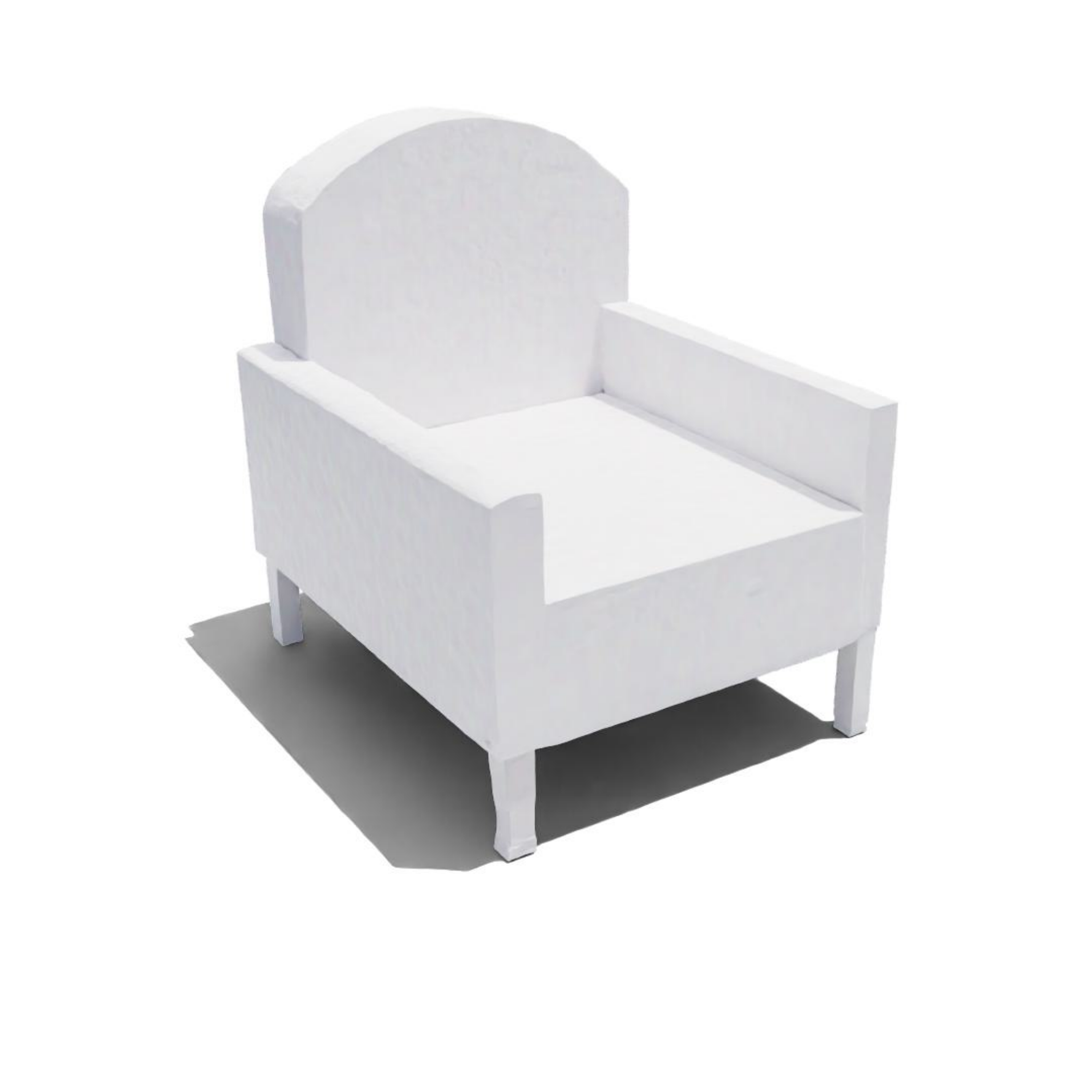}\includegraphics[width=0.08333333333333333\linewidth]{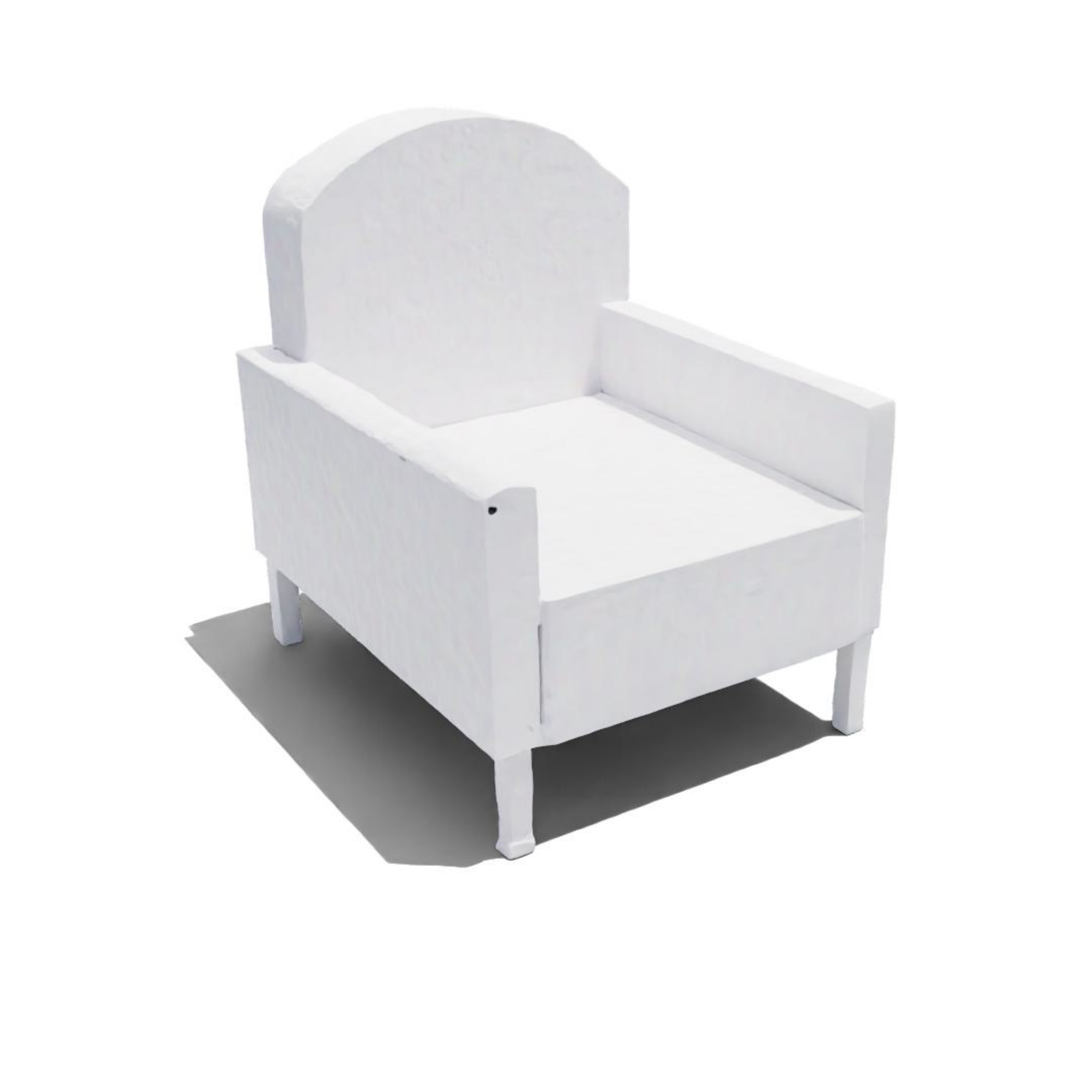}\includegraphics[width=0.08333333333333333\linewidth]{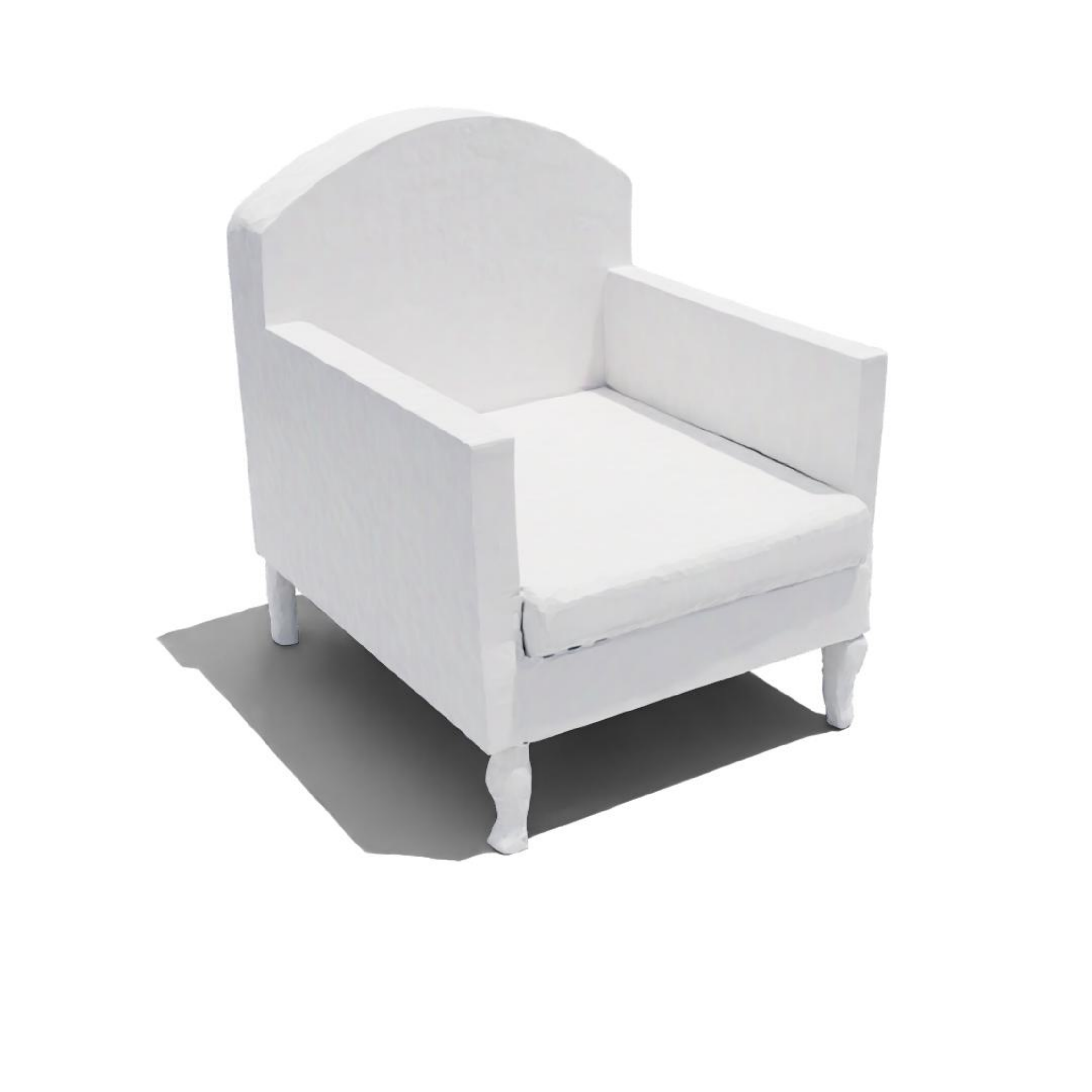}\includegraphics[width=0.08333333333333333\linewidth]{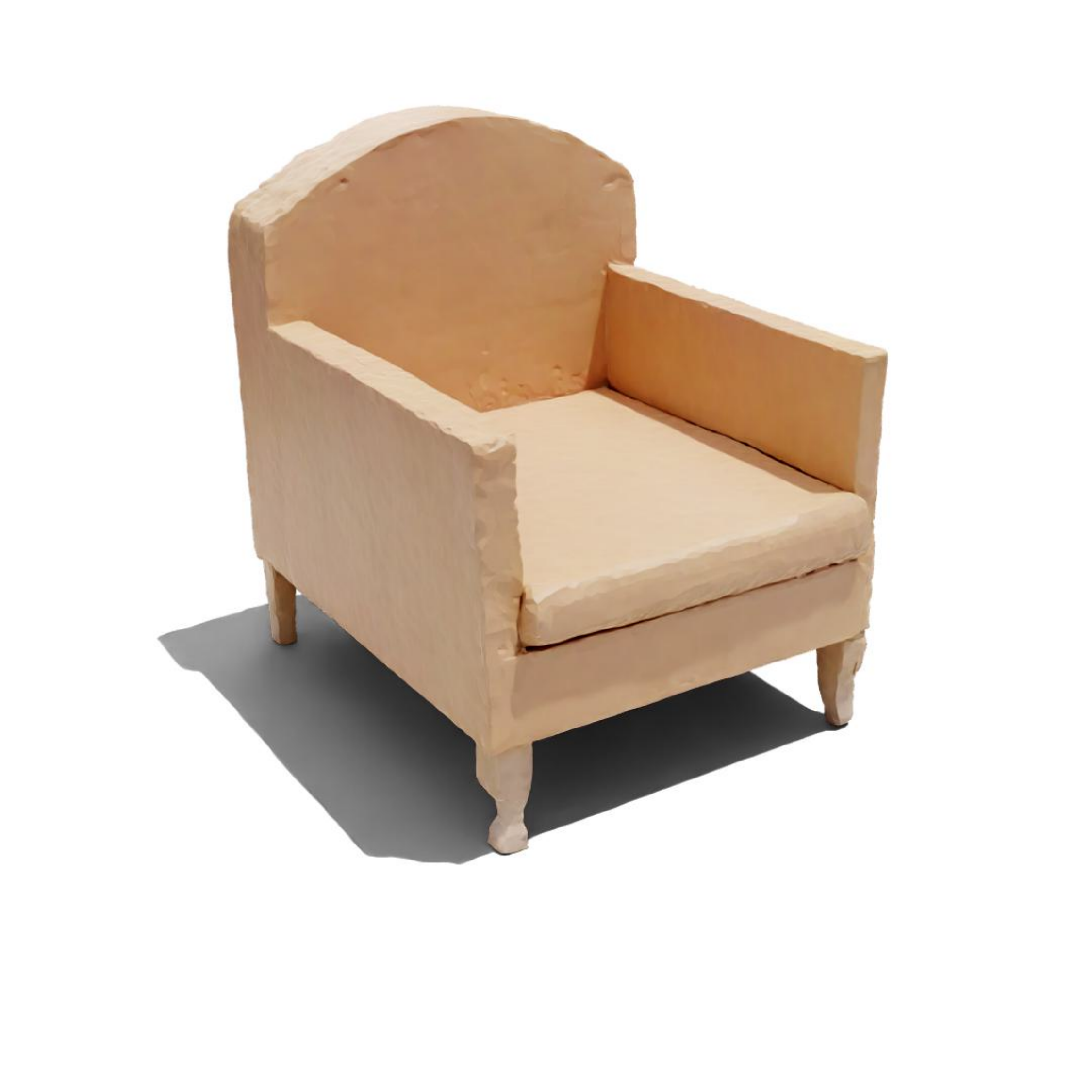}\includegraphics[width=0.08333333333333333\linewidth]{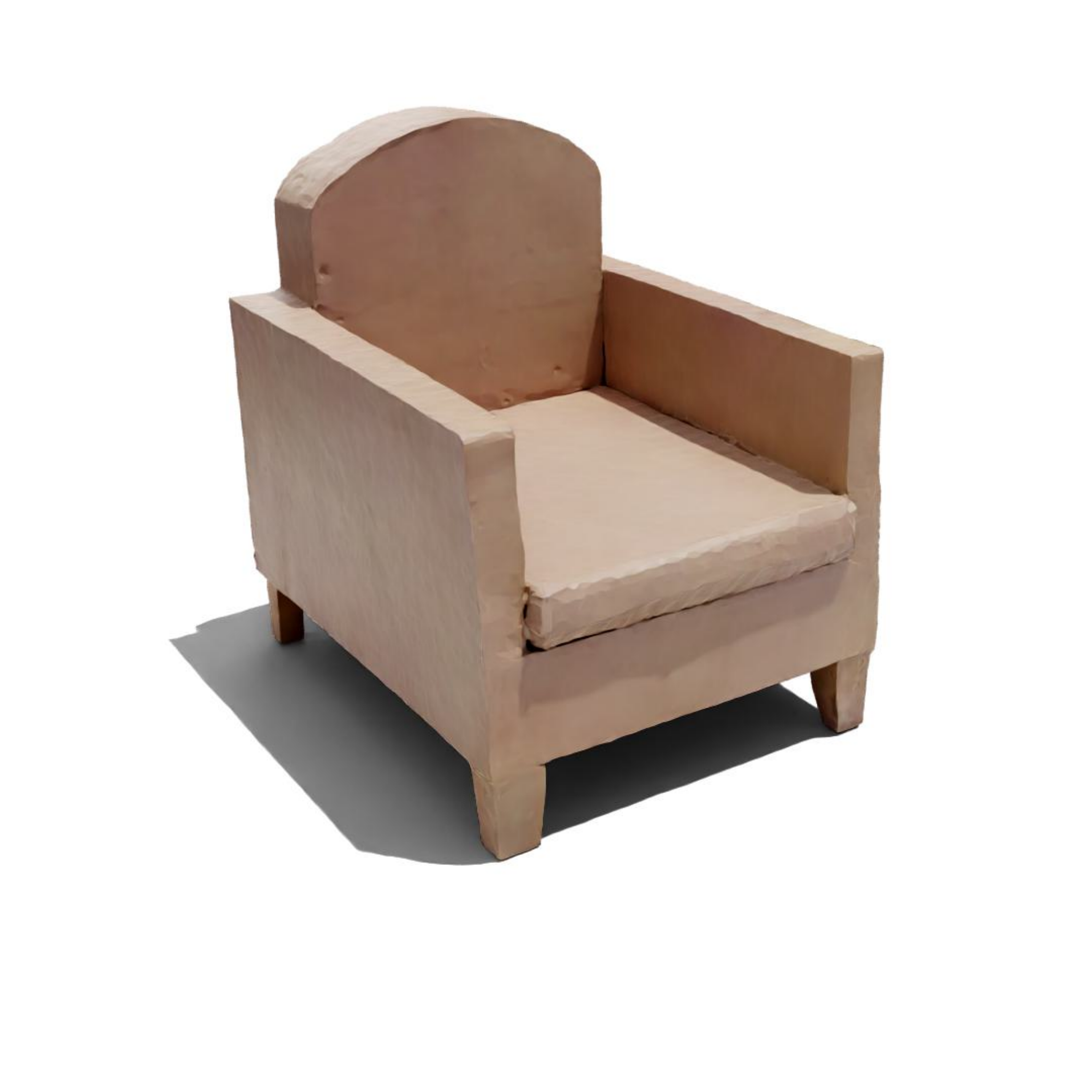}\includegraphics[width=0.08333333333333333\linewidth]{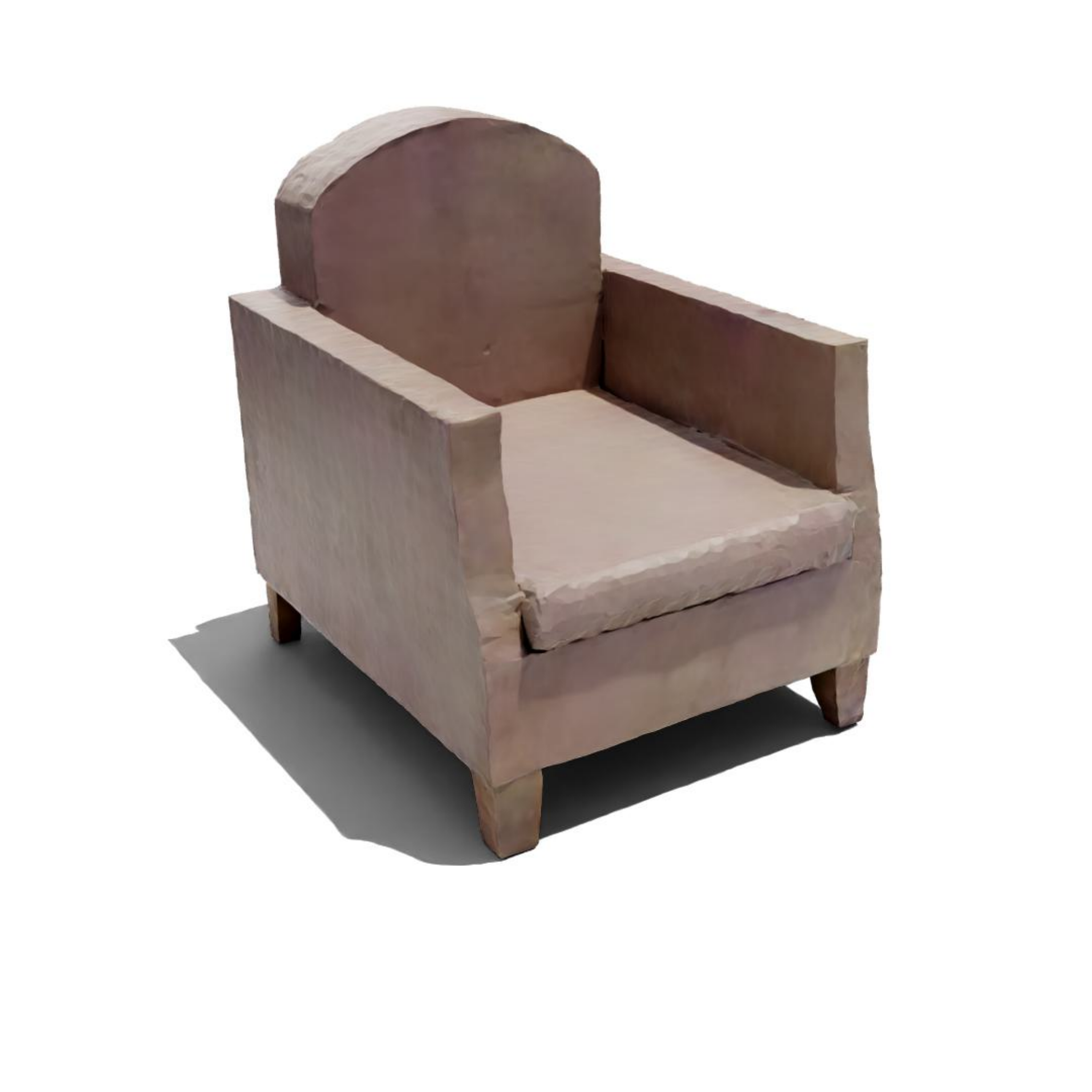}\\
\includegraphics[width=0.08333333333333333\linewidth]{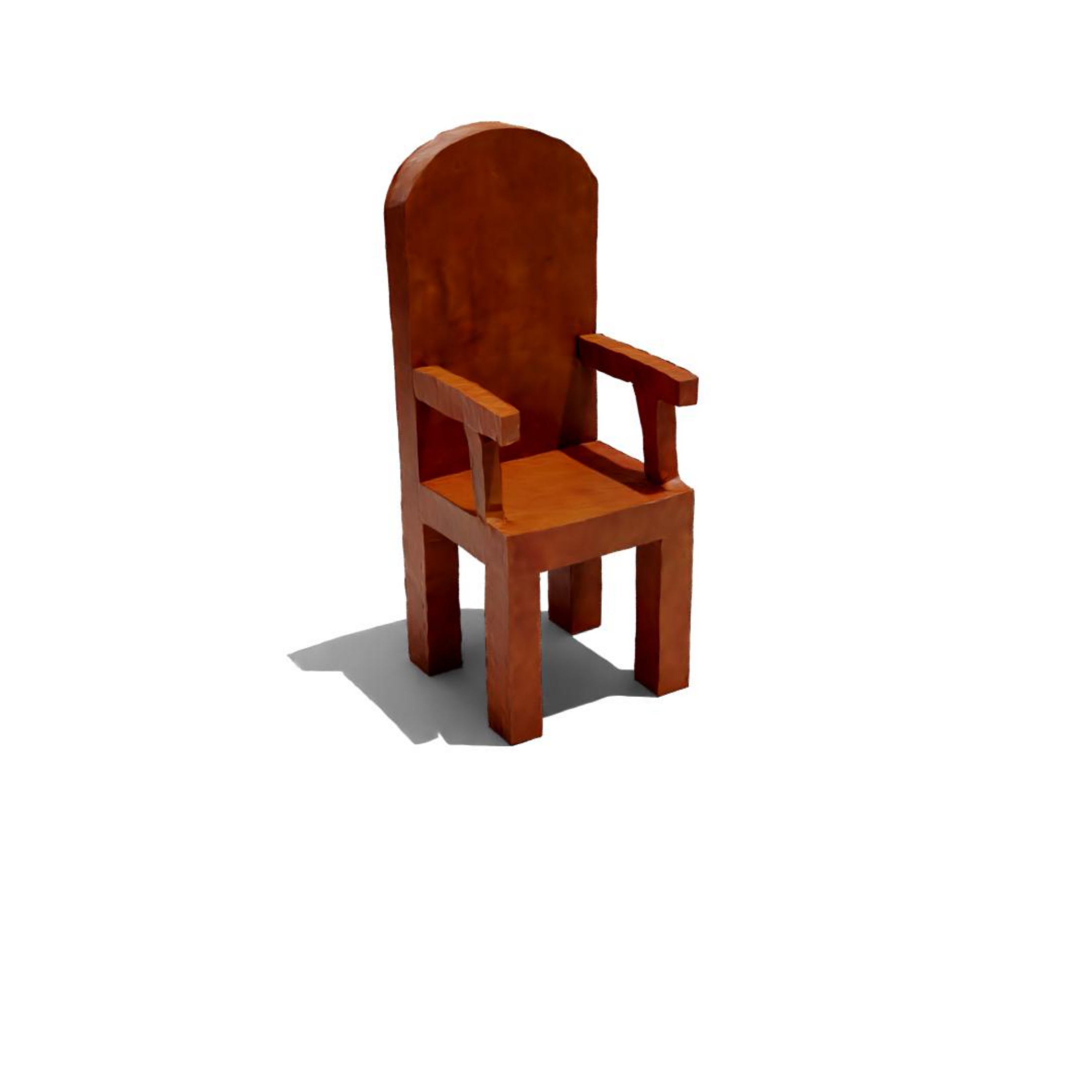}\includegraphics[width=0.08333333333333333\linewidth]{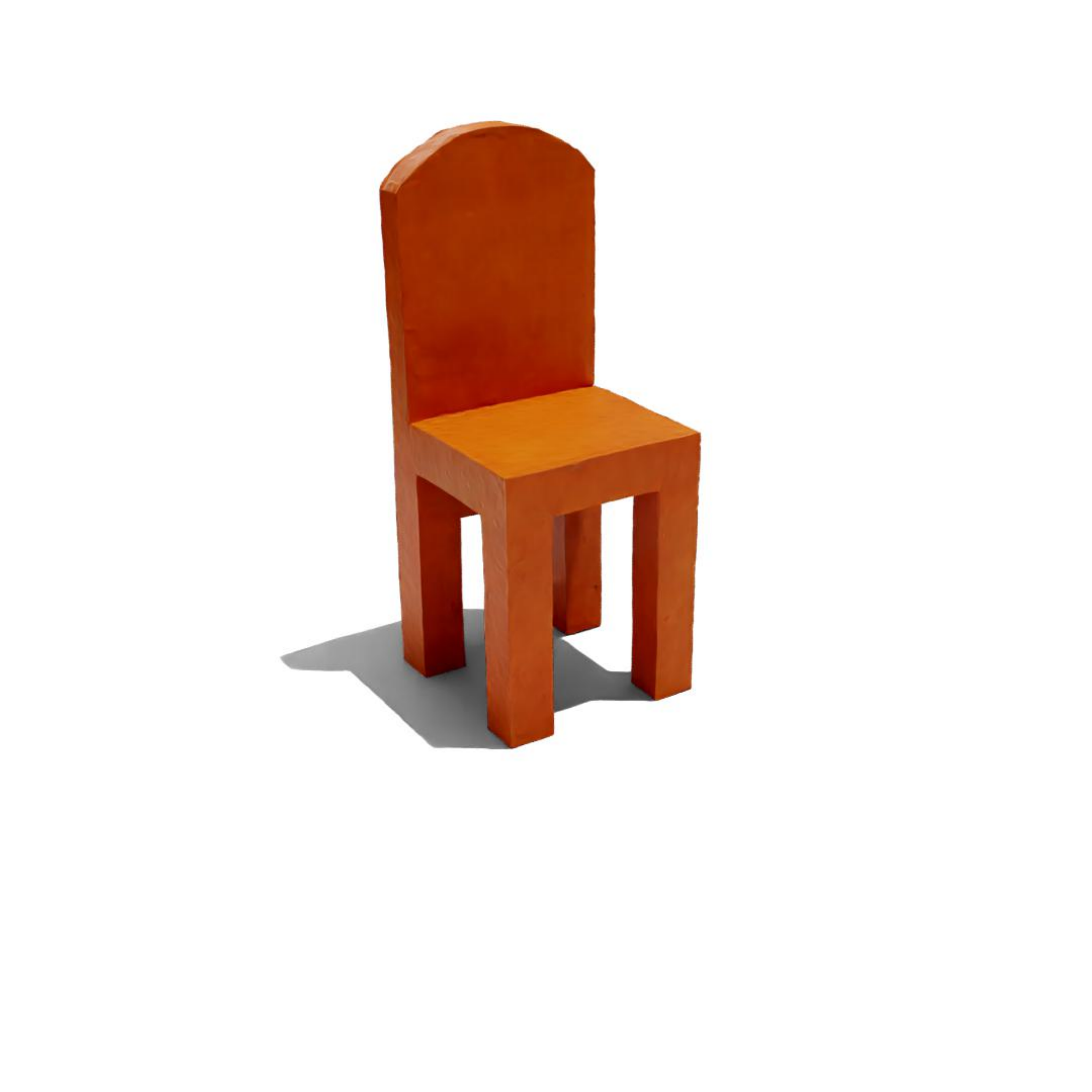}\includegraphics[width=0.08333333333333333\linewidth]{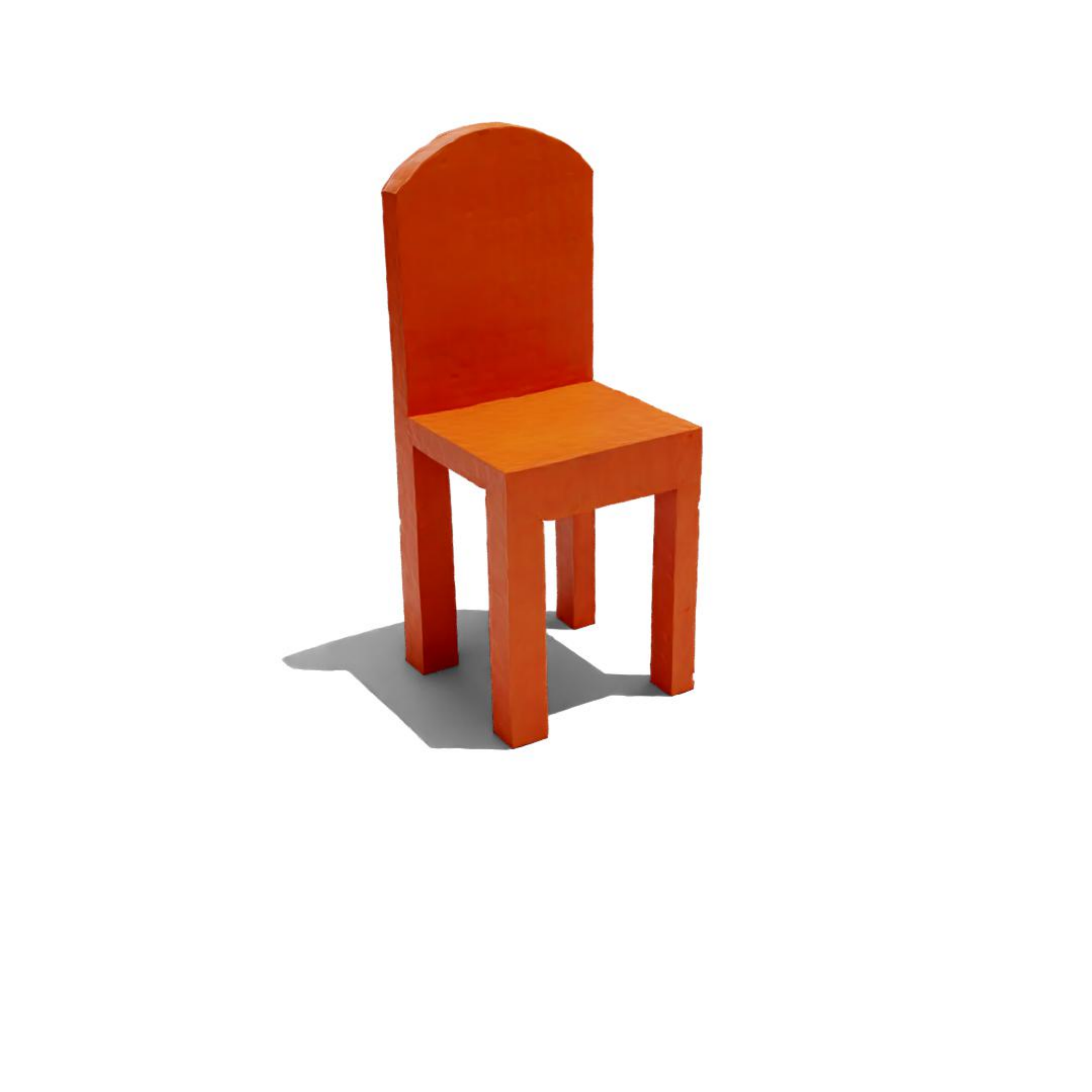}\includegraphics[width=0.08333333333333333\linewidth]{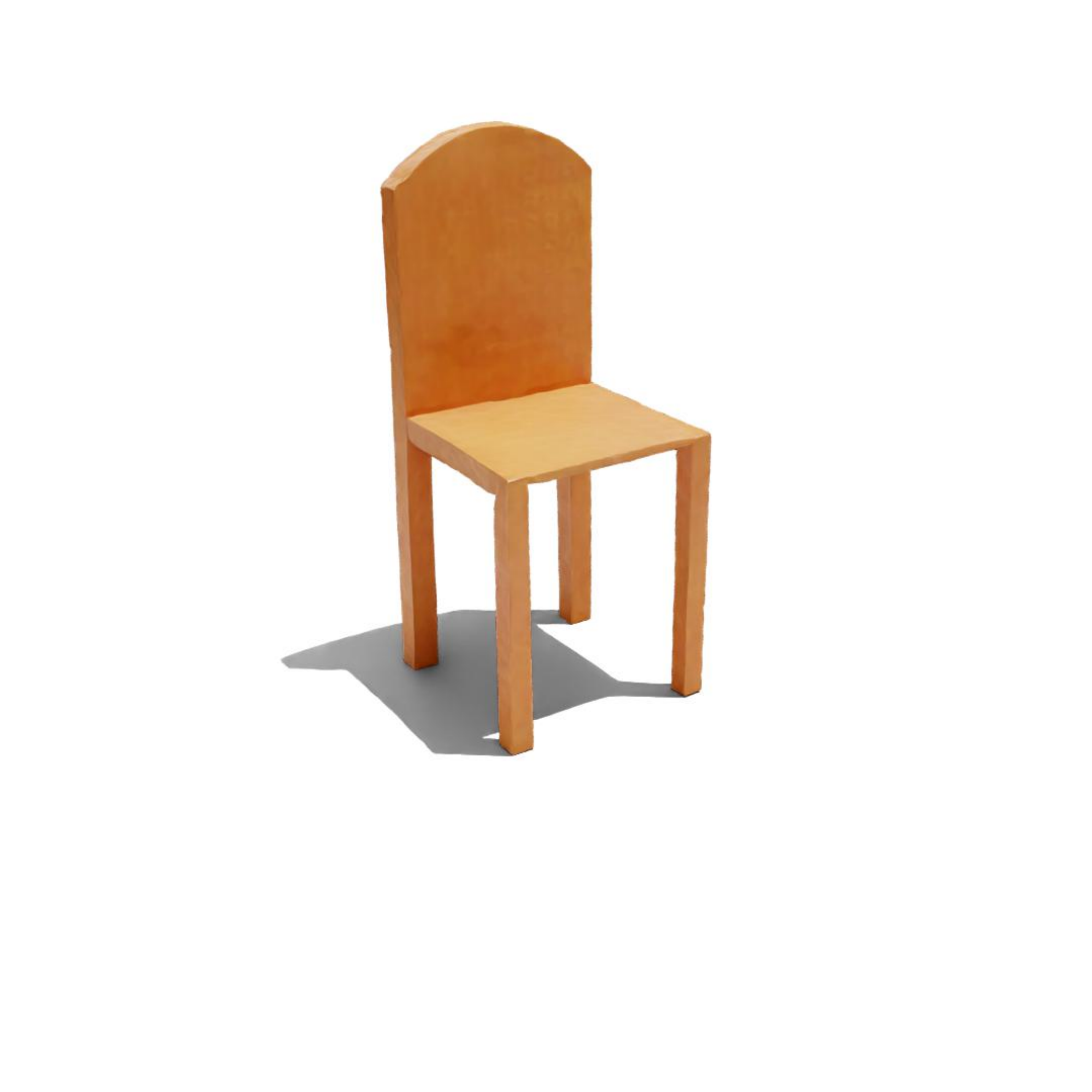}\includegraphics[width=0.08333333333333333\linewidth]{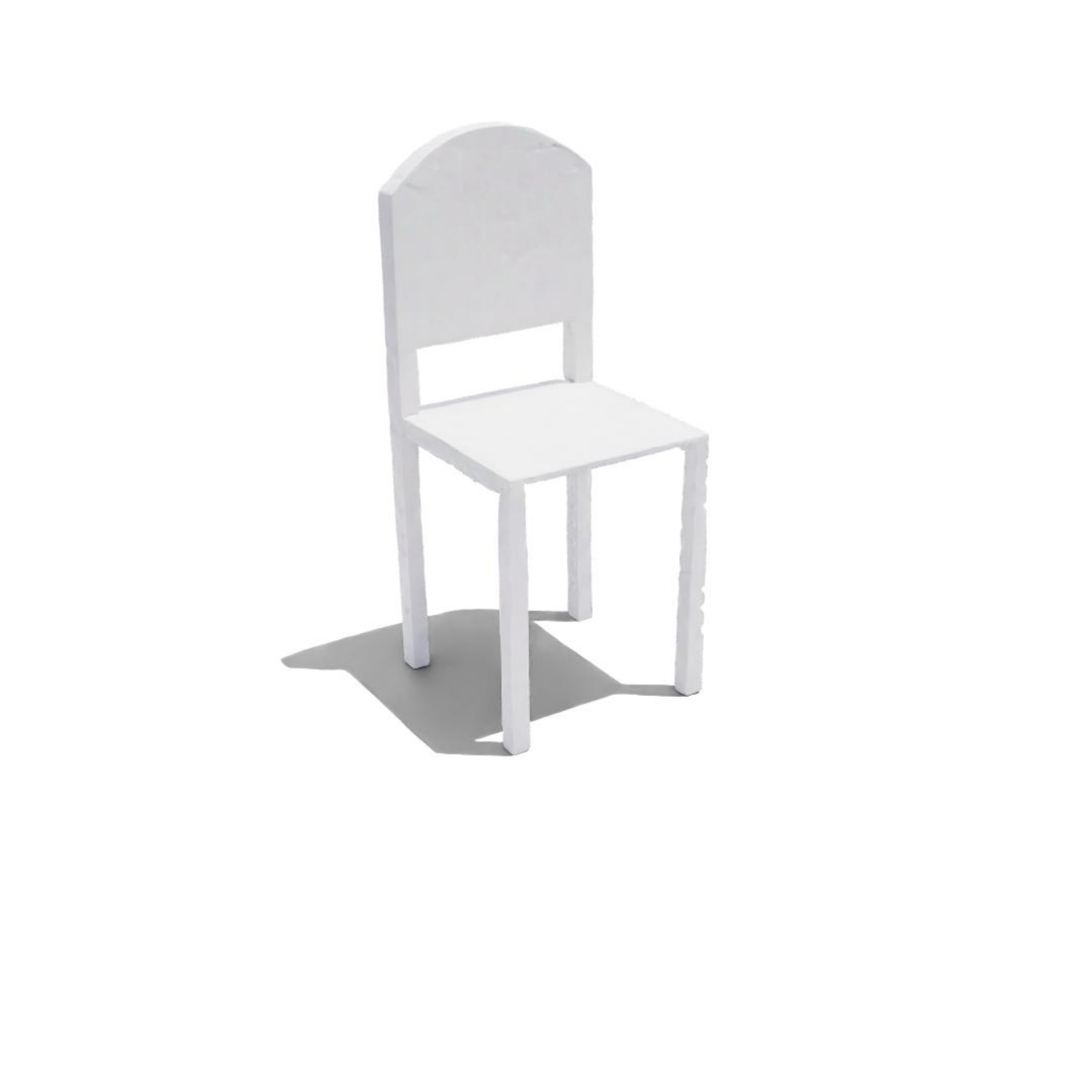}\includegraphics[width=0.08333333333333333\linewidth]{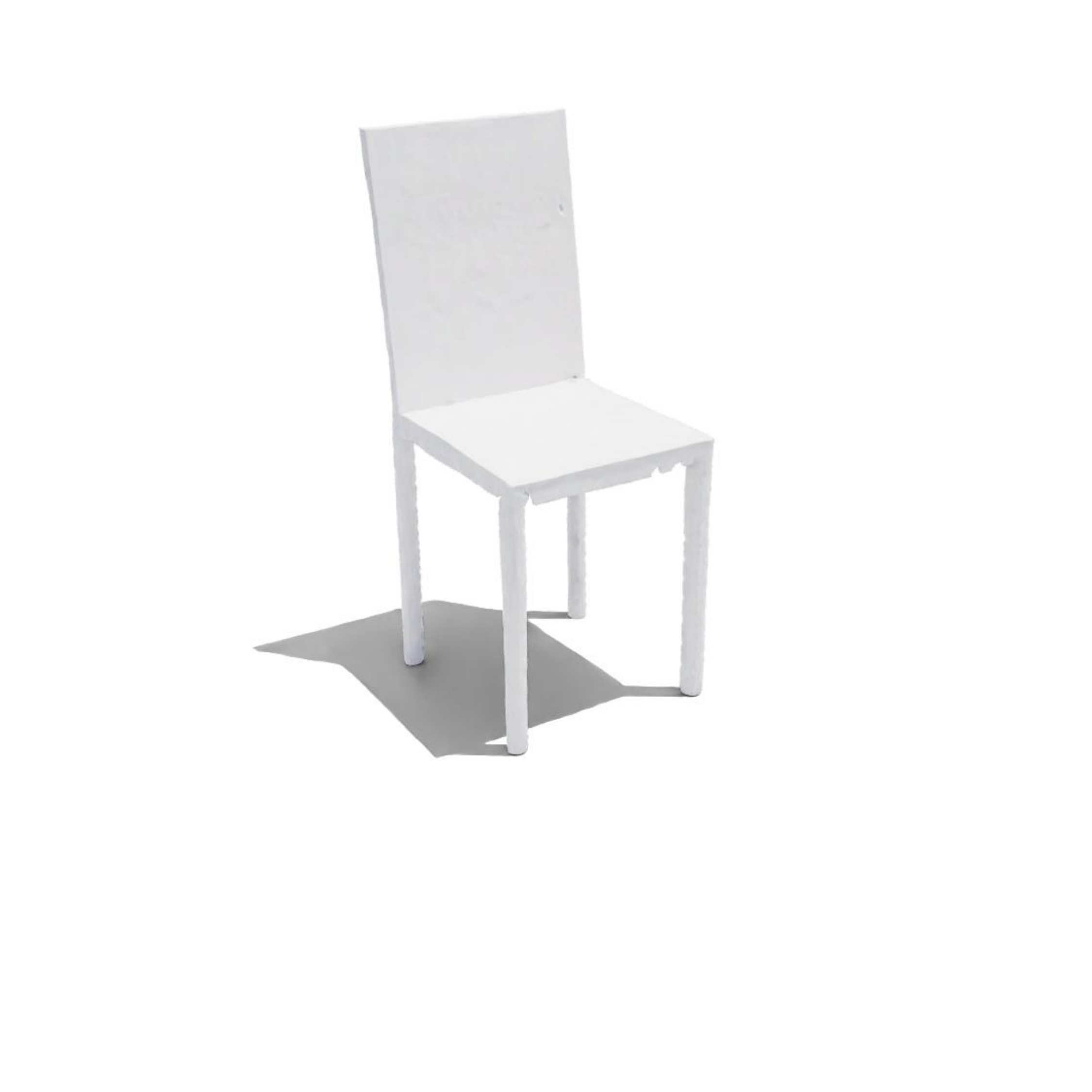}\includegraphics[width=0.08333333333333333\linewidth]{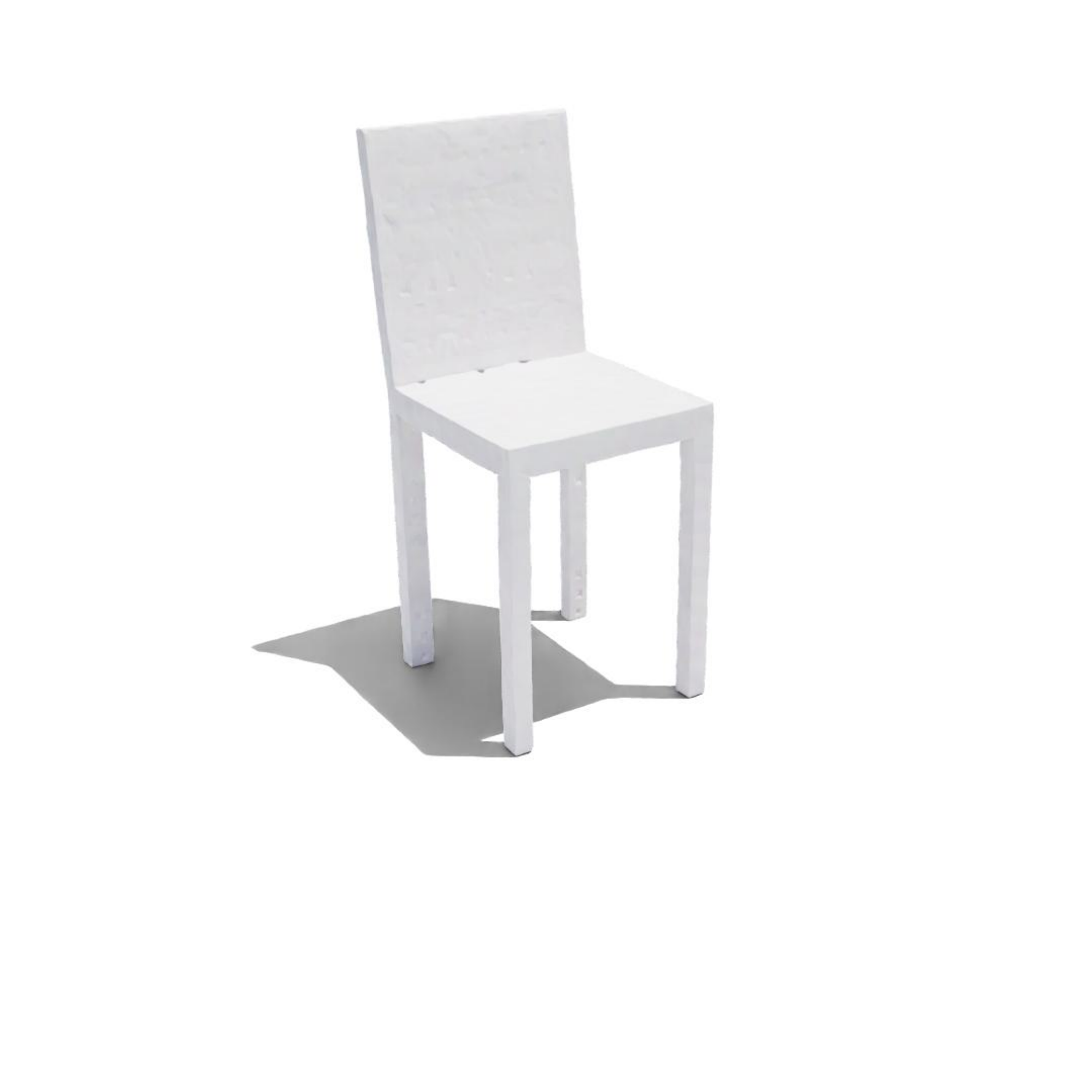}\includegraphics[width=0.08333333333333333\linewidth]{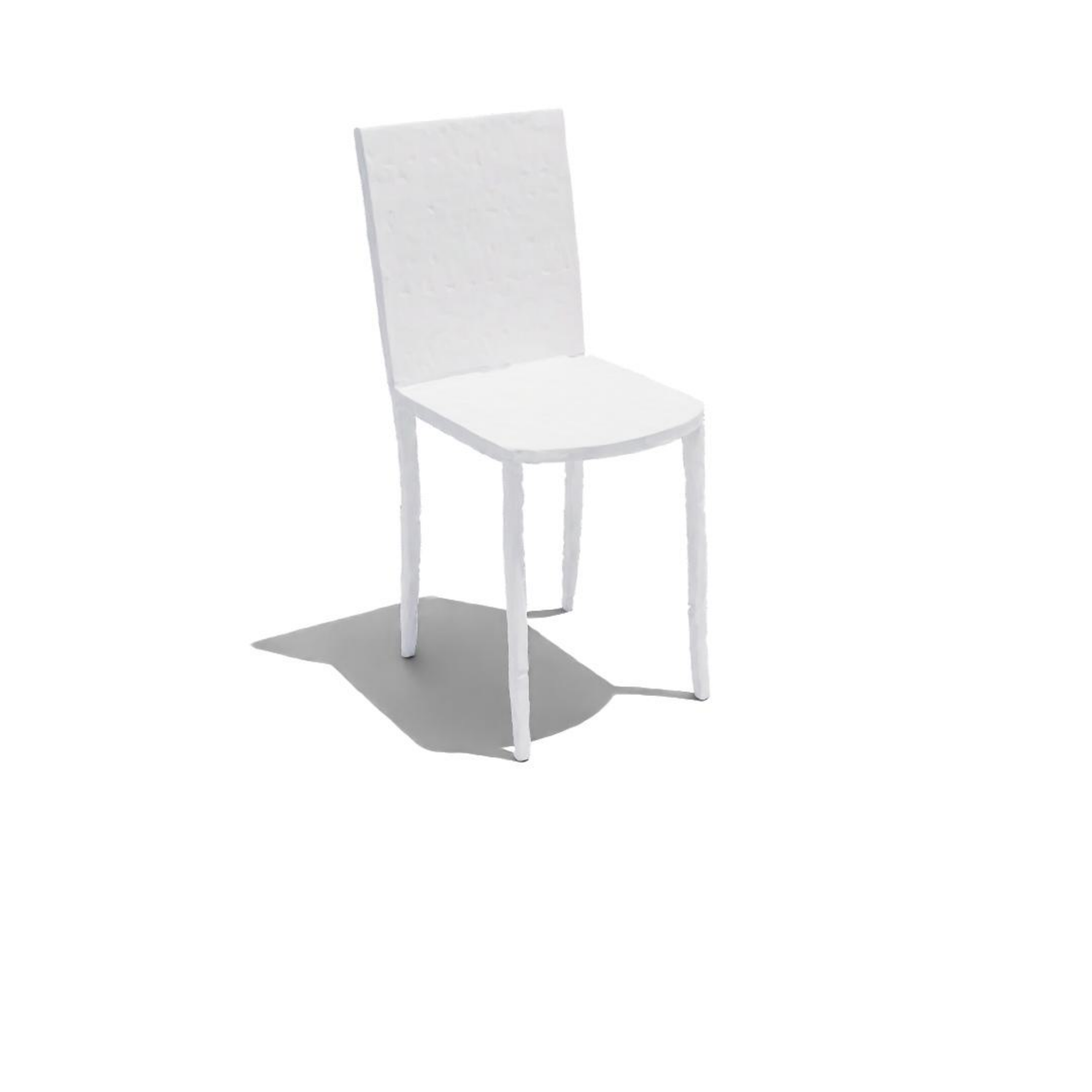}\includegraphics[width=0.08333333333333333\linewidth]{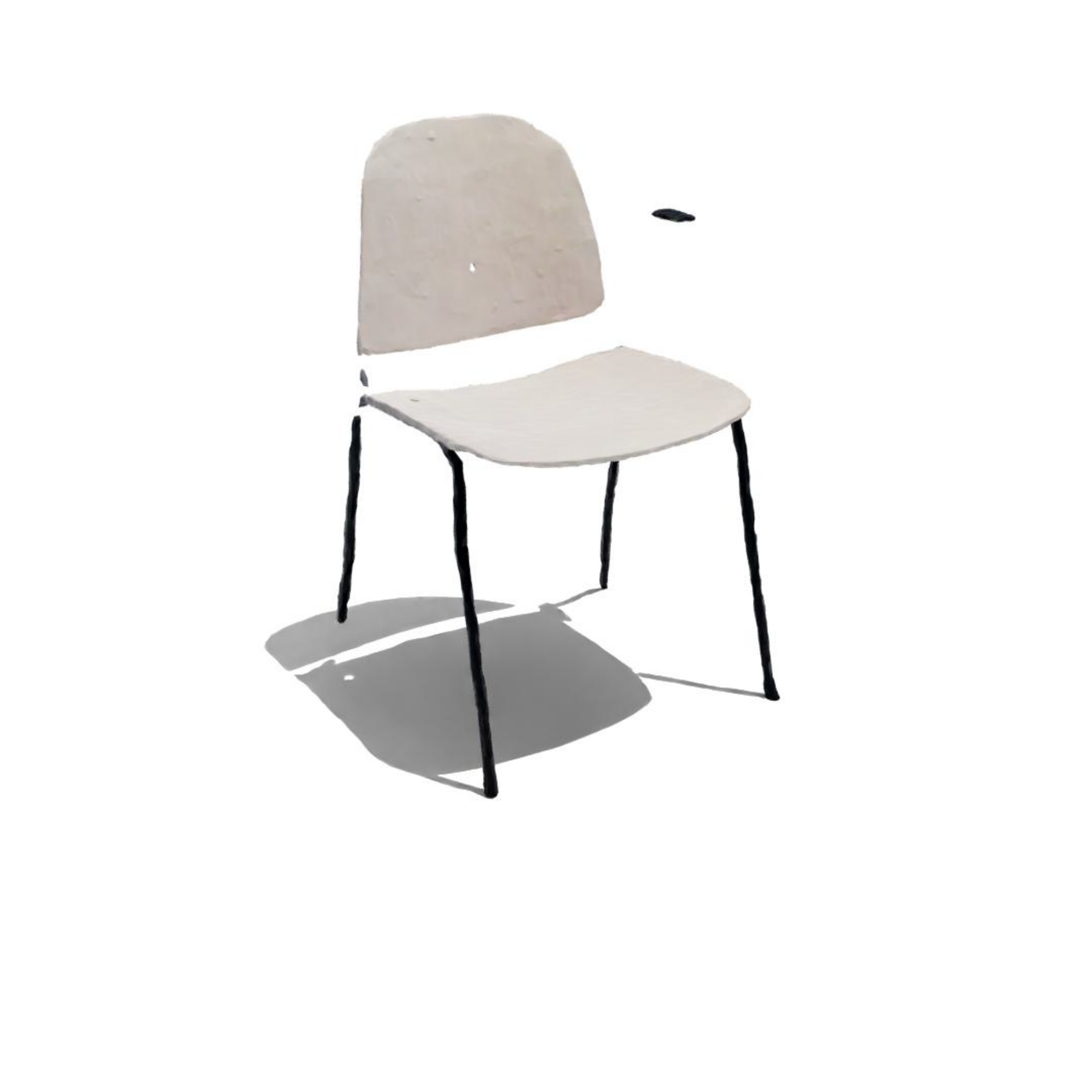}\includegraphics[width=0.08333333333333333\linewidth]{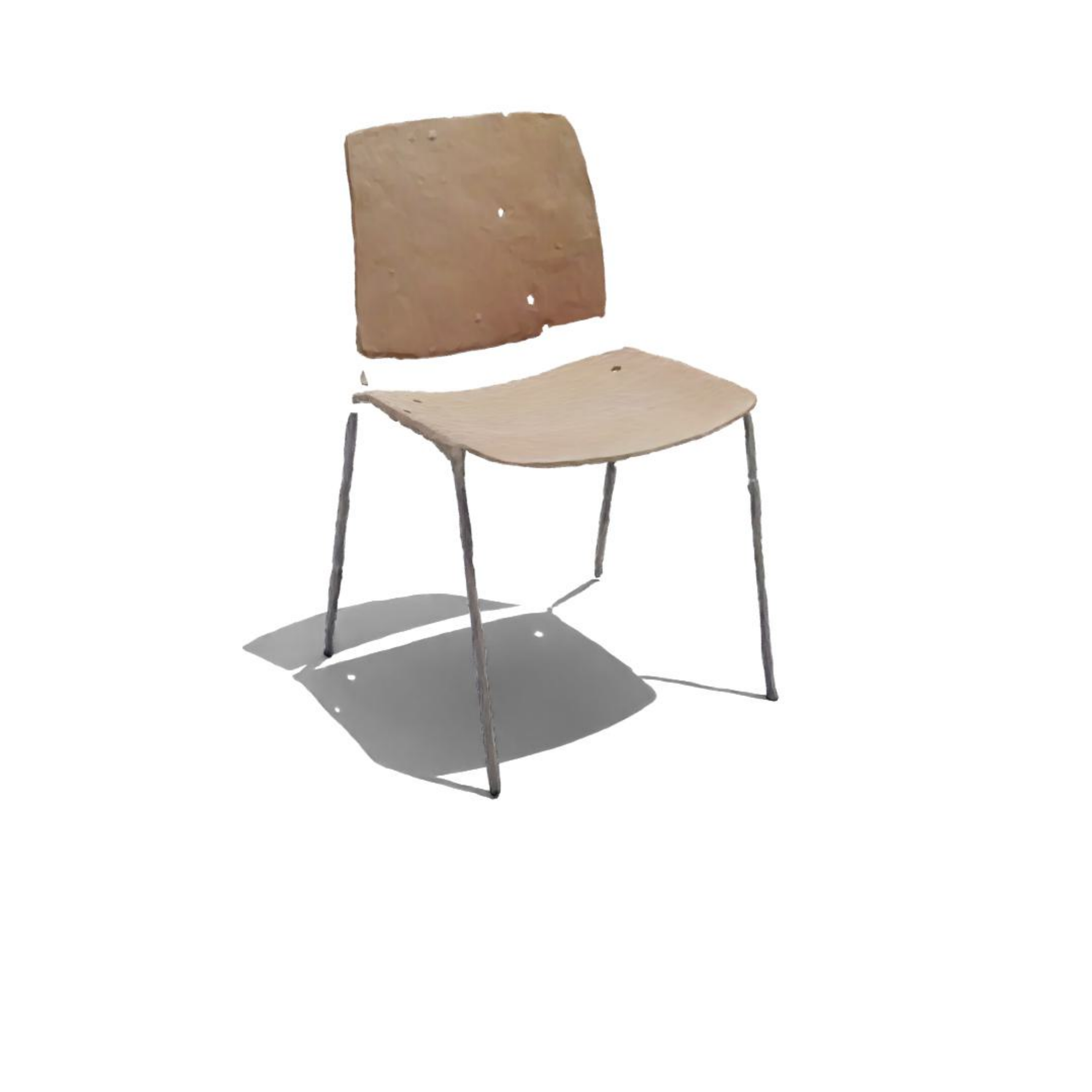}\includegraphics[width=0.08333333333333333\linewidth]{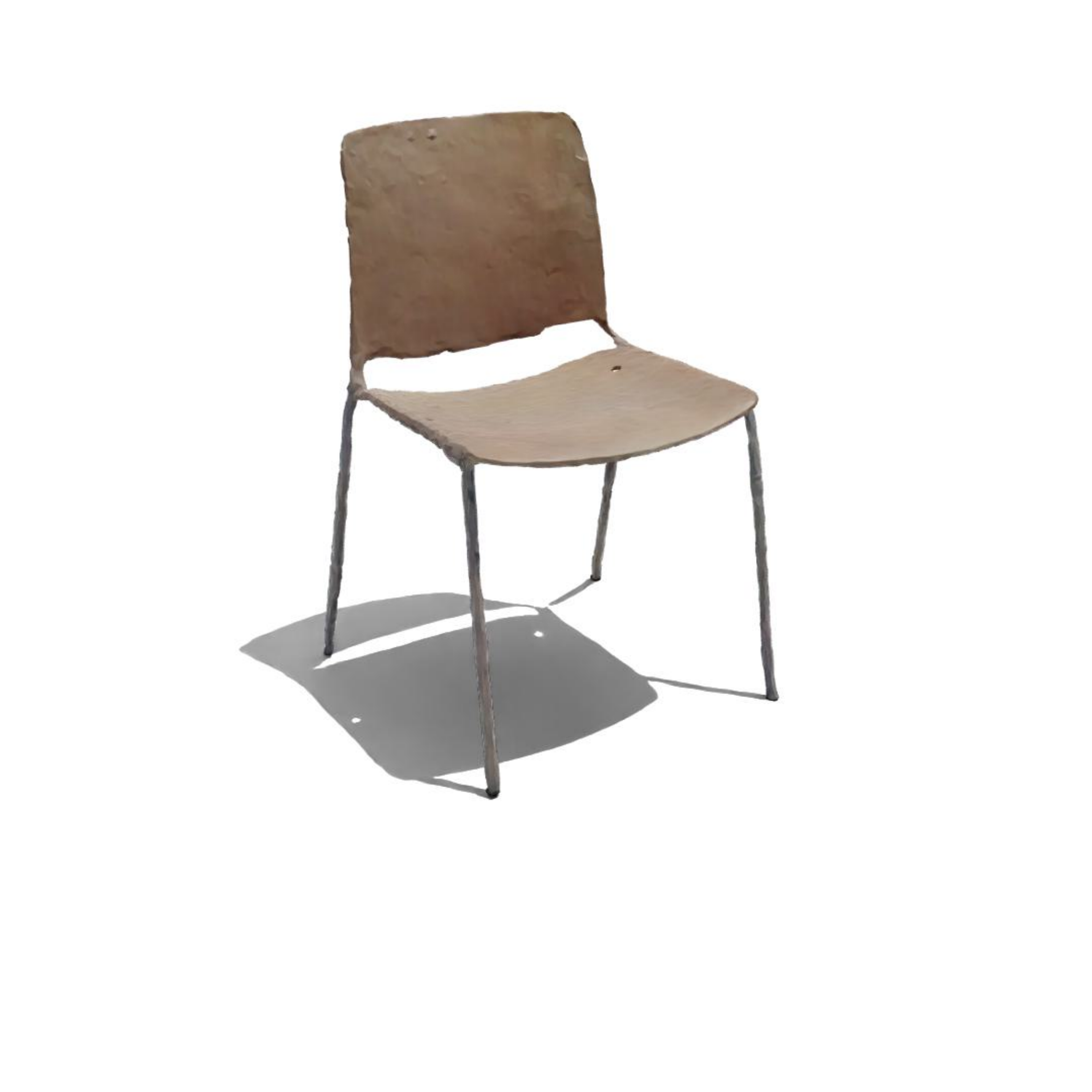}\includegraphics[width=0.08333333333333333\linewidth]{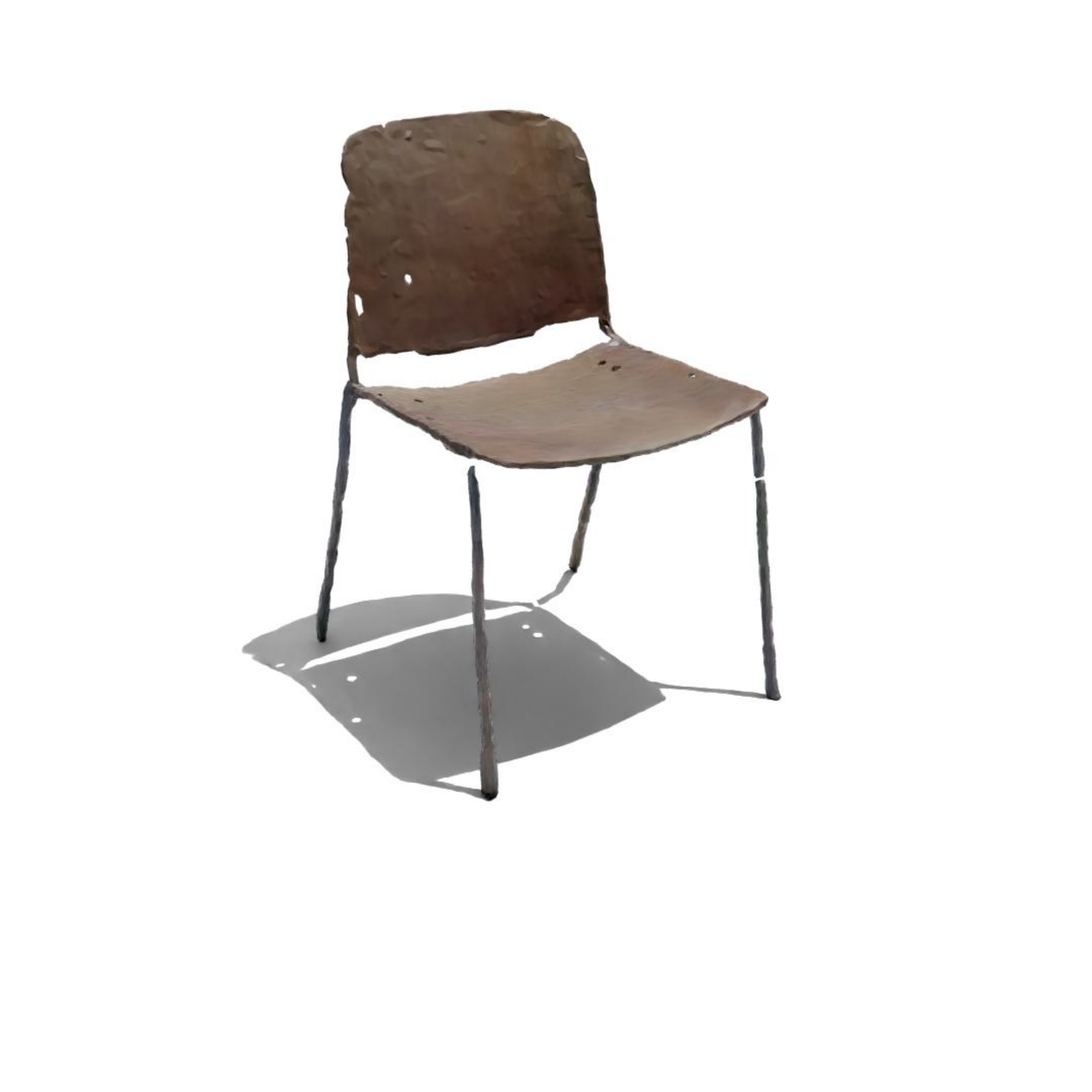}\\
\includegraphics[width=0.08333333333333333\linewidth]{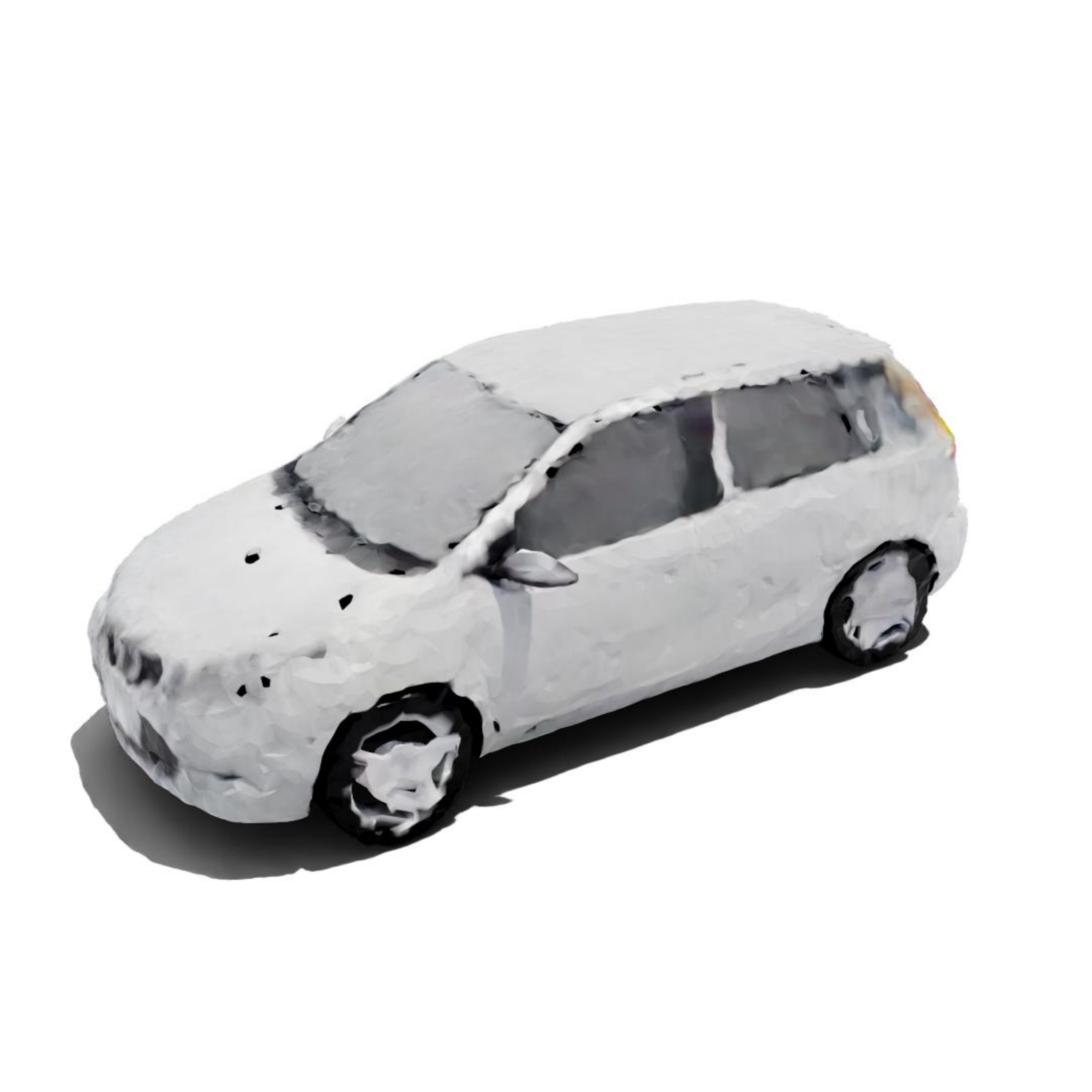}\includegraphics[width=0.08333333333333333\linewidth]{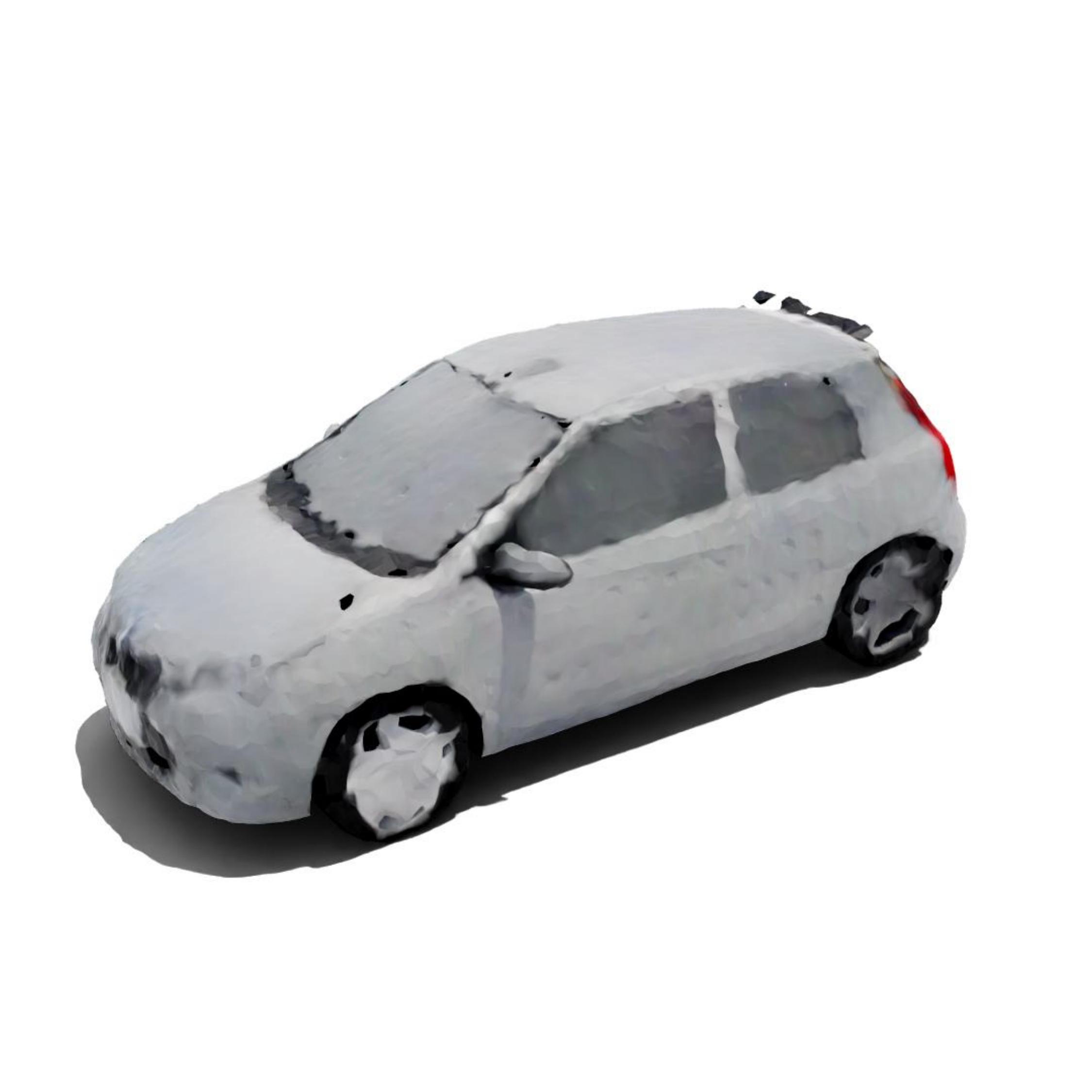}\includegraphics[width=0.08333333333333333\linewidth]{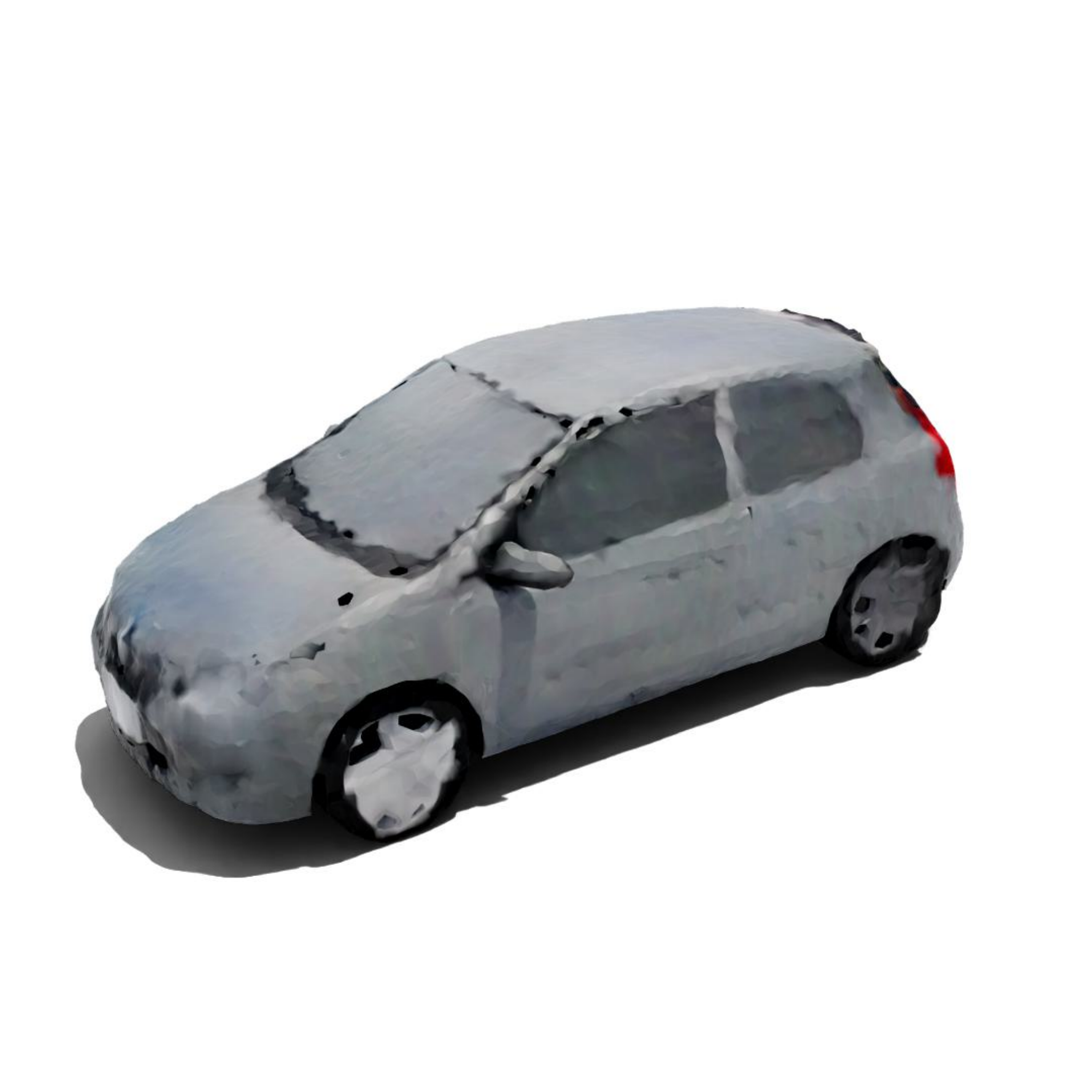}\includegraphics[width=0.08333333333333333\linewidth]{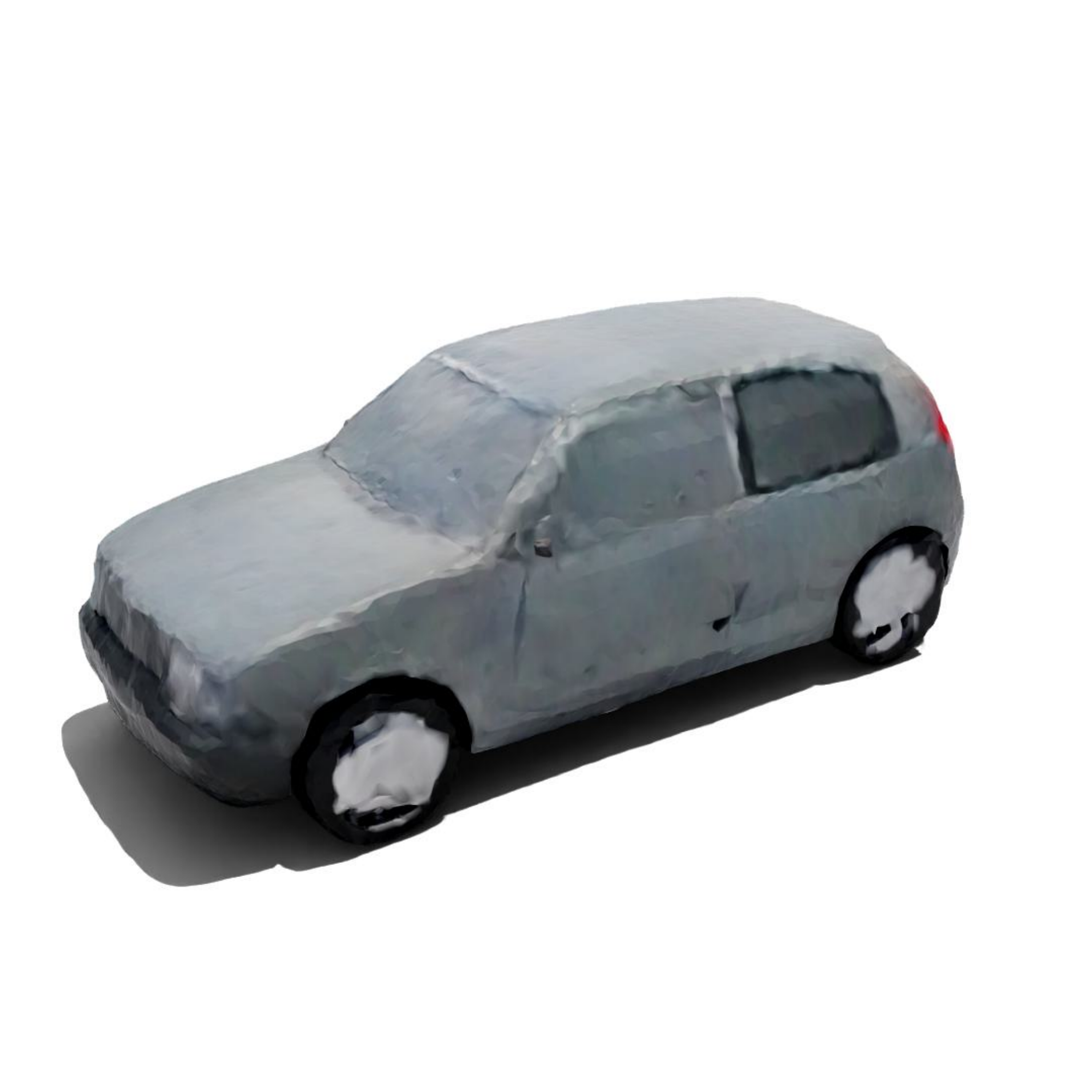}\includegraphics[width=0.08333333333333333\linewidth]{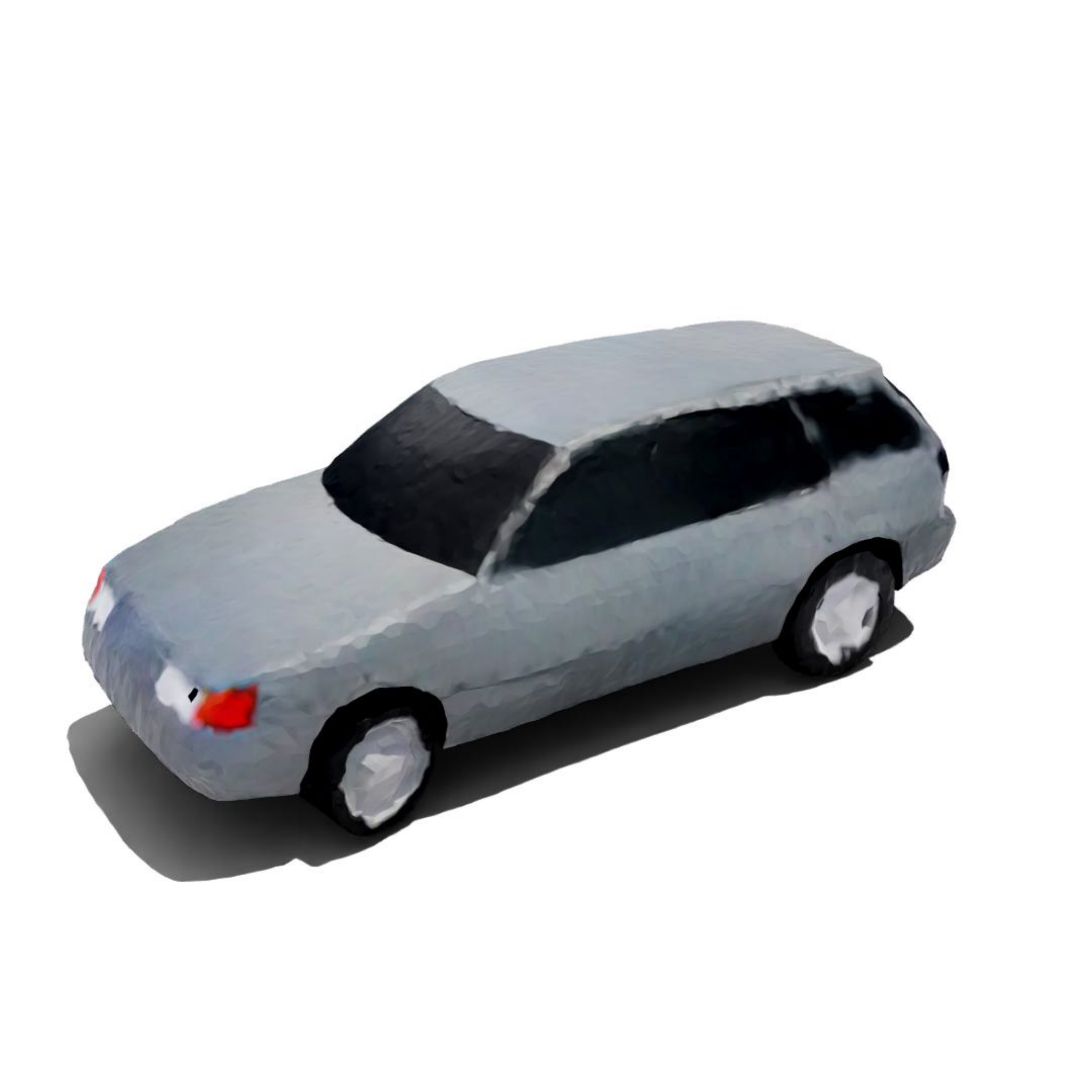}\includegraphics[width=0.08333333333333333\linewidth]{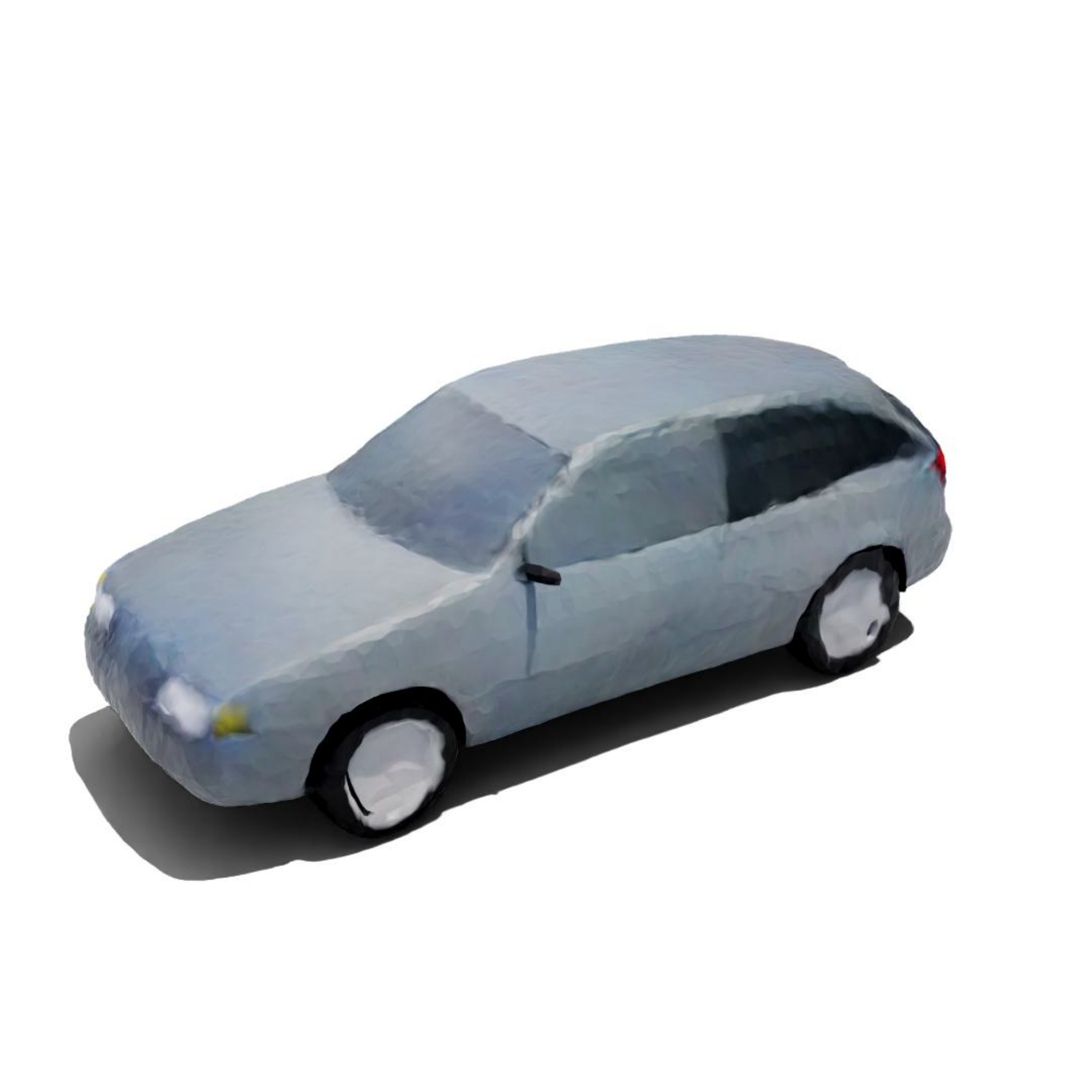}\includegraphics[width=0.08333333333333333\linewidth]{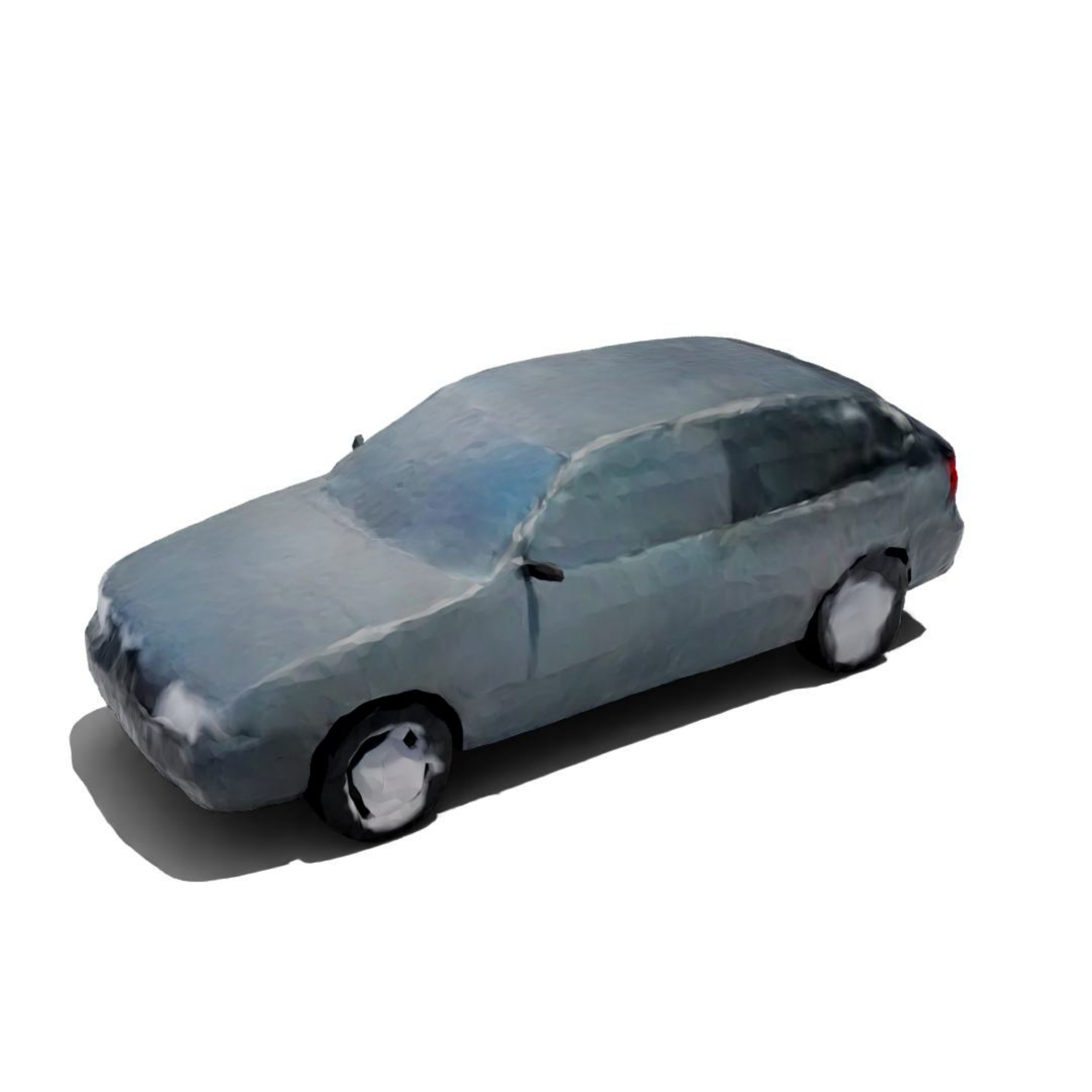}\includegraphics[width=0.08333333333333333\linewidth]{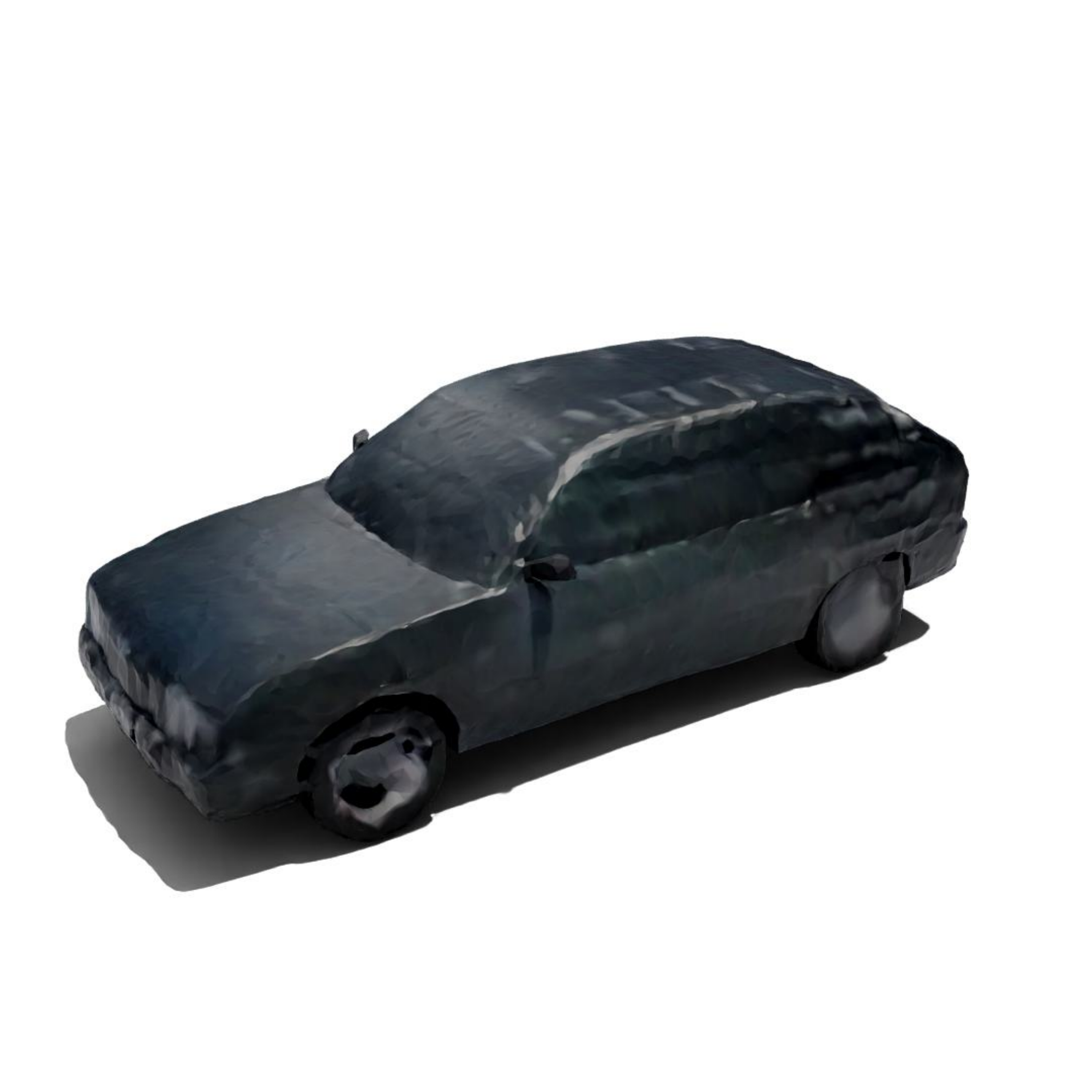}\includegraphics[width=0.08333333333333333\linewidth]{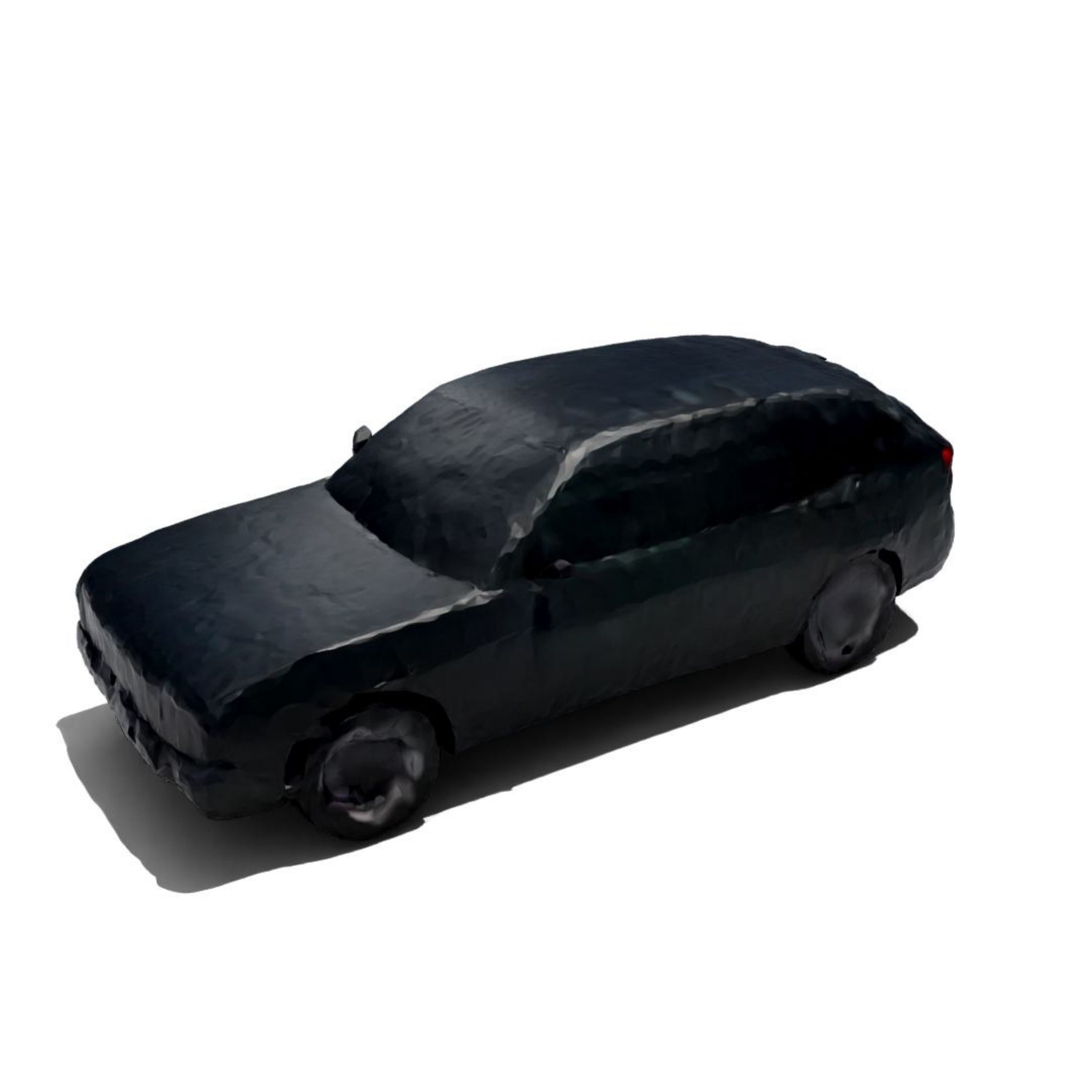}\includegraphics[width=0.08333333333333333\linewidth]{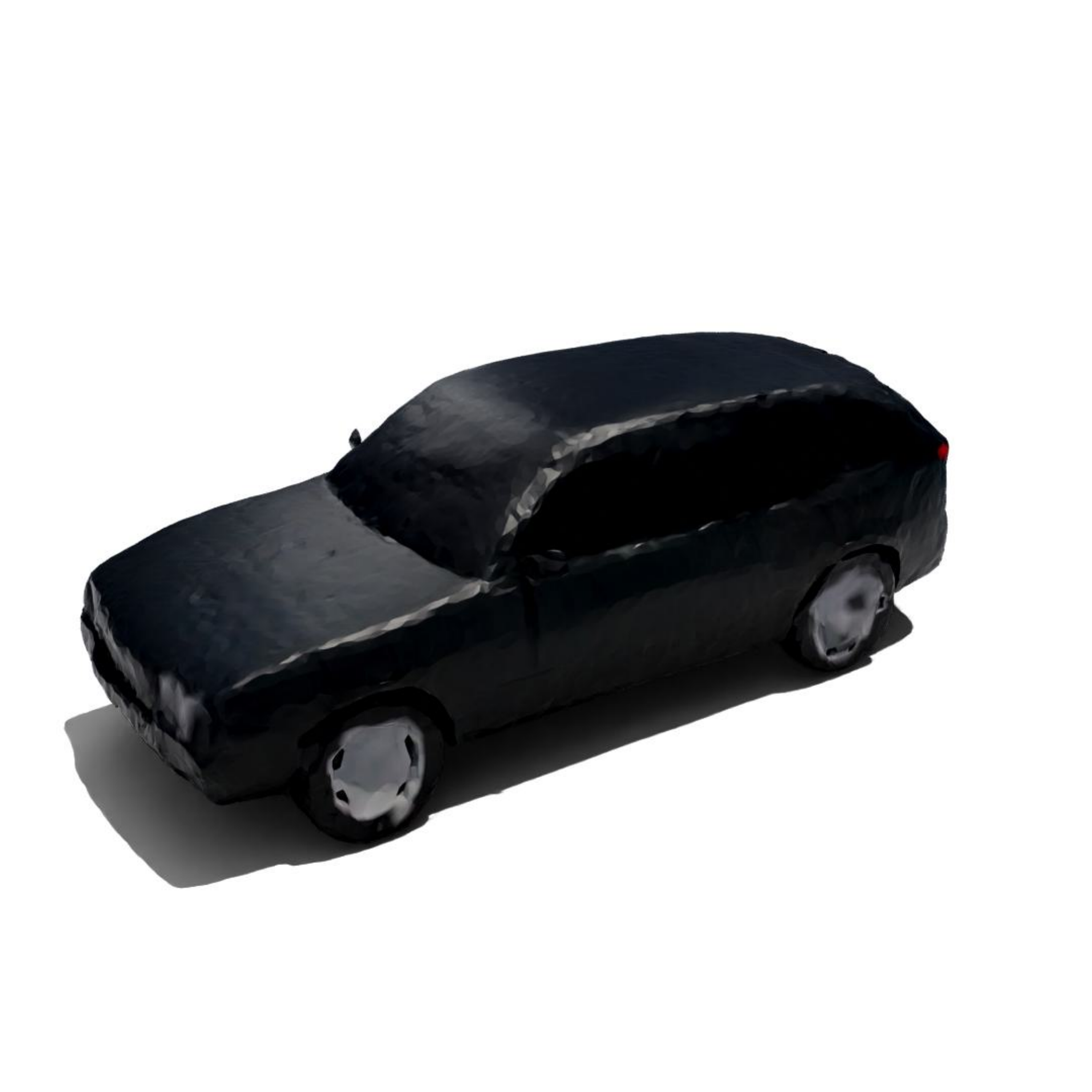}\includegraphics[width=0.08333333333333333\linewidth]{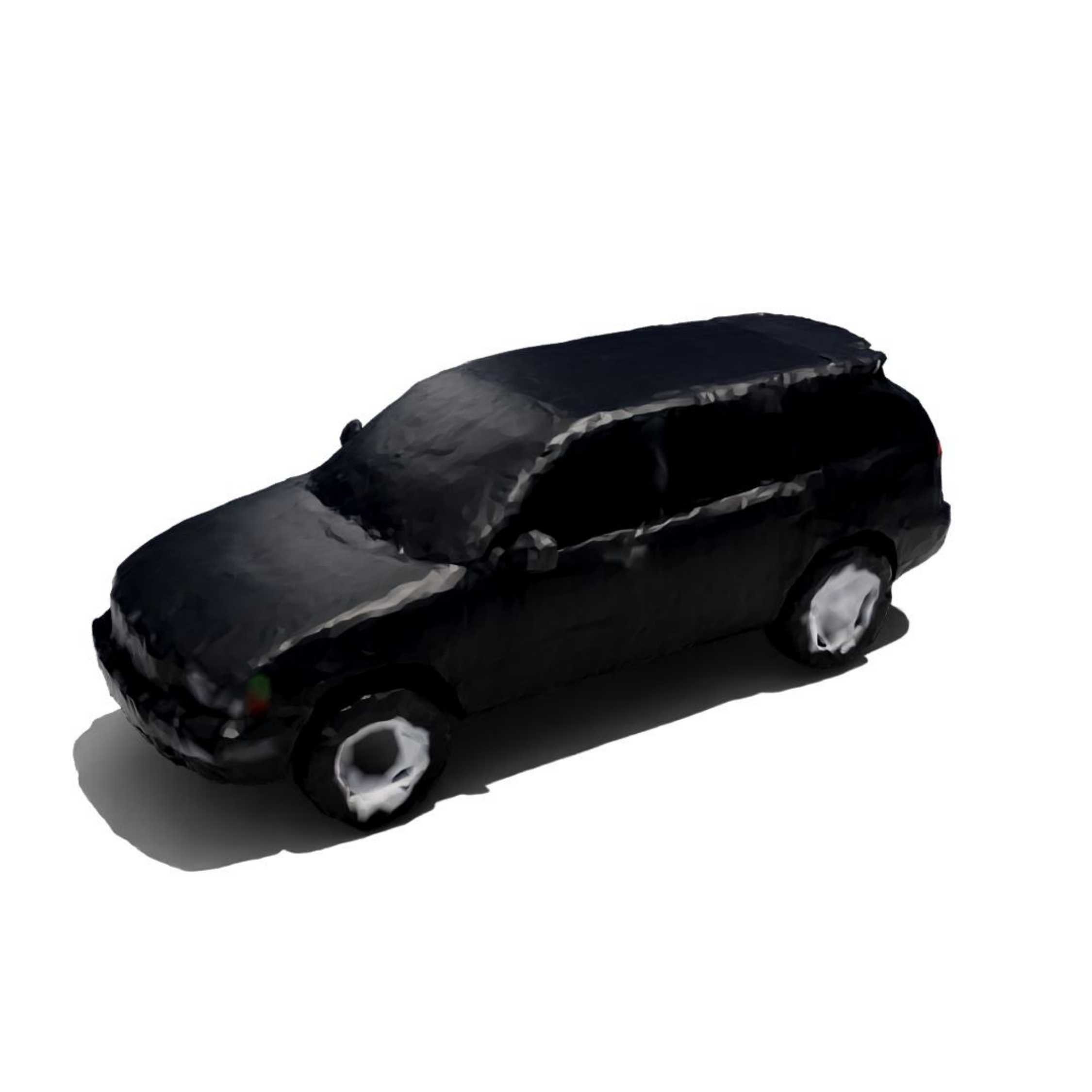}\includegraphics[width=0.08333333333333333\linewidth]{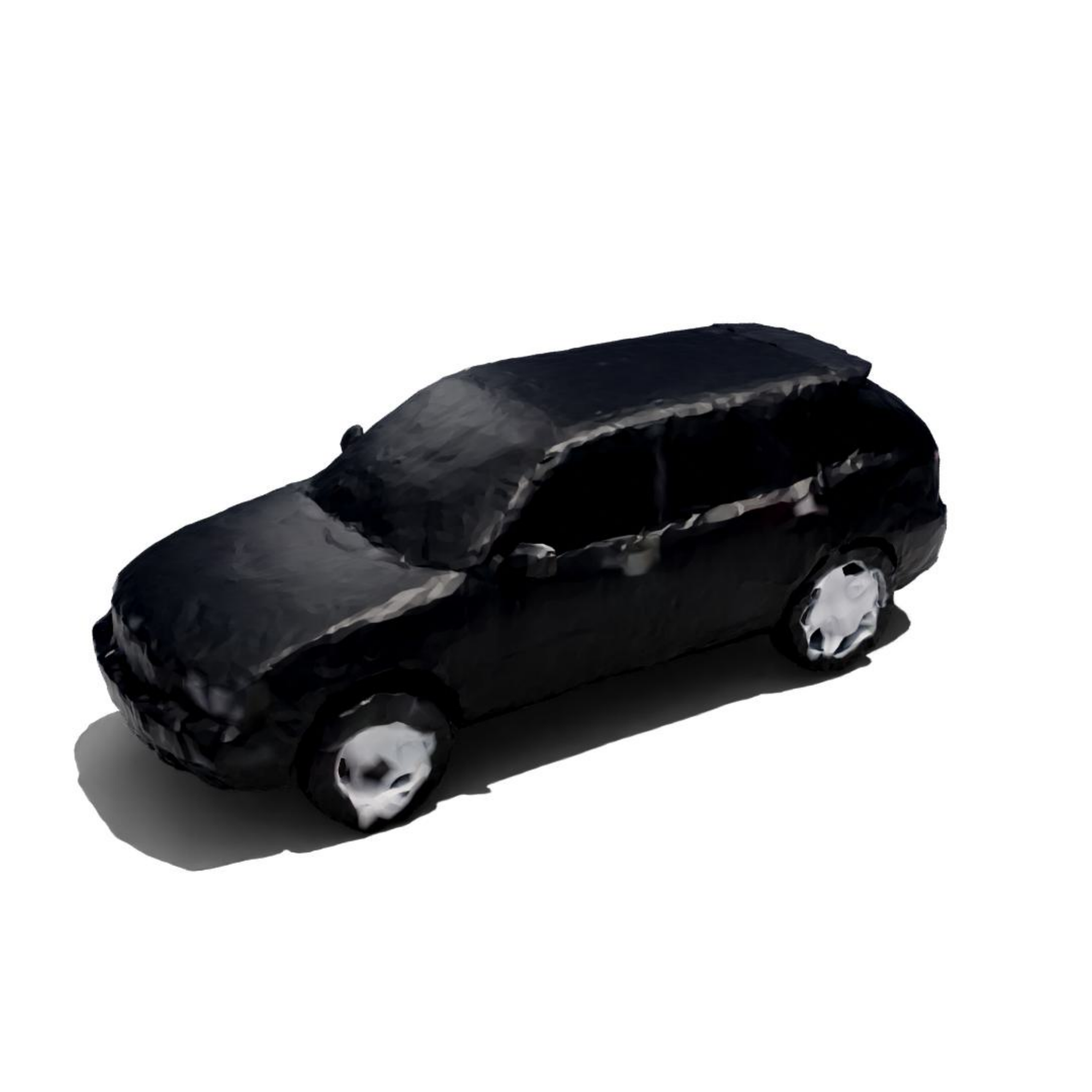}\\
\includegraphics[width=0.08333333333333333\linewidth]{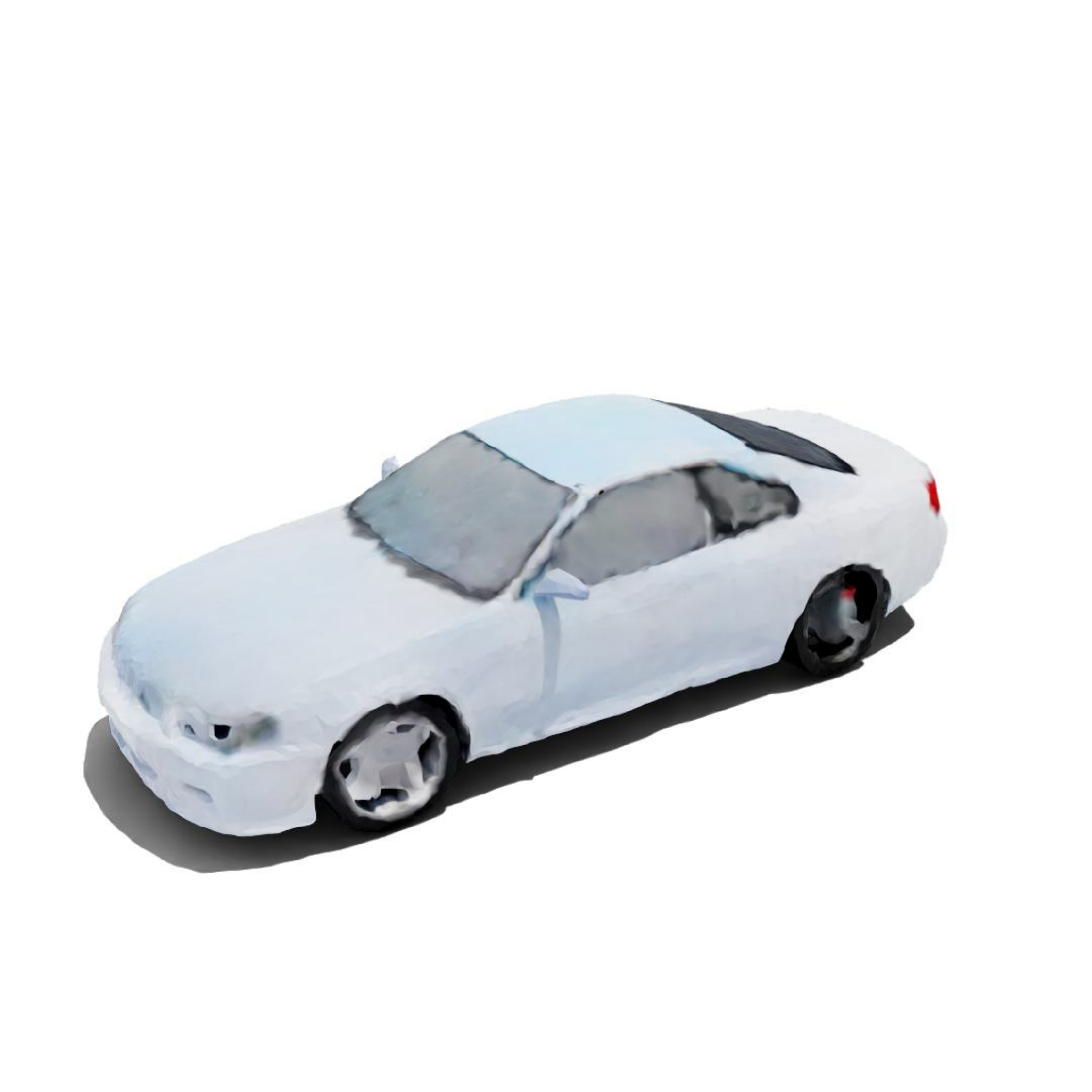}\includegraphics[width=0.08333333333333333\linewidth]{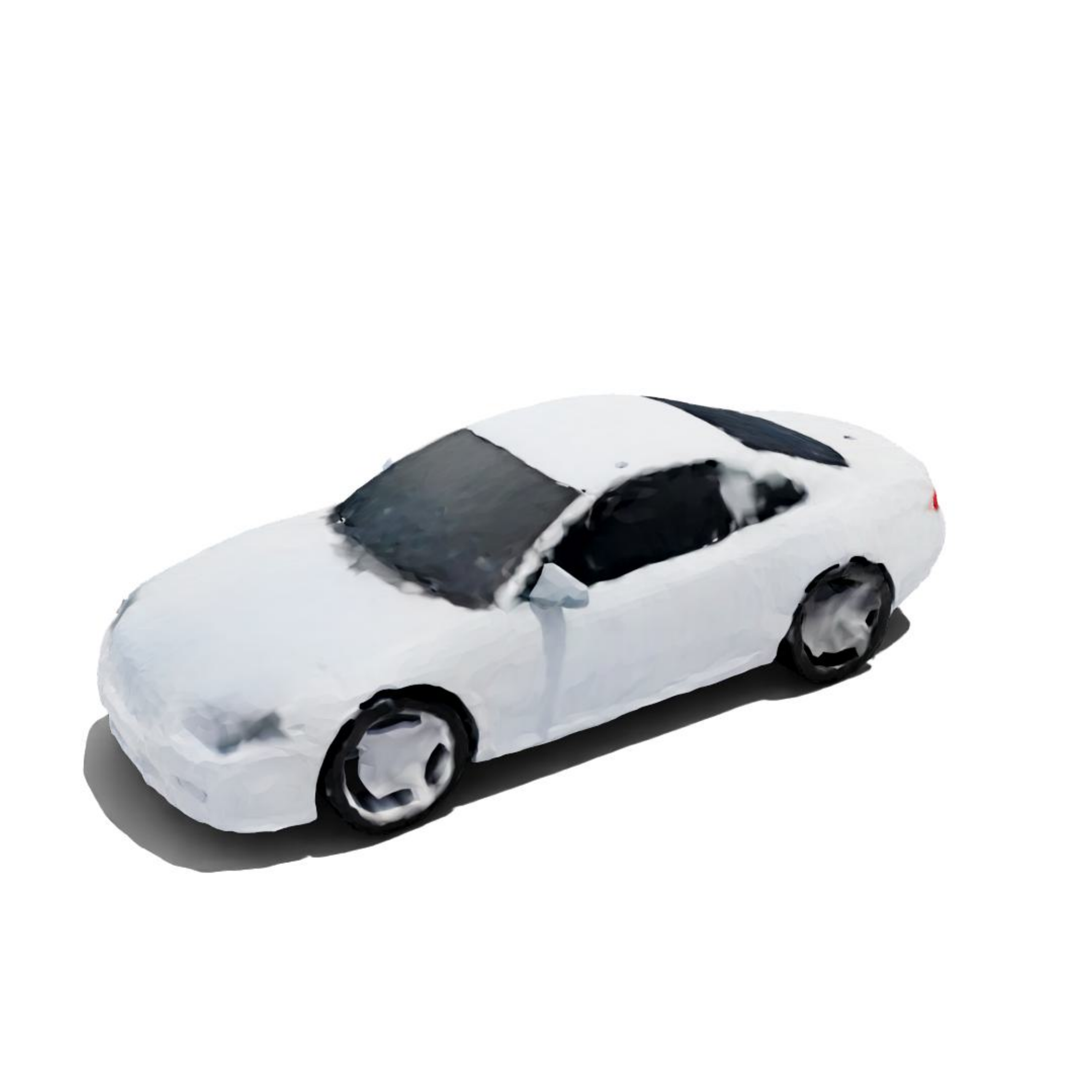}\includegraphics[width=0.08333333333333333\linewidth]{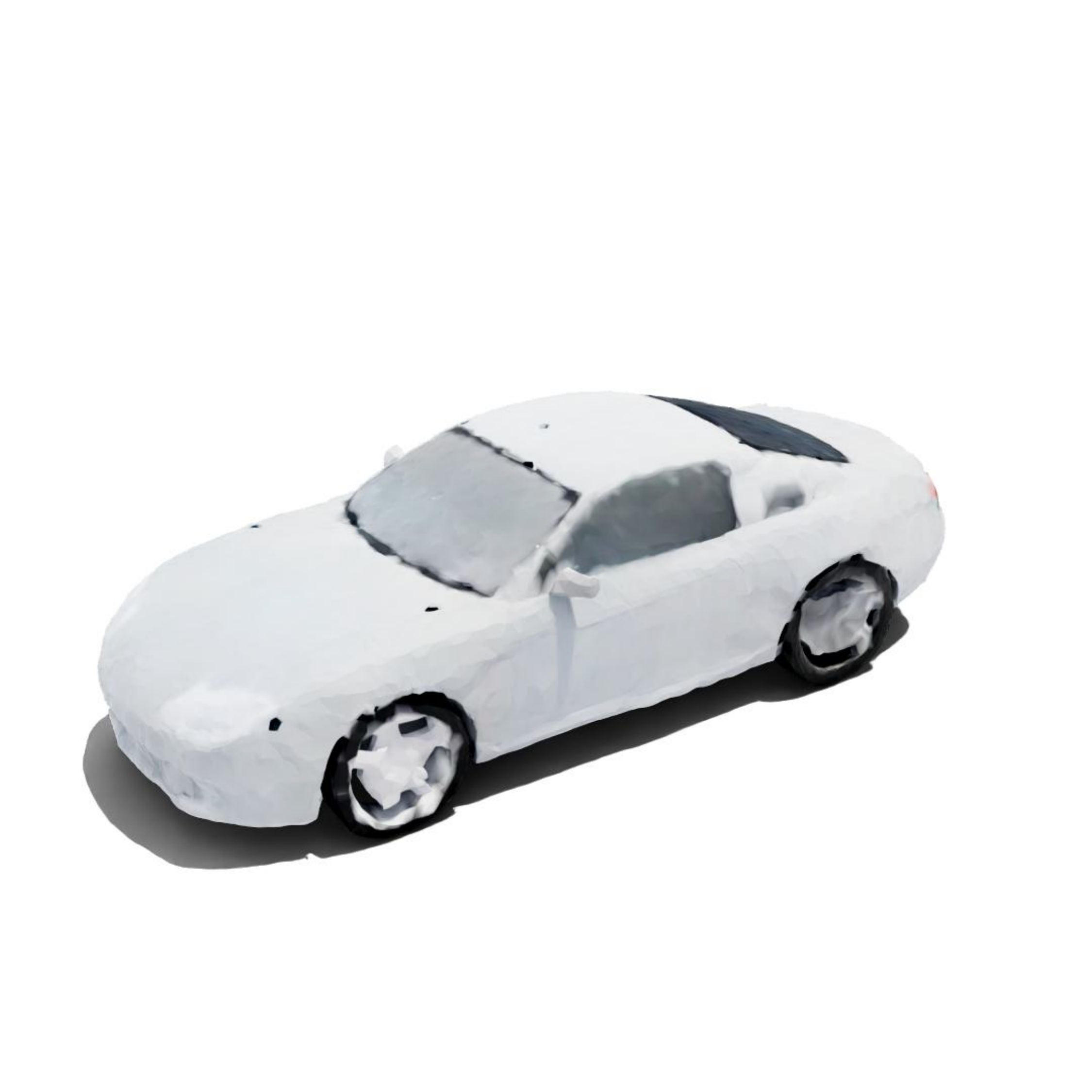}\includegraphics[width=0.08333333333333333\linewidth]{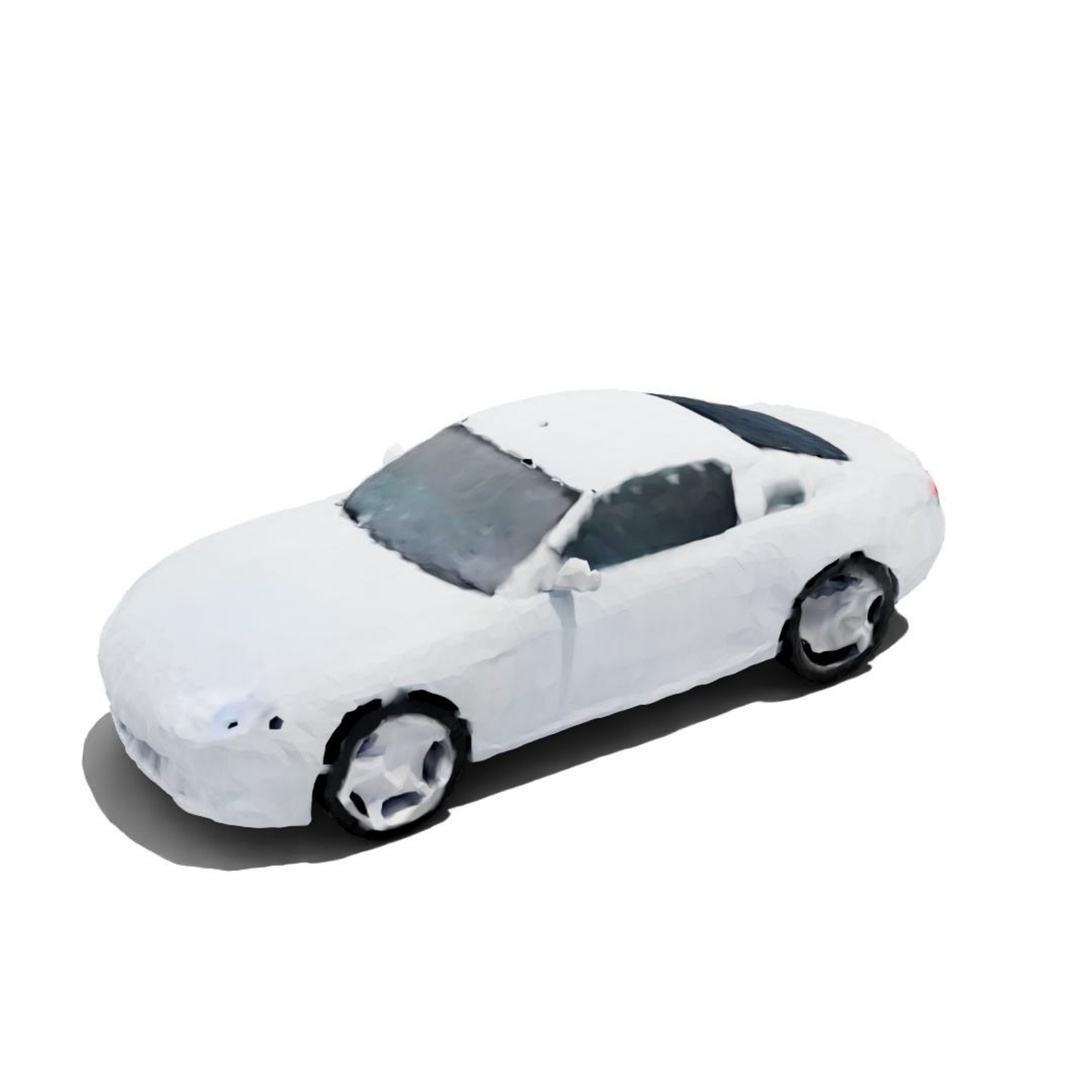}\includegraphics[width=0.08333333333333333\linewidth]{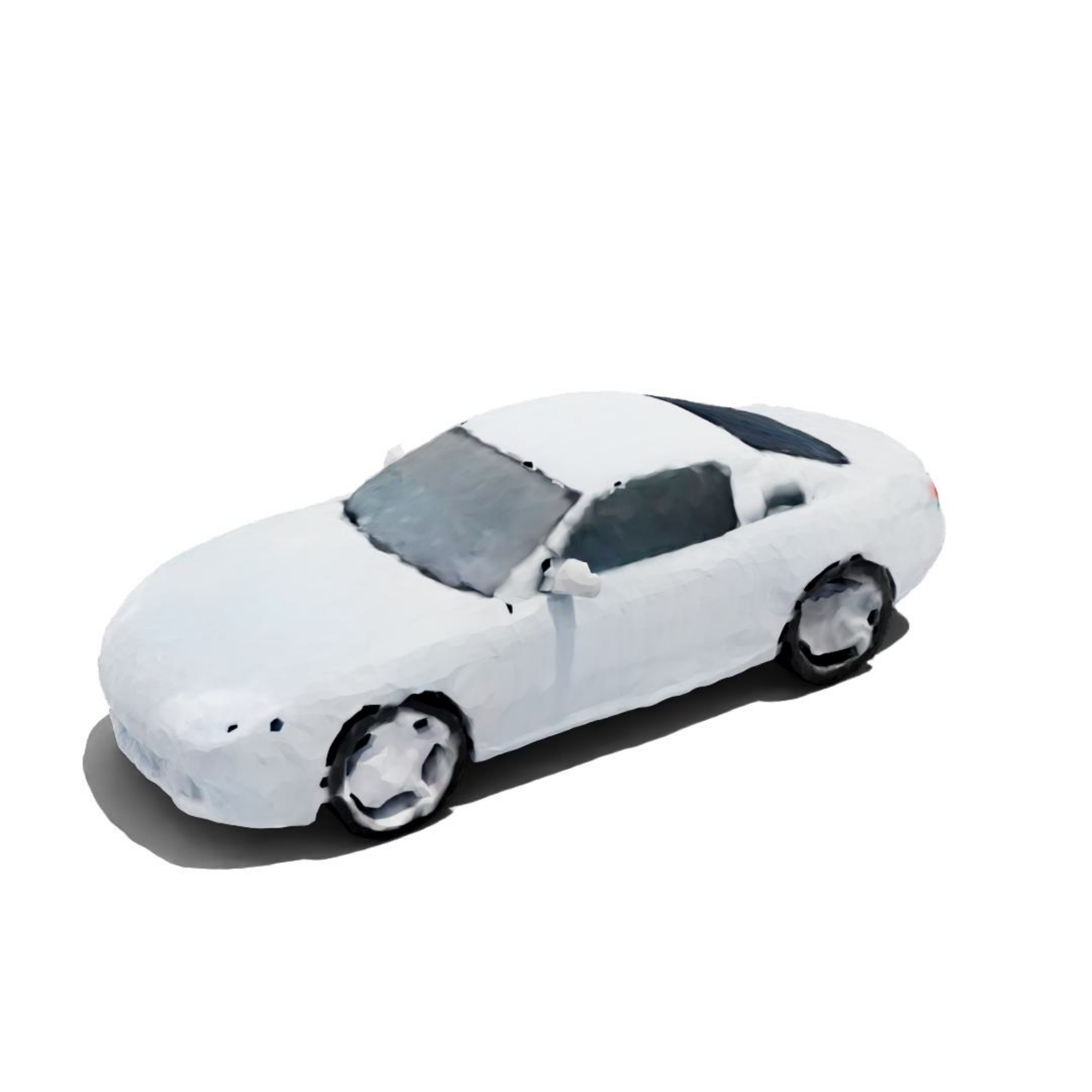}\includegraphics[width=0.08333333333333333\linewidth]{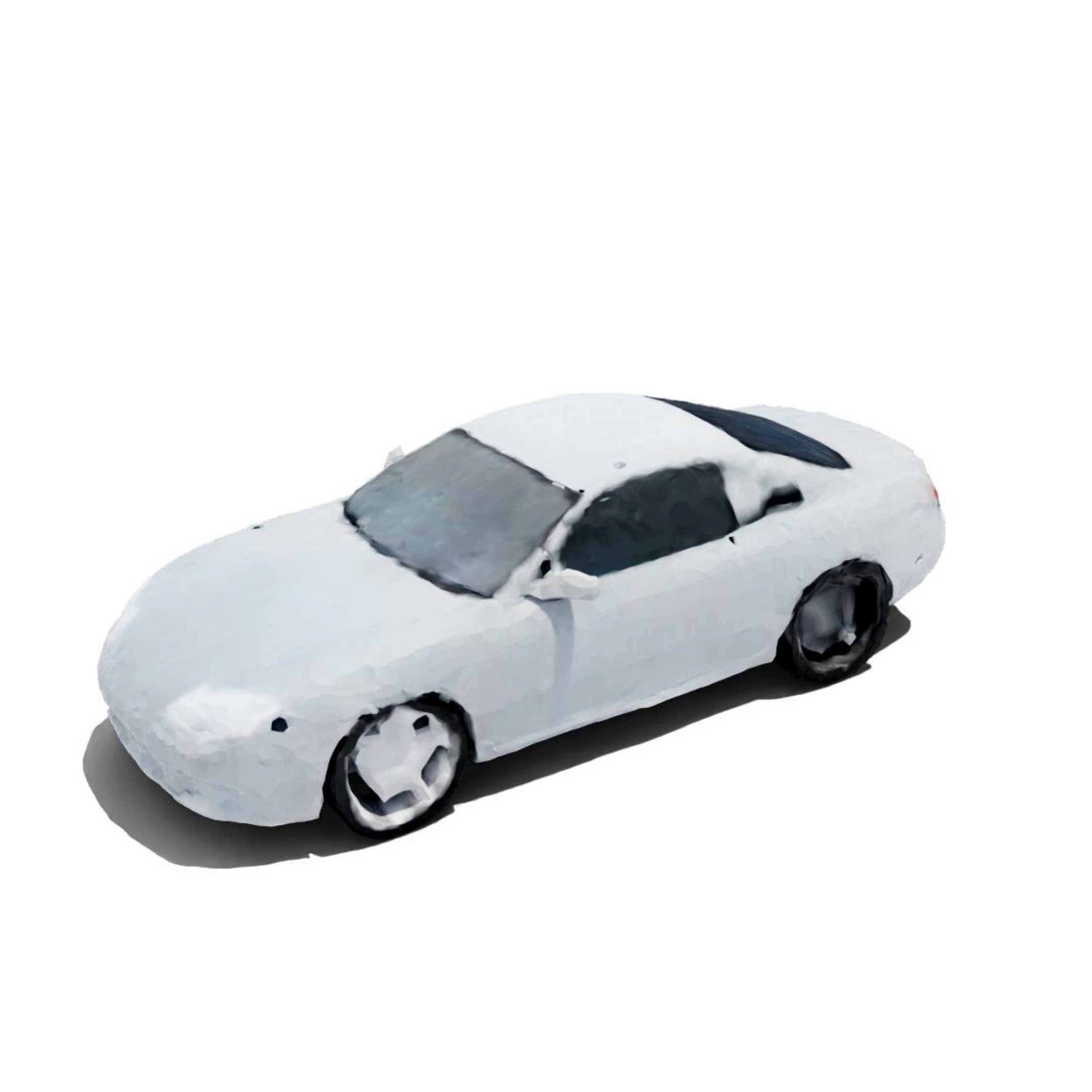}\includegraphics[width=0.08333333333333333\linewidth]{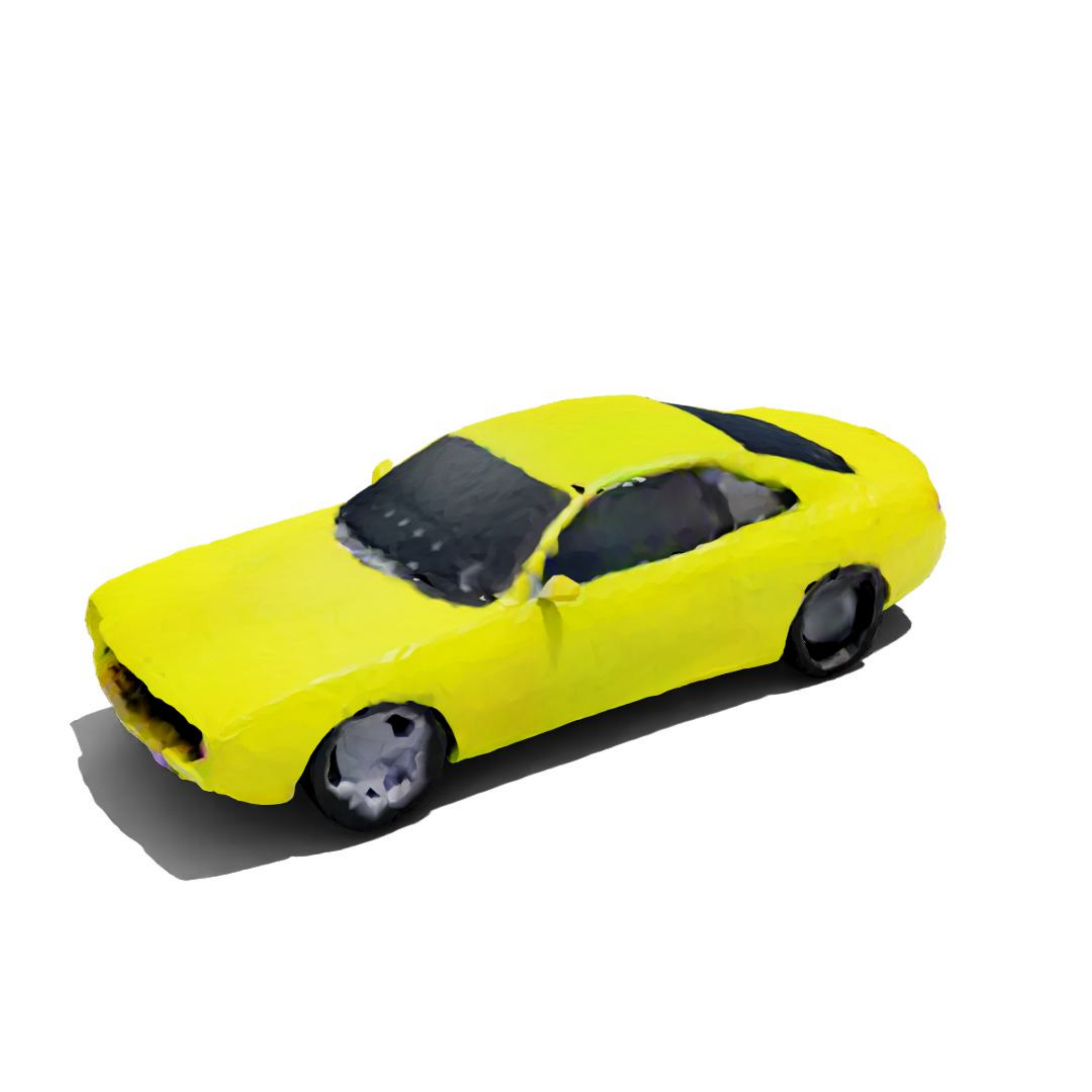}\includegraphics[width=0.08333333333333333\linewidth]{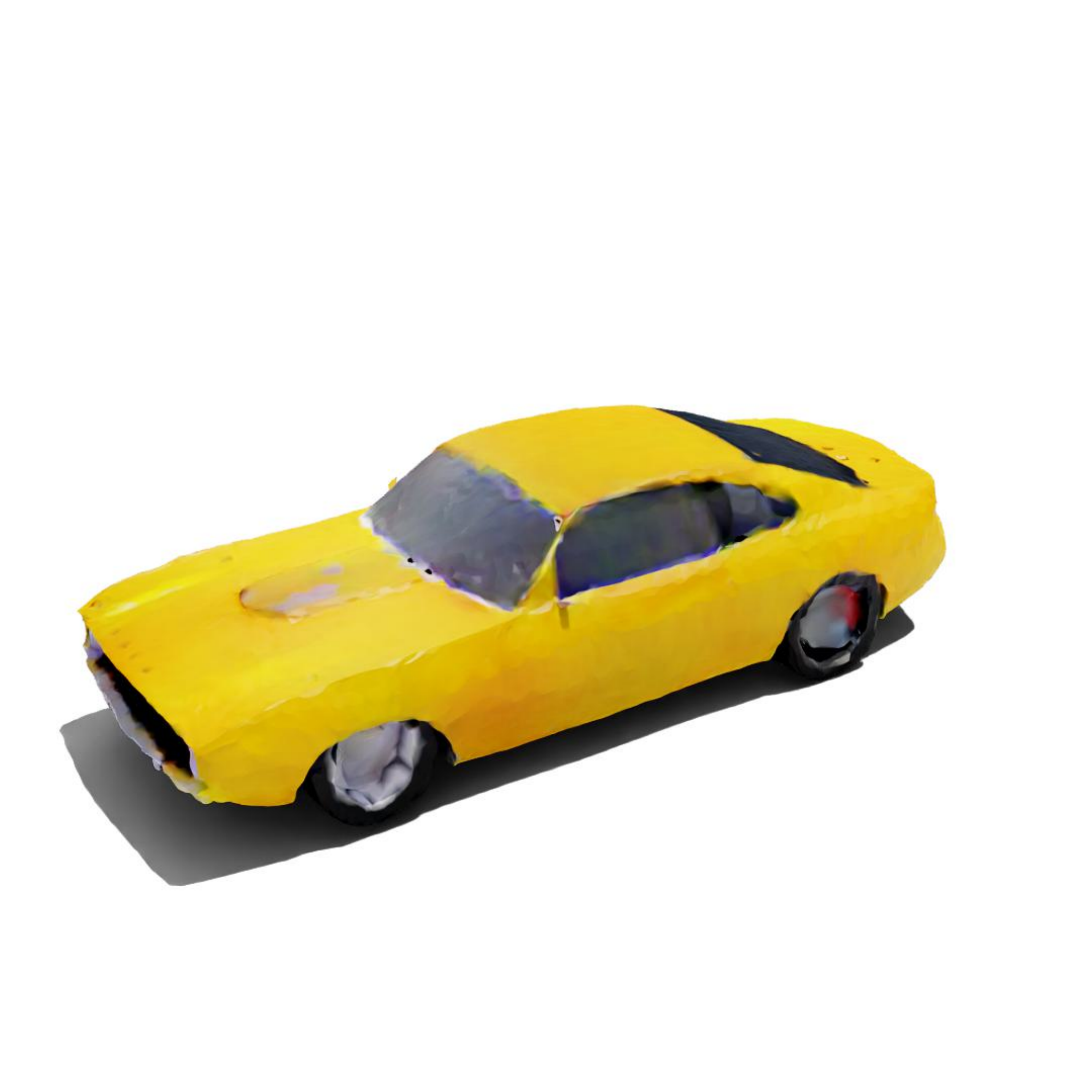}\includegraphics[width=0.08333333333333333\linewidth]{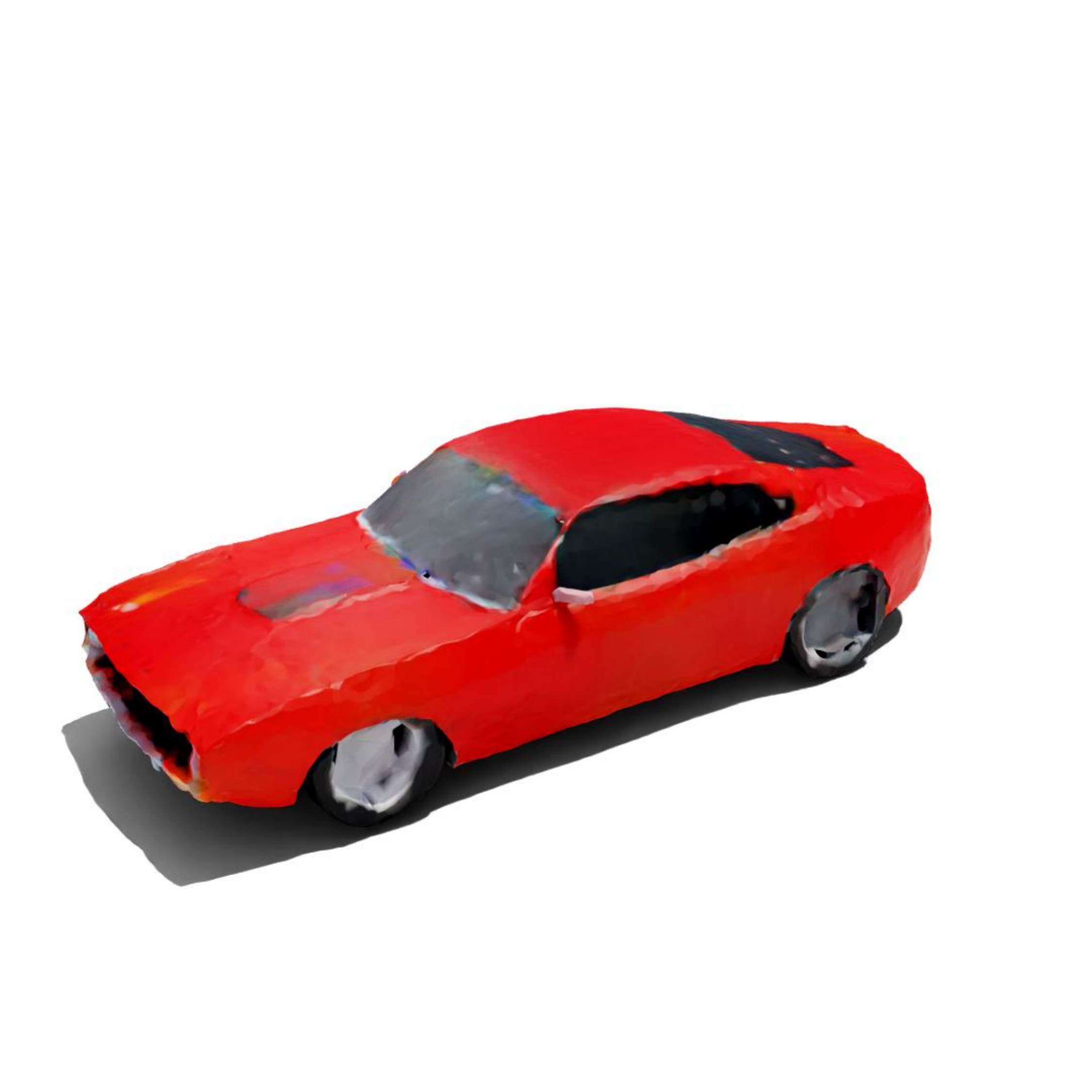}\includegraphics[width=0.08333333333333333\linewidth]{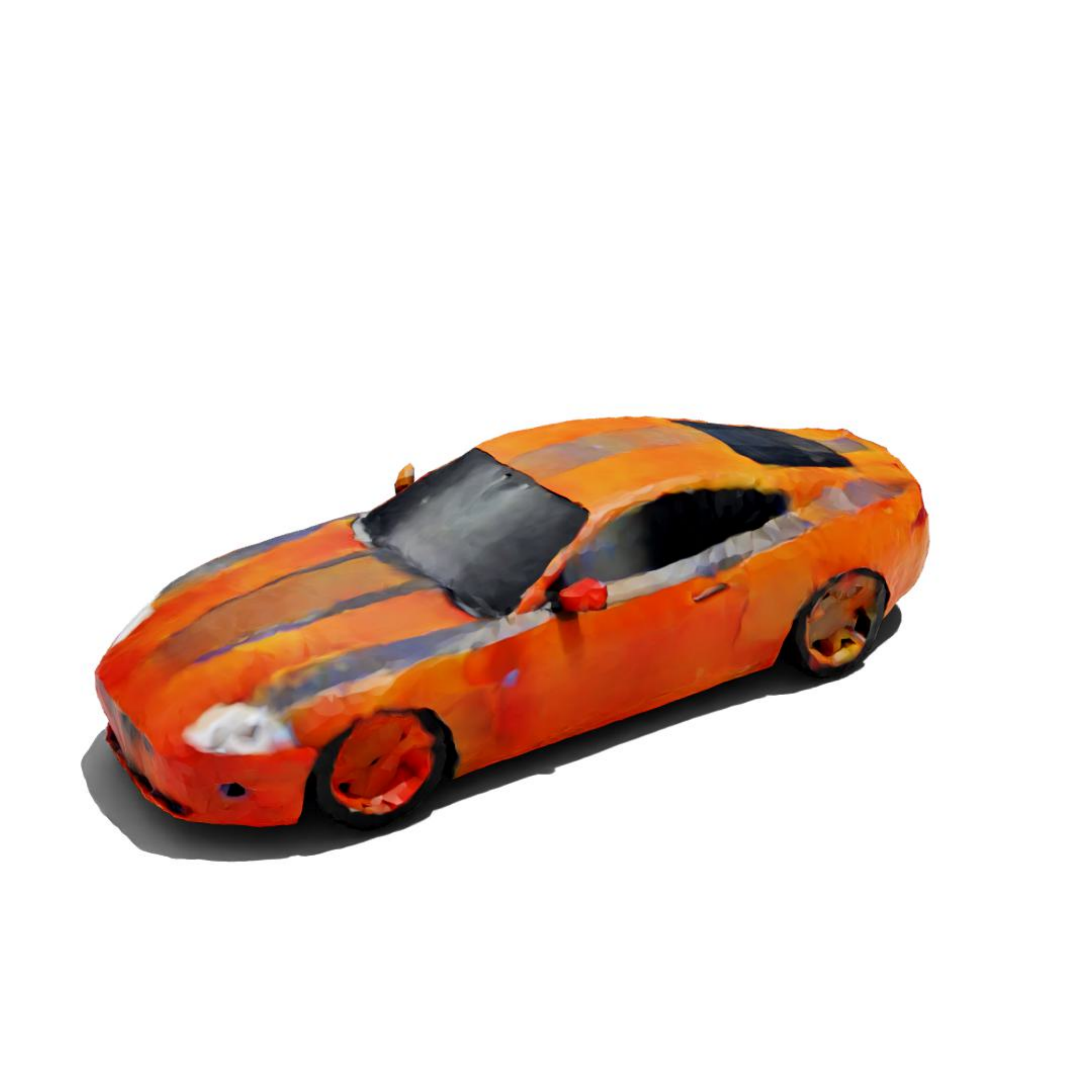}\includegraphics[width=0.08333333333333333\linewidth]{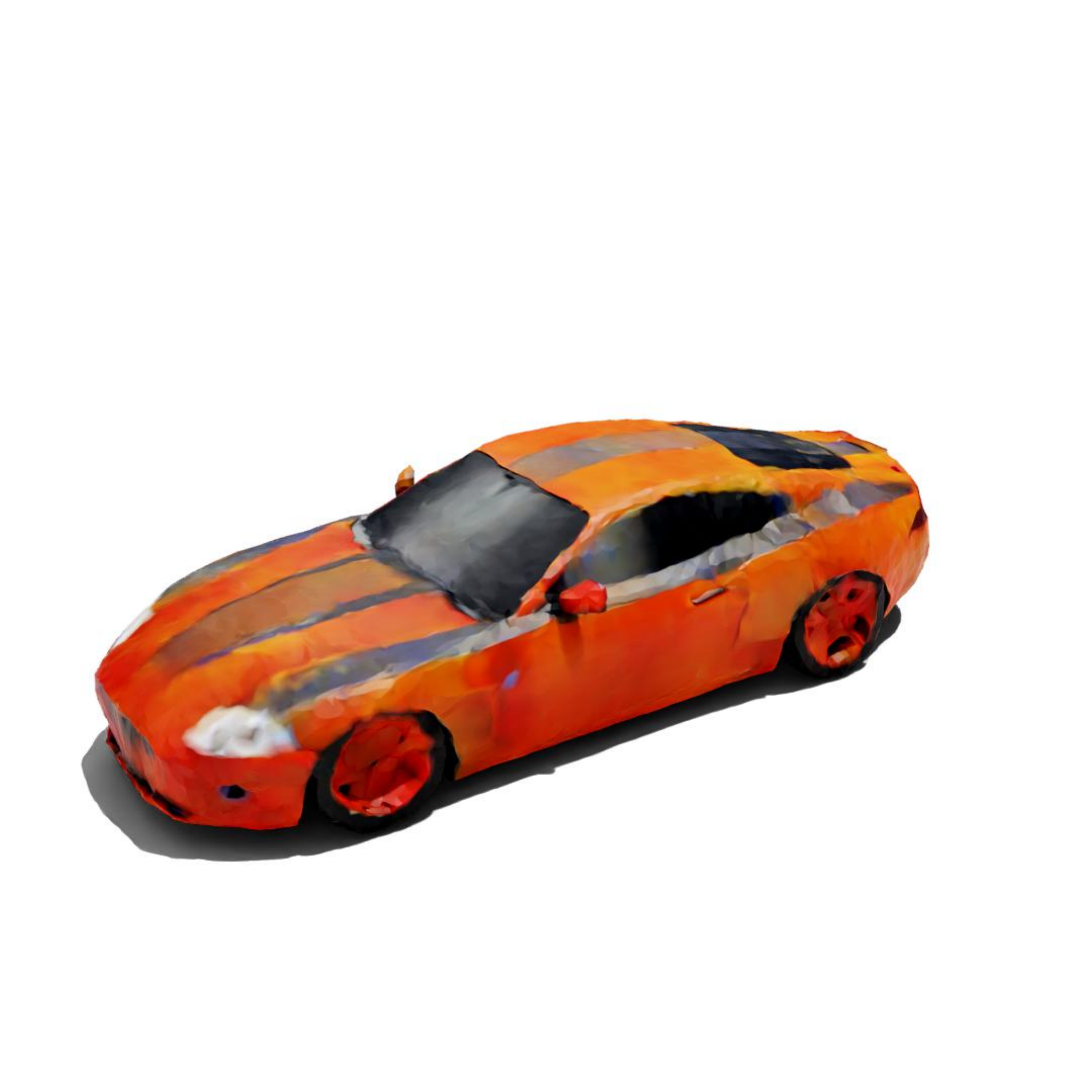}\includegraphics[width=0.08333333333333333\linewidth]{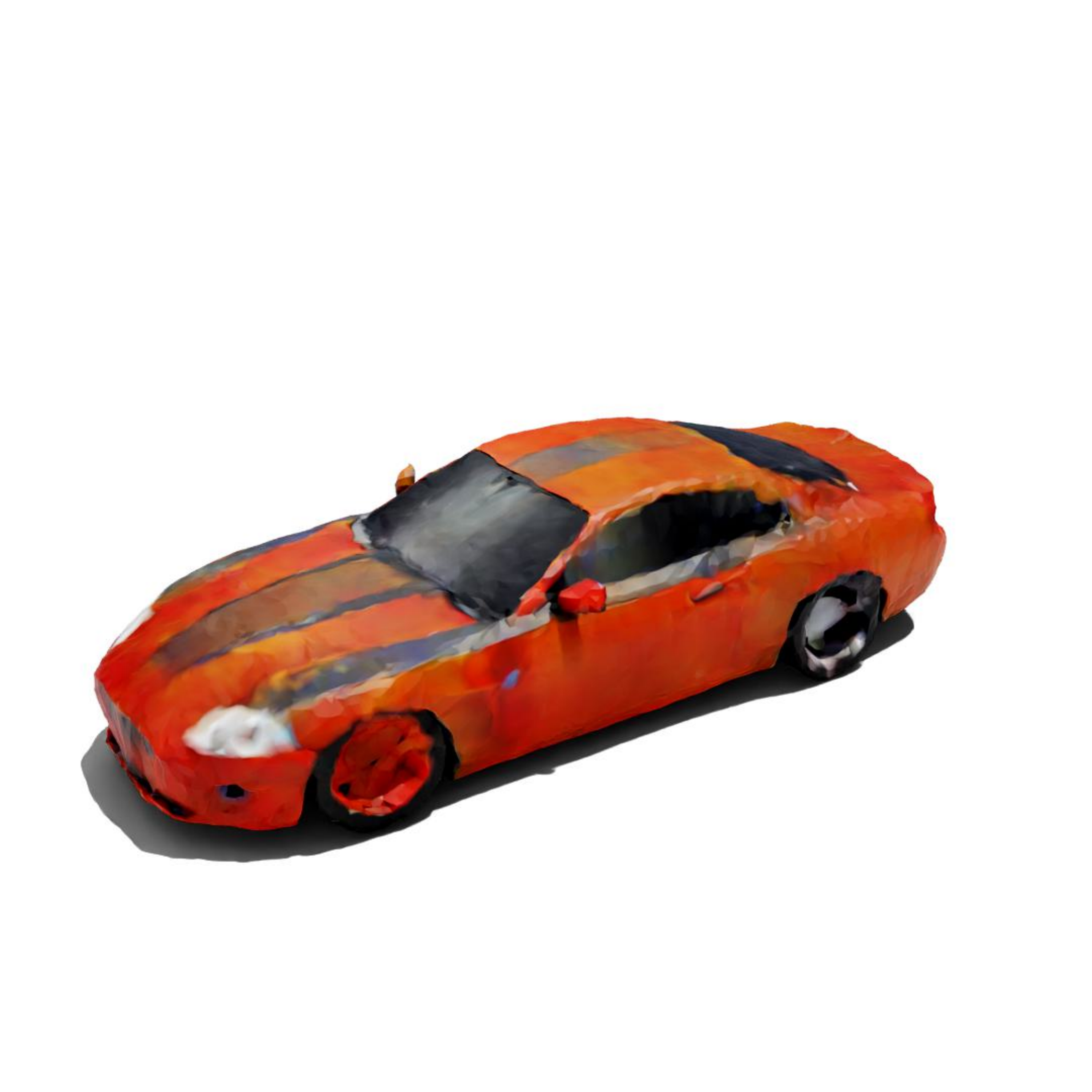}\\
\includegraphics[width=0.08333333333333333\linewidth]{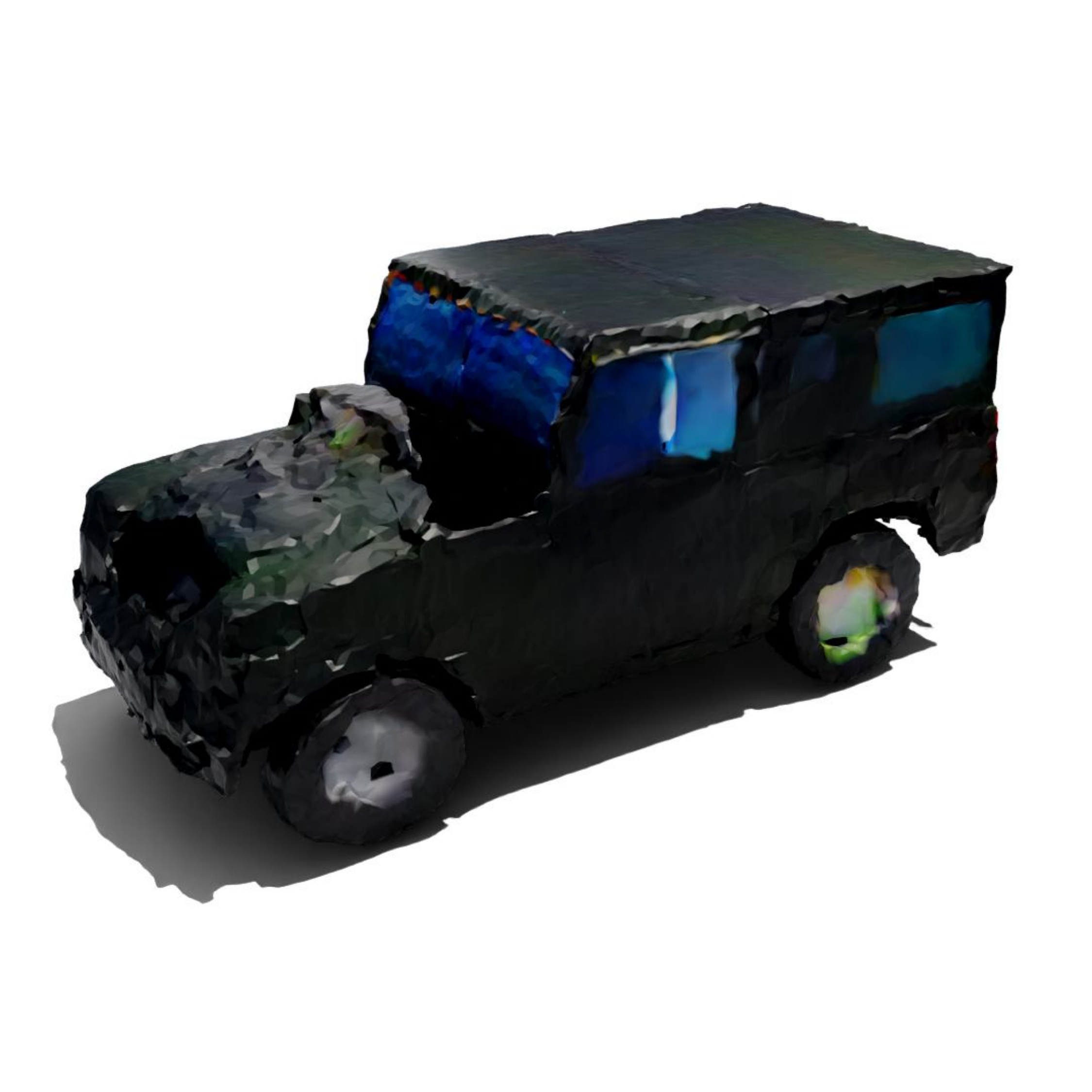}\includegraphics[width=0.08333333333333333\linewidth]{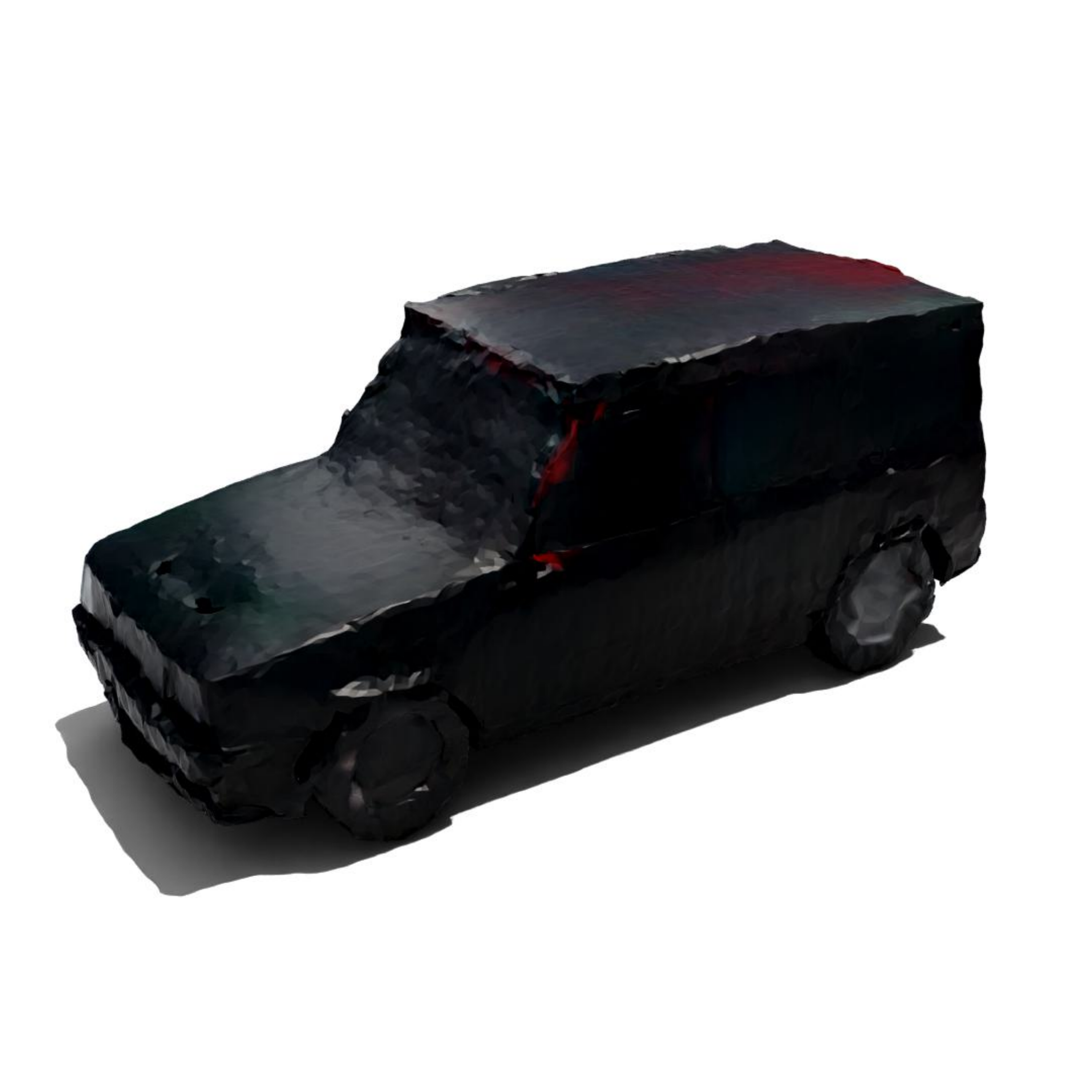}\includegraphics[width=0.08333333333333333\linewidth]{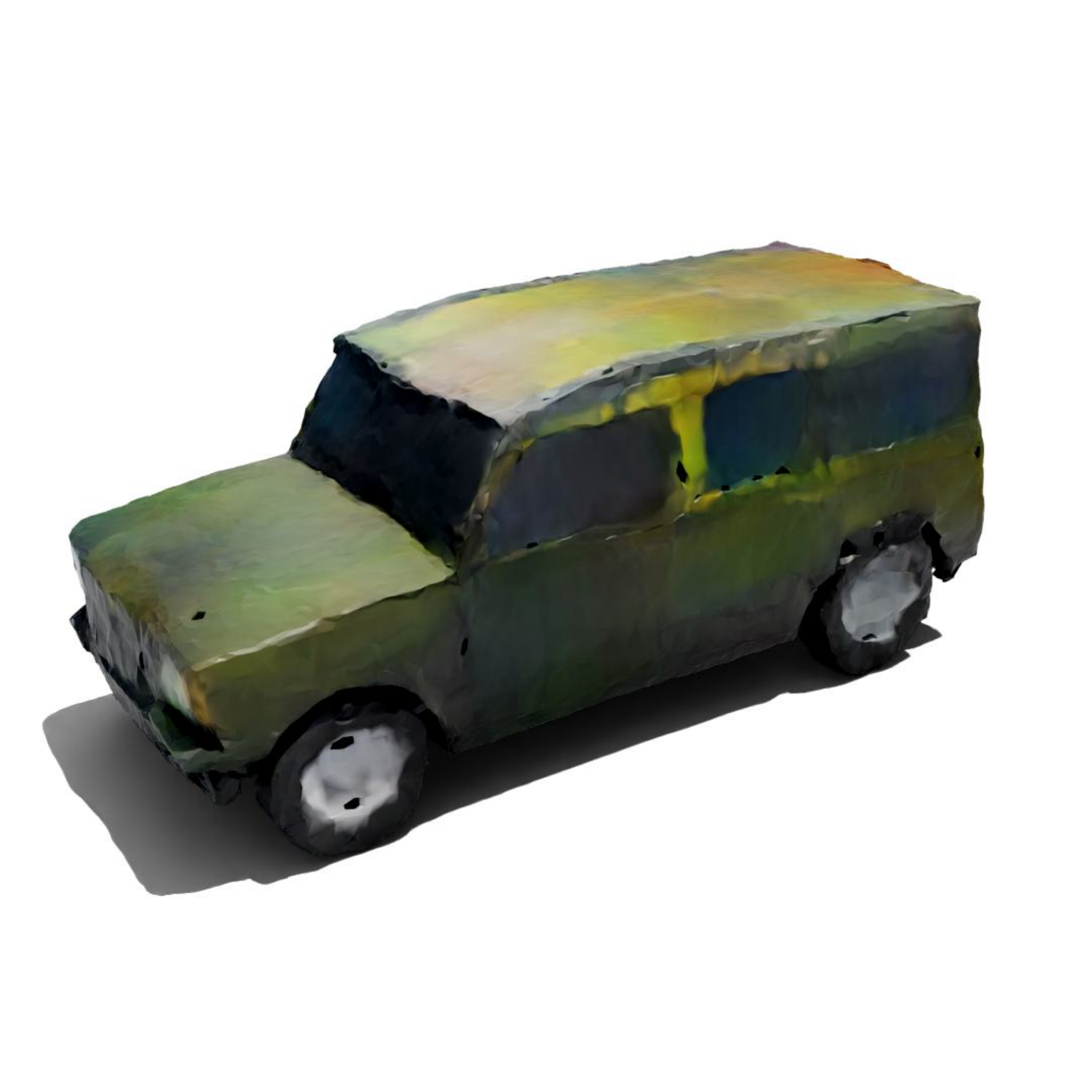}\includegraphics[width=0.08333333333333333\linewidth]{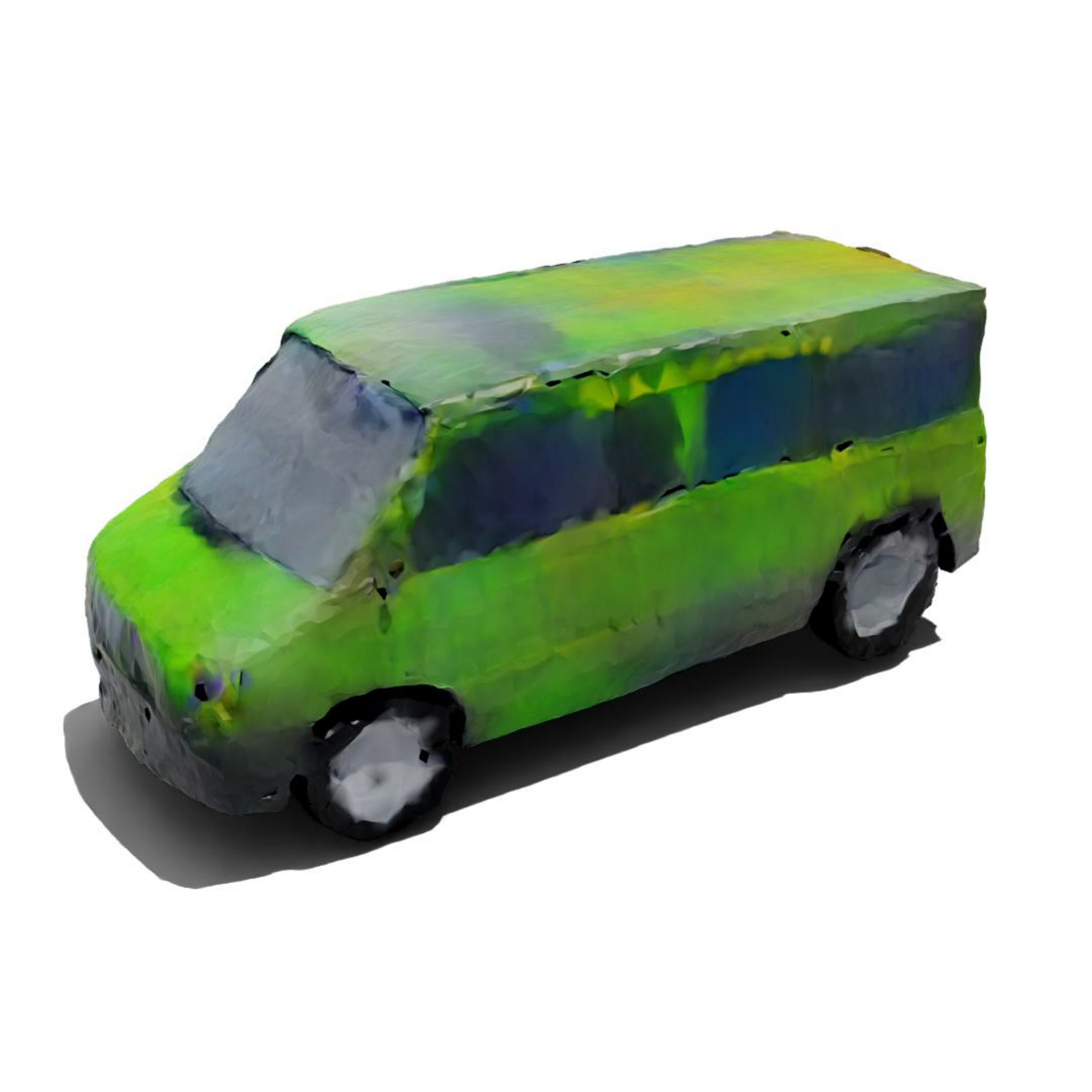}\includegraphics[width=0.08333333333333333\linewidth]{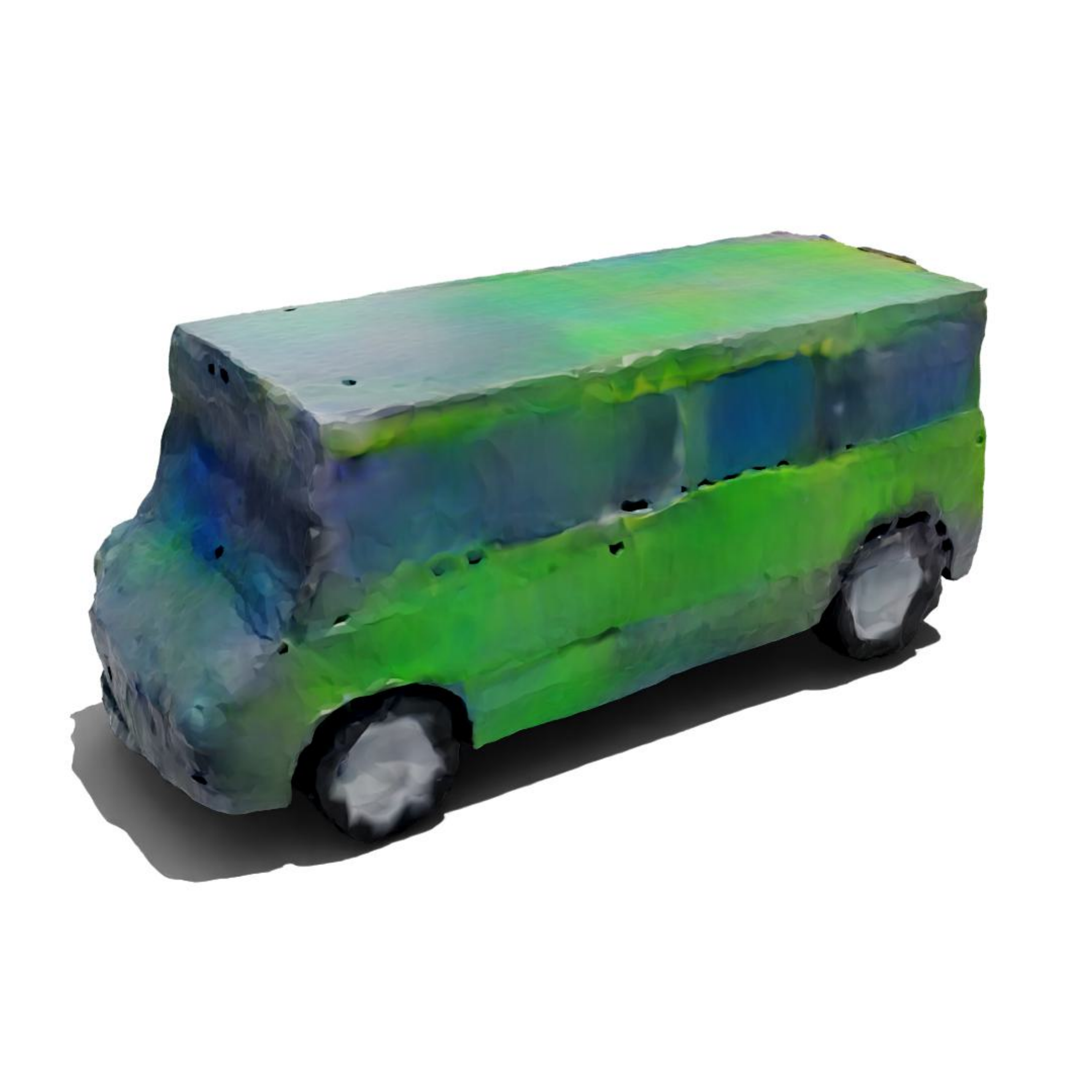}\includegraphics[width=0.08333333333333333\linewidth]{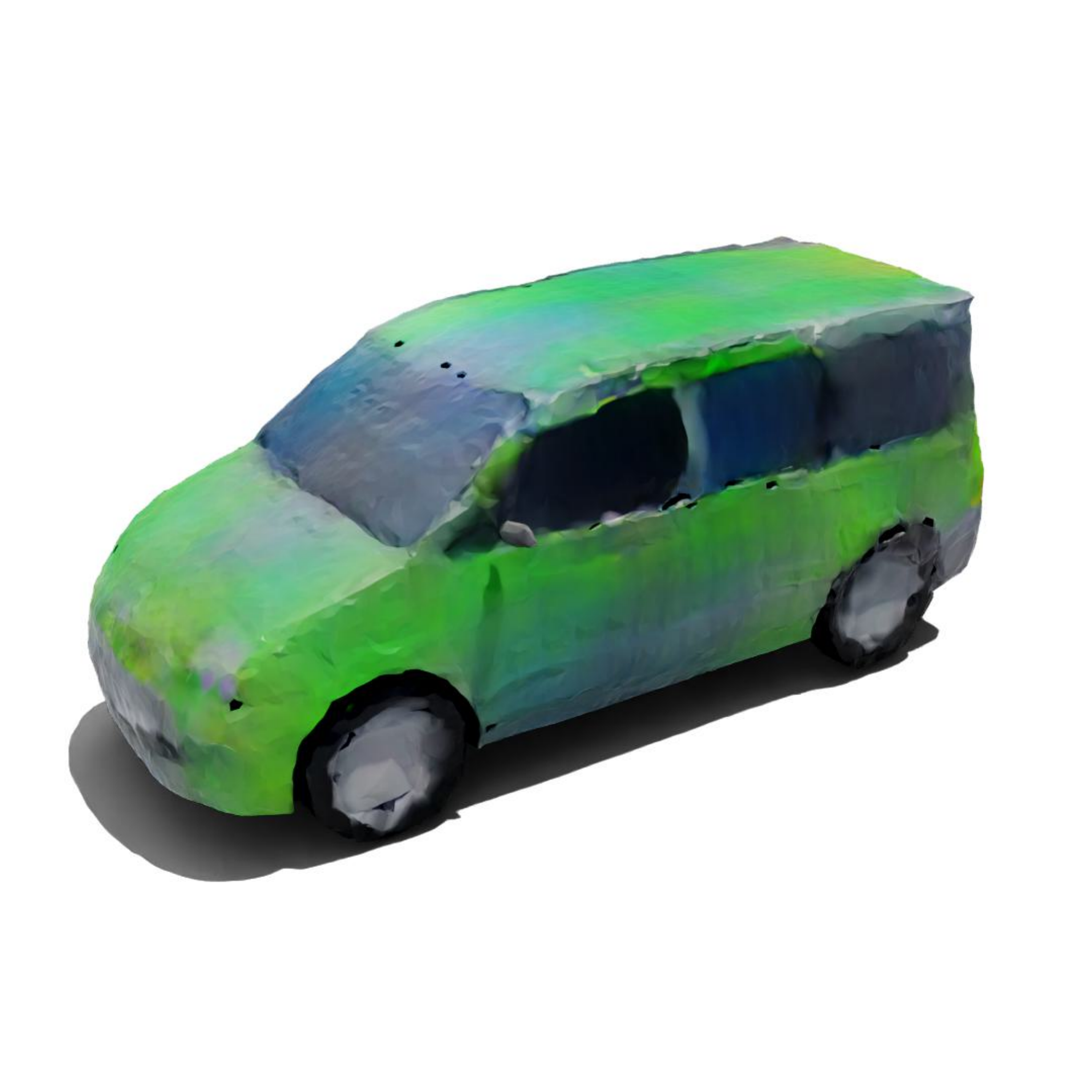}\includegraphics[width=0.08333333333333333\linewidth]{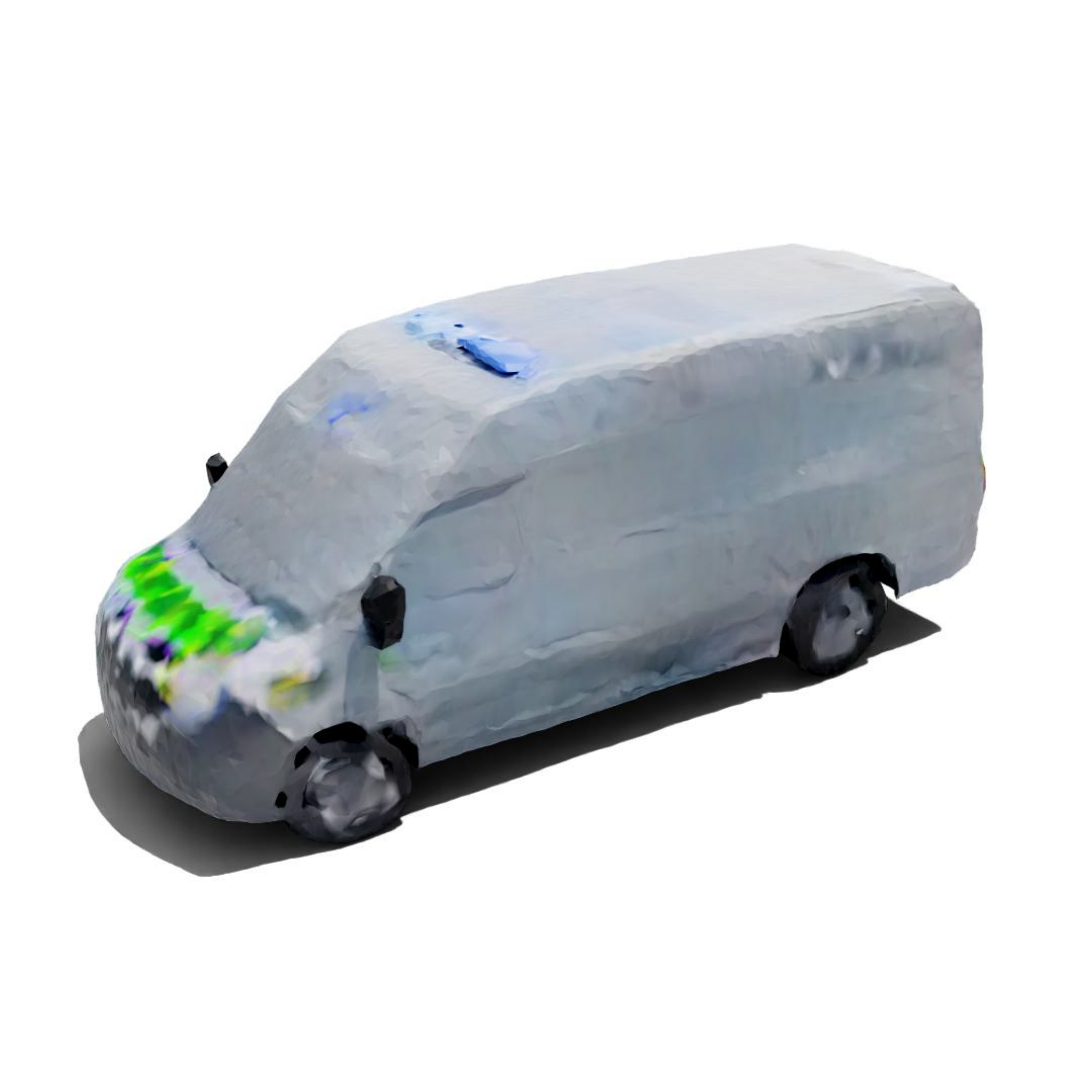}\includegraphics[width=0.08333333333333333\linewidth]{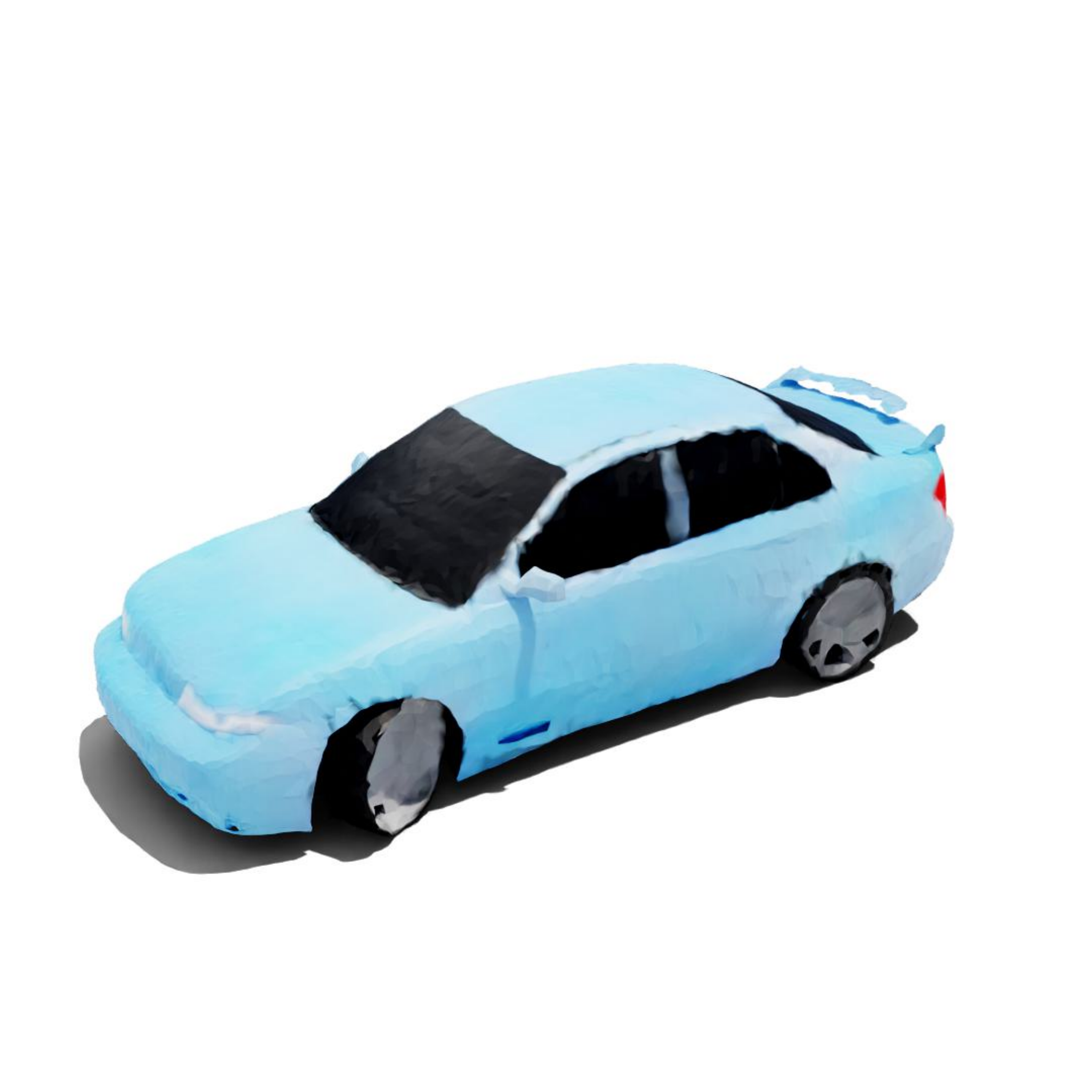}\includegraphics[width=0.08333333333333333\linewidth]{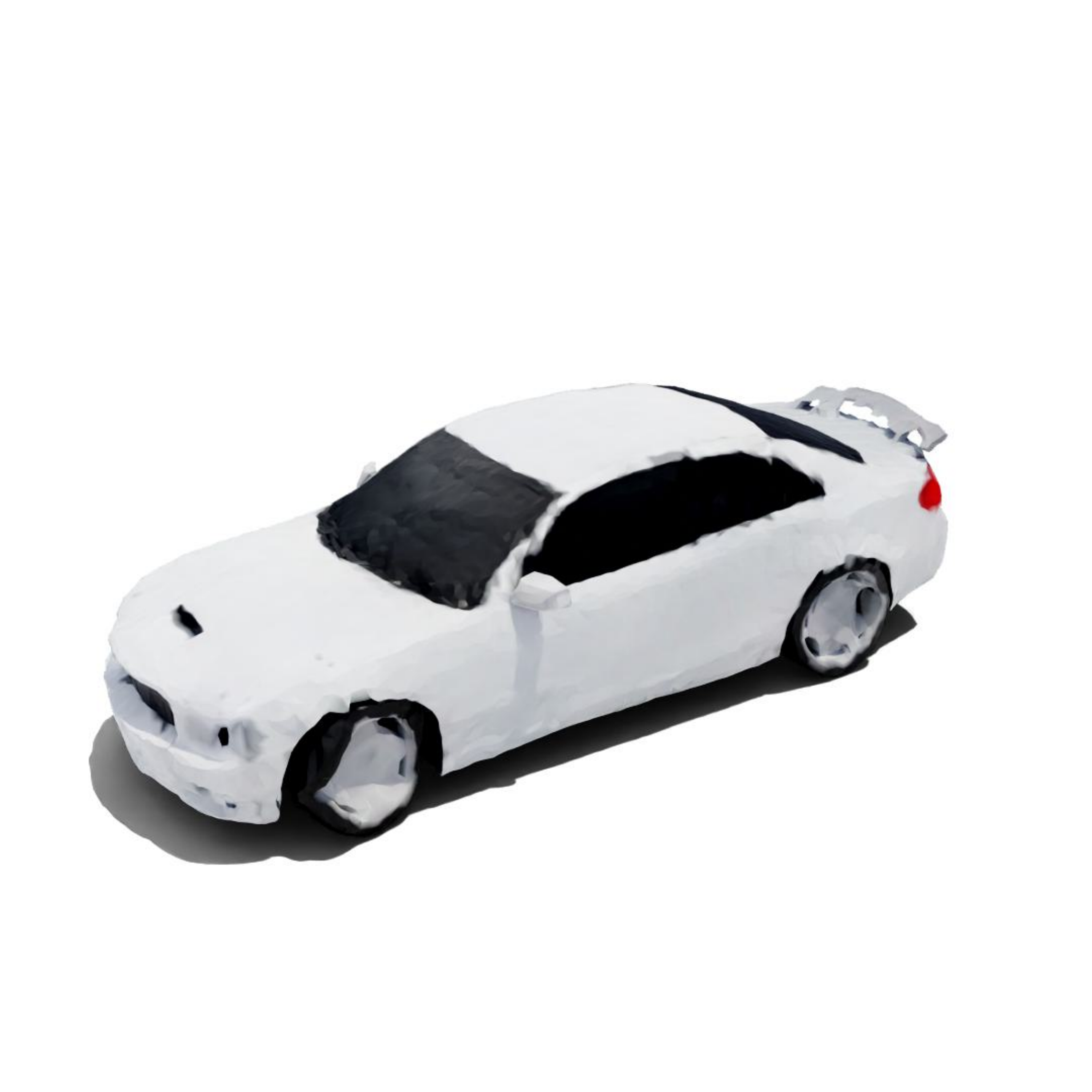}\includegraphics[width=0.08333333333333333\linewidth]{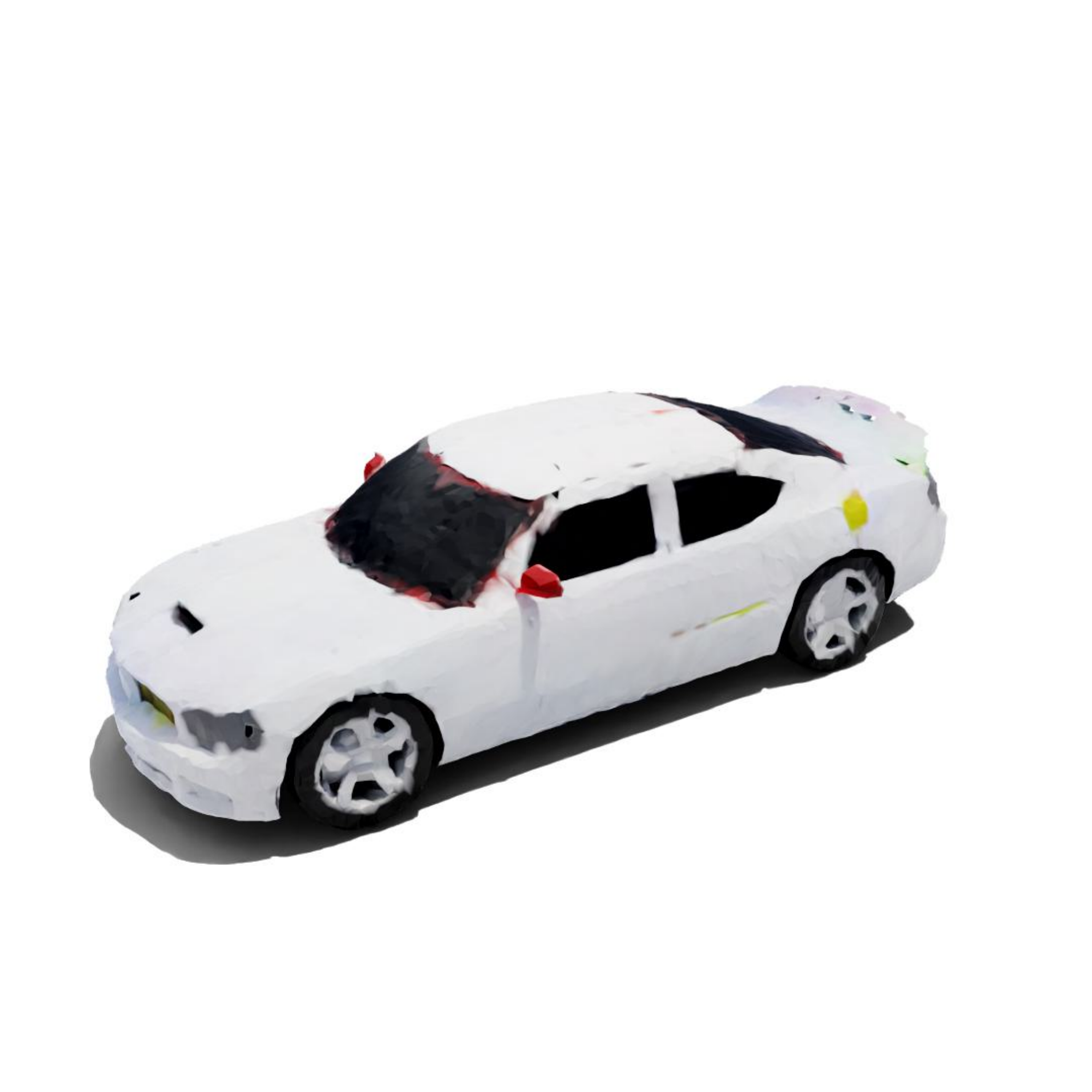}\includegraphics[width=0.08333333333333333\linewidth]{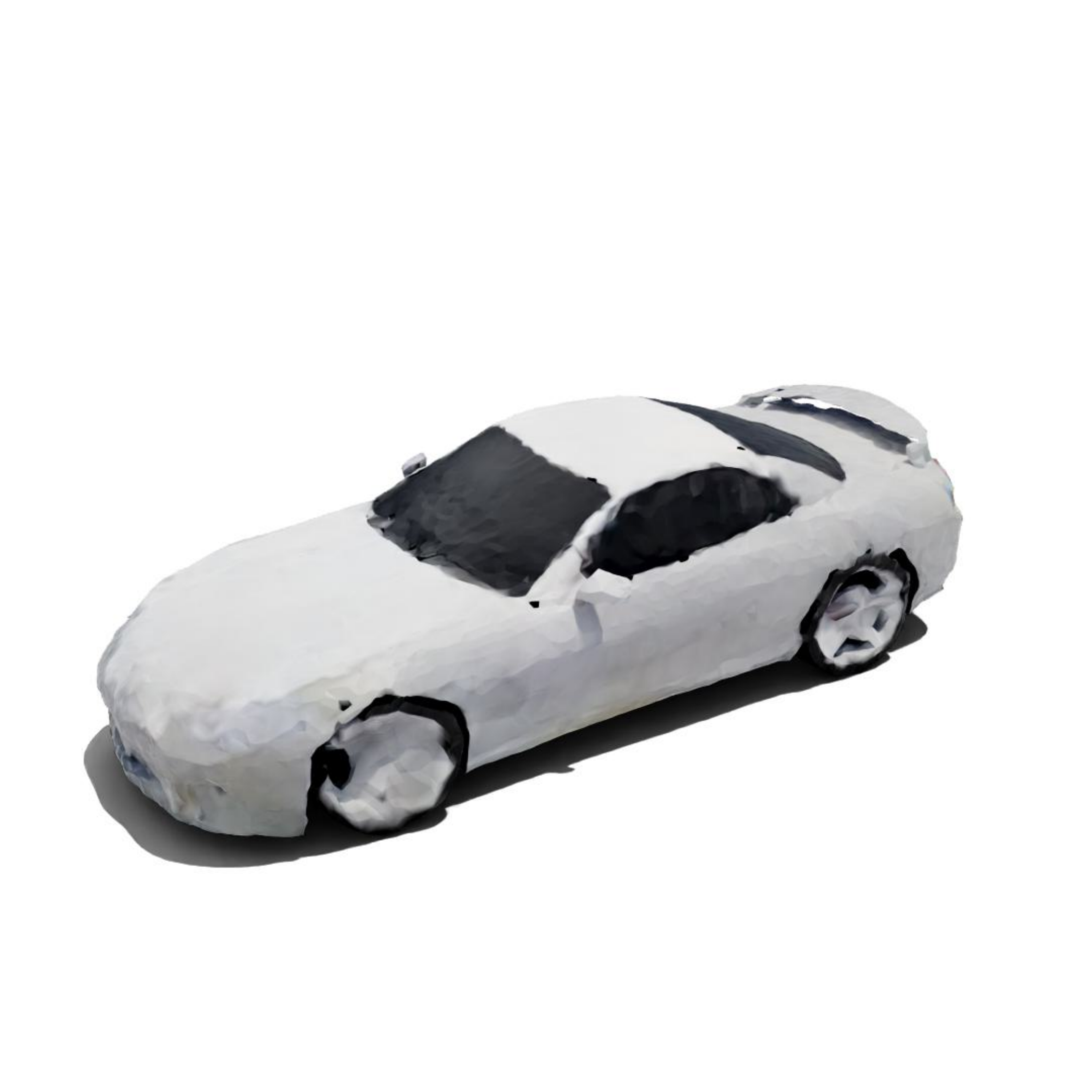}\includegraphics[width=0.08333333333333333\linewidth]{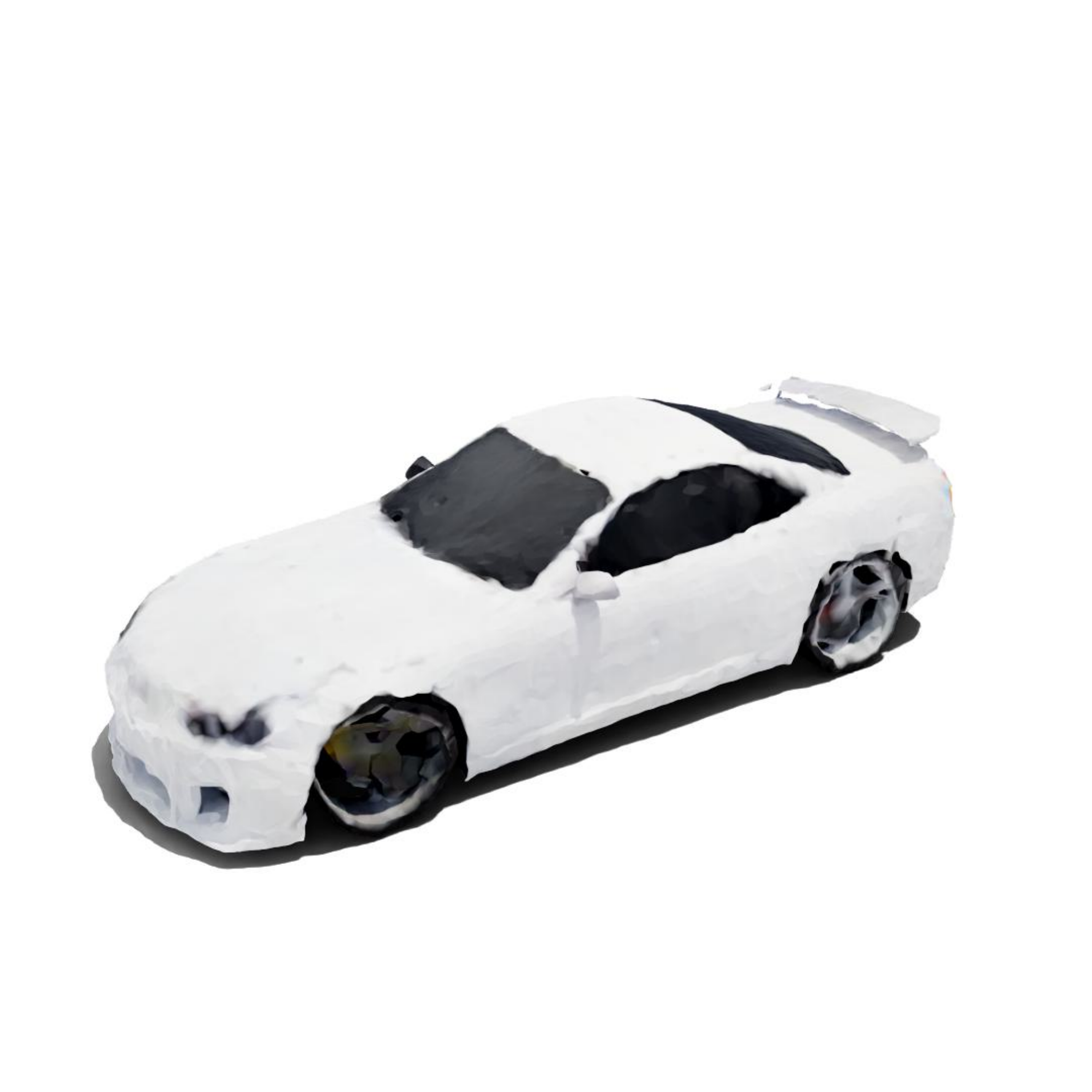}\\
\includegraphics[width=0.08333333333333333\linewidth]{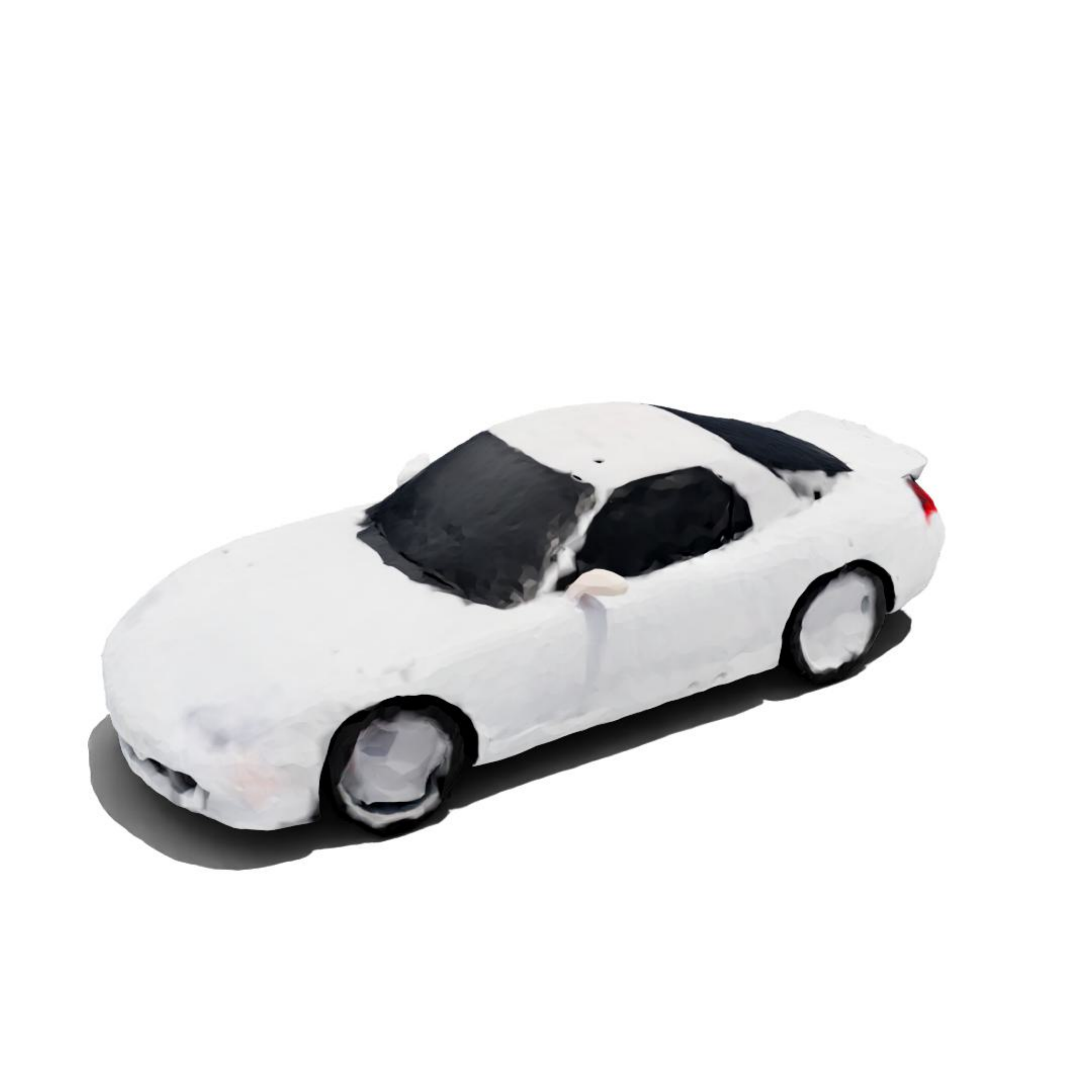}\includegraphics[width=0.08333333333333333\linewidth]{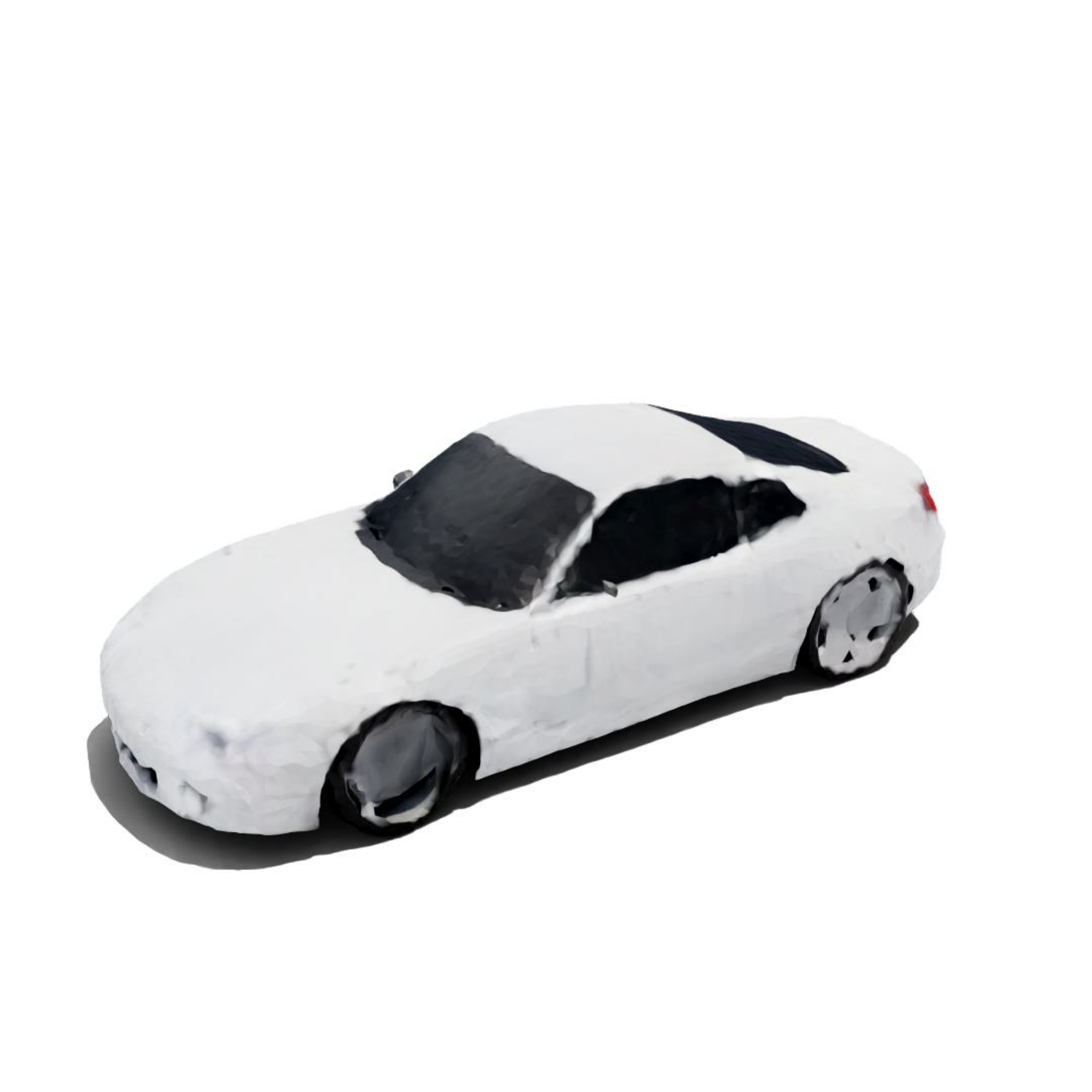}\includegraphics[width=0.08333333333333333\linewidth]{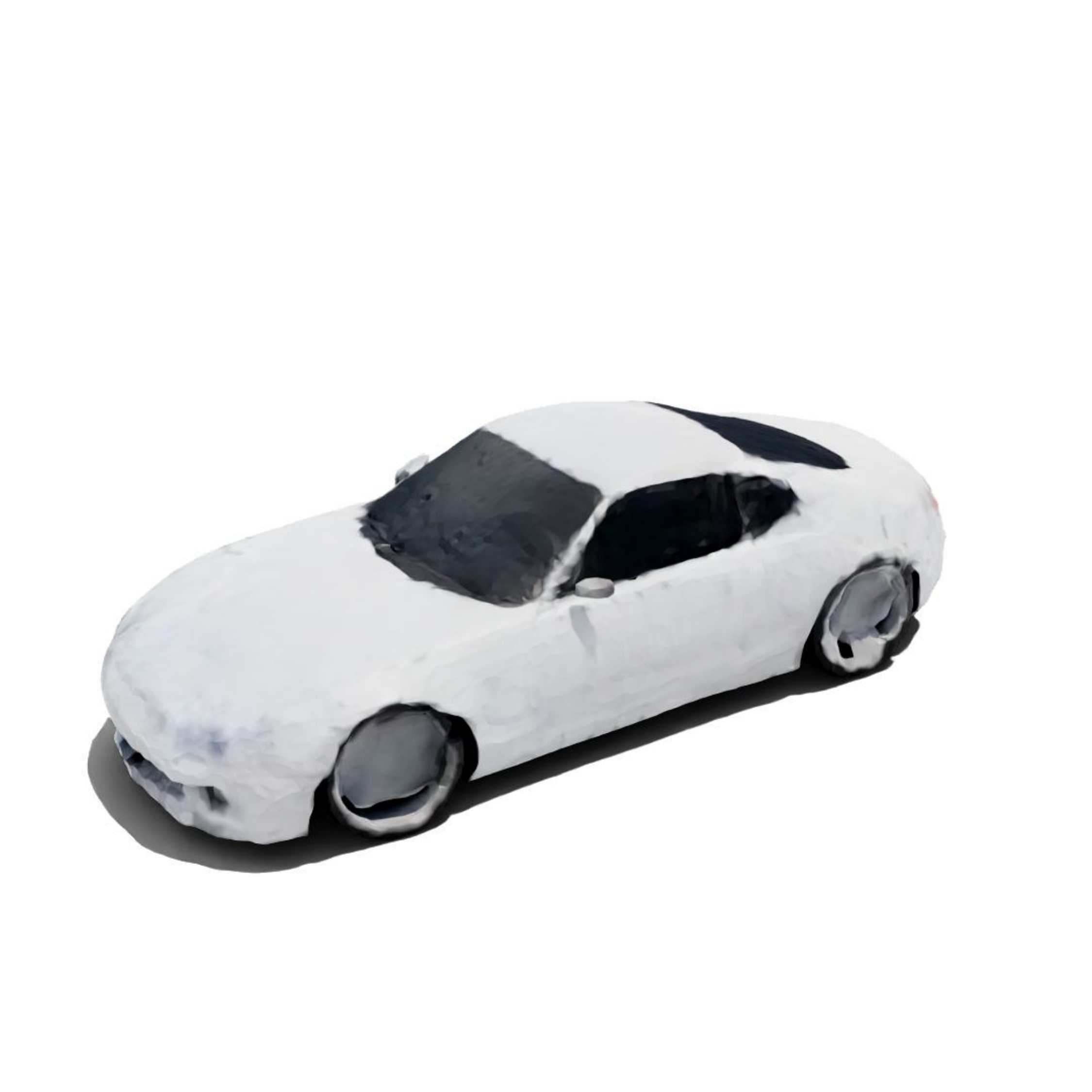}\includegraphics[width=0.08333333333333333\linewidth]{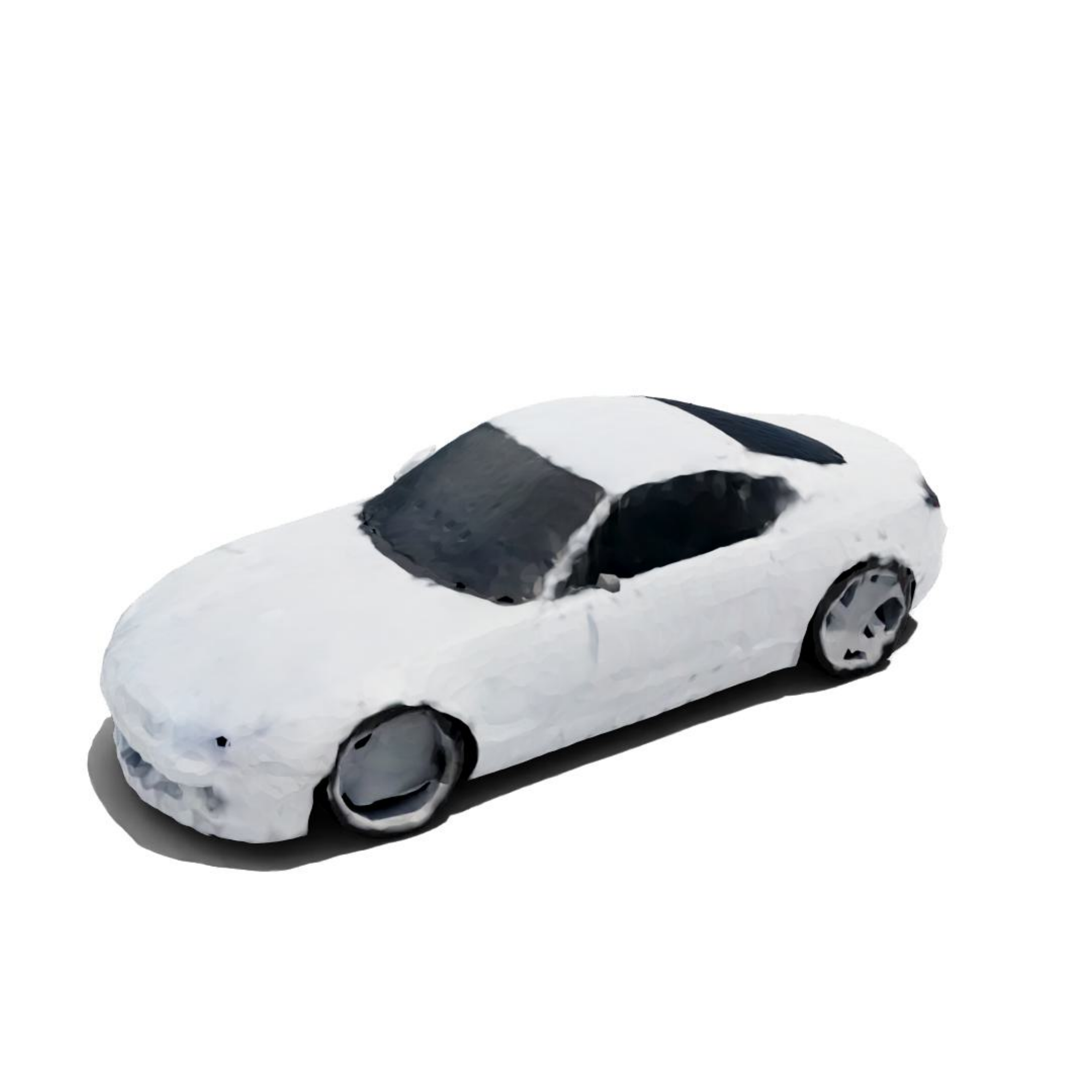}\includegraphics[width=0.08333333333333333\linewidth]{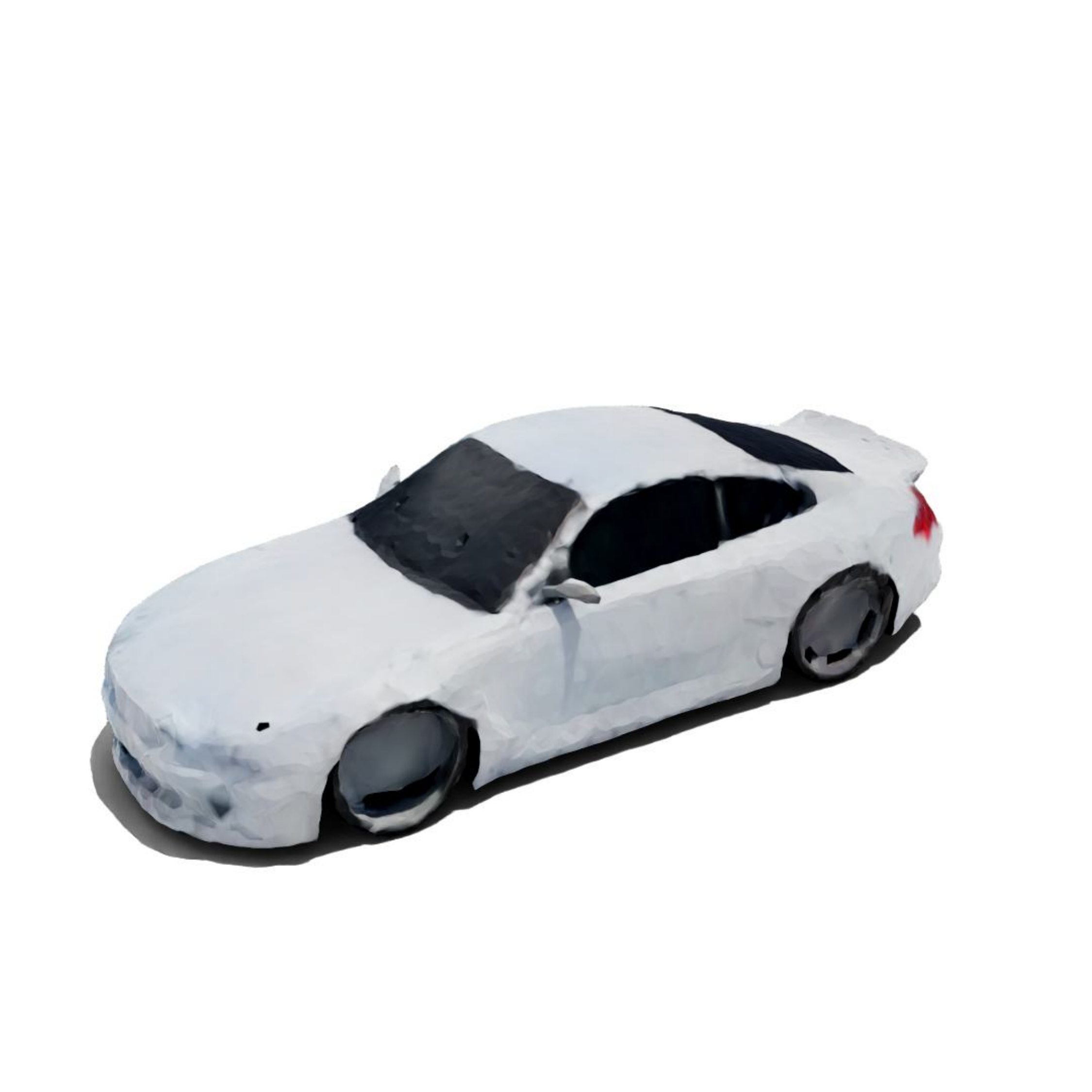}\includegraphics[width=0.08333333333333333\linewidth]{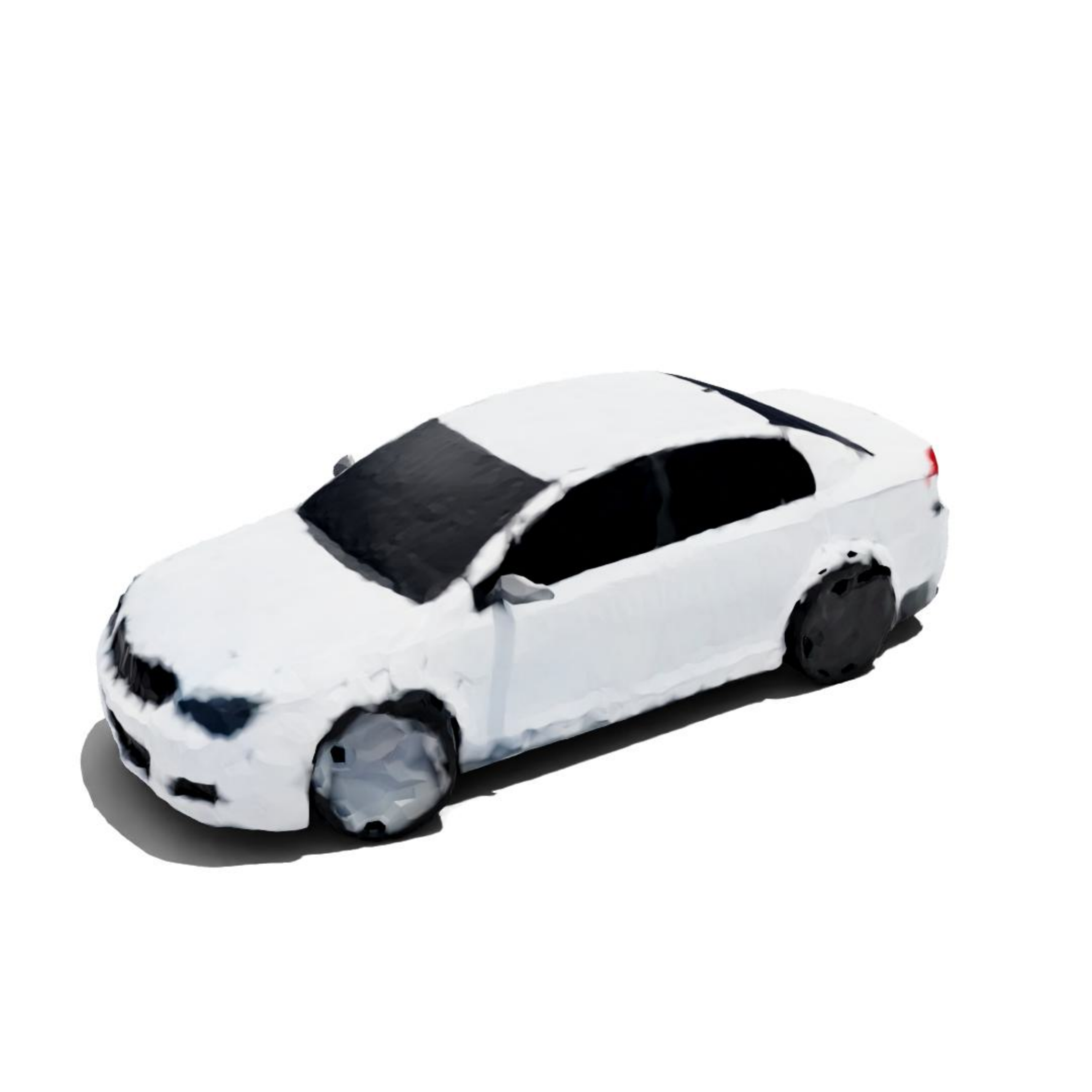}\includegraphics[width=0.08333333333333333\linewidth]{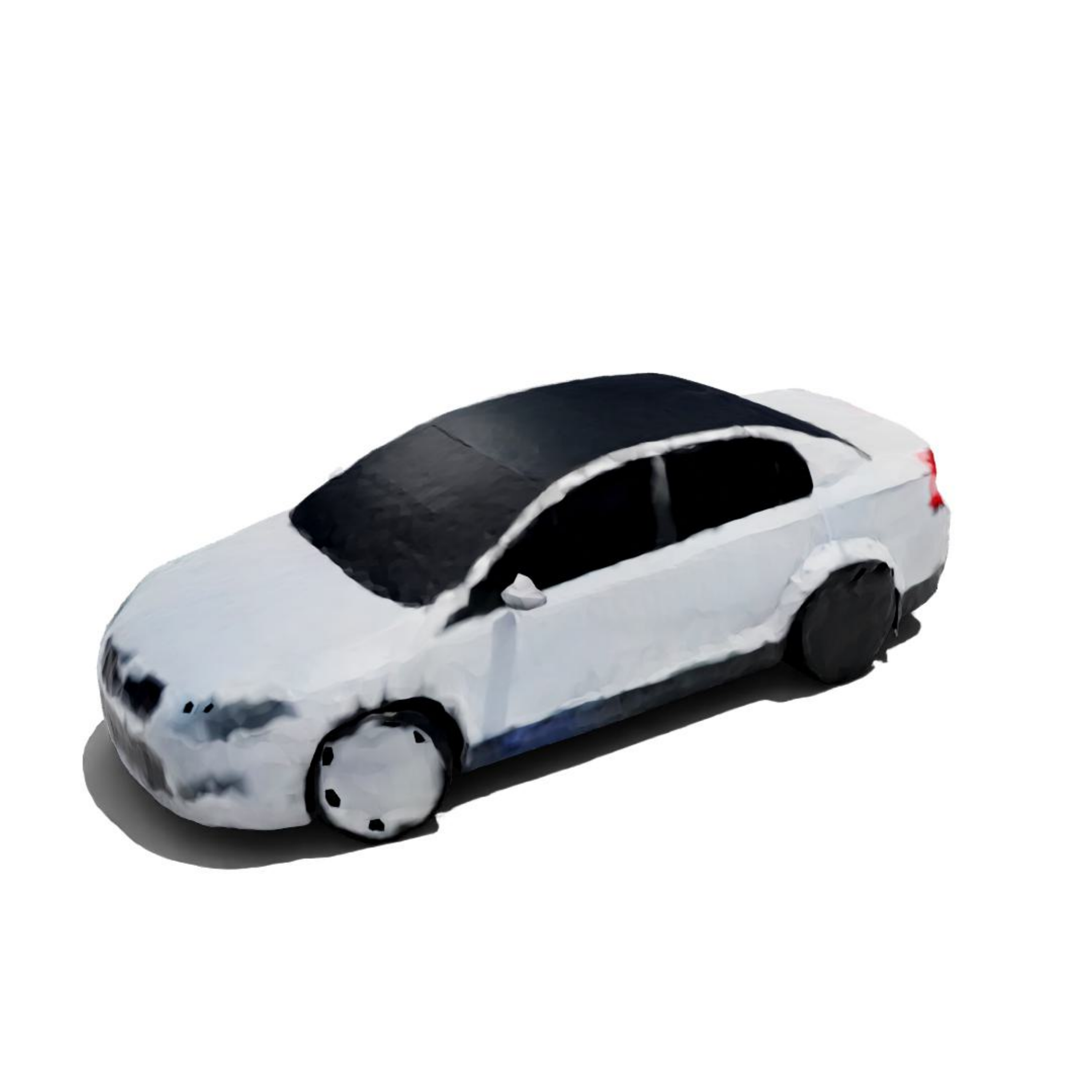}\includegraphics[width=0.08333333333333333\linewidth]{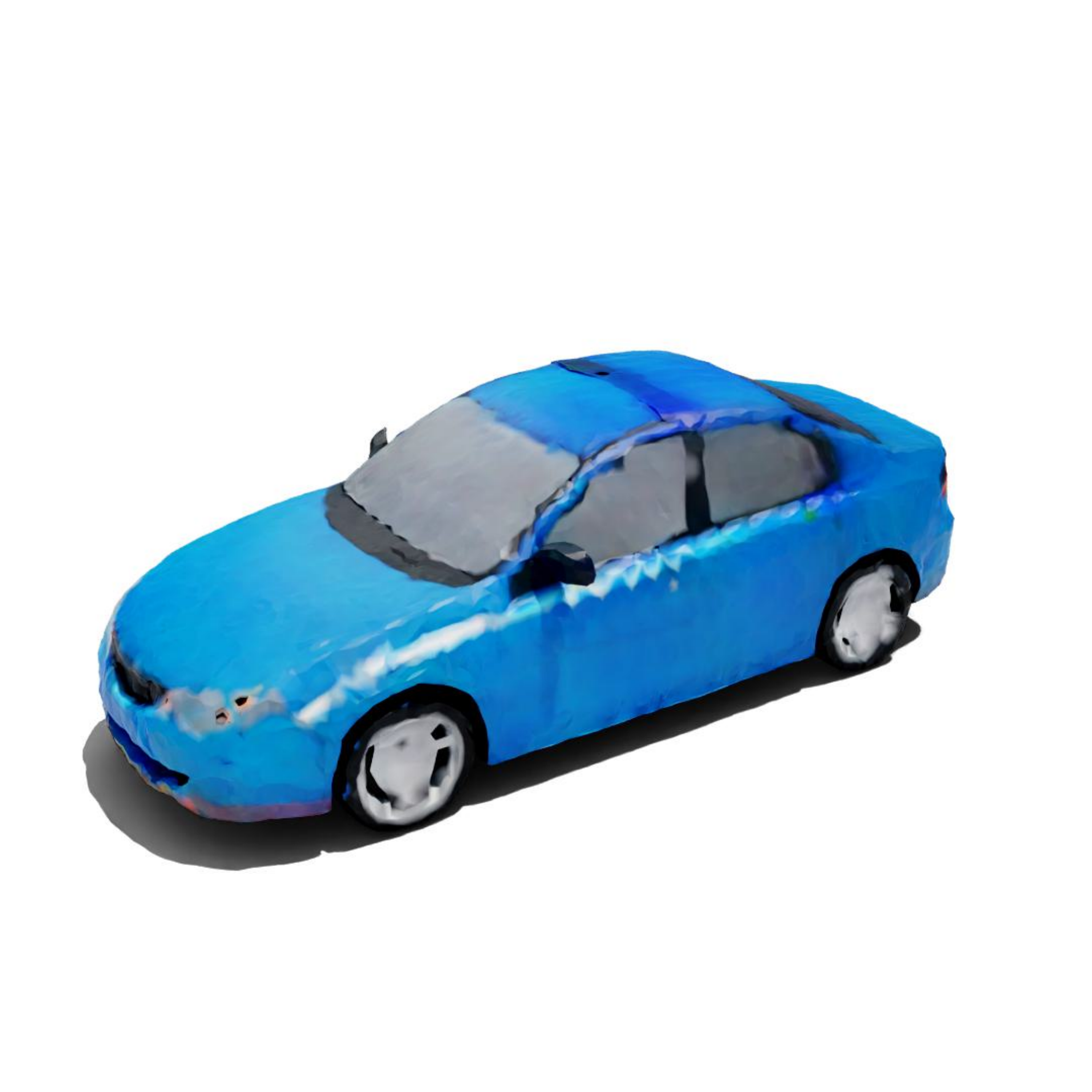}\includegraphics[width=0.08333333333333333\linewidth]{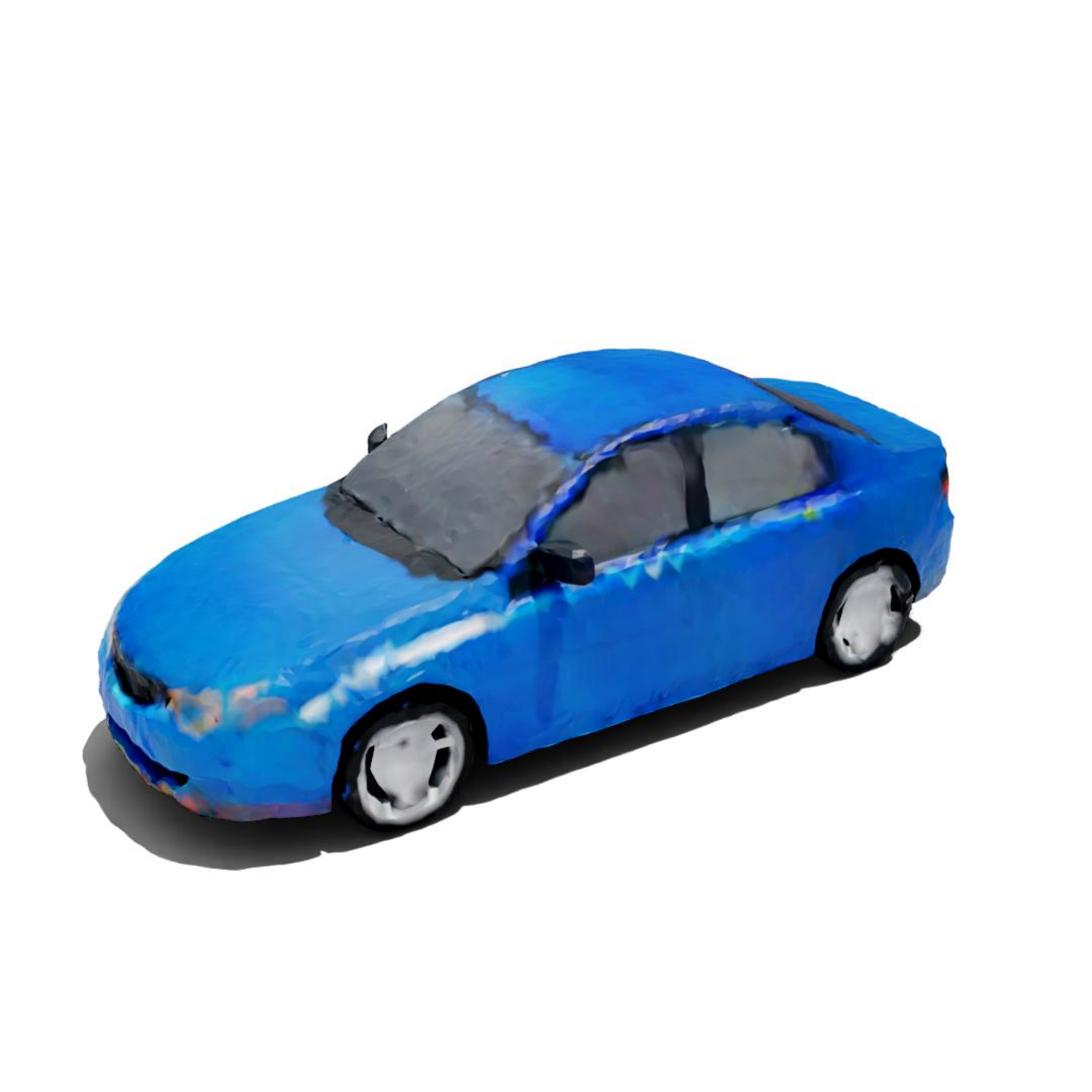}\includegraphics[width=0.08333333333333333\linewidth]{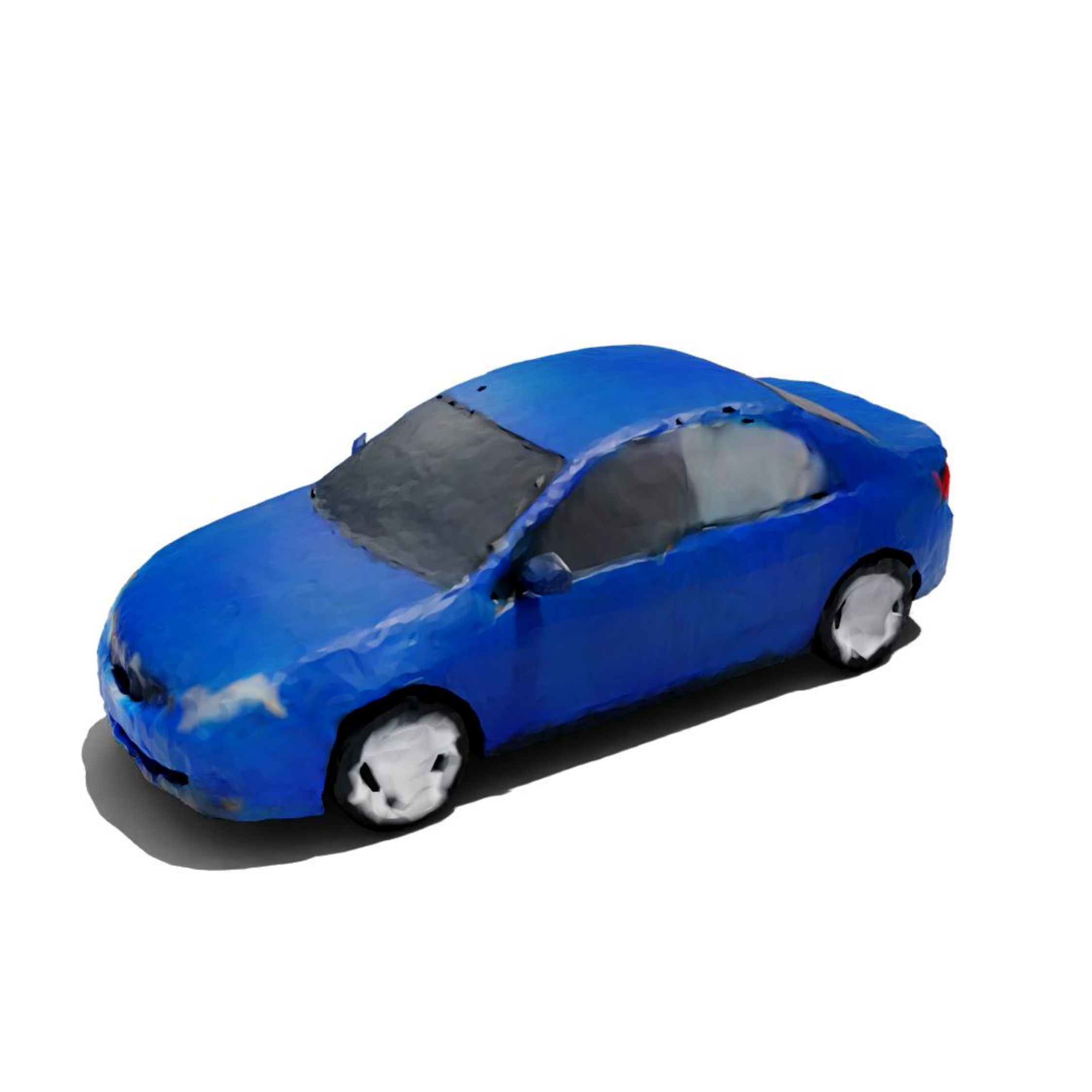}\includegraphics[width=0.08333333333333333\linewidth]{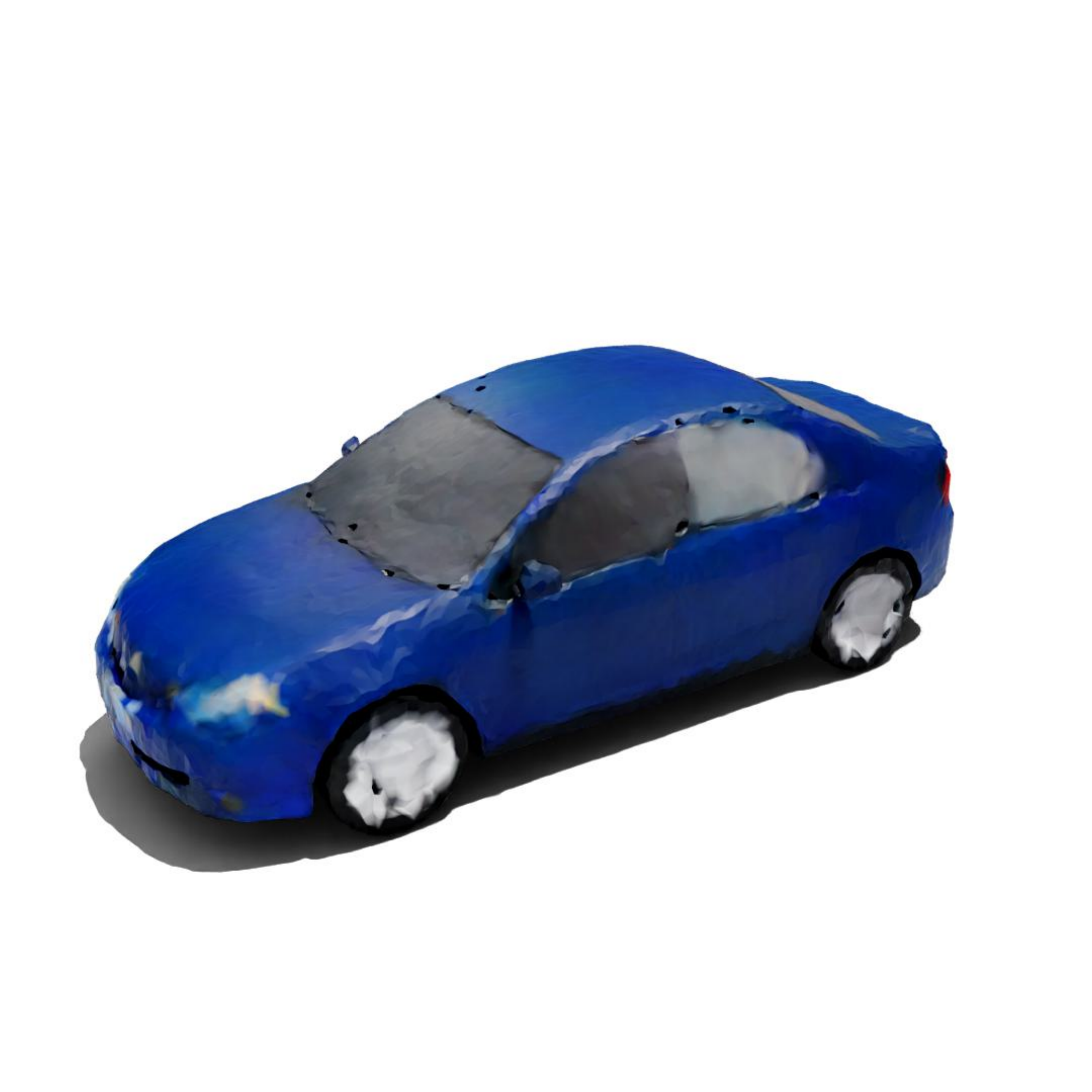}\includegraphics[width=0.08333333333333333\linewidth]{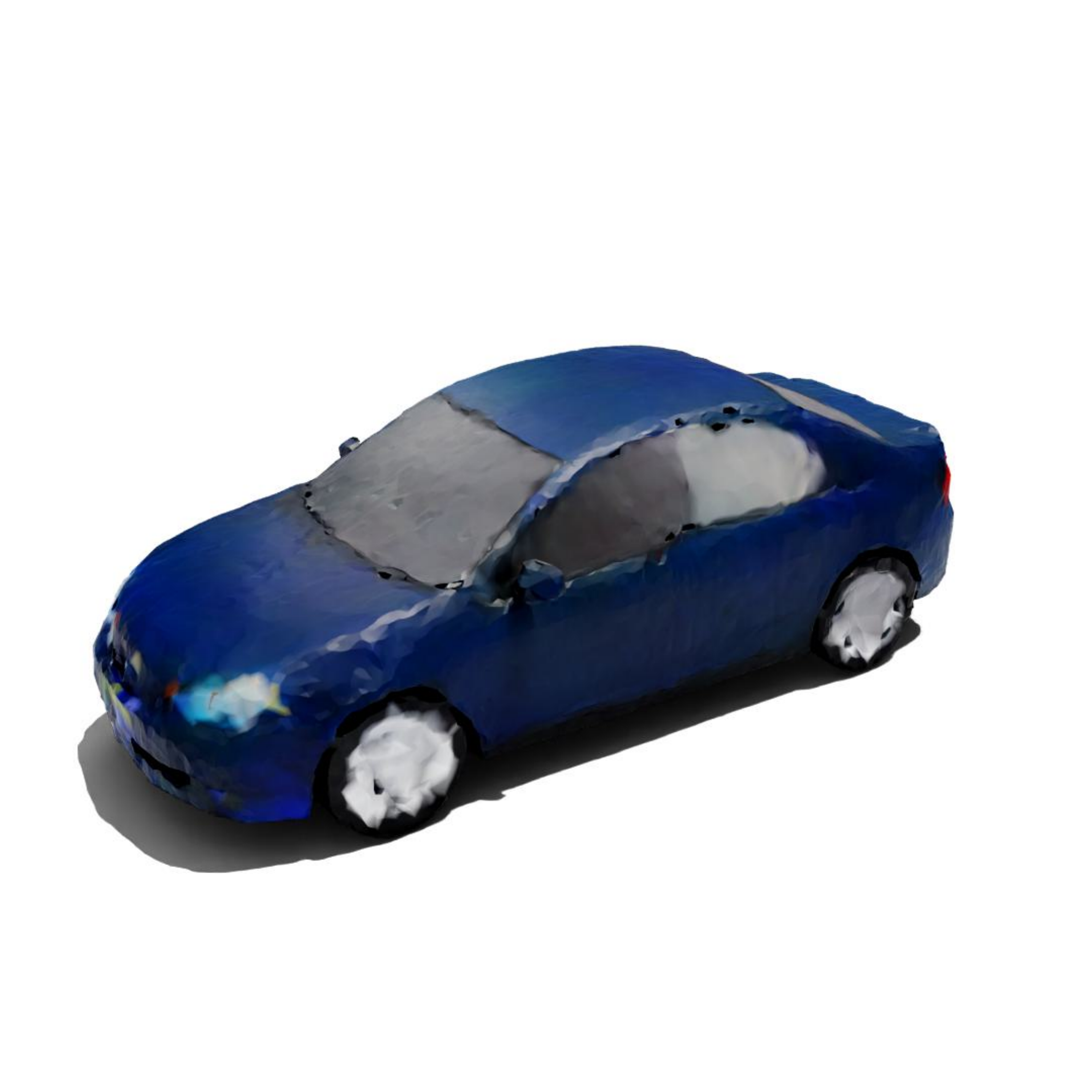}

\includegraphics[width=0.08333333333333333\linewidth]{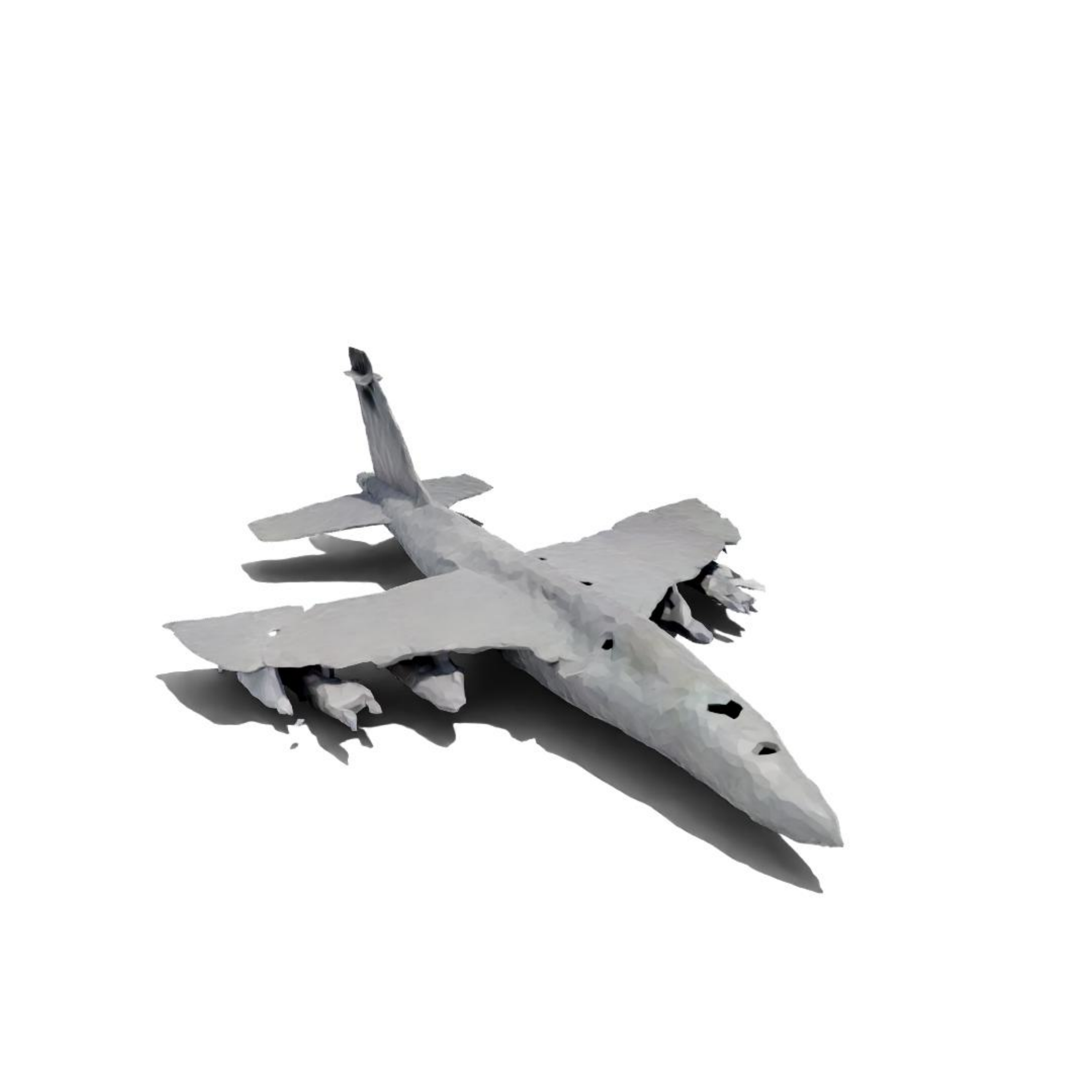}\includegraphics[width=0.08333333333333333\linewidth]{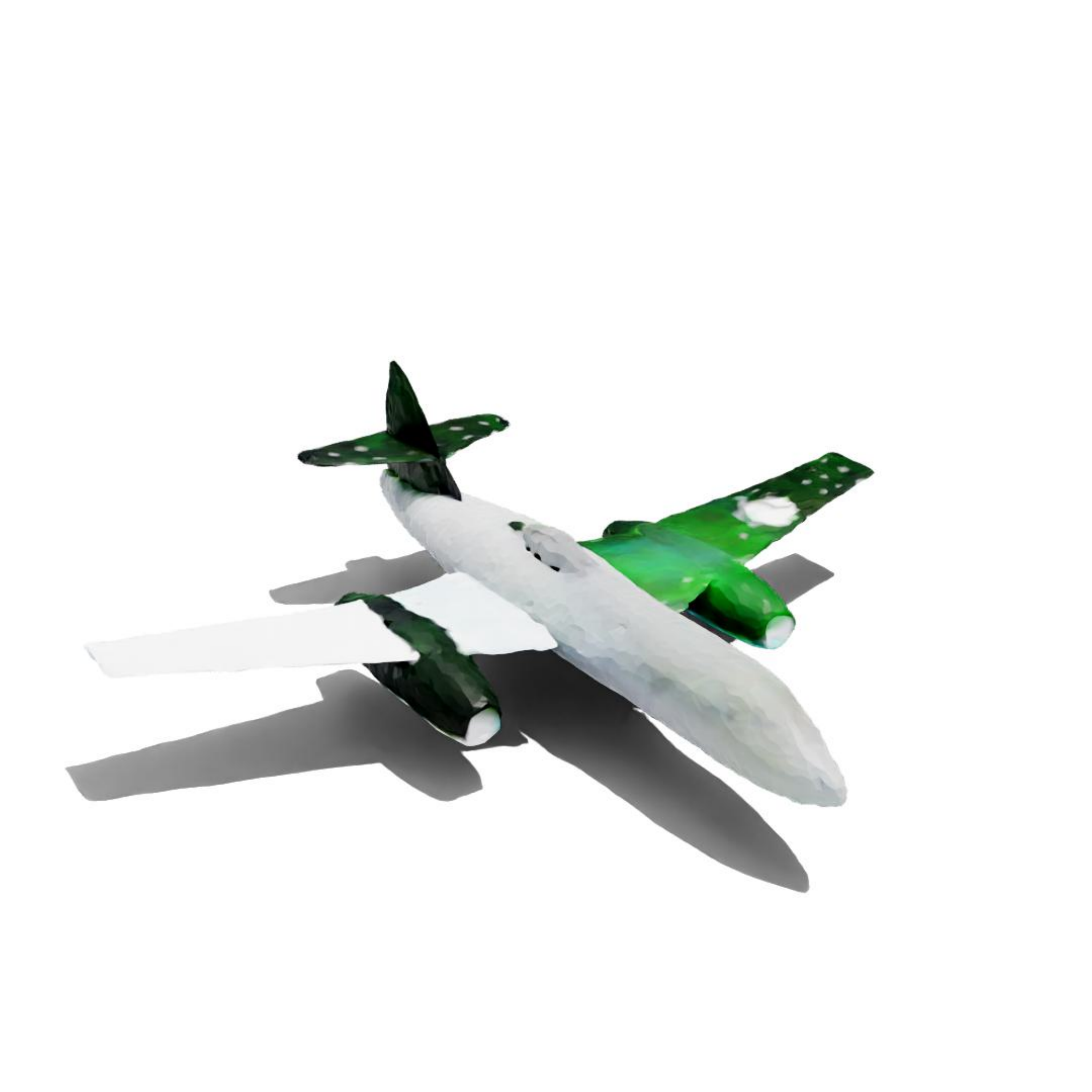}\includegraphics[width=0.08333333333333333\linewidth]{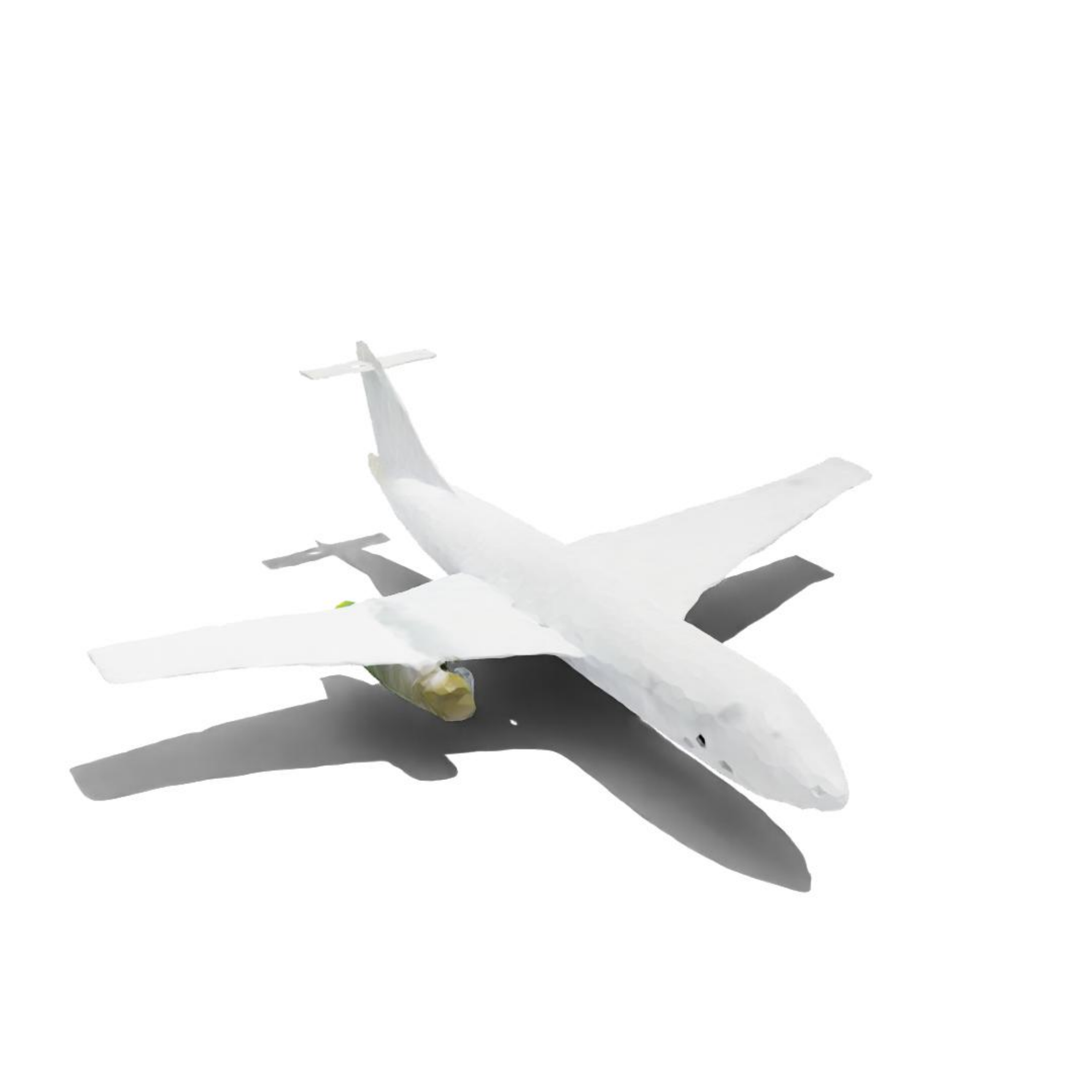}\includegraphics[width=0.08333333333333333\linewidth]{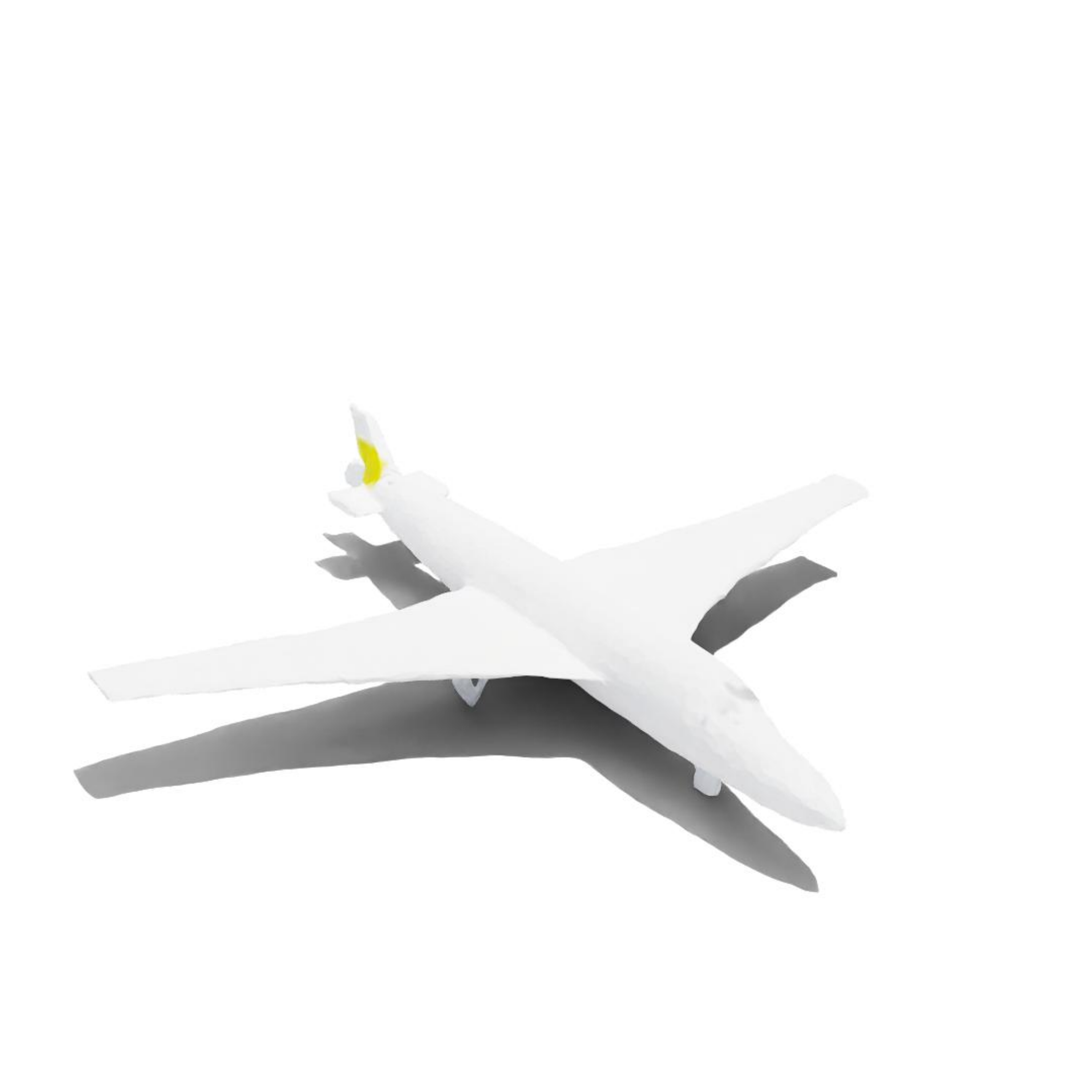}\includegraphics[width=0.08333333333333333\linewidth]{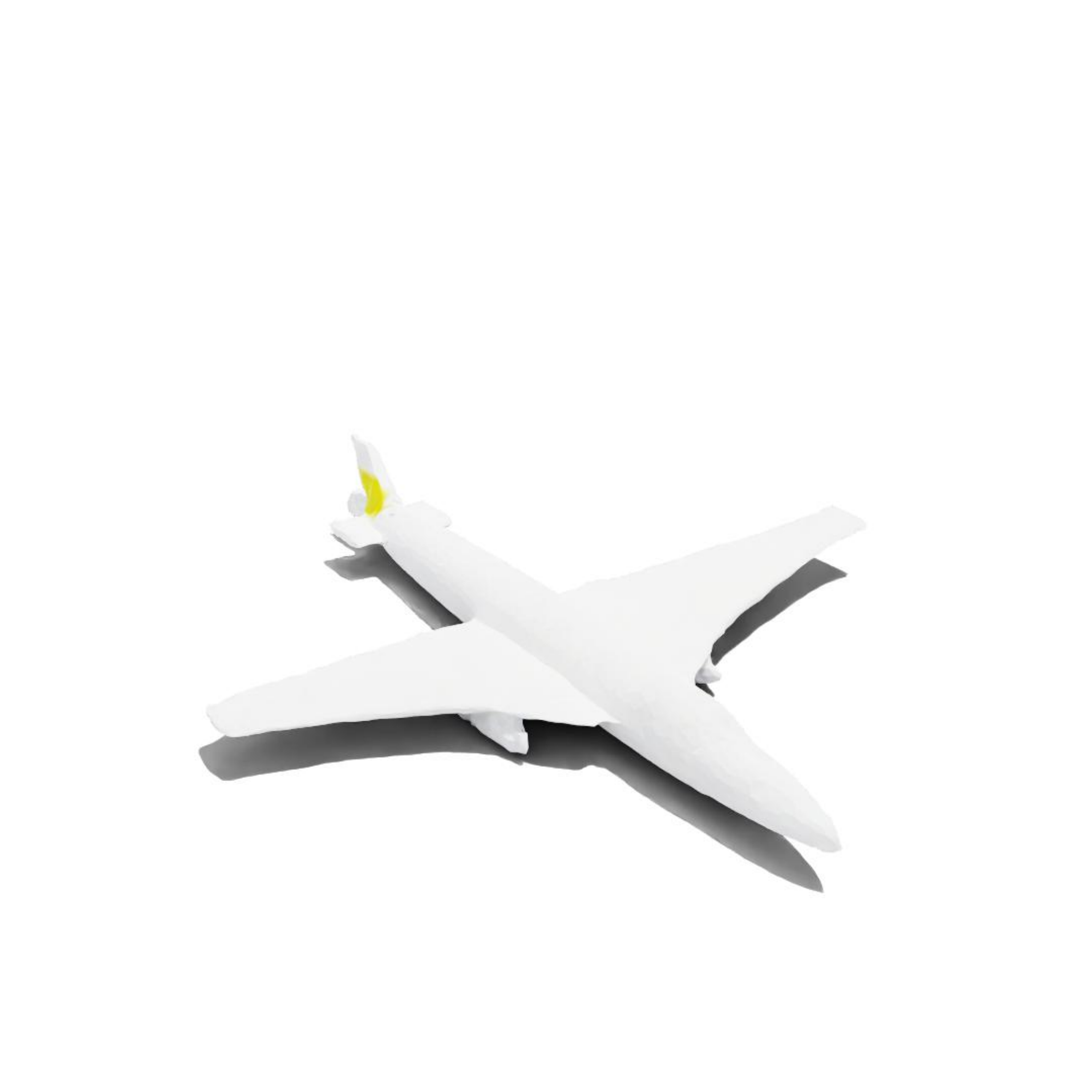}\includegraphics[width=0.08333333333333333\linewidth]{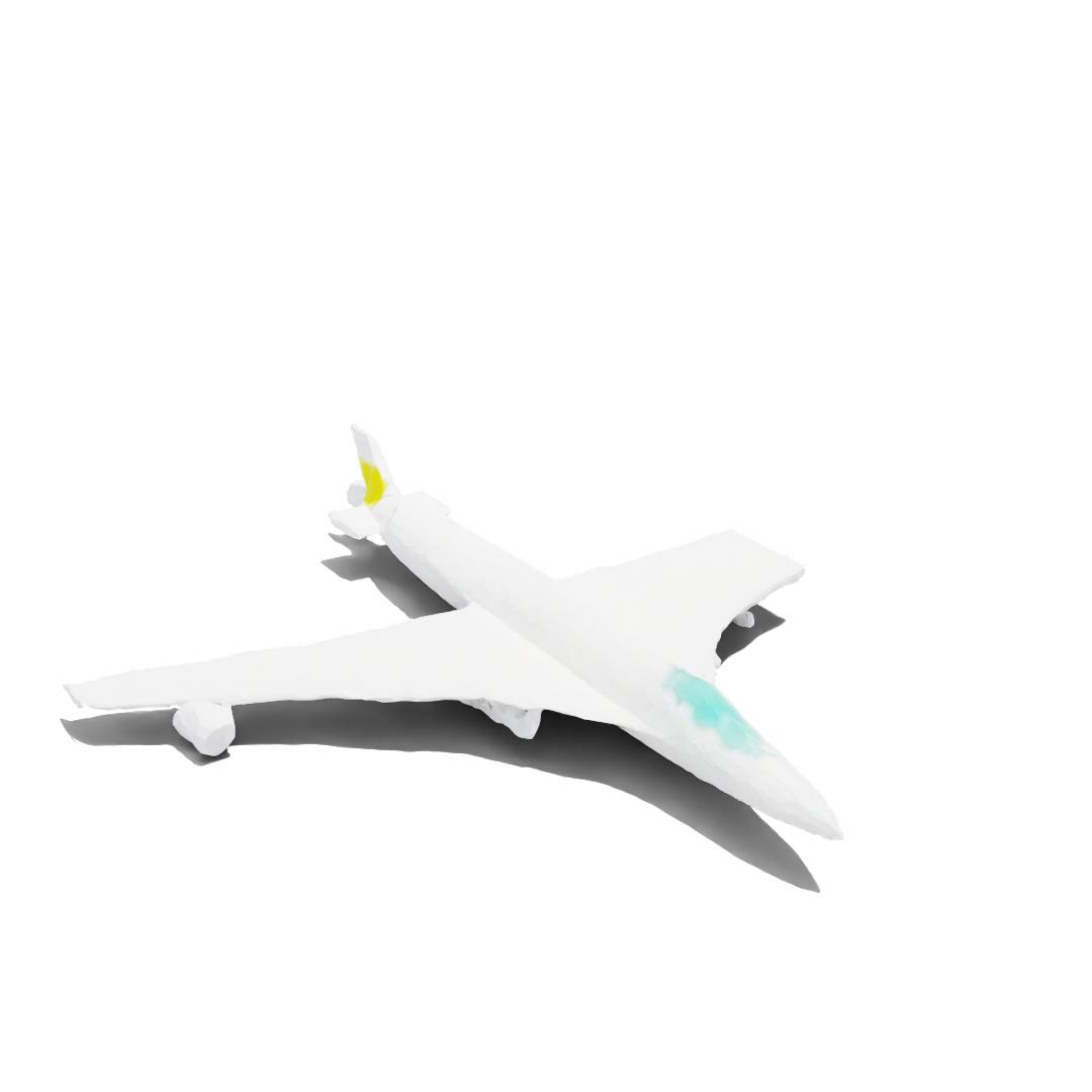}\includegraphics[width=0.08333333333333333\linewidth]{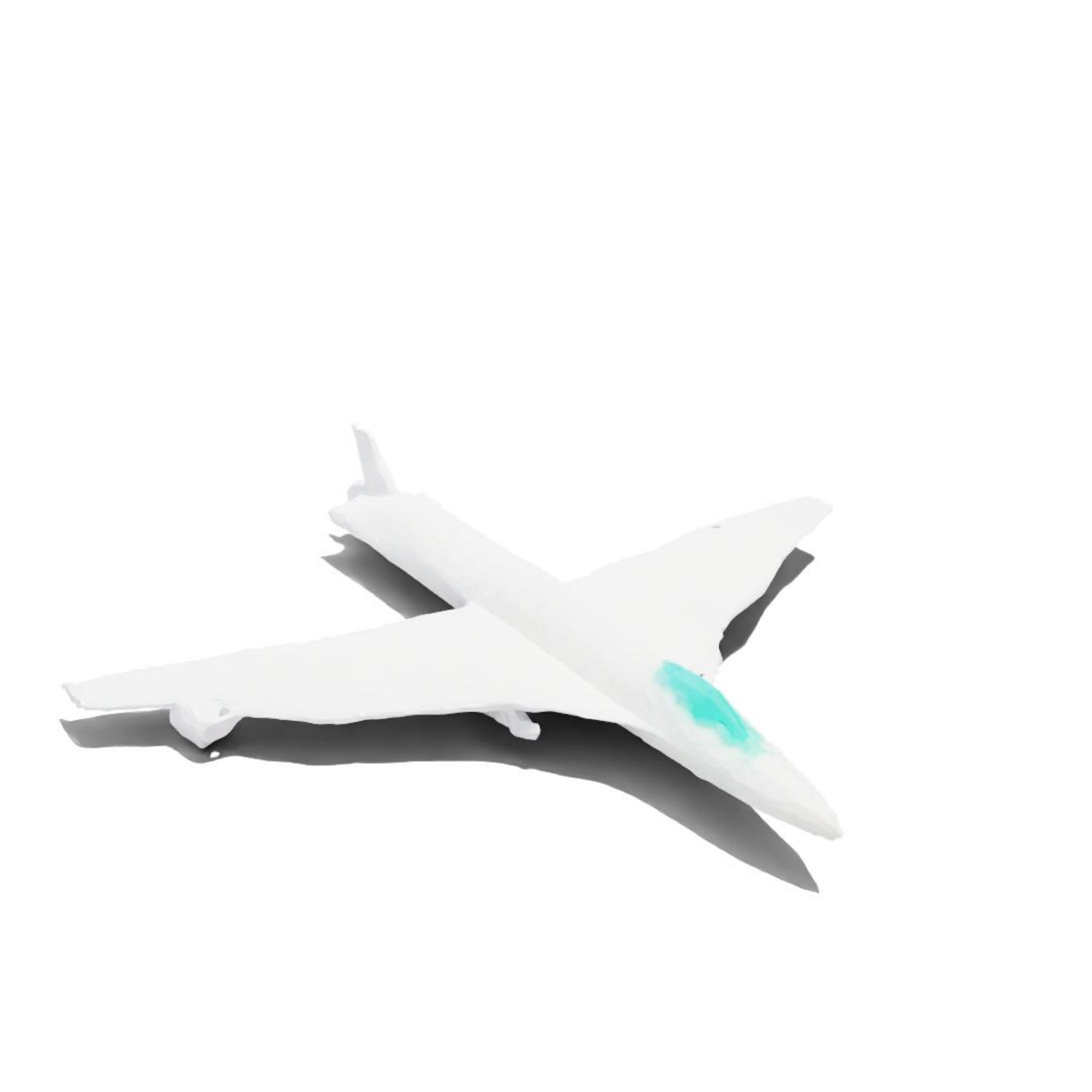}\includegraphics[width=0.08333333333333333\linewidth]{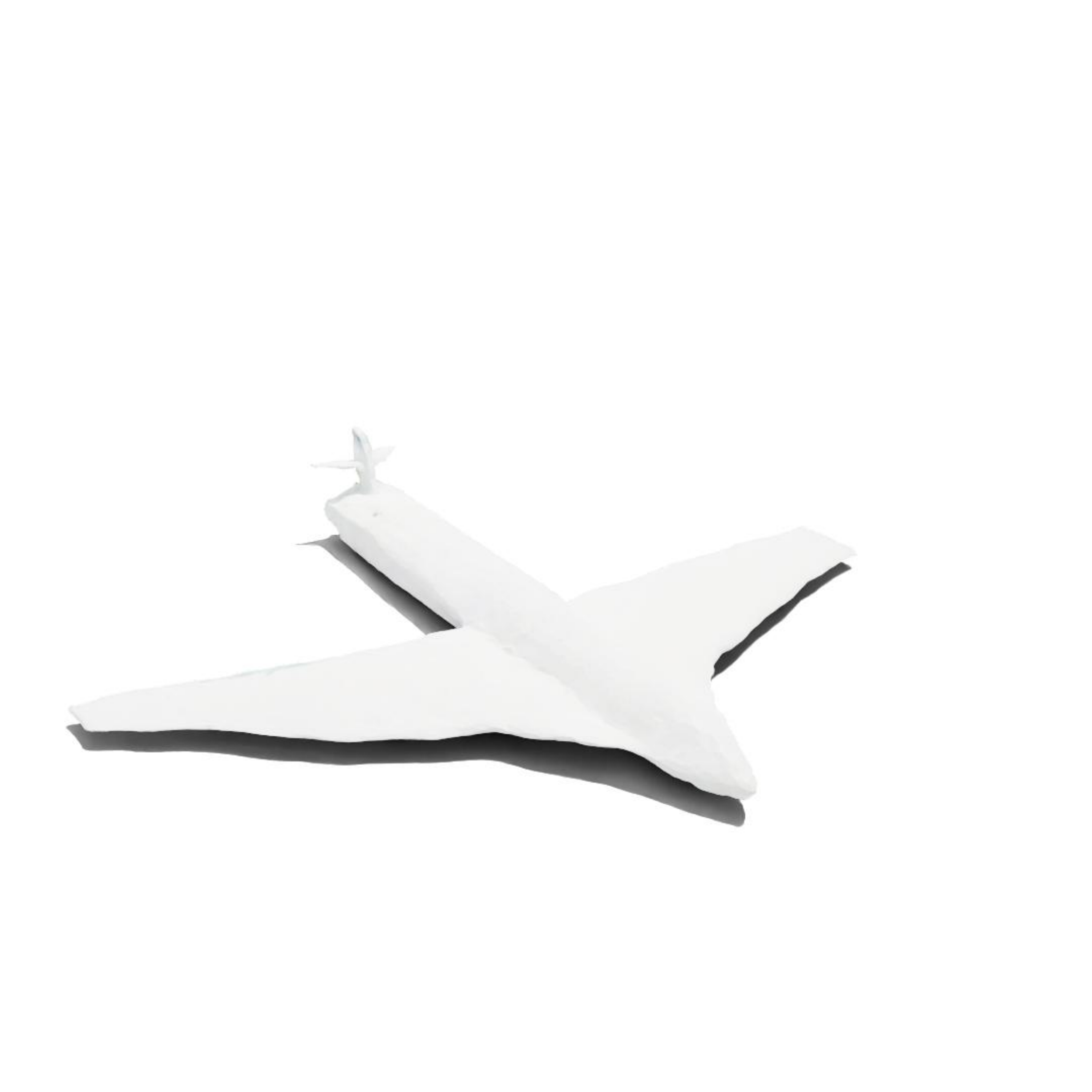}\includegraphics[width=0.08333333333333333\linewidth]{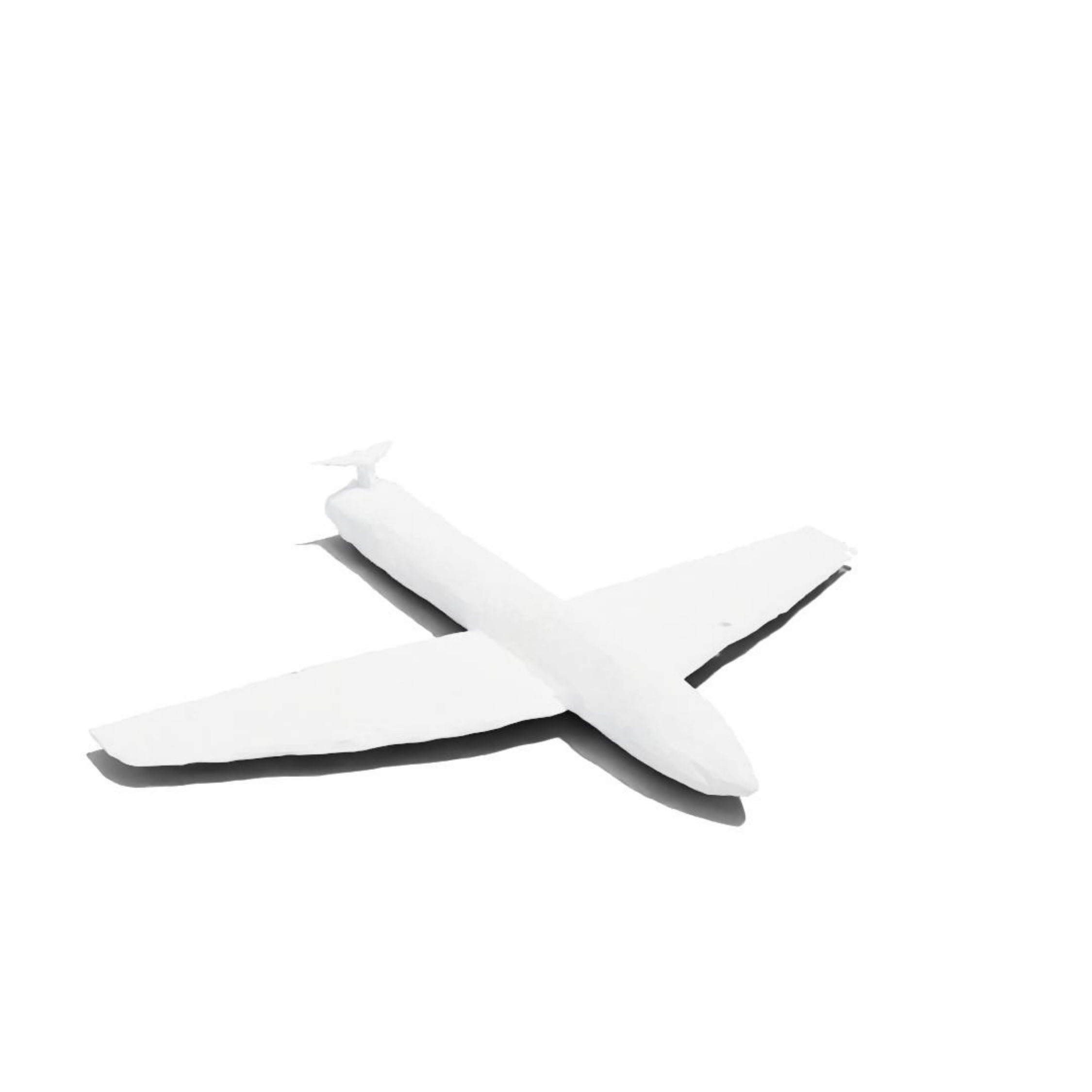}\includegraphics[width=0.08333333333333333\linewidth]{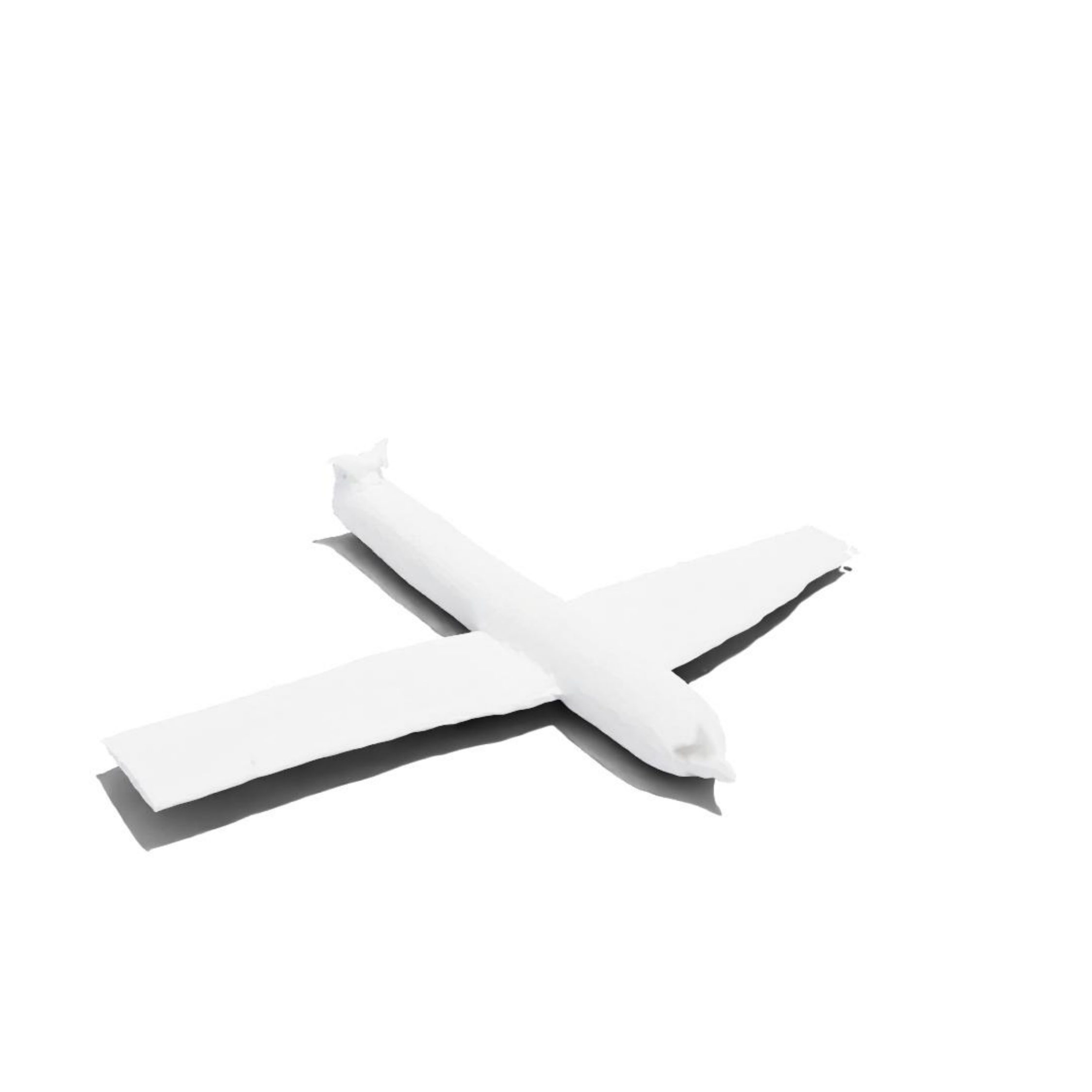}\includegraphics[width=0.08333333333333333\linewidth]{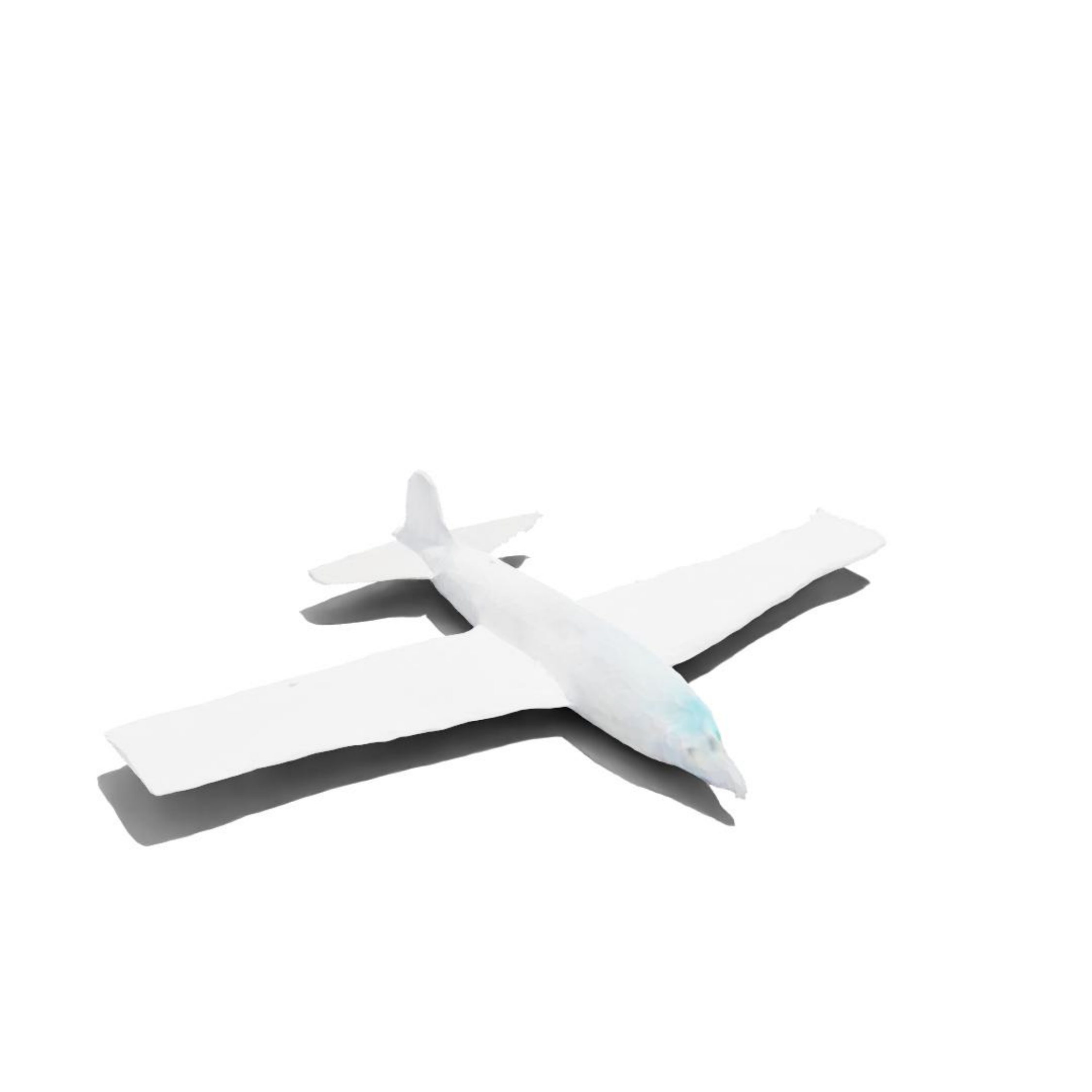}\includegraphics[width=0.08333333333333333\linewidth]{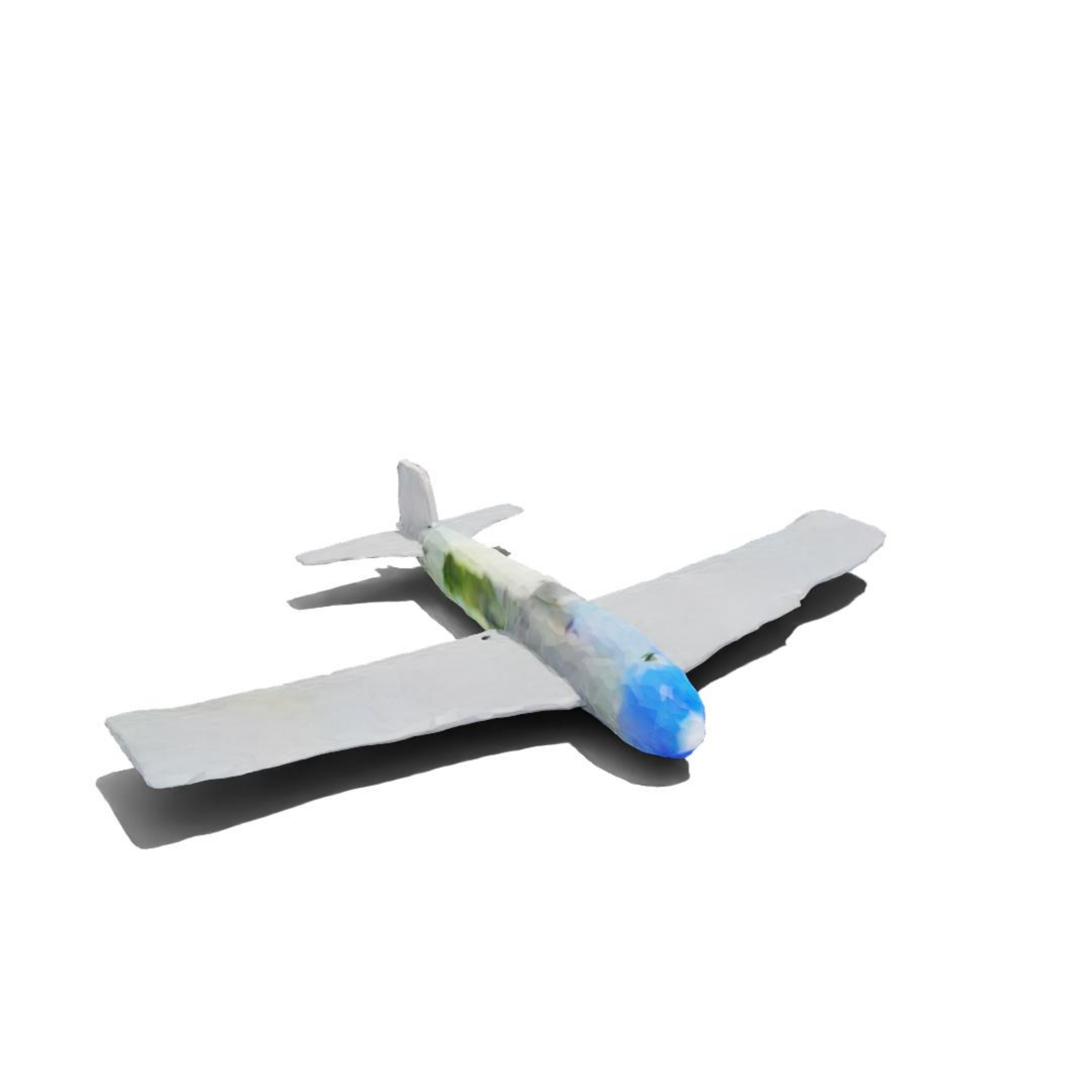}\\
\includegraphics[width=0.08333333333333333\linewidth]{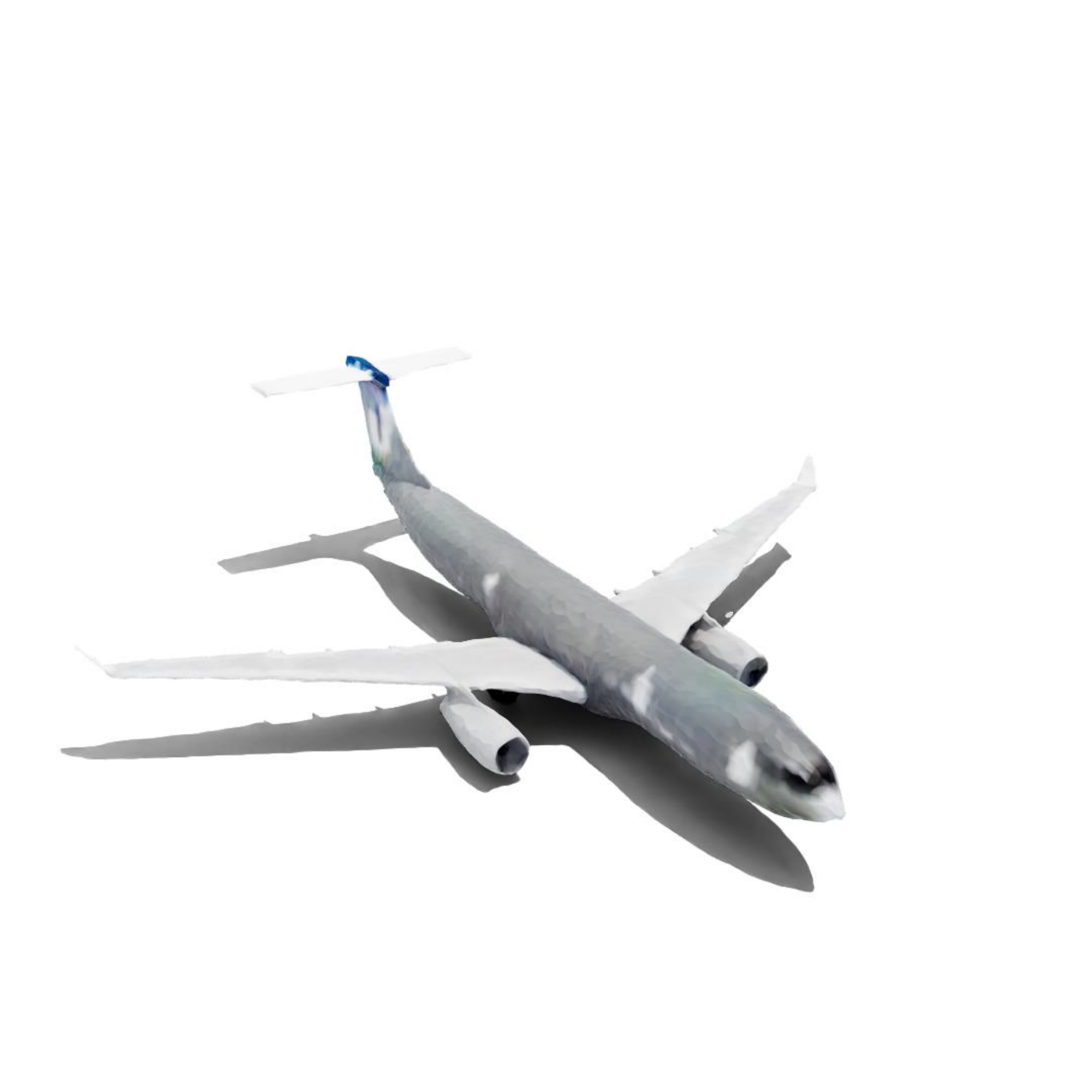}\includegraphics[width=0.08333333333333333\linewidth]{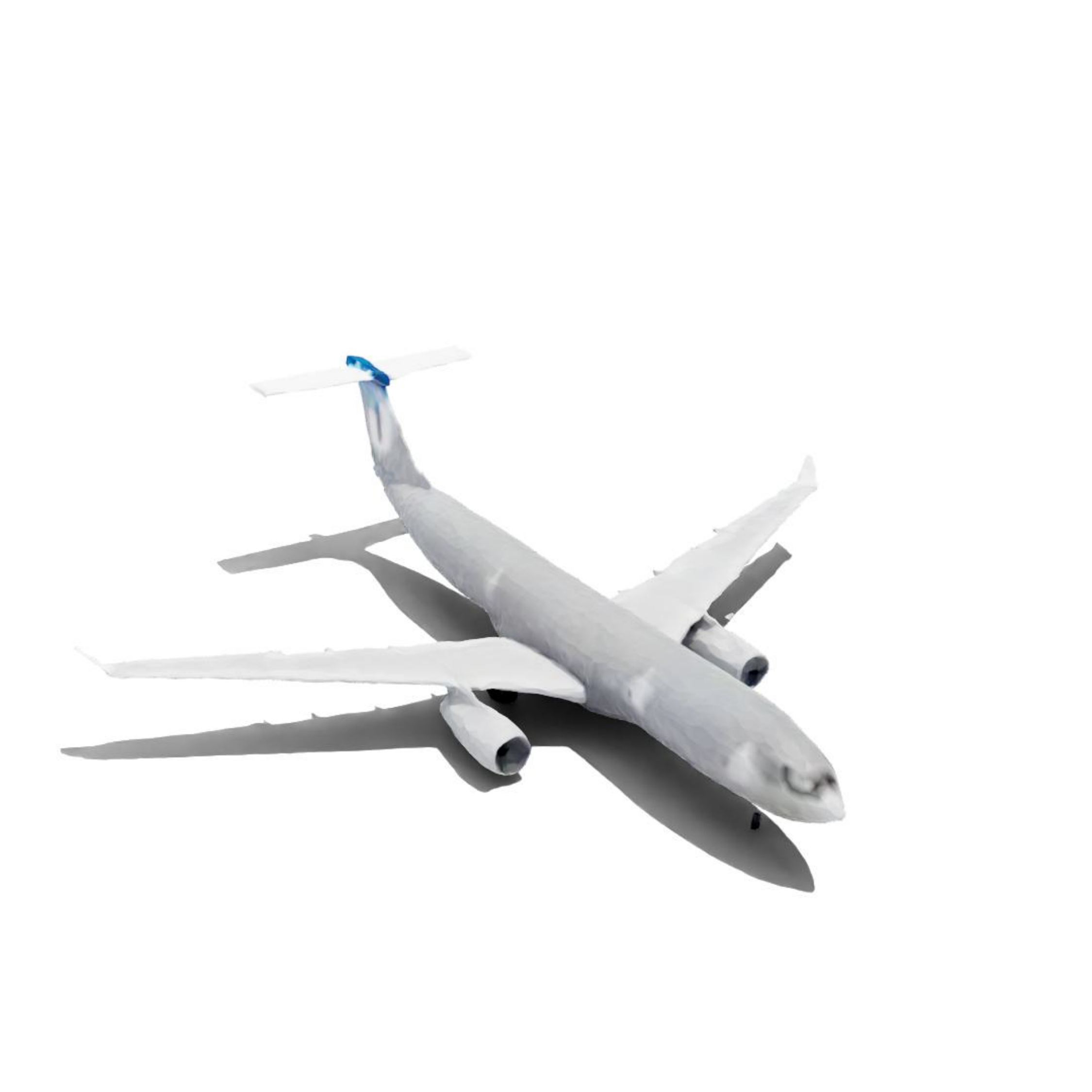}\includegraphics[width=0.08333333333333333\linewidth]{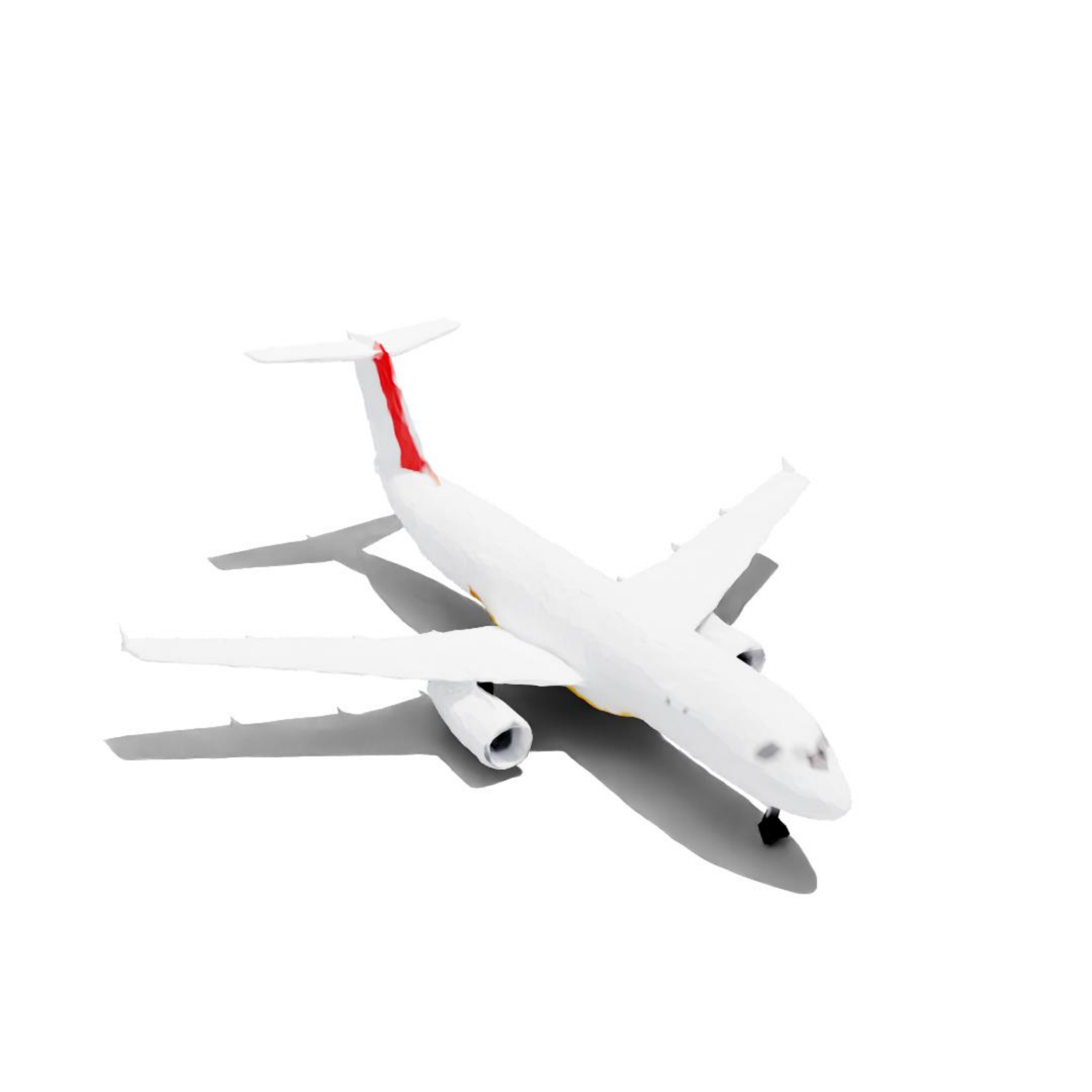}\includegraphics[width=0.08333333333333333\linewidth]{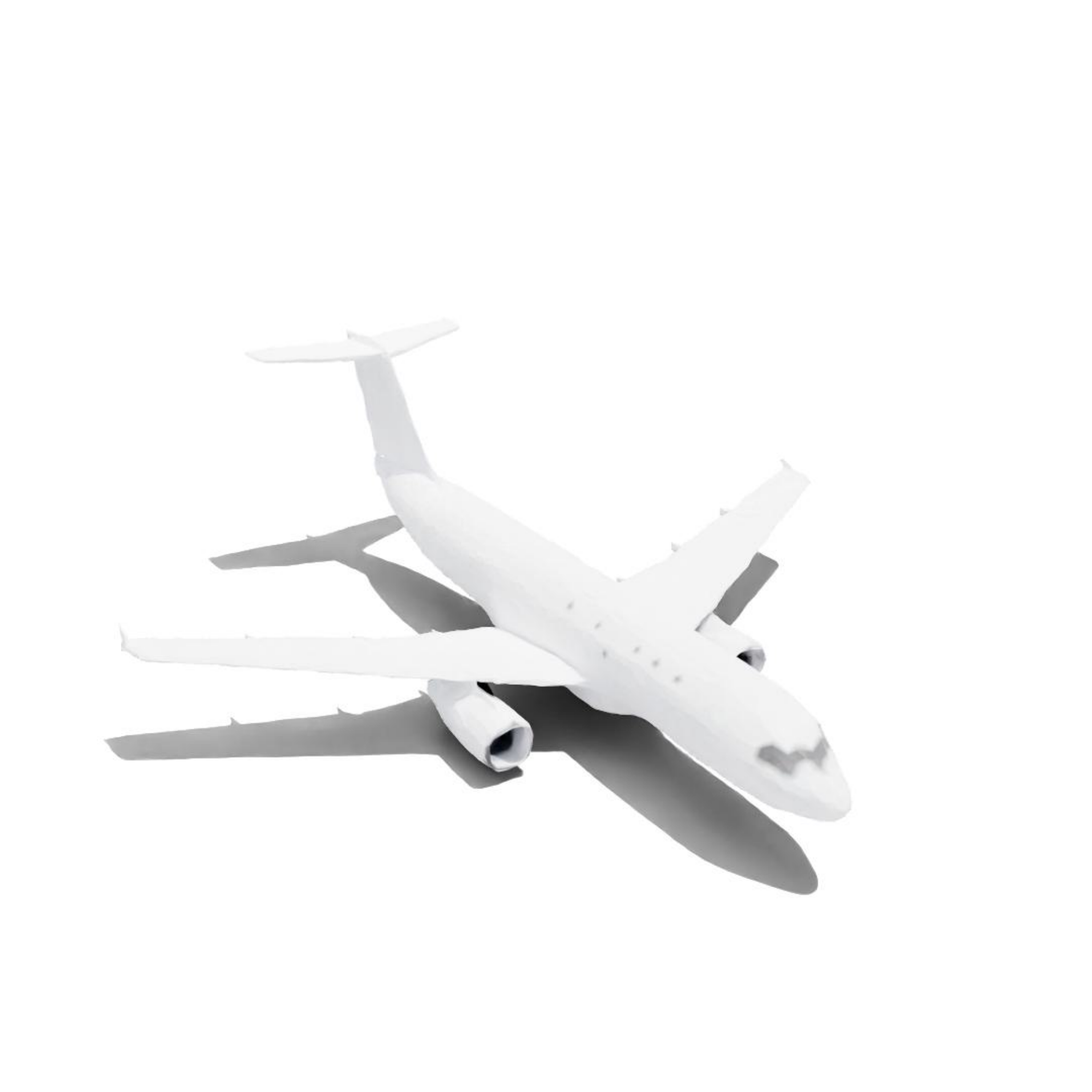}\includegraphics[width=0.08333333333333333\linewidth]{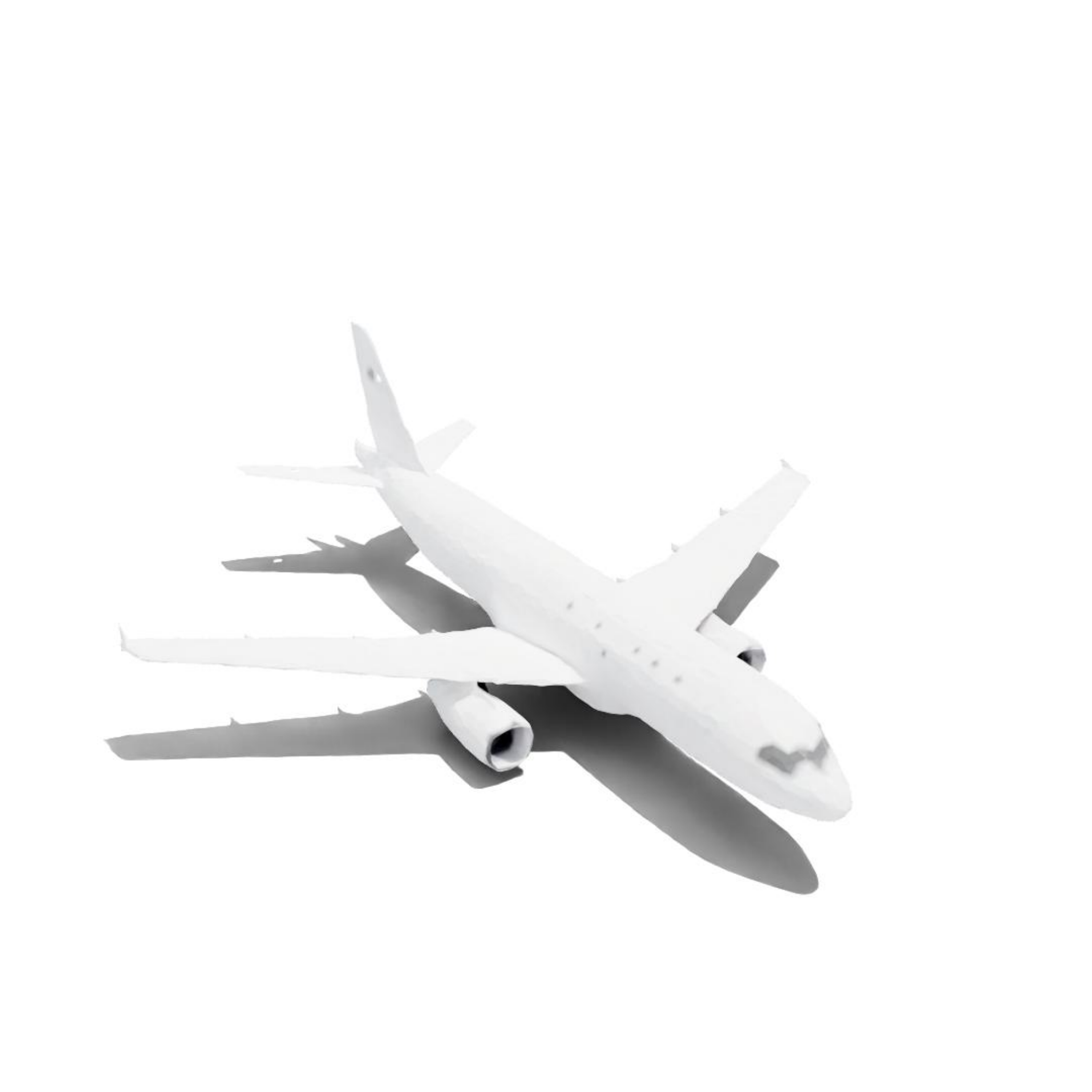}\includegraphics[width=0.08333333333333333\linewidth]{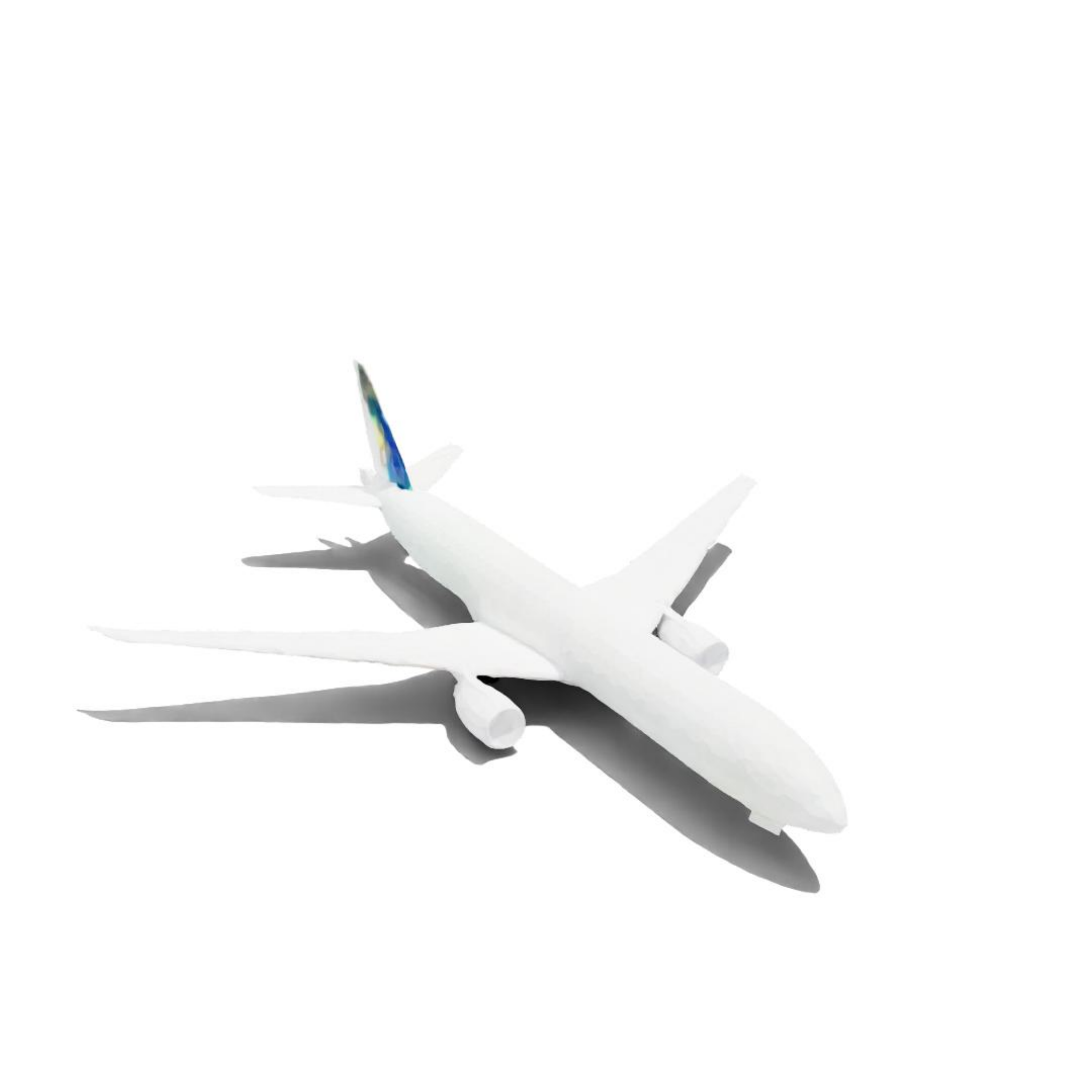}\includegraphics[width=0.08333333333333333\linewidth]{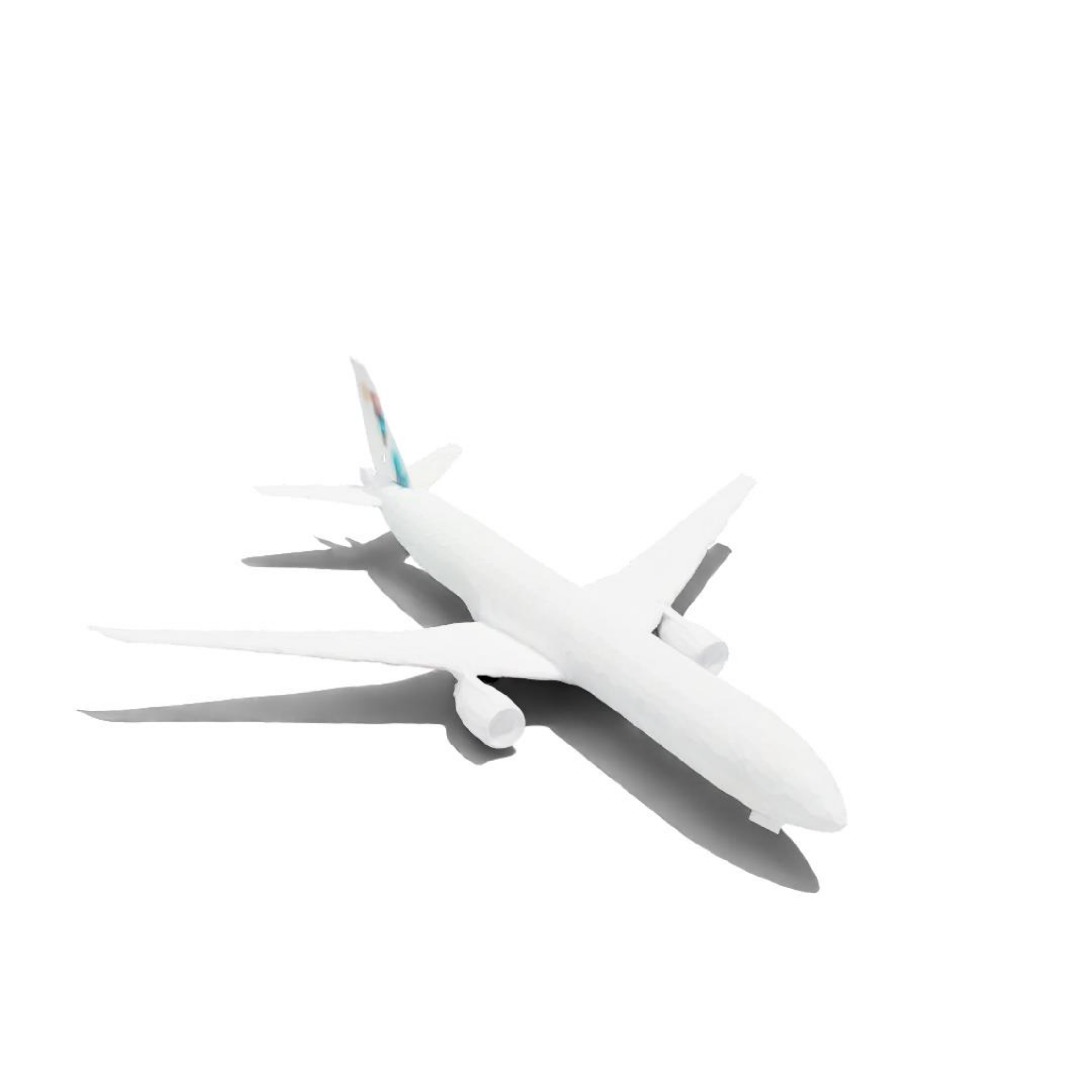}\includegraphics[width=0.08333333333333333\linewidth]{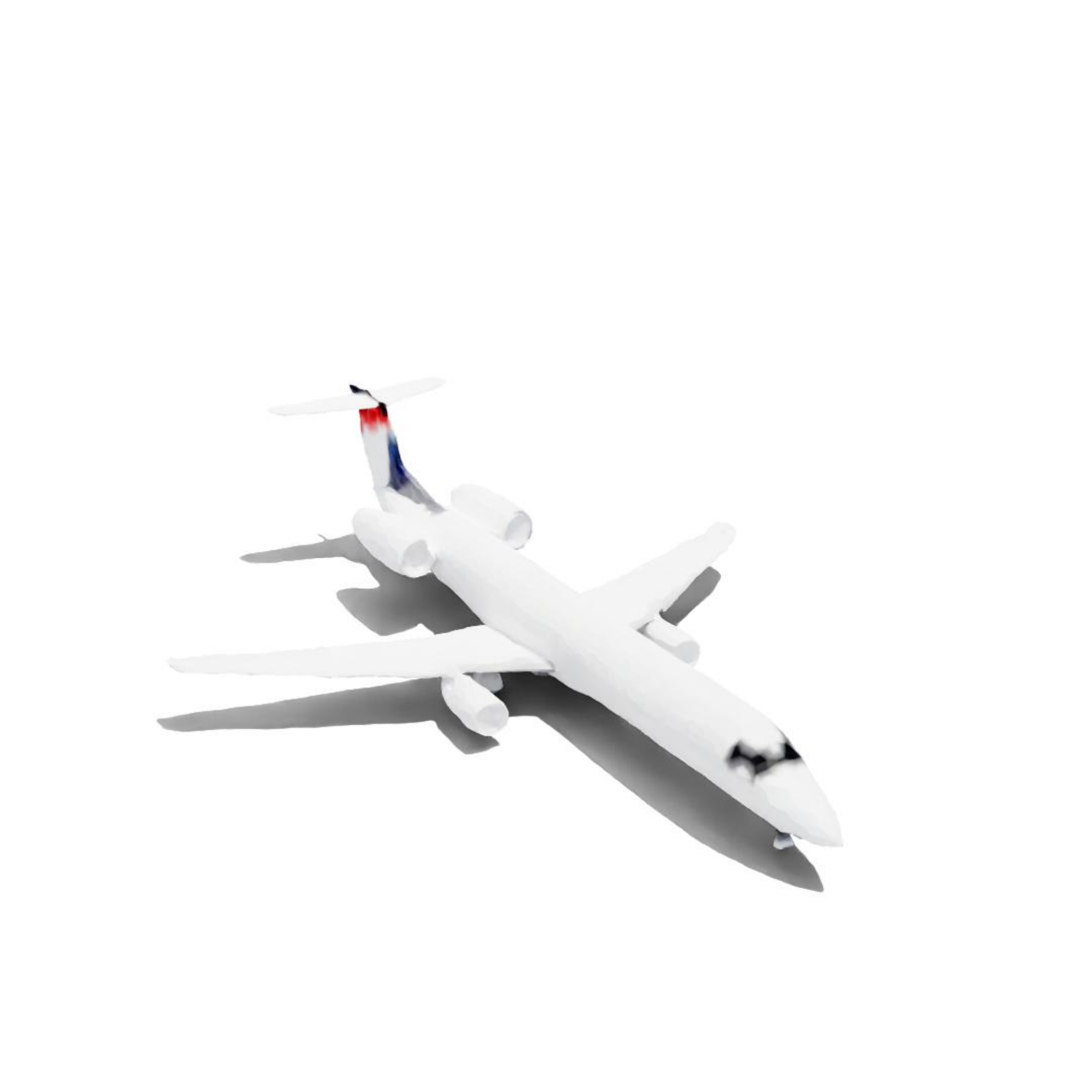}\includegraphics[width=0.08333333333333333\linewidth]{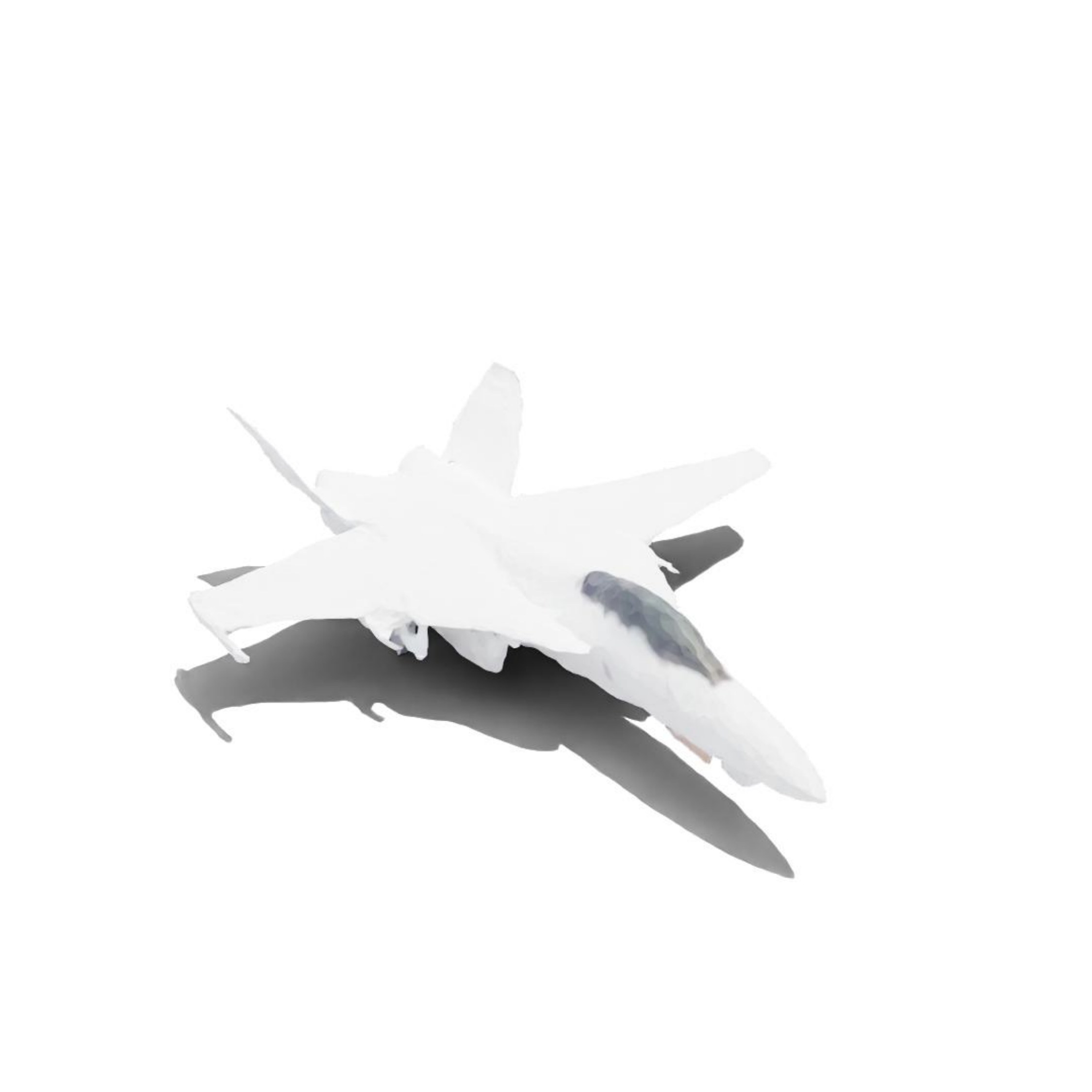}\includegraphics[width=0.08333333333333333\linewidth]{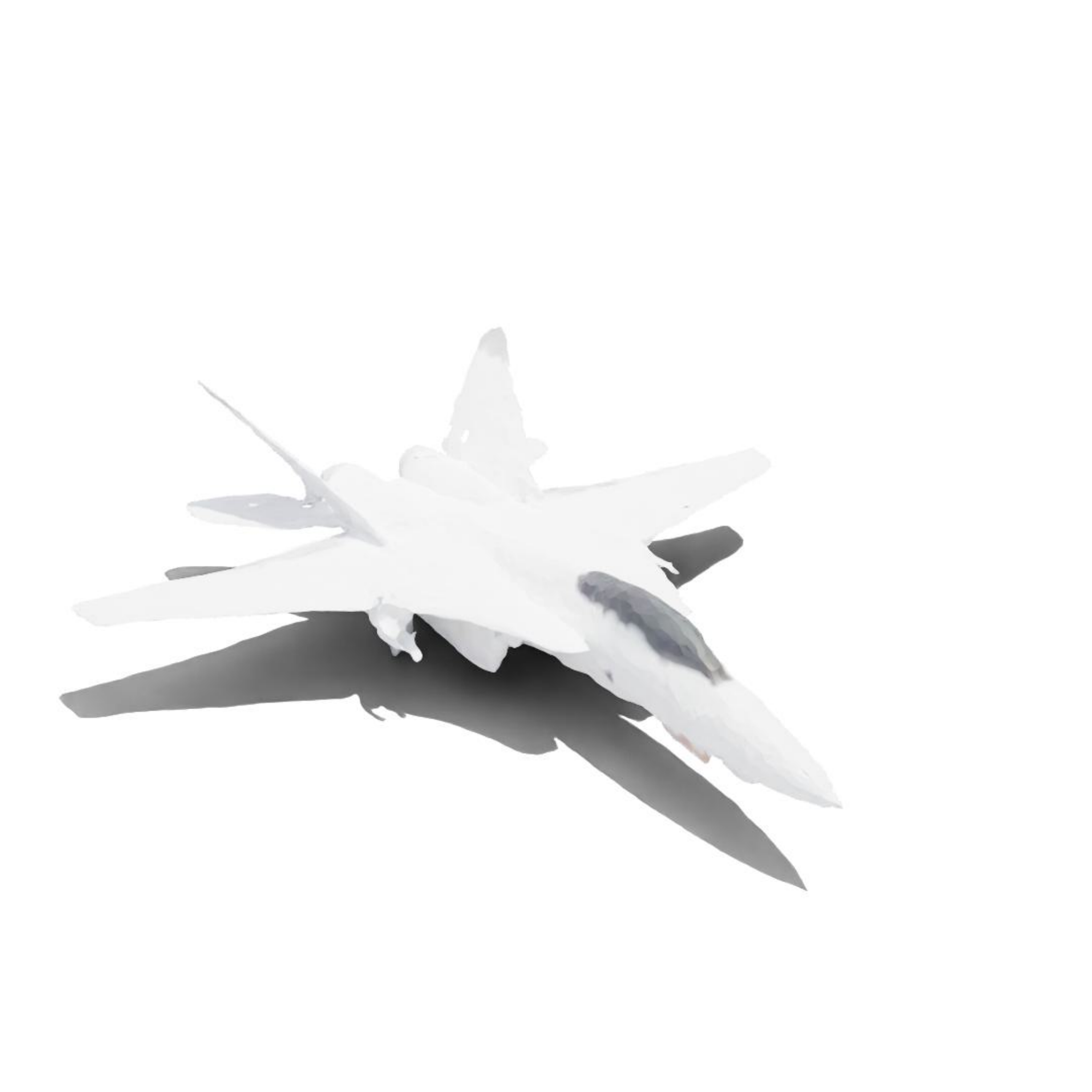}\includegraphics[width=0.08333333333333333\linewidth]{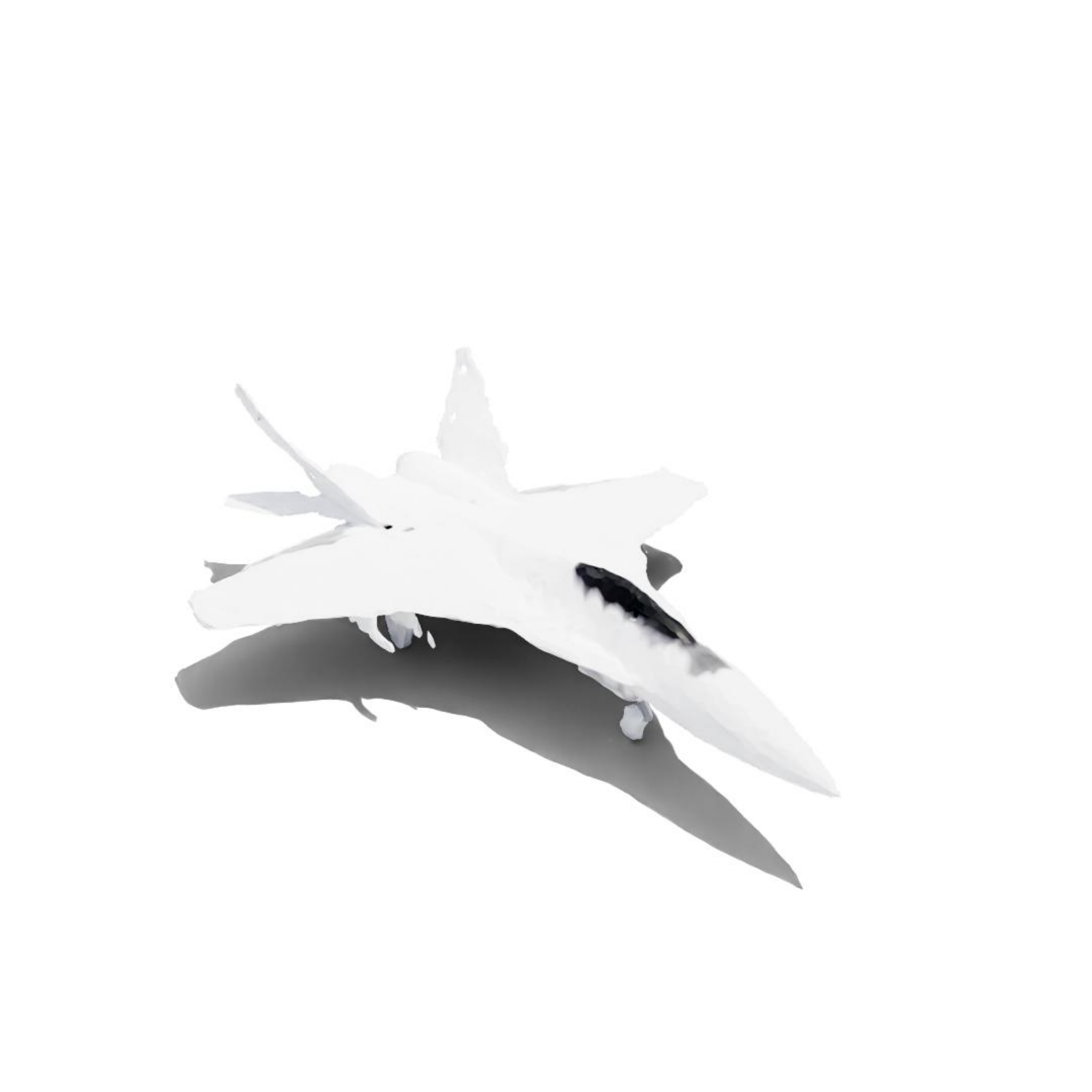}\includegraphics[width=0.08333333333333333\linewidth]{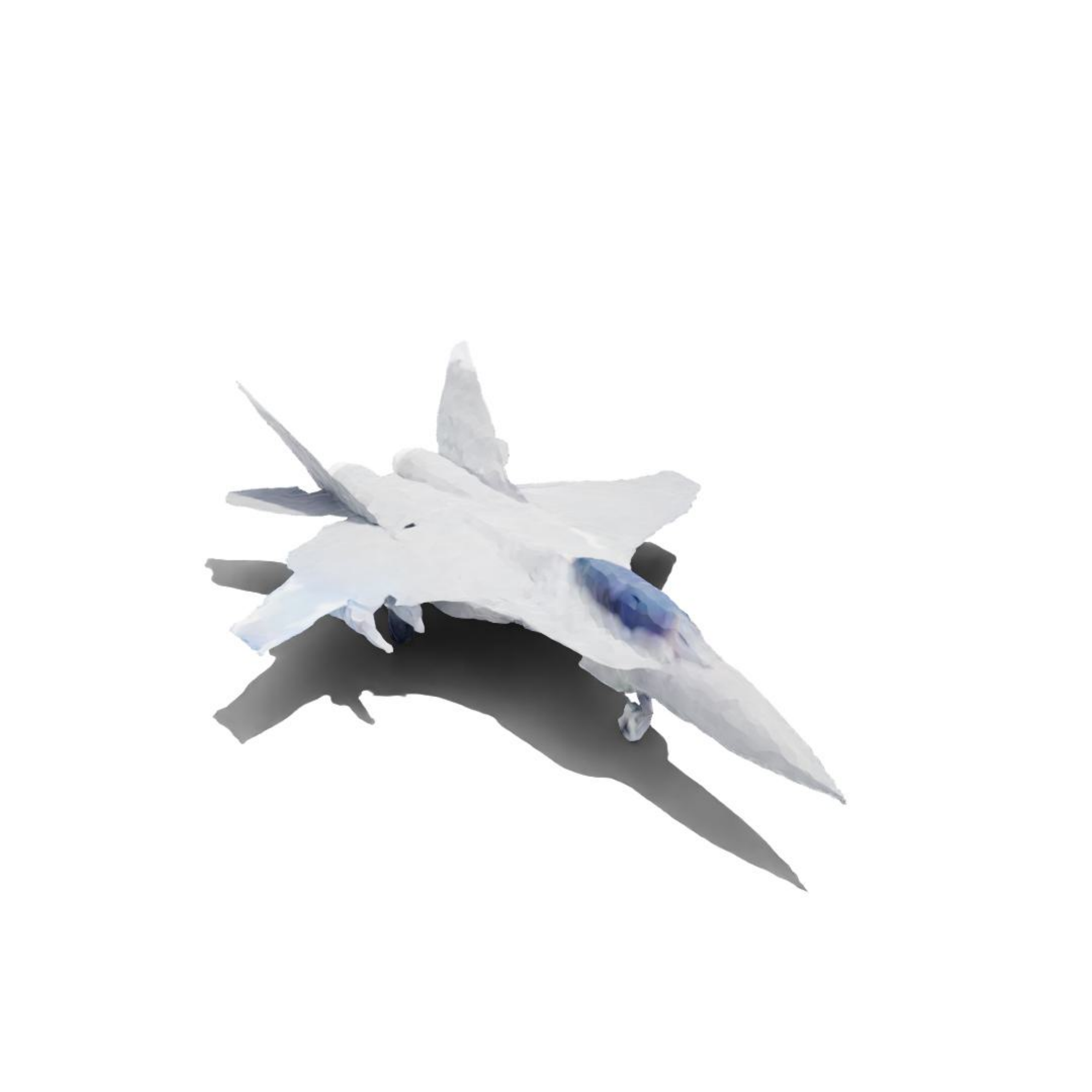}\\
\includegraphics[width=0.08333333333333333\linewidth]{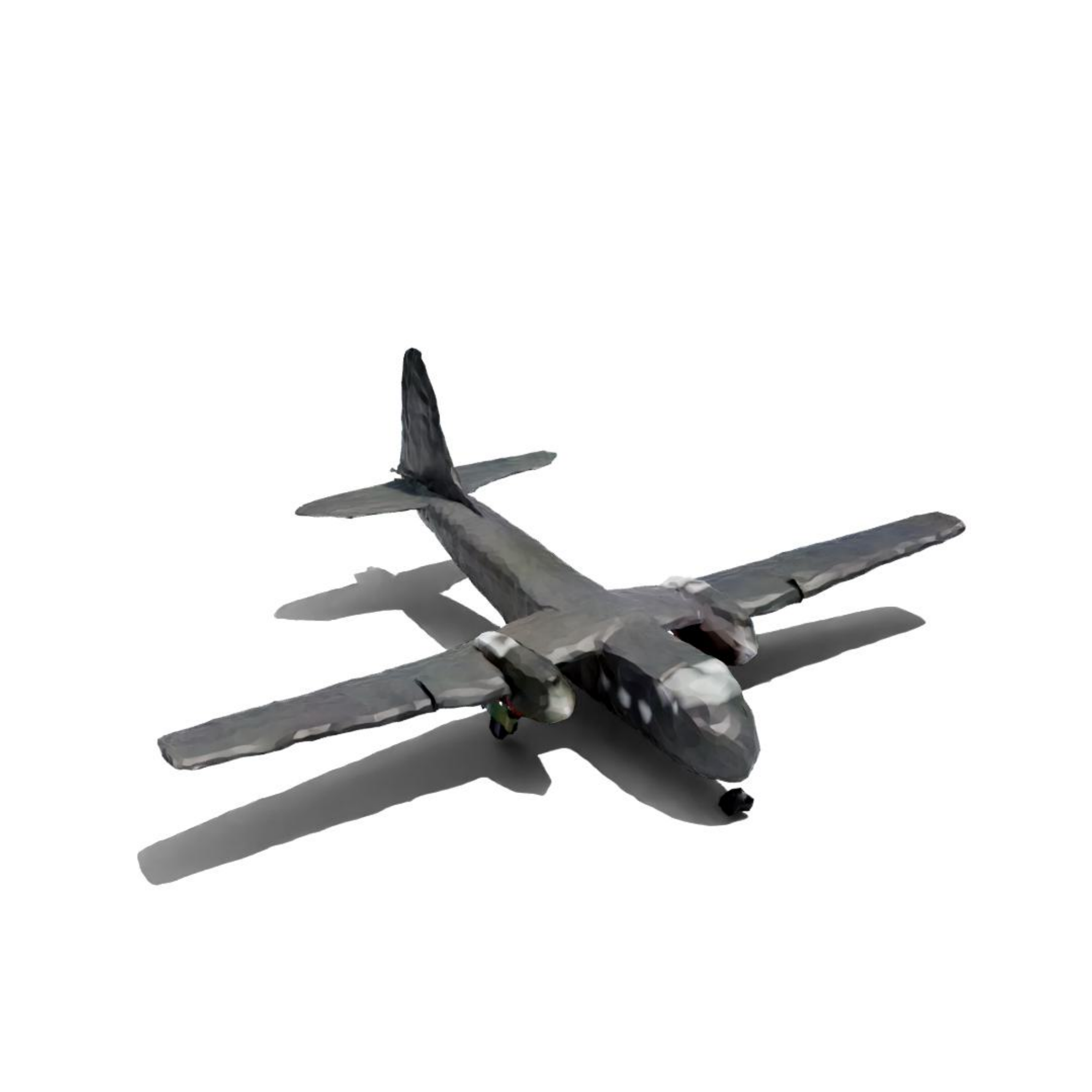}\includegraphics[width=0.08333333333333333\linewidth]{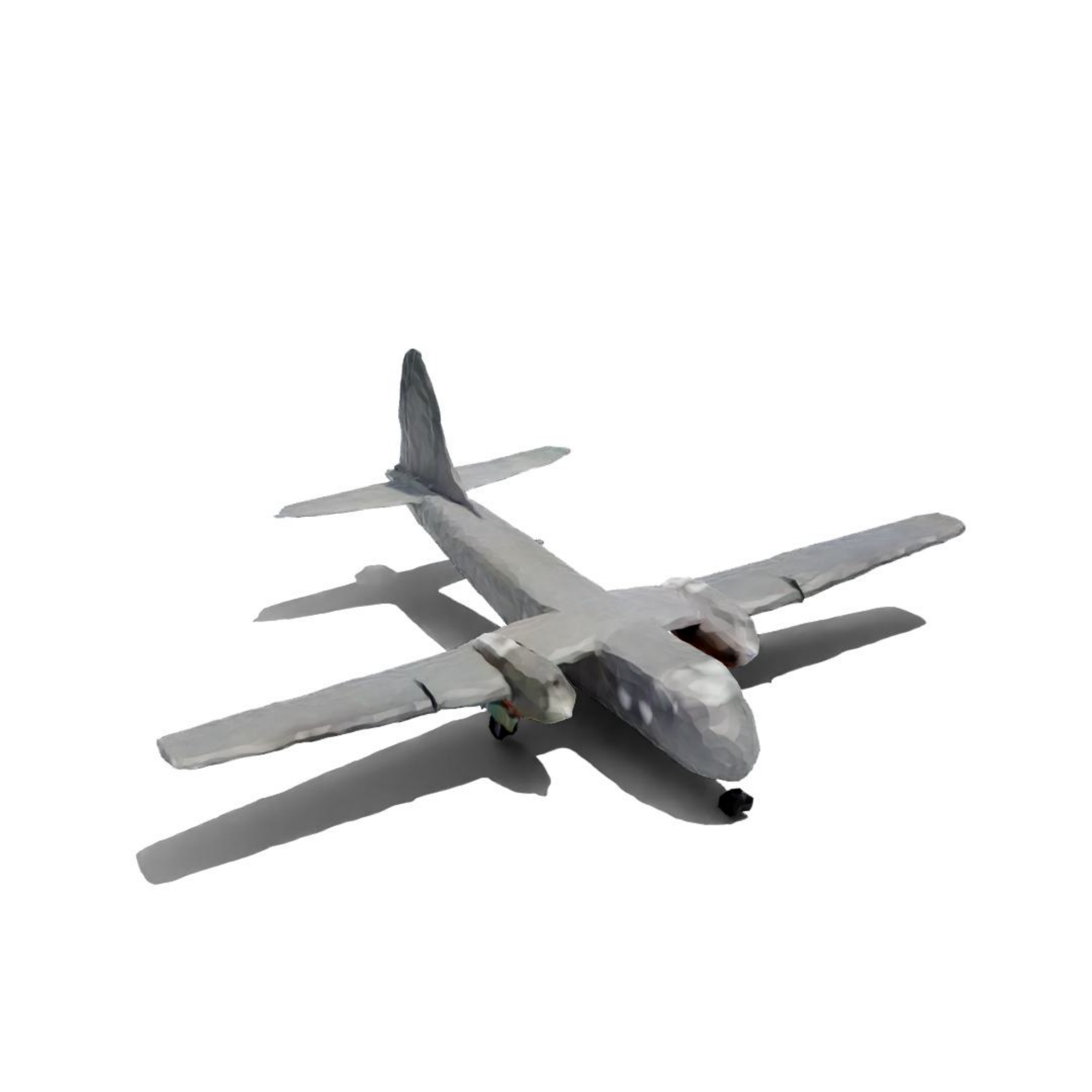}\includegraphics[width=0.08333333333333333\linewidth]{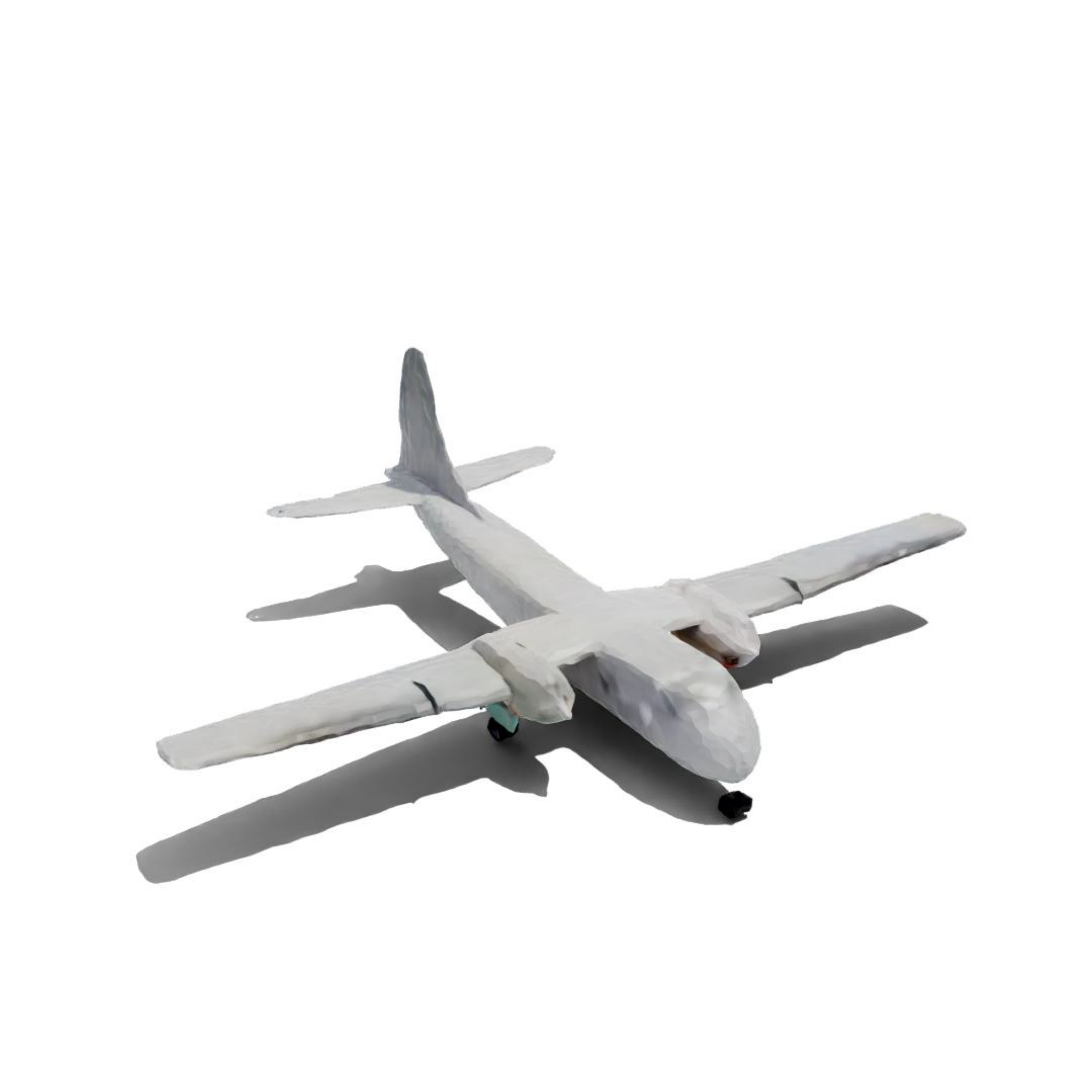}\includegraphics[width=0.08333333333333333\linewidth]{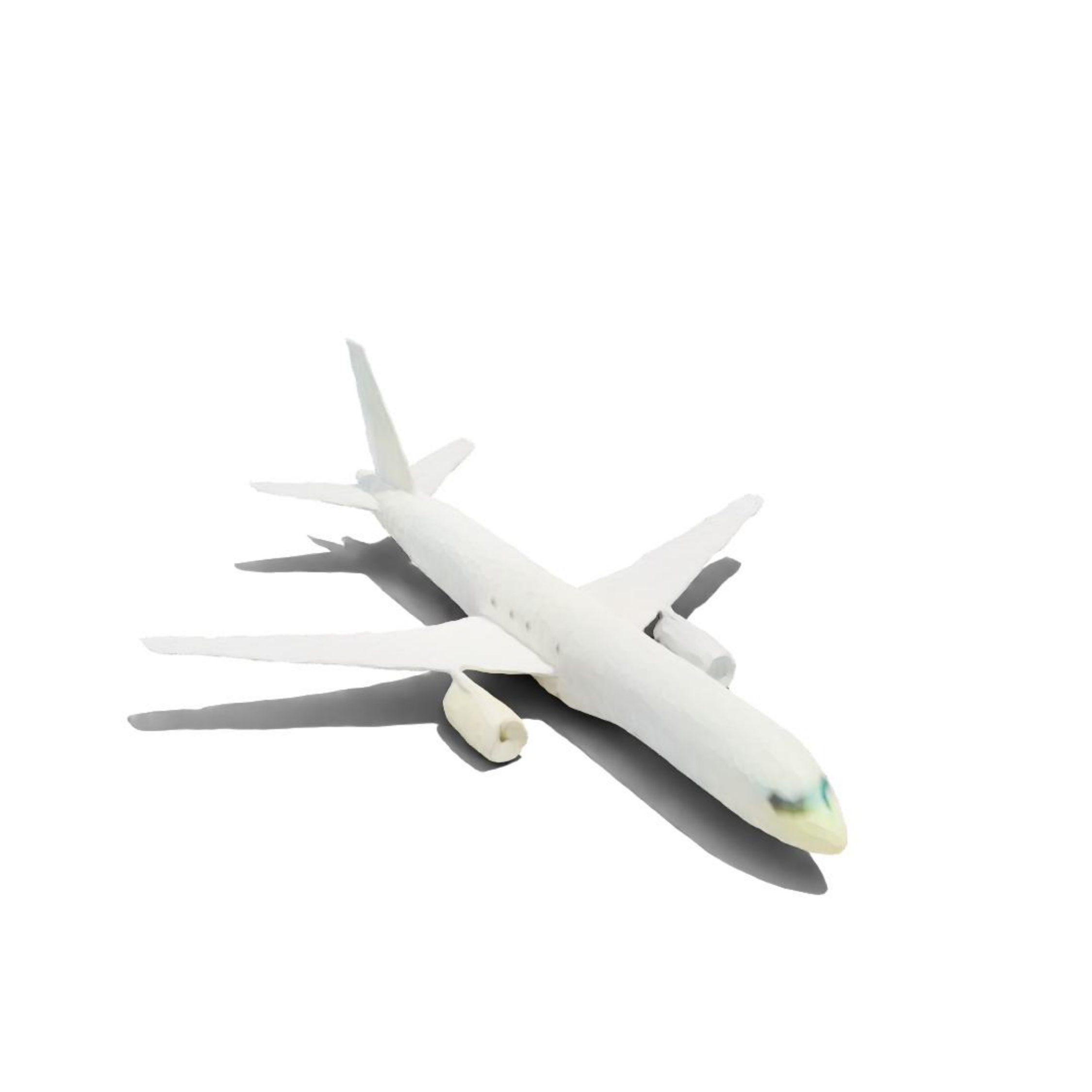}\includegraphics[width=0.08333333333333333\linewidth]{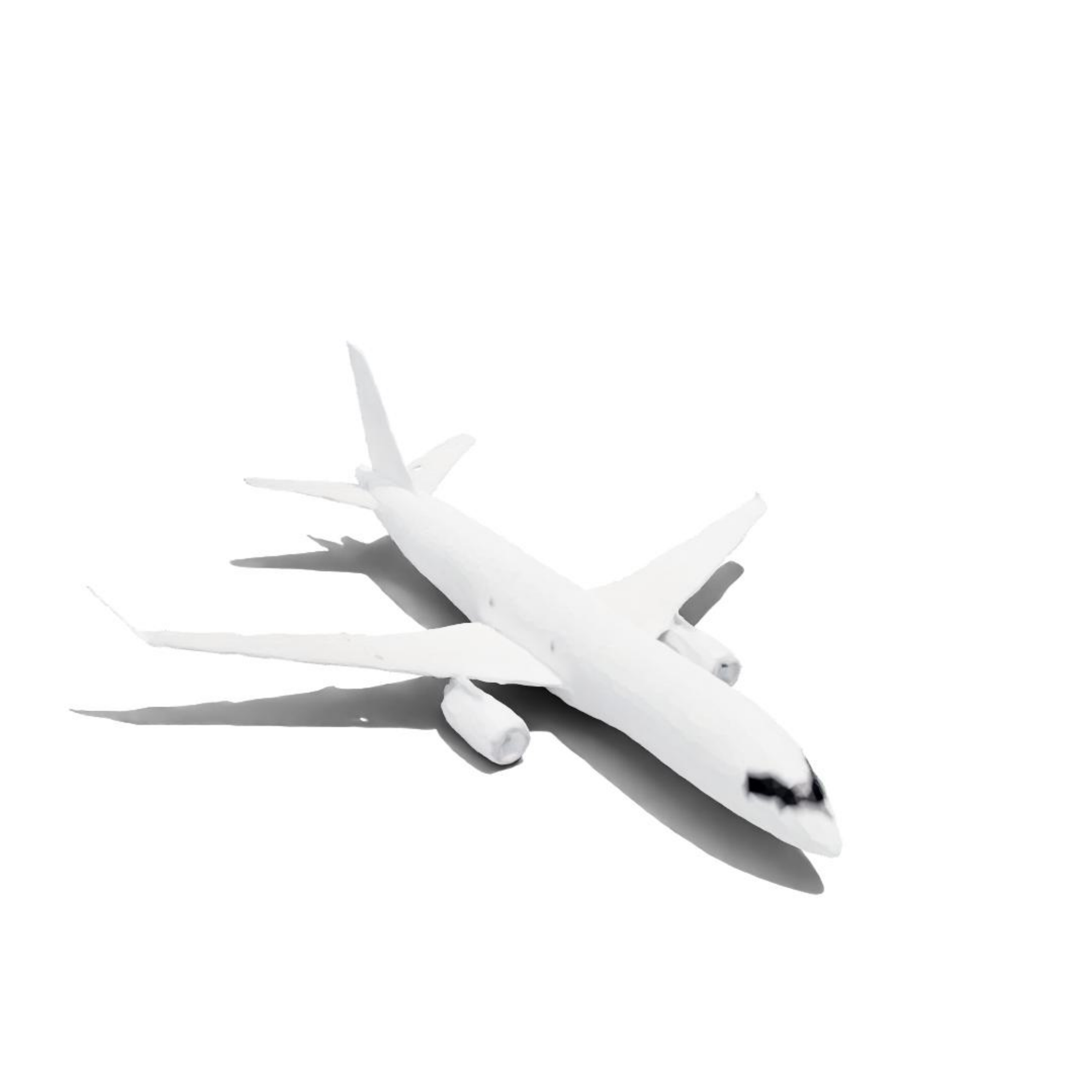}\includegraphics[width=0.08333333333333333\linewidth]{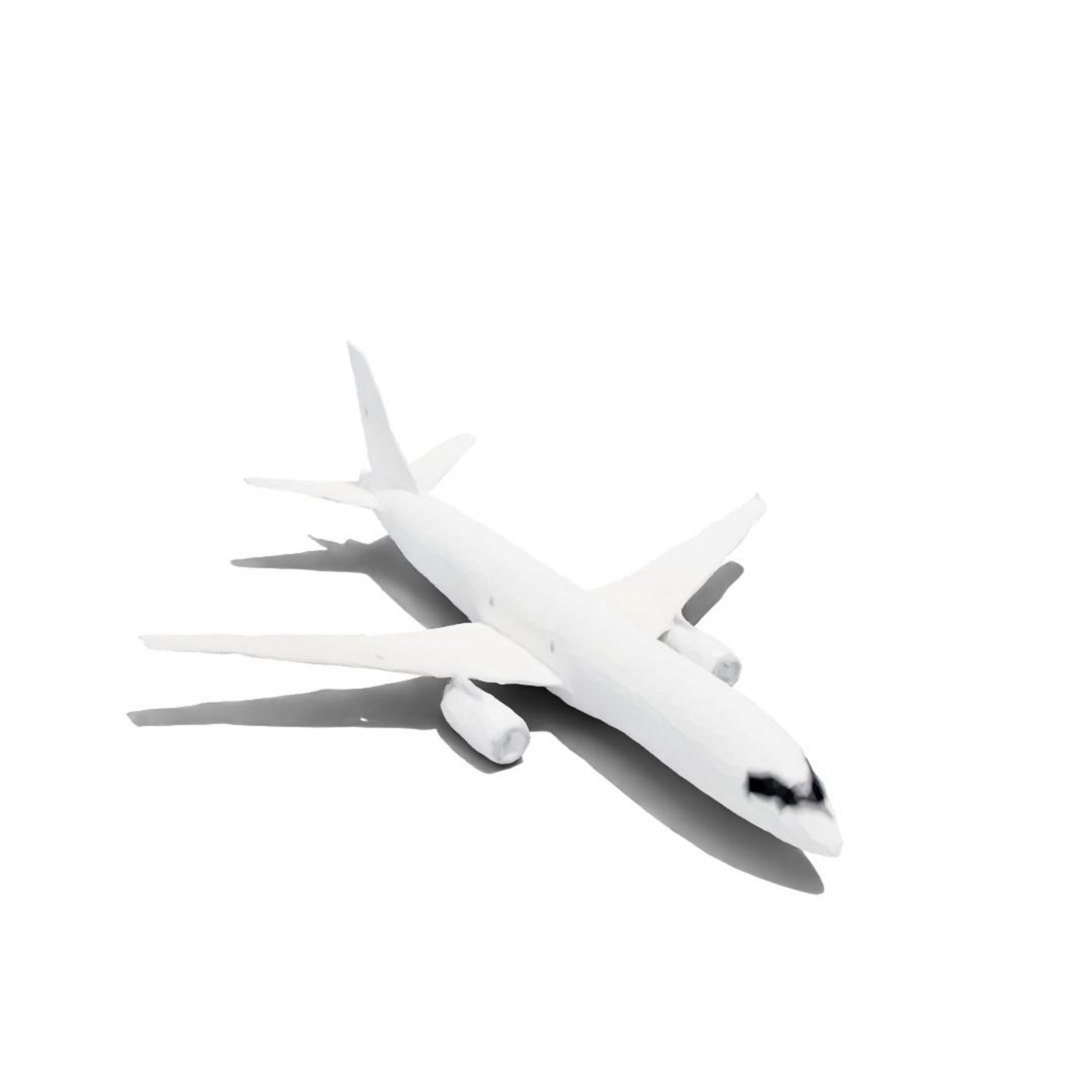}\includegraphics[width=0.08333333333333333\linewidth]{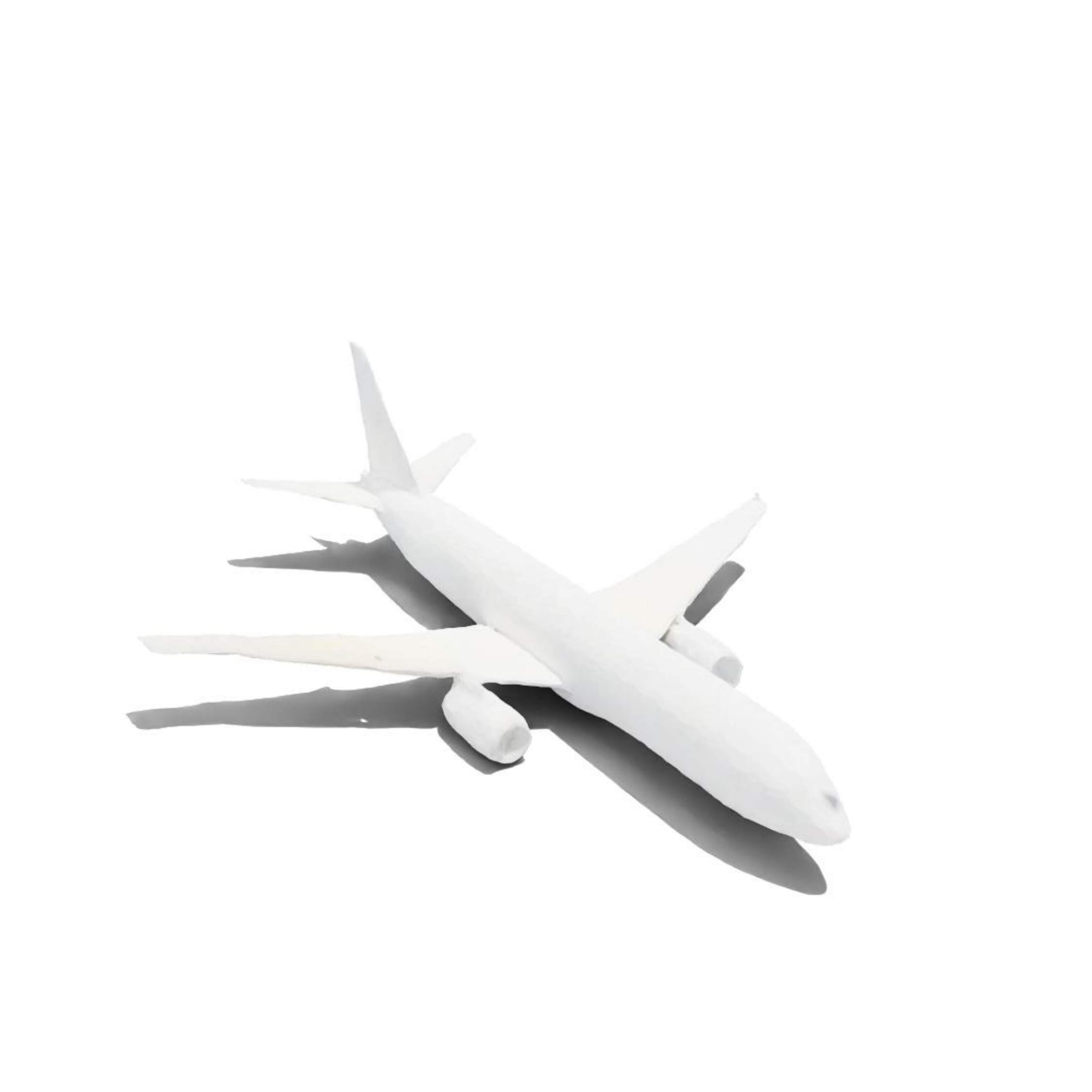}\includegraphics[width=0.08333333333333333\linewidth]{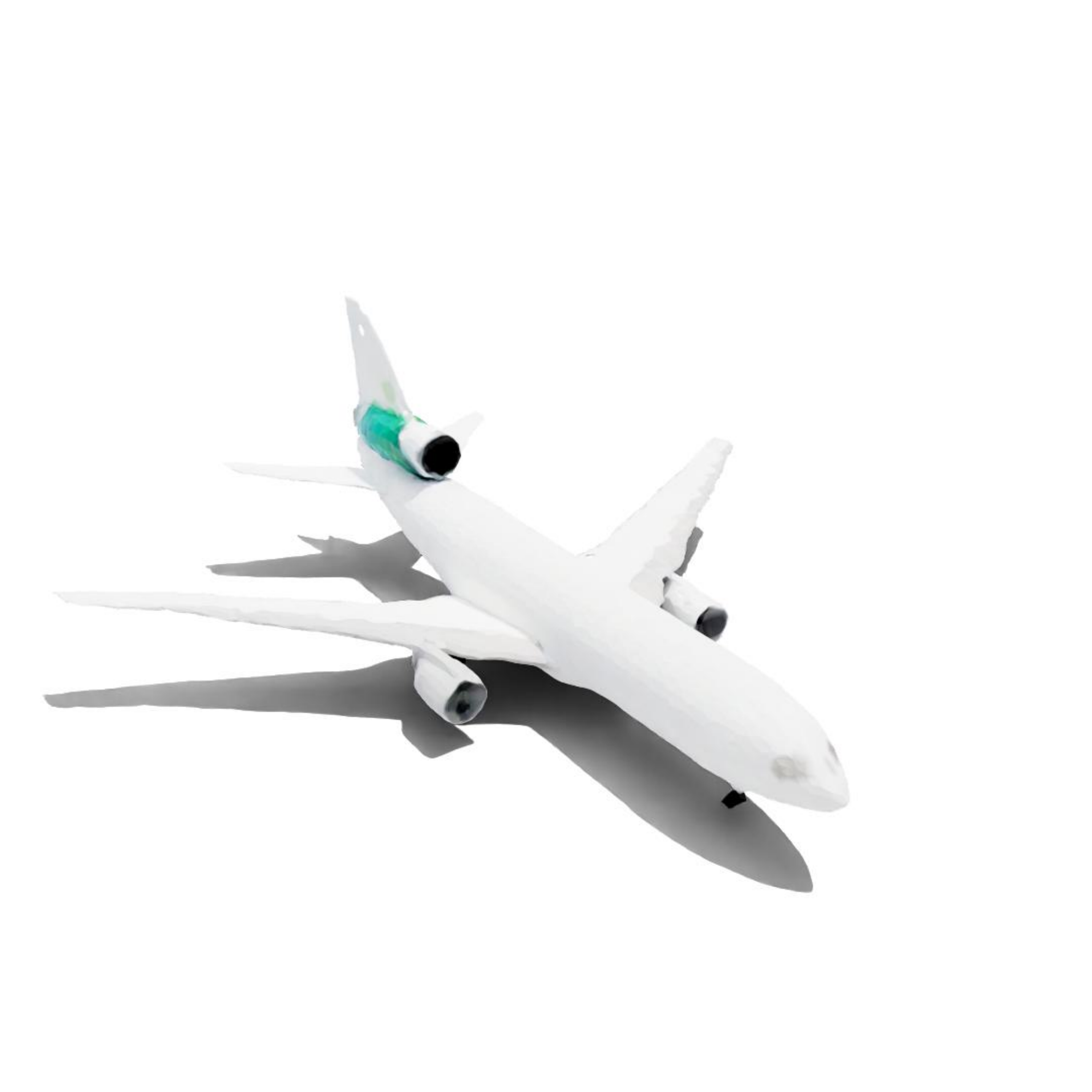}\includegraphics[width=0.08333333333333333\linewidth]{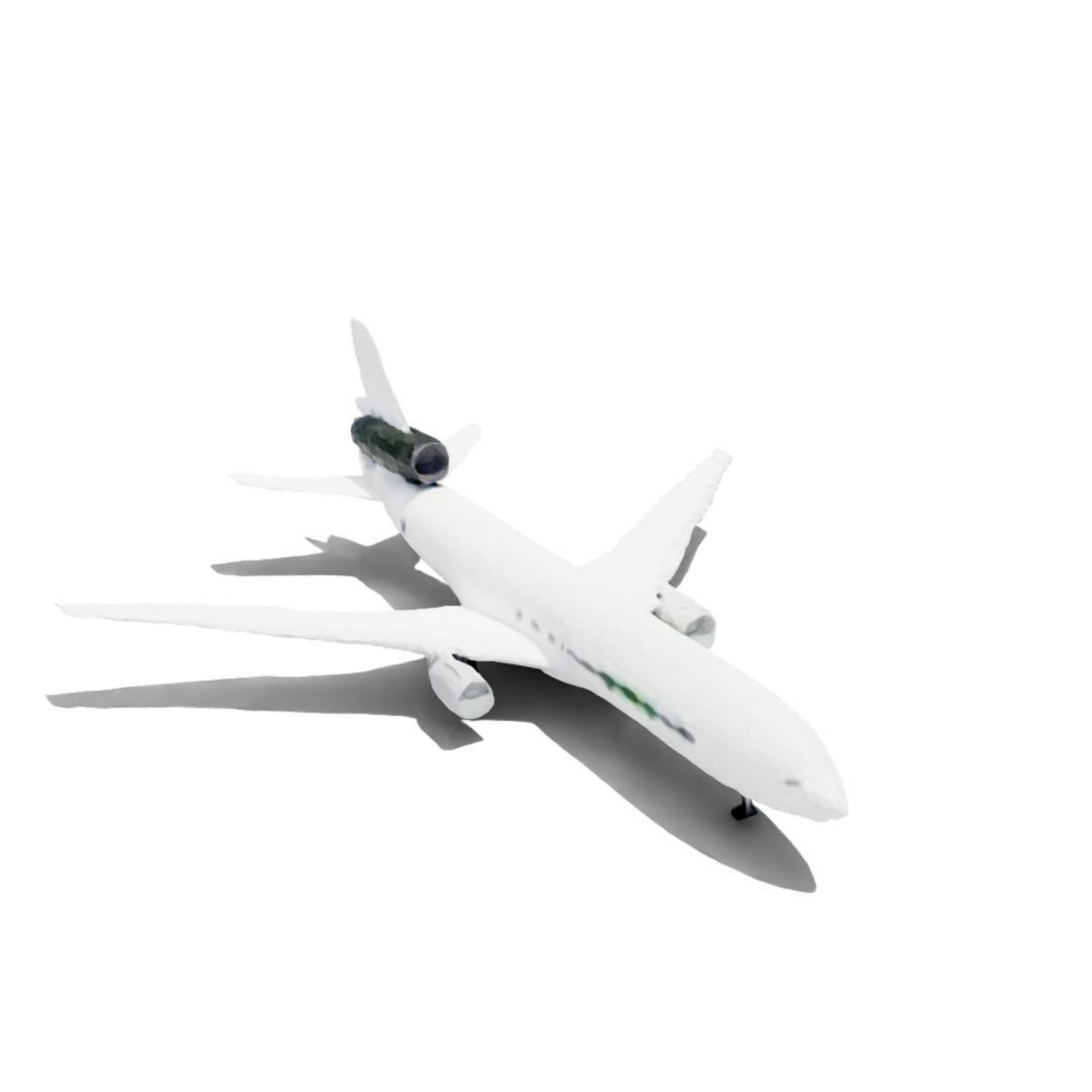}\includegraphics[width=0.08333333333333333\linewidth]{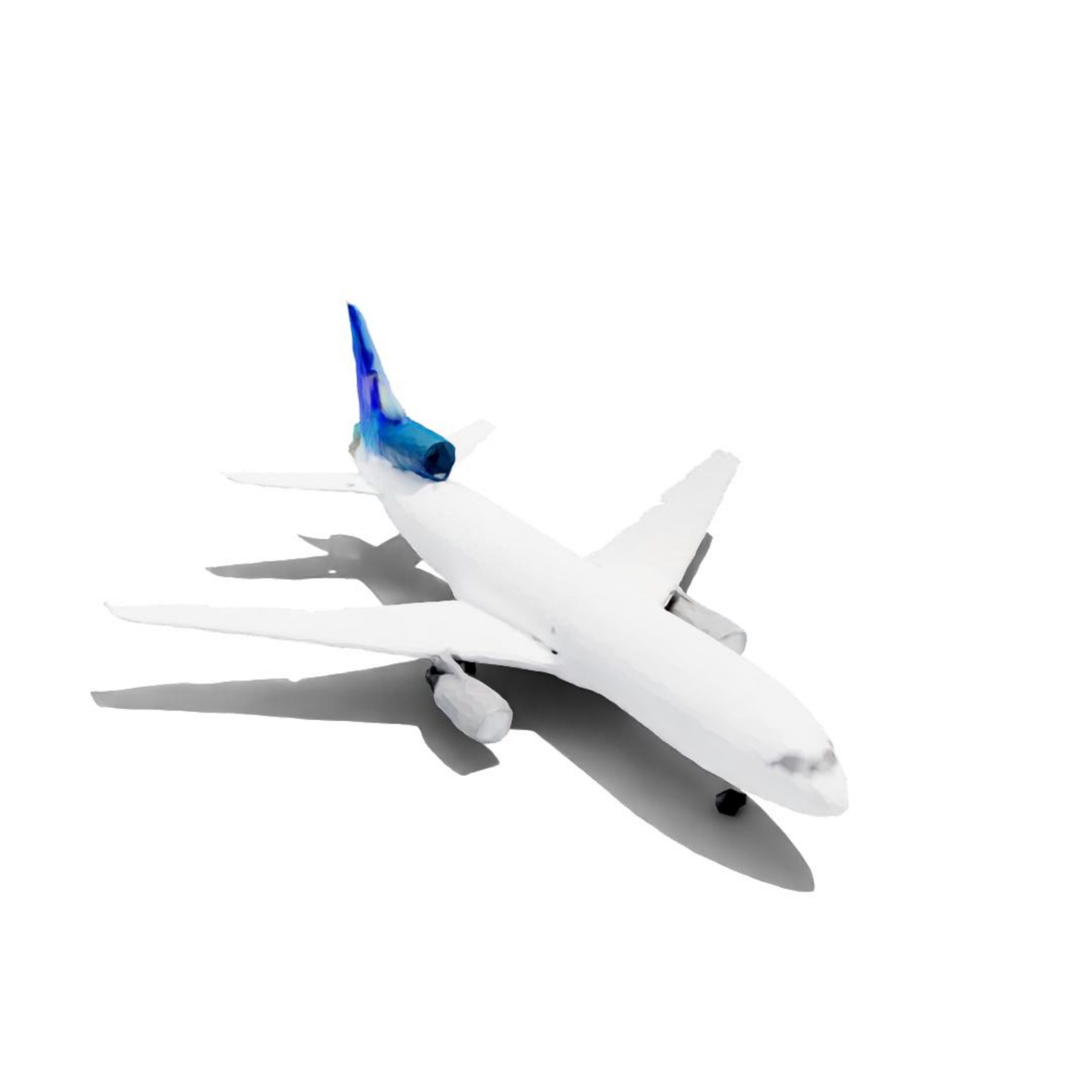}\includegraphics[width=0.08333333333333333\linewidth]{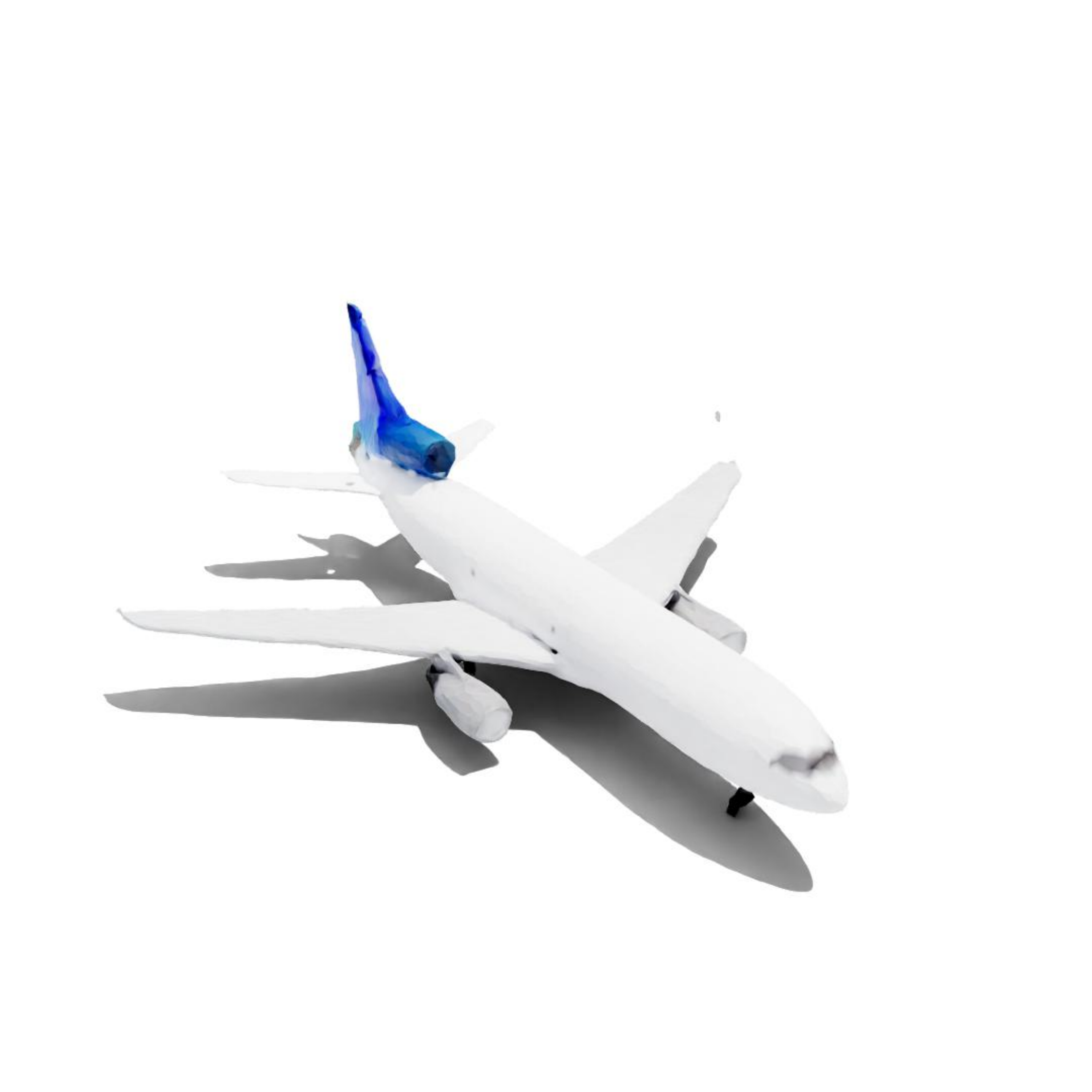}\includegraphics[width=0.08333333333333333\linewidth]{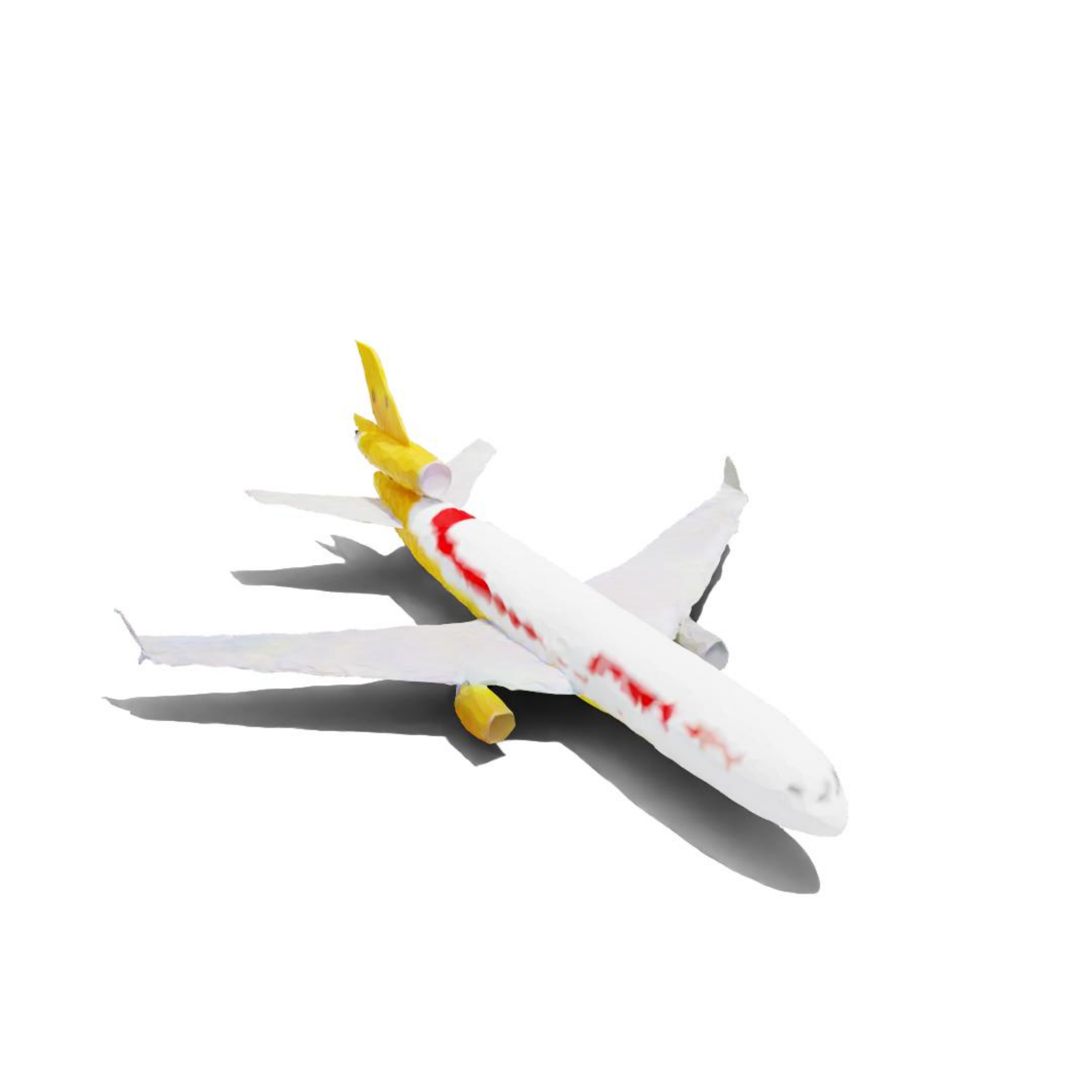}\\
\includegraphics[width=0.08333333333333333\linewidth]{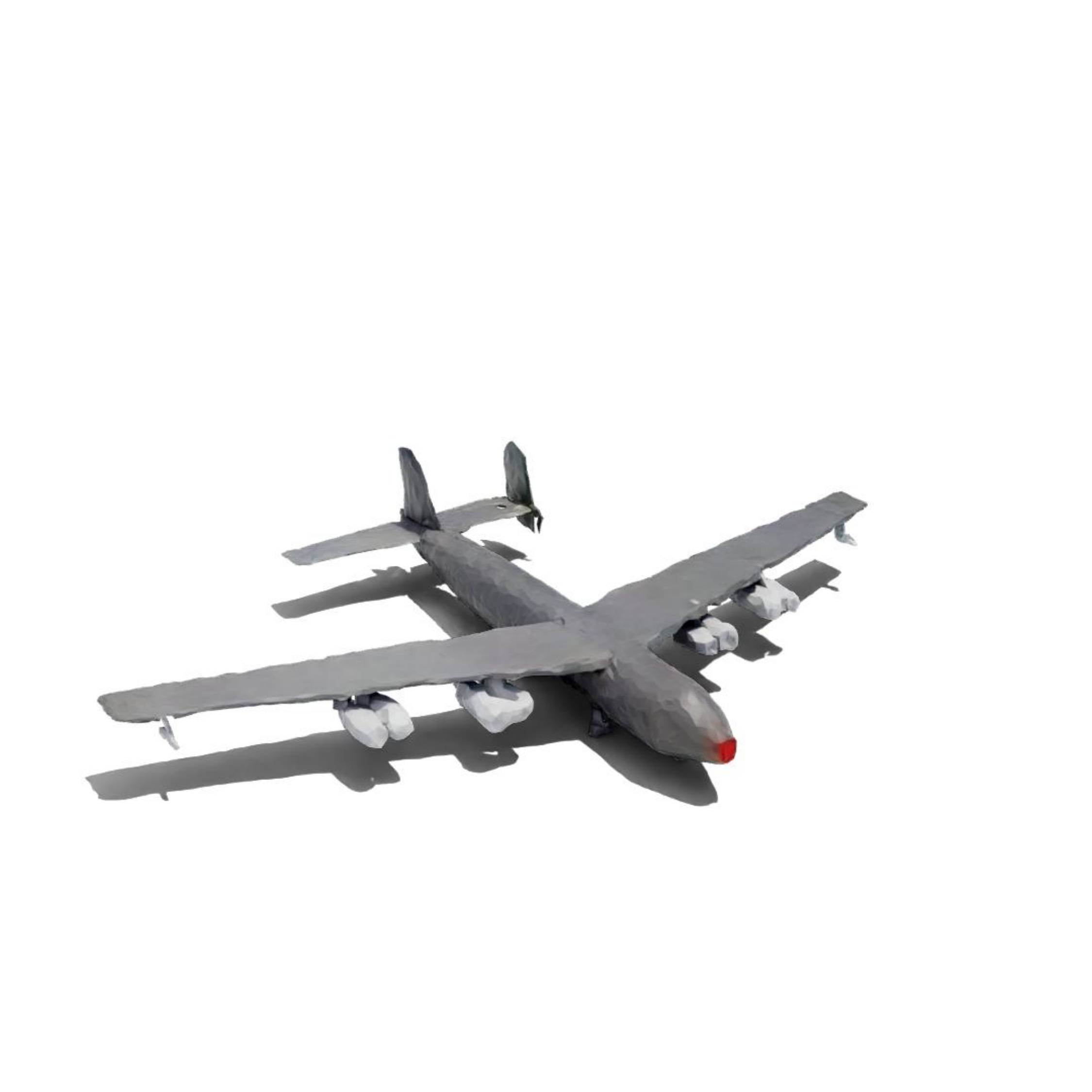}\includegraphics[width=0.08333333333333333\linewidth]{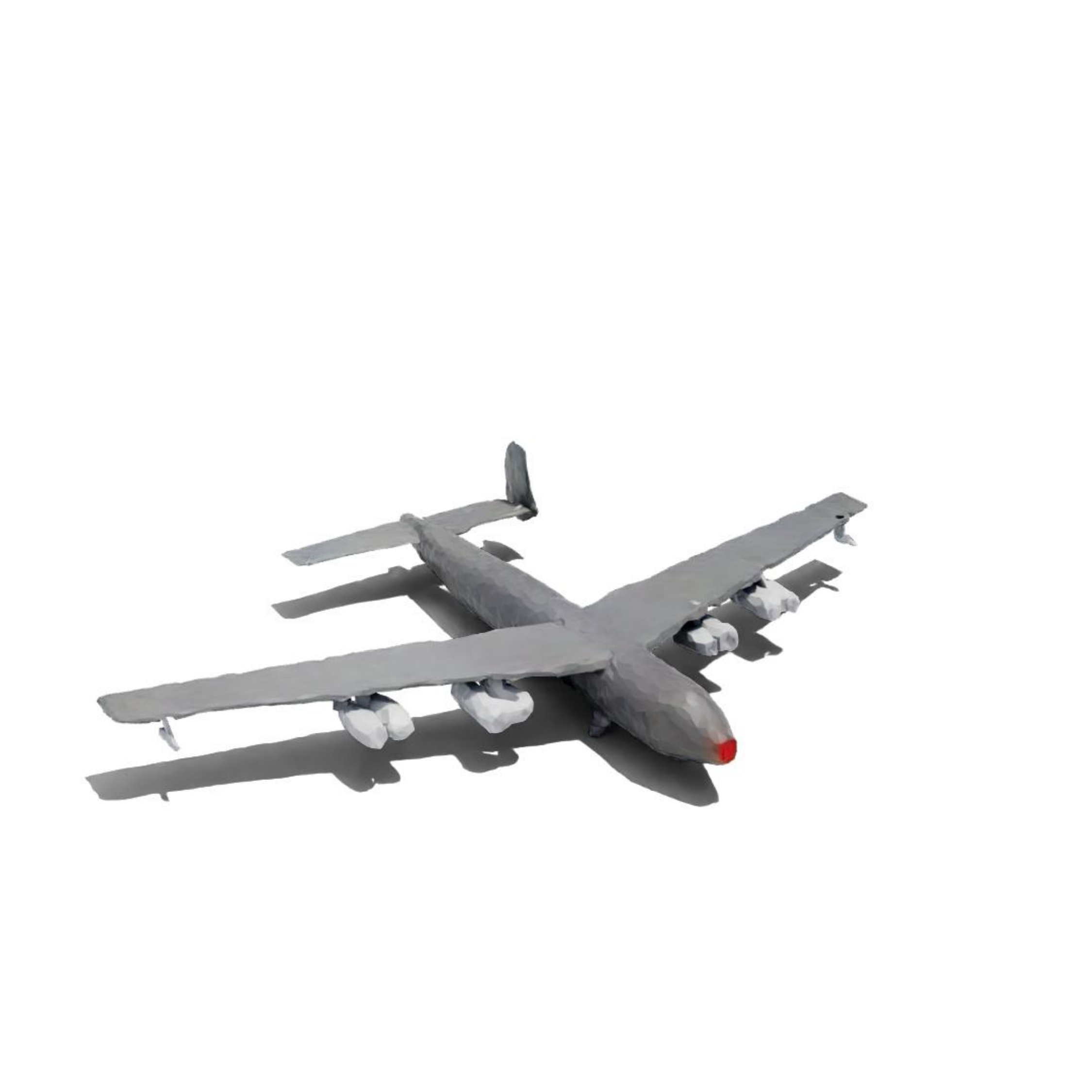}\includegraphics[width=0.08333333333333333\linewidth]{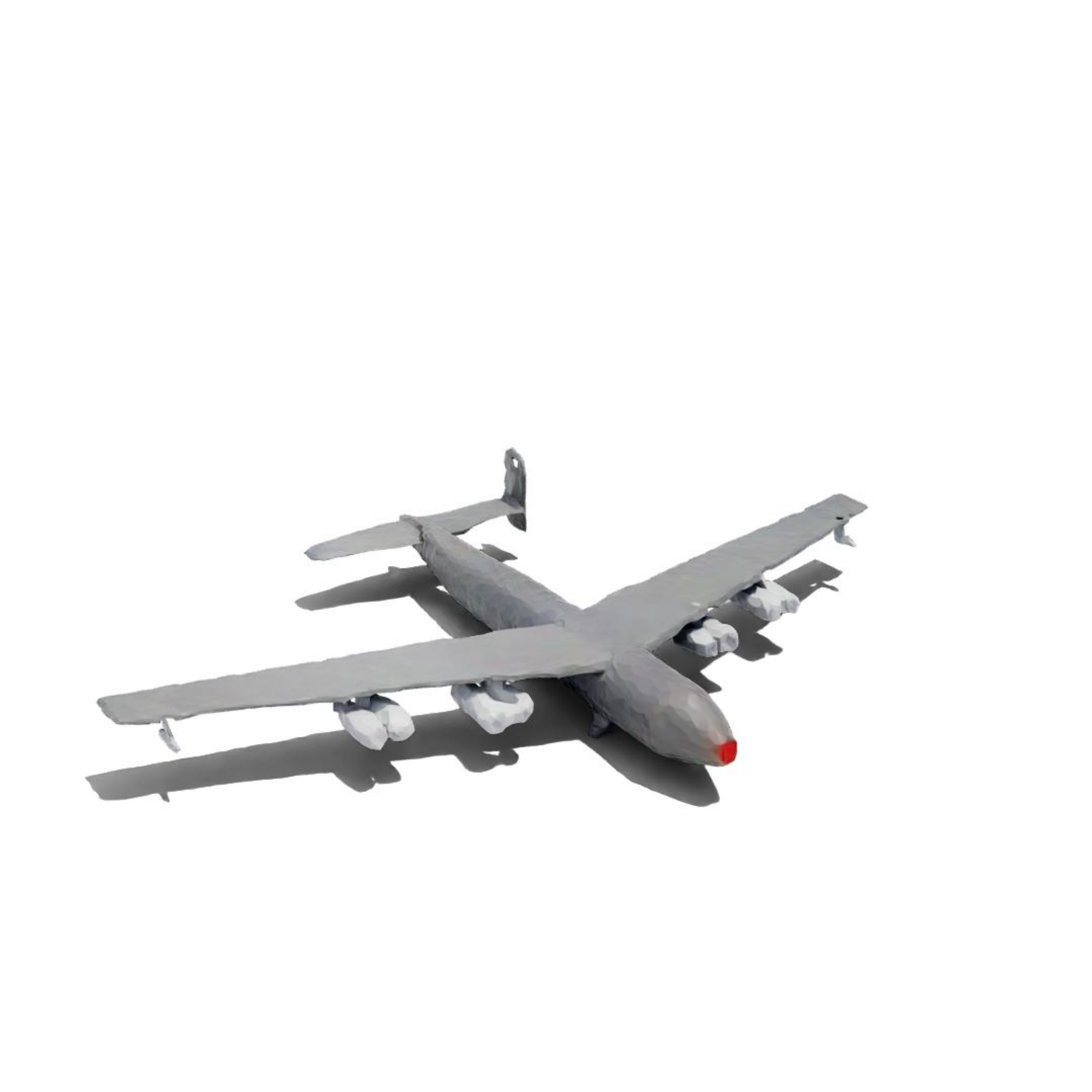}\includegraphics[width=0.08333333333333333\linewidth]{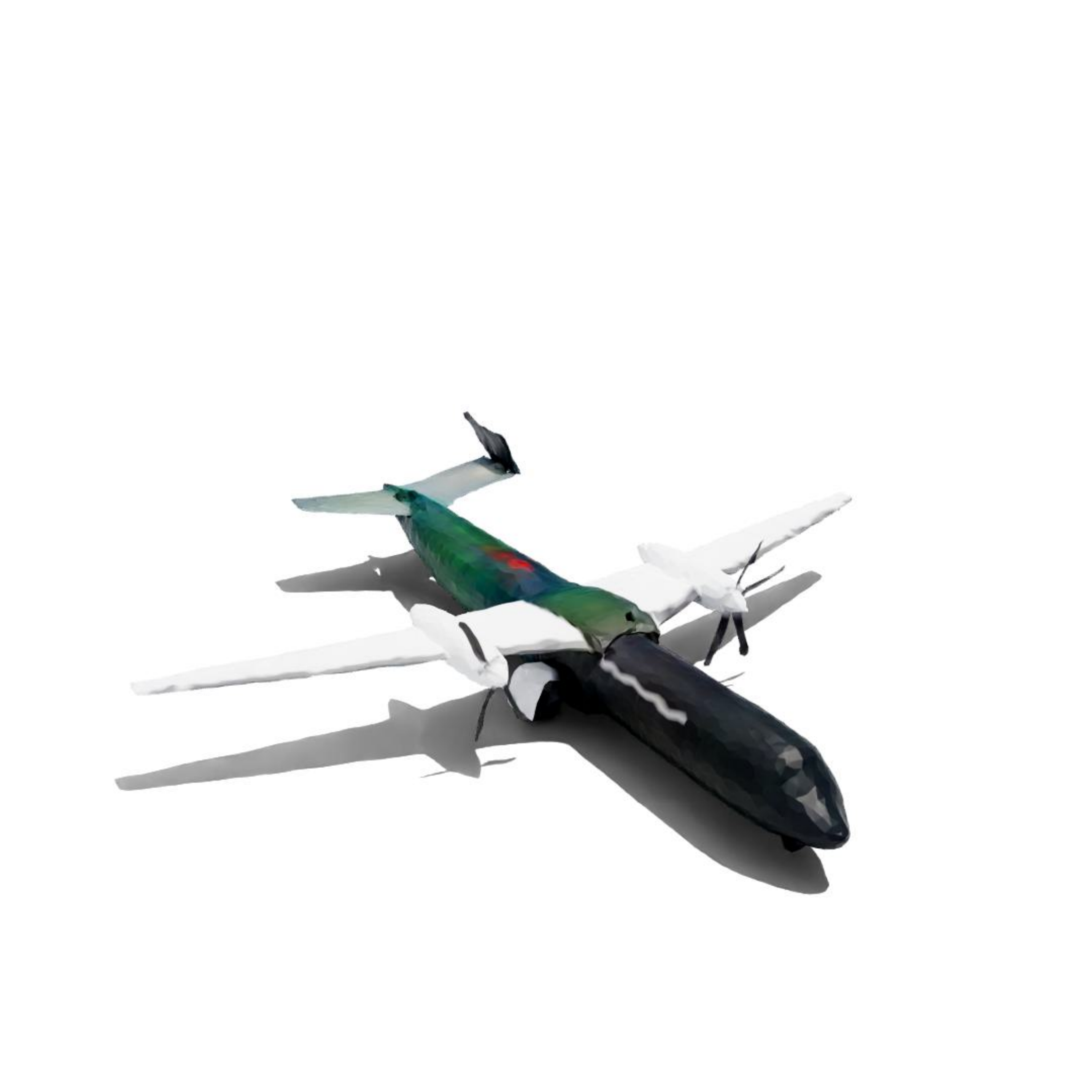}\includegraphics[width=0.08333333333333333\linewidth]{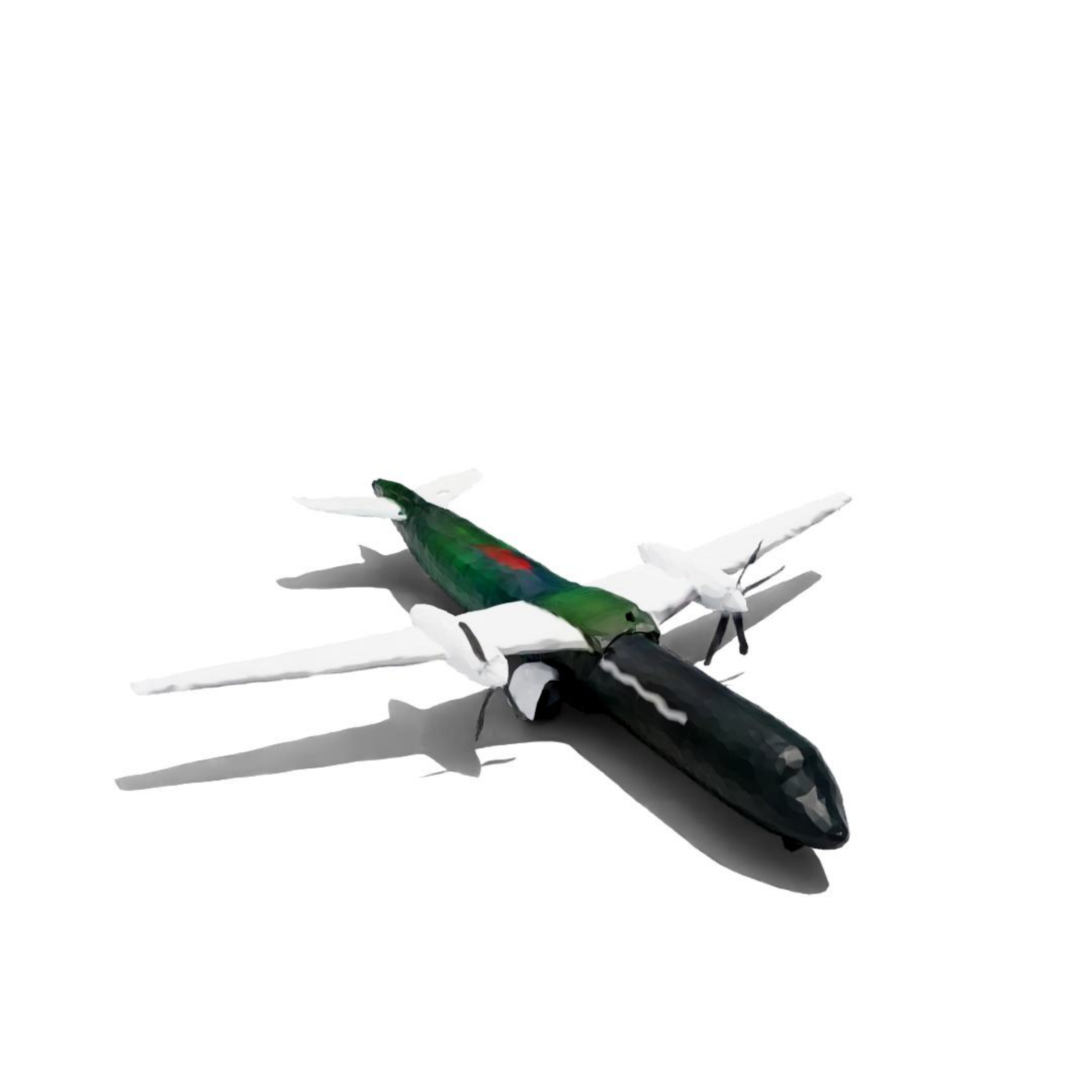}\includegraphics[width=0.08333333333333333\linewidth]{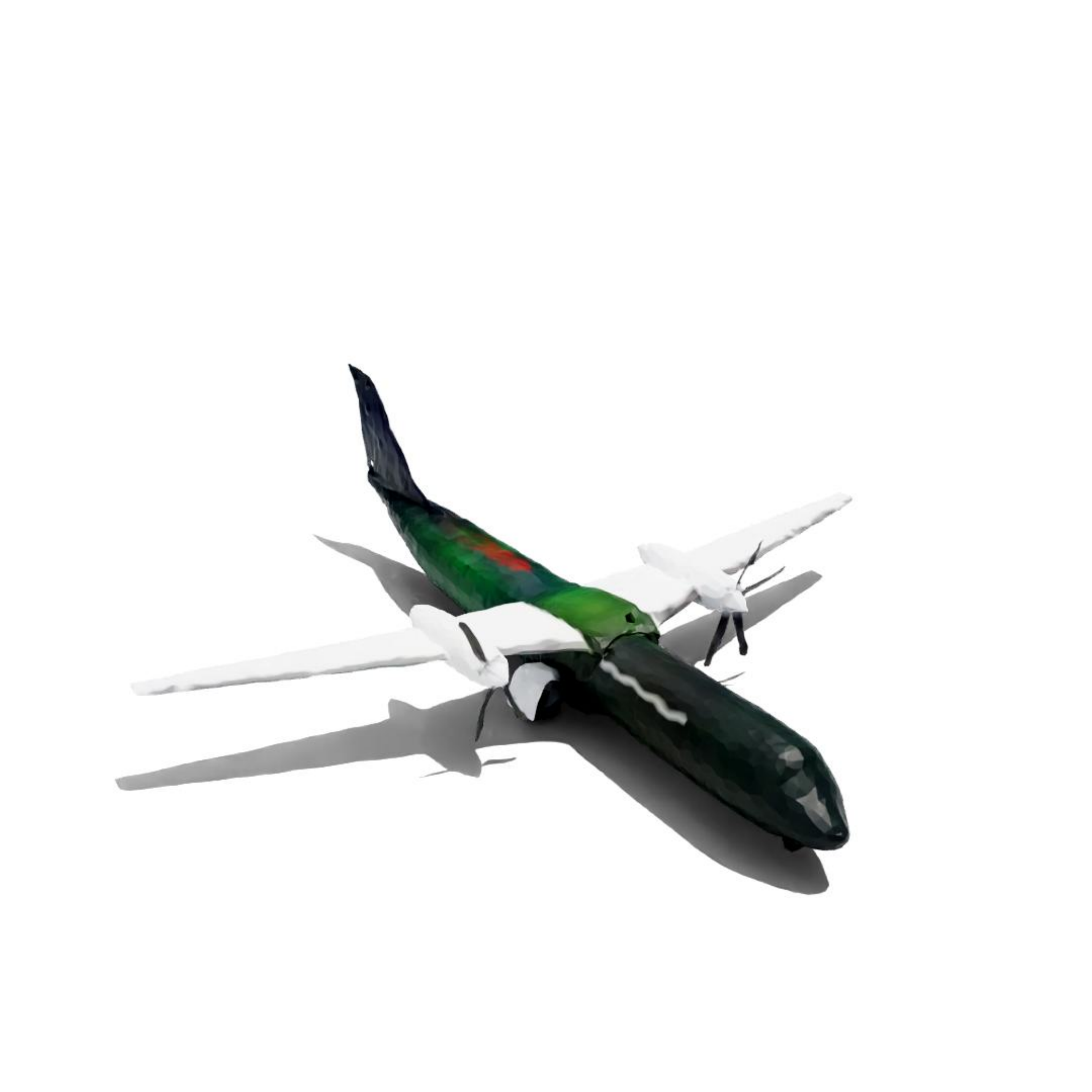}\includegraphics[width=0.08333333333333333\linewidth]{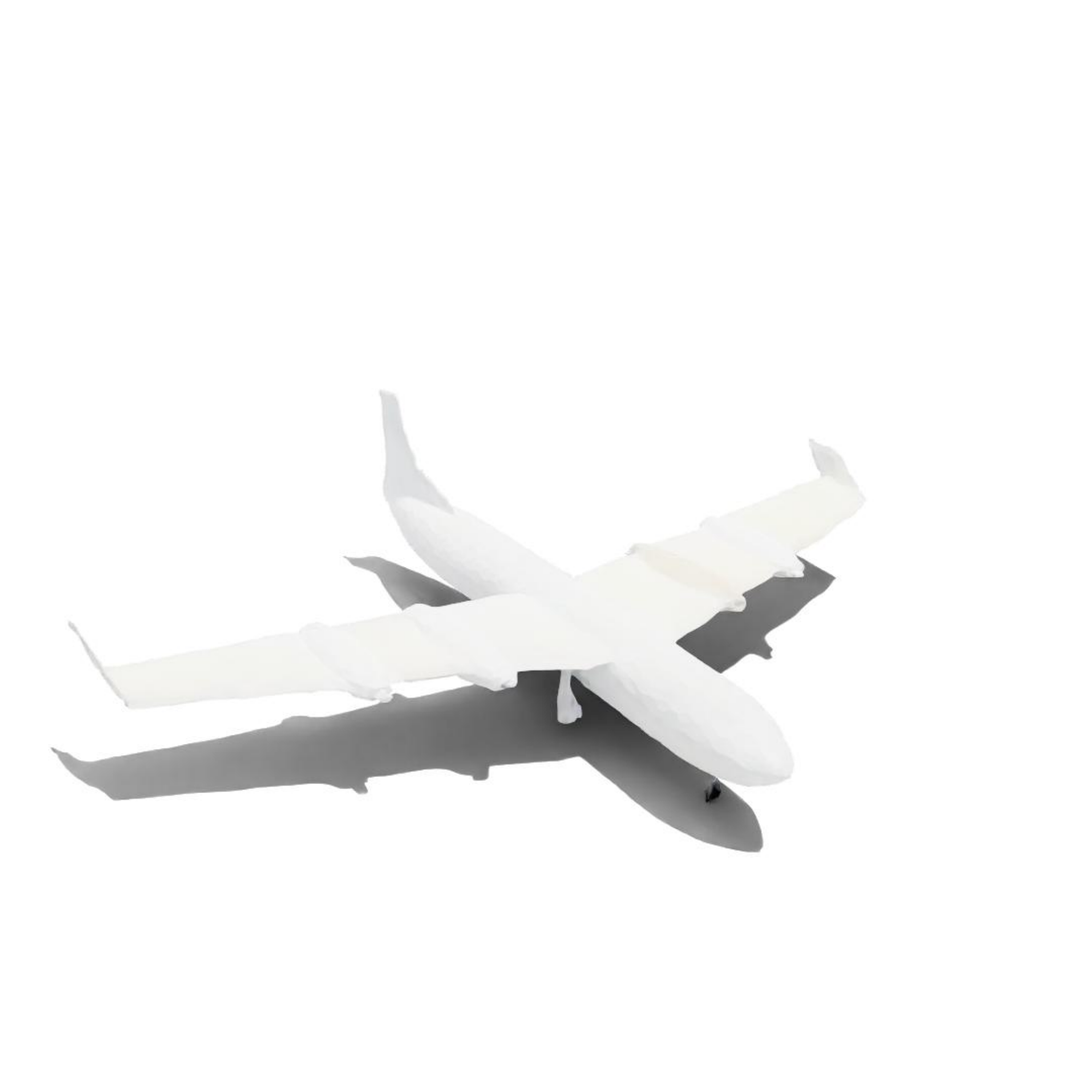}\includegraphics[width=0.08333333333333333\linewidth]{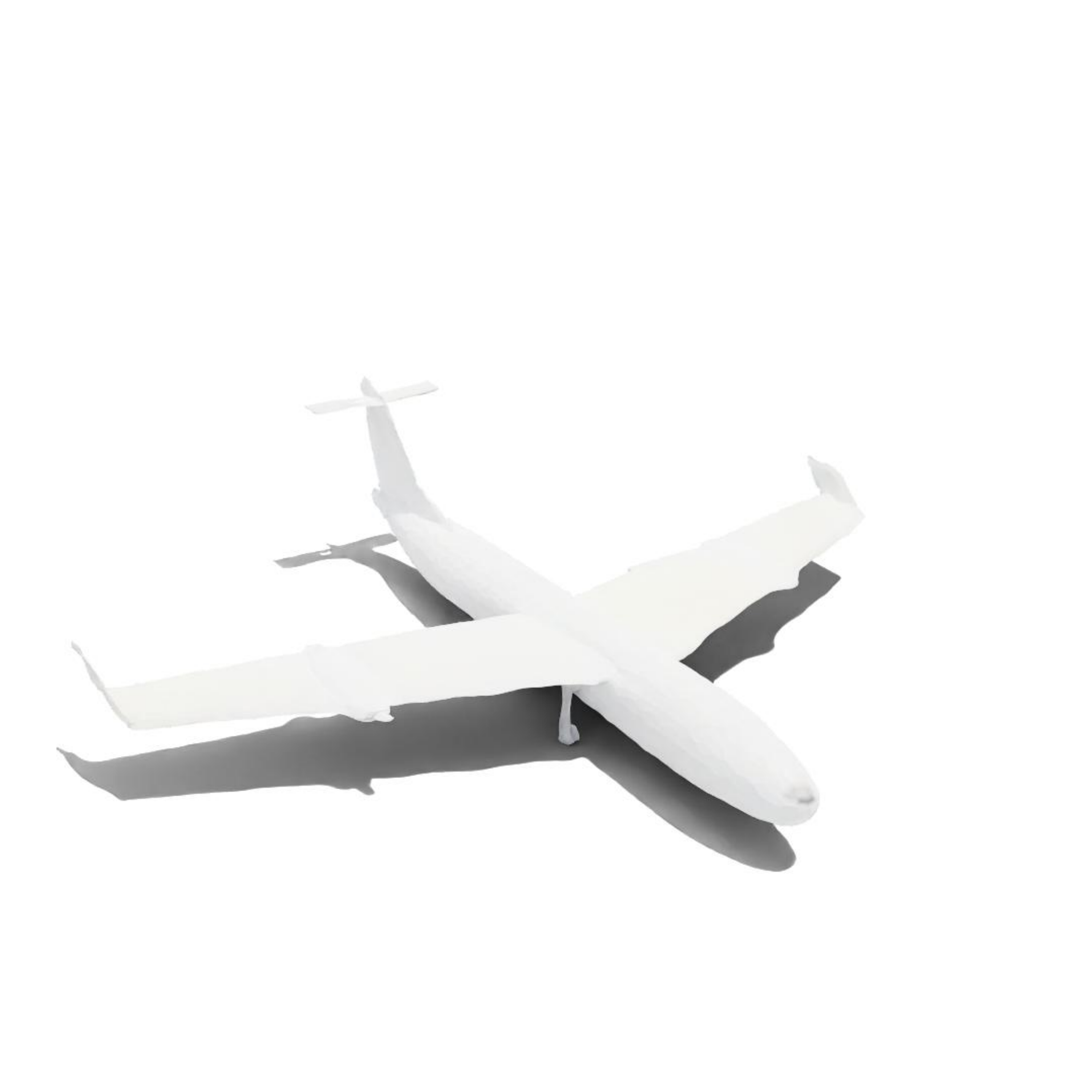}\includegraphics[width=0.08333333333333333\linewidth]{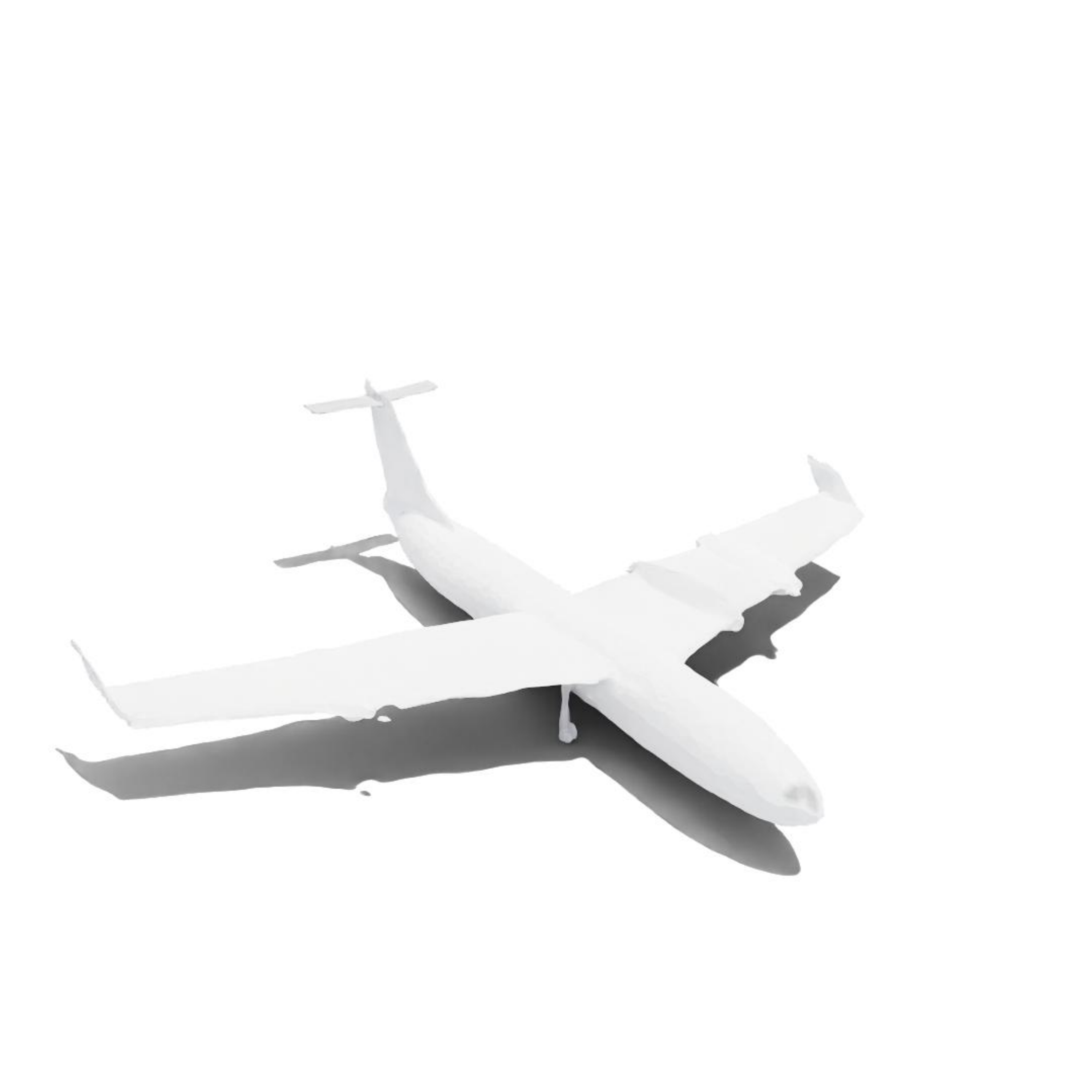}\includegraphics[width=0.08333333333333333\linewidth]{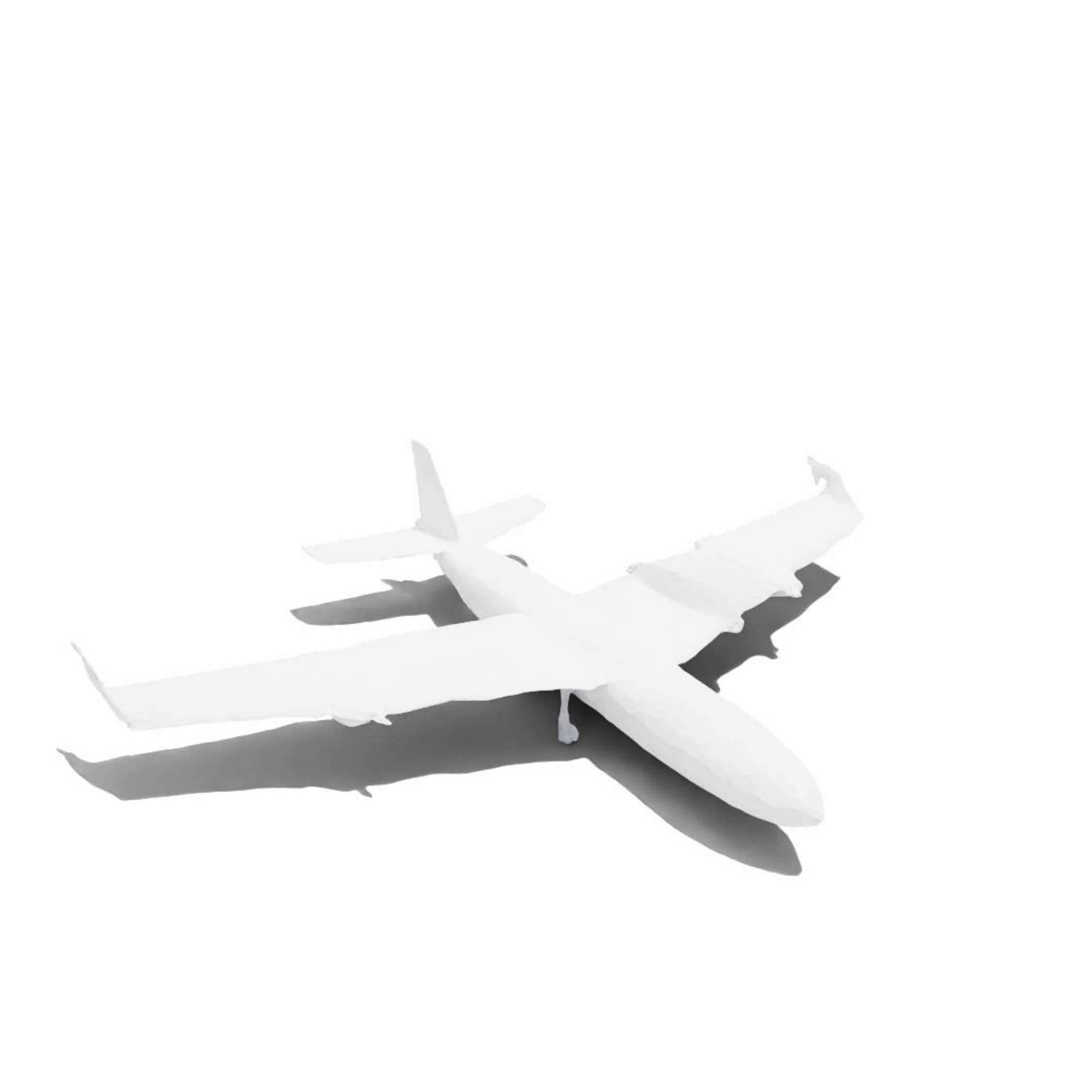}\includegraphics[width=0.08333333333333333\linewidth]{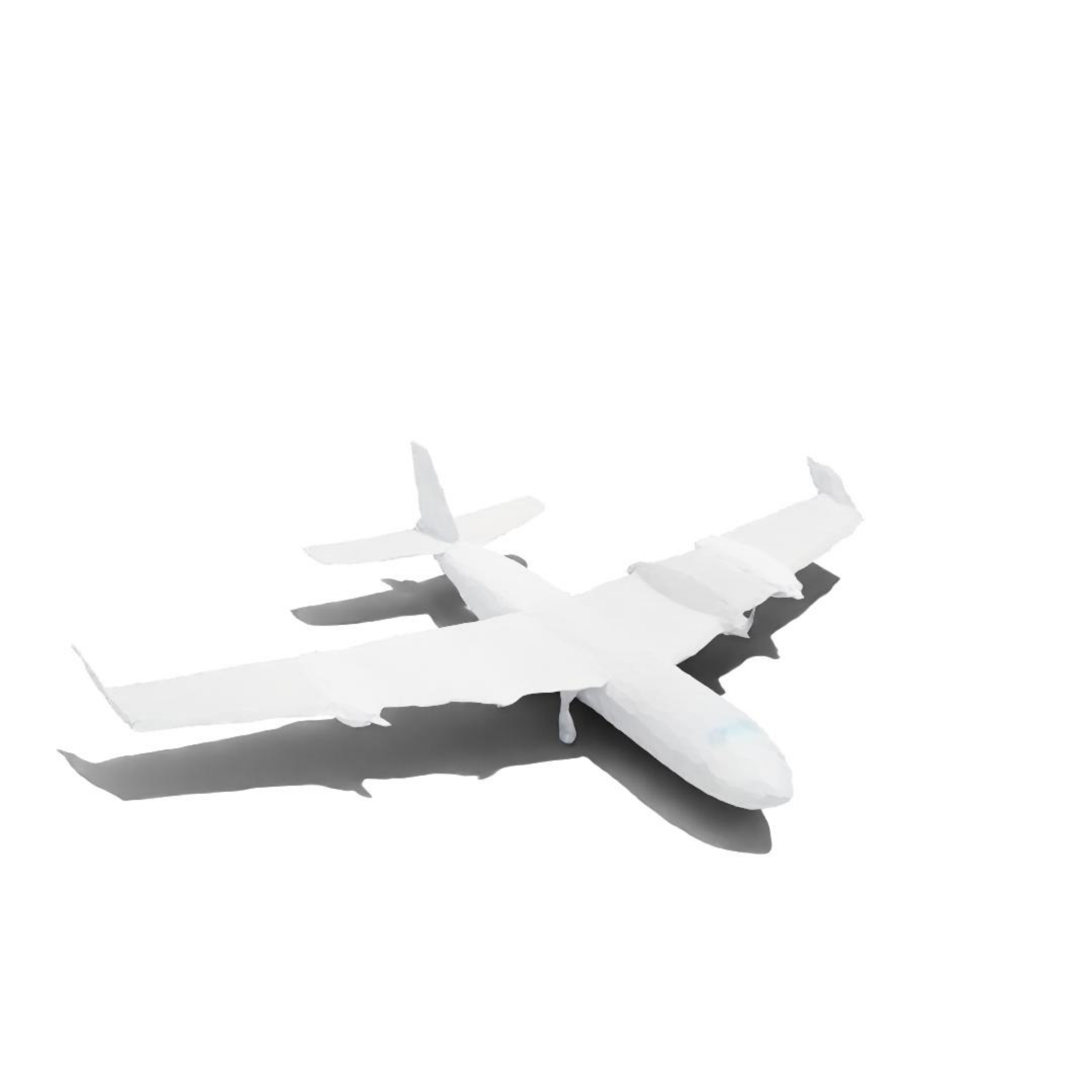}\includegraphics[width=0.08333333333333333\linewidth]{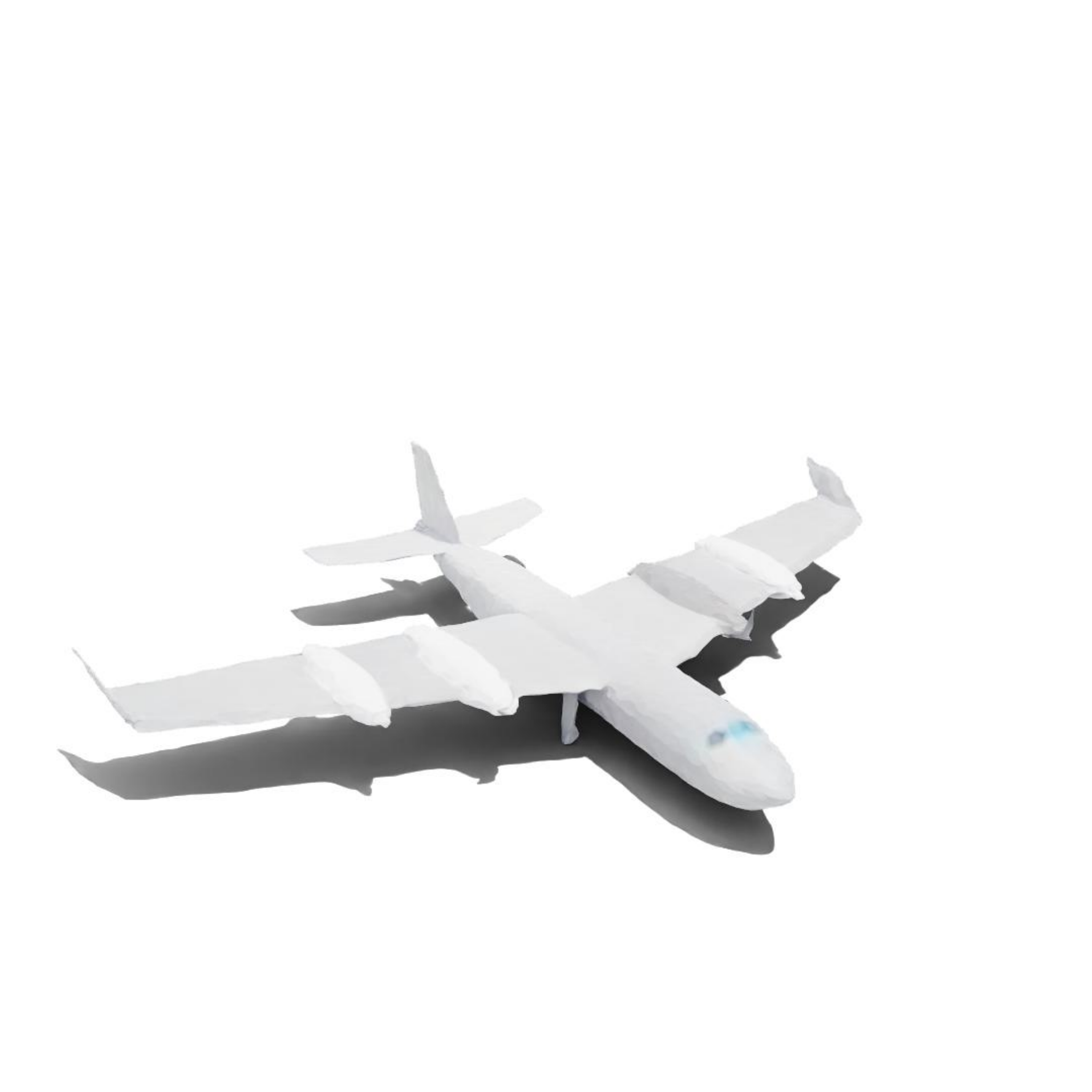}

\caption{\textbf{Shape interpolation between two anchor shapes.} We first generate the leftmost and rightmost samples and save the noise applied in every diffusion step. Intermediate shapes are then generated by spherical linear interpolation of the noise.}\label{fig:supp:interpolations}
\end{figure*}

Additionally, we found that morphing with the square root-based interpolation resulted in sudden jumps between shapes. Instead, we use spherical interpolation (Slerp) as defined in the following: 

\begin{equation}
    \hat{z}_{t, k} \coloneqq \text{Slerp}(z_{t,0},z_{t,1};k) = \frac{\sin[(1-k)\Omega]}{\sin\Omega}z_{t,0} + \frac{\sin{[k\Omega}]}{\sin{\Omega}}z_{t,1},
    \label{eq:slerp}
\end{equation}

where $\Omega$ is the subtended angle between the arc that is spanned by each $z_{t,0}$ and $z_{t,1}$ on the sphere and $k \in [0,1]$. 
In order to interpolate between two shapes we interpolate all noise vectors using equation \cref{eq:slerp} along the trajectory, resulting in smooth interpolations, as we are explicitly moving along the arc on the high dimensional sphere. See \Cref{fig:supp:interpolations} for various interpolations of airplanes, cars and chairs.

\clearpage

\subsection{Conditional generation}
While our primary goal was not to develop a strong single-view reconstruction (SVR) model, TetraDiffusion can be extended to conditional generation. We render multiple views for each ShapeNet shape and embed the images into CLIP space~\cite{radford2021learning}. To incorporate these we expand our U-Net with additional cross-attention layers~\cite{chen2021crossvit} and add the result to the output of the corresponding self-attention layer. 

\begin{figure*}[!ht]
    \setkeys{Gin}{width=\linewidth}
    \begin{tabularx}{1.0\textwidth}{>{\hsize=0.5cm}XXXXXXXXX}
         \rotatebox{90}{Condition} & 
              \includegraphics[trim={0cm 1cm 0cm 2cm}, clip, width=1.2cm]{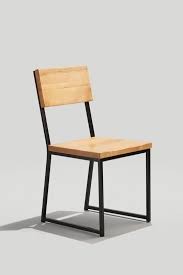} &
         \includegraphics{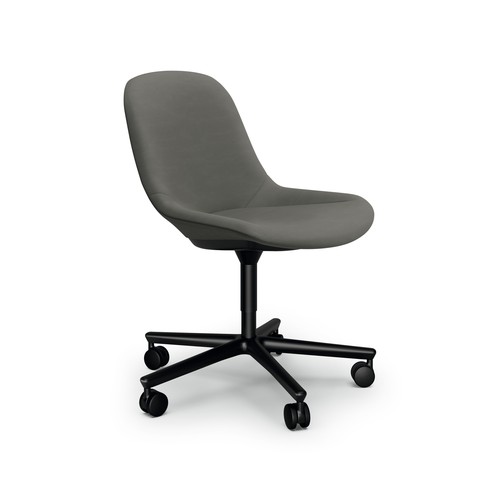} &
         \includegraphics{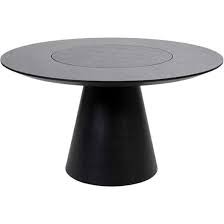} &
         \includegraphics[trim={2cm 1cm 2cm 2cm}, clip]{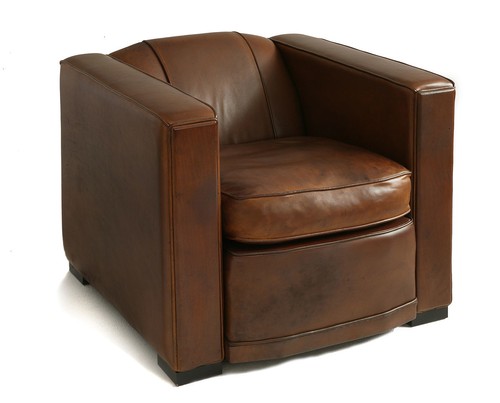} & 
         \includegraphics[trim={0cm 1cm 0cm 0cm}, clip]{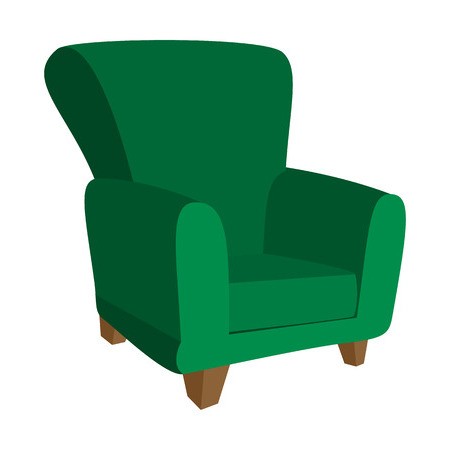} &
         \includegraphics[trim={1cm 0cm 1cm 0cm}, clip]{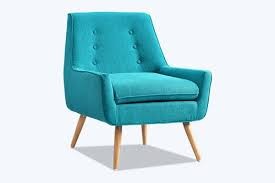} & 
              \includegraphics[trim={-1cm 0cm -1cm 0cm}, clip, width=1.3cm]{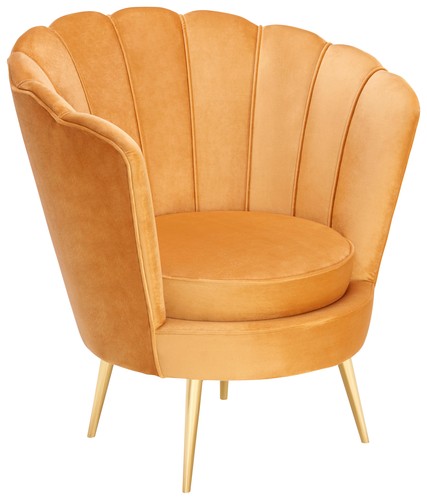} & 
             \includegraphics[trim={0cm 0.3cm 0cm 0.3cm}, clip, width=1.3cm]{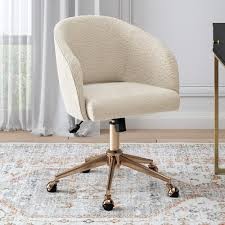}\\
         \rotatebox{90}{Generated} & 
         \includegraphics{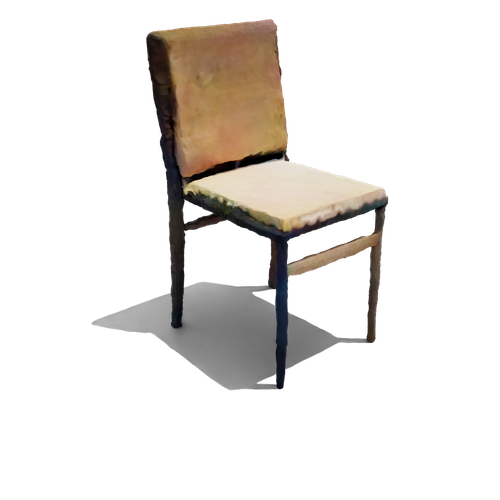} &
         \includegraphics[trim={0cm 0cm 0cm 0cm}, clip]{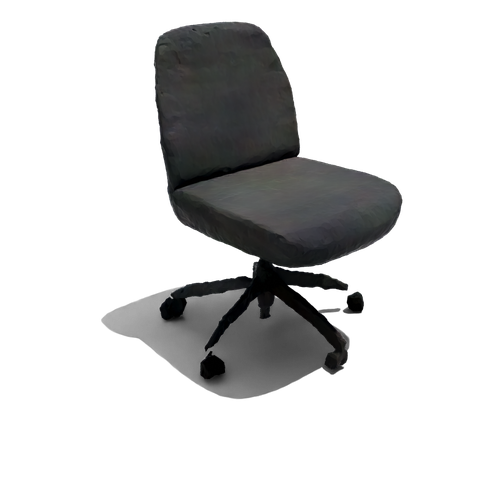} & \includegraphics{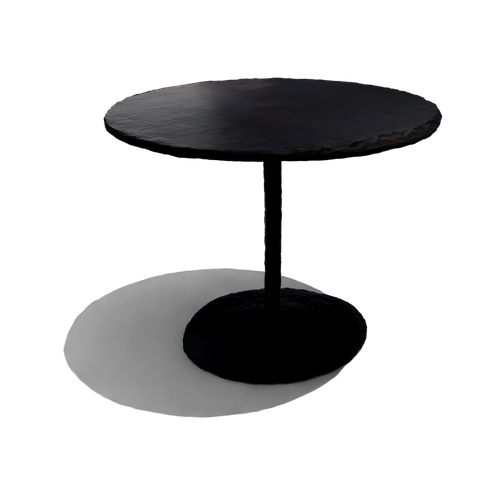} & 
         \includegraphics[trim={0cm 0cm 0cm 0cm}, clip]{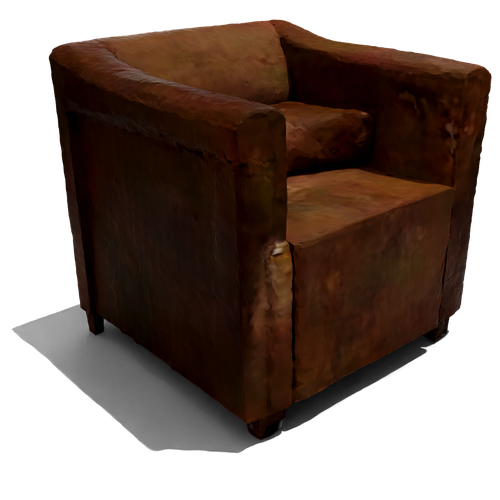} & \includegraphics[trim={0cm 0cm 0cm 0cm}, clip]{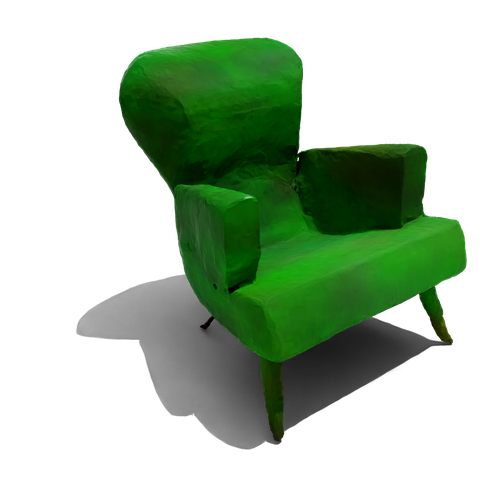} &  
         \includegraphics[trim={0cm 0cm 0cm 0cm}, clip]{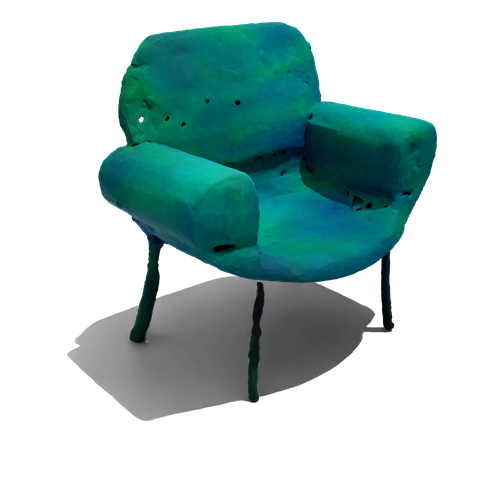} & 
         \includegraphics[trim={0cm 0cm 0cm 0cm}, clip]{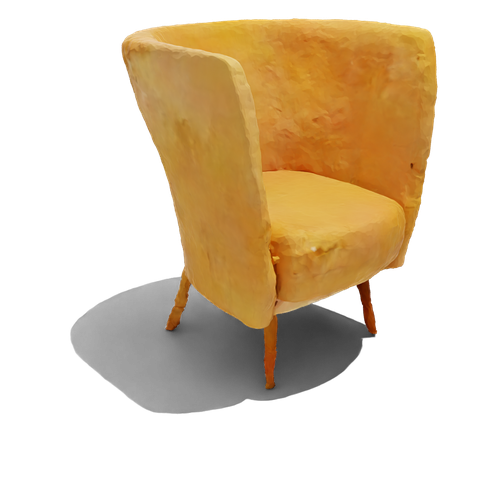} &
             \includegraphics[trim={0cm 0cm 0cm 0cm}, clip]{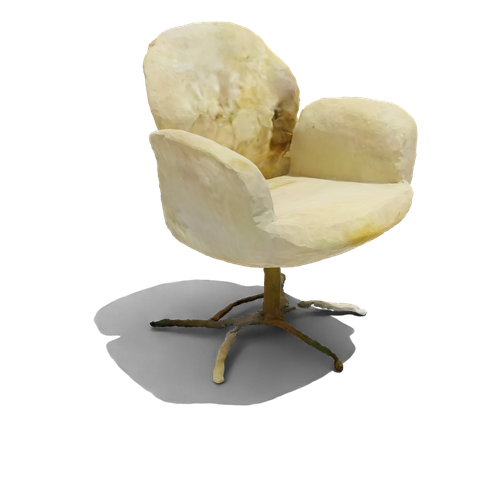}  \\    
        \midrule
         \rotatebox{90}{Condition} & 
         \includegraphics{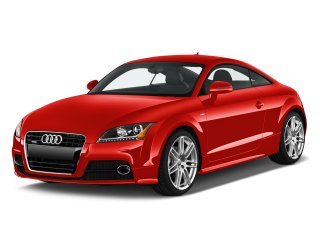}&
         \includegraphics{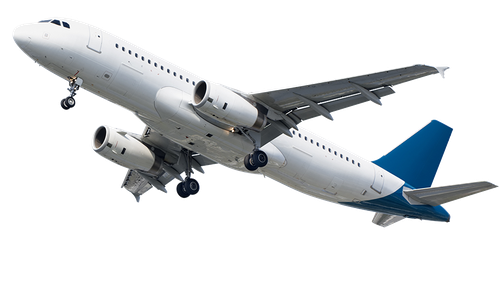} &
         \includegraphics[trim={0cm 0cm 0cm 0cm}, clip]{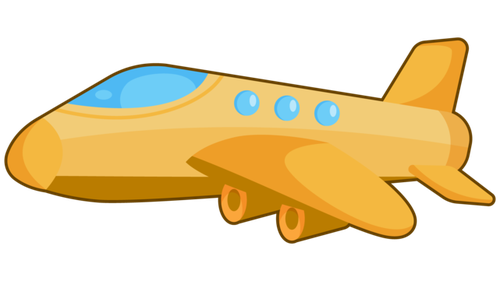} &
         \includegraphics[trim={0cm 3cm 0cm 0cm}, clip]{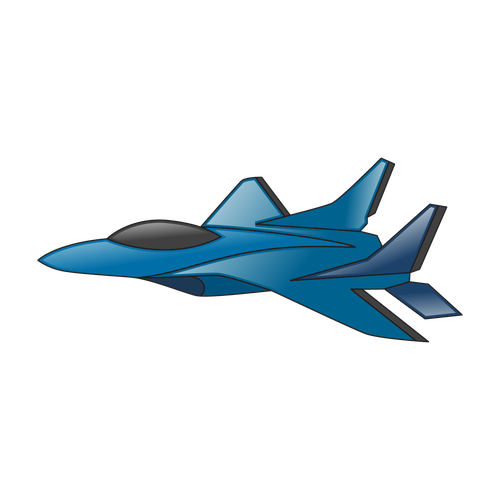}  &
         \includegraphics[trim={6cm 5cm 6cm 5cm}, clip]{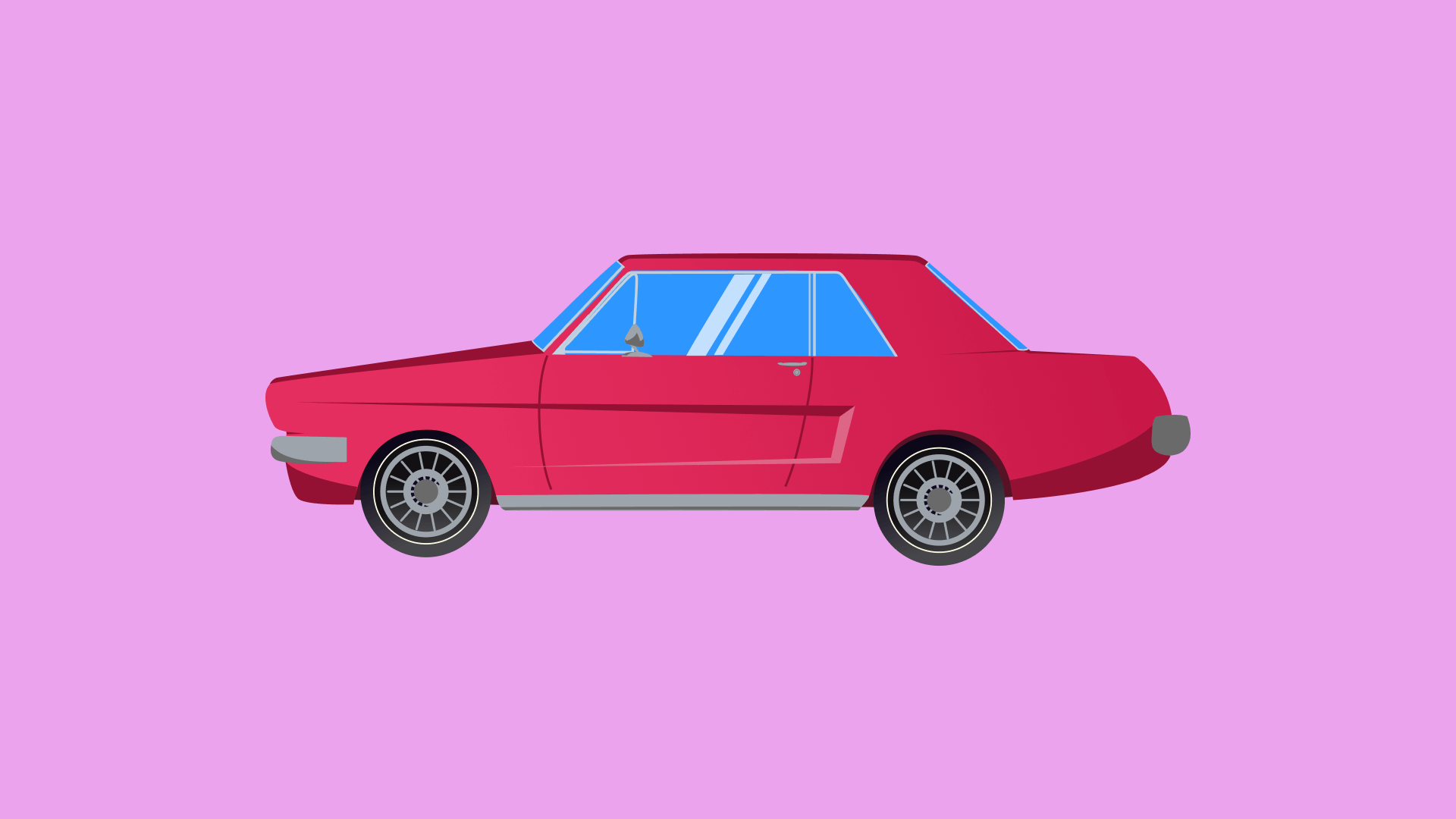} &
         \includegraphics[trim={0cm 0cm 0cm 0cm}, clip]{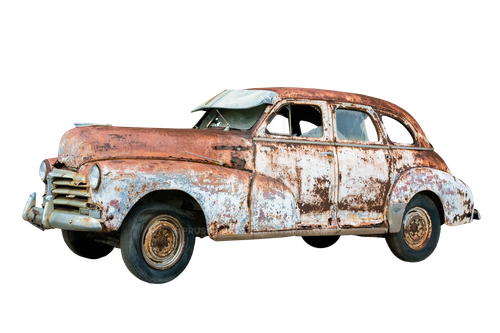} &
         \includegraphics[trim={0cm 2cm 0cm 0cm}, clip]{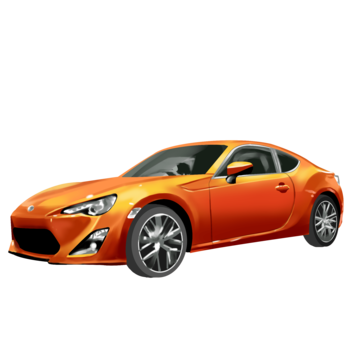} & 
         \includegraphics[trim={0cm 0cm 0cm 0cm}, clip]{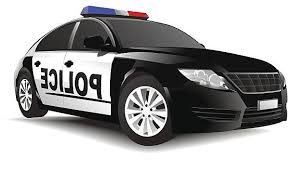}\\
         
         \rotatebox{90}{Generated} &
         \includegraphics[trim={4cm 0cm 0cm 6cm}, clip]{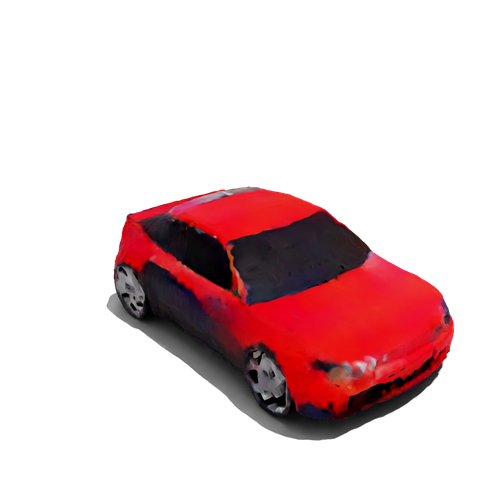}&
         \includegraphics[trim={4cm 0cm 0cm 6cm}, clip]{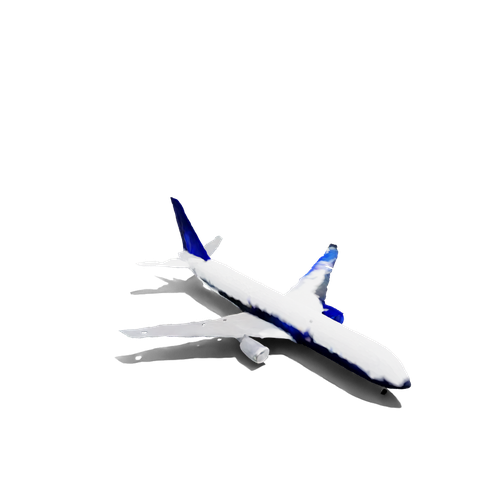} & 
         \includegraphics[trim={4cm 0cm 0cm 6cm}, clip]{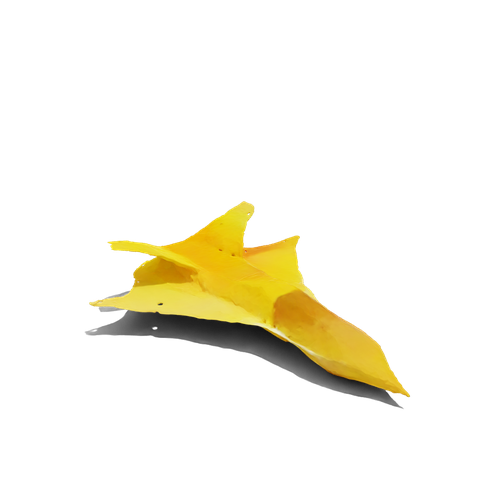} &
         \includegraphics[trim={4cm 0cm 0cm 6cm}, clip]{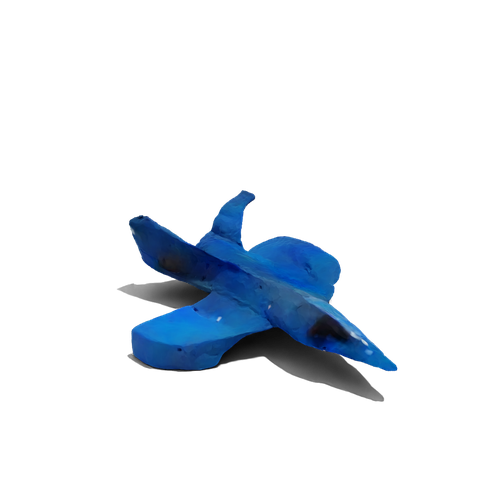} & 
         \includegraphics[trim={4cm 0cm 0cm 6cm}, clip, ]{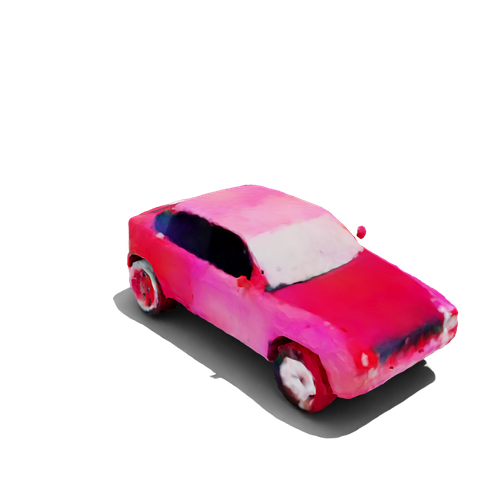} & 
         \includegraphics[trim={4cm 0cm 0cm 6cm}, clip,]{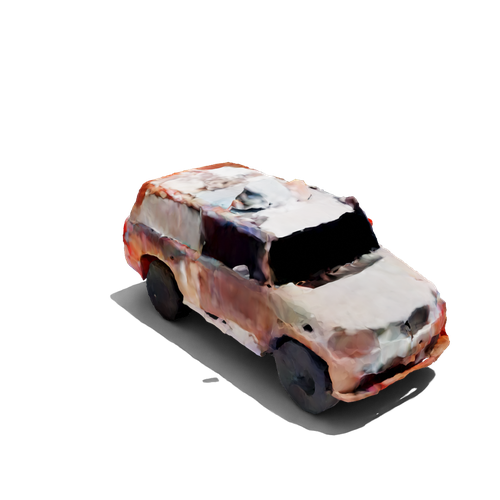} &
         \includegraphics[trim={4cm 0cm 0cm 6cm}, clip]{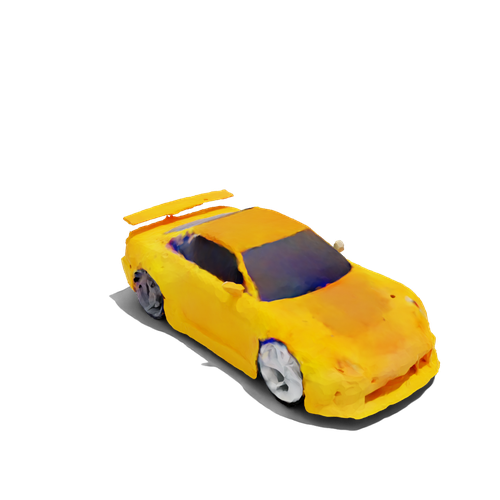} & 
         \includegraphics[trim={4cm 0cm 0cm 6cm}, clip]{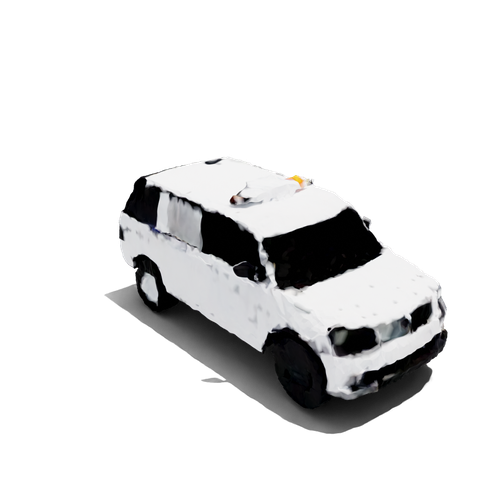}

    \end{tabularx}
    \caption{\textbf{Conditional generation of 3D shapes.} \emph{TetraDiffusion} is conditioned on CLIP embeddings of in-the-wild images during inference, while only trained on embeddings from rendered ShapeNet views. The shapes are displayed in matched pairs, with the upper image showing the condition and the lower one showing the corresponding generated mesh.}
    \label{fig:conditional}
\end{figure*}

At test time, we can then reconstruct faithful meshes from novel, unseen images by simply conditioning on the corresponding embedding. We qualitatively show the results from RGB data in \cref{fig:conditional}. Notably, a single conditional version of TetraDiffusion, trained across all classes together, is able to adapt to diverse classes, shapes and colors according to the conditioning signal. Note that we do not incorporate additional test time guidance.

\clearpage

\subsection{Test time guidance}

Similar to classifier and classifier-free guidance \cite{rombach2022high, ho2022classifier}, we incorporate test time guidance, \ie altering the mean prediction of the reverse diffusion process with the gradient of a guiding task. In image diffusion, the guiding task can be a classifier $p(y|\mathbf{z_t})$ trained to predict the correct class label $y$ given the latent variable $\mathbf{z_t}$. Instead of training an extra model or training a classifier-free conditional model, we explore the robustness of non-parametric guidance, where no additional training is required besides the unconditional model. In particular, we experiment with color and volumetric gradients, making use of our fully differentiable mesh and rendering pipeline.

In order to guide the diffusion towards a specific color, we project the color name to its CLIP embedding and compare the embedding to image embeddings obtained by rendering multiple views of $\mathbf{\hat{x}}_\theta(\mathbf{z_t}; t)$ in every step. The loss is computed as a spherical distance following \cite{crowson2021clip}:
\begin{equation}
    \mathcal{L}(e_{\text{color}}, e_{\text{render}}) = 2\cdot\bigg[\arcsin\bigg( \frac{\lVert e_{\text{color}} - e_{\text{render}} \rVert}{2}\bigg) \bigg]^2,
\end{equation}
where $e_{\text{color}} = \text{CLIP}(\text{``color"})$ is the CLIP embedding of a color in text form and $e_{\text{render}} = \Phi(\theta(\mathbf{z_t}; t))$ is a rendered view of the mesh extracted from $\mathbf{\hat{x}}_\theta(\mathbf{z_t}; t)$ with our extended marching tetrahedra algorithm.

Additionally, we may influence the volume of the diffused mesh by simply pushing towards a smaller or larger bounding box around the extracted mesh.

\begin{figure}[!hb]
    \centering
    \begin{subfigure}[b]{0.08\linewidth}
        \begin{tikzpicture}
            \node at (0.0,0.0) {};
            \draw[latex-latex] (0,0.6) -- (0,4.);
            \filldraw[fill=white] (-0.4,3.8) circle (5pt);
            \filldraw[fill=white] (-0.3,0.8) circle (2pt);
            \draw[black, thick] (-0.1,2.2) -- (0.1,2.2);
            \node at (-0.5,2.2) {Vol.};
        \end{tikzpicture}%
    \end{subfigure}
    \begin{subfigure}[b]{0.45\linewidth}
        \begin{subfigure}[b]{0.29\linewidth}
            \includegraphics[width=\linewidth]{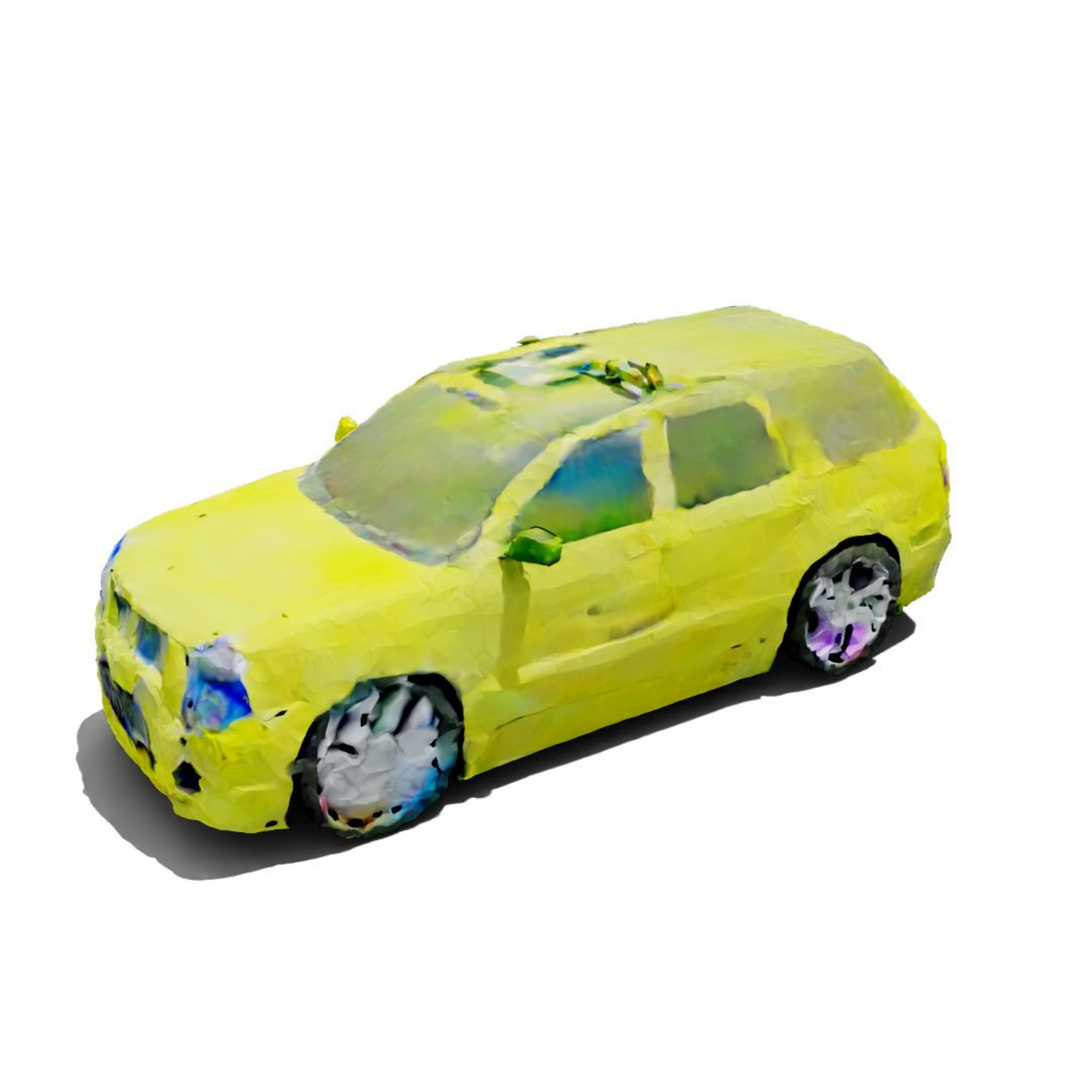}
        \end{subfigure}
        \begin{subfigure}[b]{0.29\linewidth}
            \includegraphics[width=\linewidth]{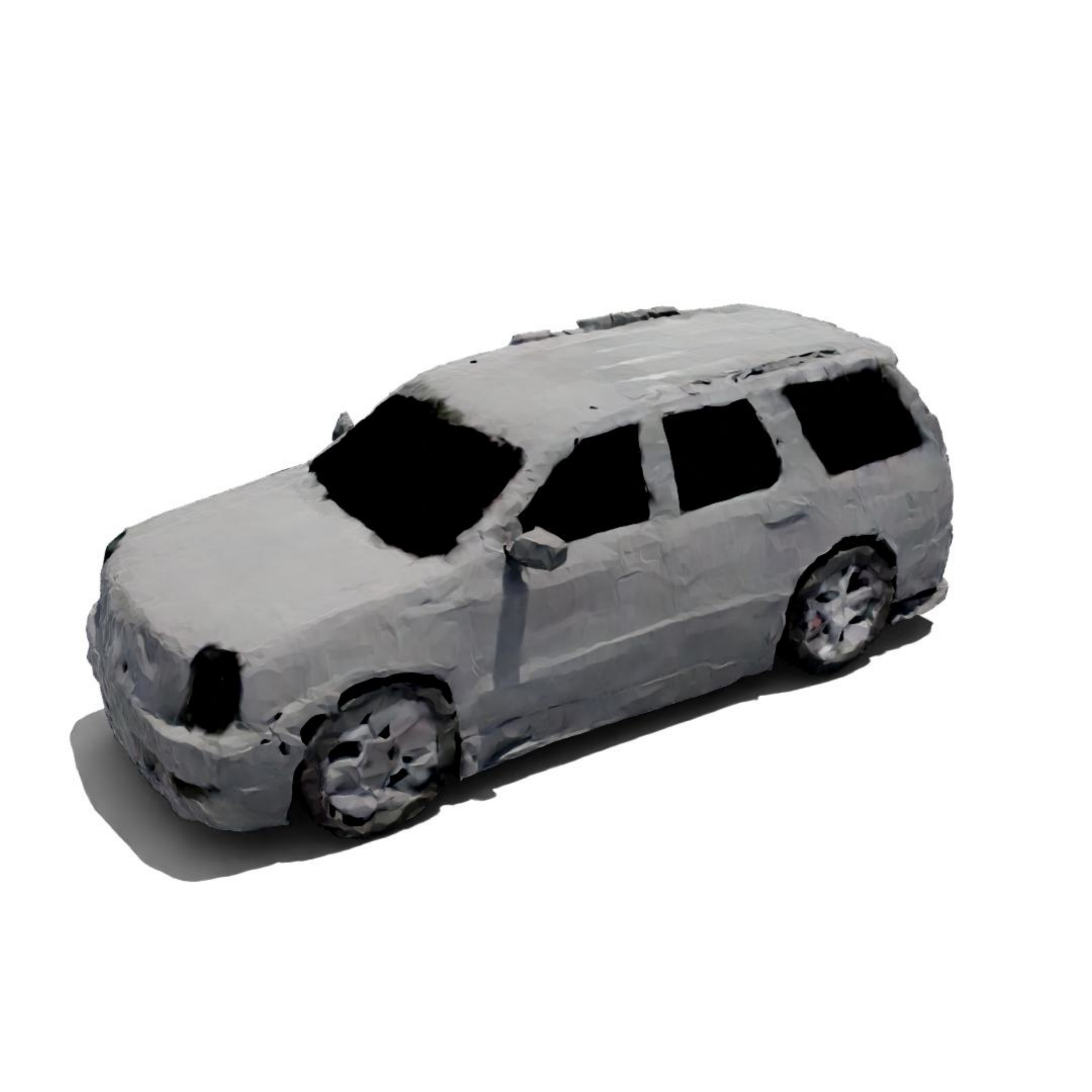}
        \end{subfigure}
        \begin{subfigure}[b]{0.29\linewidth}
            \includegraphics[width=\linewidth]{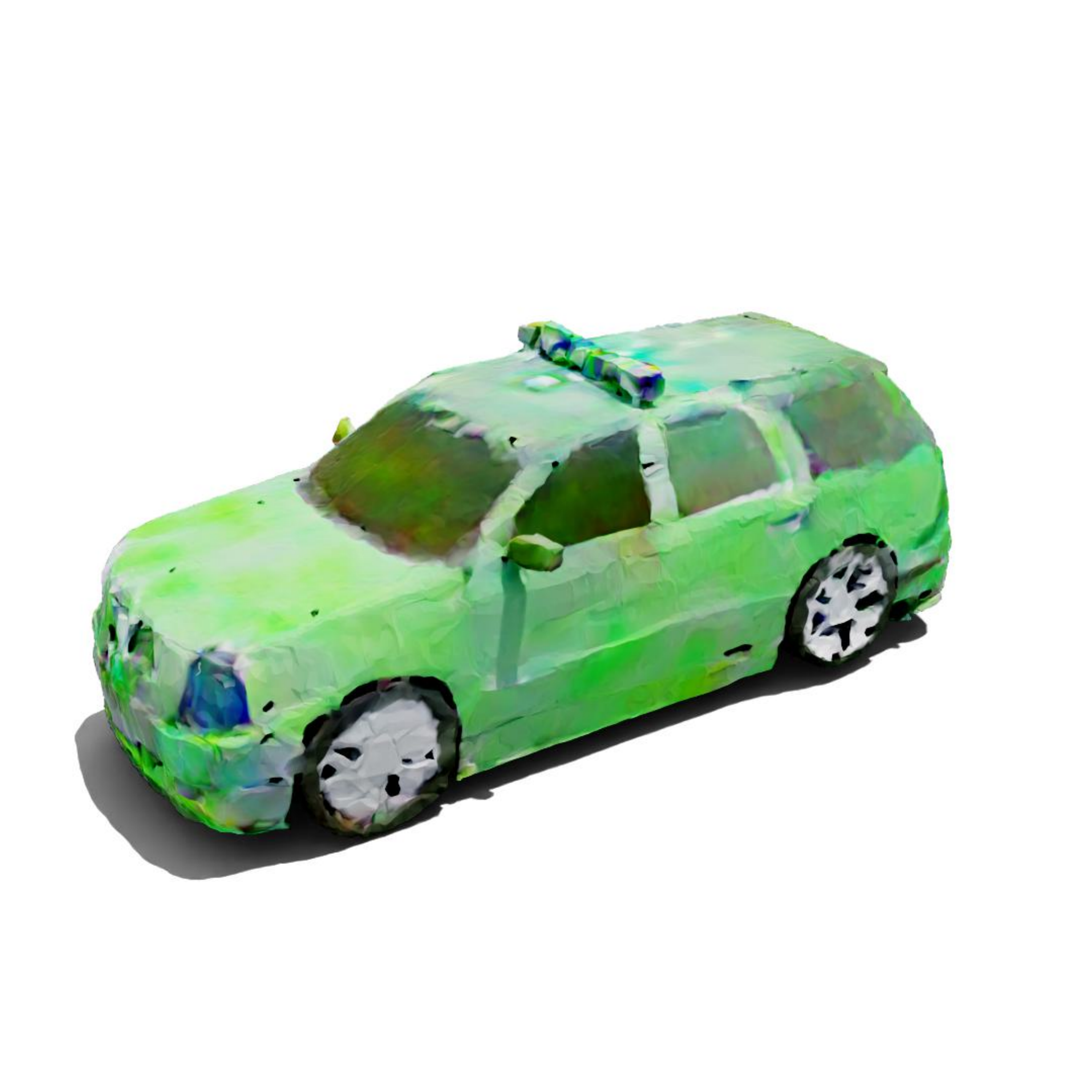}
        \end{subfigure}
        \begin{subfigure}[b]{0.29\linewidth}
            \includegraphics[width=\linewidth]{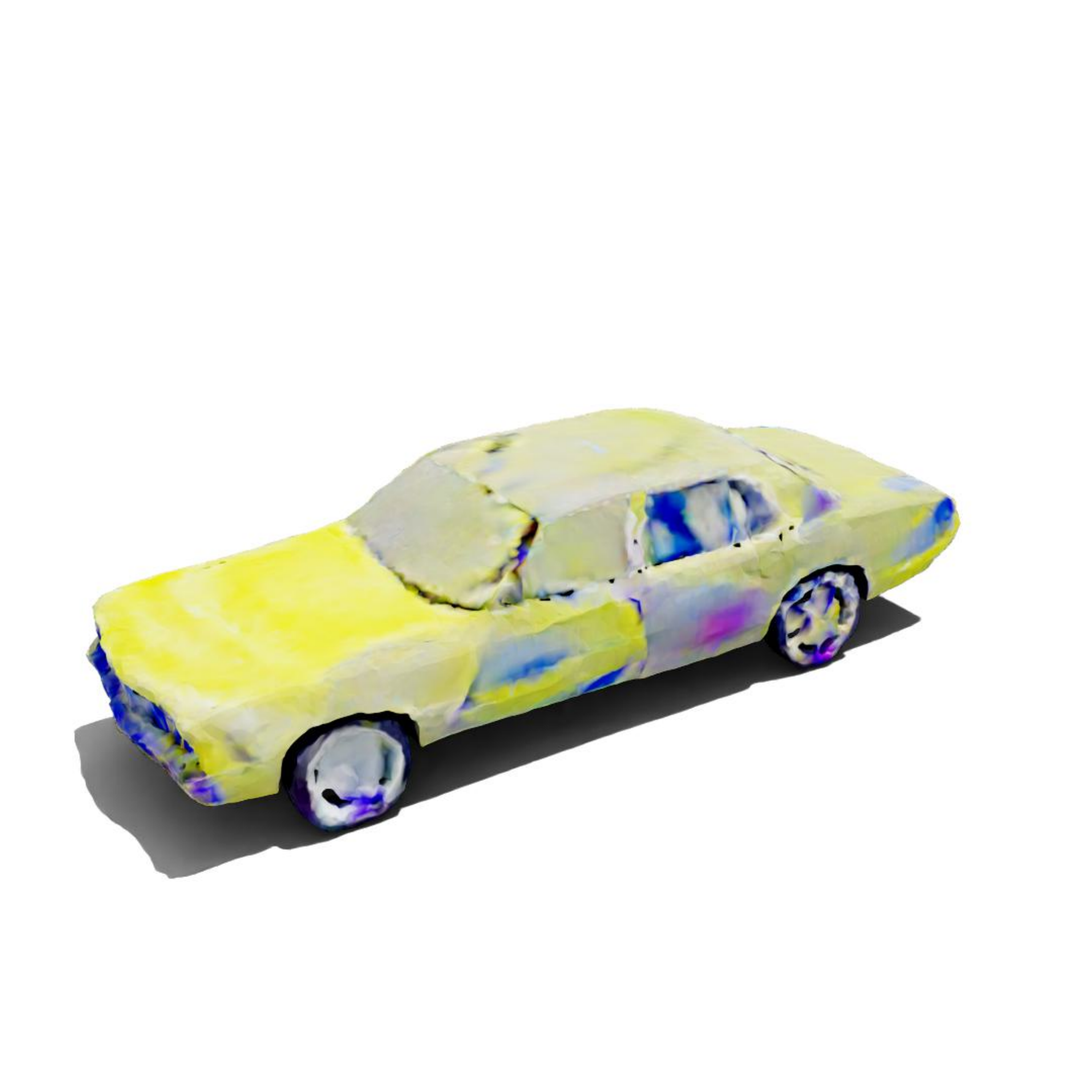}
        \end{subfigure}
        \begin{subfigure}[b]{0.29\linewidth}
            \includegraphics[width=\linewidth]{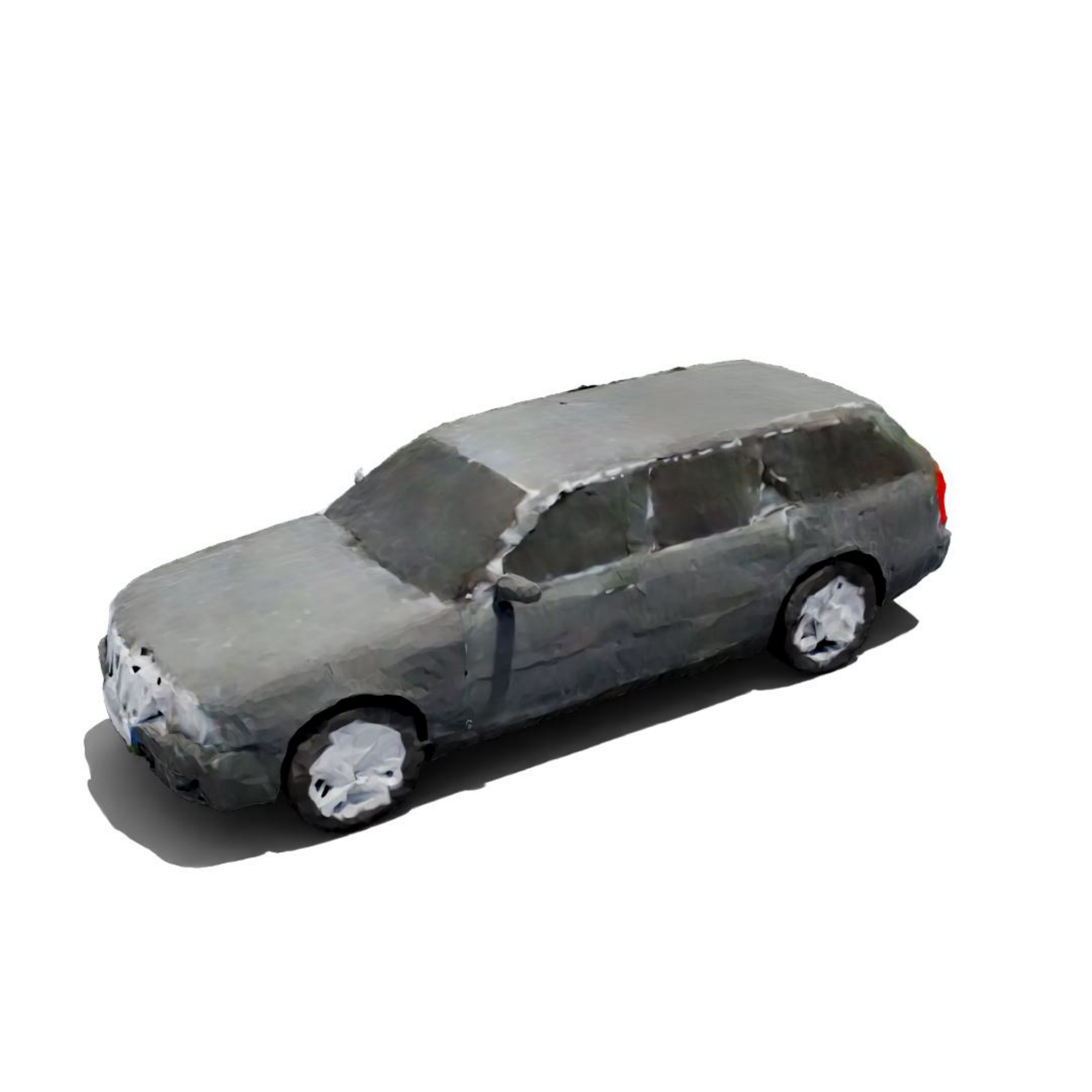}
        \end{subfigure}
        \begin{subfigure}[b]{0.29\linewidth}
            \includegraphics[width=\linewidth]{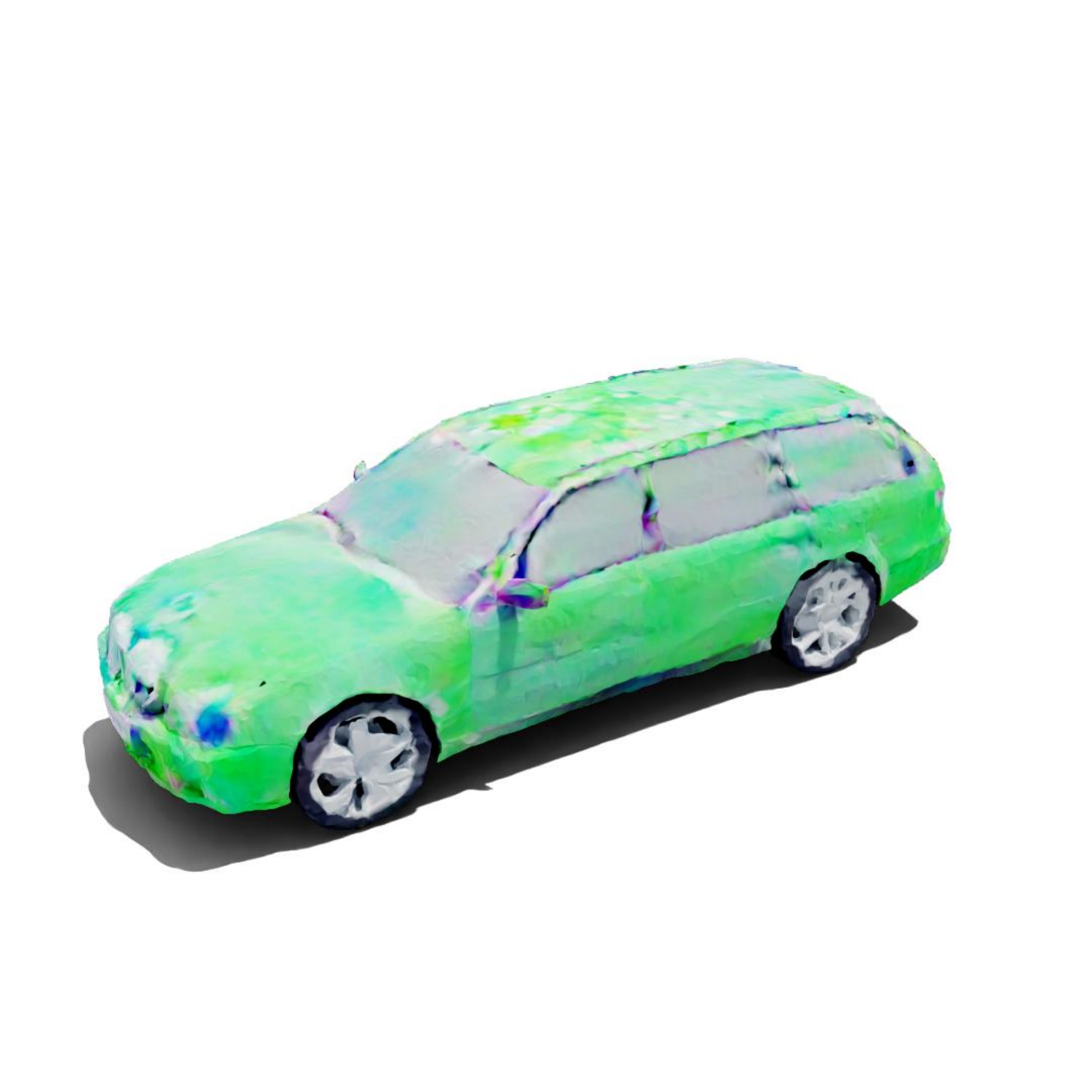}
        \end{subfigure}
        \begin{subfigure}[b]{0.29\linewidth}
            \includegraphics[width=\linewidth]{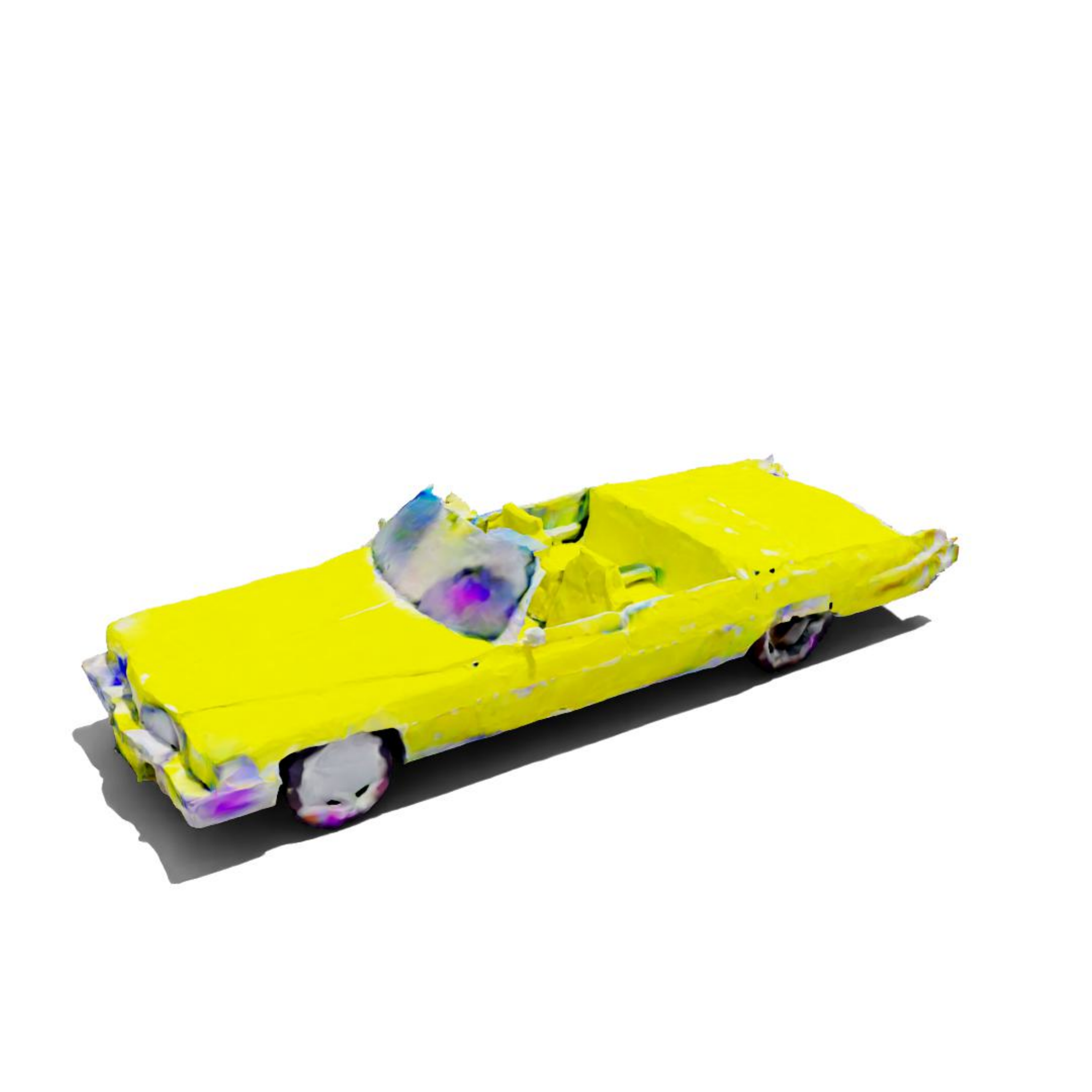}
        \end{subfigure}
        \hfill
        \begin{subfigure}[b]{0.29\linewidth}
            \includegraphics[width=\linewidth]{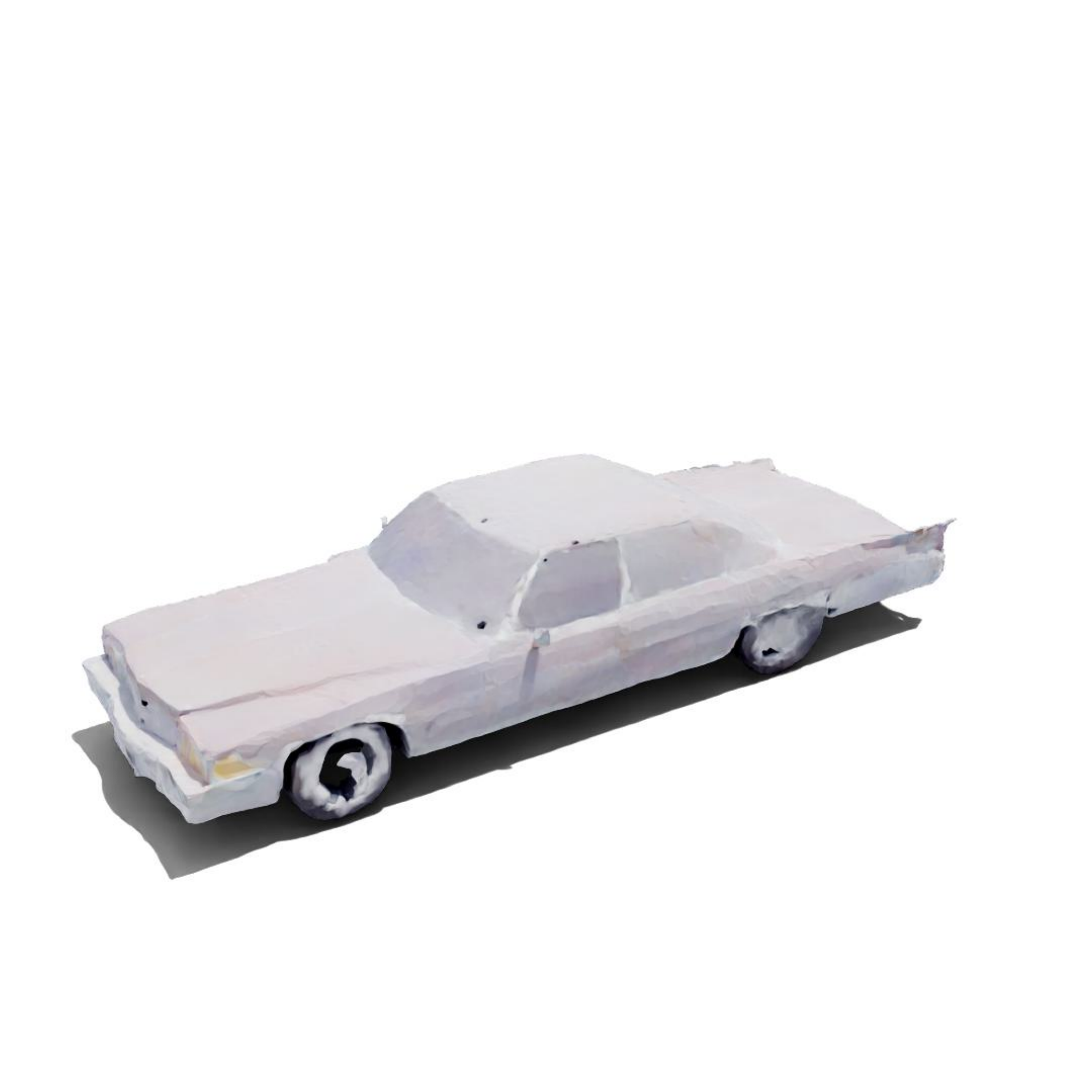}
        \end{subfigure}
        \hfill
        \begin{subfigure}[b]{0.29\linewidth}
            \includegraphics[width=\linewidth]{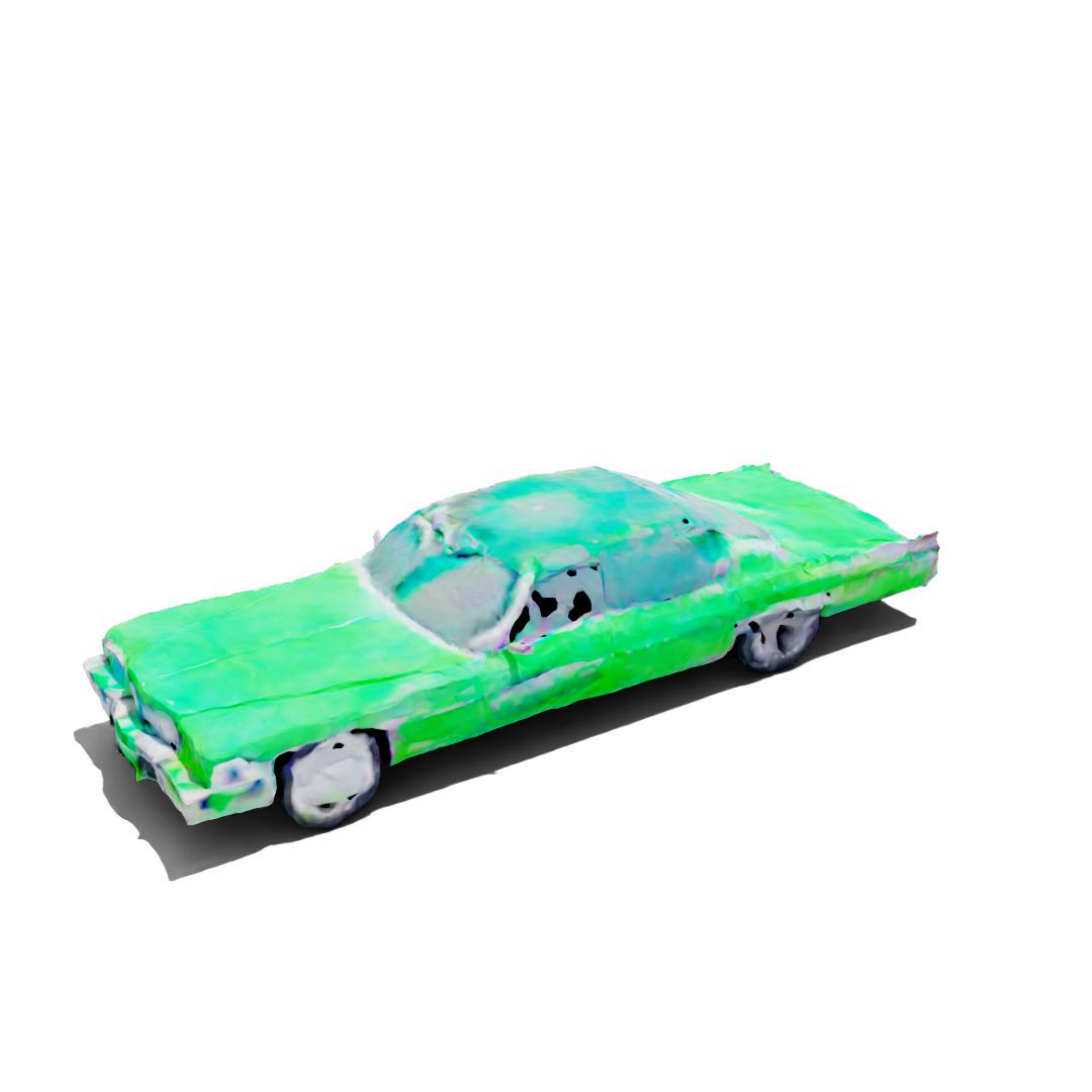}
        \end{subfigure}
    \end{subfigure}
    \hfill
    \begin{subfigure}[b]{0.45\linewidth}
        \begin{subfigure}[b]{0.29\linewidth}
            \includegraphics[width=\linewidth]{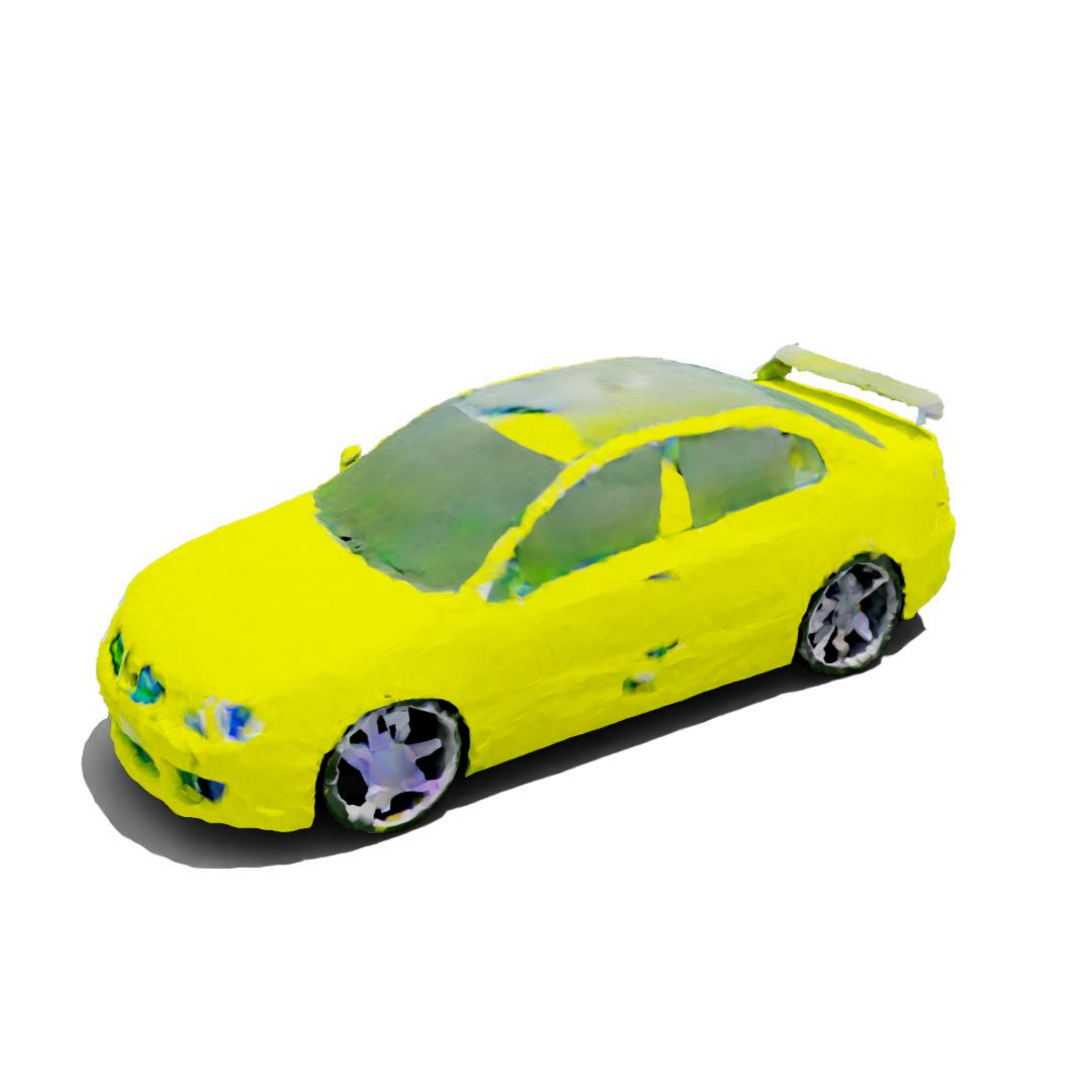}
        \end{subfigure}
        \begin{subfigure}[b]{0.29\linewidth}
            \includegraphics[width=\linewidth]{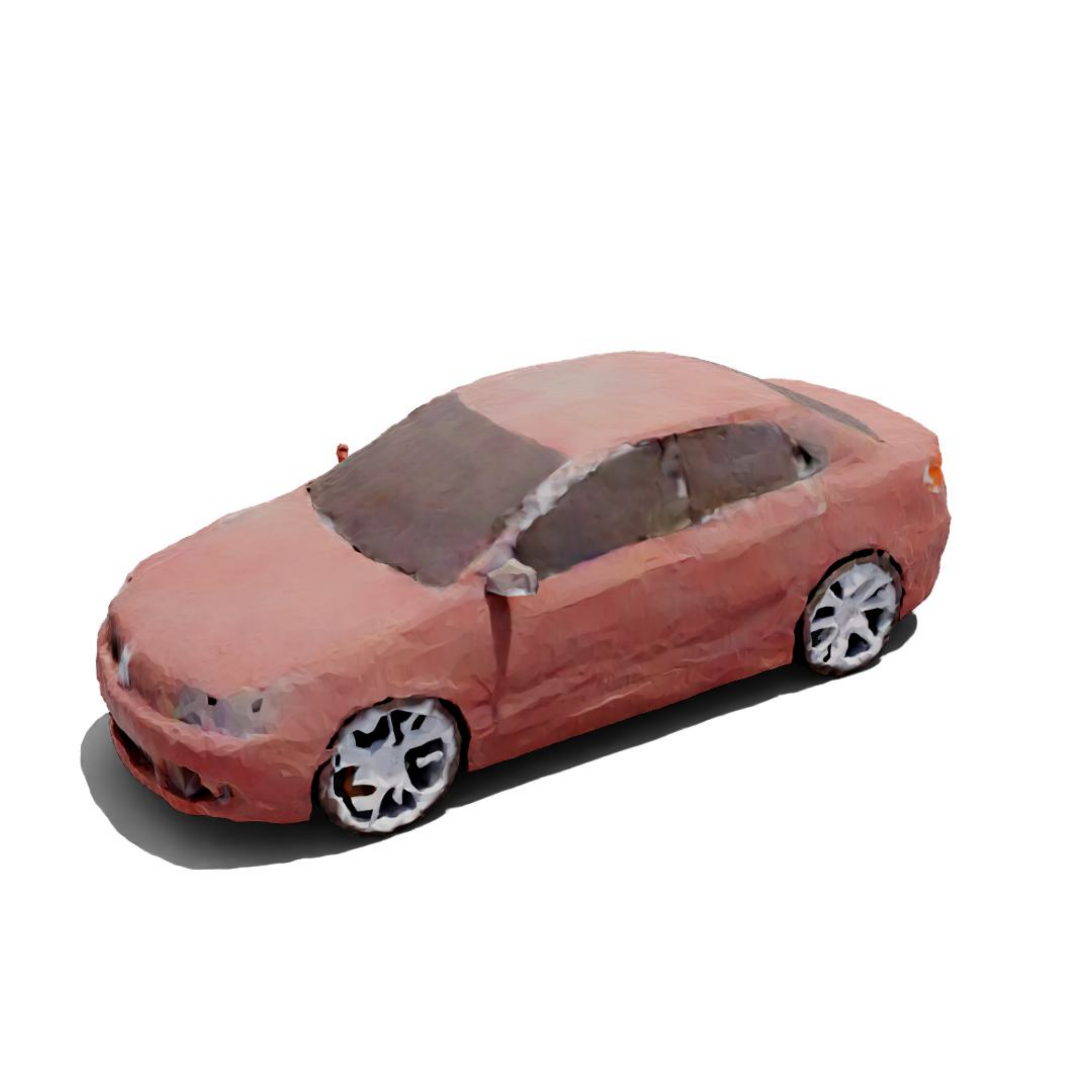}
        \end{subfigure}
        \begin{subfigure}[b]{0.29\linewidth}
            \includegraphics[width=\linewidth]{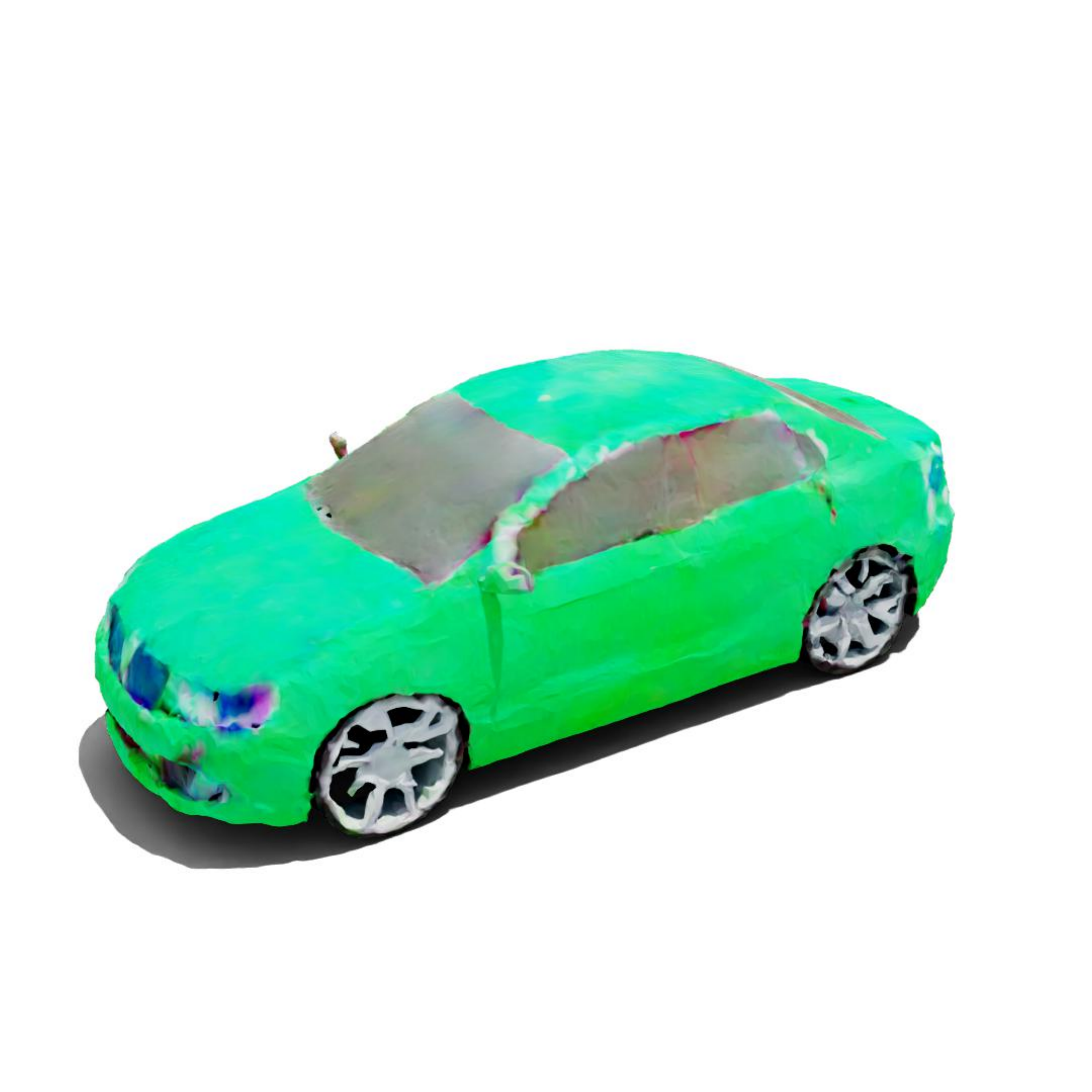}
        \end{subfigure}
        \begin{subfigure}[b]{0.29\linewidth}
            \includegraphics[width=\linewidth]{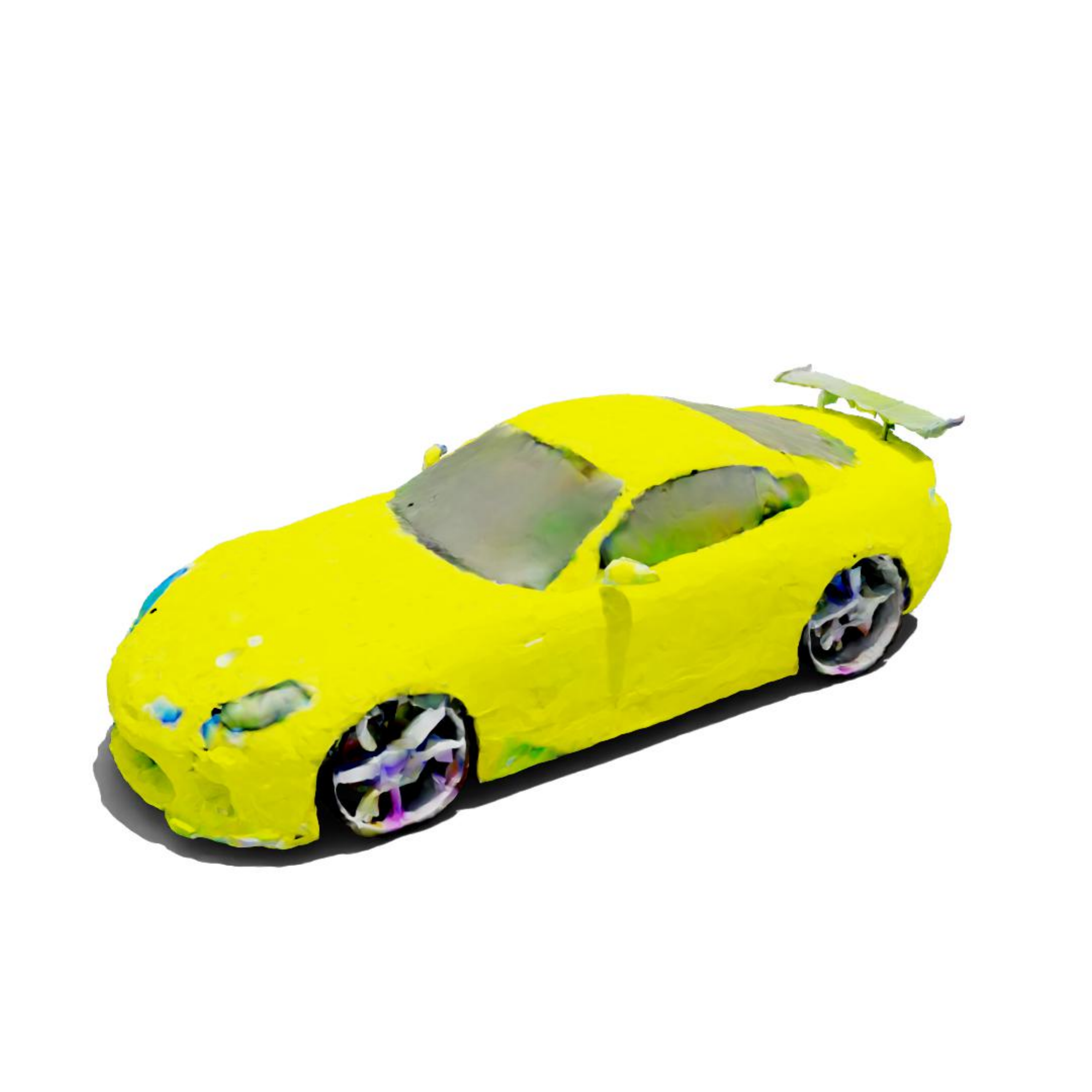}
        \end{subfigure}
        \begin{subfigure}[b]{0.29\linewidth}
            \includegraphics[width=\linewidth]{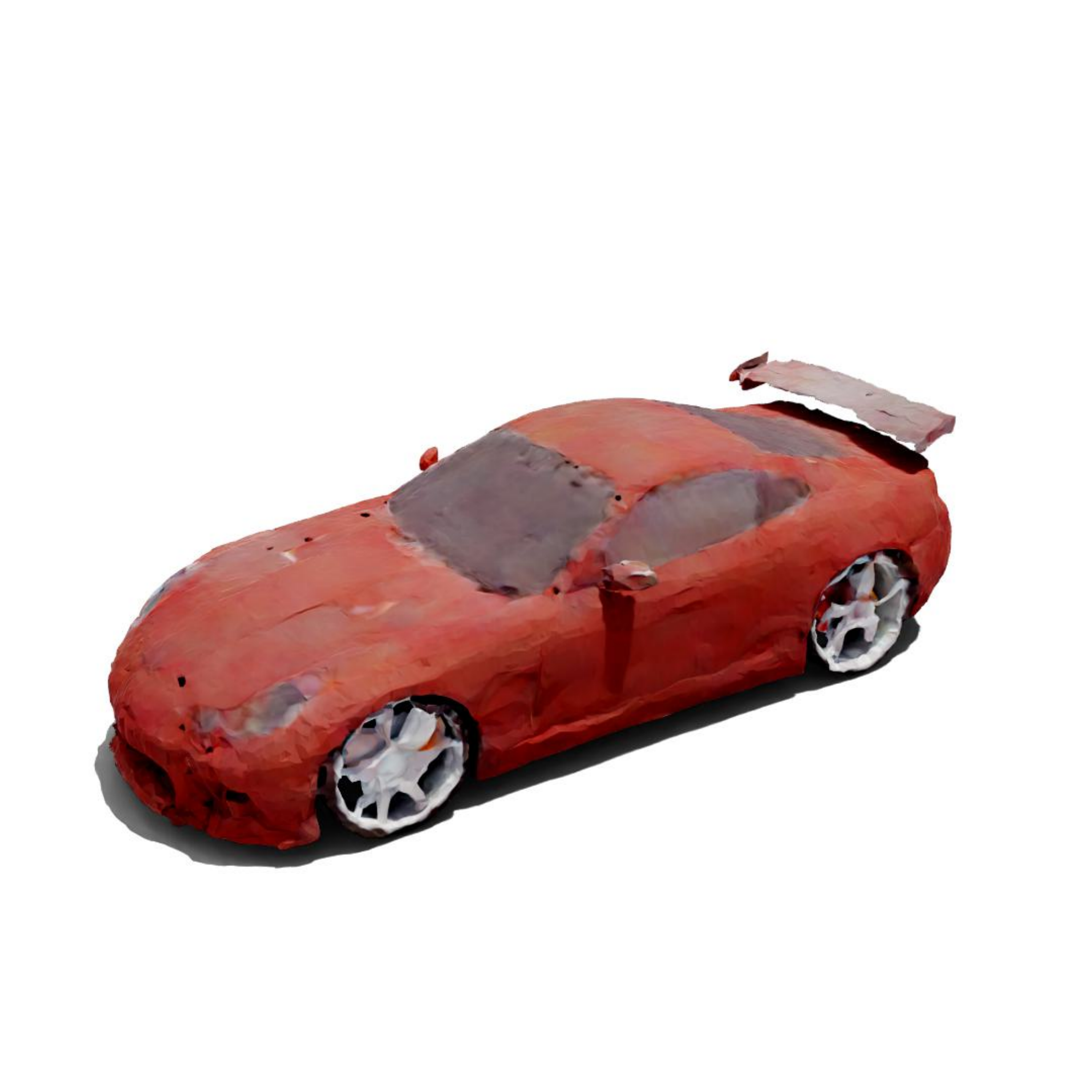}
        \end{subfigure}
        \begin{subfigure}[b]{0.29\linewidth}
            \includegraphics[width=\linewidth]{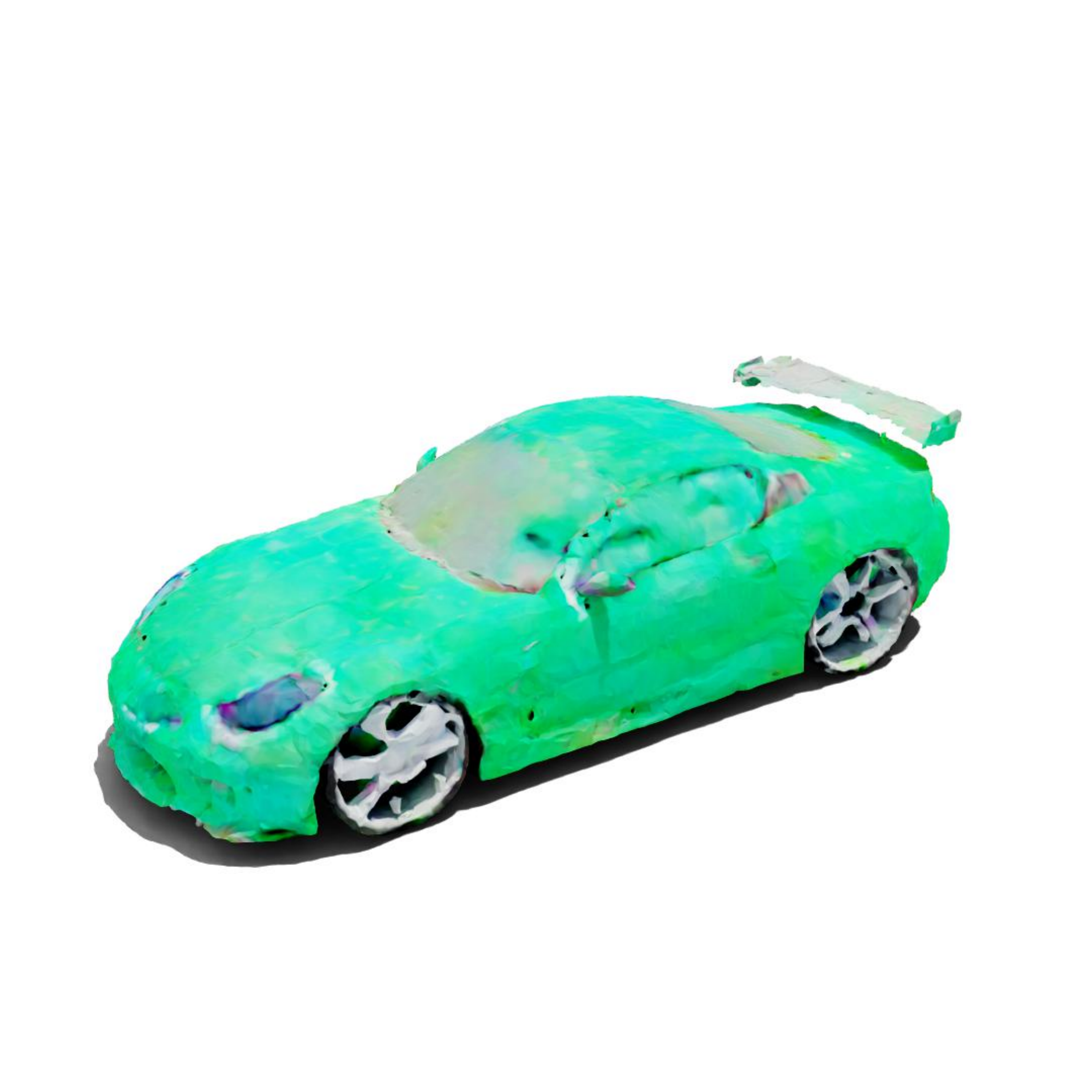}
        \end{subfigure}
        \begin{subfigure}[b]{0.29\linewidth}
            \includegraphics[width=\linewidth]{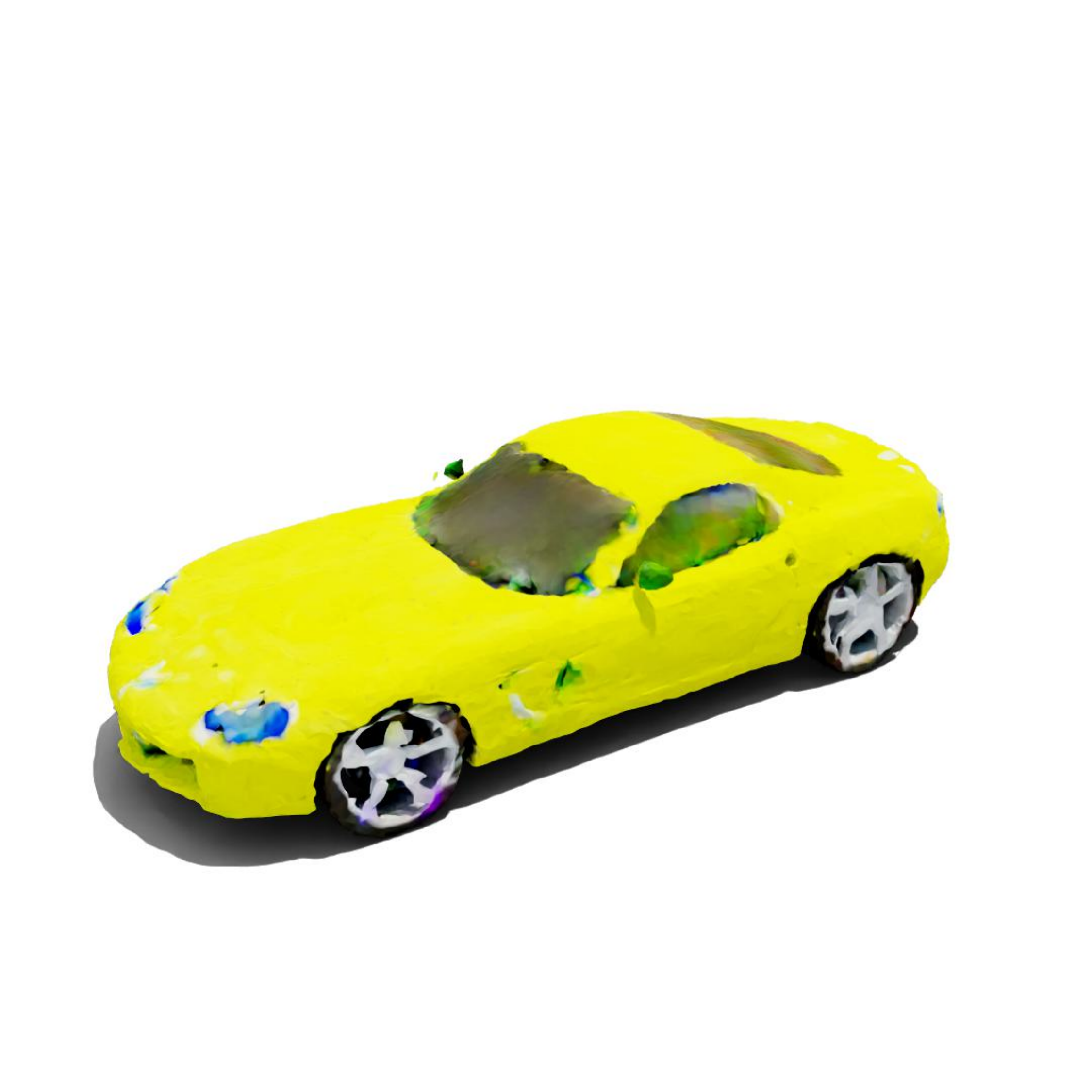}
        \end{subfigure}
        \hfill
        \begin{subfigure}[b]{0.29\linewidth}
            \includegraphics[width=\linewidth]{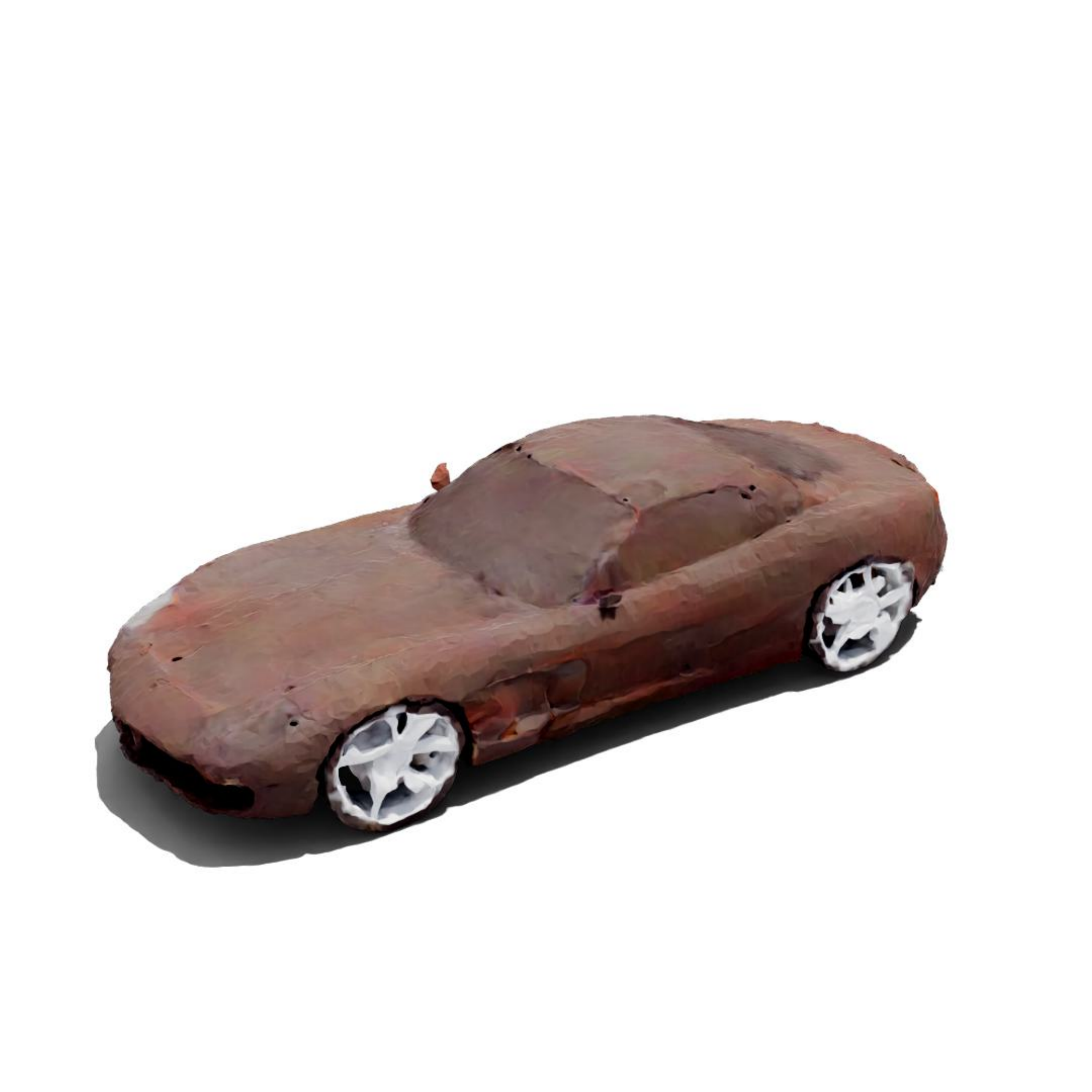}
        \end{subfigure}
        \hfill
        \begin{subfigure}[b]{0.29\linewidth}
            \includegraphics[width=\linewidth]{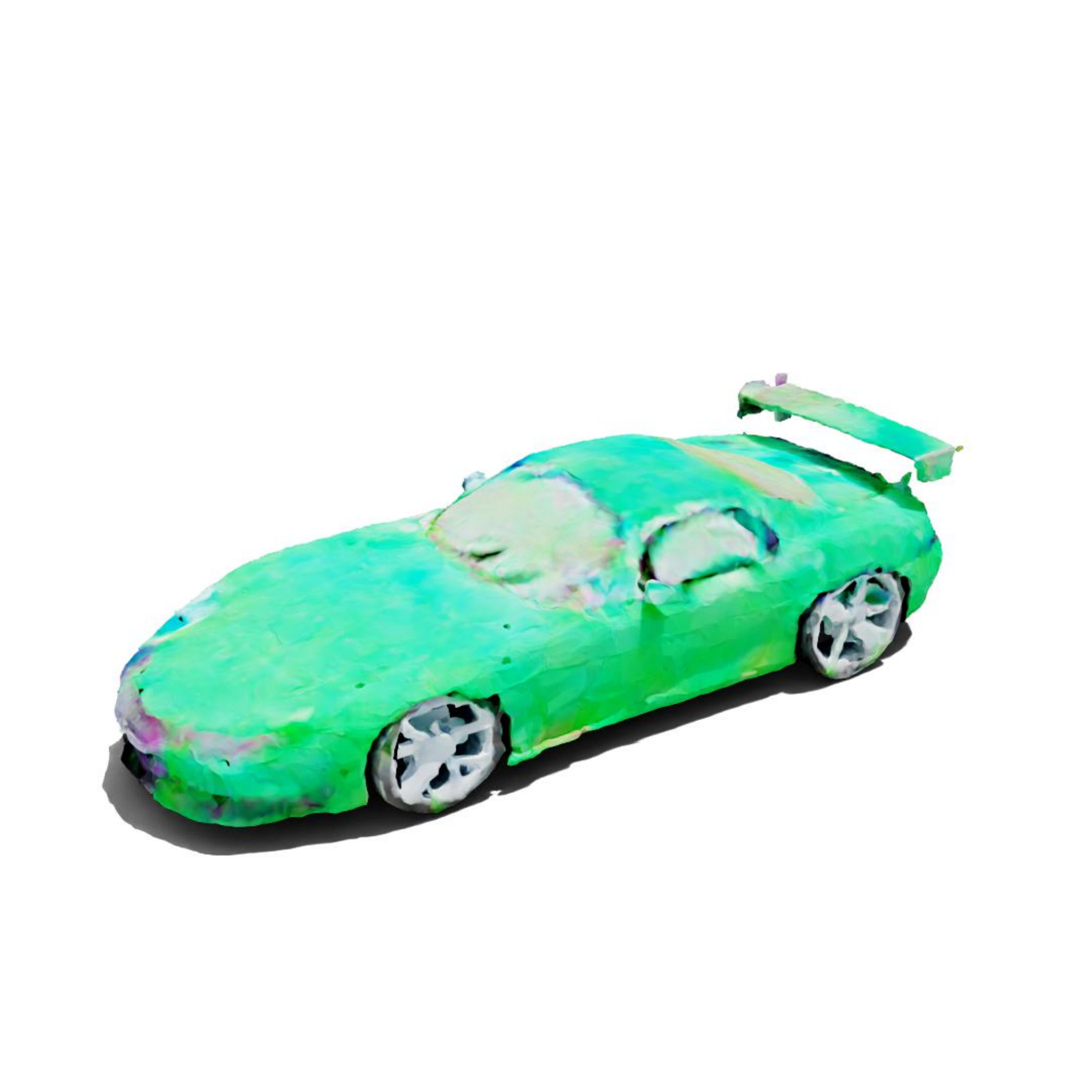}
        \end{subfigure}
    \end{subfigure}
    \begin{subfigure}[b]{0.08\linewidth}
        \begin{tikzpicture}
            \node at (0.0,0.0) {};
            \draw[latex-latex] (0,0.6) -- (0,4.);
            \filldraw[fill=white] (-0.4,3.8) circle (5pt);
            \filldraw[fill=white] (-0.3,0.8) circle (2pt);
            \draw[black, thick] (-0.1,2.2) -- (0.1,2.2);
            \node at (-0.5,2.2) {Vol.};
        \end{tikzpicture}%
    \end{subfigure}
    \begin{subfigure}[b]{0.45\linewidth}
        \begin{subfigure}[b]{0.29\linewidth}
            \includegraphics[width=\linewidth]{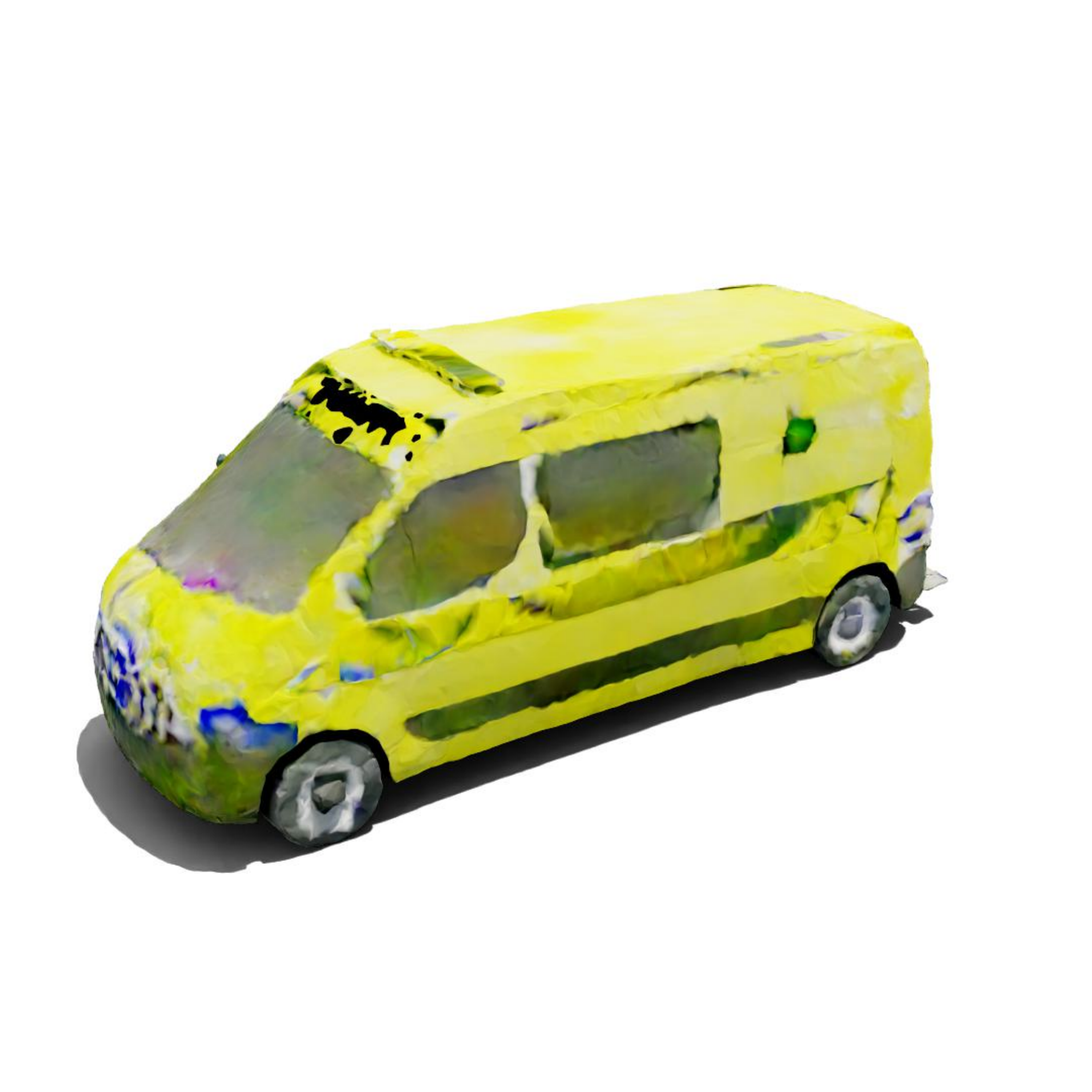}
        \end{subfigure}
        \begin{subfigure}[b]{0.29\linewidth}
            \includegraphics[width=\linewidth]{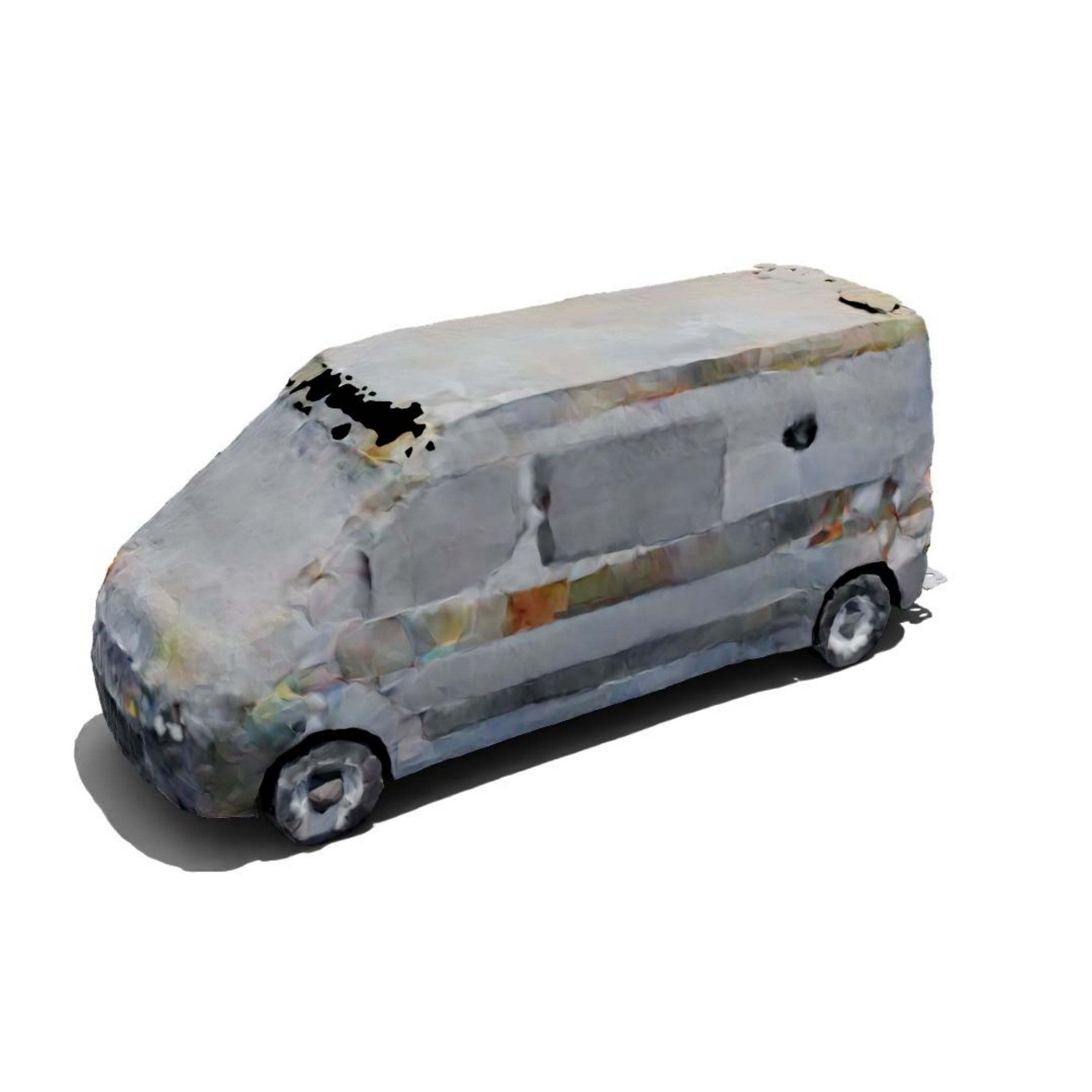}
        \end{subfigure}
        \begin{subfigure}[b]{0.29\linewidth}
            \includegraphics[width=\linewidth]{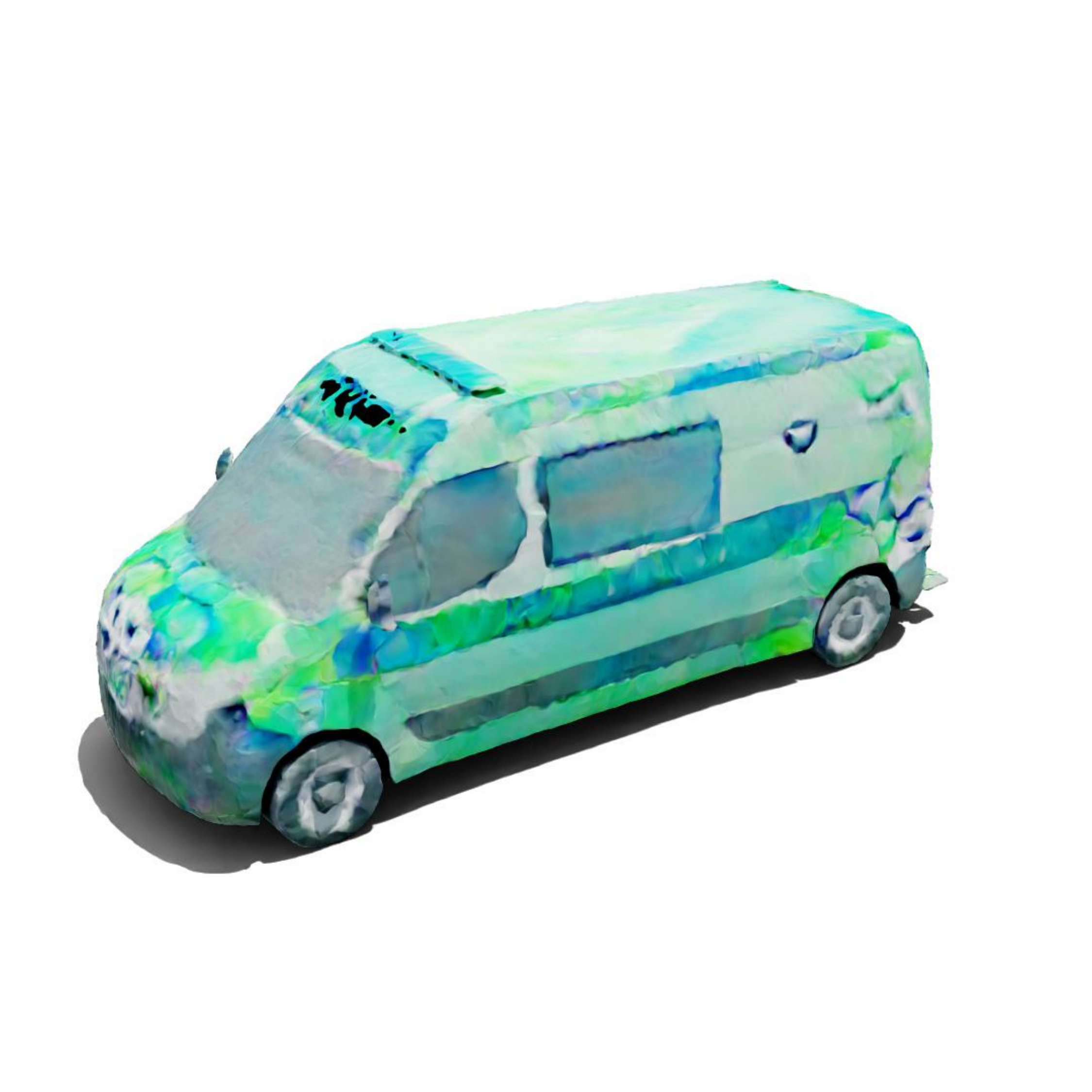}
        \end{subfigure}
        \begin{subfigure}[b]{0.29\linewidth}
            \includegraphics[width=\linewidth]{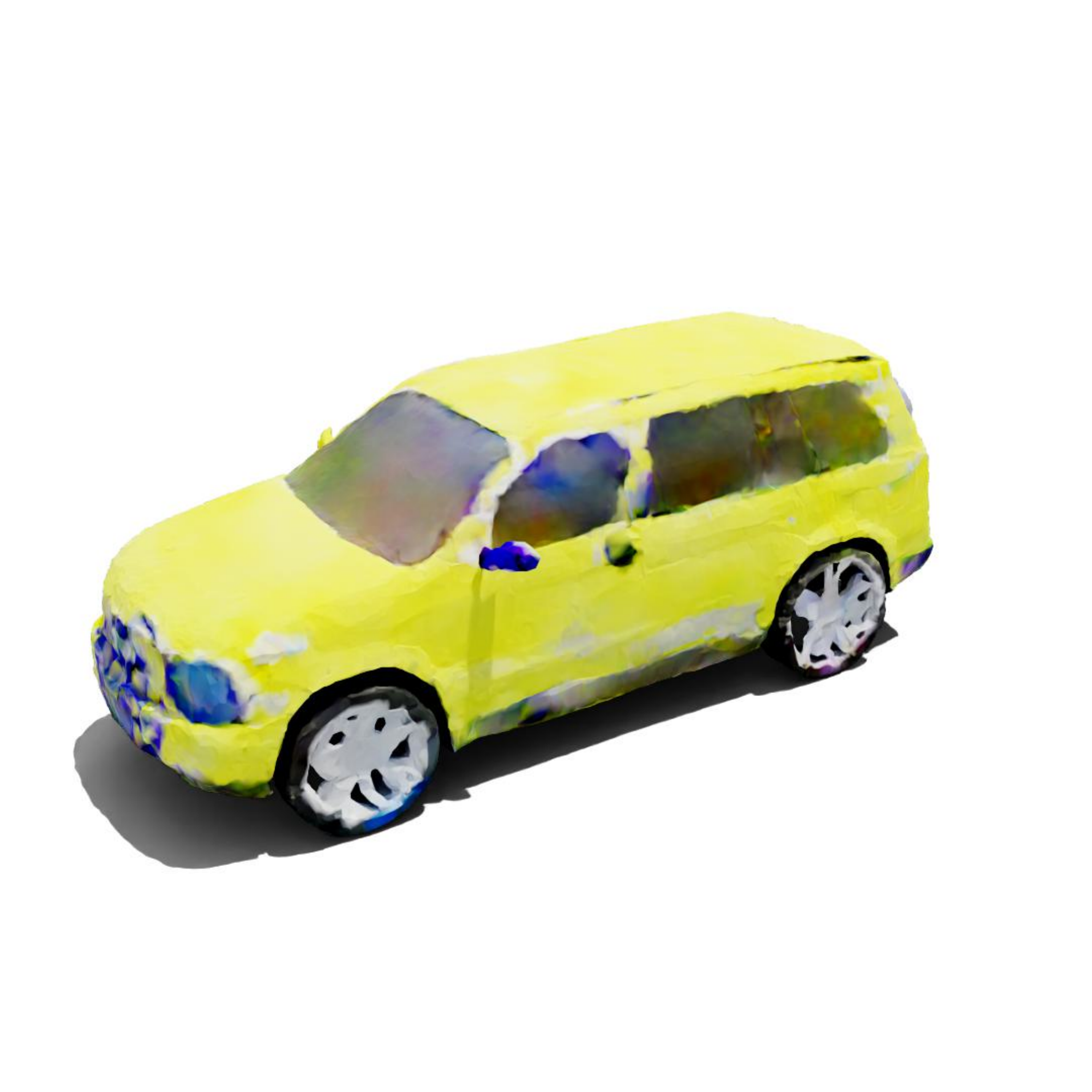}
        \end{subfigure}
        \begin{subfigure}[b]{0.29\linewidth}
            \includegraphics[width=\linewidth]{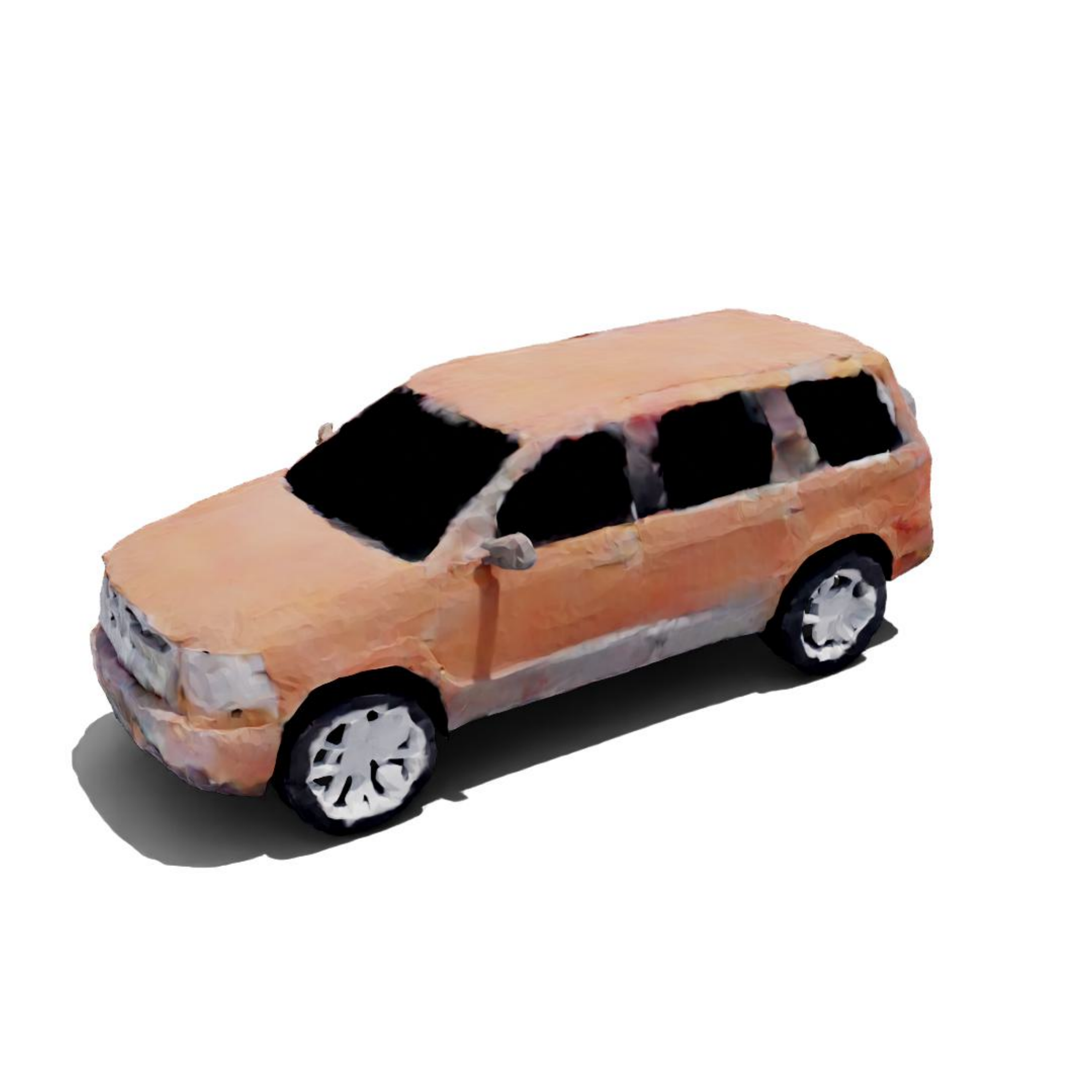}
        \end{subfigure}
        \begin{subfigure}[b]{0.29\linewidth}
            \includegraphics[width=\linewidth]{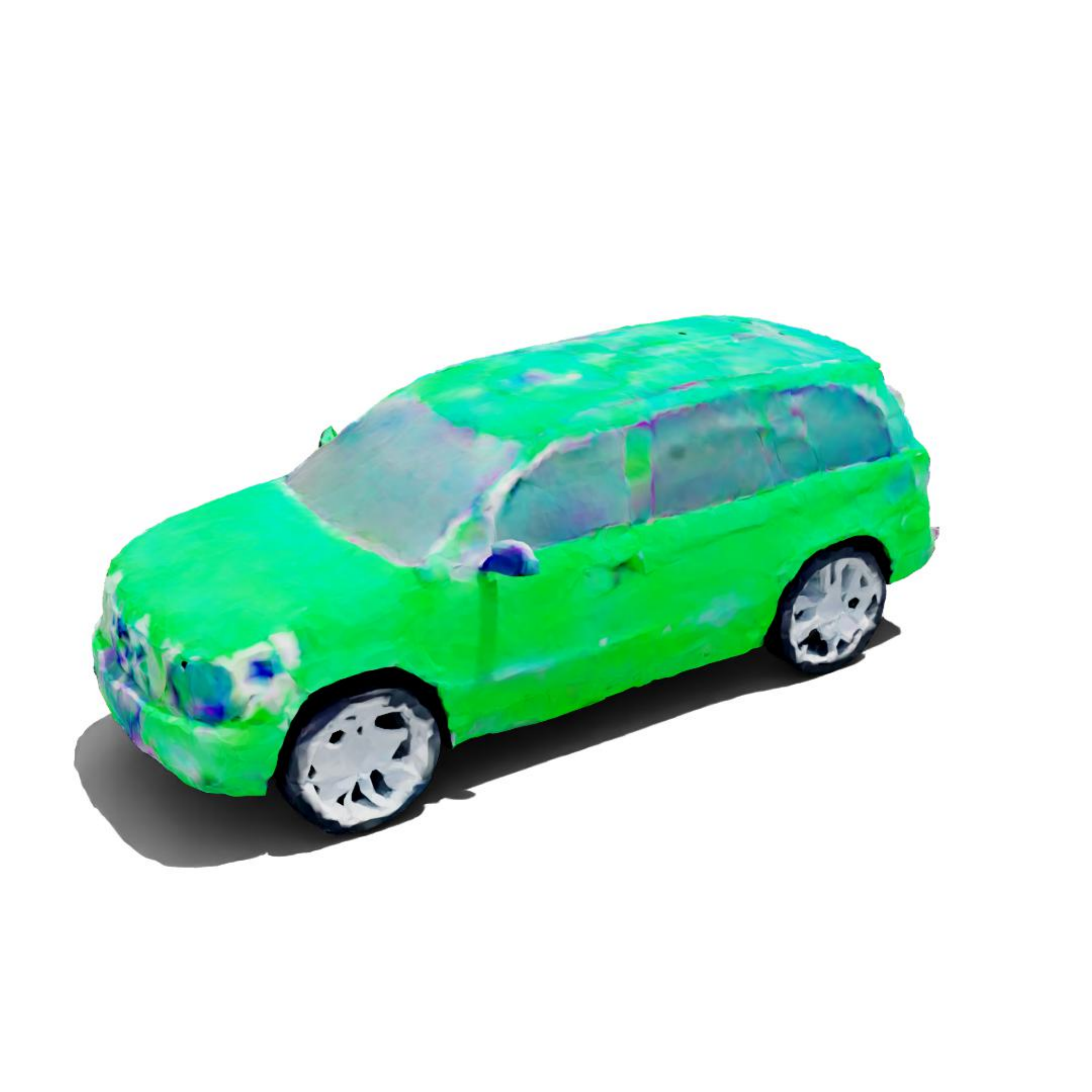}
        \end{subfigure}
        \begin{subfigure}[b]{0.29\linewidth}
            \includegraphics[width=\linewidth]{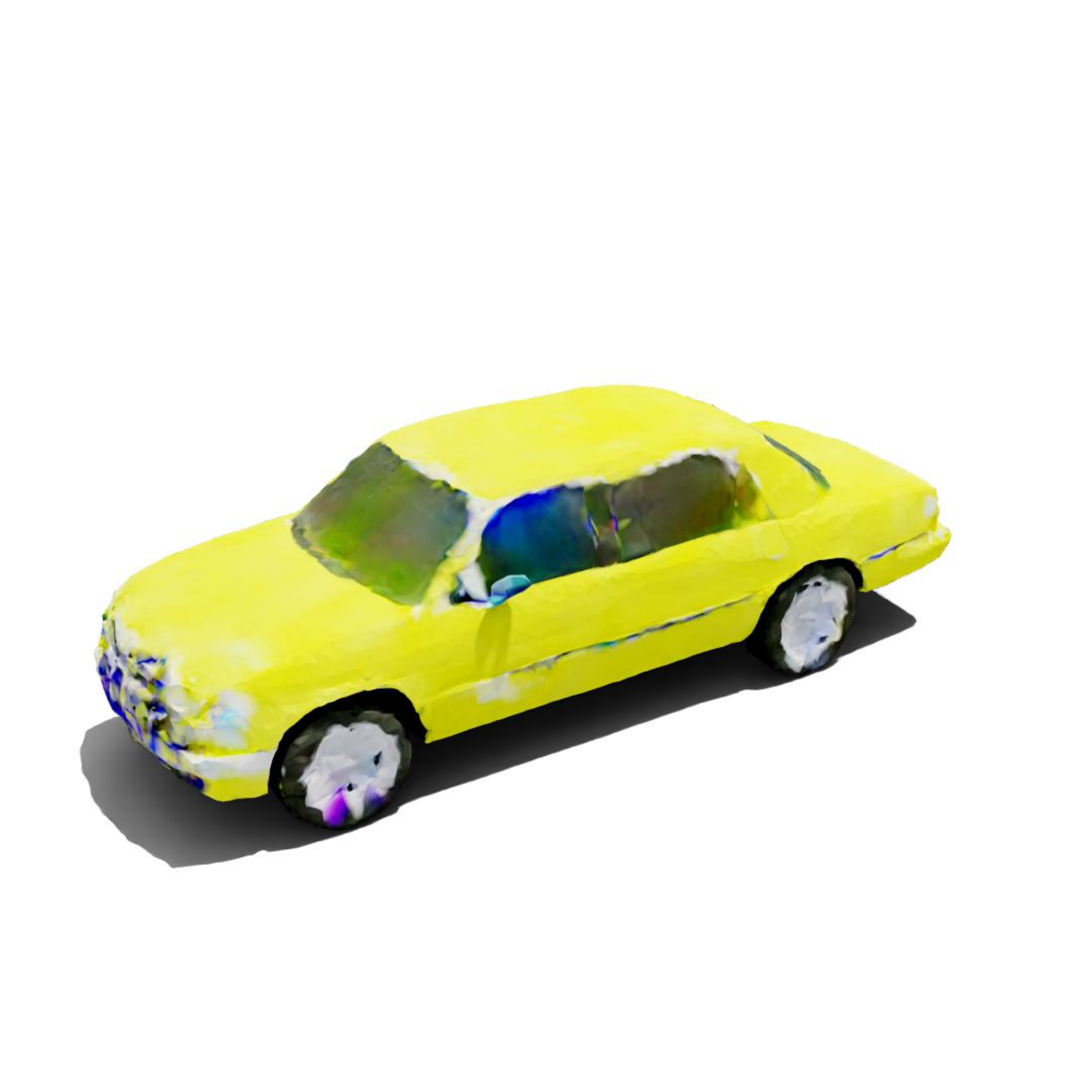}
        \end{subfigure}
        \hfill
        \begin{subfigure}[b]{0.29\linewidth}
            \includegraphics[width=\linewidth]{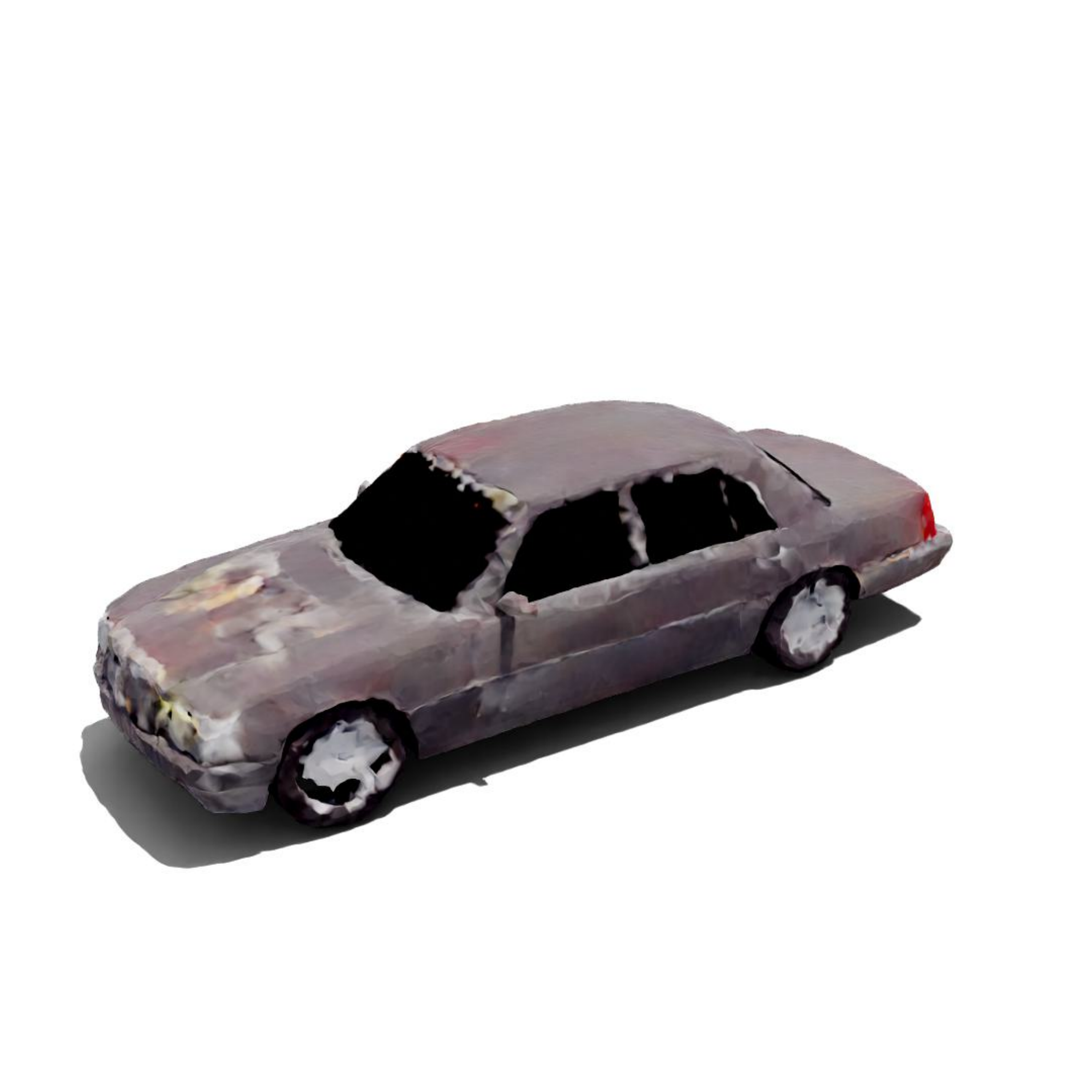}
        \end{subfigure}
        \hfill
        \begin{subfigure}[b]{0.29\linewidth}
            \includegraphics[width=\linewidth]{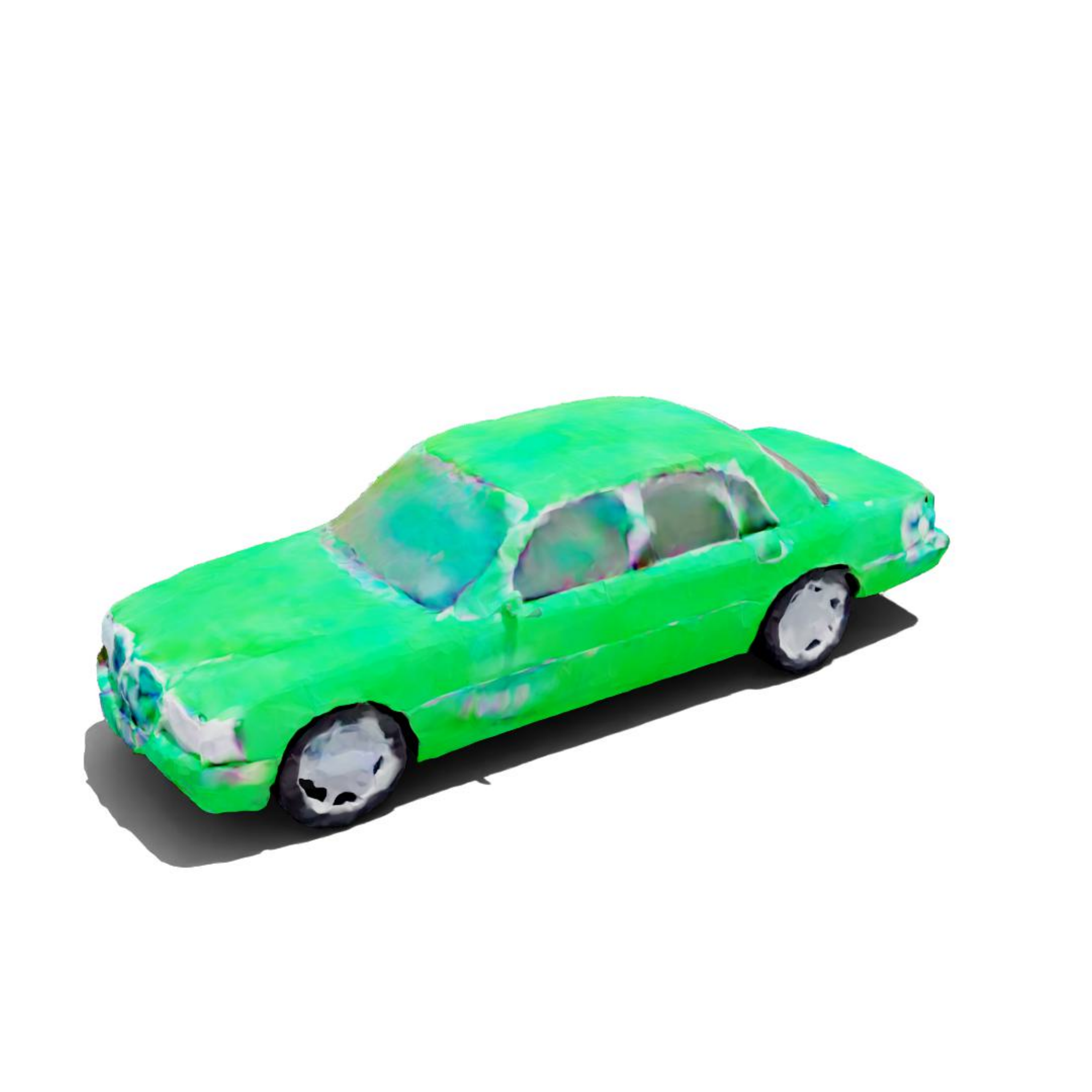}
        \end{subfigure}
    \end{subfigure}
    \hfill
    \begin{subfigure}[b]{0.45\linewidth}
        \begin{subfigure}[b]{0.29\linewidth}
            \includegraphics[width=\linewidth]{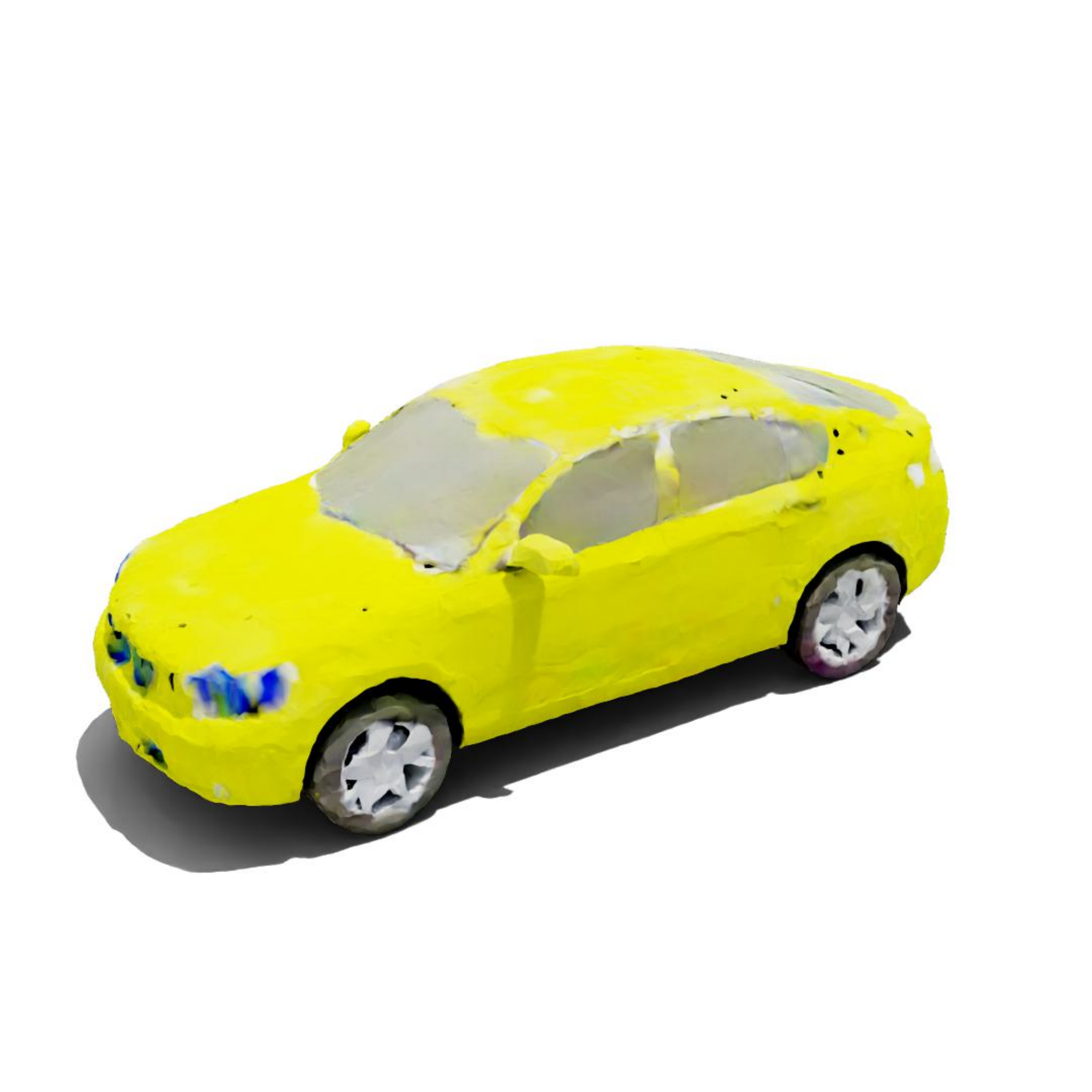}
        \end{subfigure}
        \begin{subfigure}[b]{0.29\linewidth}
            \includegraphics[width=\linewidth]{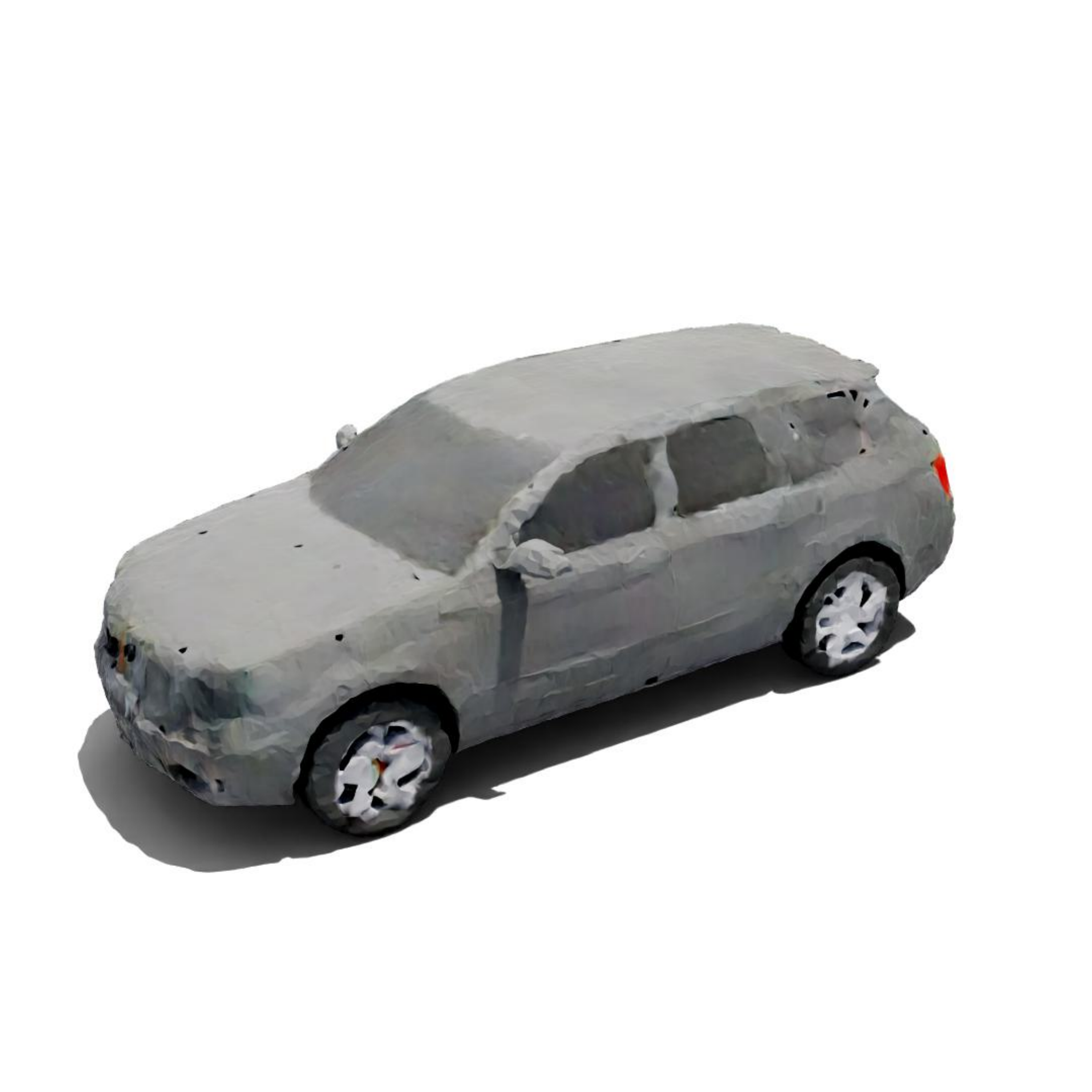}
        \end{subfigure}
        \begin{subfigure}[b]{0.29\linewidth}
            \includegraphics[width=\linewidth]{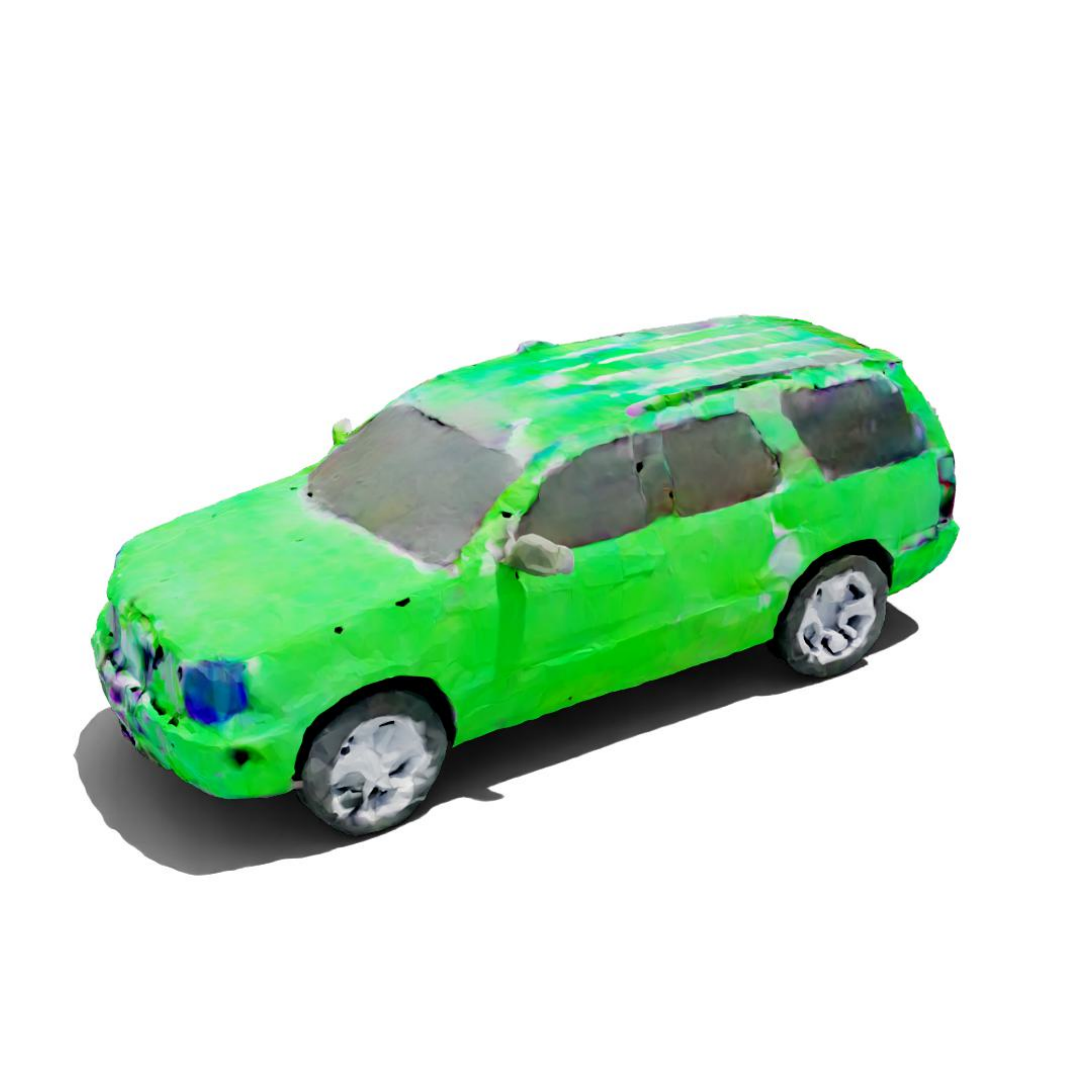}
        \end{subfigure}
        \begin{subfigure}[b]{0.29\linewidth}
            \includegraphics[width=\linewidth]{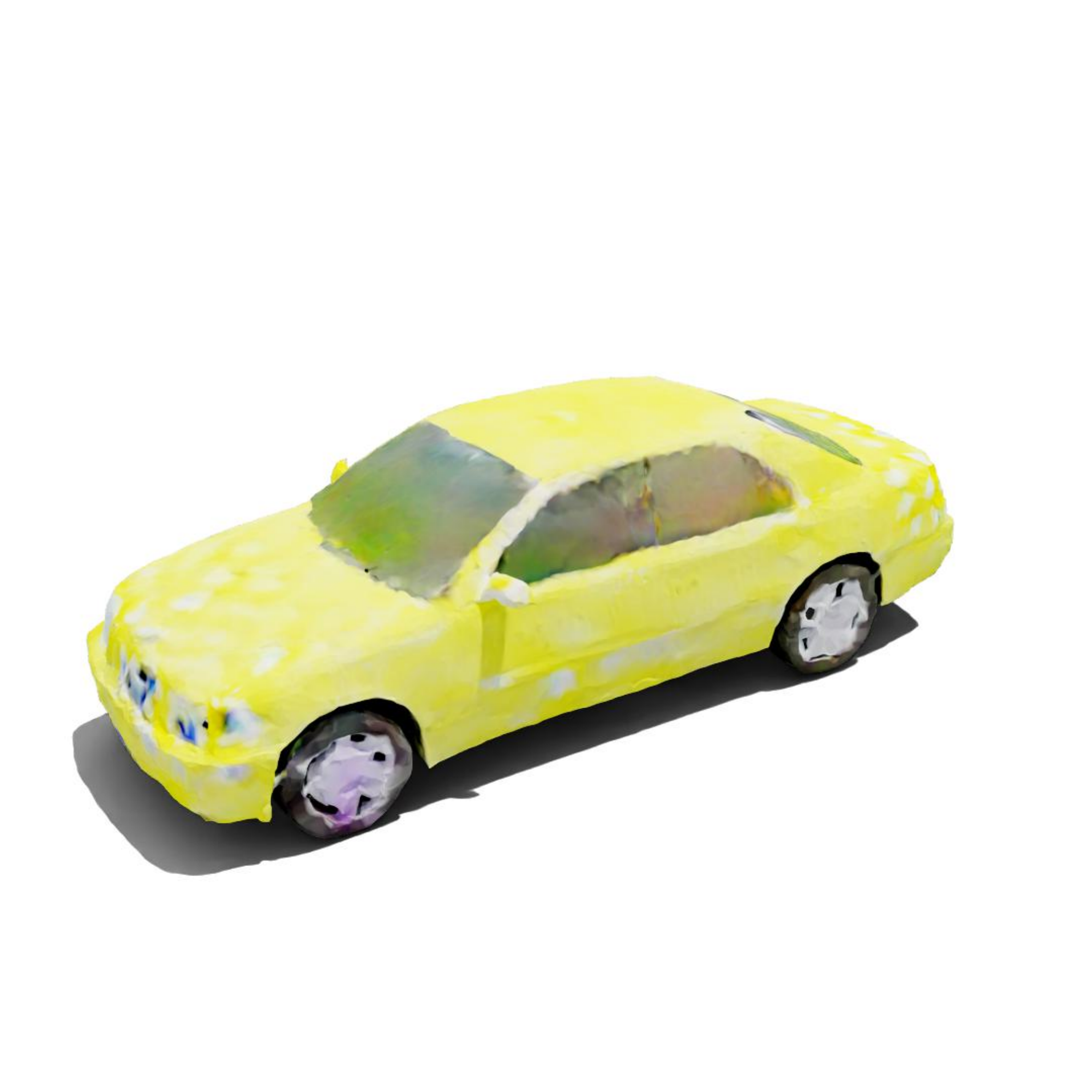}
        \end{subfigure}
        \begin{subfigure}[b]{0.29\linewidth}
            \includegraphics[width=\linewidth]{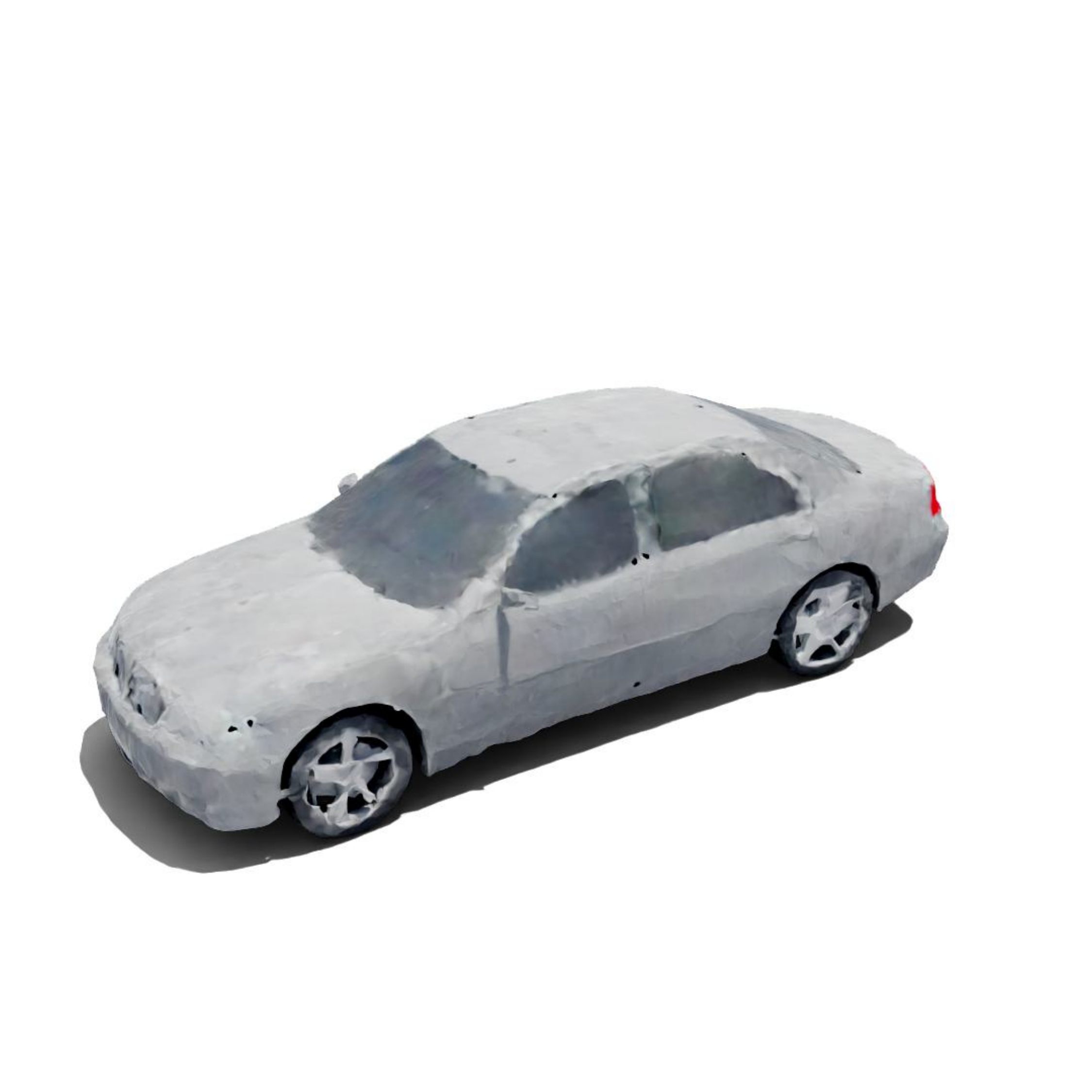}
        \end{subfigure}
        \begin{subfigure}[b]{0.29\linewidth}
            \includegraphics[width=\linewidth]{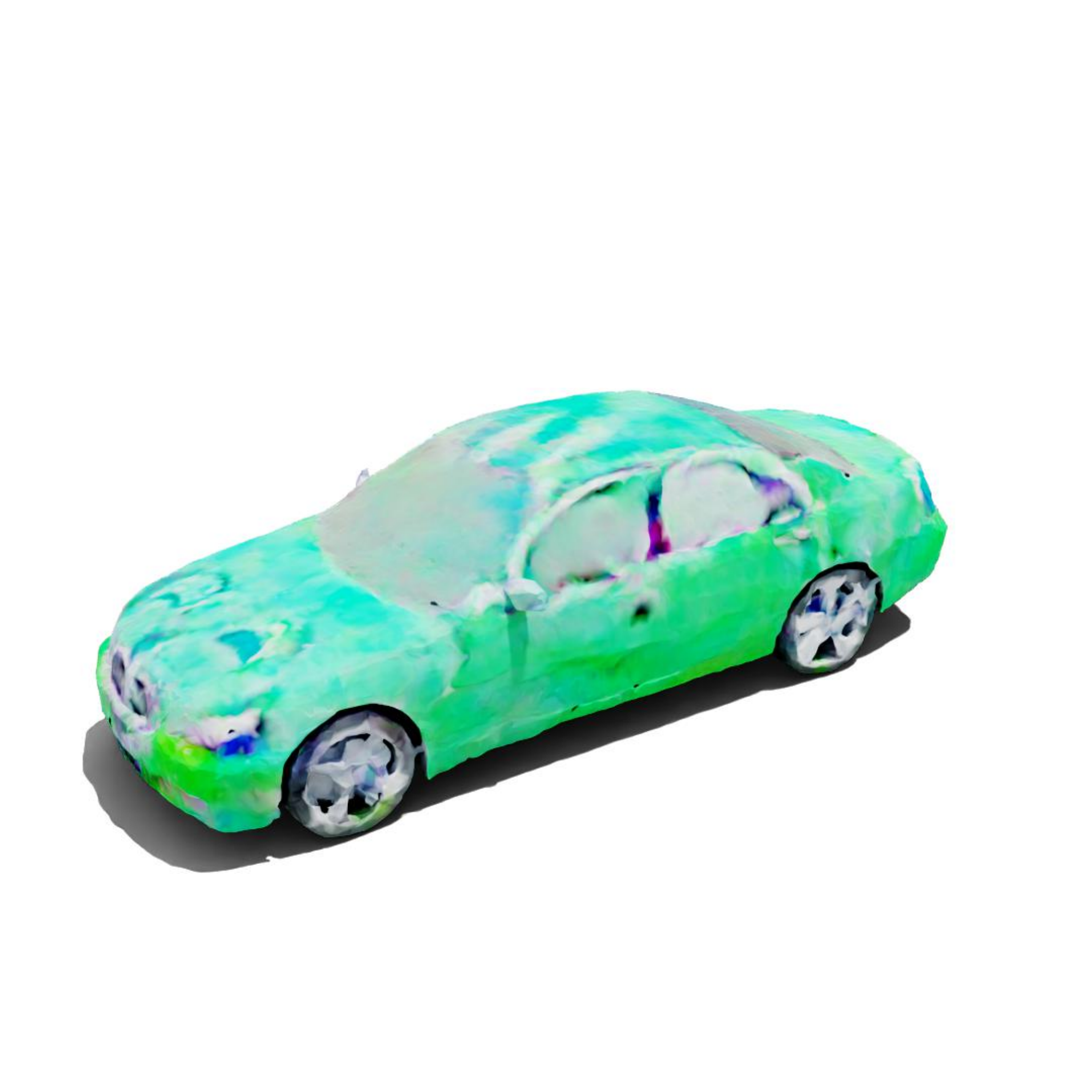}
        \end{subfigure}
        \begin{subfigure}[b]{0.29\linewidth}
            \includegraphics[width=\linewidth]{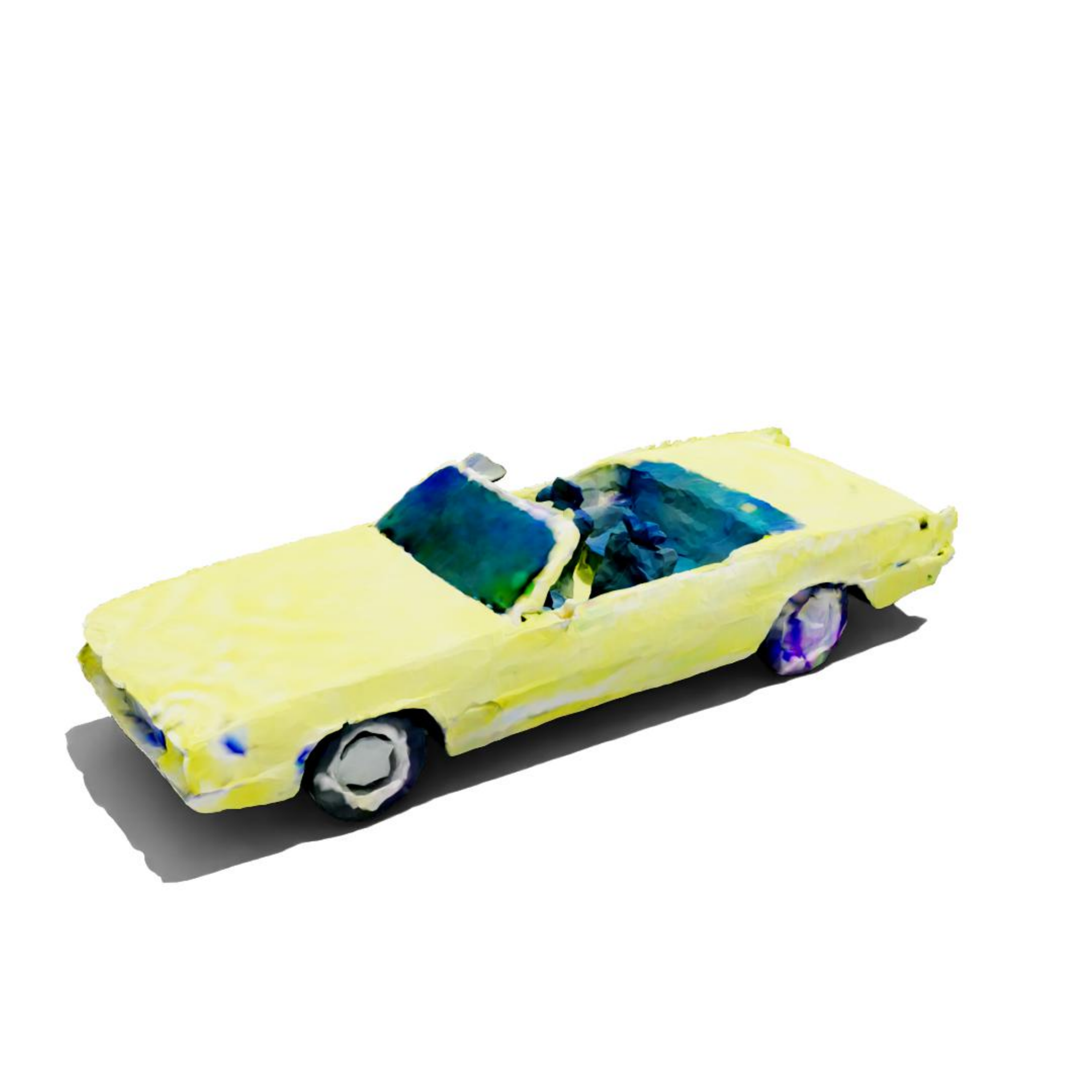}
        \end{subfigure}
        \hfill
        \begin{subfigure}[b]{0.29\linewidth}
            \includegraphics[width=\linewidth]{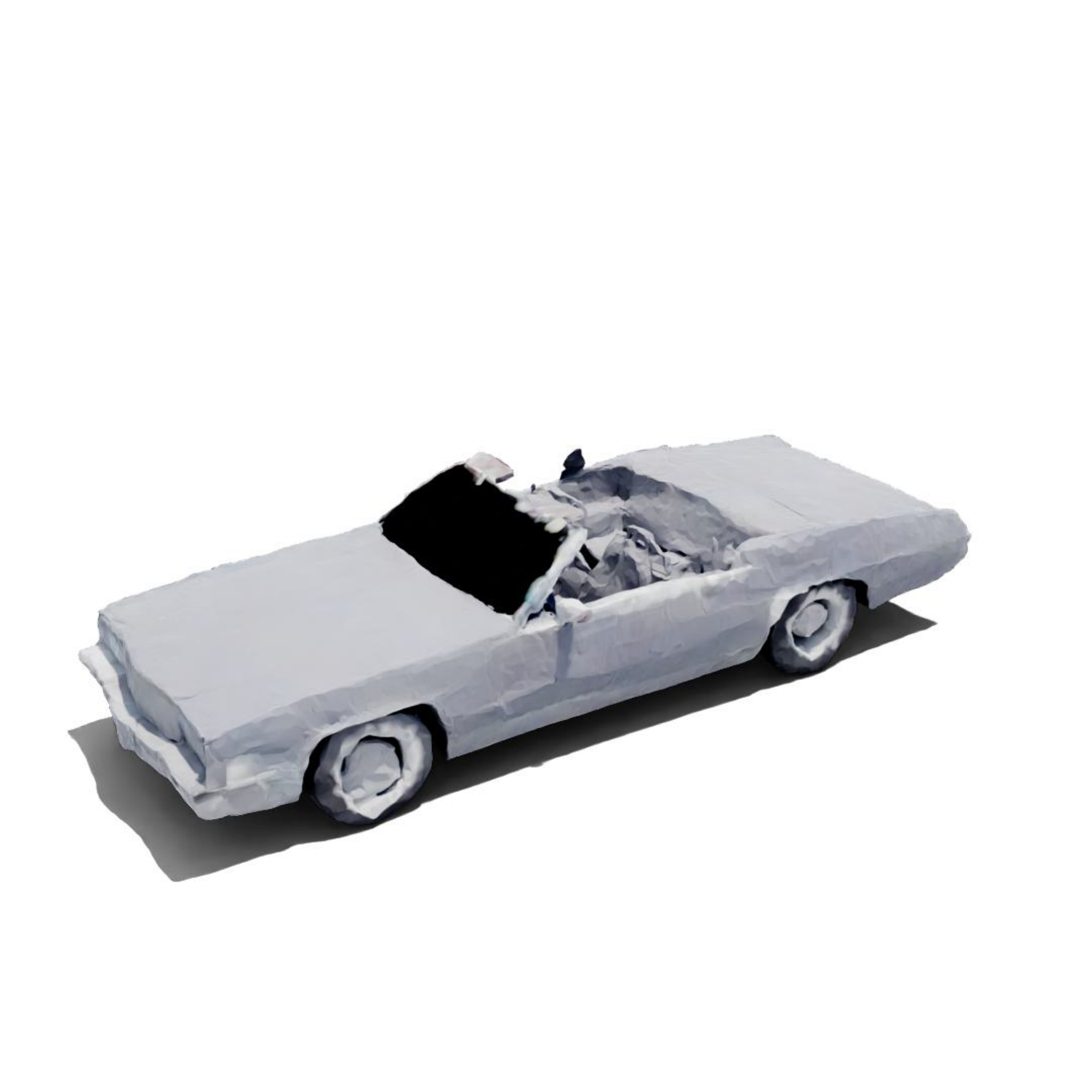}
        \end{subfigure}
        \hfill
        \begin{subfigure}[b]{0.29\linewidth}
            \includegraphics[width=\linewidth]{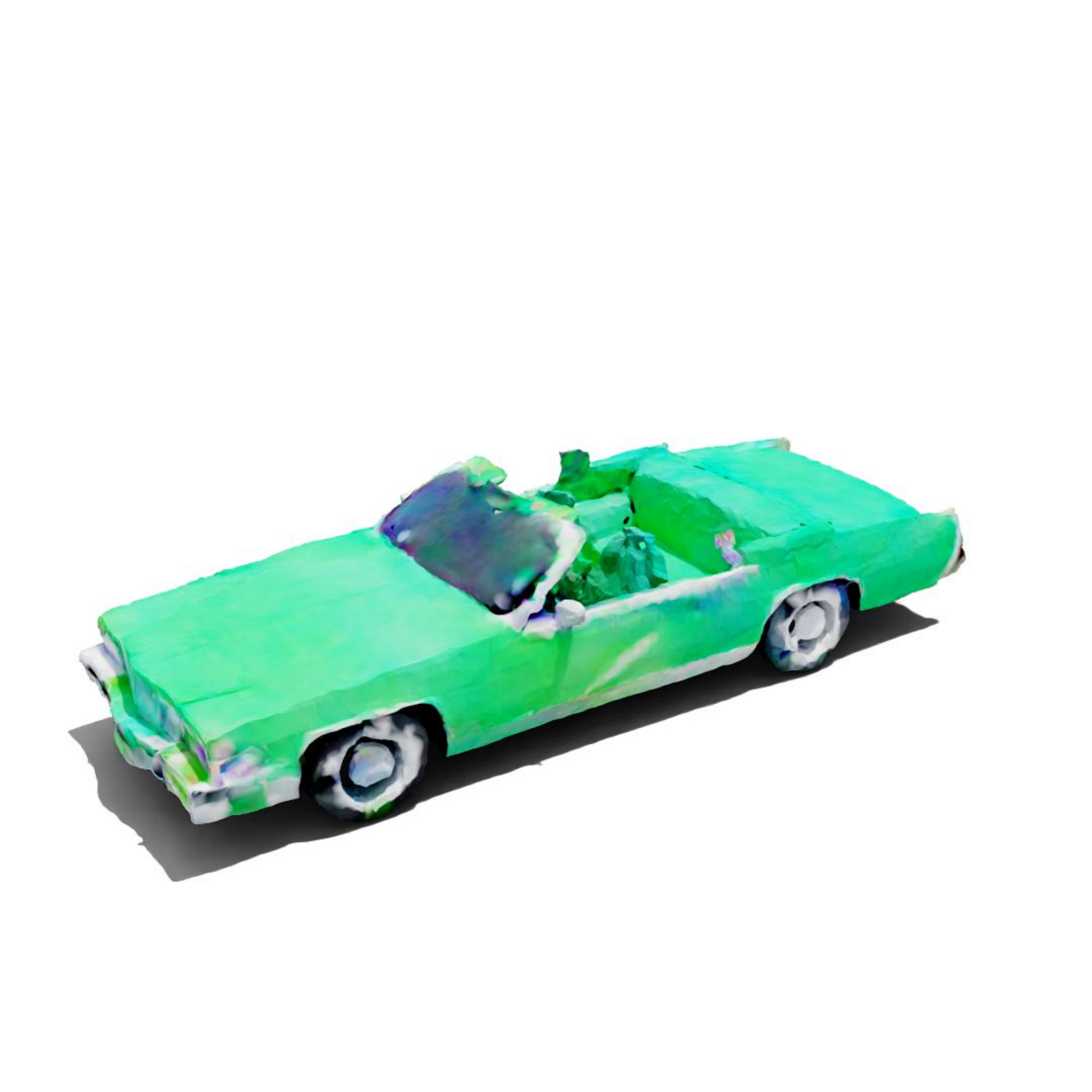}
        \end{subfigure}
    \end{subfigure}
    \begin{flushright}
        \begin{subfigure}[b]{0.94\textwidth}
            \begin{tikzpicture}
              \draw[-latex] (0,0) -- (0.40\linewidth,0);
            
              \filldraw[fill=yellow] (0,0) circle (3pt) node[below=6pt] {yellow};
              \filldraw[fill=green] (0.40\linewidth,0) circle (3pt) node[below=6pt] {green};
            
              \draw[black, thick] (0.2\linewidth,-0.1) -- (0.2\linewidth,0.1);
            \end{tikzpicture}
            \hfill
            \begin{tikzpicture}
              \draw[-latex] (0,0) -- (0.40\linewidth,0);
            
              \filldraw[fill=yellow] (0,0) circle (3pt) node[below=6pt] {yellow};
              \filldraw[fill=green] (0.40\linewidth,0) circle (3pt) node[below=6pt] {green};
            
              \draw[black, thick] (0.2\linewidth,-0.1) -- (0.2\linewidth,0.1);
            \end{tikzpicture}
        \end{subfigure}
    \end{flushright}
    \caption{\textbf{Test time guidance.} Exemplary results of text and volume guidance, using our unconditional model.}
    \label{fig:testtimeguidance}
\end{figure}

These guidance tasks are evidently orthogonal to each other and can also be combined to change volume and color at the same time, due to our somewhat disentangled representation. We depict guided samples in \Cref{fig:testtimeguidance}. As pointed out by~\cite{nichol2021glide}, this simple form of guidance may slightly degrade the results, as renderings of noisy meshes are out of distribution for the CLIP model.

\clearpage
\newpage

\section{Tetrahedral Diffusion}

\subsection{Optimized Edge Binning Algorithm}
\label{sec:EdgeBinning}
\begin{algorithm}
\caption{Optimized Binning for Tetrahedral Grid Convolution}
\begin{algorithmic}[1]
\Require Grid vertices $V$, cluster count $k = m + 1$, max number of iterations $n$
\Ensure Ordered edge sets for each vertex

\State Initialize $k$-means with $k$ clusters
\For{each vertex $v \in V$}
    \State Collect outgoing edges $E_v$ centered at $v$
\EndFor
\State Cluster all edges $E = \bigcup_v E_v$ into $k$ groups

\For{each vertex $v \in V$}
    \For{each edge $e \in E_v$}
        \State Assign $e$ to closest cluster based on cosine similarity
    \EndFor
    \State Sort assigned edges of $v$ based on cluster order
\EndFor

\State Check for collisions in edge assignments
\If{collisions exist}
    \If{current iteration $<n$}
    \State repeat clustering and sorting
    \Else 
    \State Increment $k$ and repeat clustering and sorting
    \EndIf
\EndIf
\label{alg:binning}
\end{algorithmic}
\end{algorithm}

As already outlined in the main paper, the hybrid tetrahedral representation cannot be manipulated with regular grid-based convolution, since the neighborhood of a vertex in the grid can vary. However, even though the tetrahedral grid initialized with Quartet \url{https://github.com/crawforddoran/quartet} is relatively regular, it is not straightforward to establish a discrete binning of the local edge orientations that would define a unique, collision-free ordering. To resolve this we propose to globally optimize the binning in such a way that a unique ordering is ensured, see Alg.~\ref{alg:binning}
\begin{enumerate}[leftmargin=\labelwidth]
\item Collect the set of outgoing edges at all vertices of the (undeformed) grid, center them in one point, and cluster them into $k=m+1$ groups with $k$-means. The cluster centers serve as a global basis of reference directions.
\item Arrange the basis vectors in a fixed order. At every vertex, assign the outgoing edges to basis vectors according to the cosine similarity to obtain a spatially meaningful ordering For convolution.
\item If, at any vertex, the binning would lead to a collision, increment $k$ to expand the number of basis vectors and repeat clustering and assignment.
\end{enumerate}
In practice, we have never observed a case where a collision-free basis needed more than $m+1$ reference directions.

\subsection{Tetrahedral marching extension}\label{sec:marchingtet}

The differentiable Marching Tetrahedra algorithm was popularized by~\cite{shen2021deep} and used in various works~\cite{gao2022get3d, gao2022tetgan, munkberg2022extracting, gupta20233dgen}. Similar to the Marching Cube scheme, that algorithm individually inspects every tetrahedron to determine the surface topology. Whenever the sign of the SDF changes within a tetrahedron, the surface must pass through it and one can extract the corresponding triangle by interpolating the vertex positions:
\begin{equation}\label{eq:marchingtet}
   v_{ab} = \frac{(v_a + \Delta_{v_a}) \cdot s_b - (v_b + \Delta_{v_b}) \cdot s_a}{s_b - s_a},
\end{equation}
where $s_x$ is the SDF value of vertex $x$ and $\Delta_{v_x}$ its corresponding deformation vector, $x\in \{a,b\}$.

In our case, we not only want to diffuse untextured meshes, but include texture information without using multiple networks. Luckily, extending the Marching Tetrahedra algorithm to extract any mesh vertex feature is as simple as adding features in the numerator of \cref{eq:marchingtet}, \ie $(v_x + \Delta_{v_x} + c_{v_x} + ... + *_{v_a})$, where $c_{v_x}$ can, for instance, be the color.

\subsection{Network}
\label{sec:network}

\begin{figure}[!ht]
    \centering
    \includegraphics[width=\linewidth]{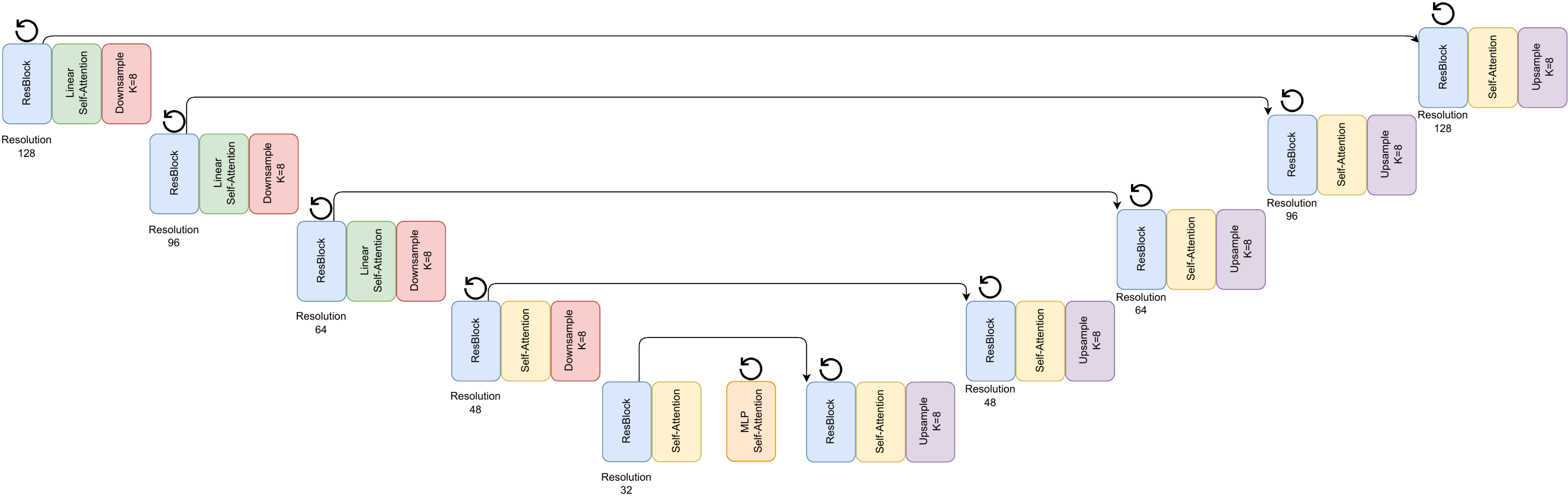}
    \caption{\textbf{Schematic of our U-Vit in standard resolution.} Every ResBlock, Downsample and Upsample contain TetraConv blocks.}
    \label{fig:networkstructure}
\end{figure}

The general structure of our network is depicted in Fig~\ref{fig:networkstructure}. Our standard network (resolution 128) has five encoder stages with $\{128,128,128,512,1024 \}$ channels. Each stage consists of two residual blocks with tetrahedral convolutions followed by a linear attention layer~\cite{wang2020linformer}. In the last two stages the linear attention layers are substituted with regular self-attention layers. Our bottleneck comprises four Transformer blocks, each with self-attention followed by a Multilayer Perceptron (MLP). The time parameter $t$ is embedded via feature modulation~\cite{perez2018film}. Our downsampling rate is 0.75, \eg going from a 128-cube to a 96-cube. The lowest resolution is 32. In the following, we list the most important training parameters.
\usemintedstyle{fruity}
\begin{minted}[frame=single, bgcolor=bg]{yaml}
cube_res: [128, 96, 64, 48, 32]
first_channel: 128
vit_depth: 4
num_blocks: [2, 2, 2, 2, 2]  
channel_multiplier: [1, 1, 1, 4, 8]
dropout: [0.0, 0.0, 0.0, 0.1, 0.1] 
linear_attentions: [True, True, True, True, False]
attentions: [False, False, False, False, True] 
group_norm_groups: 32 
knn_up_downsample: 8
ema_decay: 0.995
lr: 1.e-4
\end{minted}

Differently, for our high resolution network, \ie a resolution of 192, we directly downsample from 192 to 128, so as to reduce memory requirements. To increase information flow we use a nearest neighborhood size of 16 instead of 8 ing the downsampling operator. Additionally, we only use linear attention layers to accelerate sampling. We have not encountered a loss in performance using the higher downsampling rate and only linear attention layers. Below we list only parameters that are different in the high-resolution version.

\begin{minted}[frame=single, bgcolor=bg]{yaml}
cube_res: [192,128,64,48,32]
channel_multiplier: [1, 1, 2, 4, 6]
linear_attentions: [True, True, True, True, True]
attentions: [False, False, False, False, False] 
knn_up_downsample: 16
\end{minted}

We have found noise shift to be an important parameter when diffusing high resolution shapes. Following \cite{hoogeboom2023simple}, we shift the noise scheduler by a factor of $\frac{128}{32}=4$, which increases the noise level throughout the diffusion processes. We hypothesize that a combination of noise shift, $\nu$-parametrization and the time-continuous formulation is what makes it possible to sample high quality shapes within 24 steps.

\subsection{Pruning}
\label{sec:pruning}
Similar to other volumetric representations, the tetrahedral grid is typically sparsely populated. This reflects the fact that only a small portion of 3D space lies close to any object surface. However, our fully tetrahedral formulation allows us to alleviate this disadvantage, boosting computational efficiency. To do so, we systematically remove sections of the grid that are infrequently or never used. 
Unlike regular 3D convolutions, which permit only axis-parallel cropping, our tetrahedral convolutions enable more precise and targeted pruning. In data space, \ie in the first layer of our architecture, unused vertices are directly removed from the input grid, with subsequent updating of the vertices' neighborhoods. On the contrary, this procedure becomes more intricate in the lower layers. Here, not only must we update the vertices, neighborhoods, and tetrahedrons of the remaining vertices from the preceding layer, but we also need to ensure that vertices within the current layer's receptive field from those layers are preserved, as downsampling is defined over nearest neighbors.
Our proposed methodology starts at the data level, by applying a global occupancy mask (indicating vertex usage with 1 or 0). We propagate this mask to the lower resolution grids using our TetraConv and downsampling layers. Then, we iteratively process each level, remove unused connections (responses are all zero) and update affected vertices, including their neighborhoods and tetrahedral indices. Conceptually, deleting a vertex corresponds to removing a row, and removing an edge equates to inserting a zero in a cell within the unfolded data matrix. This entire refinement process is executed once, requiring only a few seconds, and is specific to the data set and the specific network configuration to be used. 
We point out that the pruning of vertices and connections in our methodology is lossless, meaning that it neither compromises network performance nor results in data loss. This technique is particularly beneficial for high-resolution grids, which are typically even sparser. It can significantly accelerate both training and inference times without sacrificing quality. In our experiments, the fraction of utilized data ranges from 8\% (when trained on a single class) to 82\% (when training across all available classes, such as airplanes, bikes, cars, and chairs).

\section{Quantitative evaluation}

\subsection{Metrics}
We evaluate our model in terms of 1-nearest neighbor accuracy (1NNA), minimum matching distance (MMD) and coverage (COV). All metrics compare a set of reference point clouds $S_r$ to a generated set of point clouds $S_g$. Importantly, the sets should be of same size, \ie $|S_r| = |S_g|$. In all our experiments, the individual point clouds are of size 2048.
\paragraph{1NNA} measures diversity and quality of the generated samples. It is defined as the leave-one-out accuracy of the 1-nearest neighbor classifier over the union of the two sets. In particular, a value of 50\% is considered optimal: half of the generated samples are closest to a reference sample and the other half is closest to a generated sample. Values above 50\% correlate with underfitting - generated samples are further away from the ground truth - and values below 50\% coincide with overfitting. The score can be computed with the following equation:
\begin{equation}
    \text{1NNA}(S_g, S_r) = \frac{\sum_{x\in S_g}\mathbb{1}[\mathcal{N}_x \in S_g] + \sum_{y\in S_r}\mathbb{1}[\mathcal{N}_y \in S_r]}{|S_r| + |S_g|},
\end{equation}
where $\mathcal{N}_x$ is the nearest neighbor of $x$ in the set $S_r \cup S_g - \{x\}$. The nearest neighbor can be computed in terms of Chamfer Distance (CD) or Earth Mover Distance (EMD).

\paragraph{MMD} computes the minimal distance between each reference sample and the nearest generated sample using CD or EMD and averages the score over the reference set:
\begin{equation}
    \text{MMD}(S_g, S_r) = \frac{1}{|S_r|} \sum\limits_{y\in S_r} \min\limits_{x\in S_g} D(x,y),
\end{equation}
where $D(\cdot,\cdot)$ is either CD or EMD. 

\paragraph{COV} quantifies the portion of reference samples that are matched to a generated samples. Higher values indicate diversity in the generative model. Mathematically, this translates to 
\begin{equation}
    \text{COV}(S_g, S_r) = \frac{1}{|S_g|} |\{\argmin\limits_{y\in S_r} D(x,y) | x \in S_g\}|,
\end{equation}
where $D(\cdot,\cdot)$ is either CD or EMD. 

\subsection{Quantitative results}

We additionally compare \emph{TetraDiffusion} to existing point cloud generators, namely Point-Voxel Diffusion (PVD)~\cite{zhou20213d} and Latent Point Diffusion (LION)~\cite{zeng2022lion}. While our method outperforms both in many cases (11 out of 18), comparing point cloud methods with mesh-based methods is non-trivial and should be approached with caution. The main reason is the difference in point cloud densities that emerges from the different methods. LION and PVD are trained on point clouds with strongly varying densities, \ie concentrating on smaller faces while allocating fewer points to larger faces. Conversely, our method produces point clouds with a more uniform density since the tetrahedral grid has near-uniform spacing. 

Current point cloud performance metrics are (inherently) highly influenced by sampling biases and different sampling strategies may lead to completely different numbers. In order to not favor one sampling strategy over the other, we employed Poisson disk sampling to generate ground truth point clouds from ShapeNet and compare against PVD and LION in \cref{tab:1nnasupp}.

\begin{table}[!ht]

\caption{\textbf{Quantitative evaluation of \emph{TetraDiffusion} and point cloud diffusion methods.} Metrics are computed over ShapeNet classes airplane, car and chair. MMD-CD is multiplied $\times 10^3$, MMD-EMD $\times 10^2$. We use our high resolution model for all experiments except chairs. $\circ$ denote point cloud generators and $\triangle$ mesh generators}
\label{tab:1nnasupp}

\centering
\setlength{\tabcolsep}{6pt}

\resizebox{0.85\linewidth}{!}{
\begin{tabular}{@{}llcccccc}
\toprule
    \multicolumn{2}{@{}l}{}  & \multicolumn{2}{c}{1-NNA $\downarrow$}&  \multicolumn{2}{c}{MMD $\downarrow$} &  \multicolumn{2}{c}{COV $\uparrow$} \\ 
    \hhline{~~------}
    \multicolumn{1}{@{}l}{Category} & Method & CD & EMD & CD & EMD & CD & EMD \\
    \midrule
    \multirow{3}{*}{Airplane}

    & LION $\circ$                             &87.28 & 91.73& 0.406& 0.777 & 41.48 & 27.16   \\
    & PVD  $\circ$                             &86.05 & 91.23& 0.423& 0.77 & 44.44 & 26.42  \\

     & \textbf{Ours} $\triangle$                 &\textbf{64.9} & \textbf{85.9}& \textbf{0.36}& \textbf{0.69} & 44.2 & \textbf{33.1}  \\ %
    \midrule
    
     \multirow{3}{*}{Car} 
    & LION  $\circ$                &60.94 & 68.32& \textbf{1.246}& 0.793 & \textbf{40.34} & 46.88 \\
    & PVD  $\circ$                 &60.23 & \textbf{64.49}& \textbf{1.246}& \textbf{0.776} & 38.07 &\textbf{49.15} \\
     & \textbf{Ours} $\triangle$       &\textbf{60.1} &69.9& 1.26&0.78 & 34.94 & 42.61  \\ 

    \midrule

    \multirow{3}{*}{Chair} 

     & LION $\circ$                 &61.3 & 65.01& 4.432& 2.557 & 43.15 & 43.88 \\
     & PVD $\circ$                 &61.66 & 63.25&\textbf{4.259}& 2.465 & 43.88 & 47.19 \\
     & \textbf{Ours} $\triangle$                 &\textbf{61.2} & \textbf{63.2}& 4.90&\textbf{2.66 }& \textbf{45.6} & \textbf{47.2}  \\ %

\bottomrule
\bottomrule
\vspace{0.5em}
\end{tabular}%
} %
\end{table}

\section{Ablations}

\subsection{Step size}

Treating time as continuous rather than discrete allows us to vary sampling steps during inference without relying on DDIM sampling strategies~\cite{song2020denoising}. Here, we quantitatively evaluate the number of steps needed in \emph{TetraDiffusion} to generate diverse and high quality meshes. The metrics for 24, 32, 48, 64, 96 and 128 diffusion steps can be found in \Cref{tab:steps}. Overall, the performance tends to saturate around 32 time steps.

\begin{table*}[!ht]
\centering
\setlength{\tabcolsep}{6pt}
\ra{1.2}
\caption{\textbf{Ablating different step sizes during inference.} We compare results on ShapeNet classes airplane, bike, car and chair. MMD-CD is multiplied by $\times 10^3$, EMD by $\times 10^2$.}
\label{tab:steps}

\resizebox{1.0\textwidth}{!}{
\begin{tabular}{@{}llccccccccc@{}}
\toprule
    \multicolumn{2}{@{}l}{} & \phantom{a} & \multicolumn{2}{c}{MMD $\downarrow$}& \phantom{a} & \multicolumn{2}{c}{COV $\uparrow$}& \phantom{a} & \multicolumn{2}{c}{1-NNA $\downarrow$}  \\ 
    \hhline{~~~--~--~--}
    \multicolumn{1}{@{}l}{Category} & Step size & & CD & EMD & & CD & EMD & & CD & EMD  \\
    \midrule
    \multirow{6}{*}{Airplane} 
&24	&&0.40	&0.73	&&38.77	&29.14	&&75.06	&88.40\\
&32	&&0.39	&0.78	&&41.48	&28.15	&&74.57	&88.89\\
&48	&&0.38	&0.73	&&46.91	&\textbf{31.36}	&&\textbf{66.67}	&87.65\\
&64	&&\textbf{0.37}	&0.75	&&44.69	&28.40	&&67.90	&88.40\\
&96	&&0.39	&\textbf{0.71}	&&45.19	&30.62	&&70.12	&\textbf{87.28}\\
&128&&0.38	&0.74	&&\textbf{47.41}	&27.16	&&68.77	&89.14\\
    \midrule
    \multirow{6}{*}{Bike}
&24	&&1.21	&0.95	&&45.40	&45.99	&&56.68	&65.88\\
&32	&&\textbf{1.01}	&\textbf{0.93}	&&50.45	&46.29	&&53.71	&66.02\\
&48	&&1.17	&0.96	&&\textbf{54.01}	&49.85	&&\textbf{51.04}	&64.69\\
&64	&&1.18	&0.95	&&48.96	&48.07	&&56.53	&64.84\\
&96	&&1.14	&0.95	&&52.23	&\textbf{50.15}	&&52.37	&65.28\\
&128&&1.18	&0.95	&&50.15	&48.66	&&55.34	&\textbf{64.39}\\
    \midrule
    \multirow{6}{*}{Car} 
&24	&&1.31	&0.79	&&38.64	&44.89	&&63.92	&70.74\\
&32	&&1.28  &0.79	&&38.35	&41.48	&&61.22	&70.45\\
&48	&&1.25	&0.77	&&35.23	&43.47	&&\textbf{58.81}	&\textbf{62.93}\\
&64	&&\textbf{1.24}	&0.77	&&\textbf{40.06}	&43.47	&&59.38	&67.90\\
&96	&&1.25	&\textbf{0.76}	&&37.50	&\textbf{45.17}	&&59.23	&65.48 \\
&128&&\textbf{1.24}	&0.78	&&35.51	&44.60	&&59.80	&68.47\\
     
    \midrule
    \multirow{6}{*}{Chair} 
&24	&&4.90	&2.56	&&40.09	&45.92	&&69.75	&66.33\\		
&32	&&4.48	&2.43	&&41.40	&47.96	&&65.82	&63.70\\
&48	&&4.65	&2.43	&&44.75	&47.96	&&62.24	&59.48\\
&64	&&4.63	&2.39	&&45.04	&49.27	&&61.88	&59.77\\
&96	&&4.46	&2.37	&&\textbf{45.34}	&\textbf{51.02	}&&\textbf{60.28	}&57.87\\
&128&&\textbf{4.35}&\textbf{2.30	}&&45.19	&51.31	&&61.08	&\textbf{56.41}\\
\bottomrule
\end{tabular}%
} %

\end{table*}

\subsection{Clipping and offset noise}

Additionally, we ablate two orthogonal inference strategies, namely clipping $\mathbf{\hat{x}}_\theta(\mathbf{z_t}; t)$ to the data range along the reverse diffusion chain and varying the offset noise \cite{Guttenberg2023}. Clipping the prediction in $x$-space is often found in codebases but often not emphasized~\cite{hoogeboom2023simple, saharia2022photorealistic}. As $\mathbf{\hat{x}}_\theta(\mathbf{z_t}; t)$ is iteratively used in ancestral sampling, not clipping can be seen as a train-test mismatch and may affect performance~\cite{saharia2022photorealistic}. 

An interesting observation was made in \cite{Guttenberg2023, lin2023common}. Diffusion models tend to generate images with medium brightness due to the discrepancy between training and inference signal-to-noise ratio. This becomes even more apparent in the high resolution regime as low frequency features are often not completely removed during the forward process. Conversely, the long wavelength features of the Gaussian noise at the start of the reverse diffusion process are least likely to be destroyed, which enforces the average value (the longest wavelength). A simple, yet effective solution is proposed by~\cite{Guttenberg2023}: randomizing the zero-frequency component of the noise by adding a single, independently and identically distributed sample over the entire image during training.To evaluate this method, we train a model using offset noise and conduct tests both with and without noise offsetting.

While we do not see large differences with clipping, offset noise or both, we did observe increased high-frequency detail in some classes when using clipping. On the downside, the generated meshes tend to have more holes.

\end{document}